# COMPUTING AND LEARNING ON COMBINATORIAL DATA

by

**Simon Zhang**

**A Dissertation**

*Submitted to the Faculty of Purdue University*

*In Partial Fulfillment of the Requirements for the degree of*

**Doctor of Philosophy**

Department of Computer Science

West Lafayette, Indiana

May 2025

**THE PURDUE UNIVERSITY GRADUATE SCHOOL**
**STATEMENT OF COMMITTEE APPROVAL**

**Dr. Tamal Krishna Dey, Chair**
Department of Computer Science

**Dr. David Gleich**
Department of Computer Science

**Dr. Bruno Ribeiro**
Department of Computer Science

**Dr. Saugata Basu**
Department of Mathematics

**Approved by:**
Dr. Voicu Popescu

To my parents

# ACKNOWLEDGMENTS

The journey toward completing a Ph.D. is filled with challenges and the experience of overcoming numerous difficulties, but it is also deeply rewarding. Earning the title of Doctor is a significant accomplishment and honor for me. Along the way, many extraordinary individuals have supported and guided me, helping me grow into a capable and qualified computer scientist.

My Ph.D. journey would not have been possible without the guidance and support of my advisor, Professor Dey, who believed in my potential and recognized my readiness on a rigorous research path from the start. I have taken several of his courses, where I received extensive training in algorithms and computational geometry theory. These classes have influenced my approach to research on making foundational contributions, whether the project is experimental or theoretical. Under Professor Dey's mentorship, I have gained a comprehensive understanding of the research process from studying the state-of-the-art and identifying significant problems to designing algorithms, conducting experimental evaluations, and effectively communicating results through writing. I am particularly grateful for the freedom he gave me to pursue my interests in machine learning. Moreover, Professor Dey has given me lessons in the effectiveness of both oral and written communication and encouraged me to develop independent thinking skills, which will continue to guide me throughout my career.

I would like to thank the Ph.D. Dissertation Committee members: Professors Saugata Basu, David Gleich, and Bruno Ribeiro, for their valuable insights and encouragement.

Many thanks go to my colleagues in Professor Deys research group, with whom I have had the pleasure of discussing and collaborating: Gilberto Arroyo, Andrew Haas, Dr. Tao Hou, Dr. Soham Mukherjee, Shreyas Samanga, Dr. Ryan Slechta, Aman Timalsina, and Dr. Cheng Xin. I would also like to extend my thanks to others I have interacted with at Purdue, including Hrishikesh Viswanath, Yufan Huang, Ananth Shreekumar, Yongye Su, and many more. Additionally, I really enjoyed my time participating in racquet sports such as table tennis, pickleball, badminton, and tennis at the Purdue Recreation and Wellness Center as well as participating in the table tennis club.

During my time working on two high-performance computing projects at Ohio State, I collaborated with several talented individuals, to whom I am very grateful: Ryan DeMilt, Liang Geng, Dr. Chengxin Guo, Dr. Kun Jin, Dr. Hao Wang, Professor Cathy Xia, Professor Mengbai Xiao, and Haotian Xie. I would also like to thank Professor Facundo Memoli from the Math Department at Ohio State for allowing me to attend his research group seminars. Special thanks go to my first academic advisor at Ohio State, Professor Yusu Wang, who mentioned about the usage of GPUs to accelerate computational geometry computation, starting my first step in high-performance computing research.

I would especially like to thank Dr. Greg Henselman from Pacific Northwest National Labs (PNNL) and Professor Henry Adams from Colorado State University for inviting me to present Ripser++ at various venues. I would also like to thank Professor Zhihui Zhu from Ohio State, who gave me advice on formatting my journal paper.

I completed two internships at Argonne National Labs, working on AI for science research projects. I am grateful to Dr. Zhengchun Liu and Dr. Hemant Sharma for providing me with these opportunities and for their invaluable mentorship.

My Ph.D. research has been supported through both Teaching Assistantships and Research Assistantships at Ohio State and Purdue. I am also grateful to have received a Purdue Ph.D. Dissertation Fellowship.

At the foundation of both my life and career are my parents, whose unwavering love and support have been instrumental in my journey. As well-educated individuals themselves, they have always encouraged me, and I am deeply indebted to them.

This dissertation would not have been possible without the support of all these incredible individuals. Thank you for your encouragement and for believing in me.

# TABLE OF CONTENTS

# LIST OF TABLES

# LIST OF FIGURES

# ABSTRACT

The twenty-first century is a data-driven era where human activities and behavior, physical phenomena, scientific discoveries, technology advancements, and almost everything that happens in the world resulting in massive generation, collection, and utilization of data. Connectivity in data is a crucial property. A straightforward example is the World Wide Web, where every webpage is connected to other web pages through hyperlinks, providing a form of directed connectivity. Combinatorial data refers to combinations of data items based on certain connectivity rules. Other forms of combinatorial data include social networks as simplicial complexes, molecules as graphs, sets of binary classifiers as hypergraphs, and radius graphs from a metric space, which realizes connectivity for topological data analysis. This Ph.D. dissertation focuses on computation and learning on combinatorial data.

Persistence theory asks the question of: *"What are the future dependencies of a vector in a time-indexed representation space?"*. A canonical example is given by persistent homology, which asks this question for a sequence of homology vector spaces. Homology is a measurement of "partial disconnectedness." For example, it can measure an obstruction to traveling through a 2d surface from one point to another. Since persistence is computed over a causal ordering, computing persistent homology appears as an inherently sequential process which would make computation on GPU difficult. We find that much of the algebraic computation only depends on the neighboring connectivity of the data and not on any algebraic operations, allowing for a CPU-GPU hybrid approach.

When given samples from a metric space, higher order connectivity can be realized on samples through a Vietoris-Rips complex. The persistence can be computed efficiently by exploiting the heavy hitter case of "generalized nearest neighbors." This can be computed massively in parallel on GPU with an efficient transfer of data structures between GPU and CPU. We have successfully designed and implemented algorithms and their implementations on GPU in the domain of persistent homology, achieving high performance by two open-source software, namely HYPHA and Ripser++.

The multiset of pairs of time end points when a homology generator belongs to a homology vector space is called a persistence diagram. Since any pair of times which are equal belong to the persistence diagram, when embedded in the plane, there is a diagonal with infinitely many points. To compute the earth mover's distance (EMD) between persistence diagrams, also called the Wasserstein distance between persistence diagrams, we have reduced the geometry of the plane to a graph embedded onto the surface of a finite height three dimensional pointed cone and compute an optimization problem on a sparse graph. This will not introduce any distortion in computing the EMD from the diagonal, which can be as large as 2 when only taking finitely many points from the diagonal. Indirectly, our approach gives a near linear time algorithm for a $(1+\epsilon)$-approximation of the EMD between persistence diagrams. This is an optimal lower bound if a $(1+\epsilon)$ approximate EMD cannot be solved in time $O(n^{1+o(1)-\delta})$ for any $\delta > 0$. Besides algorithms and analysis, we have developed an open- source software called PDoptFlow based on our algorithm. We perform extensive experiments and show that the actual empirical error is very low. We also show that we can achieve high performance at low guaranteed relative errors, improving upon the state of the arts.

When doing graph learning, it is conventional to use a graph neural network. Such neural networks, however, cannot see substructures on the graph. Persistent homology can be used to reconstruct the graph through an algorithm and measure the lifetime of the connected components over this graph reconstruction algorithm. The cycles of the graph, however, cannot be fully expressed through a time displacement. In extended persistence, a virtual node is introduced, and a cone is formed, creating virtual 2D sheets that can zero out the cycles with respect to the homological vector space. This works and the cone can be simulated in subquadratic time. We demonstrate the effectiveness of our method on real world datasets compared to most recent graph representation learning methods.

Hypergraphs are a collection of subsets of a set of nodes. This is a very general kind of combinatorial data. Hypergraphs are usually learned through hypergraph neural networks, which view the hypergraph as a collection of rooted trees. The automorphism group on hypergraphs would not be respected. In fact, this automorphism group forms new symmetries. We have characterized this limitation in terms of the symmetries of the data as viewed by the universal cover of a bipartite graph. We show that introducing symmetry breaking through the hyperedges lowers the symmetry introduced by GWL-1 and aids in hyperlink prediction.

# 1. INTRODUCTION

This Ph.D. dissertation focuses on learning and computing with combinatorial data, studying and examining topological and connectivity features within and across connected data to improve the performance of learning and achieve high algorithmic efficiency. The dissertation is structured into eight Chapters.

- Chapter 1. The Introduction
- Chapter 2. The background introducing all relevant notation and math. We begin with first order logic followed by set theory. We then introduce category theory which gives a general language for spaces and transformations between them. We then go through algebra, geometry and topology, which describe spaces with specific physical meaning. Finally, we review statistics and the relevant data that concerns this dissertation.
- Chapter 3. We introduce what it means to compute persistence from an algebraic perspective. In persistence theory, we are given a functor $F : \mathcal{D} \to \mathbf{Vec}$ from the category $\mathcal{D}$ to the category of vector spaces over the reals and an acyclic subcategory $Q(\mathcal{D})$ of $\mathcal{D}$. We ask the question of: "In what objects in $\mathrm{im}(F\mid_{Q(\mathcal{D})})$ does a vector in $o \in \mathrm{im}(F\mid_{Q(\mathcal{D})})$ stay independent?" We go over persistence theory in this context.
- Chapter 4. ([1]) Persistence is often viewed through the homology functor with the category $\mathcal{D}$ provided as data. This is called persistent homology. The data is given as a sequence of simplicial complexes, which are in analogy to discrete manifolds. We ask the question of persistence for the homology vector spaces over the reals. This is an entirely linear algebraic question but much of the algebra is not needed, allowing us to devise a hybrid CPU/GPU approach, termed HYPHA, to compute persistent homology as a sparse matrix problem.
- Chapter 5. ([2, 3]) This chapter is about computing persistence with GPU for connectivity that can be realized by a metric space. Ripser++ is a method to compute persistent homology induced by a metric. In Ripser++ we are in the same framework of persistent homology, but where the combinatorial data or Vietoris-Rips simplicial complex in this case, is imagined on samples from a metric space. The Vietoris-Rips complex is generated over increasing distance thresholds where a subset of samples form a simplex when their diameter is less than a distance threshold. Since the simplicial complex is being generated sequentially, it would appear that computing persistence would require a sequential search over this order of generation. However, there is a heavy hitter case where a cycle generator does not persist. In these cases only local neighborhood information, in analogy to nearest neighbor points, needs to be checked so the cycle generators are all independent of each other when viewed through each distance threshold. Along with the fact that the Vietoris-Rips complex is completely determined by the pairwise distances, we can use the GPU to efficiently eliminate these spurious generators. We also do an efficient data structure transfer between GPU and CPU to compute the few remaining cycle generators' persistence.

- Chapter 6. ([4]) Persistence diagrams are the multiset of pairs consisting of the times of creation and destruction for a generator in persistent homology. A variation of the earth mover's distance, called the 1-Wasserstein distance, can be computed between persistence diagrams. This distance looks for a matching between diagrams that optimizes an aggregation of the distances on the matching. There are assumed to be infinitely many points on a diagonal line, however. In order to maintain an accuracy below a distortion factor of 2, we view the matching over a discrete bipartite graph instead of in the original euclidean plane with the finite points projected onto the diagonal. The computation can then be done by forming a super node for the diagonal and treating the remaining points as from the Euclidean plane with distances to the super node determined by projection. If a spanner is applied on the points on the Euclidean plane, the problem reduces to a conventional min-cost flow problem on a graph with linear order input. We show that in terms of a time accuracy tradeoff this approach is effective.

- Chapter 7. ([5]) When doing graph learning, it is conventional to use a graph neural network. Such neural networks, however, cannot see substructures on the graph. Persistent homology can be used to reconstruct the graph algorithmically and measure the lifetime of the connected components. The cycles of the graph, however, cannot be fully expressed through a displacement of times. In extended persistence, a virtual node is introduced and a cone is formed, creating virtual 2-d sheets that can zero out the cycles with respect to the homological vector space. This actually works and the cone can be simulated without going to quadratic time.

- Chapter 8. ([6, 7]) Hypergraphs are a collection of subsets of a set of nodes. This is the most general kind of undirected combinatorial data. These can be viewed as bipartite graphs where the hyperedges are virtual nodes for all the nodes in the hyperedges. In hypergraph representation learning, the hypergraph neural networks still compute with respect to expanding the neighborhoods around the nodes. A single hop forward pass can be up to exponential time with respect to the number of nodes on hypergraphs and so usually more complicated neural networks are not used. We notice that for hyperlink prediction the automorphism group of the training hypergraph is not respected and in fact made more symmetric by hypergraph neural networks while the testing automorphism group is unknown. Assuming the testing automorphism group is smaller, meaning having more entropy, then it would be beneficial to break the symmetry of the training hypergraph.

# 2. BACKGROUND

We go over all of the necessary mathematical notations and definitions to understand this thesis.

## 2.1 First Order Logic

In order to have a language with which we can do reasoning, we must define a logic. We go over first order logic (FOL) [8] here in order to be able to define the later mathematical background.

In FOL, we define **truth** and **false** as two separate values that can be assigned to logical statements. These values are denoted by T and F, respectively. When expressing logic, we would like to connect true logical statements instead of false logical statements.

In FOL, we start with a domain of objects. These are in analogy to entities either physical or non-physical.

**Definition 2.1.1.** *Objects can be one of either **constant** or **variable**. Variables are changeable but constants cannot change.*

In FOL, in order to express objects in a language we define **terms**. These are in analogy to nouns in natural language. The objects "pear" and "banana" are examples of terms over the domain of tangible entitites.

When there are multiple terms together, let:

**Definition 2.1.2.** *An ordered list of terms be called a **tuple**.*

We can also define a way to convert a tuple of terms into exactly one term. For example, "the country of origin of (...)" is a function on the term (...) that returns a single country.

**Definition 2.1.3.** *A **function** in FOL is defined as taking a tuple of terms and returning a single term.*

We can now formally define terms:

**Definition 2.1.4.** *A **term** is a variable object, a constant object or a function of a tuple of terms.*

In order to assign truth or falsity value to a tuple of terms, we define a **predicate**.

**Definition 2.1.5.** *A **predicate** takes a tuple of terms and returns a single value of T or F*

Predicates are also known as **atomic logical statements**.

In FOL, we can build up atomic logical statements to form logical statements. We can do this by connecting logical statements to form more logical statements. These come about through logical connectives. We can also negate a logical statement to flip its True/False value. Quantifiers are used to assign truth value to a logical statement in terms of variables. Since logical statements are built from atomic logical statements, they must have one of either True or False value.

**Definition 2.1.6.** *The **connectives** between two logical statements $P, Q$ allow for the formation of another logical statement. These include:*

- *(conjunction) $P$ AND $Q$ means "both $P$ and $Q$ are True"*
- *(disjunction) $P$ OR $Q$ means "one of $P$ or $Q$ are True"*
- *(implication) $P \Rightarrow Q$ means "if $P$ is True, then $Q$ is True."*

**Definition 2.1.7.** *We can also take a logical statement $P$ and flip its truth value through **negation**. We denote this by $\neg P$.*

We can also apply **Quantifiers** on variables.

**Definition 2.1.8.** *A quantifier can be a **"for all"**, denoted $\forall$. This quantifier means that every possible object that the variable could instantiate, or be, is involved in the logical statement.*

**Definition 2.1.9.** *A quantifier can be an **"exists"**, denoted $\exists$. This quantifier means that some object that the variable could instantiate, or be, is involved in the logical statement.*

**Definition 2.1.10.** *A **logical statement** is defined as an atomic statement, logical statements with connectives in between them, logical statements with quantifiers on their variables, or a negated logical statement.*

We can summarize the language of FOL in the following grammar: Under finiteness assumptions this grammar becomes a context free grammar [8].

$$
\begin{aligned}
\text{Logical Statement} \rightarrow \\
&\text{Atomic Logical Statement} \\
&\mid (\text{Logical Statement Connective Logical Statement}) \\
&\mid \text{Quantifier Variable (Logical Statement)} \\
&\mid \neg\text{Logical Statement}
\end{aligned}
$$

$$\text{Atomic Logical Statements} \rightarrow \text{Predicate(Term}, \ldots)$$

$$\text{Term} \rightarrow \text{Function(Term}, \ldots) \mid \text{Constant} \mid \text{Variable}$$

$$\text{Connective} \rightarrow \ \Rightarrow \ \mid \text{AND} \mid \text{OR}$$

$$\text{Quantifier} \rightarrow \forall \mid \exists$$

$$\text{Constant} \rightarrow g \mid c \mid 1 \mid 0 \mid \cdots$$

$$\text{Variable} \rightarrow v \mid p \mid q \mid t \mid x \mid \cdots$$

$$\text{Predicate} \rightarrow \ < \ \mid \ = \ \mid \text{"connected to"} \mid \text{"contained in"} \mid \text{"caused by"} \mid \text{"happened before "} \mid \cdots$$

**Figure 2.1.** The grammar of first order logic

## 2.2 Set Theory

Foundational to the language of math is the concept of a set. Sets provide the building blocks of definitions that are usable in math. They are intimately tied to first order logic, providing a formalization to groupings of objects.

**Definition 2.2.1.** *A* ***set*** *is an unordered collection of objects called* ***elements****.*

A set can be denoted by enumerating its elements via the following notation:

$$A \triangleq \{\mathrm{e}_1, ..., \} \tag{2.2}$$

The **empty set** is the set consisting of no elements:

$$\varnothing \triangleq \{\} \tag{2.3}$$

To denote that an element e belongs to a set $A$, we use the notation e $\in A$.

Similarly, to denote that an element e does not belong to a set $A$, we use the notation e $\notin A$.

**Definition 2.2.2.** *The* ***cardinality*** *or* ***size*** *of a set is the number of elements in it. This number can be either finite or infinite. For a set $A$, let $|A|$ denote its cardinality.*

A set can also be denoted by **set-builder notation**, this is a first order logical statement that determines the set in terms of another set.

(2.4) For example, letting $X$ denote the set of all "fruits".

$$A = \{x \in X : x \text{ is a red apple}\} \tag{2.5}$$

is a set determined by the truthness of the logical statement "x is a red apple" where $x$ is viewed as a variable.

Some simple examples of sets include number systems, which are necessary for measurement:

(2.6)
- The natural numbers: $\mathbb{N} \triangleq \{0, 1, 2, ...\}$
- The integers: $\mathbb{Z} \triangleq \{..., -2, -1, 0, 1, 2, ...\}$
- The rational numbers: $\mathbb{Q} \triangleq \{\frac{p}{q} : p, q \in \mathbb{Z}, q \neq 0\}$
- The real numbers: $\mathbb{R}$
- The closed intervals: $[a, b] \triangleq \{x \in \mathbb{R} : a \leq x \leq b\}$.
- The closed-open intervals: $[a, b) \triangleq \{x \in \mathbb{R} : a \leq x < b\}$
- The complex numbers: $\mathbb{C}$
- The integers from 1 to $n$: $[n] \triangleq \{1, ..., n\}, n \in \mathbb{N}$

- The integer interval: $\{\mathrm{i}, ..., \mathrm{j}\}, \mathrm{i}, \mathrm{j} \in \mathbb{N}, \mathrm{i} \leq \mathrm{j}$

The natural numbers $\mathbb{N}$ can be bootstrapped from:

1. The empty set, which represents $0 \in \mathbb{N}$
2. The singleton of an emptyset: $\{\varnothing\}$, which represents $1 \in \mathbb{N}$
3. And the successor operation, which takes a natural number and increments it by 1. This is done by taking a set-theoretically constructed natural number and its union with $\{\varnothing\}$.

The other number systems are derivable from $\mathbb{N}$, see [9, 10].

**Definition 2.2.3.** *A set $A$ is a* ***subset*** *of $B$, denoted $A \subseteq B$ iff:*

$$x \in A \Rightarrow x \in B, \forall x \in A, \tag{2.7}$$

This is a first order logic statement that expresses that every element in $A$ belongs to $B$ as well, see Section 2.1 for what first order logic is.

**Definition 2.2.4.** *If $A \subseteq X$, we denote $X \setminus A$ as the* ***complementary set*** *of $A$ with respect to $X$.*

$$X \setminus A \triangleq \{x \in X : x \notin A\} \tag{2.8}$$

**Definition 2.2.5.** *The* ***union*** *of two sets $A, B$ is the set formed by all the elements of $A$ or of $B$:*

$$A \cup B \triangleq \{x : x \in A \ \textit{or} \ x \in B\} \tag{2.9}$$

**Definition 2.2.6.** *The* ***intersection*** *of two sets $A, B$ is the set formed by all the elements belonging to both $A$ and $B$:*

$$A \cap B \triangleq \{x : x \in A \ \textit{and} \ x \in B\} \tag{2.10}$$

**Definition 2.2.7.** *A* ***disjoint union*** *of two sets $A, B$ is defined by:*

$$A \sqcup B \triangleq \{(a, 1), a \in A\} \cup \{(B, 2), b \in B\} \tag{2.11}$$

### 2.2.8 Cartesian Products

Given two sets, we can also combine them to form a set of orderings over all pairings:

**Definition 2.2.9.** *A* ***cartesian product*** *between two sets* $A, B$ *is defined by the set of ordered pairs of elements from both sets:* $A \times B \triangleq \{(a, b) : a \in A, b \in B\}$

The cartesian product can be applied multiple times across sets. The elements of a $k$-fold cartesian product, denoted:

$$\Pi_{i=1}^{k} A_i \triangleq A_1 \times ... \times A_k \tag{2.12}$$

are called **$k$-tuples**, or just tuples when the $k$ is implicit. A $k$-tuple is often denoted $(a_1, ..., a_k), a_i \in A_i$. This definition is compatible with the definition of a tuple from FOL.

A **sequence** over a set $A$ is defined as an element of $\Pi_{i \in \mathbb{N}} A$ where $\mathbb{N}$ is the set of natural numbers. Sequences are denoted by $(a_i)_{i \in \mathbb{N}}$ where each $a_i \in A$.

**Definition 2.2.10.** *We can define a* ***power*** *of a set* $B$ *by* $A$ *by the cartesian product:*

$$B^A \triangleq \Pi_{a \in A} B \tag{2.13}$$

In particular, we define the power set of a set $X$, denoted $2^X$ by:

$$2^X \triangleq \Pi_{x \in X} \{0, 1\} \tag{2.14}$$

**Proposition 2.2.11.** $2^X = \{A : A \subseteq X\}$

*Proof.* Let elements of $2^X$ be denoted by $(i_x)_{x \in X}, i_x \in \{0, 1\}, \forall x \in X$

The following one to one correspondence between $2^X$ and $\{A : A \subseteq X\}$ gives the equality:

$$(i_x)_{x \in X} \mapsto \{x : i_x = 1\} \tag{2.15}$$

Certainly this is injective due to the definition of set equality:

$$\{x : i_x = 1\} = \{x : j_x = 1\} \Rightarrow i_x = j_x \tag{2.16}$$

Similarly the map is surjective since for any set $A \subseteq X$, we can define an indicator function $\mathrm{i}_\bullet : X \to \{0,1\}$ with $A = \{x : \mathrm{i}_x = 1\}$ $\square$

A collection of subsets of a set $X$ can thus be considered as a subset of the power set $2^X$. We denote the collection of all subsets of size $k$ of $X$, with $0 \leq k \leq |X|$ as follows:

$$\binom{X}{k} \tag{2.17}$$

We also define a special subset of $2^X$ with a joint separation property.

**Definition 2.2.12.** *A **partition** $P$ of $X$ is a collection of subsets $P \subseteq 2^X$ with the property that:*

1. $\bigcap_{\mathrm{i}} S_{\mathrm{i}} = \varnothing, \forall S_{\mathrm{i}} \in P$
2. $\bigcup_{S \in P} S = X$

A partition $P$ of $X$ can also be denoted by the following disjoint union:

$$\bigsqcup_{S \in P} S = X \text{ (up to bijection)} \tag{2.18}$$

Over one set, we can then define a relation, or set of pairs over that set:

**Definition 2.2.13.** *Let a **relation** $R$ on a set $P$ be a subset of the cartesian product $P \times P$. It is customary to use the notation $aRb$ if $a, b \in R \subseteq P \times P$*

A pair of elements in $P$ that do not belong to $R$ are called **incomparable** with respect to relation $R$.

For a set, we can define equality amongst elements. This can be defined by a special kind of relation called an equivalence relation:

**Definition 2.2.14.** *Given a set $S$, an **equivalence relation** is a relation $\sim$ on set $S$ satisfying:*

- $a \sim a, \forall a \in S$ *(reflexivity)*

- *$a \sim b$ iff $b \sim a, \forall a, b \in S$ (symmetry)*

- *$\forall a, b, c \in S$, If $a \sim b$ and $b \sim c$ then $a \sim c$*

A set $S$ with an equivalence relation imposed on it forms a new set. We denote such a set by $\frac{S}{\sim}$.

The elements of $\frac{S}{\sim}$ are denoted $[s], s \in S$ where $[s] = [s']$ if $s \sim s', \forall s, s' \in S$. These elements $[s] \in \frac{S}{\sim}$ can be considered as subsets of $S$ and $\frac{S}{\sim}$ as a collection of subsets of $S$. We call the elements of $\frac{S}{\sim}$ as equivalence classes.

**Proposition 2.2.15.** *The equivalence classes of $\frac{S}{\sim}$ form a partition of $S$.*

*Proof.* See the chapter on Relations in [11]. □

### 2.2.16 Multisets

We also have a notation for sets that can have repeated elements:

**Definition 2.2.17.** *A **multiset** is an unordered collection of objects called **elements** that can repeat.*

Multisets are often also called "bags." These are useful to express an enumeration of objects with repetition.

### 2.2.18 Functions

Functions can be defined in set theory beyond the definition of function in FOL. Of course, this definition is still compatible with the definition of a function from FOL.

**Definition 2.2.19.** *A **function** from domain set $X$ to target set $Y$ is an element of the set formed by $Y$ powered by $X$, $Y^X$. These elements are denoted $f : X \to Y$.*

Since a function $f : X \to Y$ belongs to $Y^X$, it maps a single $x \in X$ to a unique $y \in Y$. This can be seen by:

$$Y^X = \Pi_{x \in X} Y \tag{2.19}$$

so an element of $Y^X$ assigns only one $y \in Y$ for each $x \in X$.

The **domain** $X$ of $f : X \to Y$ is denoted $\text{dom}(f)$.

The **codomain** $Y$ of $f : X \to Y$ is denoted $\text{codom}(f)$.

The range or **image** of $f : X \to Y$ is denoted

$$\text{im}(f) \triangleq \{y \in Y : y = f(x)\} \tag{2.20}$$

Certainly $\text{im}(f) \subseteq Y$.

**Definition 2.2.20.** *A function $f : X \to Y$ is called **injective** if:*

$$f(x) = f(y) \Rightarrow x = y, \forall f(x), f(y) \in im(f) \tag{2.21}$$

**Definition 2.2.21.** *A function $f : X \to Y$ is called **surjective** if:*

$$im(f) = Y \tag{2.22}$$

### 2.2.22 Order

In the language of math, sets are often given order. This models direction from the physical world, such as causality.

We can define an order that models locally pairwise orderings:

**Definition 2.2.23.** *A **partial order**, is a relation $\leq$ on a set $P$ that is reflexive, antisymmetric, and transitive. That is, for all $a, b, c \in P$ it must satisfy:*

- *Reflexivity: $a \leq a$, i.e. every element is related to itself.*
- *Antisymmetry: if $a \leq b$ and $b \leq a$ then $a = b$, i.e. no two distinct elements precede each other.*
- *Transitivity: if $a \leq b$ and $b \leq c$ then $a \leq c$.*

(2.23) An example of a partial order includes a **binary decision tree** $(T, \leq)$. The set $T$ is a set of elements with an exponentially growing hierarchy.

$$T \triangleq \{x_{(\mathrm{i},\mathrm{j})} : \mathrm{i} \in \mathbb{N}, \mathrm{j} \in \{0, ..., 2^{\mathrm{i}} - 1\}\} \tag{2.24}$$

The index i is called the level in the decision tree and the integer j is the index in the hierachy.

The partial order $\leq \subseteq T \times T$ models the binary decisions at each level. This can be defined by:

$$x_{(\mathrm{i},\mathrm{j})} \leq x_{(\mathrm{i}+1,2\mathrm{j})} \text{ and } x_{(\mathrm{i},\mathrm{j})} \leq x_{(\mathrm{i}+1,2\mathrm{j}+1)} \tag{2.25}$$

In a partial order $(P, \leq)$, we can define meets and joins in analogy to the min and max of numbers as follows:

**Definition 2.2.24.** *A **join** of a poset* $(P, \leq)$ *is an element* $\mathrm{j} \in P$ *with no* $\mathrm{j}' \in P, \mathrm{j}' \leq \mathrm{j}$ *and* $\mathrm{j}' \neq \mathrm{j}$.

**Definition 2.2.25.** *A **meet** of a poset* $(P, \leq)$ *is an element* $m \in P$ *with no* $m' \in P, m \leq m'$ *and* $m' \neq m$.

**Definition 2.2.26.** *A **total order*** $\leq$ *on set* $P$ *is a partial order on* $P$ *that satisfies strong connectivity, meaning:*

$$a \leq b \text{ or } b \leq a, \forall a, b \in P \tag{2.26}$$

(2.27) An example of a total order includes the real number line: $(\mathbb{R}, \leq)$.

As a line, at any $x \in \mathbb{R}$, all $y \in \mathbb{R}, y \neq x$, we must have only one of $y \leq x$ or $y \geq x$ be true.

Compare this in contrast to the binary decision tree $(T, \leq)$ that involves branching, meaning for some $a \in T, \exists b, c$ with $a \leq b, a \leq c$ with $b, c$ incomparable.

**Definition 2.2.27.** *A **lattice** is a poset* $(P, \leq)$ *with two additional operations between pairs:*

- *The meet* $a \wedge b \in P$ *between* $a, b \in P$ *has that:*
  - $a \wedge b \leq a, a \wedge b \leq b$

- $\nexists c, c \leq a, c \leq b, a \wedge b \leq c$

- *The join $a \vee b \in P$ between $a, b \in P$ has that:*
  - $a \leq a \vee b, b \leq a \vee b$
  - $\nexists c, a \leq c, b \leq c, c \leq a \vee b$

(2.28) A useful lattice is the **boolean algebra** consisting of a set $B$ containing the special elements $0, 1$ and the operations AND, OR, NOT. These operations are define below:

- $b_1$ AND $b_2$ is the meet of $b_1, b_2$ in $B$
- $b_1$ OR $b_2$ is the join of $b_1, b_2$ in $B$
- NOT $b \in B$
- 1 AND $b = 1, \forall b \in B$
- 0 OR $b = b, \forall b \in B$

### 2.3 Category Theory

We will briefly review category theory [12]. The definition of a category is motivated by classes of sets and the functions between these sets.

In order to define a category, we must define a generalization of a set called a class:

**Definition 2.3.1.** *A class $C$ is defined by its members, which are sets. A member $c$ belonging to $C$ is denoted $c \in C$*

*A subclass $S \subseteq C$ of a class $C$ has every member of $S$ belonging to $C$.*

In terms of observables in the real world, which is of more concern in computer science, we can assume that classes are just sets.

We define a category by classes of objects and morphisms, also called arrows, in analogy to sets and functions. Formally, we define a category $\mathcal{C}$ as:

**Definition 2.3.2.** *A category $\mathcal{C} \triangleq (ob(\mathcal{C}), mor(\mathcal{C}), dom : mor(\mathcal{C}) \to ob(\mathcal{C}), codom : mor(\mathcal{C}) \to ob(\mathcal{C}), \circ)$ where $ob(\mathcal{C})$ is a class of objects, $mor(\mathcal{C})$ are a class of morphisms, or arrows, and:*

- $\forall a, b, c \in ob(\mathcal{C}), \circ : hom(a,b) \times hom(b,c) \rightarrow hom(a,c)$ *where* $hom(b,c) \subseteq mor(\mathcal{C})$ *is a subclass of* $mor(\mathcal{C})$ *where all* $f \in mor(\mathcal{C})$ *have* $dom(f) = b$, $codom(f) = c$, *denoted* $f : b \rightarrow c$

- *The composition operation* $\circ$ *is associative:* $\forall a, b, c \in ob(\mathcal{C}), \forall f : a \rightarrow b, g : b \rightarrow c, h : c \rightarrow d, h \circ g \circ f = (h \circ g) \circ f = h \circ (g \circ f)$

- $\forall a \in ob(\mathcal{C}), id_a : a \rightarrow a, \forall f \in mor(\mathcal{C}), id_a \circ f = f \circ id_a = f$

We can view categories as sets with the following definitions:

**Definition 2.3.3.** *A category* $\mathcal{C}$ *where the class of objects* $ob(\mathcal{C})$ *is a set is called a* ***small category****.*

*Furthermore, if the category where the class of morphisms is also a set then* $\mathcal{C}$ *is called a* ***locally small category****.*

We will attempt to make all categories locally small categories.

We will often denote locally small categories $\mathcal{C}$ by a pair $(V, E)$ where $V$ is a set of objects and $E$ is a set of morphisms between objects. We can also recover the objects and morphisms of $\mathcal{C}$ with the notation: $V(\mathcal{C})$ and $E(\mathcal{C})$. These are the sets of objects and morphisms of $\mathcal{C}$.

**Definition 2.3.4.** *A category* $\mathcal{C}$ *is* ***finite*** *if its class of objects is a finite set and the class of morphisms is also a finite set.*

A finite category $\mathcal{C}$ is a locally small category and a locally small category is a small category. The converses to any of these statements are not true, however.

A subcategory $\mathcal{S}$ of a category $\mathcal{C}$ is a category derived from the same objects and morphisms of category $\mathcal{C}$ while preserving the same properties of $\mathcal{C}$ using $\mathcal{S}$. This is formally defined below:

**Definition 2.3.5.** *For category*

$$\mathcal{C} \triangleq (ob(\mathcal{C}), mor(\mathcal{C}), dom : mor(\mathcal{C}) \rightarrow ob(\mathcal{C}), codom : mor(\mathcal{C}) \rightarrow ob(\mathcal{C}), \circ) \tag{2.29}$$

*a* ***subcategory*** $\mathcal{S} \subseteq \mathcal{C}$ *has:*

$ob(\mathcal{S}) \subseteq ob(\mathcal{C})$ *as a subclass and* $mor(\mathcal{S}) \subseteq mor(\mathcal{C})$ *as a subclass and:*

- $\forall a \in ob(\mathcal{S}), id_a \in mor(\mathcal{S})$
- $\forall f : X \to Y, f \in mor(\mathcal{S})$ *both* $X, Y \in ob(\mathcal{S})$
- $\forall f, g \in mor(\mathcal{S}), f \circ g \in mor(\mathcal{S})$

**Definition 2.3.6.** *Given a locally small category $\mathcal{C}$, the **set** of all subcategories of $\mathcal{C}$ is denoted $2^{\mathcal{C}}$.*

Two categories $\mathcal{C}, \mathcal{D}$ can naturally form a category called the product category.

**Definition 2.3.7.** *The **product category** $\mathcal{C} \times \mathcal{D}$ has objects being the pairs of objects from $ob(\mathcal{C})$ and $ob(\mathcal{D})$.*

*The morphisms are the pairs $(f, g) : (a, b) \to (c, d)$ where $f, g \in mor(\mathcal{C}), mor(\mathcal{D})$ which satisfy:*

$$(f_2, g_2) \circ (f_1, g_1) = (f_2 \circ f_1, g_2 \circ g_1) \tag{2.30}$$

*The identity morphism is defined as*

$$id_{(a,b)} \triangleq (\mathrm{id}_a, \mathrm{id}_b), a \in ob(\mathcal{C}), b \in ob(\mathcal{D}) \tag{2.31}$$

### 2.3.8 The Morphisms of a Category:

We define morphisms of a category that satisfy particular properties.

Morphisms that generalize injective functions over sets are given a special name:

**Definition 2.3.9.** *For a category $\mathcal{C}$, $f \in hom(a, b)$ for $hom(a, b) \subseteq mor(\mathcal{C})$,*

*The morphism $f$ is a **monomorphism** if it is left-cancellative:*

$$f \circ g_1 = f \circ g_2 \Rightarrow g_1 = g_2, \forall g_1, g_2 \in mor(c, a) \tag{2.32}$$

*Denote the class of monomorphisms by $Mono(\mathcal{C})$.*

(2.33) For the case of the category of finite sets: **FinSet**, Mono(**FinSet**) is the set of all injective functions over finite sets.

Morphisms on the same object are given a special name.

**Definition 2.3.10.** *For a category $\mathcal{C}$, $a \in ob(\mathcal{C})$,*

*Any morphism $f : a \to a$ is called an **endomorphism**. These morphisms belong to $hom(a,a) \subseteq mor(\mathcal{C})$, which we denote as $Endo(a)$.*

A special morphism that is used to represent a form of equality between objects is called an isomorphism:

**Definition 2.3.11.** *For a category $\mathcal{C}$, $a, b \in ob(\mathcal{C})$.*

*A morphism $f \in hom(a,b) \subseteq mor(\mathcal{C})$ is an **isomorphism** if there exists a morphism $g : b \to a$ such that $f \circ g = id_b$ and $g \circ f = id_a$*

*The set of such morphisms is denoted $Iso(a,b)$.*

Any morphism that is both an endomorphism and an isomorphism is given a special name:

**Definition 2.3.12.** *For a category $\mathcal{C}$, $a \in ob(\mathcal{C})$*

*A morphism $f \in Iso(a,a)$ is called an **automorphism**. We rename $Iso(a,a)$ by $Aut(a)$.*

Automorphisms are helpful since they define symmetries on individual objects in the category, meaning that for object $a \in \text{ob}(\mathcal{C})$, composing with an automorphism $f \in Aut(a)$ does not change the object $a \in \text{ob}(\mathcal{C})$. We will use this concept on data viewed as objects.

For a category $\mathcal{C}$ we can define subobjects of an object $o \in \text{ob}(\mathcal{C})$ through monomorphisms:

**Definition 2.3.13.** *For an object $o \in ob(\mathcal{C})$, a subobject is the following equivalence class of monomorphisms from $Mono(\mathcal{C})$*

$$(m_a : a \to o) \sim (m_b : b \to o) \text{ monomorphisms iff } \exists \phi : a \to b, m_b \circ \phi = m_a \tag{2.34}$$

(2.35) For the category of all sets: **Set**, the subobjects of a set $T$ are determined by all the injective maps $m : S \to T$ with codomain $T$.

(2.36) For the subcategory of all sets: **Set** that have only inclusion maps as the morphisms, the subobjects of a set $T$ are determined by all the subsets $S \subseteq T$.

Given a category $\mathcal{C}$, this allows us to define the category of subobjects of an object $C \in \text{ob}(\mathcal{C})$:

**Definition 2.3.14.** *Let $\mathcal{C}$ be a category.*

*The category $\boldsymbol{Sub}_{\mathcal{C}}(C)$ has objects consisting of subobjects of $C$ and morphisms between subobjects $m_a, m_b$ of the form:*

$$f : a \to b \text{ s.t. } m_b \circ f = m_a \tag{2.37}$$

### 2.3.15 Functors

Between two categories we may define a "translation" between the two categories. This is defined via a functor, which can be intuitively thought of as a map between categories that preserves the arrows. This is defined formally below:

**Definition 2.3.16.** *Given two categories $\mathcal{C}, \mathcal{D}$, a (covariant/contravariant) functor $F : \mathcal{C} \to \mathcal{D}$ is function between mor($\mathcal{C}$) and mor($\mathcal{D}$) where:*

- $F(a) \in \mathcal{D}, \forall a \in \mathcal{C}$
- $F(\text{id}_a) = id_{F(a)}$
- *covariant:* $F(g \circ f) = F(g) \circ F(f), \forall f : a \to b, g : b \to c$
- *OR contravariant:* $F(g \circ f) = F(f) \circ F(g), \forall f : a \to b, g : b \to c$

When not stated explicitly, a functor is assumed to be covariant.

(2.38) A useful example of a functor is the **forgetful functor**. Given a category $\mathcal{C}$ and the category **Set**, let $\text{Forget} : \mathcal{C} \to \textbf{Set}$ be defined as follows:

1. $\text{Forget}(o) \triangleq \{x : x \in o\}, o \in \text{ob}(\mathcal{C})$
2. $\text{Forget}(f) : \text{Forget}(a) \to \text{Forget}(b)$
   s.t. $(\text{Forget}(f) : x \mapsto f(x) : x \in a), f \in \text{hom}(a, b) \subseteq \text{mor}(\mathcal{C})$

This is a covariant functor.

We can compose two functors $F : \mathcal{C} \to \mathcal{D}, G : \mathcal{D} \to \mathcal{E}$, this defines functor composition:

**Definition 2.3.17.** *The composition* $G \circ F$ *of two covariant functors* $F : \mathcal{C} \to \mathcal{D}, G : \mathcal{D} \to \mathcal{E}$ *is defined by:*

1. $G \circ F(o) \triangleq G(F(o)), \forall o \in ob(\mathcal{C})$
2. $G \circ F(f) \triangleq G(F(f)), \forall f \in hom(a, b) \subseteq mor(\mathcal{C})$

**Proposition 2.3.18.** *The composition* $G \circ F$ *is a covariant functor if both* $G : \mathcal{D} \to \mathcal{E}, F : \mathcal{C} \to \mathcal{D}$ *are covariant.*

*Proof.* We can check that:

$$G \circ F(o) \in \mathcal{E}, \forall o \in \mathcal{C} \tag{2.39}$$

If both $F$ and $G$ are covariant, we can check the composition property:

$$G \circ F(g \circ f) = G \circ (F(g) \circ F(f)) = G(F(g)) \circ G(F(f)) \tag{2.40}$$

as well as the identity property:

$$G \circ F(\mathrm{id}_{\mathcal{C}}) = G(\mathrm{id}_{\mathcal{D}}) = \mathrm{id}_{\mathcal{E}} \tag{2.41}$$

□

A functor can have an **image**, in analogy to the image of a function on sets:

**Definition 2.3.19.** *Let* $F : \mathcal{C} \to \mathcal{D}$ *be a functor. The* ***image*** *of a functor* $F$*, denoted im*$(F)$ *is the subcategory of* $\mathcal{D}$ *with objects consisting of* $F(u), u \in \mathcal{C}$ *and morphisms of the form:* $F(f) : F(u) \to F(v), f : u \to v \in hom(u, v)$.

According to the definition of $\mathrm{im}(F)$, the objects in $\mathrm{ob}(\mathrm{im}(F))$ do not equate from equality in $\mathcal{D}$. So, for example, if $F(u) = F(v)$ in $\mathcal{D}$, we do not view them as equivalent in $\mathrm{im}(F)$. This means that $\mathrm{im}(F)$ preserves the arrow structure of $\mathcal{C}$.

**Definition 2.3.20.** *For a functor* $F : \mathcal{C} \to \mathcal{D}$ *and a subcategory* $Q(\mathcal{C})$ *of* $\mathcal{C}$*, we can define a new functor*

$$F\,|_{Q(\mathcal{C})} : Q(\mathcal{C}) \to \mathcal{D} \tag{2.42}$$

*called the restriction of $F$ to $Q(\mathcal{C})$.*

The restriction of $F$ to $Q(\mathcal{C})$ forms a functor by inheriting the functor properties of $F$ on the subcategory $Q(\mathcal{C})$.

We define acyclic categories from Combinatorial Algebraic Topology in the following:

**Definition 2.3.21.** *([13]) A locally small category $\mathcal{C}$ is acyclic if only identity morphisms have inverses, and any morphism from an object to itself is an identity.*

This is equivalent to saying that for any $X \in \text{ob}(\mathcal{C})$ there does not exist a sequence of morphisms $f_1 : X \to X_2, ..., f_n : X_n \to X,$ with $f \triangleq f_1 \circ ... \circ f_n \in \text{mor}(Q(\mathcal{C}))$.

For a locally small acyclic category $\mathcal{C}$, we can view $\mathcal{C}$ as a pair of sets $(V, E)$ where $V$ is a set of objects and $E$ is a set of morphisms between any two objects. This pair $(V, E)$ can be viewed as an acyclic quiver [14], or multi-directed acyclic graph (see Definition 2.8.19).

Thus, in analogy to simple directed graphs, see Definition 2.8.7, we can define a **simple** acyclic category:

**Definition 2.3.22.** *A locally small category $\mathcal{C}$ is a **simple** acyclic category if it is:*

1. *Acyclic and*

2. *(Simple) For any two objects $a, b \in ob(\mathcal{C})$ where there is no $c \in ob(\mathcal{C}), c \neq a, c \neq b$ with $f : a \to c, g : c \to b \in mor(\mathcal{C})$, then:*

$$|hom(a, b)| \leq 1 \tag{2.43}$$

The **simplicity assumption** of Equation 2.43 for a simple acyclic category prevents multiple morphisms between two spaces when none of these morphisms can factor into a composition of two morphisms. In contrast, a posetal, or thin category is a small category whose objects form a **poset** through the morphisms. In other words, a thin category has commuting morphisms.

**Definition 2.3.23.** *A small category $\mathcal{C}$ is called **posetal**, or **thin** if for $a, b \in ob(\mathcal{C})$ and any $f, g \in hom(a, b)$,*

$$a \underset{g}{\overset{f}{\rightrightarrows}} b \tag{2.44}$$

*then* $f = g$.

**Definition 2.3.24.** *Let* $D : J \to \mathcal{U}$ *be a covariant functor where* $J$ *is a category called the indices, a* ***diagram*** *is the subcategory* $D(J)$ *of* $\mathcal{U}$. *The functor* $D$ *is called a diagram of type* $J$.

For a category $\mathcal{C}$, we can "thin out" the category $\mathcal{C}$ by forcing the posetal condition on all morphisms, this is denoted thin$(\mathcal{C})$.

If the index $J$ category is posetal, we say the diagram **commutes**. Otherwise, we will be explicit about a diagram **not** commuting. We define what the commuting property means formally below:

**Definition 2.3.25.** *For a diagram* $D$ *of type* $J$, *it must* ***commute***, *which means:*

*For any two objects* $o_1, o_n \in ob(U)$ *any composition of arrows* $f = f_n \circ \cdots \circ f_1, g = f_n' \circ \cdots \circ f_1'$ *with* $f_k : o_k \to o_{k+1}$ *and* $f_k' : o_k' \to o_{k+1}'$, *and* $o_1' = o_1, o_n' = o_n$, *then* $f = g$

Usually a diagram is only partially shown depending on the commuting property of interest. The category of its objects and the posetal category $J$ are usually implicit from the partially shown diagram.

A functor $F : \mathcal{C} \to \mathcal{D}$ can be visualized, with slight abuse of terminology, as a functor diagram on spaces over any pair of $X, Y \in \text{ob}(\mathcal{C})$:

$$\begin{array}{ccc} X & \xrightarrow{f} & Y \\ \downarrow & & \downarrow \\ F(X) & \xrightarrow[F_{X,Y}(f)]{} & F(Y) \end{array} \tag{2.45}$$

where $F_{X,Y}(f) : \hom_{\mathcal{C}}(X,Y) \to \hom_{\mathcal{D}}(F(X), F(Y))$ is the map from morphisms between $X, Y$ and $F(X), F(Y)$. The map $F_{X,Y}(f)$ is called the map **induced** by $f$ through the functor $F$.

For a functor $F : \mathcal{C} \to \mathcal{D}$, the category $\mathcal{C}$ is often called the "upstairs" category to the "downstairs" category $F(\mathcal{C})$.

A functor that can injectively map morphisms from one category to another is given a special name:

**Definition 2.3.26.** *A functor* $F : \mathcal{C} \to \mathcal{D}$ *between two categories* $\mathcal{C}, \mathcal{D}$ *is called* ***faithful*** *if* $\forall X, Y \in ob(\mathcal{C}), F_{X,Y}$ *is injective. Where* $F_{X,Y}$ *is the map that takes the morphism* $f : X \to Y$ *to the induced morphism* $F(f) : F(X) \to F(Y)$ *as shown in diagram 2.45.*

Similarly, a functor that can surjectively map to all morphisms in another category is given a special name:

**Definition 2.3.27.** *A functor* $F : \mathcal{C} \to \mathcal{D}$ *between two categories* $\mathcal{C}, \mathcal{D}$ *is called* ***full*** *if* $\forall X, Y \in ob(\mathcal{C}), F_{X,Y}$ *is surjective. Where* $F_{X,Y}$ *is the map that takes the morphism* $f : X \to Y$ *to the induced morphism* $F(f) : F(X) \to F(Y)$ *as shown in the diagram 2.45.*

Functors that can satisfy both faithfulness and fullness are called **full and faithful**.

Two categories $\mathcal{C}, \mathcal{D}$ are considered **isomorphic** if there are functors $F : \mathcal{C} \to \mathcal{D}, G : \mathcal{D} \to \mathcal{C}$ with:

$$G \circ F = \mathrm{id}_{\mathcal{C}}; \text{ and } F \circ G = \mathrm{id}_{\mathcal{D}} \tag{2.46}$$

Two functors $F : \mathcal{C} \to \mathcal{D}, G : \mathcal{C} \to \mathcal{D}$ can be mapped to one another through a natural transformation:

**Definition 2.3.28.** *Given two covariant functors* $F : \mathcal{C} \to \mathcal{D}, G : \mathcal{C} \to \mathcal{D}$*, a natural transformation* $\eta : F \Rightarrow G$ *has the property that:*

*For every* $X \in ob(\mathcal{C})$*,* $\eta_X : F(X) \to G(X)$ *is a morphism in* $\mathcal{D}$

*and that* $\eta_X$ *satisfies the following:*

$$\begin{array}{ccc} F(X) & \xrightarrow{\eta_X} & G(X) \\ {\scriptstyle F(f)}\downarrow & & \downarrow{\scriptstyle G(f)} \\ F(Y) & \xrightarrow[\eta_Y]{} & G(Y) \end{array} \tag{2.47}$$

*If* $F, G$ *are contravariant, then we flip the direction of the arrows* $F(f), G(f)$*.*

**Definition 2.3.29.** *A natural transformation is called a* ***natural isomorphism****, or an isomorphism between functors, if for every* $X \in ob(\mathcal{C}), \eta_X$ *is an isomorphism in* $mor(\mathcal{D})$*.*

## 2.4 Algebra

Algebra is a branch of math that involves sets with binary operations defined on them, called algebraic structures. These algebraic structures generalize arithmetic over number systems.

### 2.4.1 Monoids

One of the simplest algebraic structures are monoids. These are simply sets with a binary operation along with a designated identity element.

**Definition 2.4.2.** *A **monoid** $(M, \cdot)$ has that $M$ is a set and $\cdot$ is a binary operation that takes two element $a, b \in M$ to form an element $a \cdot b \in M$ has two properties:*

1. *(Associativity)* $a \cdot (b \cdot c) = (a \cdot b) \cdot c$
2. *(Identity Element)* $\exists \mathrm{e} \in M$ *with* $\mathrm{e} \cdot a = a, \forall a \in M$

Due to a monoid's simplicity, most other algebraic structures inherit it's structure. Monoids generalize the natural numbers $\mathbb{N}$ with addition.

### 2.4.3 Groups

A slightly more structured algebraic structure is a group. These are generalizations of the integers.

**Definition 2.4.4.** *A **group** $(G, \cdot)$ consists of a non-empty set $G$ together with a binary operation on $G$, here denoted $\cdot$, that combines any two elements $a, b \in G$ to form an element of $G$. This binary operation satisfies the following axioms:*

- *Associativity:* $\forall a, b, c \in G$, *one has* $(a \cdot b) \cdot c = a \cdot (b \cdot c)$.
- *Identity element: There exists a unique element* $\mathrm{e} \in G$ *such that, for every* $a \in G$, *one has* $\mathrm{e} \cdot a = a$ *and* $a \cdot \mathrm{e} = a$. *It is called the identity element.*

- *Inverse element: For each* $a \in G$*, there exists an element* $b \in G$ *such that* $a \cdot b = \mathrm{e}$ *and* $b \cdot a = \mathrm{e}$*, where* $\mathrm{e}$ *is the identity element. For each* $a$*, the element* $b$ *is unique and denoted* $a^{-1}$*.*

The category **Grp** has objects as groups and morphisms called homomorphisms.

A group is called **abelian** if its binary operation has the commuting property:

$$\forall a, b \in G : a \cdot b = b \cdot a \tag{2.48}$$

These are special objects in the category **Grp**.

**Definition 2.4.5.** *For two groups* $(G_1, \cdot_1), (G_2, \cdot_2)$*, a group* ***homomorphism*** $\phi : (G_1, \cdot_1) \to (G_2, \cdot_2)$ *is a map that satisfies the following commuting diagram:*

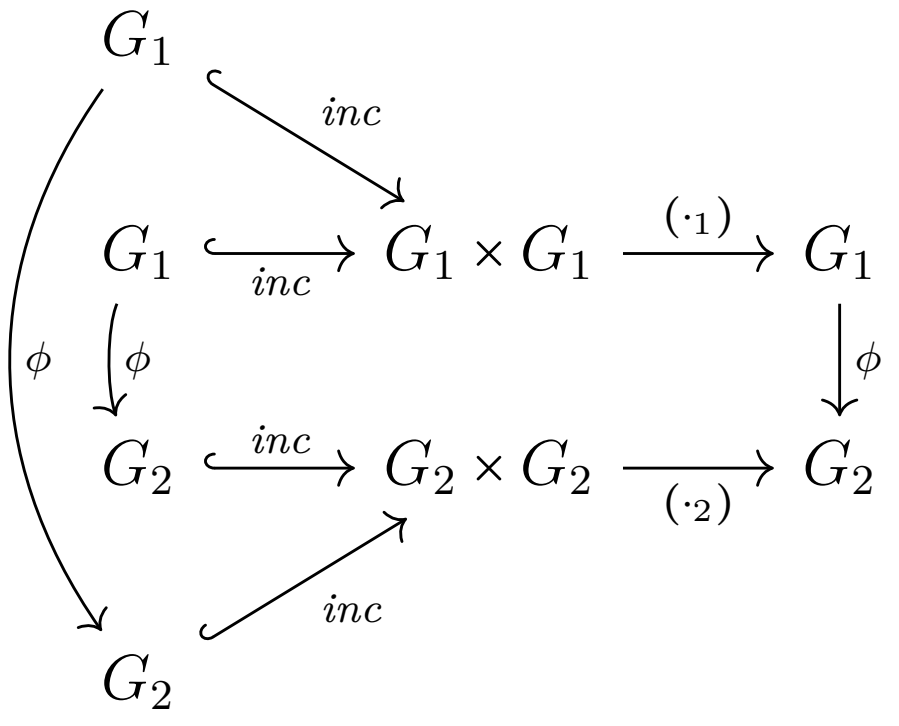

(2.49)

*This can be equivalently written as:*

$$\phi(a \cdot_1 b) = \phi(a) \cdot_2 \phi(b), \forall a, b \in G_1 \tag{2.50}$$

As in category theory, the group isomorphism and group automorphism are defined based on the group homomorphism.

**Definition 2.4.6.** *A* ***subgroup*** $(H, \cdot)$ *of a group* $(G, \cdot)$*, denoted* $H \leq G$ *has* $H \subseteq G$ *as a subset where* $(H, \cdot)$ *is a group with the same binary operation* $\cdot$ *of* $(G, \cdot)$*.*

A special subgroup that has a partial commuting property can be defined. This is helpful for defining an equivalence relation on a group that respects its binary operation:

**Definition 2.4.7.** *A* ***normal subgroup*** $H \le G$ *of the group* $(G, \cdot)$ *has the property that:*

$$\forall a \in G, \exists h, h' \in H \text{ s.t. } a \cdot h = h' \cdot a \tag{2.51}$$

We can thus define a quotient group. These groups act like a "set difference" between sets in set theory while still maintaining group structure.

**Definition 2.4.8.** *A* ***quotient group*** *is defined by a group* $(G, \cdot)$ *and a normal subgroup* $H \le G$*. This is defined as:*

$$\frac{G}{H} \triangleq (\frac{G}{\sim}, \times) \tag{2.52}$$

*where the equivalence relation has* $a \sim b$ *iff* $a \cdot b^{-1} \in H, \forall a, b \in G$.

*Thus, all elements of* $\frac{G}{H}$ *are equivalence classes called cosets, denoted* $gH, \forall g \in G$ *and the binary operation* $\times$ *is defined by:*

$$aH \times bH = (ab)H; \forall aH, bH \in \frac{G}{\sim} \tag{2.53}$$

### 2.4.9 Rings

Given an additive abelian group $(R, +)$, we can introduce a multiplicative binary operation to it. This is in analogy to defining multiplication to the integers with addition. Such a group with a multiplication operation is called a ring.

**Definition 2.4.10.** *A* ***ring*** $(R, +, \cdot)$ *is an abelian group* $(R, +)$ *with an additional multiplication operation which has the property that:*

- *(Distributivity)* $r \cdot (r_1 + r_2) = r \cdot r_1 + r \cdot r_2 \in R, \forall r, r_1, r_2 \in R$
- *(Associativity)* $r_1 \cdot (r_2 \cdot r_3) = (r_1 \cdot r_2) \cdot r_3, \forall r_1, r_2, r_3 \in R$

**Definition 2.4.11.** *A* ***unital*** *ring* $(R, +, \cdot)$ *is a ring with a unit, which we denote by* 1*. It has the property that:*

$$1 \cdot r = r \cdot 1 = r, \forall r \in R \tag{2.54}$$

(2.55) Common examples of rings include the ring of integers $(\mathbb{Z}, +, \cdot)$ and real valued functions $(\mathbb{R}^{\mathbb{R}}, +, \cdot)$ with addition and multiplication in the codomain of reals.

**Definition 2.4.12.** *A* ***commutative ring*** *is a ring where multiplication commutes:*

$$r_1 \cdot r_2 = r_2 \cdot r_1, \forall r_1, r_2 \in R \tag{2.56}$$

The previous examples are both commutative rings.

**Definition 2.4.13.** *A* ***subring*** *of a ring $R$ is a subset $S \subseteq R$ where $S$ is a ring with the two inherited binary operations of $R$ on $S$.*

A looser notion of a subring is an ideal, which we define below:

**Definition 2.4.14.** *An left/right* **ideal** *$I$ in a ring $R$ is an abelian subgroup of $R$ where $I$ is closed under left/right multiplication:*

- *$I \subseteq R$, $I$ is a subgroup of $R$*
- *(multiplicatively left closed) $\forall r \in R, \mathrm{i}_1, \mathrm{i}_2 \in I, r \cdot (\mathrm{i}_1 + \mathrm{i}_2) = r \cdot \mathrm{i}_1 + r \cdot \mathrm{i}_2 \in I$*
- *and/or: (multiplicatively right closed) $\forall r \in R, \mathrm{i}_1, \mathrm{i}_2 \in I, (\mathrm{i}_1 + \mathrm{i}_2) \cdot r = \mathrm{i}_1 \cdot r + \mathrm{i}_2 \cdot r \in I$*

Certain ideals can be generated by an element $r \in R$. A left ideal generated by some $r \in R$ is denoted:

$$(r) \triangleq \{c \cdot r : c \in R\} \tag{2.57}$$

A right ideal can be defined analogously.

The product of two ideals $I_1, I_2 \subseteq R$ is denoted:

$$I_1 \cdot I_2 \triangleq \{ \sum_{\text{finite number of j}} (\mathrm{i}_{1,\mathrm{j}} \cdot \mathrm{i}_{2,\mathrm{j}}) : \mathrm{i}_{1,\mathrm{j}} \in I_1, \mathrm{i}_{2,\mathrm{j}} \in I_2\} \tag{2.58}$$

This forms an ideal of $R$.

For a ring $(R, +, \cdot)$, we can define the ring of **formal polynomials** $R[t]$, in analogy to the real-valued single variable polynomials.

We call $t$ an **indeterminant**, which is a variable that cannot belong in the ring $R$. It can be powered to form higher power indeterminates of the form $t^k$. Indeterminates can also carry coefficients to form **monomials** $r \cdot t^k, r \in R$.

**Definition 2.4.15.** *The ring of formal polynomials are thus:*

$$R[t] \triangleq \{ \sum_{i \in \mathbb{N} \cup \{0\}} c_i t^i : c_i \neq 0 \text{ for only finitely many } i, c_i \in R \} \tag{2.59}$$

*where we can define the addition operation by:*

$$\sum_{i \in \mathbb{N} \cup \{0\}} c_i t^i + \sum_{i \in \mathbb{N} \cup \{0\}} d_i t^i = \sum_{i \in \mathbb{N} \cup \{0\}} (c_i + d_i) t^i \tag{2.60}$$

*and multiplication by:*

$$( \sum_{i \in \mathbb{N} \cup \{0\}} c_i t^i) \cdot ( \sum_{j \in \mathbb{N} \cup \{0\}} d_j t^j) = \sum_{i,j \in \mathbb{N} \cup \{0\}} (c_i \cdot d_j) t^{i+j} \tag{2.61}$$

*where the coefficients from $R$ can be added and multiplied according to the operations in the ring $R$.*

The polynomial ring $R[t]$ must form a **grading** by monomial degree, meaning that

$$\begin{gathered} R[t] = \bigoplus_{k \in \mathbb{N} \cup \{0\}} (t^k) \text{ s.t.:} \\ (t^i) \cdot (t^j) \subseteq (t^{i+j}), i, j \in \mathbb{N} \cup \{0\} \end{gathered} \tag{2.62}$$

where the direct sum simply separates out the monomials by the formal addition defined for $R[t]$ and the product respects the degree of the monomials.

We can compose the process of forming a polynomial ring by introducing indeterminants in sequence. Starting from a ring $R$, we can define:

$$R[t_1, t_2] \triangleq (R[t_1])[t_2] \tag{2.63}$$

And thus, for any $k \in \mathbb{N}$:

$$R[t_1, t_2, ..., t_k] \triangleq (R[t_1, ..., t_{k-1}])[t_k] \tag{2.64}$$

is called the ring on $k$ indeterminants over $R$.

We define a common kind of ring with a finiteness condition for its ideals:

**Definition 2.4.16.** *A commutative ring $R$ is* ***Noetherian*** *iff for any sequence of ideals ordered by inclusion:*

$$I_1 \subseteq \ldots \subseteq I_k \subseteq \ldots \tag{2.65}$$

*there exists a $n \in \mathbb{N}$ with $I_n = I_{n+1} = \ldots$*

(2.66) The ring of integers $(\mathbb{Z}, +, \cdot)$ is Noetherian.

**Theorem 2.4.17.** *([15] Hilbert's Basis Theorem)*

*If $R$ is Noetherian, then $R[t]$ is also Noetherian.*

(2.67) By the Hilbert's Basis Theorem, the ring $\mathbb{Z}[t_1, \ldots, t_k]$ is Noetherian.

### 2.4.18 Fields

With more structure a ring can form a field. A field is a group with another invertible binary operation. This is in analogy to the formation of the rational numbers from the integers.

**Definition 2.4.19.** *A* ***field*** *$(F, +, \times)$ is an algebraic structure that has two operations $+, \times$ satisfying the following axioms.*

- *Associativity of addition and multiplication: $a + (b + c) = (a + b) + c$, and $a \times (b \times c) = (a \times b) \times c$.*
- *Commutativity of addition and multiplication: $a + b = b + a$, and $a \times b = b \times a$.*
- *Additive and multiplicative identity: there exist two distinct elements $0$ and $1$ in $F$ such that $a + 0 = a$ and $a \times 1 = a$.*
- *Additive inverses: for every $a$ in $F$, there exists an element in $F$, denoted $-a$, called the additive inverse of $a$, such that $a + (-a) = 0$.*
- *Multiplicative inverses: for every $a \neq 0$ in $F$, there exists an element in $F$, denoted by $a^{-1}$ or $1 \div a$, called the multiplicative inverse of $a$, such that $a \times a^{-1} = 1$.*

- *Distributivity of multiplication over addition:* $a \times (b + c) = (a \times b) + (a \times c)$.

*In other words,* $(F, +)$ *is a group with* $0$ *the identity element and where* $+$ *is commutative. Furthermore,* $(F, \times)$ *is also a group with* $1$ *the identity element and* $\times$ *is commutative. Multiplication distributes over addition.*

A field is in analogy to the arithmetic of the real number system. Some common fields that we use include:

(2.68)
1. The field of real numbers $\mathbb{R}$ with the usual arithmetic operations of $+, -, \times, \div$
2. The field of rational numbers $\mathbb{Q}$ with the usual arithmetic operations of $+, -, \times, \div$ on the reals restricted to fractions of integers.
3. The field $\mathbb{Z}_p \triangleq \{0, ..., p-1\}$ with the arithmetic of the integers modulo $p \in \mathbb{Z}$, a prime:
   - $x + y = (x + y \mod p), x, y \in \mathbb{Z}_p$
   - $x - y = (x - y + p \mod p), x, y \in \mathbb{Z}_p$
   - $x \times y = (x \times y \mod p), x, y \in \mathbb{Z}_p$
   - $x \div y = (x \times y^{-1} \mod p), x, y \in \mathbb{Z}_p$

   where $y^{-1} \in \mathbb{Z}_p$ is the unique solution to the equation $y \times y^{-1} \equiv 1 \mod p$. This solution exists uniquely if $p$ is prime.

   Where the modulo operation on integers is defined by:

$$(a \mod p) \triangleq \begin{cases} \max_k (a - k \times p) \\ \text{s.t. } 0 \leq a - k \times p \leq p - 1 \end{cases} \tag{2.69}$$

Of the previous examples, we have

$$\mathbb{Z}_p \subseteq \mathbb{Q} \subseteq \mathbb{R} \tag{2.70}$$

as sets. Thus any of these fields can be represented by a floating point type. Of course, $\mathbb{Z}_p$ does not respect the arithmetic operations of $\mathbb{R}$ and thus the algebraic structure of $\mathbb{Z}_p$ is not preserved in $\mathbb{R}$.

### 2.4.20 Vector Spaces

A vector space is another common algebraic structure. This is in analogy to Euclidean space, see Section 2.5 which is closed under scaling and translation. A vector space can be formalized algebraically as follows:

**Definition 2.4.21.** *A vector space $(V, F, +, \cdot)$ is defined by a group of vectors $V$ with the addition operation $+$ and a field of scalars $(F, +_F, \times_F)$ that act on the group of vectors through scalar multiplication $\cdot$. The additive group $V$ and the action by $F$ is defined by the following axioms:*

- *Associativity of vector addition: $u + (v + w) = (u + v) + w$*
- *Commutativity of vector addition $u + v = v + u$*
- *Identity element of vector addition: There exists an element $\overrightarrow{0} \in V$, called the zero vector, such that $v + \overrightarrow{0} = v$ for all $v \in V$.*
- *Inverse elements of vector addition: For every $v \in V$, there exists an element $-v \in V$, called the additive inverse of $v$, such that $v + (-v) = \overrightarrow{0}$.*
- *Closed under Linear Combinations: For any $n \in \mathbb{N}$, $\sum_{\mathrm{i}=1}^{n} c_{\mathrm{i}} \cdot v_{\mathrm{i}} \in V$ for $c_{\mathrm{i}} \in F, v_{\mathrm{i}} \in V, \mathrm{i} = 1, \ldots, n$.*
- *Compatibility of scalar multiplication with field multiplication: $a \cdot (b \cdot v) = (a \times_F b) \cdot v$*
- *Identity element of scalar multiplication: $1 \cdot v = v$, where $1$ denotes the multiplicative identity in $F$.*
- *Distributivity of scalar multiplication with respect to vector addition: $a \cdot (u + v) = a \cdot u + a \cdot v, \forall a \in F$*

We use the notation $\mathbf{0} \triangleq \{\overrightarrow{0}\}$ is the $F$-vector space consisting of the zero vector alone.

Often times we denote a vector space simply by $V$, where $F$ and the binary operations are implicit. Otherwise, when the field is ambiguous $V$ is called an $F$-vector space.

**Definition 2.4.22.** *The category of vector spaces under the field* $\mathbb{R}$ *is denoted* ***Vec****.*

1. *Its objects are the vector spaces over the reals*
2. *Its morphisms are linear maps between vector spaces*

*When not stated, we will assume that all the objects in* ***Vec*** *are finite dimensional. This makes the category locally small.*

**Definition 2.4.23.** *A map* $L : V \to W$ *between* $F$*-vector spaces* $V, W$ *is a linear map if it satisfies* ***linearity****:*

$$L(\sum_i c_i \cdot v_i) = \sum_i c_i \cdot L(v_i), \forall c_i \in F, \forall v_i \in V \tag{2.71}$$

**Definition 2.4.24.** *A subset* $W$ *of vector space* $V$ *that is still a vector space is called a subvector space, or* ***linear subspace****.*

We can combine linear subspaces of $V$ to form more linear subspaces of $V$.

**Definition 2.4.25.** *For two linear subspaces* $V_1, V_2 \subseteq V$ *of a vector space* $V \in ob(\textbf{Vec})$*, let their sum be defined by:*

$$V_1 + V_2 \triangleq \{v_1 + v_2 : v_1 \in V_1, v_2 \in V_2\} \tag{2.72}$$

$V_1 + V_2$ *is a linear subspace of* $V$*, as can be checked.*

Similarly,

**Definition 2.4.26.** *For two linear subspaces* $V_1, V_2 \subseteq V$ *of a vector space* $V \in ob(\textbf{Vec})$*, let their difference be defined by:*

$$V_1 - V_2 \triangleq \{v_1 - v_2 : v_1 \in V_1, v_2 \in V_2\} \tag{2.73}$$

$V_1 - V_2$ *is a linear subspace of* $V$*, as can be checked.*

**Definition 2.4.27.** *For two linear subspace* $V_1, V_2 \subseteq V$ *of a vector space* $V \in ob(\textbf{Vec})$*, let their intersection be defined by:*

$$V_1 \cap V_2 \triangleq \{v : v \in V_1, v \in V_2\} \tag{2.74}$$

$V_1 \cap V_2$ *is a linear subspace of* $V$*, as can be checked.*

We can also take complements with respect to the sum operation.

**Definition 2.4.28.** *For a linear subspace* $U \subseteq V$ *of a vector space* $V \in ob(\textbf{\textit{Vec}})$*, let the* ***complementary subspace*** *of* $V$ *with respect to* $U \subseteq V$ *be defined as the linear subspace* $W \subseteq V$ *that solves the following equation:*

$$U + W = V \text{ and } U \cap W = \mathbf{0} \tag{2.75}$$

*The space* $W$ *is often given by the notation* $V \ominus U$*.*

**Proposition 2.4.29.** *For a vector space* $V$ *and linear subspace* $U \subseteq V$*, the complementary subspace* $V \ominus U$ *is unique.*

*Proof.* Given a linear subspace $W \subseteq V$ with

$$U + W = V \text{ and } U \cap W = \mathbf{0} \tag{2.76}$$

any other linear subspace $W' \subseteq V$ also satifying

$$U + W' = V \text{ and } U \cap W' = \mathbf{0} \tag{2.77}$$

must have

$$U + W = V = U + W' \Rightarrow W = W' \tag{2.78}$$

□

**Definition 2.4.30.** *For two vector spaces* $V, W$*, let the direct sum between* $V, W$ *be defined as follows:*

$$V \bigoplus W \triangleq \{(v, w) \in V \times W\} \tag{2.79}$$

*is a vector space of pairs of vectors with addition and scalar multiplication defined element-wise, meaning:*

$$(c_1 v + c_2 v', c_1 w + c_2 w') = c_1(v, w) + c_2(v', w'), \forall c_1, c_2 \in \mathbb{R}, (v, w), (v', w') \in V \bigoplus W \tag{2.80}$$

*with the zero vector defined as:*

$$(\overrightarrow{0}, \overrightarrow{0}) \tag{2.81}$$

**Proposition 2.4.31.** *If for two vector spaces $V, W$ if we have that the vector space $U = V + W$ and $V \cap W = \boldsymbol{0}$, then $U \cong V \bigoplus W$*

*Proof.* Define the map $\Phi : V \bigoplus W \to U$ with: $(v, w) \mapsto v - w$.

This is an isomorphism since $\ker(\Phi) = \{(v, w) : v - w = \overrightarrow{0}\} = (V \cap W) \bigoplus (V \cap W) = \mathbf{0}$ and $\Phi$ is surjective. Thus by the first isomorphism theorem of linear maps [16], we have that $\Phi$ is an isomorphism. □

As for vector spaces, a direct sum can also be taken between linear maps.

**Definition 2.4.32.** *For two linear maps $L_1 : V_1 \to W_1, L_2 : V_2 \to W_2$ the direct sum of the two linear maps:*

$$L_1 \oplus L_2 : V_1 \bigoplus V_2 \to W_1 \bigoplus W_2 \tag{2.82}$$

*is defined by:*

$$(L_1 \oplus L_2)(v, w) \triangleq (L_1(v), L_2(w)) \tag{2.83}$$

*This is a linear map, as can be checked.*

Proposition 2.4.31 can also apply to linear maps.

**Proposition 2.4.33.** *For a vector space $U$ with a decomposition:*

$$U = V_1 + V_2, V_1 \cap V_2 = \boldsymbol{0} \tag{2.84}$$

*For linear maps $L_1 : V_1 \to W, L_2 : V_2 \to W$, we can define:*

$$L_1 \oplus L_2 : V_1 \bigoplus V_2 \to W \tag{2.85}$$

*by:*

$$(L_1 \oplus L_2)(v_1, v_2) \triangleq L_1(v_1) + L_2(v_2) \tag{2.86}$$

*This is a linear map and can be written up to isomorphism as:*

$$(L_1 \oplus L_2) : U \to W \tag{2.87}$$

*with* $L_1(V_1) \cap L_2(V_2) = \boldsymbol{0}$.

*Proof.* Checking for linearity:

$$(L_1 \oplus L_2)(c_1 \cdot v_1 + c_2 \cdot v_2, c_1 \cdot v_1' + c_2 \cdot v_2') \tag{2.88a}$$

$$= c_1 \cdot L_1(v_1) + c_1 \cdot L_1(v_1') + c_2 \cdot L_2(v_2) + c_2 \cdot L_2(v_2') \tag{2.88b}$$

$$= c_1 \cdot (L_1 \oplus L_2)(v_1, v_1') + c_2 \cdot (L_1 \oplus L_2)(v_2, v_2') \tag{2.88c}$$

We know that any $u \in U$ can be written as $u = v_1 + v_2, v_1 \in V_1, v_2 \in V_2$.

Since $U \cong V_1 \bigoplus V_2$, we must have that the linear map $L_1 \bigoplus L_2$ can be defined on $U$ up to isomorphism via the following commuting diagram:

$$\begin{array}{ccc} V_1 \bigoplus V_2 & \xrightarrow{L_1 \bigoplus L_2} & W \\ \cong & \nearrow & \\ U & & \end{array} \tag{2.89}$$

Furthermore, since $V_1 \cap V_2 = \mathbf{0}$, we must have that $L_1(V_1) \cap L_2(V_2) = \mathbf{0}$. □

**Spanning a Vector Space:**

**Definition 2.4.34.** *For a set $S$, the span of $S$ is the linear space of all linear combinations of vectors from $S$.*

$$span(S) \triangleq \{\sum_{i=1}^{n} c_i \cdot v_i : c_i \in F, v_i \in S, n \in \mathbb{N}\} \tag{2.90}$$

When $S \subseteq V$, then $\text{span}(S) \subseteq V$ is a linear subspace of $V$.

When $V = \text{span}(S)$, we say that $S$ spans $V$.

**Definition 2.4.35.** *We define a category **SpanVec**. This category has:*

1. *Objects consisting of triples $(S, V)$ with $V$ a vector space and $span(S) = V$.*

*2. Morphisms consisting of maps of the form:*

$$(\phi_{V\to W}\mid_S : S \to S', \phi_{V\to W}) : (S,V) \to (S',W) \tag{2.91}$$

*where the map $\phi_{V\to W}$ is a linear map between $V$ and $W$.*

We define a notion of independence in a vector space.

**Definition 2.4.36.** *A finite set of vectors $S \subseteq V$ are called* ***linearly independent*** *if there are no $c_i \neq 0, c_i \in F$ with*

$$\sum_{v_i \in S} c_i \cdot v_i = \overrightarrow{0} \tag{2.92}$$

*An infinite set of vectors are linearly independent if any finite subset is linearly independent.*

*If the set $S$ is not linearly independent, then we call the set* ***linearly dependent****.*

We also use the notation

$$v_1 \perp_V \ldots \perp_V v_{|S|} \tag{2.93}$$

to denote that the subset of vectors $\{v_1, \ldots, v_{|S|}\} \subseteq V$ from vector space $V$ are linearly independent.

**Definition 2.4.37.** *A set of vectors $S$ that is both linearly independent and spans a vector space $V$ is called a* ***basis*** *of $V$.*

The cardinality of a basis of a vector space $V$ is called the dimension of $V$. This is denoted by $\dim(V)$.

**Proposition 2.4.38.** *The dimension of a vector space is invariant to the choice of basis.*

*Proof.* Proof by contradiction:

Say there are two bases $S_V$ and $S'_V$ with $|S_V| < |S'_V|$.

Since both $S_V$ and $S'_V$ are bases, $\text{span}(S_V) = \text{span}(S'_V) = V$.

Since $|S_V| < |S'_V|$, there must exist a $v \in S'_V \setminus S_V$.

Thus, $\text{span}(S'_V) \neq \text{span}(S_V)$ since $v \notin \text{span}(S_V)$ but $v \in \text{span}(S'_V)$. However this contradicts that $\text{span}(S_V) = \text{span}(S'_V) = V$.

Thus, $\dim(V)$ is independent of basis. □

Every finite dimensional vector space must have a basis. If we specify this basis along with the vector space, we can form the following category.

**Definition 2.4.39.** *We define a category* ***BasisVec****. This category has:*

1. *Objects consisting of pairs $(B, V)$ with a $V$ a vector space and $B$ is a finite basis of $V$.*
2. *Morphisms consisting of maps of the form:*

$$(\phi_{V\to W}\,|_B : B \to B', \phi_{V\to W}) : (B, V) \to (B', W) \tag{2.94}$$

*where the map $\phi_{V\to W}$ is a linear map between $V$ and $W$.*

It can be checked that **BasisVec** is a subcategory of **SpanVec**. This follows since a basis of a vector space is a spanning set of that same vector space.

**Proposition 2.4.40.** *The dimension is additive on a direct sum of vector spaces $V, W$:*

$$dim(V) + dim(W) = dim(V \bigoplus W) \tag{2.95}$$

*Proof.* This follows since if $\mathcal{B}_V, \mathcal{B}_W$ are the bases of $V, W$. There is a basis of

$$V \bigoplus W \tag{2.96}$$

of the form

$$\mathcal{B}_{V\oplus W} \triangleq \{(b_V, \vec{0}) : b_V \in \mathcal{B}_V\} \cup \{(\vec{0}, b_W) : b_W \in \mathcal{B}_W\} \tag{2.97}$$

Certainly,

$$\{(b_V, \vec{0}) : b_V \in \mathcal{B}_V\} \cap \{(\vec{0}, b_W) : b_W \in \mathcal{B}_W\} = \{(\vec{0}, \vec{0})\} \tag{2.98}$$

Thus by additivity of disjoint sets:

$$\dim(V) + \dim(W) = \dim(V \bigoplus W) \tag{2.99}$$

□

A standard theorem of linear algebra is that linear maps are determined by their behavior on bases. This is stated below:

**Theorem 2.4.41.** *A linear map $L : V \to W$ between finite dimensional vector spaces $V$ and $W$ is determined by how it maps a basis vector from $V$ into $W$.*

*Proof.* Let $S_V$ be a basis of $V$.

If $L(b_i) = d_i \in W$ for each $b_i \in S_V$, then:

**1. $d_i$ are linearly independent:**

We check that $\sum_{b_i \in S_V} c_i \cdot b_i = \overrightarrow{0}, c_i \in F$ iff $c_i = 0$ for all $i = 1, ..., \dim(V)$.

By linearity of $L$:

$$\sum_{b_i \in S_V} c_i \cdot b_i = \sum_{b_i \in S_V} c_i \cdot L(b_i) = L(\sum_{b_i \in S_V} c_i \cdot b_i) = \overrightarrow{0} \Leftrightarrow \sum_{b_i \in S_V} c_i \cdot b_i = \overrightarrow{0} \tag{2.100}$$

Since $S_V$ is a basis, $S_V$ is linearly independent thus:

$$\sum_{b_i \in S_V} c_i \cdot b_i = \overrightarrow{0} \text{ iff } c_i = 0 \text{ for all } i = 1, ..., \dim(V) \tag{2.101}$$

**2. Any vector $v \in V$ maps to linear combinations of $b_i \in S_V$:**

For $v \in V$, we can write it as $v = \sum_{b_i \in S_V} c_i \cdot b_i$ since $S_V$ forms a basis.

By linearity of $L$:

$$L(\sum_{b_i \in S_V} c_i \cdot b_i) = \sum_{b_i \in S_V} c_i \cdot L(b_i) = \sum_{i} c_i \cdot b_i \in W \tag{2.102}$$

This determines the image of $L$ on $V$.

This shows that knowing the behavior of $L$ on $S_V$ determines $L$ itself. □

Let $V \in \text{ob}(\mathbf{Vec})$ be a vector space. For a linear subspace $U \subseteq V$, we can define the following two maps:

**Definition 2.4.42.** *The map lin-inc $: U \hookrightarrow V$ is the linear map that acts as the inclusion map on some basis of $U$.*

The map lin-inc $: U \hookrightarrow V$ is well-defined by Theorem 2.4.41.

**Definition 2.4.43.** *The linear projection map* $\pi_U : V \to U$ *is the linear map satisfying:*

$$\pi_U(v) \triangleq \begin{cases} id & \text{if } v \in U \\ \overrightarrow{0} & \end{cases} \tag{2.103}$$

The rank of a linear map is defined by the rank of its image. This is defined below:

**Definition 2.4.44.** *For* $V, W \in ob(\textbf{Vec})$ *with both* $rank(V), rank(W) < \infty$

*Let* $L : V \to W$ *be a linear map, then:*

$$rank(L) \triangleq rank(im(L)) \tag{2.104}$$

We can use these maps to extend from subspaces not just bases.

**Proposition 2.4.45.** *(Subspace Extension on Linear Maps)*

*Let* $V \subseteq \tilde{V}, W \subseteq \tilde{W}$ *as linear subspaces. Any linear map* $L : V \to W$ *can be extended to its* ***linear extension map*** $\tilde{L} : \tilde{V} \to \tilde{W}$ *with* $\tilde{L}\,|_V = (lin\text{-}inc : W \hookrightarrow \tilde{W}) \circ L$*.*

*Proof.* For the morphisms $f : V \to W$, we can uniquely form the map $\tilde{f} : \tilde{V} \to \tilde{W}$ by composition with:

$$\tilde{f} \triangleq (\text{lin-inc} : W \hookrightarrow \tilde{W}) \circ f \circ \pi_V \tag{2.105}$$

As a composition of linear maps, $\tilde{f} : \tilde{V} \to \tilde{W}$ must be a linear map. □

Another interesting space determined by a linear map is its **kernel**. We define this as follows:

**Definition 2.4.46.** *For two vector spaces* $V, W \in ob(\textbf{Vec})$ *and for a linear map* $L : V \to W$*:*

$$ker(L) \triangleq \{x : L(x) = \overrightarrow{0}\} \tag{2.106}$$

The vectors in the kernel are closed under linear combinations and thus form a linear subspace of $V$.

The **nullity** of a linear map is defined as the rank of the kernel. This is defined below:

**Definition 2.4.47.** *For $V, W \in ob(\textbf{Vec})$ with both $rank(V), rank(W) < \infty$*

*Let $L : V \to W$ be a linear map, then:*

$$nullity(L) \triangleq rank(ker(L)) \tag{2.107}$$

For computational purposes it will be of interest to consider only finite dimensional vector spaces over a field $F$. We denote this category by $F$-**FinDimVec**.

A standard theorem of linear algebra is the rank-nullity theorem. This is stated in the category $F$-**FinDimVec** as follows:

**Theorem 2.4.48.** *([17]) For $V, W \in ob(F\text{-}\textbf{FinDimVec})$ and $L \in hom(V, W)$,*

$$rank(L) + nullity(L) = dim(V) \tag{2.108}$$

*Proof.* Theorem 2 page 71 of [17] □

### Duality

We can translate vectors to linear maps in a one to one fashion. We call these linear maps dual vectors or covectors. The translation happens through a higher order linear map called a dual operator.

**Definition 2.4.49.** *A dual operator $\bullet^{\perp}$ on a basis $\{\mathrm{e_i}\}_{\mathrm{i}=1}^{n}$ of vector space $V$ of dimension $dim(V) = n$ is defined as follows:*

$$\bullet^{\perp} : \mathrm{e_i} \mapsto \phi_{\mathrm{e_i}}, \text{ where } \phi_{\mathrm{e_i}}(\mathrm{e_j}) = \delta_{\mathrm{i,j}}, \text{ for any basis } \{\mathrm{e_i}\}_{\mathrm{i}=1}^{n} \text{ with } span(\{\mathrm{e_i}\}_{\mathrm{i}=1}^{n}) = V \tag{2.109}$$

*where the map $\phi_{\mathrm{e_i}} : V \to \mathbb{R}, \phi_{\mathrm{e_i}} \in Lin(V, \mathbb{R})$ satisfies linearity:*

$$\phi_{\mathrm{e_i}}(\sum_{\mathrm{j}} c_{\mathrm{j}} \cdot \mathrm{e_j}) = \sum_{\mathrm{j}} c_{\mathrm{j}} \cdot \phi_{\mathrm{e_i}}(\mathrm{e_j}), \forall c_{\mathrm{j}} \in \mathbb{R} \tag{2.110}$$

*We call these dual basis vectors $\phi_{\mathrm{e_i}} : V \to \mathbb{R}$ covectors.*

*We can extend the map $\bullet^\perp$ by linearity to get that:*

$$\bullet^\perp(v) = \sum_{\mathrm{i}=1}^{n} c_\mathrm{i} \cdot \phi_{\mathrm{e_i}} \text{ for } v = \sum_{\mathrm{i}=1}^{n} c_\mathrm{i} \cdot \mathrm{e_i}, c_\mathrm{i} \in \mathbb{R} \tag{2.111}$$

We can check that we have the following composition property:

$$(L_2 \circ L_1)^\perp = L_1^\perp \circ L_2^\perp \tag{2.112}$$

Similarly, we can use the covectors to span (still as a vector space) a new space, called the dual vector space:

$$V^\perp \triangleq \mathrm{span}(\{\phi_{\mathrm{e_i}}\}_{\mathrm{i}=1}^{n}) \tag{2.113}$$

If we compose twice, $(\bullet^\perp)^\perp : V^\perp \to (V^\perp)^\perp$, we recover $V$ up to isomorphism. This isomorphism $(V^\perp)^\perp \cong V$ follows according to [18].

We say that the operator $\bullet^\perp$ is an **involution**. It is its own inverse. This means that $\bullet^\perp$ is an isomorphism.

We can now define the **annihilator** of a vector space $V$ through the dual space $V^\perp$.

**Definition 2.4.50.** *The **annihilator** is defined by the space of dual vectors that zero the vector space:*

$$Ann(V) \triangleq \{u \in V^\perp : u \cdot v = 0, \forall v \in V\} \tag{2.114}$$

We can also take the **dual** on linear maps $L : V \to W$ as follows:

$$\begin{aligned} L^\perp &: W^\perp \to V^\perp \\ L^\perp(\phi_\mathrm{i}) &\triangleq \phi_\mathrm{i} \circ L \end{aligned} \tag{2.115}$$

These maps are still linear and are called dual linear maps.

We also have that $\text{im}((L^{\perp})^{\perp}) \cong \text{im}(L)$. This follows by the following non-commuting diagram:

$$\begin{array}{ccc} V & \xrightarrow[L]{} & W \\ {\scriptstyle \bullet^{\perp}}\updownarrow & & \updownarrow{\scriptstyle \bullet^{\perp}} \\ V^{\perp} & \xleftarrow[L^{\perp}]{} & W^{\perp} \\ {\scriptstyle \bullet^{\perp}}\updownarrow & & \updownarrow{\scriptstyle \bullet^{\perp}} \\ (V^{\perp})^{\perp} & \xrightarrow[(L^{\perp})^{\perp}]{} & (W^{\perp})^{\perp} \end{array} \tag{2.116}$$

**The Dual Functor:**

We thus have a category for the dual vector spaces and the dual maps. We call this category the dual vector space category.

**Definition 2.4.51.** *The dual vector space category* $\textbf{\textit{Vec}}^{\perp}$ *has:*

1. *Objects consisting of dual vector spaces*

2. *Morphisms consisting of dual linear maps*

We can then relate **Vec** and $\textbf{Vec}^{\perp}$ with the functor generalization of the dual operator on vector spaces.

**Definition 2.4.52.** *The* ***dual functor****:*

$$\bullet^{\perp} : \textbf{\textit{Vec}} \to \textbf{\textit{Vec}}^{\perp} \tag{2.117}$$

*from vector spaces to dual vector spaces. This is defined as follows:*

1. *On objects:* $\bullet^{\perp}(V) \triangleq V^{\perp}$

2. *On morphisms:* $\bullet^{\perp}(L : V \to W) \triangleq (L^{\perp} : W^{\perp} \to V^{\perp})$

*This is a contravariant functor.*

Since the dual operators on vector spaces and morphisms are involutions, we have that the dual functor is as well.

### Column and Matrix Representation

In the following, we define a computationally useful way to represent vector spaces and the maps between them.

A **column** vector is defined by a finite sequence of elements from a field $F$. The length of this sequence is called the length of a column. A column of length $d$ thus belongs to $F^d$. The ith entry of a column vector $v \in F^d$ is denoted $v[\mathrm{i}]$.

Let the category $F$-**Mat** have objects consisting of vector spaces spanned by column vectors and morphisms of linear maps represented as matrices.

For integer $d$, the vector space $F^d$ consists of column vectors over the field $F$. Let the **standard basis** vectors $\mathrm{e_i} \in F^d$ be defined by:

$$\mathrm{e_i}[\mathrm{j}] \triangleq \begin{cases} 1 & \mathrm{j} = \mathrm{i} \\ 0 & \mathrm{j} \neq \mathrm{i} \end{cases} \tag{2.118}$$

Certainly these form a basis of $F^d$.

A **matrix** is a table of field elements indexed by a pair of integers. A matrix $M$ at index $\mathrm{i}, \mathrm{j}$ is a field element denoted by $M[\mathrm{i}, \mathrm{j}]$. A $d_2 \times d_1$ matrix $M$ can represent a linear map from $F^{d_1}$ to $F^{d_2}$ via matrix multiplication.

The matrix $M$, viewed as a linear map, can compose on a $d_1$-lengthed column vector $v \in F^{d_1}$ by **matrix multiplication** via the following equation:

$$(M \cdot v)[\mathrm{i}] \triangleq \sum_{\mathrm{j}} M[\mathrm{i}, \mathrm{j}] \cdot v[\mathrm{j}] \tag{2.119}$$

It can be easily checked that $v \mapsto M \cdot v$ is a linear map. In fact, its behavior is determined by the columns $M[\bullet, \mathrm{j}]$ since $M[\bullet, \mathrm{j}] = M \cdot \mathrm{e_j}$ and $\{\mathrm{e_j}\}_{\mathrm{i}=1}^{d_1}$ form a basis.

A **submatrix** $S \in F^{d_2' \times d_1'}, S \subseteq M$ of a matrix $M \in F^{d_2 \times d_1}$ is a matrix whose entries are indexed by a subset of the column and row indices. This means:

$$S[a, b] = M[\mathrm{i}_a, \mathrm{j}_b], \forall a \in [d_1'] \subseteq [d_1], \forall b \in [d_2'] \subseteq [d_2], \exists! \mathrm{i}_a \in [d_1], \exists! \mathrm{i}_b \in [d_2] \tag{2.120}$$

When given many $d$-lengthed columns we can concatenate them to form a matrix. This is defined by the following notation:

$$[v_1 \| \cdots \| v_n], v_i \in F^d \tag{2.121}$$

is the concatenation of $n$ $d$-lengthed columns $v_1, ..., v_n$, forming a $d \times n$ matrix.

**Definition 2.4.53.** *Let the category $F$-**Mat** have:*

1. *Objects consisting of the vector spaces $F^d, d \geq 1$ spanned by column vectors all of the same length $d$ with entries from $F$.*

2. *Morphisms between vector spaces $F^{d_1}$ and $F^{d_2}$ are defined by matrices from $F^{d_2 \times d_1}$. Composition is defined by matrix multiplication.*

A standard theorem [17] of linear algebra states that there exists a full and faithful covariant functor from the category $F$-**FinDimVec** to the category $F$-**Mat**, which depends on a basis for each vector space. This means if a basis is chosen for every finite dimensional vector space. Then all of $F$-**FinDimVec** can be represented by column vectors and matrices, fully preserving algebraic structure.

**Definition 2.4.54.** *Let*

$$B : ob(F\text{-}\boldsymbol{FinDimVec}) \to ob(\boldsymbol{Set}) \tag{2.122}$$

*be the **basis selection map**. This exists since a finite dimensional vector space always has some basis. For each finite dimensional vector space $V \in ob(F$-**FinDimVec**$)$, let*

$$V \mapsto B(V) \subseteq \binom{V}{dim(V)} \text{ have } B(V) \text{ a basis for } V \tag{2.123}$$

We can thus explicitly define the column-matrix functor:

**Definition 2.4.55.** *The column-matrix functor $[\bullet]_{B(\bullet)} : F$-**FinDimVec** $\to F$-**Mat** is defined as follows:*

1. $[V]_{B(\bullet)} \triangleq F^{dim(V)}, \forall V \in ob(F\text{-}\boldsymbol{FinDimVec})$

   *We call* $[V]_{B(\bullet)}$ *the column vector representation of* $F$*-vector space* $V$*.*

2. $[L]_{B(\bullet)} \triangleq [[L(b_1)]_{B(\bullet)} \| \cdots \| [L(b_{dim(V)})]_{B(\bullet)}]$,

   $\forall L \in hom(V,W); hom(V,W) \subseteq mor(F\text{-}\boldsymbol{Mat})$ *where* $B(V) = \{b_1, ..., b_{dim(V)}\}$

   *We call* $[L]_{B(\bullet)}$ *the matrix representation of the linear map* $L : V \to W$*.*

When performing linear algebraic computations on the computer, there must exist some finite representation(s). This is why a choice of basis is essential for well-defined computation.

If the basis of the matrix representation of the linear map $L : V \to W$ is clear, then we omit the subscript for the basis selection map $B(\bullet)$ and just use $[L]$.

**Definition 2.4.56.** *We can define a row-matrix category* $F$-$\boldsymbol{Mat}^T$ *as follows:*

1. *Objects consisting of the covector spaces* $F^{d\times 1}, d \geq 1$ *spanned by row vectors all of the same length* $d$ *with entries from* $F$*.*

2. *Morphisms between vector spaces* $F^{d_1\times 1}$ *and* $F^{d_2\times 1}$ *are defined by matrices from* $F^{d_1\times d_2}$*. These are defined by row-matrix multiplication:*

$$(v \cdot M)[\mathrm{i}] \triangleq \sum_{\mathrm{j}} v[\mathrm{j}] \cdot M[\mathrm{j}, \mathrm{i}] \tag{2.124}$$

We can connect $F$-**Mat** with $F$-$\mathbf{Mat}^T$ through a functor which we define below:

**Definition 2.4.57.** *We can define a dual functor on the* $F$-$\boldsymbol{Mat}$ *category, denoted* $\bullet^T : F\text{-}\boldsymbol{Mat} \to F\text{-}\boldsymbol{Mat}^T$ *for the* $F$-$\boldsymbol{Mat}$ *category. This is defined as:*

1. *On objects:* $[v]^T \in F^{d\times 1}, \forall [v] \in F^d$

2. *On morphisms:* $[L]^T \in F^{d_2\times d_1}, \forall [L] \in F^{d_1\times d_2}$ *where:*

$$[L]^T[\mathrm{j}, \mathrm{i}] = [L][\mathrm{i}, \mathrm{j}], \forall (\mathrm{i}, \mathrm{j}) \in [d_1] \times [d_2] \tag{2.125}$$

   *With composition through matrix multiplication:*

$$([L_2] \cdot [L_1])[\mathrm{i}, \mathrm{j}] = \sum_k [L_2][\mathrm{i}, k] \cdot [L_1][k, \mathrm{j}] \tag{2.126}$$

*This is a contravariant functor.*

We can compose the matrix representation functor and the transpose functor to get the following functor

$$[\bullet]_{B(\bullet)}^{T} : F\text{-}\mathbf{FinDimVec} \to F\text{-}\mathbf{Mat} \tag{2.127a}$$

Similarly, we can compose the dual functor with the matrix representation functor to get the following functor:

$$[\bullet^{\perp}]_{B(\bullet)} : F\text{-}\mathbf{FinDimVec} \to F\text{-}\mathbf{Mat} \tag{2.127b}$$

**Proposition 2.4.58.** *We have the following relationship between the dual functor $\bullet^T$ on $F$-**Mat** to the dual functor on finite dimensional vector spaces $\bullet^T$ :$F$-**FinDimVec** →$F$-**FinDimVec**$^T$ :*

$$[\bullet^{\perp}]_{B(\bullet)} = [\bullet]_{B(\bullet)}^{T} \tag{2.128}$$

*Proof.* This follows since:

1. For the objects: $V^{\perp} = \mathrm{Lin}(V, \mathbb{R})$, which consists of covectors with a matrix representation of a row vector acting as a matrix.

2. For the morphisms: $\hom(W^{\perp}, V^{\perp})$, these must map from row vectors to row vectors. This can be written as the following matrix transpose: $[L]^T : F^{d_2\times 1} \to F^{d_1\times 1}$ that acts on row vectors as follows:

$$[v]^T \cdot [L]^T = ([L]\cdot[v])^T \tag{2.129}$$

where

$$\sum_{\mathrm{i}\in[d_2]} [v]^T[\mathrm{i}]\cdot[L]^T[\mathrm{i},\mathrm{j}] = \sum_{\mathrm{i}\in[d_2]} [L][\mathrm{j},\mathrm{i}]\cdot[v][\mathrm{i}] \tag{2.130}$$

□

Due to Proposition 2.4.58, we can rederive that $(M^T)^T = M, \forall M \in F^{d_1\times d_2}$ since:

$$[\bullet] = [(\bullet^{\perp})^{\perp}] = ([\bullet^{\perp}])^T = ([\bullet]^T)^T \tag{2.131}$$

where the first equality follows from the fact that $(\bullet^{\perp})^{\perp} \cong \bullet$, for $\bullet$ a vector space, since $\bullet^{\perp}$ is an involution functor.

### 2.4.59 Modules

Given a unital ring $R$, we can view it as if it were the scalars of a vector space.

**Definition 2.4.60.** *Given a commutative group $(M, +)$ with the identity element $\overrightarrow{0} \in M$, we can make $M$ a (left)$R$-**module** by introducing an action of $R$ on $M$, $\cdot : R \times M \to M$ as follows:*

- *(Associativty)* $r_1 \cdot (r_2 \cdot m) = (r_1 \cdot r_2) \cdot m \in M, \forall m \in M, \forall r_1, r_2 \in R$
- *(Distributivity)* $r_1 \cdot (m_1 + m_2) = r_1 \cdot m_1 + r_1 \cdot m_2 \in M, \forall m_1, m_2 \in M, \forall r_1 \in R$
- *(Additivity)* $(r_1 + r_2) \cdot m_1 = r_1 \cdot m_1 + r_2 \cdot m_1 \in M, \forall r_1, r_2 \in R, \forall m_1 \in M$
- *(Identity)* $1 \cdot m_1 = m_1, \forall m_1 \in M$

As for linear maps between vector spaces, we can define the maps between $R$-modules as follows:

**Definition 2.4.61.** *For a ring $R$ and two $R$-modules $M, N$, a $R$-**linear map** $f : M \to N$ is defined as:*

1. $f(x + y) = f(x) + f(y), x, y \in M$
2. $r \cdot f(x) = f(r \cdot x), x \in M, r \in R$

Similar to quotient groups, the **quotient** of two $R$-modules can be defined by the $R$-module formed by the quotient group of their respective additive groups.

**Definition 2.4.62.** *The **quotient module** between a $R$-module $M$ and a submodule $N$ is defined as:*

$$\frac{M}{N} \triangleq \{m + N : m \in M\} \tag{2.132}$$

*where the elements $m+N$ for $m \in M$ satisfy:*

$$m + N \triangleq \{m + n : n \in N\} \tag{2.133}$$

*and are called* ***cosets*** *or quotient classes.*

*Picking an element $m+n^* \in m+N$ for each coset $m+N$, determines a* ***representative*** *of that coset.*

### Generators and Relations

For a unital ring $R$, we can take any finite set $S$ and make it into a **free $R$-module**, this is denoted:

$$< S > \triangleq R^{|S|} \tag{2.134}$$

The set $S$ is called a set of **generators** that generate the $R$-module $< S >$.

The elements of $R^{|S|}$ are the formal sums of the elements of $S$. The formal sums can be written as:

$$\sum_{i=1}^{|S|} c_i \cdot s_i \in R^{|S|}, \text{ with } c_i \in R \text{ i} = 1, ..., |S|, S = \{s_i\}_{i=1}^{|S|} \tag{2.135}$$

**Proposition 2.4.63.** *A free $R$-module generated by a finite set $S$ satisfies the definition of an $R$-module as given in Definition 2.4.60.*

*Proof.* The formal sums form a $R$-module as follows:

1. $(R^{|S|}, +)$ forms a group:

   (a) (Zero) $\sum_i 0 \cdot s_i \in R^{|S|}$

   (b) (Inverse) $\sum_i (-c_i) \cdot s_i \in R^{|S|}$ and is the inverse of $\sum_i c_i \cdot s_i \in R^{|S|}$

   (c) (Closure) $\sum_i a_i \cdot s_i + \sum_i b_i \cdot s_i = \sum_i (a_i + b_i) \cdot s_i \in R^{|S|}$ when $\sum_i a_i \cdot s_i, \sum_i b_i \cdot s_i \in R^{|S|}$

   (d) (Associativity)

$$(\sum_i a_i \cdot s_i + \sum_i b_i \cdot s_i) + \sum_i c_i \cdot s_i = \sum_i a_i \cdot s_i + (\sum_i b_i \cdot s_i + \sum_i c_i \cdot s_i) \tag{2.136}$$

   where $\sum_i a_i \cdot s_i, \sum_i b_i \cdot s_i, \sum_i c_i \cdot s_i \in R^{|S|}$

2. With the following action of $R$ on $(R^{|S|}, +)$:

$$r \cdot (\sum_i c_i \cdot s_i) \triangleq \sum_i (r \cdot c_i) \cdot s_i \tag{2.137}$$

It can be checked that the four properties of an action of $R$ on $R^{|S|}$ to form a (left) $R$-module are true.

□

**Set-Theoretic Operations on the elements of a free $R$-module:**

Since the elements of a free $R$-module behave as tuples of $R$-coefficients, we can define set-theoretic operations on the formal sums.

The elements of $< S >$ are tuples and thus we can define their intersection as follows:

**Definition 2.4.64.** *For $< S >$ a **free** $R$-**module** is an $R$-module in the sense of Definition 2.4.60*

$$s \cap t \triangleq \sum_i \begin{cases} c_i \cdot s_i & c_i = c'_i \\ 0 & c_i \neq c'_i \end{cases} \text{ where } s = \sum_i c_i \cdot s_i, t = \sum_i c'_i \cdot s_i; \forall s, t \in < S > \tag{2.138}$$

This allows us to also define the intersection between two free $R$-modules:

**Definition 2.4.65.** *For $< S >, < T >$ two free $R$-modules, the intersection between $< S >, < T >$:*

$$< S > \cap < T > \triangleq \{s \cap t : s \in < S >, t \in < T >\} \tag{2.139}$$

We can also say that one formal sum is contained in another:

**Definition 2.4.66.** *For $< S >$ a **free** $R$-**module**,*

*We say that $s \subseteq t$ where $s = \sum_i c_i s_i, t = \sum_i c'_i s_i; \forall s, t \in < S >$ if:*

$$\forall c_i, \exists c'_j \text{ with } c_i = c'_j \tag{2.140}$$

**Definition 2.4.67.** *We will use the notation* $s \in r$ *for* $r \in < S >$ *if:*

$$r = c_s \cdot s + \sum_{s' \in S \smallsetminus \{s\}} c_{s'} \cdot s', c_s \neq 0 \textit{ and } c_{s'} \in R, \forall s' \in S \tag{2.141}$$

If the ring $R$ is a field, free $R$-modules are vector spaces over the field $R$.

A $R$-module can be finitely generated. This is defined as follows:

**Definition 2.4.68.** *A* $R$*-module* $M$ *is* ***finitely generated*** *iff there exists an* $n \in \mathbb{N}$ *and a surjective* $R$*-linear map* $\phi : R^n \twoheadrightarrow M$*, called the* ***presentation map****, denoted:*

$$R^n \xrightarrow{\phi} \!\!\!\!\twoheadrightarrow M \tag{2.142}$$

Since $M$ is finitely generated, we have that we can form the following sequence of maps:

$$\ker(\phi) \longrightarrow R^n \overset{\phi}{\twoheadrightarrow} M \twoheadrightarrow \mathbf{0} \tag{2.143}$$

where the map $\ker(\phi) \to R^n$ is defined by:

$$\begin{aligned} k_{\mathrm{i}} &\mapsto (g_1, ..., g_n)_{\mathrm{i}} \\ \text{s.t. } \phi(k_{\mathrm{i}}) = \sum_{\mathrm{j}} c_{\mathrm{j}} \cdot \phi(g_{\mathrm{j}}) &= \overrightarrow{0} \text{ for some } c_{\mathrm{j}} \in R \end{aligned} \tag{2.144}$$

Certainly the quotient of a free $R$-module with any of its submodules is a $R$-module. By the first isomorphism theorem for modules [16], we have that:

$$M \cong \frac{R^n}{\ker(\phi)} \tag{2.145}$$

The elements of the submodule $\ker(\phi)$ are called the **relations** of $M$.

We can give an equivalent characterization of a Noetherian ring $R$ using any finitely generated module over $R$:

**Theorem 2.4.69.** *([19])* $R$ *is a Noetherian ring if and only if every submodule of a finitely generated* $R$*-module* $M$ *is also finitely generated.*

### Hilbert's Syzygies

When $R$ is a polynomial ring over a field, we would like to find all of the generators and relations of a $R$-module $M$.

Let $k$ be a field. We can show that for a finitely generated $R$-module $M$ with $R = k[t]$, $M$ can be written as a presentation, or quotient of a free $R$-module $R^n$ and a free submodule of $R^n$, where $n \in \mathbb{N}$.

By the Hilbert Basis theorem, 2.4.17, we know that $R = k[t]$ is Noetherian. By Theorem 2.4.69, $\ker(\phi)$ is a finitely generated submodule of $R^n$. The $R$-module $\ker(\phi)$ is an $R$-module of **relations** and the module $R^n$ forms a free $R$-module of **generators**. We call the decomposition of $M$ into generators and relations as the presentation of $M$.

The $R$-module $\ker(\phi)$ is called the 1st syzygy of $M$ and denoted $\text{Syzygy}_1(R^n, M)$.

If the $R$-module $\text{Syzygy}_1(R^n, M)$ is finitely generated, this "pullback" process can continue with higher order relations formed by finding a free module that can map surjectively onto $\ker(\phi)$ and then finding the kernel of the map from this free module to $\ker(\phi)$. Each of these higher order kernels are denoted by

$$\text{Syzygy}_{\text{i}}(R^n, M) \triangleq \text{Syzygy}_1(\text{Syzygy}_{\text{i}-1}(R^n, M), M) \tag{2.146}$$

where i denotes the number of pull backs off of the free module $R^n$ of $M$.

In fact, for any polynomial ring over $m$ indeterminants, there can exist up to $m$ pull back operations.

**Theorem 2.4.70.** *([15] Hilbert's Syzygy Theorem)*

*Let $R = k[t_1, ..., t_m]$ where $k$ is a field and $M$ a $R$-module.*

*There exists a $n \in \mathbb{N}$ and the following sequence of maps:*

$$Syzygy_m(R^n, M) \longrightarrow \cdots \longrightarrow Syzygy_1(R^n, M) \longrightarrow R^n \twoheadrightarrow M \tag{2.147}$$

*where:*

$$Syzygy_{\text{i}}(R^n, M) = R^{k_{\text{i}}}, \text{i} = 1, ..., m \tag{2.148}$$

*is a finitely generated free R-module with $k_i$ generators.*

There is a natural corollary to Hilbert's Syzygy Theorem for finite dimensional vector spaces over a field.

**Corollary 2.4.71.** *Let $k$ be a field and $M$ a finite dimensional vector space over $k$ of dimension $m$.*

*For any $n > m$, there is the following sequence of maps:*

$$\mathbf{0} \longrightarrow Syzygy_1(R^n, M) \longrightarrow R^n \twoheadrightarrow M \longrightarrow \mathbf{0} \tag{2.149}$$

*where:*

$$Syzygy_1(R^n, M) \cong R^k \tag{2.150}$$

*is a vector space of dimension $k = n - m$.*

*Proof.* Let $\phi : R^n \to R^m$ be a linear map of rank $m$. This means $\mathrm{im}(\phi) \cong R^m$. We then have by the rank-nullity Theorem 2.4.48:

$$\ker(\phi) \oplus \mathrm{im}(\phi) \cong R^k \oplus R^m \cong R^n \tag{2.151}$$

Letting $\mathrm{Syzygy}_1(R^n, M) = \ker(\phi) \cong R^k$, the Corollary follows. □

**Corollary 2.4.72.** *(The $Syzygy_1$ of a spanning set of vectors of a vector space from Corollary 2.4.71 )*

*Let $k$ be a field and $V$ a finite dimensional vector space over $k$ of dimension $m$.*

*Let $S \triangleq \{v_i : v_i \in V, i = 1, ..., n\}$ be a subset of vectors from $V$ with $span(S) = V, n > m$.*

*There is a **unique** presentation map $\phi :< S > \twoheadrightarrow V$ with*

$$\phi(\sum_{i=1}^{n} c_i \cdot v_i) = \sum_{i=1}^{n} c_i \cdot v_i \tag{2.152}$$

*acting as the identity on $S$, where*

$$ker(\phi) = Syzygy_1(< S >, V) \subseteq Syzygy_1(< \boldsymbol{Forget}(V) >, V) \tag{2.153}$$

*Proof.* By Corollary 2.4.71, we know that a map $\phi :< S > \twoheadrightarrow V$ exists.

The map given in Equation 2.152 is linear:

$$\phi(\sum_{i=1}^{n} c_i \cdot v_i) = \sum_{i=1}^{n} c_i \cdot \phi(v_i) = \sum_{i=1}^{n} c_i \cdot v_i \tag{2.154}$$

The map $\phi$ is surjective since $\text{span}(S) = V$ so any $v \in V$ can be expressed as $v = \sum_{i=1}^{n} c_i \cdot v_i$.

It is unique since on the basis $S$ of $< S >$, we have that $\phi(v_i) = v_i, \forall v_i \in S$. This uniquely defines any linear map.

Certainly we have the following subspace relationship:

$$\ker(\phi) = \{\sum_{i=1}^{n} c_i \cdot v_i \in < S >: \sum_{i=1}^{n} c_i \cdot v_i = \vec{0}\} \tag{2.155a}$$

$$\subseteq \{ \sum_{u \in U: U \subseteq \textbf{Forget}(V), |U| < \infty} d_u \cdot u \in < \textbf{Forget}(V) >: \sum_{u \in U: U \subseteq \textbf{Forget}(V), |U| < \infty} d_u \cdot u = \vec{0}\} \tag{2.155b}$$

this is because $S$ satisfies $S \subseteq V$ as a subset of vectors. □

### 2.5 Euclidean Geometry

Euclidean geometry in $d_{amb}$-dimensions is a space that models the physical world. It can be defined as an affine space. As an **affine space** it is defined by the pair $(\mathbb{E}, \vec{\mathbb{E}})$ where $\mathbb{E}$ is a set of points determined by a tuple of $d_{amb}$ real numbers and $\vec{\mathbb{E}}$ is represented as a real vector space $\mathbb{R}^{d_{amb}}$ of $d_{amb}$ dimensions. The space of points $\mathbb{E}$ has an origin point, denoted $\mathbf{0}$, which is the all 0's tuple. The vectors in $\vec{\mathbb{E}}$ are called Euclidean vectors.

The vectors $v \in \vec{\mathbb{E}}$ act on $\mathbb{E}$ via the following map:

$$(v, p) \mapsto p + v \in \mathbb{E} \tag{2.156}$$

where the addition between $p$ and $v$ is an entry-wise addition of the two $d_{amb}$ lengthed tuples.

For any $p \in \mathbb{E}$ this map is a free and transitive group action [16], meaning it satisfies the following properties:

1. (Right identity) $p + \vec{0} = p, \forall p \in \mathbb{E}, \vec{0} \in \vec{\mathbb{E}}$

2. (Associativity) $p + (v + w) = (p + v) + w, \forall v, w \in \overrightarrow{\mathbb{E}}$

3. (Free and transitive action) $v \mapsto p + v, \forall p \in \mathbb{E}$ is a bijection

According to property (3) of the group action, for any two points $p, q \in \mathbb{E}$, there is a unique **displacement vector** $\overrightarrow{qp} \triangleq p - q \in \overrightarrow{\mathbb{E}}$ determined by the points $p$ and $q$.

$\mathbb{E}$ **forms a vector space:** Viewing points as vectors displaced from the origin, $\mathbb{E}$ can form a vector space with the addition between two points $p, q \in \mathbb{E}$ defined as

$$p + q \triangleq (p - \mathbf{0}) + (q - \mathbf{0}) + \mathbf{0} \in \mathbb{E} \tag{2.157}$$

and scalar multiplication by $c \in \mathbb{R}$ on a point $p \in \mathbb{E}$ as:

$$cp \triangleq c(p - \overrightarrow{0}) + \overrightarrow{0} \in \mathbb{E} \tag{2.158}$$

The physical meaning of the Euclidean vector is as a direction. This means for a Euclidean vector $v \in \overrightarrow{\mathbb{E}}$ that for any point $p \in \mathbb{E}$, the points $q(t) = p + t \cdot v, \forall t \in [0, \infty)$ are all displaced away from $p$ in the direction $v$.

**Definition 2.5.1.** *An **affine subspace** of Euclidean space* $(\mathbb{E}, \overrightarrow{\mathbb{E}})$ *is defined by the pair* $(\{a + v : v \in \overrightarrow{\mathbb{E}'}\}, \overrightarrow{\mathbb{E}'})$ *where* $a \in \mathbb{E}$ *is a single point and* $\overrightarrow{\mathbb{E}'} \subseteq \overrightarrow{\mathbb{E}}$ *is a linear subspace of* $\overrightarrow{\mathbb{E}}$*. We say that a point* $q \in \mathbb{E}$ *belongs to a p-dimensional affine subspace if there exists* $a \in \mathbb{E}$ *and* $\overrightarrow{\mathbb{E}'} \subseteq \overrightarrow{\mathbb{E}}$ *so that* $q \in \{a + v : v \in \overrightarrow{\mathbb{E}'}\}$ *and* $dim(\overrightarrow{\mathbb{E}'}) = p$

The space $\overrightarrow{\mathbb{E}}$ is equipped with the Euclidean inner product, called the dot product, between Euclidean vectors. The dot product between two Euclidean vectors is defined by:

$$u \cdot v = \sum_{\mathrm{i}=1}^{d_{amb}} u_{\mathrm{i}} \cdot v_{\mathrm{i}}, \forall u, v \in \overrightarrow{\mathbb{E}} \tag{2.159}$$

The norm of a vector in Euclidean space $\overrightarrow{\mathbb{E}}$ is defined through the dot product:

$$\|v\|_2 \triangleq \sqrt{v \cdot v} \tag{2.160}$$

Any vector $v \in \overrightarrow{\mathbb{E}}$ with norm 1 is called a unit vector.

We can also define a Euclidean distance between points $p, q \in \mathbb{E}$ by the norm as follows:

$$d(p,q) \triangleq \|p - q\|_2 \tag{2.161}$$

**Definition 2.5.2.** *Let a **curve** $C : [0,1] \to \mathbb{E}$ be a continuously differentiable function of the unit interval to $\mathbb{E}$.*

The length of a curve $C$ is defined by:

$$L(C) \triangleq \int_0^1 \|\frac{d}{dt}C(t)\|_2 dt \tag{2.162}$$

Curves between two points $p, q \in \mathbb{E}$ acting as linear equations can be defined as follows:

$$SL_{p,q} : t \mapsto tp + (1-t)q \tag{2.163}$$

A **straight line segment** between $p, q$ is the image of $SL_{p,q}$ on $[0,1]$, defined as:

$$\overline{pq} = \{tp + (1-t)q : t \in [0,1]\} \tag{2.164}$$

**Proposition 2.5.3.** *For $p, q \in \mathbb{E}$,*

*The length of the curve $SL_{p,q}$ traversing $\overline{pq}$ is exactly $d(p,q)$.*

*The curve $SL_{p,q}$ is the shortest path between $p, q \in \mathbb{E}$*

*Proof.* **For the first part:**

$$L(SL_{p,q}) = \int_0^1 \|\frac{d}{dt}SL_{p,q}(t)\|_2 dt = \int_0^1 \|p-q\|_2 dt = \|p-q\|_2 \tag{2.165}$$

**For the second part:**

Letting $N : p \mapsto \|p\|_2$ be denoted by $N : \mathbb{E} \to \mathbb{R}$.

We can thus say that $N$ is a $d_{amb}$-variable function on the $d_{amb}$ variables of $\dot{C} \in \mathrm{im}(\frac{d}{dt}C)$ where $\mathbb{E}$ has dimension $d_{amb}$ and that

$$L(C) = \int_0^1 N(\frac{d}{dt}(C(t)))dt \tag{2.166}$$

The operator $L : \{C : [0,1] \to \mathbb{E} \text{ a curve}\} \to \mathbb{R}$ is called an action functional and $N$ is called a Lagrangian since it is defined on the time $t$, position $C(t)$ and velocity $\frac{d}{dt}C(t)$ of the curve $C$ and has units of "energy", namely length.

The shortest path is the curve that minimizes the following optimization equation:

$$\min_{C:[0,1]\to\mathbb{E}:C(0)=p,C(1)=q} L(C) \tag{2.167}$$

The minimizing $C^*$ curve between $p, q \in \mathbb{E}$ satisfies the Euler-Lagrange Equations for the action functional [20]:

$$\begin{aligned} &\frac{\partial N}{\partial C_{\mathrm{i}}}(t, C_1(t), ..., C_{d_{amb}}(t), \dot{C}_1(t), ..., \dot{C}_{d_{amb}}(t)) \\ -&\frac{d}{dt}(\frac{\partial N}{\partial \dot{C}_{\mathrm{i}}}(t, C_1(t), ..., C_{d_{amb}}(t), \dot{C}_1(t), ..., \dot{C}_{d_{amb}}(t))) = 0, \forall \mathrm{i} = 1, ..., d_{amb} \end{aligned} \tag{2.168}$$

We have that:

$$\frac{\partial N}{\partial \dot{C}_{\mathrm{i}}}(\dot{C}(t)) = \frac{\dot{C}_{\mathrm{i}}(t)}{N(\dot{C}(t))} \tag{2.169a}$$

$$\frac{\partial N}{\partial C_{\mathrm{i}}} = 0 \tag{2.169b}$$

Solving for $C$ from the Euler Lagrange equation:

$$0 - \frac{d}{dt}\frac{\dot{C}_{\mathrm{i}}(t)}{N(\dot{C}(t))} = 0, \forall \mathrm{i} = 1, ..., d_{amb} \tag{2.170}$$

We thus have,

$$\frac{\dot{C}_{\mathrm{i}}(t)}{N(\dot{C}(t))} = K_{\mathrm{i}}, K_{\mathrm{i}} \in \mathbb{R}, \forall \mathrm{i} = 1, ..., d_{amb} \tag{2.171}$$

Solving:

$$(1 - K_{\mathrm{i}}^2)(\dot{C}_{\mathrm{i}}^2(t)) - \sum_{\mathrm{j}\neq\mathrm{i}} K_{\mathrm{i}}^2(\dot{C}_{\mathrm{j}}^2(t)) = 0 \tag{2.172}$$

Setting $K_i = 0$, we must have that $\dot{C}_i^2 = 0$, so that $C_i(t)$ is linear of the form $C_i(t) = at + b$.

By the boundary conditions, we must have: $C^*(t) = tp + (1-t)q$.

Thus, the straight line $SL_{p,q} = C^*(t)$ is the shortest path between $p, q \in \mathbb{E}$ □

Besides the straight lines between two points in $\mathbb{R}^{d_{amb}}$, we can also minimize a set of line segments over many points. If we let these line segment endpoints require a connectivity condition on a set of points $P \subseteq \mathbb{R}^{d_{amb}}$, we can define the following graph:

**Definition 2.5.4.** *A **Euclidean Minimum Spanning Tree** (EMST) is the minimizer of the following optimization equation:*

$$\min_{(P,E) \text{ is a spanning tree on } P} \sum_{(u,v)\in E} \|u - v\|_2 \tag{2.173}$$

This connectivity structure is crucial to viewing the zero dimensional homology vector space on Vietoris-Rips complexes on $\mathbb{R}^{d_{amb}}$.

From here on out, we denote Euclidean space by $\mathbb{R}^{d_{amb}}$. This overloads the notation of $\mathbb{E}$ and $\overrightarrow{\mathbb{E}}$ so that $\mathbb{R}^{d_{amb}}$ denotes $\mathbb{E}$ viewed as a vector space. We go over some standard definitions for sets of points in Euclidean space.

### 2.5.5 Isometric Symmetries of Euclidean space

We can define a special subset of the set of endomorphisms on $\mathbb{R}^{d_{amb}}$, the maps from $\mathbb{R}^{d_{amb}}$ back to itself that preserve the Euclidean metric. We call this set of endomorphisms as the **Euclidean** group, denoted $E(d_{amb})$. This can be formalized as follows:

**Definition 2.5.6.**

$$E(d_{amb}) \triangleq \{f : \mathbb{R}^{d_{amb}} \to \mathbb{R}^{d_{amb}} : \|f(x) - f(y)\|_2 = \|x - y\|_2, \forall x, y \in \mathbb{R}^{d_{amb}}\} \tag{2.174}$$

These maps are known as isometries. The isometries can be considered as a form of symmetry. Isometries in Euclidean space form a group by the Myers-Steenrod Theorem [21].

The Translation group $\mathbb{T}(d_{amb})$ is a subgroup of the isometry group $E(d_{amb})$ that involves only translations:

**Definition 2.5.7.**

$$\mathbb{T}(d_{amb}) \triangleq \{T : \mathbb{R}^{d_{amb}} \to \mathbb{R}^{d_{amb}} : T(x_1, x_2, ..., x_{d_{amb}}) = (x_1 + \tau_1, x_2 + \tau_2, ..., x_{d_{amb}} + \tau_{d_{amb}})\} \quad (2.175)$$

The Rotational group $SO(d_{amb})$ is a subgroup of the isometry group $E(d_{amb})$ that involves only rotations:

**Definition 2.5.8.**

$$SO(d_{amb}) \triangleq \{r : \mathbb{R}^{d_{amb}} \to \mathbb{R}^{d_{amb}} : r(x) = Q \cdot x, Q \cdot Q^T = Q^T \cdot Q = I \ \ and\ det(Q) = 1\} \quad (2.176)$$

The Special Euclidean group is the subgroup of the group of isometries on $\mathbb{R}^{d_{amb}}$ that involve the motion of rigid bodies. This means that they cannot change orientation of a rigid body. We can thus define this as follows:

**Definition 2.5.9.**

$$SE(d_{amb}) \triangleq \{f : \mathbb{R}^{d_{amb}} \to \mathbb{R}^{d_{amb}} : \exists k, f = f_k \circ ... \circ f_1, \ \ where\ f_i \in \mathbb{T}(d_{amb}) \cup SO(d_{amb})\} \quad (2.177)$$

### 2.5.10 Convex Shapes in Euclidean Space

We would like to define subsets of Euclidean space which are closed under straight-line segments. Straight-line segments in Euclidean space are shortest paths between the two endpoints. The length of a line segment is the Euclidean distance between its two end points. This means that the subset preserves the distance between its points.

**Definition 2.5.11.** *A* ***convex set*** $S \subseteq \mathbb{R}^{d_{amb}}$ *has the property that the straight line segment between any two points stays in* $S$*. More formally:*

$$\forall x, y \in S, \forall t \in [0, 1], tx + (1 - t)y \in S \quad (2.178)$$

When $|S| = 1$, then the $S$ itself is convex.

A simple $d_{amb} - 1$-dimensional convex set can be defined as follows:

**Definition 2.5.12.** *A* ***half-plane*** *in* $\mathbb{R}^{d_{amb}}$ *is defined as the set:*

$$\{x \in \mathbb{R}^{d_{amb}} : v \cdot x + b = 0\} \tag{2.179}$$

*where* $v, b \in \mathbb{R}^{d_{amb}}$.

The half-planes define another convex set called half-spaces:

**Definition 2.5.13.** *A* ***half-space*** *is defined as the set:*

$$\{x : v \cdot x + b \leq 0\} \tag{2.180}$$

We can also define a convex set determined by a point and a distance.

**Definition 2.5.14.** *For* $x \in \mathbb{R}^{d_{amb}}$ *and* $r \in \mathbb{R}^{+}$*, the* $d_{amb}$***-dimensional ball*** *of radius* $r$ *centered at* $x$*, denoted* $B(x,r)$*, is the following set:*

$$B(x,r) \triangleq \{z \in \mathbb{R}^{d_{amb}} : \|x - z\|_2 \leq r\} \tag{2.181}$$

Certainly it can be checked that $d_{amb}$-dimensional balls in Euclidean space are convex.

**Definition 2.5.15.** *The boundary of a* $d_{amb}$***-dimensional ball*** *of radius* $r$ *is defined by all the points exactly* $r$ *distance away from* $x$*.*

$$\partial(B(x,r)) \triangleq \{z \in \mathbb{R}^{d_{amb}} : \|x - z\|_2 = r\} \tag{2.182}$$

*This is also called the* $d_{amb} - 1$*-dimensional sphere,* $S(x,r)$*.*

For an arbitrary number of points, we can define a centroid as:

**Definition 2.5.16.** *For* $p + 1$ *points* $S \subseteq \mathbb{R}^{d_{amb}}$*, the* ***centroid*** *of* $S$ *is defined as:*

$$c_S \triangleq \frac{1}{p+1} \sum_{s \in S} s \tag{2.183}$$

We can also define a convex shape on a set of points:

**Definition 2.5.17.** *A convex hull of a set of points $S \subseteq \mathbb{R}^{d_{amb}}$ is defined as the intersection of all convex subsets of $\mathbb{R}^{d_{amb}}$ that contain $S$. This is denoted as follows:*

$$conv(S) \triangleq \bigcap_{S' \supset S : S' \text{ convex}} S' \tag{2.184}$$

*The set of points $S$ is said to* ***span*** *conv($S$).*

**Proposition 2.5.18.** *A convex hull of $S \subseteq \mathbb{R}^{d_{amb}}$ is the intersection of the half-spaces that contain $S$.*

*Proof.* As a convex hull:

$$\text{conv}(S) = \bigcap_{C \supseteq S:\text{ C convex}} C \tag{2.185}$$

By the hyperplane separation theorem [22], for any point $x$ outside conv($S$), we can find a hyperplane $H$ that separate $x$ from conv($S$). Let the unique halfspace $H_{x,\text{conv}(S)}$ be the halfspace determined by $H$ that covers conv($S$).

Certainly halfspaces are convex so:

$$\text{conv}(S) = \bigcap_{C \supseteq S:\text{ C convex}} C \subseteq \bigcap_{H_{x,\text{conv}(S)}: x \notin \text{conv}(S)} H_{x,\text{conv}(S)} \tag{2.186}$$

Since $x$ was arbitrary from the complement, we must have that:

$$\bigcup_{H_{x,\text{conv}(S)}: x \notin \text{conv}(S)} (\mathbb{R}^{d_{amb}} \setminus H_{x,\text{conv}(S)}) \subseteq \mathbb{R}^{d_{amb}} \setminus \text{conv}(S) \tag{2.187a}$$

$$\Rightarrow \text{conv}(S) \supseteq \bigcap_{H_{x,\text{conv}(S)}: x \notin \text{conv}(S)} H_{x,\text{conv}(S)} \tag{2.187b}$$

and thus:

$$\text{conv}(S) = \bigcap_{H_{x,\text{conv}(S)}: x \notin \text{conv}(S)} H_{x,\text{conv}(S)} \tag{2.188}$$

□

**Definition 2.5.19.** *If $p \leq d_{amb}$ with $S \subseteq \mathbb{R}^{d_{amb}}$ being $p+1$ points. If for any $q+2$ points $S' \subseteq S$ does not belong to a $q$-dimensional affine subspace with $0 \leq q < p$, then conv($S$) is called a $p$-dimensional* ***simplex*** *on a set of $p+1$ points $S \subseteq \mathbb{R}^{d_{amb}}$.*

*Any subset $S'$ of $q+1, q \leq p$ points from simplex conv$(S)$ spans conv$(S') \subseteq$ conv$(S)$. This convex hull spanned by $S'$ is called a q-dimensional face of simplex conv$(S)$.*

*These faces are q-dimensional simplices.*

In Definition 2.5.19, since the points $S$ uniquely define the simplex spanned by $S \subseteq \mathbb{R}^{d_{amb}}$, we call $S$ as a $p$-dimensional simplex as well.

We can piece together simplices to form a polytope. In the literature, a polytope is usually defined as an intersection of half spaces or as a convex hull [23]. We give a more relaxed definition of a polytope in terms of simplices in the following:

**Definition 2.5.20.** *For any $p : p \leq d_{amb}, d_{amb} \in \mathbb{N}$, a **polytope** on a finite set of points $S \subseteq \mathbb{R}^{d_{amb}}$ is defined as a union of p-dimensional simplices where*

- *Any intersection can only be at a $p-1$ dimensional face.*
- *Every simplex intersects atleast one other simplex.*

If we impose that the polytope on $S \subseteq \mathbb{R}^{d_{amb}}$ in Definition 2.5.20 is convex, it satisfies the usual definition of a convex hull of its point set $S$. Such a polytope is called a convex polytope.

**Proposition 2.5.21.** *A convex polytope on $S \subseteq \mathbb{R}^{d_{amb}}$ is a convex hull on $S$.*

*Proof.* Since the convex polytope of $S$ is a union of $p$-dimensional simplices, it would suffice to show that it is the intersection of half-spaces according to Proposition 2.5.18.

Let the polytope on $S$ be the following union of $p$-dimensional simplices:

$$P(S) = \bigcup_{\sigma:\sigma=\text{conv}(S'),|S|=p+1,\bigsqcup S'=S} \sigma \tag{2.189}$$

Consider the union of all $p-1$-dimensional faces that do not belong to more than one $p$-dimensional simplex:

$$B = \bigcup_{\sigma'\subseteq\sigma:\nexists\tau,\tau\neq\sigma \text{ s.t. } \tau\cap\sigma=\sigma'} \text{conv}(\sigma') \tag{2.190}$$

For each of the $p-1$-dimensional simplices $\sigma'$, there exists at least one half-space $H_{\sigma'}$ that passes through it.

Any half-space $H$ that contains $P(S)$ contains an intersection of some of these half-spaces $H_{\sigma'}$. This is because any $H$ has a nearest point $s^* \in S$ with $s^* \in H$. If we consider the intersection of all $H_{\sigma'}$ where $s^* \in \sigma'$:

$$H_{s^*} = \bigcap_{\sigma':s^*\in\sigma'} H_{\sigma'} \tag{2.191}$$

then $H_{s^*} \subseteq H$.

Consider any $x \in \mathbb{R}^{d_{amb}} \setminus H$. There exists a nearest half-plane coming from the $H_{\sigma'}$ of $H_{s^*}$ that separates $x$ from $P(S)$. Since $x$ was arbitrary, taking a union:

$$\mathbb{R}^{d_{amb}} \setminus H \subseteq \bigcup_{\sigma':s^*\in\sigma'} H_{\sigma'} \tag{2.192}$$

Taking complements on both sides gives the desired inclusion.

According to Proposition 2.5.18 we know that the convex hull of $P(S)$ satisfies:

$$\text{conv}(P(S)) = \bigcap_{H:P(S)\subseteq H} H \tag{2.193}$$

Taking the intersection over all of these half-spaces recovers $P(S)$:

$$P(S) \subseteq \bigcap_{s\in S} H_s \subseteq \bigcap_{H:P(S)\subseteq H} H = \text{conv}(P(S)) \subseteq P(S) \tag{2.194}$$

Thus $P(S) = \text{conv}(P(S))$, meaning it is a convex hull of $S$. □

**Definition 2.5.22.** *A* ***normal*** *to a p-dimensional simplex $\sigma = conv(\{x_1, ..., x_{p+1}\})$ is defined as a unit vector $\vec{n}_\sigma$ having the property that for any $q \in \sigma$:*

$$\vec{n}_\sigma \cdot (x_\text{i} - q) = 0, \forall \text{i} = 1, ..., p+1 \tag{2.195}$$

*The set of all normals to $\sigma$ is denoted by $N_\sigma$.*

*The normal $n_\sigma$ defines an orthogonal axis to $\sigma$, namely the set of all points $\{c_\sigma + t\vec{n}_\sigma : t \in \mathbb{R}\}$*

We would like to parameterize a set of points by a single number through the Euclidean distance.

**Definition 2.5.23.** *For $t > 0$, we say a set of points $P \subseteq \mathbb{R}^{d_{amb}}$ are $t$-**equidistant** iff*

$$\|x_i - x_j\|_2 = t, \forall x_i, x_j \in P, x_i \neq x_j \tag{2.196}$$

Certainly taking the convex hull of a set of $p+1$ $t$-equidistant points forms a $p$-dimensional simplex. We call such a set as a **$t$-equilateral $p$-dimensional simplex**.

We can connect $p$-dimensional simplices with their next higher $p+1$-dimension via a single point, which we call the **apex**.

**Definition 2.5.24.** *Let $p \leq d_{amb}$.*

*For a $t$-equilateral $p$-dimensional simplex $\sigma \subseteq \mathbb{R}^{d_{amb}}$, we call any point $a \in \mathbb{R}^{d_{amb}}$ that makes $conv(\sigma \cup \{a\})$ a $t$-equilateral $p+1$-dimensional simplex as a $t$-equilateral-**apex** of $\sigma$.*

*Let **t-equilateral-apexes**$(\sigma)$ denote the set of all $t$-equilateral-apex points for $\sigma$.*

**Proposition 2.5.25.** *Amongst all apexes for a $t$-equilateral $p$-dimensional simplex $\sigma$, there are exactly two apexes that pass through the line passing through the centroid of a $p$-dimensional simplex $\sigma$ in the direction of one of its normals $\vec{n}_\sigma \in N_\sigma$.*

*Proof.* The line passing through the centroid of a $p$-dimensional simplex $\sigma$ in the direction of its normal comprises the set of points: $\{c_\sigma + t\vec{n}_\sigma : t \in \mathbb{R}\}$.

Since the line only has two directions, it can have atmost two intersection points with **t-equilateral-apexes**$(\sigma)$. Both intersection points are apexes since the line is centered at $c_\sigma$ as can be checked. □

**Proposition 2.5.26.** *Let $t > 0$ and $p, d_{amb} \in \mathbb{N}, 0 \leq p \leq d_{amb}$.*

*For $p+1$ $t$-equidistant points $\sigma_p = \{x_1, ..., x_{p+1}\}$, there exists a polytope $P$ with $P \subseteq \bigcap_{i=1}^{p+1} B(x_i, t) \subseteq \mathbb{R}^{d_{amb}}, P \neq \varnothing$ spanning $2^{d_{amb}-p}(p+1)$ many points.*

*Proof.* Let

$$S_{p+1} = \bigcap_{i=1}^{p+1} \partial(B(x_i, t)) \tag{2.197}$$

for $p \leq d_{amb}$.

Certainly by definition of the boundary of a ball, all $s \in S_{p+1}$ have the property that $\|s - x_{\mathrm{i}}\|_2 = t, \forall \mathrm{i} = 1, ..., p+1$.

We construct $P$ inductively.

Let $P_q, p \leq q \leq d_{amb}$ be a collection of point sets we construct from $\sigma_p$. This can be defined inductively:

1. $P_p \leftarrow \{\sigma_p\}$.

2. $P_{q+1} \leftarrow P_q \cup \{\text{conv}(\{u\} \cup \tau) : \forall u \in (S_{p+1} \cap \text{apexes}(\tau) \cap \{c_\tau + t\vec{n}_\tau : t \in \mathbb{R},\text{ for a single } \vec{n}_\tau \in N_\tau\}), \forall \tau \in P_q\}$

From $q$ to $q$+1, we add $t$-equilateral apex points for each simplex $\tau$ in $P_q$ on the orthogonal axis passing through the centroid $c_\tau$ of $\tau$. By Proposition 2.5.25, we are adding two diametrically opposing apex points.

Since taking the convex hull of a $t$-equidistant simplex with its apex maintains $t$-equidistance, all simplices in $P_q$ are $t$-equidistant simplices. Furthermore, since $\tau \subseteq \text{conv}(\{u\} \cup \tau), \forall \tau \in P_q, p \leq q \leq d_{amb}$ and $\sigma_p \in P_p$, we must have that $\sigma_p \subseteq \tau, \forall \tau \in P_q, p \leq q \leq d_{amb}$. Thus, each $\tau$ is the convex hull of $\sigma_p$ with points $t$-equidistant from all of the points $x_1, ...., x_{p+1}$.

Since $\bigcap_{\mathrm{i}=1}^{p+1} B(x_{\mathrm{i}}, t)$ is convex, each $\tau \in P_q$ is the smallest convex set containing $x_1, ..., x_{p+1}$ and the apex points used to form $\tau$, and all apex points belong to $S_{p+1}$. We must have that $\tau \subseteq \bigcap_{\mathrm{i}=1}^{p+1} B(x_{\mathrm{i}}, t)$.

Since constructing $P_{q+1}$ from $P_q$ introduces $2|P_q|$ many new apex points. From $P_p$ to $P_{d_{amb}}$ we must double the size of the starting set of $p+1$ equidistant points $d_{amb} - p$ times.

Let $P \leftarrow \bigcup_{\tau \in P_{d_{amb}}} \tau$.

There are in total $2^{d_{amb}-p}(p+1)$ many points used to span $P$.

It is a union of $d_{amb}$-dimensional simplices where by construction of $P_{d_{amb}}$, each simplex intersects another at a $d_{amb} - 1$ dimensional face and all simplices were constructed from $d_{amb} - 1$-dimensional faces. Thus $P$ is a polytope. □

We define a collection of simplices that allow it to be embedded in some ambient space $\mathbb{R}^{d_{amb}}$. Such a collection is called a simplicial complex, which is defined as follows:

**Definition 2.5.27.** *A simplicial complex $K$ is a collection of simplices embedded in $\mathbb{R}^{d_{amb}}$ such that*

1. *If $K$ contains a $p$-dimensional simplex $\sigma$, then $K$ also contains every $p-1$ dimensional face of $\sigma$.*
2. *If two simplices in $K$ intersect, then their intersection is a face of each of them*

The **top dimension** of a simplicial complex is the maximum dimension of any simplex in it. A simplicial complex is $d$-dimensional if its top dimension is $d$.

A simplicial complex is a union of polytopes whose pairwise intersections are amongst lower dimensional faces of the simplices in each polytope or is empty.

Conversely, a $p$-dimensional polytope of Definition 2.5.19 is a simplicial complex consisting of $p$-dimensional simplices and all their faces where each $p$-dimensional simplex is connected to atleast one other through a $p-1$ dimensional face.

Let the category $\mathbf{Simp} \triangleq (\text{ob}(\mathbf{Simp}), \text{mor}(\mathbf{Simp}))$ have objects as simplicial complexes and arrows determined by simplicial maps. A simplicial map between two simplicial complexes $K, L$ is defined as follows:

**Definition 2.5.28.** *A **simplicial map** $m : K \to L$ is a function $f : V(K) \to V(L)$ such that whenever $\sigma \subseteq V(K)$ belongs to $K$, the image $f(\sigma)$ belongs to $L$.*

Besides simplices, which tightly cover a set of points in $\mathbb{R}^{d_{amb}}$. We can define a simpler cover of a set of points, which is axis-align with the standard basis. This standard shape from computational geometry that is used to span a set of points $S \subseteq \mathbb{R}^{d_{amb}}$ is called a bounding box. It is often used as a hypothesis shape that tightly covers a more complicated shape inside the convex hull of $S$. It is useful for deriving closed forms and computation since it respects the orthogonality of the standard basis of Euclidean space.

**Definition 2.5.29.** *Let $S \subseteq \mathbb{R}^{d_{amb}}$ be a set of points in $d_{amb}$-dimensional Euclidean space.*

*A $d_{amb}$-dimensional bounding box is defined by the $d_{amb}$-dimensional hypercube spanned by the two corner points $(\max_{s\in S}(s_k))_{k=1}^{d_{amb}}$ and $(\min_{s\in S}(s_k))_{k=1}^{d_{amb}}$. This can be expressed as:*

$$BB(S) \triangleq [(\min_{s\in S}(s_k))_{k=1}^{d_{amb}}, ((\mathbf{1}_{k=\mathrm{j}}[\max_{s\in S}(s_k))_{k=1}^{d_{amb}}] + (\mathbf{1}_{k\neq \mathrm{j}}[\min_{s\in S}(s_k)])_{k=1}^{d_{amb}})]^{d_{amb}} \tag{2.198}$$

**Definition 2.5.30.** *Denote* $vol\colon 2^{(\mathbb{R}^{d_{amb}})} \to \mathbb{R}$ *as the map that returns the volume of a subset of* $d_{amb}$*-dimensional space using the Euclidean volume form*

$$\omega_{\text{euc}} \triangleq \Pi_{\text{i}=1}^{d_{amb}} dx^{\text{i}} \tag{2.199}$$

*This can be expressed as follows:*

$$vol(S) \triangleq \int_S \omega_{\text{euc}} \tag{2.200}$$

A $p$-dimensional simplex can be embedded in $\mathbb{R}^{d_{amb}}$ as a convex hull of $p$ standard basis vectors. We call such a simplex a $p$-dimensional standard simplex $\Delta_p$. This has a nice closed form for its volume:

$$\text{vol}(\Delta_p) \triangleq \frac{1}{p!} \tag{2.201}$$

If we scale each standard basis vector in $\Delta_p$ by $r \in \mathbb{R}$, then we obtain a $p$-dimensional standard $r$-lengthed simplex $\Delta_{p,r}$, which has volume $\frac{1}{p!}r^p$

## 2.6 Topology

The category of topological spaces is denoted **Top**. The objects are topological spaces and the morphisms are continuous maps between topological spaces.

### 2.6.1 Point Set Topology

Central to topology is the concept of connectedness, which can be defined through open sets. An open set generalizes the concept of a neighborhood ball about a point from geometry.

**Definition 2.6.2.** *A topological space* $X$ *is defined as a set of points with a topology, denoted* $\tau(X) \subseteq 2^X$*. The elements of* $\tau(X)$ *are called open sets, which are subsets of* $X$*.*

*We say that a subset* $U \subseteq X$ *is open iff* $U \in \tau(X)$*.*

*A topology satisfies the following axioms:*

- $\varnothing \in \tau(X)$
- *Any union of open sets is an open set*

- *Any finite intersection of open sets is an open set*

A continuous map between two topological spaces is defined as follows:

**Definition 2.6.3.** *Let $X, Y$ be two topological spaces.*

*A map $f : X \to Y$ is continuous iff*

*for any $U \subseteq Y$, if $U$ is open, then $f^{-1}(U) \subseteq X$ is open.*

Continuous maps are generalizations of the common notion of a real valued function without discontinuities from calculus.

The isomorphisms in **Top** are called homeomorphisms. These are defined as follows:

**Definition 2.6.4.** *Let $X, Y$ be two topological spaces.*

*A map $f : X \to Y$ is a homeomorphism if:*

- *$f$ is a bijection*
- *$f$ is continuous*
- *$f^{-1}$ is continuous*

Homeomorphisms fall into the general class of "continuous deformations." These are often colloquially thought of as transformations that do not "tear" a space. A common theme in topology is to characterize spaces up to homeomorphisms.

### 2.6.5 The Category Simp and the Boundary Map:

For a $d$-dimensional simplicial complex, as defined in Definition 2.5.27, for each dimension $p, 0 \leq p \leq d$ the set of all $p$-simplices freely generate a $p$-chain vector space over $\mathbb{R}$.

**Definition 2.6.6.** *Given a simplicial complex $K$, a $p$-chain $\tau = \sum_{\sigma \in K : |\sigma| = p+1} m_\sigma \cdot \sigma$ with coefficients $m_\sigma \in \mathbb{R}$ is a formal sum of $p$-simplices from $K$.*

*A $p$-chain vector space $(C_p(K), +)$ consists of the set of $p$-chains of $K$ with the formal sum operation $+$ and a coefficient field of $\mathbb{R}$.*

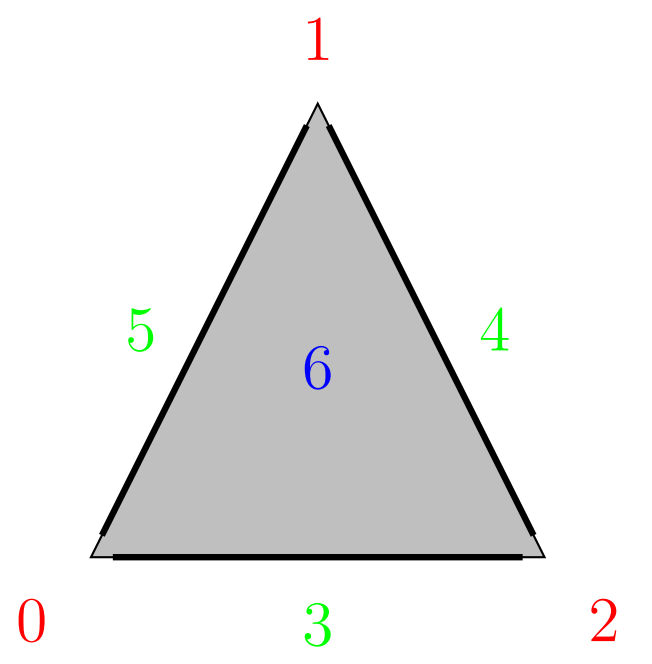

| − | 0 | 1 | 2 | 3 | 4 | 5 | 6 |
|---|---|---|---|---|---|---|---|
| 0 | 0 | 0 | 0 | -1 | 0 | -1 | 0 |
| 1 | 0 | 0 | 0 | 0 | -1 | 1 | 0 |
| 2 | 0 | 0 | 0 | 1 | 1 | 0 | 0 |
| 3 | 0 | 0 | 0 | 0 | 0 | 0 | -1 |
| 4 | 0 | 0 | 0 | 0 | 0 | 0 | 1 |
| 5 | 0 | 0 | 0 | 0 | 0 | 0 | -1 |
| 6 | 0 | 0 | 0 | 0 | 0 | 0 | 0 |

**Figure 2.2.** (a) A simplicial 2-dimension complex composed of 6 simplices. The points 0, 1, 2 are 0-simplices, the line-segments 3, 4, 5 are 1-simplices, and the triangle 6 is a 2-simplex. (b) The corresponding boundary matrix. In the matrix, a column representing a simplex is encoded by the simplices in its boundary, e.g., the triangle 6 has the boundary composed of line-segments 3, 4, and 5.

The formal sum between two $p$-chains geometrically represents a union of two $p$-chains where at their intersection the coefficients add in the reals.

If there are $n$ simplices, then a simplex $\sigma_i$ can be represented as a standard basis vector $e_i \in \mathbb{R}^n$ and a $p$-chain $\sigma = \sum_{i \in [n]} m_i \cdot \sigma_i$ can be represented by $\sum_{i \in [n]} m_i \cdot e_i \in \mathbb{R}^n$. This is since the $p$-chains form a vector space. Thus any $p$-chain can be viewed as a column vector. We denote this column vector by $[\sigma] \in \mathbb{R}^n$. When a simplex $\sigma_i$ is a summand of $p$-chain $\sigma$ then we denote this by $\sigma_i \in \sigma$.

A simplicial map $f : K_1 \to K_2$ can be extended to a chain vector space linear map as follows:

$$\begin{aligned} &\tilde{f}_p : C_p(K_1) \to C_p(K_2) \\ \text{s.t. } &\tilde{f}_p(\sum_i m_i \cdot \sigma_i) = \sum_i m_i \cdot f(\sigma_i) \end{aligned} \tag{2.202}$$

Connectivity amongst these simplices can be defined algebraically through a boundary map. We define this formally here:

**Definition 2.6.7.** *A **boundary map** $\partial^K_{p+1} : C_{p+1}(K) \to C_p(K)$ is defined by*

$$\partial^K_{p+1}(\tau) \triangleq \sum_{\mathrm{i}=1}^{p+1}(-1)^{\mathrm{i}} \cdot [\tau_{-\mathrm{i}}], \forall \tau = \sum_{\sigma \in K : \sigma = \{v_0, \ldots, v_{p+1}\}} m_\sigma \cdot \sigma \in C_{p+1}(K) \tag{2.203}$$

*where* $\tau_{-\mathrm{i}} = \sum_{\sigma \in K : \sigma = \{v_0, \ldots, v_{p+1}\}} m_\sigma \cdot (\sigma \setminus \{v_{\mathrm{i}}\})$

The formal sum from Equation 2.6.7 is called the $p$-dimensional **boundary** of the $p+1$-dimensional boundary.

When it is clear what the simplicial complex $K$ is for a boundary map $\partial^K_{p+1}$, we often omit the $K$ when denoting the boundary map: $\partial_{p+1}$. We can also denote the $p$-dimensional boundaries with the simplicial complex $K$ as follows:

$$\partial_{p+1}(\sigma), \forall \sigma \in K, |\sigma| = p+2 \tag{2.204}$$

This boundary map has a matrix representation, which is the concatenation of all the column representations of $\partial_{p+1}(\sigma), \sigma \in K, |\sigma| = p+2$. These column representations are denoted $[\partial_{p+1}(\sigma)]$, and their column-wise concatenation is called the $p+1$-boundary matrix. Their concatenation into a matrix is denoted $[\partial_{p+1}]$. If we concatenate all the $p$-boundary matrices $[\partial_p]$ for $0 \le p \le \dim(K)$, we form the boundary matrix

$$[\partial] \triangleq [\partial_0 \| \cdots \| \partial_{\dim(K)}] \tag{2.205}$$

To explain the boundary map geometrically, a surface in $\mathbb{R}^3$ has a boundary map which maps continuous surface patches to their geometric boundaries, which are loops.

In Figure 2.2, we give an example of a simplicial complex $K$ and its boundary matrix $[\partial]$.

### 2.6.8 The $H_\bullet$ Functor:

Consider the following differential equation whose solutions $z_p$ are a subset of the $p$-chain vector space subject to an equivalence relation:

$$\begin{aligned} z_p \in C_p(K) : \partial_p(z_p) = \overrightarrow{0}, p \in \mathbb{N} \\ \text{s.t. } z_p \equiv z'_p \text{ if } \exists b_{p+1} \in C_{p+1}(K) : z_p = z'_p + \partial_{p+1}(b_{p+1}) \end{aligned} \tag{2.206}$$

To explain geometrically, for $K$ a surface with boundary embedded in $\mathbb{R}^3$, the differential equation expresses the set of holes on the surface with each hole having attached to it any amount of connected 2-dimensional subset of the surface. In particular, the holes of a surface are determined by the boundary of the surface.

The first equation expresses that for a solution $z_1$, all $\sigma \in z_1$ must encompass a hole. The constraint for the first equation states that two 1-chains belong to the same solution to the differential equation if they are connected through the formal sum, which means they are connected through a 2-dimensional sheet.

The homology functor is describing "partial disconnectivity" since the surface is path connected but lacks the 2-dimensional connectivity **it could have**. Furthermore, homology is an invariant to continuous deformations since the holes cannot change without tearing the surface.

We can thus give the following physical interpretation of $p$-dimensional homology:

> *$p$-dimensional homology measures $p+1$-dimensional disconnectivity when there is still $p$-dimensional connectivity*

The functor $H_\bullet : \textbf{Simp} \to \textbf{Vec}$ for $\bullet \in \mathbb{N}$ maps on objects as follows:

$$H_\bullet(K) \triangleq \frac{\ker(\partial_\bullet^K)}{\operatorname{im}(\partial_{\bullet+1}^K)} \tag{2.207}$$

**Observation 2.6.9.** *$H_\bullet(K)$ forms a vector space consisting of all the solutions to the differential equation.*

*Proof.* The operator $\partial_\bullet, \forall \bullet \in \mathbb{N}$ is linear so its kernel and image are vector spaces.

The double composition satisfies $\partial_\bullet \circ \partial_{\bullet+1}(\sigma) = \overrightarrow{0}, \forall \sigma \in C_{\bullet+1}(K)$. Thus $\text{im}(\partial_{\bullet+1}) \subseteq \ker(\partial_\bullet)$. In particular, the kernel of $\partial_\bullet$ summed with any vectors from $\text{im}(\partial_{\bullet+1})$ are all solutions. In other words, all vectors of the form $z + b, z \in \ker(\partial_\bullet), b \in \text{im}(\partial_{\bullet+1})$ satisfy $\partial_p(z + b) = \overrightarrow{0}$.

Since this is exactly the solution space of the differential equation of Equation 2.206 we must have that $H_\bullet(K)$ is the solution space. □

On morphisms, the functor acts on simplicial maps naturally as follows:

$$H_\bullet(f : K_1 \to K_2) \triangleq \frac{\pi_{\ker(\partial_\bullet^{K_2})}(\tilde{f}_\bullet |_{H_\bullet(K_1)})}{\text{im}(\partial_{\bullet+1}^{K_2})} \tag{2.208}$$

where:

$$\tilde{f}_\bullet |_{H_\bullet(K_1)} (z_\bullet + \text{im}(\partial_{\bullet+1}^{K_1})) = \tilde{f}_\bullet(z_\bullet) + \tilde{f}_\bullet(\text{im}(\partial_{\bullet+1}^{K_1})), \forall z_\bullet + \text{im}(\partial_{\bullet+1}^{K_1}) \in H_\bullet(K_1) \tag{2.209}$$

The map $H_\bullet(f : K_1 \to K_2)$ is called an induced map because it uses $f$ on homology classes, or coset of $\bullet$-dimensional chains, to map generators across homology groups. The relations come from the functor alone.

For simplicial complexes of surfaces in $\mathbb{R}^3$ with $p = 1$ such solutions $\sigma$ enclose a disk.

For $\bullet \geq 0$, the $H_\bullet$ functor is not a faithful functor 2.3.26. This is easily demonstrated by the following example:

(2.210)

$$K \triangleq \{\{1\}, \{2\}, \{3\}\}, L \triangleq \{\{1\}, \{3\}, \{1, 3\}\} \tag{2.211}$$

We can define the following functor diagram:

$$\begin{array}{ccc} K & \longrightarrow & L \\ \downarrow & & \downarrow \\ H_0(K) & \longrightarrow & H_0(L) \end{array} \tag{2.212}$$

We see that the downstairs induced map maps from a rank 3 homology group to a rank 1 homology group. However, the upstairs map can take on many forms, depending on how to map the element 2: e.g. either $2 \mapsto 1$ or $2 \mapsto 3$.

### 2.7 Statistics

We go over the statistical background needed to understand data.

For any problem, we are usually given data as input in the form of $n$ independent and identical (i.i.d) observations $S_{\text{i}}$ sampled from a probability distribution $P$, denoted $S_{\text{i}} \sim P$. The set of observations $S \triangleq \{S_{\text{i}}\}_{\text{i}=1}^{n}$ is called a **data set**. We go over what a probability distribution is in the following:

#### 2.7.1 Probability Spaces

A probability distribution is defined by a **probability space**, which is the tuple $(\Omega, \mathcal{F}, P)$.

A **sample space** $\Omega$ represents the space of all possible outcomes. For example, a coin flip has a sample space $\{H, T\}$, meaning a coin can either be heads or tails. From the sample space, we can then define a **$\sigma$-Algebra** $\mathcal{F} \subseteq 2^{\Omega}$.

**Definition 2.7.2.** *The **$\sigma$-Algebra** $\mathcal{F}$ is a collection of subsets of $\Omega$, called **measurable sets** which are observable which we define as follows:*

- *(The sample space is in $\mathcal{F}$):* $\Omega \in \mathcal{F}$
- *(Closed under Complement):* $A \in \mathcal{F} \Rightarrow \Omega \setminus A \in \mathcal{F}$
- *(Closed under Countable Unions):* $A_{\text{i}} \in \mathcal{F}$ *for* $\text{i} \in \mathbb{N} \Rightarrow \bigcup A_{\text{i}} \in \mathcal{F}$

Intuitively a $\sigma$-**Algebra** $\mathcal{F}$ can be viewed as a space of logical statements of the form $Q : \Omega \to \{T, F\}$ where these logical statements can take on either Truth or Falsity. Also, any countable disjunction of these logical statements is also a valid set of outcomes.

We formalize this idea in category theoretic terms:

Let **$\sigma$-Algebra** be the category with objects consisting of all $\sigma$-algebras $\mathcal{F}$ of a sample space $\Omega$ where $\mathcal{F} \subseteq 2^{\Omega}$ and morphisms consisting of **$\sigma$-homomorphisms**.

**Definition 2.7.3.** *Let $\Omega_1, \Omega_2$ be two sample spaces. A $\sigma$-**homomorphism** between two $\sigma$-algebras $\mathcal{F}_1 \subseteq 2^{\Omega_1}$, $\mathcal{F}_2 \subseteq 2^{\Omega_2}$ is a map $f : \mathcal{F}_1 \to \mathcal{F}_2$ that preserves the properties of a $\sigma$-algebra via homomorphism. This can be formalized as follows:*

- $f(\Omega_1) = \Omega_2$
- $f(\Omega_1 \setminus A) = \Omega_2 \setminus f(A), \forall A \in \mathcal{F}_1$
- $f(\bigcup_i A_i) = \bigcup_i f(A_i), \forall A_i \in \mathcal{F}_1, i \in \mathbb{N}$

**A Measurable Space can be viewed as a Space of Logical Statements:** According to FOL with domain being a set, a logical statement assigns only one of Truth or False to samples from a sample space $\Omega$. This can be expressed as a boolean function $Q : \Omega \to \{T, F\}$. We show that measurable spaces of a sample space $\Omega$ can be viewed as a FOL with domain $\Omega$.

Let **FOL** be the category with:

1. Objects consisting of the pair $(\Omega, L)$ where $\Omega$ is a set and $L$ is a set of logical statements with domain $\Omega$. A logical statement is represented by a boolean functions $Q : \Omega \to \{T, F\}$.
2. The morphisms in **FOL** are the **logical morphisms**.

**Definition 2.7.4.** *For two predicate spaces: $(\Omega_1, L_1), (\Omega_2, L_2)$.*

*A **logical morphism** is just a map $f : L_1 \to L_2$ mapping logical statements to logical statements.*

We define a map $F : \sigma$-**Algebra** $\to$ **FOL** as follows:

$$F(\mathcal{F}) \triangleq \{Q_A : A \in \mathcal{F}\}, \forall \mathcal{F} \in \text{ob}(\sigma\textbf{-Algebra}) \text{ s.t.}$$
$$Q_A(x) = \begin{cases} T, \forall x \in A \\ F, \forall x \in \Omega \setminus A \end{cases} \tag{2.213}$$

and where $F(f) : \{Q_A : A \in \mathcal{F}\} \to \{Q_B : B \in \mathcal{G}\}$ is the predicate morphism defined as:

$$F(f)(Q_A) \triangleq Q_{f(A)} \tag{2.214}$$

for any $f : \mathcal{F} \to \mathcal{G}$ being a $\sigma$-homomorphism.

**Proposition 2.7.5.** *The map $F : \sigma$-**Algebra** $\to$ **FOL** of Equation 2.213 is a faithful functor.*

*Proof.* **1. $F$ is a functor:**

We check the axioms for $F : \sigma$-**Algebra** $\to$ **FOL** to be a functor:

- $F(A) = \{Q_A : A \in \mathcal{F}\} \in \text{ob}(\textbf{FOL})$
- $F(\text{id}_{\mathcal{F}}) = \text{id}_{(\{Q_A : A \in \mathcal{F}\})} = \text{id}_{F(\mathcal{F})}$
- $F(f \circ g)(Q_A) = Q_{f \circ g(A)} = F(f)(Q_{g(A)}) = F(f) \circ F(g)(Q_A)$

**2. $F$ is faithful:**

Checking injectivity:

Let $f, g \in \text{mor}(\sigma\textbf{-Algebra})$

$$F(f) = F(g) \Rightarrow F(f)(Q_A) = F(g)(Q_A), \forall A \in \mathcal{F}, B \in \mathcal{G} \tag{2.215a}$$

$$\Rightarrow Q_{f(A)} = Q_{g(A)}, \forall A \in \mathcal{F}, B \in \mathcal{G} \tag{2.215b}$$

$$\Rightarrow f(A) = \{x : Q_{f(A)}(x) = T\} = \{x : Q_{f(B)}(x) = T\} = f(B), \forall A \in \mathcal{F}, B \in \mathcal{G} \tag{2.215c}$$

$$\Rightarrow f = g \tag{2.215d}$$

□

We give the pair $(\Omega, \mathcal{F})$ of sample space and corresponding $\sigma$-algebra the name of **measurable space**. For short, we also say that a set $\Omega$ that has a $\sigma$-algebra $\mathcal{F} \subseteq 2^{\Omega}$ is called a measurable space.

**Probability Measures:** On a $\sigma$-algebra we can define a notion of the chance for which a set $A$ can occur during a sampling process. We can define chance by a probability measure $P : \mathcal{F} \to [0, 1]$ which must satisfy the following two criterion:

- ($\sigma$-additive): $A_{\text{i}} \in \mathcal{F}$, $\text{i} \in \mathbb{N}$ which are pairwise disjoint, meaning $A_{\text{i}} \cap A_{\text{j}} = \varnothing$, has:

$$P(\bigsqcup_{\text{i}} A_{\text{i}}) = \sum_{\text{i}} P(A_{\text{i}}) \tag{2.216}$$

- (The sample space gives probability 1): $P(\Omega) = 1$

For a probability space $(\Omega, \mathcal{F}, P)$, we often only use the probability measure $P$. When this is the case, we often call $P$ a **probability distribution**. The measurable space $(\Omega, \mathcal{F})$ is implicit when using the probability distribution $P$.

With a probability measure defined on a $\sigma$-algebra $\mathcal{F}$, every set in $\mathcal{F}$ is called an **event** and often viewed as a logical statement through the functor $F$.

For probability distribution $P$, we use the notation $P(A)$ to denote the probability that an event $A \subseteq \Omega$ occurs.

We define a notion of dependency through the probability measure:

**Definition 2.7.6.** *Two events $A$ and $B$ are **independent**, denoted $A \perp B$ if:*

$$P(A \cap B) = P(A)P(B) \tag{2.217}$$

The concept of independence allows us to define a probability measure that measures the independence between two events. This is called the conditional probability:

**Definition 2.7.7.** *For events $A$ and $B$ from probability space $(\Omega, \mathcal{F}, P)$, we can define the **conditional probability** of $A$ conditioned on $B$ by:*

$$P(A \mid B) \triangleq \frac{P(A \cap B)}{P(B)} \tag{2.218}$$

**Proposition 2.7.8.** *For a probability space $(\Omega, \mathcal{F}, P)$ and $B \in \mathcal{F}$ a measurable set.*

*$P(\bullet \mid B)$ is a probability measure on the measurable space $(\Omega, \mathcal{F})$.*

*Proof.* Checking the definition:

**1. $P(\bullet \mid B)$ has a unit interval codomain:**

We know that $0 \leq P(A \cap B) \leq P(B)$ since by $\sigma$-additivity,

$$P(B) = P(A \cap B) + P(B \setminus (A \cap B)) \geq P(A \cap B) \tag{2.219}$$

Thus, upon dividing $P(A \cap B)$ by $P(B)$:

$0 \leq P(\bullet \mid B) \leq 1$

**2.** $\sigma$**-additivity:**

$P(\sqcup A_i \mid B) = \frac{P((\sqcup A_i) \cap B)}{P(B)} = \frac{P(\sqcup (A_i \cap B))}{P(B)} = \frac{\sum_i P(A_i \cap B)}{P(B)} = \sum_i \frac{P(A_i \cap B)}{P(B)} = \sum_i P(A_i \mid B)$

**3.** $P(\Omega \mid B) = 1$**:**

$P(\Omega \mid B) = \frac{P(\Omega \cap B)}{P(B)} = \frac{P(B)}{P(B)} = 1$

□

With this definition, we see that we can redefine independence between events $A$ and $B$ as follows:

$$P(A \mid B) = P(A) \tag{2.220}$$

It is also possible to define independence amongst multiple events. This is defined as follows:

**Definition 2.7.9.**

$$A_1 \perp \ldots \perp A_n \textit{ iff } \Pi_i P(A_i) = P(A_1 \textit{ and } A_2 \textit{ and } \cdots \textit{ and } A_n) \tag{2.221}$$

Defining probability through logical statements is very convienent when the sample space and $\sigma$-algebra are not really necessary or known exactly in closed form. For example:

(2.222) Let $P(\text{"It rains tomorrow"}) = 0.5$.

This probability definition avoids having to define anything at the set-theoretic level and replaces it with the logical statement of: "It rains tomorrow," which must be either True or False but not both.

We can also define independence through logical statements this way. For example,

(2.223) The statement "It rains tomorrow" and the statement "The presidential election is tomorrow" are independent events.

Using logical statements, we can thus understand independence without actually going to the set-theoretical level.

**Random Variables:** We often require expressing probability in terms of a space of observable data which can follow a chosen probability distribution. We can achieve this by mapping a probability space to the measurable space of observable data. In order to preserve measurable sets of observable data, we require that this map be **measurable**. We define this formally:

**Definition 2.7.10.** *For two measurable spaces* $(\Omega_1, \mathcal{F}_1), (\Omega_2, \mathcal{F}_2)$

*A map* $m : \Omega_1 \to \Omega_2$ *is* ***measurable*** *if:*

$$m^{-1}(A) \in \mathcal{F}_1, \forall A \in \mathcal{F}_2 \tag{2.224}$$

**Definition 2.7.11.** *A* ***random variable*** $X : \Omega \to \mathcal{O}$ *is a measurable map from a sample space* $\Omega$ *of some probability space* $(\Omega, \mathcal{F}, P)$ *to a measurable entity space* $(\mathcal{O}, \mathcal{G})$*.*

*The probability space* $(\Omega, \mathcal{F}, P)$ *is tied to random variable* $X : \Omega \to \mathcal{O}$*. We use the notation* $X \sim P$ *to denote this. The distribution* $P$ *is called the* ***law*** *of the random variable* $X$*.*

Two random variables $X : \Omega \to \mathcal{O}_1, Y : \Omega \to \mathcal{O}_2$ are **independent**, denoted $X \perp Y$, if:

$$P(X \in A)P(Y \in B) = P(X \in A \text{ and } Y \in B), \forall A, B \in \mathcal{O}_1, \mathcal{O}_2 \tag{2.225}$$

The notation $P(X = x)$ where $X : \Omega \to \mathcal{O}$ is a random variable and $x \in \mathcal{O}$. Similarly, $P(X \in A)$ denotes $P(X^{-1}(A))$, where $A \in \mathcal{G}$ is a measurable set in the measurable space $(\mathcal{O}, \mathcal{G})$.

**Remark 2.7.12.** *In the statistics literature, random variables are usually denoted by real valued functions. This is not necessary. We use the more general definition where the entity space* $\mathcal{O}$ *can belong to any (small) category.*

For two random variables, we can denote the conditional distribution $P(A \mid B)$ by:

$$P(X \in A \mid Y \in B) = \frac{P(X \in A, Y \in B)}{P(Y \in B)} \tag{2.226}$$

### 2.7.13 Causality

We can model the concept of cause and effect, or causality, from the physical world using the language of probability. In order to do this, we define a discrete directed structure called a directed graph. This can define a distribution in terms of random variables and the conditionals determined by the directions:

Let $V$ be a set. A **directed** graph is a pair $G = (V, E), E \subseteq V \times V$.

A directed graph $G = (V, E)$ is **acyclic**, called a DAG if any path $v_1, ..., v_n$ with $(v_i, v_{i+1}) \in E$ does not have $v_n = v_1$

A structural causal model is a way to define causality through a directed graph of causation amongst random variables.

Let **RV** be the category of random variables as objects with morphisms being conditional distributions so that $X_j = f(X_i, U_i)$ where $U_i$ is a uniform distribution on $[0, 1]$ and $f$ is any function on $X_i$ and $U_i$

**Definition 2.7.14.** *A **structural causal model** (SCM) is defined by a directed acyclic graph (DAG) $G = (V, E)$ where each node $v \in V$ indexes a random variable $X_v \in ob(\boldsymbol{RV})$ called an endogenous variable and each directed edge $\mathrm{e} \in E, \mathrm{e} = (u, v)$ indexes the conditional $X_v = f(X_u, U_v)$ where $U_v$ is called an exogenous variable.*

A SCM must have the structure of a DAG so that it can have a well defined joint.

In particular, if the exogenous variables $U_1 \perp U_2 \perp ... \perp U_n$ (joint independence), then the SCM is called **Markovian**.

**Theorem 2.7.15.** *([24]) For a **Markovian** SCM the joint distribution can be factored into conditional distributions for each endogenous variable:*

$$P(V_1, ..., V_n) = \Pi_{i=1}^{n} P(V_i \mid V_{pa_i}), \tag{2.227}$$

**Definition 2.7.16.** *For a SCM on $G = (V, E)$.*

*Let the **parents** of a random variable $X_v$ denoted by $pa(X_v)$, denote the tuple of all random variables $X_u$ (of some chosen order) where there are directed edges $(u, v) \in E$.*

According to Pearl [25], causality between a cause and effect can be measured through interventional distributions. An **intervention** in probability theory is determined by eliminating the parents of a random variable. The purpose of an intervention is to isolate a cause for an effect. Interventions are often used in experiments through controlled experimental settings.

(2.228) For example, to determine whether *any country that has a large GDP for chocolate causes it to win more nobel prizes*, we should intervene on every country's chocolate eating habits.

This means every country would be forced to have a prespecified GDP for chocolate. Given this intervention, we then measure the country's probability of winning nobel prizes.

The interventional relationship of the probability of nobel prizes for any country depending on the intervened chocolate GDP determines the intended causal relationship.

**Definition 2.7.17.** *For two random variables $E, C$, let the interventional distribution $P(E \mid do(C = c))$ denote the distribution of $E$ conditioned on random variable $C$ taking on value $c$ independent of all the parents of random variable $C$.*

$$P(E \mid do(C = c)) \triangleq \int_{d=(d_1,\dots,d_k)\in supp(pa(C))} P(E \mid C = c, pa(C) = d) P(pa(C) = d) \tag{2.229}$$

*The event $pa(C) = (d_1, \dots, d_k)$ denotes setting every random variable $D_\mathrm{i}$ in the tuple $pa(C) \triangleq (D_1, \dots, D_k)$ to the value $d_\mathrm{i}$ from $pa(C)$.*

**Definition 2.7.18.** *A* ***confounder*** *$C$ in causality denotes a random variable that acts as the common cause between atleast two other random variables $A_\mathrm{i}$, $\mathrm{i} \in I$ with $I$ some index set of size atleast $2$. This means:*

$$P(A_\mathrm{i} \mid C) = P(A_\mathrm{i} \mid do(C)), \forall \mathrm{i} \in I \tag{2.230}$$

When measuring causality between random variables $E, C$, we see that the interventional distribution $P(E \mid \text{do}(C))$ eliminates any common cause, or confounder, $C'$ between $E, C$. In particular, $\text{do}(C)$ disconnects its dependency with $C'$. We call this eliminating confounders through intervention.

A **counterfactual** represents asking "what-if?" questions. These can be defined through interventions on existing observations. Say we have observed that it has rained today and that the ground is wet. We could ask the counterfactual question of: *what is the chance of the ground being wet if it hadn't rained, knowing that it has already rained today.*

In medicine a counterfactual patient often denotes a patient that is most similar to some other patient. This counterfactual patient, however, is chosen to have a different feature. For example, one patient could undergo treatment, and another would not. This fits into the concept of a "what-if?" since the counterfactual patient represents an alternative to the patient in question, conditioned on the patient in question.

We can formally define counterfactuals through interventions as follows:

**Definition 2.7.19.** *A counterfactual distribution for a sequence of random variable $X_1, ..., X_n$, with observables $x_1, ..., x_n$ with intervention on $X_1$ to be $x_1'$ is defined as $P(do(X_1 = x_1'), ..., X_n = x_n)$.*

*This can also be defined for conditional distributions: $P(E \mid do(X_1 = x_1'), ..., X_n = x_n)$ is the counterfactual distribution with response variable $E$.*

Counterfactuals are important in science since they represent a "neighborhood" of alternative realities for observable data. Counterfactuals are a way to express the connection between observed data and "similar" data.

Counterfactuals can be used to make better decisions beyond the observable data as well as create meaningful alternative data for learning.

**Remark 2.7.20.** *Causality does not necessarily have to involve uncertainty. Although probability provides a convenient language for causality, an event $C$ causing the effect $E$ does not have to require any randomness.*

## 2.8 Combinatorial Data

Data often comes with connections. This connectivity can be represented in various ways. We formally define various ways that data can be structured through connections. We then introduce various objects that can be computed on such data.

At the most general level a set can be connected through its subsets. This is defined by an undirected hypergraph:

**Definition 2.8.1.** *An undirected hypergraph* $\mathcal{H} \triangleq (\mathcal{V}, \mathcal{E})$ *consists of a set of nodes* $\mathcal{V}$ *along with the collection of hyperedges* $\mathcal{E} \subseteq 2^{\mathcal{V}}$*, which is a collection of subsets of* $\mathcal{V}$*.*

(2.231) An example of hypergraphs include:

- a $\sigma$-algebra
- a space of binary classifiers
- a collection of social groups
- knowledge graphs
- entities and their relations

A hypergraph can also be directed, this is defined by directed hyperedges. A hyperedge is directed if it is represented by two (ordered) disjoint subsets of nodes. We will consider all hypergraphs as undirected in the following.

Let **HyperGraph** denote the category of hypergraphs. The objects are hypergraphs and the morphisms between two hypergraphs are called hypergraph morphisms which are maps defined on the vertices that preserve connectivity. We define this below:

**Definition 2.8.2.** *For two undirected hypergraphs* $\mathcal{H}_1 = (\mathcal{V}_1, \mathcal{E}_1), \mathcal{H}_2 = (\mathcal{V}_2, \mathcal{E}_2)$*, a hypergraph morphism* $f : \mathcal{H}_1 \to \mathcal{H}_2$ *is defined by the following two criterion:*

- *(Bijection)* $f(v) \in \mathcal{V}_2, \forall v \in \mathcal{V}_1$
- *(Edge-preserving)* $\{f(v)\}_{v \in \mathrm{e}} \in \mathcal{E}_2, \forall \mathrm{e} \in \mathcal{E}_1$

In this category **HyperGraph**, a hypergraph morphism is called an isomorphism when the morphism introduces a bijection on the vertices. This is defined below:

**Definition 2.8.3.** *A hypergraph isomorphism between two hypergraphs* $\mathcal{H}_1 = (\mathcal{V}_1, \mathcal{E}_1), \mathcal{H}_2 = (\mathcal{V}_2, \mathcal{E}_2)$ *is a hypergraph morphism* $\rho : \mathcal{H}_1 \to \mathcal{H}_2$ *in the category* ***HyperGraph*** *where* $\rho\,|_{\mathcal{V}_1}: \mathcal{V}_1 \to \mathcal{V}_2$ *is a bijection.*

As an alternative view of connectivity, given a point set $X$, successively higher-order connections can be defined inductively through closure. For example, pairs of points can form edges. When a triple of edges form a complete graph, a triple of points can form a triangle etc. This is a combinatorial generalization of simplicial complexes from the category **Simp** which are intended to be embeddable in $\mathbb{R}^{d_{amb}}$ for some ambient dimension $d_{amb} \in \mathbb{N}$. Simplicial complexes are discrete analogs to $d$-dimensional manifolds over a point set, where a point set topology has every point locally Euclidean. We can define a $d$-dimensional abstract simplicial complex on a finite point set.

**Definition 2.8.4.** *Given a set* $X$*, let a* $p$*-dimensional abstract simplex be a subset of* $X$ *of cardinality* $p+1$*. A* $p$*-dimensional abstract simplex* $\tau$ *is called a face of a* $p+1$*-dimensional abstract simplex* $\sigma$ *if* $\tau \subseteq \sigma$ *as sets.*

*A* $d$*-dimensional* ***abstract simplicial complex*** $\mathcal{S}$ *is a collection of* $p$*-dimensional abstract simplices* $0 \le p \le d$ *which satisfies the* ***downward closure property****:*

$$\forall \sigma \in \mathcal{S}, \tau \subseteq \sigma, \forall \tau \in \mathcal{S} \tag{2.232}$$

The downward closure relationship forms a partial ordering amongst simplices. A $d$-dimensional simplex cannot exist until all its $d-1$-dimensional faces exist. This partial order is often represented by a Hasse diagram [26].

The $p$-chain group can be defined as in simplicial complexes. A collection of linearly independent $p$-simplices span a vector space over $\mathbb{R}$ freely, meaning there are no relations. These are called the $p$**-chain** vector spaces $C_p(\mathcal{S})$.

Abstract simplicial maps can also be defined:

**Definition 2.8.5.** *Between two $d, d'$-dimensional abstract simplicial complexes $\mathcal{S}, \mathcal{S}'$.*

*An* ***abstract simplicial map*** *$f : \mathcal{S} \to \mathcal{S}'$ is defined by:*

$$f(\{v_1, ..., v_{p+1}\}) \triangleq \{f(v_{\mathrm{i}})\}_{\mathrm{i}=1}^{p+1} \tag{2.233}$$

.

The category of abstract simplicial complexes **AbstractSimp** with objects of abstract simplicial complexes and abstract simplicial maps between abstract simplicial complexes. Isomorphisms are defined by bijective simplicial maps.

(2.234) Examples of abstract simplicial complexes include:

- Set systems on a set $P$ of the form $(P, \leq)$ which forms a downward closed partial order.
- Triangular meshes, which consist of triangles, edges and points which connected together at the edges.

Of course the simplest kind of connectivity is defined by pairwise relationships. These are called graphs, which can also be defined as 2-uniform hypergraphs.

**Definition 2.8.6.** *A graph is a pair $G \triangleq (V, E)$ where $G$ is a hypergraph with all $\mathrm{e} \in E$ having $|\mathrm{e}| = 2$*

To make it clear that between any two nodes there can only be one edge, we often call graphs **simple** graphs.

Let **Graph** denote the category of graphs with objects of graphs and hypergraph morphisms between graphs forms a subcategory of the category **HyperGraph**. Isomorphisms, of course, are also defined through hypergraph isomorphisms.

When there is direction on the edges, we can define a directed graph:

**Definition 2.8.7.** *A (directed) graph $G \triangleq (V, E)$ where $V$ is a set and $E$ is a subset of pairs from $V \times V$*

(2.235) Examples of graphs include:

- Molecules
- Graph Genomes
- Intermediate code representations
- Decision trees
- Knowledge graphs
- Data structures

It is common to define sequences of nodes connected by edges in a graph $G = (V, E)$.

**Definition 2.8.8.** *A **walk** is a (finite) sequence of $n$ nodes, $n \in \mathbb{N}$ with $(v_i)_{v_i \in V, i \in [n+1]}$ connected by edges, meaning:*

- $\{v_i, v_{i+1}\} \in E, \forall i \in [n-1]$ *for the case of $G$ undirected*
- *and* $(v_i, v_{i+1}) \in E, \forall i \in [n-1]$ *for the case of $G$ directed.*

The length of a walk is the number of edges on the walk.

**Definition 2.8.9.** *A **trail** is a walk where no edge repeats*

A special kind of trail is one that starts and ends the same.

**Definition 2.8.10.** *A **cycle** in a graph $G = (V, E)$ is a trail that starts and ends on the same vertex.*

*A **directed cycle** is a cycle on a directed graph.*

**Definition 2.8.11.** *A **path** is a walk where no node repeats.*

When a graph has no cycles, we can give it a name:

**Definition 2.8.12.** *A directed acyclic graph (DAG) is defined as a directed graph that has no directed cycles.*

We can use these sequences on the graph to give a continuity property to the graph:

**Definition 2.8.13.** *A graph $G = (V, E)$ is **path connected**, or just **connected**, if any two nodes $u, v \in V$ have a path on $G$ between them.*

**Definition 2.8.14.** *The **shortest path** between two nodes $u, v \in V$ is defined as follows:*

$$SP_G(u,v) \triangleq \arg \min_{(p_1,\ldots,p_k) \text{ is a path in } G, p_1=u, p_k=v} k \tag{2.236}$$

We can also define a maximal shortest path on a graph, called its diameter:

**Definition 2.8.15.** *The **diameter** of a graph $G = (V, E)$ is defined as:*

$$diam(G) \triangleq \max_{u,v \in V} |SP_G(u,v)| \tag{2.237}$$

**Definition 2.8.16.** *A **tree** is a graph that is:*

1. *Connected*

2. *Has no cycles*

*The nodes in a tree that have degree 1 are called **leaves***

For a tree, its **diameter** must come from a shortest path between two leaves.

Let us define the following graph $T = (P, E)$ on a set $P$.

**Definition 2.8.17.** *A **spanning tree** on a set $P$ is defined by a set of unordered pairs from $P$: $E \subseteq P \times P$ which satisfy:*

1. *Every $p \in P$ has $\exists \{u, v\} \in E : p \in \{u, v\}$*

2. *There does not exist a trail (see Definition 2.8.9) of the form: $p_1, \ldots, p_k \in P$ with $\{p_i, p_{i+1}\} \in E$ and $p_1 = p_k$.*

*Any edge $\{u, v\} \in (P \times P) \setminus E$ is called a **complementary edge**.*

We can also define a special kind of graph that has the properties of an injective function between two sets:

**Definition 2.8.18.** *Let $V_1, V_2$ be two sets.*

*A **matching** $M \subseteq V_1 \times V_2$ is defined so that:*

$$\text{(Vertex Disjointness): } u_1 \neq v_1 \text{ and } u_2 \neq v_2, \forall (u_1, u_2), (v_1, v_2) \in M \tag{2.238}$$

If we expand a graph to have a multiset of (directed) edges. We call this a multigraph.

**Definition 2.8.19.** *A (directed) **multigraph** $G \triangleq (V, E)$ where $V$ is a set and $E$ is a multiset of pairs from $V \times V$.*

A directed multigraph is undirected if the multiset is made up of sets of pairs instead. Multigraphs are meaningful for representing multiple connections between nodes. We will use multigraphs to describe locally small categories where the directed edges are morphisms.

When the number of edges between two nodes in a (directed) multigraph is atmost one we recover simple (directed) graphs.

We can define an algebra on the space of paths on a directed multigraph. This is a convenient algebraic way of representing the space of paths.

**Definition 2.8.20.** *([27]) A **path algebra** $\mathbb{R}\Gamma$ over the reals on a directed multigraph $G = (V, E)$ is defined as a vector space generated by the basis of paths $(v_{\mathrm{i}}, v_{\mathrm{i}+1})_{\mathrm{i}=1}^{n-1}$, s.t. $(v_{\mathrm{i}}, v_{\mathrm{i}+1}) \in E$ for $n \geq 1$ from $G$ with a multiplication operation determined by path composition.*

*The elements of $\mathbb{R}\Gamma$ are of the form:*

$$\sum_{\mathrm{j}<\infty} c_{\mathrm{j}} \cdot p_{\mathrm{j}}, c_{\mathrm{j}} \in \mathbb{R}, p_{\mathrm{j}} = ((v_{\mathrm{i}}, v_{\mathrm{i}+1}))_{\mathrm{i}=1}^{n_{\mathrm{j}}-1}, (v_{\mathrm{i}}, v_{\mathrm{i}+1}) \in E, n_{\mathrm{j}} \geq 1 \tag{2.239}$$

*with multiplication defined on basis vectors as follows:*

$$((v_{\mathrm{i}}, v_{\mathrm{i}+1}))_{\mathrm{i}=1}^{n_{\mathrm{j}}-1} \circ ((w_{\mathrm{i}}, w_{\mathrm{i}+1}))_{\mathrm{i}=1}^{n_k-1} = \begin{cases} (w_{\mathrm{i}}, w_{\mathrm{i}+1}) & 1 \leq \mathrm{i} \leq n_k - 1 \\ (v_{\mathrm{i}}, v_{\mathrm{i}+1})_{\mathrm{i}} & n_k \leq \mathrm{i} \leq n_k + n_{\mathrm{j}} - 2 \\ \vec{0} & \textit{otherwise} \end{cases}, \textit{ when } w_{n_k} = v_1 \tag{2.240}$$

*and with multiplication between elements on* $\mathbb{R}\Gamma$ *defined by:*

$$\sum_{\mathrm{j}<\infty} c_{\mathrm{j}} \cdot ((v_{\mathrm{i}}, v_{\mathrm{i}+1}))_{\mathrm{i}=1}^{n_{\mathrm{j}}-1} \circ \sum_{k<\infty} d_k \cdot ((w_{\mathrm{i}}, w_{\mathrm{i}+1}))_{\mathrm{i}=1}^{n_k-1} \triangleq \sum_{\mathrm{j},k} c_{\mathrm{j}} \cdot d_k \cdot ((v_{\mathrm{i}}, v_{\mathrm{i}+1}))_{\mathrm{i}=1}^{n_{\mathrm{j}}-1} \circ ((w_{\mathrm{i}}, w_{\mathrm{i}+1}))_{\mathrm{i}=1}^{n_k-1} \quad (2.241)$$

*where the following relations between nodes and paths exist on the algebra:*

- $p_v^0 \circ p_v^0 = p_v^0, \forall 0$*-lengthed paths* $p_v^0 \in \mathbb{R}\Gamma$
- $p_u^0 \circ p_v^0 = \overrightarrow{0}, \forall 0$*-lengthed paths* $p_v^0 \in \mathbb{R}\Gamma$
- $p_v^0 \circ p = p \circ p_u^0, \forall 0$*-lengthed paths* $p_u^0, p_v^0 \in \mathbb{R}\Gamma$ *and for any* $p \in \mathbb{R}\Gamma$ *which starts on* $u \in V$ *and ends on* $v \in V$*.*

The category **PathAlg** has objects of path algebras over the reals and morphisms consisting of linear maps $\phi : \mathbb{R}\Gamma_1 \to \mathbb{R}\Gamma_2$ that preserve path composition, meaning:

$$\phi(p_2 \circ p_1) = \phi(p_2) \circ \phi(p_1) \quad (2.242)$$

# 3. COMPUTING PERSISTENCE

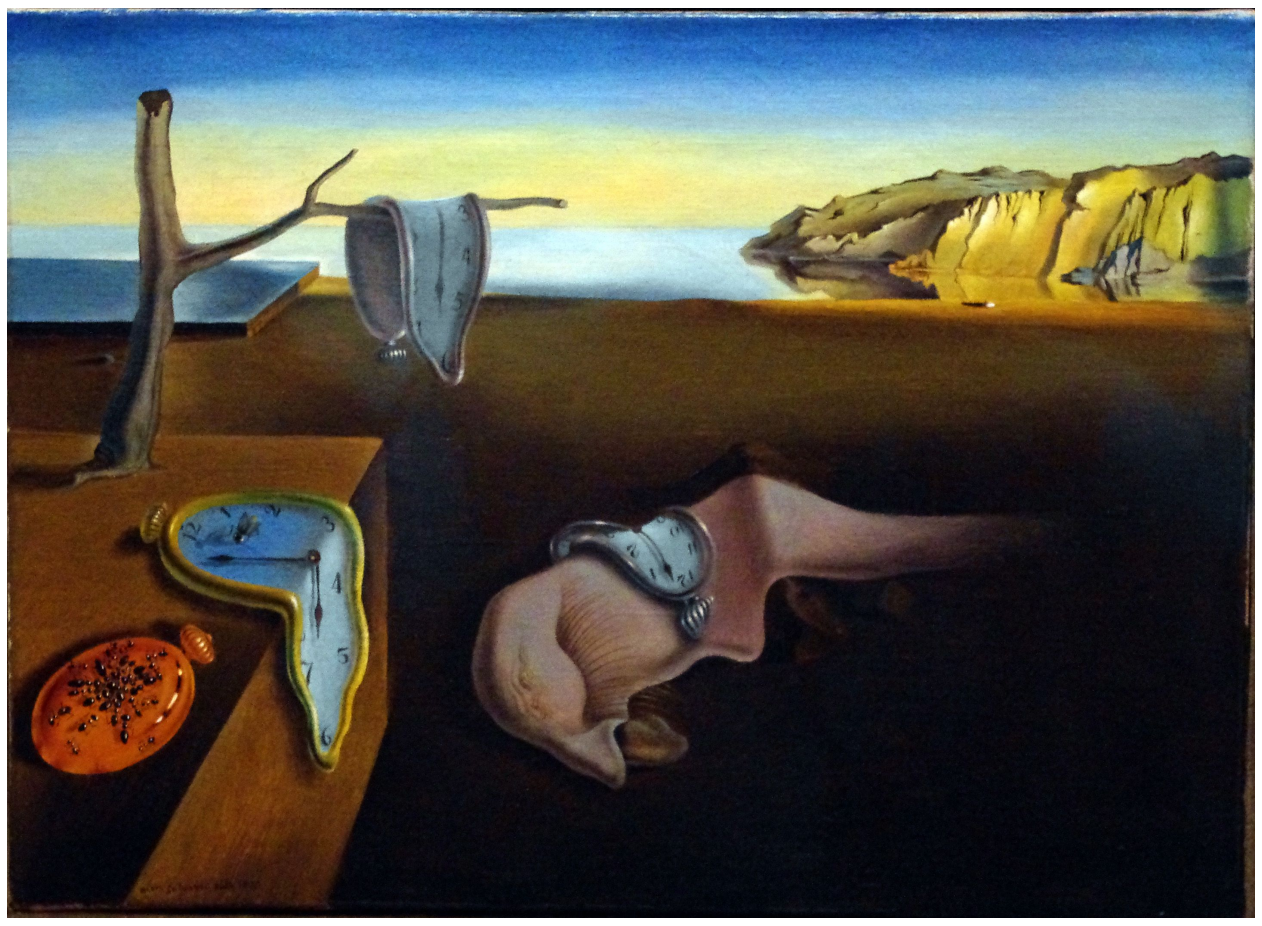

**Figure 3.1.** The painting Persistence of Memory presents a Surrealist example of the passage of time. (Salvador Dali, 1931) [28]

We go over persistence theory here. Originally motivated by the decomposition theorem of Gabriel [29], we give a more general interpretation in terms of causality.

We would like there to exist a global synchronized notion of time over the objects of a category. We can achieve this by simulating a simple directed acyclic graph on the category. For simple directed acyclic graphs, there always exists a topological sort [30] on the nodes. This topological sort of the objects forms a total ordering, which can be interpreted as having units of time.

To simulate simple directed acyclic graphs on a category, we can bring back Definition 2.3.22 as a finite subcategory.

Given a category $\mathcal{D}$, we define a simple finite simple acyclic subcategory in the following:

**Definition 3.0.1.** *A **simple finite simple acyclic subcategory** $Q(\mathcal{D}) \subseteq \mathcal{D}$ is a finite subcategory of $\mathcal{D}$ that is simple and acyclic.*

Let there be two categories $\mathcal{D}, \mathcal{E}$, a functor $F : \mathcal{D} \to \mathcal{E}$, and a **finite simple acyclic subcategory** of $\mathcal{D}$, denoted $Q(\mathcal{D}) = (V, E), V \subseteq \text{ob}(\mathcal{D}), E \subseteq \text{mor}(\mathcal{D})$ over $\mathcal{D}$.

We want to track the lifetime of an "independent generator" in $\text{im}(F|_{Q(\mathcal{D})})$. It would suffice to track the future merging and zeroing events of the generator composed across the morphisms of $\text{im}(F|_{Q(\mathcal{D})})$.

This is in analogy to checking for independence by Gaussian elimination over a matrix.

(3.1) A matrix $M \in \mathbb{R}^{n\times m}, n, m \in \mathbb{Z}^+$ can be viewed as a sequence of column vectors $v_\text{i} \in \mathcal{D} = \mathbb{R}^n, \text{i} \in [m]$ with $v_0 = \overrightarrow{0} \in \mathbb{R}^n$. The first i columns, $0 \leq \text{i} \leq m$, form a vector space $V_\text{i} \triangleq \text{span}(\{v_1, ..., v_\text{i}\})$. In Gaussian elimination, every column $v_\text{i}$ is checked for independence with its previous columns $v_1, ..., v_{\text{i}-1}$. Either it is independent or dependent with $\text{span}(\{v_1, ..., v_{\text{i}-1}\})$.

1. If it is independent, then $v_\text{i}$ can act as a generator.
2. If it is dependent, then we would like to know which vectors it is dependent with. Each time a vector $v_\text{i}$ is dependent on previous columns, $v_\text{i}$ is "merging" with the columns that span $V_{\text{i}-1}$.

   This **relation** is determined by **some linear combination** of $v_\text{i}$ with the previous $\text{i} - 1$ columns. A single column $v_s, s < \text{i}$ will belong to many such relationships into the future.

This is because by the generator-relation decomposition, also known as the presentation of $V_\text{i}$ we have:

$$V_\text{i} \cong \frac{< S_\text{i} >}{Syzgy_1(S_\text{i}, V_\text{i})} \tag{3.2}$$

where $S_\text{i} \triangleq \{v_1, ..., v_\text{i}\}$ and $< S_\text{i} >$ is a free $\mathbb{R}$-module with free generating set $S_\text{i}$ and the relations are determined by the 1st Syzygy subspace as given in Section 2.4.59 in the Algebra Background.

We can formulate the Gaussian elimination algorithm into the category theoretic framework:

Let $\mathcal{D}$ = **TotOrderedSet** where objects are totally ordered sets and morphisms are inclusions between totally ordered sets.

Let $\mathcal{E} = \mathbf{Sub}_{\mathbf{SpanVec}}((\text{Forget}(\mathbb{R}^n), \mathbb{R}^n))$ consisting of objects that are the cartesian product of sets of vectors from $\mathbb{R}^n$ and vector subspaces of $\mathbb{R}^n$ (up to isomorphism) for some $n < \infty$ with the morphisms being the pair of functions between sets and linear maps between them.

Define the set $S \triangleq \{v_{\text{i}} \in \mathbb{R}^n : \text{i} \in [m]\}$.

Let $S_{\text{i}} = \{v_1, ..., v_{\text{i}}\} \subseteq S$ and let $V_{\text{i}} \triangleq \text{span}(\{v_1, ..., v_{\text{i}}\}) \subseteq \mathbb{R}^n; V_0 \triangleq \mathbf{0} \subseteq \mathbb{R}^n$.

By the generator-relation decomposition of $V_{\text{i}}$ with respect to $< S_{\text{i}} >$ we can write $V_{\text{i}}$ by its presentation as:

$$V_{\text{i}} \cong \frac{< S_{\text{i}} >}{\text{Syzygy}_1(< S_{\text{i}} >, V_{\text{i}})} \tag{3.3}$$

Let $F : \mathbf{TotOrderedSet} \to \mathbf{Sub}_{\mathbf{SpanVec}}((\text{Forget}(\mathbb{R}^n), \mathbb{R}^n))$ be defined by:

- $F(S_{\text{i}}) = (S_{\text{i}}, V_{\text{i}})$ on objects.
- $F(\text{inc} : S_{\text{i}} \hookrightarrow S_{\text{j}}) = (\text{inc} : S_{\text{i}} \hookrightarrow S_{\text{j}}, \phi_{\text{i,j}} : V_{\text{i}} \to V_{\text{j}})$ where:

$$\phi_{\text{i,j}}(v) = v \in V_{\text{j}} \tag{3.4}$$

  and thus:

$$\text{inc}_{S_{\text{i}} \hookrightarrow S_{\text{j}}} = \phi_{\text{i,j}} \,|_{S_{\text{i}}} \tag{3.5}$$

Let the matrix $M = [v_1 \| \cdots \| v_m]$ be a concatentation of $m$ column vectors. Each time a column $v_{\text{i+1}}$ is examined in $M$ during Gaussian elimination, we can view this as an inclusion map $\text{inc} : S_{\text{i}} \hookrightarrow S_{\text{i+1}}$.

We would like to know:

1. (New Independence) $v_{\text{i+1}} \perp_{\mathbb{R}^n} \text{im}(\phi_{\text{i,i+1}})$? or
2. (New Dependencies) $v_{\text{i+1}} \not\perp_{\mathbb{R}^n} \text{im}(\phi_{\text{i,i+1}})$

   In the case of (2): $v_{\text{i+1}} \not\perp_{\mathbb{R}^n} \text{im}(\phi_{\text{i,i+1}})$, we can say that $v_{\text{i+1}} = \sum_{\text{j} \leq \text{i}} c_{\text{j}} v_{\text{j}}$ for $c_{\text{j}} \in \mathbb{R}$. We would like to know how these linear combinations relate to the vectors $v_{\text{j}}$ that "happened before" $v_{\text{i+1}}$, namely $\text{j} \leq \text{i}$.

In terms of generators and relations, using the presentation given by Equation 3.3, this is equivalent to adding a free generator $v_{\text{i+1}}$ to $< S_{\text{i}} >$ and then extending the 1st Syzygy modules from $\text{Syzygy}_1(< S_{\text{i}} >, V_{\text{i}})$ to $\text{Syzygy}_1(< S_{\text{i+1}} >, V_{\text{i+1}})$.

These two vector subspaces of $< S_{\mathrm{i}+1} >$ satisfy:

$$\mathrm{Syzygy}_1(< S_{\mathrm{i}} >, V_{\mathrm{i}}) \subseteq \mathrm{Syzygy}_1(< S_{\mathrm{i}+1} >, V_{\mathrm{i}+1}) \tag{3.6}$$

up to isomorphism. If the two vector spaces are isomorphic, then (1) occurs, otherwise (2) occurs.

Persistence theory on the tuple

$$\begin{aligned}(F : \mathbf{TotOrderedSet} \to \mathbf{Sub}_{\mathbf{SpanVec}}((\mathrm{Forget}(\mathbb{R}^n), \mathbb{R}^n)), Q(\mathbf{TotOrderedSet})\\ \triangleq (V \triangleq \{S_{\mathrm{i}}\}_{\mathrm{i}=0}^m, E \triangleq \{\mathrm{inc} : S_{\mathrm{i}} \hookrightarrow S_{\mathrm{j}}\}_{\mathrm{i},\mathrm{j} \in \{0,\ldots,m\}:\mathrm{i} \leq \mathrm{j}})\end{aligned} \tag{3.7}$$

asks the question:

**Question 3.0.1.** *For any $v_s \in \mathbb{R}^n, 1 \leq s \leq m$, for which* i *which "happening after" $s$, $s < \mathrm{i} \leq m$, do we have that:*

1. *(New Independence)* $v_{\mathrm{i}} \perp im(\phi_{\mathrm{i}-1,\mathrm{i}})$ *or*
2. *(New Dependencies)* $\exists c_s \neq 0$ *with* $v_{\mathrm{i}} = c_s v_s + \sum_{\mathrm{j}<\mathrm{i}} c_{\mathrm{j}} v_{\mathrm{j}}$

## 3.1 Persistence

The above question motivated by Gaussian elimination hints at a generalization.

For the Gaussian elimination case we had a composable sequence of maps on subspaces of $\mathbb{R}^n$. On the ith space $V_i \subseteq \mathbb{R}^n$, the next space $V_{i+1} \subseteq \mathbb{R}^n$ can be viewed as adding more generators and relations to the $V_i$. For a vector $u_s \in F(v_s)$ we would also like to track the merging of generators from the past j that "happened before" index i + 1 into the i + 1-th space due to a composition of maps from space j to space i + 1.

We can then generalize the above to the following definition:

**Definition 3.1.1.** *A* ***data representation functor on a finite simple acyclic subcategory*** *is defined by the pair:*

$$(F : \mathcal{D} \to \boldsymbol{SpanVec}, Q(\mathcal{D}) = (V, E)) \tag{3.8}$$

*where the image of the restricted functor must satisfy:*

$$im(F|_{Q(\mathcal{D})}) = \boldsymbol{Sub}_{\boldsymbol{SpanVec}}(\mathcal{W}) \tag{3.9}$$

*This is the category of subobjects of a common object $\mathcal{W} \triangleq (Forget(W), W) \in ob(\boldsymbol{SpanVec})$ where $W$ is a vector space. The object $\mathcal{W}$ is called an embedding space.*

The functor $F : \mathcal{D} \to$ **SpanVec** is a functor from some data category $\mathcal{D}$ to the category of vector spaces with spanning sets, **SpanVec**.

The category **SpanVec** is defined in Definition 2.4.35.

As in the case of Gaussian elimination, each new column vector provides new information to a sequence of linear maps. In analogy, for an object $(S, V) \in$ ob(**SpanVec**), $S$ provides a set of free generators to the presentation of a vector space from a vector representation of ob($\mathcal{D}$). If these free generators can be determined without ambiguity, then the target category of the functor $F$ can be replaced with **Vec**.

The finite simple acyclic subcategory $Q(\mathcal{D})$ represents some relevant observable portion of the data which can be measured by the functor $F$. Due to acyclicity of $Q(\mathcal{D})$, finite paths of objects can be measured unambiguously. This is discussed in Section 2.3.15. We will use this to measure the persistence of generators across a sequence of composable arrows from $\mathrm{im}(F\mid_{Q(\mathcal{D})})$.

Since the category $\mathrm{im}(F\mid_{Q(\mathcal{D})})$ is a category of subobjects of a single object $\mathcal{W}$, we can view the category $\mathrm{im}(F\mid_{Q(\mathcal{D})})$ as an "embedding space" for all of the vector representations of the data from $Q(\mathcal{D})$. As defined in Definition 2.4.35, a subobject of $\mathcal{W} = (\mathrm{Forget}(W), W)$ consists of a subset $S$ of vectors from $W$ and an injective linear map $\psi : V \to W$ where $\mathrm{span}(S) = V$.

(3.10) Let $\mathcal{W} \triangleq (\mathrm{Forget}(\mathbb{R}^n), \mathbb{R}^n)$. The subobjects are equivalent to all the subspaces of $\mathbb{R}^n$ up to the isomorphisms of $\mathcal{W}$ from the **SpanVec** category.

For example, any vector space of the form: $V = \frac{\mathbb{R}^m}{\mathbb{R}^d}$ with $0 \le m - d \le n$ spanned by some subset $S$ of vectors from $V$ has an injective linear map

$$\psi : V \to \mathbb{R}^n \tag{3.11}$$

The map $\psi$ determines the representatives of each coset from $V$. However, the map and its domain, which is spanned by $S$, are determined up to isomorphism.

In particular, for any isomorphic space $V' \cong V$ spanned by $S' \subseteq V'$ with $|S'| = |S|$ and any injective linear map $\psi' : V' \to W$, there exists an isomorphism $\Phi : V' \to V$ with:

$$\psi \circ \Phi = \psi' \tag{3.12}$$

Since both $\psi'$ and $\psi$ are injective linear maps on isomorphic domains, we must have:

1. $\psi'$ is invertible on the linear subspace $\mathrm{im}(\psi') \subseteq W$.

   Similarly, $\psi$ is invertible on $\mathrm{im}(\psi) \subseteq W$.
2. $\mathrm{im}(\psi') \cong \mathrm{im}(\psi)$ through an isomorphism.
3. $\exists \rho : \mathbb{R}^n \to \mathbb{R}^n$ an automorphism with:

$$\rho\mid_{\mathrm{im}(\psi')} : \mathrm{im}(\psi') \to \mathrm{im}(\psi) \tag{3.13}$$

coming from the isomorphism of (2) and:

$$\rho\mid_{\mathbb{R}^n - \mathrm{im}(\psi')}: (\mathbb{R}^n - \mathrm{im}(\psi')) \to (\mathbb{R}^n - \mathrm{im}(\psi')) = \mathrm{id}_{\mathbb{R}^n - \mathrm{im}(\psi')} \tag{3.14}$$

This makes $\rho = \rho\mid_{\mathrm{im}(\psi')} \oplus \mathrm{id}_{\rho|_{\mathrm{im}(\psi')}}$

Using these facts, we can define the map $\Phi : V' \to V$ as the following linear map:

$$\Phi \triangleq \psi^{-1} \circ \rho \circ \psi' \tag{3.15}$$

This means that the subobjects of $\mathcal{W}$ are the subspaces of $\mathbb{R}^n$ spanned by some subset $S \subseteq \mathrm{Forget}(\mathbb{R}^n)$ up to isomorphism.

We can thus ask the following question on a data representation functor:

**Question 3.1.1.** *(Persistence across Vector Spaces)*

*Let $(F : \mathcal{D} \to \boldsymbol{Span\,Vec}, Q(\mathcal{D}) = (V, E))$ be a data representation functor on a finite simple acyclic subcategory and let $s, t \in \mathbb{N}$ with $t > s$.*

*From the subcategory $Q(\mathcal{D}) \subseteq \mathcal{D}$, consider a sequence of $t - s$ composable maps starting from an object $u \in V(Q(\mathcal{D}))$ and ending at an object $v \in V(Q(\mathcal{D}))$.*

*Relabeling $v_s \triangleq u, v_t \triangleq v$ and letting the composable maps from $v_s$ to $v_t$ be denoted by $\mathrm{e}_{s,s+1}, \ldots, \mathrm{e}_{t-1,t}$ with $\mathrm{e}_{\mathrm{j},\mathrm{j}+1} : v_{\mathrm{j}} \to v_{\mathrm{j}+1} \in E(Q(\mathcal{D}))$.*

*Letting*

$$(\phi_{\mathrm{j},\mathrm{j}+1}\mid_{S_{\mathrm{j}}}, \phi_{\mathrm{j},\mathrm{j}+1}) \triangleq F(\mathrm{e}_{\mathrm{j},\mathrm{j}+1}) : F(v_{\mathrm{j}}) \to F(v_{\mathrm{j}+1}), \mathrm{j} = s+1, \ldots, t-1 \tag{3.16a}$$

$$\textit{for } (S_{\mathrm{j}}, V_{\mathrm{j}}) = F(v_{\mathrm{j}}), \mathrm{j} = s+1, \ldots, t \tag{3.16b}$$

*For any vector $u_s \in S_s$ from $(S_s, V_s) = F(v_s)$:*

*For each index $\mathrm{i}$ from index $s+1$ to index $t$, we can ask the following:*

1. (***New Independences***)

*Compute the set of vectors*

$$\tilde{S}_{\rm i} \triangleq S_{\rm i} \bigcap (V_{\rm i} \setminus im(\phi_{\rm i-1,i})) \tag{3.17}$$

*which must be vectors from* $V_{\rm i}$ *to add to* $\bigcup_{\rm j \le i-1} \tilde{S}_{\rm j}$ *for the following presentation of* $V_{\rm i}$*:*

$$V_{\rm i} \cong \frac{< \bigcup_{\rm j=\mathit{s}+1}^{\rm i} \phi_{\rm i-1,i} \circ \cdots \circ \phi_{\rm j,j+1}(\tilde{S}_{\rm j}) >}{Syzygy_1(< \bigcup_{\rm j=\mathit{s}+1}^{\rm i} \phi_{\rm i-1,i} \circ \cdots \circ \phi_{\rm j,j+1}(\tilde{S}_{\rm j}) >, V_{\rm i})} \tag{3.18}$$

*where the* $Syzygy_1$ *subspace is uniquely defined by Corollary 2.4.72 to Hilbert's Syzygy Theorem.*

2. (***New Dependencies***)

   (a) (***Dependencies from*** $V_{\rm i}$) *Are there linear dependencies amongst the vectors* $\tilde{S}_{\rm i}$ *in* $V_{\rm i}$*? Find them.*

   (b) (***Merging of independent generators from the past***) *Let* $w_{s,\rm i} = \phi_{\rm i-1,i} \circ \cdots \circ \phi_{s,s+1}(u_s)$. *Which vectors* $v_{{\rm j};k} \in \tilde{S}_{\rm j} : s < {\rm j} < {\rm i}$, *where* $k$ *indexes the vectors in* $\tilde{S}_{\rm j}$, *have that* $\phi_{\rm i-1,i} \circ \cdots \circ \phi_{\rm j,j+1}(v_{{\rm j};k}) \nmid_{V_{\rm i}} w_{s,\rm i}$*?*

*In Part (2.b.), we notice that once a free generator in* $\tilde{S}_{\rm j}$ *becomes dependent with* $w_{s,\rm i}$*, it loses its independence into the future for all* ${\rm i}' > {\rm i}$.

We can visualize the above three parts to the question in Figure 3.2. The three colors: green, red and blue, correspond to the three questions being asked.

1. Green represents the creation of independent vectors.
2. Red represents the vector space and its own algebraic structure.
3. Blue represents the passing of vector signal across the composition of maps.

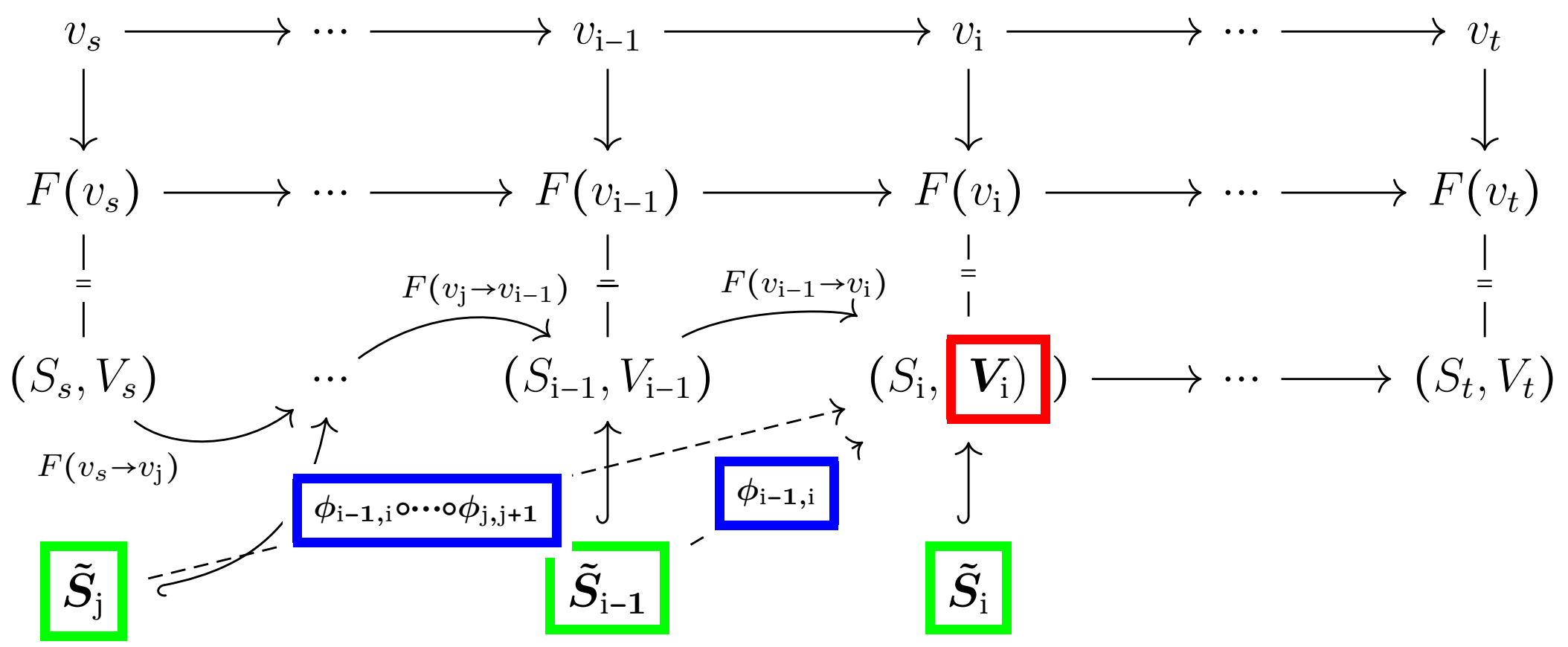

**Figure 3.2.** An illustration of a sequence of composable maps $e_{s,s+1} : v_s \to v_{s+1}, \ldots, e_{t-1,t} : v_{t-1} \to v_t$ in the context of the data representation functor and the independence/dependence Question being asked at index i, $s < i \leq t$

### 3.1.2 Examples

We go over here examples that illustrate this question of persistence

(3.19) When $\mathcal{D} = Q(\mathcal{D})$ is a finite directed multigraph where the objects are the elements, called nodes, of a finite set and the morphisms are the (finite in number of) directed edges, called arrows, between nodes. Assume that all self-loops, or directed edges from a node to itself, exist. Then $F|_{Q(\mathcal{D})} : Q(\mathcal{D}) \to \mathbf{Vec}$ is called a **quiver representation** [31]. The path algebra $\mathbb{R}\Gamma$ on $Q(\mathcal{D})$ can act on the following vector space:

$$V \triangleq \bigoplus_{V_i \in \mathrm{ob}(\mathrm{im}(F|_{Q(\mathcal{D})}))} V_i \tag{3.20}$$

by the map:

$$(\alpha_{u,v}) \cdot w \triangleq \begin{cases} \mathrm{inc}_{v,V} \circ F|_{Q(\mathcal{D})}(u \to v) \circ \pi_u(w) & u \neq v \\ \mathrm{inc}_{u,V} \circ \pi_u(w) & u = v \end{cases}, \forall w \in V \tag{3.21}$$

where $\alpha_{u,v}$ is a path basis vector in $\mathbb{R}\Gamma$, see Definition 2.8.20, which allows us to define a (left)$\mathbb{R}\Gamma$-module [32].

This allows us to have an algebraic correspondence theorem for quiver representations:

**Proposition 3.1.3.** *([31]) The category of all quiver representations is isomorphic to the category of (left)* $\mathbb{R}\Gamma$*-modules.*

When $Q(\mathcal{D})$ has special properties, the quiver representation often recovers many decomposition theorems common to algebraic systems. This is the primary contribution of quiver representation theory such as Kac's theorem [14, 33, 34] and Gabriel's Theorem [29].

**Persistent homology:**

We will show in later chapters the case of a quiver representation from a homology functor $H_\bullet : \mathbf{Simp} \to \mathbf{Vec}$ over a filtered simplicial complex

$$Q(\mathbf{Simp}) \triangleq \left(\{K_0 = \varnothing\} \cup \{K_\mathrm{i} : |K_\mathrm{i} \setminus K_{\mathrm{i}-1}| = 1\}_{\mathrm{i}=1}^{n}, \{K_\mathrm{i} \hookrightarrow K_\mathrm{j}\}_{\mathrm{i},\mathrm{j} \in ([n] \cup \{0\}) : \mathrm{i} \leq \mathrm{j}}\right) \tag{3.22}$$

We show that the quiver representation is equivalent to a $\mathbb{R}[t]$-module. The decomposition of this $\mathbb{R}[t]$-module into one dimensional subspaces, see Theorem 6.0.2, can provide an answer to the "Question of Persistence across Vector Spaces" on:

$$(H_\bullet : \mathbf{Simp} \to \mathbf{Vec}, Q(\mathbf{Simp})) \tag{3.23}$$

The assumption of simplex-wise inclusions:

$$|K_\mathrm{i} \setminus K_{\mathrm{i}-1}| = 1 \tag{3.24}$$

is needed in order to not have to define an accumulating set of free generators explicitly.

**Example:** For example, at time index 1 say we have:

$$K_1 \triangleq \{\{v_1\}, \{v_2\}, \{v_3\}, \{v_4\}, \{v_1, v_2\}, \{v_2, v_3\}, \{v_3, v_4\}\} \tag{3.25}$$

and, alternatively, let:

$$K_1' \triangleq \{\{v_1\},\{v_2\},\{v_3\},\{v_4\},\{v_1,v_2\},\{v_1,v_3\},\{v_1,v_4\}\} \tag{3.26}$$

both be simplicial complexes. The 0-dimensional homology group of either $K_1$ or $K_1'$ through their boundary maps are rank 1 and thus isomorphic. However, $K_1 \ncong K_1'$ since only $K_1$ forms a path. However, both consist of the same number of nodes and edges. In fact, a single connected component could have infinitely many possible simplicial complexes that form it.

With a simplex-wise filtration, it becomes clear that if a new homology vector space generator appears, this is the only generator that could form since

$$\dim(\ker(\partial_p^{K_i})) - \dim(\ker(\partial_p^{K_{i-1}})) \leq 1 \tag{3.27}$$

since there is only a single $\sigma_i \in K_i \setminus K_{i-1}$ that could contribute to a rank change.

(3.28) We give an example from language modeling: Let there be a sequence of words $X_1, ..., X_n \in \mathcal{W}^n$ where $\mathcal{W}$ is a set of words in some language.

**The Data and Vector Space Categories:**

Let **Language** be a category where:

1. ob(**Language**) consists of the collection of all sets of words
2. and mor(**Language**) is the set of all functions over sets of words.

Let **Vec** be the category where:

1. ob(**Vec**) is the set of all $d$-dimensional vector spaces
2. and mor(**Vec**) consists of all the linear maps between $d$-dimensional vector spaces.

**The Functor:** Let $F$ : **Language** → **SpanVec** denote a covariant functor between these two categories.

1. It can be defined on objects through a word embedding $\text{emb} : \mathcal{W} \to \mathbb{R}^d$, $d \in \mathbb{Z}^+$ as follows:

   For $S = \{X_1, ..., X_s\}$,

$$F(S) \triangleq (\{\text{emb}(X_1), ..., \text{emb}(X_s)\}, \text{span}(\{\text{emb}(X_1), ..., \text{emb}(X_s)\})) \tag{3.29}$$

2. For two sets of words $S = \{X_1, ..., X_s\}, T = \{Y_1, ..., Y_t\}; S, T \in \text{ob}(\textbf{Language})$ and any function $f : S \to T$, let the functor map as an induced map as follows:

$$F(f : S \to T) : F(S) \to F(T) \tag{3.30}$$

   have:

$$F(f : S \to T) \triangleq (g, \tilde{g}) \tag{3.31}$$

   where

$$\begin{aligned} g : \{\text{emb}(X_1), ..., \text{emb}(X_s)\} \to \{\text{emb}(f(X_1)), ..., \text{emb}(f(X_s))\} \\ g(\text{emb}(X_\text{i})) \triangleq \text{emb}(f(X_\text{i})) \end{aligned} \tag{3.32}$$

   and $\tilde{g}$ is a linear extension of $g$ from $\text{span}(\text{emb}(X_1), ..., \text{emb}(X_s))$ to $\text{span}(\text{emb}(f(X_1)), ..., \text{emb}(f(X_s)))$

It can be checked that $F$ is a covariant functor.

**The Finite Simple Acyclic Subcategory:** Define the finite simple acyclic subcategory $Q(\textbf{Language}) = (V, E)$ with

$$V \triangleq \{S_\text{i}\}_{\text{i}=0}^n, E \triangleq \{\text{inc}_{\text{i,j}} : S_\text{i} \hookrightarrow S_\text{j}\}_{\text{i,j} \in \{0,...,n\}:\text{i} \leq \text{j}} \tag{3.33}$$

which is a filtration of sets of words: $\varnothing \subseteq S_1 \subseteq ... \subseteq S_n$, with $S_\text{i} \triangleq \{X_1, ..., X_\text{i}\}$

The question that persistence theory asks on the tuple:

$$(F : \textbf{Language} \to \textbf{SpanVec}, Q(\textbf{Language})) \tag{3.34}$$

**Question 3.1.2.** *Given the s-th word embedding* $emb(X_s) \in \mathbb{R}^d$*, for a* $t > s$*, the future words indexed at* $\mathrm{i}, s < \mathrm{i} \le t$ *"happening after" word s we can ask:*

1. *(**New Independences**) Let the word embedding* $emb(X_\mathrm{i})$ *be a free generator for the following presentation of* $V_\mathrm{i}$ *for* $(E_\mathrm{i}, V_\mathrm{i}) = F(S_\mathrm{i})$ *where* $E_\mathrm{i} = \{emb(X_\mathrm{j})\}_{\mathrm{j}=1}^{\mathrm{i}}$ *and* $V_\mathrm{i} = span(E_\mathrm{i})$

$$V_\mathrm{i} \cong \frac{< E_\mathrm{i} >}{Syzygy_1(< E_\mathrm{i} >, V_\mathrm{i})} \tag{3.35}$$

*where the isomorphism follows by Corollary 2.4.72 to Hilbert's Syzygy Theorem.*

2. *(**New Dependencies from** $V_\mathrm{i}$) Is* $emb(X_\mathrm{i})$ *linearly dependent to* $emb(X_s)$*?*

*We do not ask about the dependencies from the maps. They do not exist since the maps are injective.*

(3.36) Consider the fork-join model [35] from parallel computing, a generalization of the bulk-synchronous parallel model [36–38]:

A thread can be defined as a basic unit of sequential execution. Every thread has a thread id, called the tid. This is a positive integer that uniquely identifies the order in time for when the thread was launched relative to all possible threads. This includes an ordering for the threads from forking operations. The largest tid has value at most $L$, $L < \infty$.

**The Data Category:**

Let **ThreadPool** denote the category of thread pools where:

1. ob(**ThreadPool**) consists of all thread pools of living threads timestamped by a number $t \in \mathbb{R}$. The objects are the thread pools at time $t$, denoted $T_{t,a}$. Letting

$$\mathrm{tid}^*(t) = \max_{t \in T_{t,a}}[\mathrm{tid}(t)] \tag{3.37}$$

We have that this is a strictly monotonically increasing function in $t$, meaning:

$$\mathrm{tid}^*(\hat{t}) > \mathrm{tid}^*(t) \text{ if } \hat{t} > t \tag{3.38}$$

Each thread pool cannot surpass the maximum number of threads the machine can run at a time, which is less than $L$.

Let $T$ denote the set of all possible threads that could be issued over all of time of size $L$. At each time $t$, we have the following separation of $T$:

(a) (Activated threads) $T_t \subseteq T$ denotes the activated threads from the past and present. All threads in $T_t$ have that their tid is less than or equal to $\mathrm{tid}^*(t)$. The activated threads can be broken up into these two types:

   i. $T_{t,a} \subseteq T_t$ denotes all threads that have not completed execution, or are "alive." These form the thread pool at time $t$.

   ii. $T_t \setminus T_{t,a}$ denotes threads that have completed execution.

(b) (Unactivated threads) $T_{t,b} \triangleq \{u : \mathrm{tid}(u) > \mathrm{tid}^*(t)\}$ denotes all unactivated threads. An unactivated thread is a potential thread that could be added to a thread pool in the future.

2. The set $\mathrm{mor}(\mathbf{ThreadPool})$ is defined by the fork-join operations at each time $t$, which guarantee that at time $t + \epsilon$, for some $\epsilon > 0$, the threads will achieve their desired behavior. A "living" thread from $T_{t,a}$ can (exclusively) either fork, join with another thread, or continue staying "alive."

   - When it **forks**, it activates unactivated threads from $T_{t,b}$, issuing a different tid for each new thread larger than $\mathrm{tid}^*(t)$ and continues its own thread as active.
   - When it is related to a **joining event**, either it:
     (a) Completes its execution.
     (b) Or, it unblocks and **joins with a (unique) completing thread**, where the other thread completes its execution. This thread then continues execution after unblocking.
   - When it **does neither**, it stays "alive."

   We assume that at time $t + \epsilon$ all living threads must come from exactly one of these three operations and no other. In other words, there are no "external living threads" that need to be kept track of. Thus we have the following disjoint decomposition of the thread pool $T_{t,a}$:

$$T_{t,a} = \text{Forking}_t \bigsqcup \text{Joining}_t \bigsqcup \text{Staying}_t \tag{3.39}$$

We denote a fork-join-stay map, which represents joining, forking, and staying alive separately as the following surjective map:

$$(\text{Fork} \oplus \text{Join} \oplus \text{Iden})_{t,t+\epsilon}(u) \triangleq \begin{cases} (U \circ \widetilde{\text{Fork}}_{t,t+\epsilon})(u) & u \in \text{Forking}_t \\ \text{Join}_{t,t+\epsilon}(u) & u \in \text{Joining}_t \\ \text{Iden}_{t,t+\epsilon}(u) & u \in \text{Staying}_t \end{cases} \tag{3.40}$$

These maps are explicitly defined below:

Forking for a thread results in non-deterministic behavior along with a deterministic identity behavior for the thread.

$$\begin{aligned} \widetilde{\text{Fork}}_t : \text{Forking}_t \to 2^{T_{t,b}} \cup \{\text{Forking}_t\} \\ \widetilde{\text{Fork}}_t(u) \triangleq \{u' \in T_{t,b} : u' \text{ are newly launched threads by } u\} \cup \{\text{Iden-Fork}(u)\} \end{aligned} \tag{3.41a}$$

$$\begin{aligned} \text{Iden-Fork}_{t,t+\epsilon} : \text{Forking}_t \to T_{t+\epsilon,a} \\ \text{Iden-Fork}_{t,t+\epsilon}(u) \triangleq u \end{aligned} \tag{3.41b}$$

$$\begin{aligned} U : 2^T \to T_{t+\epsilon,a} \\ U(V) \triangleq \bigcup_{v \in V} v \end{aligned} \tag{3.41c}$$

$$\begin{aligned} \text{Join}_{t,t+\epsilon} : \text{Joining}_t \to T_{t+\epsilon,a} \\ \text{Join}_{t,t+\epsilon}(u) \triangleq \begin{cases} u'(u) & \text{if } u \text{ is a terminating thread} \\ u & \text{if } u \text{ is a joining thread} \end{cases} \end{aligned} \tag{3.41d}$$

where $u'(u)$ is the unique thread that joins with thread $u$.

$$\begin{aligned} \mathrm{Iden}_{t,t+\epsilon} &: \mathrm{Staying}_t \to T_{t+\epsilon,a} \\ \mathrm{Iden}_{t,t+\epsilon}(u) &\triangleq u \end{aligned} \tag{3.41e}$$

Forking only affects the threads from $\mathrm{Forking}_t$ and newly launched threads which have thread id's strictly greater than all threads in $T_{t,a}$. For joining, $\mathrm{Join}_{t,t+\epsilon}$ is a merging map and for identity, $\mathrm{Iden}_{t,t+\epsilon}$ is an identity map. The images of both maps do not intersect and both cannot involve new threads. Thus we have the following decomposition of $T_{t+\epsilon,a}$:

$$T_{t+\epsilon,a} = \mathrm{im}(U \circ \widetilde{Fork}_{t,t+\epsilon}) \bigsqcup \mathrm{im}(\mathrm{Join}_{t,t+\epsilon}) \bigsqcup \mathrm{im}(\mathrm{Iden}_{t,t+\epsilon}) \tag{3.42}$$

which is a disjoint union.

**The Representation Category:**

The set $(\mathbb{Z}_2)^L$ consists of all bit vectors of length $L$. A bit vector is a $L$-tuple of zeros and ones.

The set of bit vectors $(\mathbb{Z}_2)^L$ can be viewed as a vector space of dimension $L$ over the field $\mathbb{Z}_2$ spanned by the standard basis of $L$-length columns $\mathrm{e_i}, \mathrm{i} = 1, ..., L$.

$$\mathrm{e_i[j]} \triangleq \begin{cases} 1 & \mathrm{j = i} \\ 0 & \mathrm{j \neq i} \end{cases}, \forall \mathrm{j} \in [L] \tag{3.43}$$

Let $\mathbf{Sub}_{\mathbf{SpanVec}}(\{\mathrm{e}_1, ..., \mathrm{e}_L\}, (\mathbb{Z}_2)^L)$ be a category having:

1. Objects of the form:

$$B_S \triangleq (E_S, \mathrm{span}(E_S)) \text{ for } E_S \triangleq \{\mathrm{e_i}\}_{\mathrm{i} \in S}, S \subseteq [L] \tag{3.44}$$

2. Morphisms being linear maps $(f|_{E_S}, f : B_S \to B_{\tilde{S}})$ which have $f$ map on the standard basis of $(\mathbb{Z}_2)^L$ as follows:

$$\forall \mathrm{i} \in S, \exists! \mathrm{j} \in \tilde{S} : f(\mathrm{e_i}) = \mathrm{e_j} \tag{3.45a}$$

satisfying linearity:

$$f(\sum_{\mathrm{i}\in S} c_{\mathrm{i}} \cdot \mathrm{e_i}) = c_{\mathrm{i}} \cdot \sum_{\mathrm{i}\in S} f(\mathrm{e_i}), c_{\mathrm{i}} \in \mathbb{Z}_2 \tag{3.45b}$$

**Matrix Representation of the Morphisms**

We have by Proposition 2.4.45 that we can extend $f : B_S \to B_{S'}$ to $\tilde{f} : (\mathbb{Z}_2)^L \to (\mathbb{Z}_2)^L$ since $B_S, B_{S'} \subseteq (\mathbb{Z}_2)^L$ are linear subspaces. Furthermore, by Section 2.4.20 if we choose a basis for the domain and codomain of $\tilde{f} : (\mathbb{Z}_2)^L \to (\mathbb{Z}_2)^L$, then there is a unique matrix representation in the category $\mathbb{Z}_2$-**Mat**.Let the basis for $(\mathbb{Z}_2)^L$, which is both the domain and the codomain, be the set $E_{[L]} \subseteq (\mathbb{Z}_2)^L$.

This matrix is denoted by $[\tilde{f}]_{E_{[L]}} \in (\mathbb{Z}_2)^{L\times L}$. It is called a $L \times L$ **boolean matrix**, which must be a table of bits. This $L \times L$ boolean matrix **uniquely** identifies a linear map $f$. We will thus define the morphisms by $L \times L$ matrices.

**The Functor:** Let $F : \textbf{ThreadPool} \to \textbf{Sub}_{\textbf{SpanVec}}((\{\mathrm{e}_1, ..., \mathrm{e}_L\}, (\mathbb{Z}_2)^L))$ be a functor that:

1. On its objects, maps a thread pool at time $t$ to a set of length $L$ bit vectors. This is defined as follows:

$$F(T_{t,a}) \triangleq (E_S, B_S) \text{ where } S = \{\mathrm{tid}(q) : q \in T_{t,a}\} \subseteq [L] \tag{3.46}$$

   and $\mathrm{tid}(q) \in [L]$ is the thread id of thread $q \in T_{t,a}, T_{t,a} \subseteq T_t$.

2. On the morphisms of **ThreadPool**:

$$F((\mathrm{Fork} \oplus \mathrm{Join} \oplus \mathrm{Iden})_{t,t+\epsilon}) \triangleq [\mathrm{Iden\text{-}Fork}_{t,t+\epsilon}] \oplus [\mathrm{Join}_{t,t+\epsilon}] \oplus [\mathrm{Iden}_{t,t+\epsilon}] \tag{3.47}$$

which is a $L \times L$ boolean matrix, corresponding to a morphism in $\mathbf{Sub}_{\mathbf{SpanVec}}((\{\mathrm{e}_1,...,\mathrm{e}_L\},(\mathbb{Z}_2)^L))$. We only model the identity part of the forking operation. The non-deterministic part cannot be modeled by a function.

**Instead, the newly forked threads at time $t+\epsilon$ will come from $F(T_{t+\epsilon,a})$.**

The behavior of $[\text{Iden-Fork}_{t,t+\epsilon}] \oplus [\text{Join}_{t,t+\epsilon}] \oplus [\text{Iden}_{t,t+\epsilon}]$ is defined for each standard basis column $\mathrm{e_i} \in F(T_{t,a})$ as follows:

$$([\text{Fork}_{t,t+\epsilon}] \oplus [\text{Join}_{t,t+\epsilon}] \oplus [\text{Iden}_{t,t+\epsilon}]) \cdot \mathrm{e_i} \triangleq \begin{cases} \overrightarrow{0} & \text{if thread j joins with thread i} \\ \mathrm{e_i} & \text{Otherwise} \end{cases} \tag{3.48}$$

which defines how each thread in $T_{t,a}$ joins or stays alive.

Since all objects in $\mathrm{ob}(\mathbf{Sub}_{\mathbf{SpanVec}}((\{\mathrm{e}_1,...,\mathrm{e}_L\},(\mathbb{Z}_2)^L)))$ are free modules over $\mathbb{Z}_2$, all forked tids at each time $t$ are unique and each of them larger than the monotonically increasing $\mathrm{tid}^*(t)$.

**The Finite Simple Acyclic Subcategory:** Let $Q(\mathbf{ThreadPool}) = (V,E)$ be a sequence of $n$ thread pools indexed by time $t \in [0,T\epsilon], T \in \mathbb{N}$ where times are of the form $\mathrm{i}\epsilon, \mathrm{i} \in \mathbb{N}$, each separated by a time quanta of $\epsilon > 0$. We then have:

$$\begin{aligned} V(Q(\mathbf{ThreadPool})) &\triangleq \{T_{t,a}\}_{t=\hat{t}\epsilon,\hat{t}\in\{0\}\cup[T]}, \\ E(Q(\mathbf{ThreadPool})) &\triangleq \{(\text{Fork} \oplus \text{Join} \oplus \text{Iden})_{t,t+\epsilon}\}_{t=\hat{t}\epsilon,\hat{t}\in\{0\}\cup[T]} \end{aligned} \tag{3.49}$$

The question that persistence theory asks on the tuple

$$(F : \mathbf{Thread\ Pool} \to \mathbf{Sub}_{\mathbf{SpanVec}}((\{\mathrm{e}_1,...,\mathrm{e}_L\},(\mathbb{Z}_2)^L)), Q(\mathbf{ThreadPool})) \tag{3.50}$$

is the following:

**Question 3.1.3.** *At time $s = \hat{s}\epsilon, \hat{s} \in \{0\} \cup [T]$ we have that*

$$(E_{S_s}, B_{S_s}) = F(T_s) \tag{3.51}$$

*Let* $\mathrm{e_j} \in E_{S_s}$ *be some thread with tid* $\mathrm{j} \in [L]$.

*Let* $t = \hat{t}\epsilon, \hat{s} < \hat{t} \leq T$ *"happening after" s, meaning* $t > s$ *do we have that each time* $\mathrm{i} = \hat{\mathrm{i}}\epsilon, \hat{s} \leq \hat{\mathrm{i}} \leq \hat{t}$ *from time* $s + \epsilon$ *to time* $t = \hat{t}\epsilon$*, we ask the following:*

$$(E_{S_\mathrm{i}}, B_{S_\mathrm{i}}) = F(T_{\mathrm{i},a}), \mathrm{i} = s + \epsilon, \ldots, t \tag{3.52}$$

1. (***New Independences***) *Compute the set of vectors*

$$\tilde{E}_{S_\mathrm{i}} \triangleq E_{S_\mathrm{i}} \bigcap (B_{S_\mathrm{i}} \setminus im(\phi_{\mathrm{i}-1,\mathrm{i}})) \tag{3.53}$$

   *which are considered free generators to add to* $\bigcup_{\mathrm{j} \leq \mathrm{i}-1} \tilde{E}_{S_\mathrm{j}}$ *for the presentation of* $B_{S_\mathrm{i}}$. *These free generators are standard basis vectors of the form* $\mathrm{e}_{k^*} \in (\mathbb{Z}_2)^L$ *with a single nonzero at* $\mathrm{j} \in [L]$.

   ***These nonzero indices are the tids of the newly forked threads at time*** $\mathrm{i}$***.***

$$B_{S_\mathrm{i}} \cong \frac{< \bigcup_{\mathrm{j}=s+\epsilon}^{\mathrm{i}} \tilde{E}_{S_\mathrm{j}} >}{Syzygy_1(< \bigcup_{\mathrm{j}=s+\epsilon}^{\mathrm{i}} \tilde{E}_{S_\mathrm{j}} >, B_{S_\mathrm{i}})} \tag{3.54}$$

   *where the isomorphism follows by Corollary 2.4.72 to Hilbert's Syzygy Theorem.*

2. (***Merging of independent generators from the past***) *Let* $w_{s,\mathrm{i}} = \phi_{\mathrm{i}-1,\mathrm{i}} \circ \cdots \circ \phi_{s,s+1}(\mathrm{e}_{k^*})$. *Which free generators* $\mathrm{e}_k \in \tilde{E}_{S_\mathrm{j}}, s < \mathrm{j} < \mathrm{i}$ *have that* $\phi_{\mathrm{i}-1,\mathrm{i}} \circ \cdots \circ \phi_{\mathrm{j},\mathrm{j}+1}(\mathrm{e}_{k^*}) \not\perp_{B_{S_\mathrm{i}}} w_{s,\mathrm{i}}$*?*

*We do not have any dependencies from* $B_{S_\mathrm{i}}$ *for any* $\mathrm{i} : s < \mathrm{i} \leq t$ *since all newly introduced generators are standard basis vectors and linearly independent.*

### 3.1.4 Multiparameter Persistence

In the literature "multiparameter persistence" [39, 40], which is a part of quiver representation [14] theory, is often mentioned. This is given by the data representation functor $(F : \mathcal{D} \to \mathbf{Vec}, Q(\mathcal{D}))$ where $Q(\mathcal{D}) \subseteq \mathcal{D}$ is a finite posetal subcategory of category $\mathcal{D}$. As usual, **Vec** consists of only finite dimensional vector spaces for objects and thus is a locally small category.

Multiparameter persistence is motivated by the quiver isomorphism problem, which asks whether two quiver representations $F : \mathcal{D} \to \textbf{Vec}$, $G : \mathcal{D} \to \textbf{Vec}$ are naturally isomorphic. This is usually solved by finding a **quiver representation invariant**.

**Definition 3.1.5.** *An invariant of the quiver representation* $F|_{Q(\mathcal{D})}$ *is a map of the form* $f_{Thin(\textbf{Vec})} : 2^{Thin(\textbf{Vec})} \to \{0,1\}$ *for the binary target set* $\{0,1\}$ *and where* $2^{Thin(\textbf{Vec})}$ *is the category of all posetal subcategories of* ***Vec****. An invariant must also satisfy the property that the following diagram commutes for any other quiver representations* $G : \mathcal{D} \to \textbf{Vec}$*.*

$$\begin{array}{ccc} F|_{Q(\mathcal{D})} & \cong & G|_{Q(\mathcal{D})} \\ \downarrow & & \downarrow \\ im(F|_{Q(\mathcal{D})}) & & im(G|_{Q(\mathcal{D})}) \\ \downarrow {\scriptstyle f_{Thin(\textbf{Vec})}} & \swarrow {\scriptstyle f_{Thin(\textbf{Vec})}} & \\ \{0,1\} & & \end{array} \tag{3.55}$$

*Where the isomorphism is a natural isomorphism between functors (see Definition 2.3.29).*

Some common invariants include:

- The rank function [39] on every posetal subcategory, which is called the rank invariant:

$$\text{rank}(F(M_a \to M_b)), a, b \in \text{ob}(Q(\mathcal{D})) \tag{3.56}$$

- The Hilbert function

$$\text{rank}(F(M_a)), a \in \text{ob}(Q(\mathcal{D})) \tag{3.57}$$

- Hilbert polynomial and Hilbert Series [41]

- Generalized rank invariant [42, 43]

A complete invariant for functor $F|_{Q(\mathcal{D})}$ is defined as an invariant $f_{\text{Thin}(\mathbf{Vec})} : 2^{\text{Thin}(\mathbf{Vec})} \to \{0,1\}$ which has the property that

$$f(\text{im}(F|_{Q(\mathcal{D})})) \neq f(\text{im}(G|_{Q(\mathcal{D})})) \text{ if } \text{im}(F|_{Q(\mathcal{D})}) \ncong \text{im}(G|_{Q(\mathcal{D})}) \tag{3.58}$$

**Problem 3.1.1.** *(Complete invariant posetal quiver representation problem)*

*Given a quiver representation $F : \mathcal{D} \to \mathbf{Vec}$ and posetal subcategory $Q(\mathcal{D}) \subseteq \mathcal{D}$, find a complete invariant $f_{Thin(\mathbf{Vec})} : 2^{Thin(\mathbf{Vec})} \to \{0,1\}$ for $F|_{Q(\mathcal{D})}$.*

A complete invariant map for the functor $F|_{Q(\mathcal{D})}$ is defined as a map $f : 2^{Thin(\mathbf{Vec})} \to \mathcal{T}$ which has the property that

$$f(\text{im}(F|_{Q(\mathcal{D})})) \cong f(\text{im}(G|_{Q(\mathcal{D})})) \text{ iff } \text{im}(F|_{Q(\mathcal{D})}) \cong \text{im}(G|_{Q(\mathcal{D})}) \tag{3.59}$$

for some target category $\mathcal{T}$.

(3.60) Let $\mathcal{D}$ be the locally small subcategory of **Set** with:

1. Objects consisting of finite sets.
2. Morphisms consisting of inclusion maps between finite sets.

Let us define the following 3 object subcategory $Q(\mathcal{D})$ of $\mathcal{D}$:

1. With the class of objects forming the following set: $\{\{a\},\{b\},\{a,b\}\}$.
2. And morphisms being the two inclusions $\text{inc} : \{a\} \hookrightarrow \{a,b\}$, $\text{inc} : \{b\} \hookrightarrow \{a,b\}$.

We can write $Q(\mathcal{D}) = (\{\{a\},\{b\},\{a,b\}\}, \subseteq)$.

We show that in the simple case of just the two morphisms $f_1, f_2 \in \text{mor}(\text{im}(F|_{\mathbf{Q}(\mathcal{D})}))$, for some quiver representation $F|_{Q(\mathcal{D})}$ when there exists a merging of the two maps, meaning that their codomains are identical, then the rank invariant on $F|_{Q(\mathcal{D})}$ may not be a complete invariant.

Thus, letting $F|_{Q(\mathcal{D})}\colon (\{\{a\},\{b\},\{a,b\}\},\subseteq) \to \mathbf{Vec}$ be a covariant functor on $Q(\mathcal{D})$ to **Vec**, forming a posetal category $\mathrm{im}(F|_{Q(\mathcal{D})})$:

$$\begin{array}{ccc} & \mathbb{R}^2 & \\ {\scriptstyle 1\mapsto(1,0)}\nearrow & & \nwarrow{\scriptstyle 1\mapsto(1,0)} \\ \mathbb{R} & & \mathbb{R} \end{array} \tag{3.61}$$

This is not isomorphic to the image of a similar functor:
$G|_{Q(\mathcal{D})}\colon (\{a,b,\{a,b\}\},\subseteq) \to \mathbf{Vec}$

$$\begin{array}{ccc} & \mathbb{R}^2 & \\ {\scriptstyle 1\mapsto(1,0)}\nearrow & & \nwarrow{\scriptstyle 1\mapsto(0,1)} \\ \mathbb{R} & & \mathbb{R} \end{array} \tag{3.62}$$

by a natural isomorphism, however the rank invariant cannot distinguish these two functors.

## 3.2 Persistent Homology

Persistence theory was originally formulated in terms of the homology functor over a finite filtered simplicial complex. This can be denoted by:

(3.63) The functor $(H_\bullet : \mathbf{Simp} \to \mathbf{Vec}, Q(\mathbf{Simp}))$ from Section 2.6.8 where the finite simple acyclic subcategory $Q(\mathbf{Simp}) = (V, E)$ has:

$$V \triangleq \{K_\mathrm{i}\}_{\mathrm{i}=0}^{n}, E \triangleq \{\mathrm{inc}_{\mathrm{i,j}} : K_\mathrm{i} \hookrightarrow K_\mathrm{j}\}_{\mathrm{i,j}\in\{0,\ldots,n\}:\mathrm{i}\leq\mathrm{j}} \tag{3.64}$$

This is called a **filtration** on simplicial complexes and denoted $K_0 \subseteq K_1 \subseteq \ldots \subseteq K_n$.

If $K_0 \triangleq \varnothing$ and $K_{\mathrm{i}+1} \triangleq K_\mathrm{i} \cup \{\sigma_{\mathrm{i}+1}\}, \sigma_{\mathrm{i}+1}$ a new simplex so that $K_{\mathrm{i}+1}$ is still an abstract simplicial complex. We then call this finite simple acyclic subcategory as a **simplex-wise** filtration. For a simplex-wise filtration, for any two simplices $\sigma_\mathrm{j}, \sigma_\mathrm{i} \in K_n$ we can define the total order:

$$\sigma_\mathrm{j} < \sigma_\mathrm{i} \text{ iff } \mathrm{j} < \mathrm{i} \tag{3.65}$$

**Fact 3.2.1.** *If the filtration is simplex-wise then the* $H_\bullet : \boldsymbol{Simp} \to \boldsymbol{Vec}$ *functor restricted to the subcategory* $Q(\boldsymbol{Simp}) \subseteq \boldsymbol{Simp}$*, denoted*

$$H_\bullet \mid_{Q(\boldsymbol{Simp})}: Q(\boldsymbol{Simp}) \to \boldsymbol{Vec}, \tag{3.66}$$

*is faithful.*

*Proof.* The inclusion map $\mathrm{inc}_{\mathrm{i,i+1}} : K_{\mathrm{i}} \hookrightarrow K_{\mathrm{i+1}}$ is the only map that can exist between $K_{\mathrm{i}}$ and $K_{\mathrm{i+1}}$ in $Q(\mathbf{Simp})$ so $\hom(H_\bullet(K_{\mathrm{i}}), H_\bullet(K_{\mathrm{i+1}}))) : \mathbf{Simp} \to \mathbf{Vec}$ can only have a single morphism. Thus, the functor $H_\bullet$ is faithful. □

Such a finite simple acyclic subcategory satisfying Equation 3.64 is also known as a filtration of simplicial complexes.

**The Spanning Set of each Homology Vector Space:**
Following the framework of the "Question of Persistence across Vector Spaces," we implicitly know the spanning set of generators for each homology vector space. According to the discussion of the example on persistent homology of Example 3.19 at most a single new generator is added to an accumulating set of generators that span each vector space $H_\bullet(K_{\mathrm{i}})$. This is due to the simplex-wise assumption. These generators come from a common set $\mathrm{Forget}(W)$, where $W$ is a $n$-dimensional vector space. We also assume that the map $\phi_{\mathrm{i,i+1}}$ induced by inclusion restricts to an inclusion between the generators from $\mathrm{Forget}(W)$.
By identifying the space $W$, the generators can be distinguished.

This gives us the downstairs finite simple acyclic subcategory

$$\mathrm{im}(H_\bullet \mid_{Q(\mathbf{Simp})}) = (\{H_\bullet(K_{\mathrm{i}})\}_{\mathrm{i}=0}^{n}, \{\phi_{\mathrm{i,j}} : H_\bullet(K_{\mathrm{i}}) \to H_\bullet(K_{\mathrm{j}})\}_{\mathrm{i,j} \in \{0,\dots,n\} : \mathrm{i} \le \mathrm{j}}) \tag{3.67}$$

where $\phi_{\mathrm{i,j}} : H_\bullet(K_{\mathrm{i}}) \to H_\bullet(K_{\mathrm{j}})$ is induced by inclusion from the commuting diagram:

$$\begin{array}{ccc} K_{\mathrm{i}} & \xhookrightarrow{\mathrm{inc}_{\mathrm{i,j}}} & K_{\mathrm{j}} \\ \downarrow & & \downarrow \\ H_\bullet(K_{\mathrm{i}}) & \xrightarrow{\phi_{\mathrm{i,j}}} & H_\bullet(K_{\mathrm{j}}) \end{array} \tag{3.68}$$

According to [44], the finite simple acyclic subcategory $\mathrm{im}(H_\bullet |_{Q(\mathbf{Simp})})$ is a more general instance of a persistence module, which is defined below:

**Definition 3.2.2.** *([44]) A persistence module over a poset $(A, \leq)$ is any collection $V = \{V_a\}_{a \in A}$ of vector spaces $V_a$ together with linear maps $v_{a,a} : V_a \to V_{a'}$ so that $v_{a,a} = \mathrm{id}$ and $v_{a',a''} \circ v_{a,a'} = v_{a,a''}$ for all $a, a', a'' \in A$ where $a \leq a' \leq a''$. Sometimes we write $V = \{V_a \xrightarrow{v_{a,a'}} V_{a'}\}_{a \leq a'}$ to denote this collection with the maps.*

The definition in [44] requires the vector spaces and linear maps follow the structure of a poset. A generic quiver representation, however would not require this.

**Discussion: Is the Poset Assumption Necessary?**

We do not believe the poset assumption is necessary if the goal is to measure the **persistence** of individual vectors across a composition of maps. The posetal symmetry that is imposed by the poset $(A, \leq)$ on the persistence module $V$ makes maps only indexable by pairs of elements from the poset. This prevents distinguishing different sequences of composable maps with the same source and target.

A finite simple DAG-like structure, on the other hand, can express a sequence of finitely many composable maps. It can also be totally ordered by a topological sort, allowing for a global total order over all sequences of composable maps. This allows us to unify the notion of persistence across all possible sequences of composable maps across the entire DAG.

Two persistence modules are considered equivalent when there is an isomorphism between them. This is defined as follows:

**Definition 3.2.3.** *([44]) We say two persistence modules* $U = \{U_a \xrightarrow{u_{a,a'}} U_{a'}\}_{a \le a', a, a' \in A}$ *and* $V = \{V_a \xrightarrow{v_{a,a'}} V_{a'}\}_{a \le a', a, a' \in A}$ *indexed over an index set* $A$ *are **isomorphic**, denoted* $U \cong V$*, if the following two conditions hold:*

1. $U_a \cong V_a$ *for every* $a \in A$ *as vector spaces, and*

2. *for every* $x \in U_a$*, if* $x$ *is mapped to* $y \in V_a$ *by the vector space isomorphism, then* $u_{a,a'}(x) \in U_{a'}$ *is mapped to* $v_{a,a'}(y) \in V_{a'}$ *also by the isomorphism.*

We define the summation of two persistence modules $M_1, M_2$, both indexed by a poset $(A, \le)$ as follows:

$$(M_1 \oplus M_2)_a = (M_1)_a \oplus (M_2)_a, \forall a \in A \tag{3.69}$$

It will be helpful to define an algebraic unit for persistence modules. We call these persistence modules as interval modules.

**Definition 3.2.4.** *([44]) Given a partially ordered index set* $(A, \le)$ *and a pair of indices* $b, d \in A$*,* $b \le d$*, an interval module denoted* $I[b, d) = \{V_a \xrightarrow{v_{a,a'}} V_{a'}\}_{a, a' \in A}$ *is a persistence module where:*

- $V_a = \mathbb{R}$ *for all* $a \in [b, d) \cap A$ *and* $V_a = \boldsymbol{0}$ *otherwise,*

- $v_{a,a'}$ *is identity map for* $b \le a \le a' < d$ *and zero map otherwise*

These persistence modules can have the indecomposability property, meaning that they cannot be decomposed into summands any further. This is a desired property of an interval since it would mean that it cannot be shrunk any further without introducing a decomposition into two interval modules.

**Definition 3.2.5.** *We say a persistence module* $M$ *is indecomposable if* $M \cong M_1 \oplus M_2 \Rightarrow M_1 = \boldsymbol{0}$ *or* $M_2 = \boldsymbol{0}$ *where* $\boldsymbol{0}$ *denotes the trivial vector space of* $\{\overrightarrow{0}\}$*.*

According to Gabriel's theorem [29], if persistence module $V$ is totally ordered by a finite totally ordered poset $A$ and $\dim(V_a) < \infty$, then there exists a decomposition of any persistence module $V$ into indecomposable interval persistence modules $I[b, d)$ up to isomorphism, this can be written as:

$$V \cong \bigoplus_{b,d \in A \setminus B : b < d} I[b, d) \oplus \bigoplus_{b \in B : B \subseteq A} I[b, \infty) \tag{3.70}$$

Gabriel's theorem also applies to zigzag posets, which are posets $(A, \leq)$ where there is a bijection from $A$ to a set $[n], n \in \mathbb{N}$ and $a_i \leq a_{i+1}$ or $a_i \geq a_{i+1}$. When a persistence module is ordered by a zigzag poset, then it is also said to be a simply laced Dynkin diagram of type $A_n$ [45, 46].

The decomposition from Gabriel's theorem is unique up to a permutation of the intervals. This follows by Krull-Schmidt [47].

**Remark 3.2.6.** *Notice that Gabriel's theorem is in direct analogy to the Chinese Remainder Theorem [11], which in general form is called the "Structure Theorem for Finitely Generated Modules over a Principal Ideal Domain" [16]. In particular, indecomposables are "primes" that can be factored out by linear algebraic operations. See Theorem 6.0.2 for a proof.*

For the example of persistent homology:

**Observation 3.2.7.** *Given a pair $(H_\bullet : \boldsymbol{Simp} \to \boldsymbol{Vec}, Q(\boldsymbol{Simp}))$ where $Q(\boldsymbol{Simp}) = (V, E)$ is a simplex-wise filtration*

$$V \triangleq \{K_i\}_{i=0}^n, E \triangleq \{inc_{i,j} : K_i \hookrightarrow K_j\}_{i,j \in \{0,\ldots,n\}: i \leq j} \tag{3.71}$$

*with a total order on the simplices of $K_n$.*

*We can form a persistence module $V$ from:*

$$im(H_\bullet |_{Q(\boldsymbol{Simp})}) \tag{3.72}$$

*by simply taking*

$$V \triangleq \{H_\bullet(K_i) \xrightarrow{v_{i,j}} H_\bullet(K_j)\}_{i,j \in \{0,\ldots,n\}: i \leq j} \tag{3.73}$$

*as the persistence module.*

**Observation 3.2.8.** *Using the decomposition of Gabriel's theorem of $V$ from Observation 3.2.7, there exists a set of indecomposable interval modules $I[b, d)$ of Equation 3.70 for V with:*

$$V \cong \bigoplus_{b,d \in Q(\boldsymbol{Simp}) \setminus G : b < d} I[b, d) \oplus \bigoplus_{b \in G : G \subseteq Q(\boldsymbol{Simp})} I[b, \infty) \tag{3.74}$$

*These interval modules form a sequence of arrow compositions*

$$id_{b,c_n} = (\mathrm{id}_{c_{n-1},c_n} \circ \cdots \circ id_{b,c_1}), c_n < d \tag{3.75}$$

*where a cycle generator* $\mathrm{e} \in H_\bullet(K_b)$ *is linearly independent in all of* $H_\bullet(K_a), \forall a : b \leq a < d$.

From Observation 3.2.7 we see that for persistent homology, denoted $(H_\bullet : \mathbf{Simp} \to \mathbf{Vec}, Q(\mathbf{Simp}))$ where $Q(\mathbf{Simp})$ is filtration, Gabriel's theorem answers our original question of "Persistence across Vector Spaces," Question 3.1.1, about persistence in the downstairs category, which for persistent homology is **Vec**.

The converse to Observation 3.2.7 only holds if the homology vector space is viewed up to isomorphism as the span of a particularly chosen **basis** at each index i. If a basis is not chosen, then when the rank drops by 1, it is ambiguous as to which generators become "dead". Of course, when generators merge no generator is actually deleted. The generators are only merged into a common coset.

Explicitly, when the rank drops by exactly 1, our homology vector space is now of the form:

$$\frac{\ker(\partial_p^{K_\mathrm{i}})}{\mathrm{im}(\partial_{p+1}^{K_\mathrm{i}})} \cong \frac{\ker(\partial_p^{K_\mathrm{i}})}{\mathrm{span}(\mathrm{im}(\partial_{p+1}^{K_{\mathrm{i}-1}}) \cup \{\partial_{p+1}(\sigma_\mathrm{i})\})} \tag{3.76}$$

However, we can define a rule across all indices i each time the rank drops by exactly 1 by exactly one relation (an independent boundary vector $\partial_{p+1}(\sigma_\mathrm{i})$ is introduced). This rule will determine a **basis** at an index i that depends on the previous generators and relations. With this **basis**, the intervals from Gabriel's theorem coincide with generators that persist over time in the persistence framework of Question 3.1.1.

We can summarize the following connection between the intervals of Gabriel's theorem and the general framework of Question 3.1.1:

**Theorem 3.2.9.** *In the general framework of Question 3.1.1: "The Question of Persistence across Vector Spaces,"*

1. *The intervals from Gabriel's theorem can be simulated by replacing the accumulating set of spanning generators from Example 3.63 with some* ***basis*** *for* $H_\bullet(K_\mathrm{i})$ *from* $Forget(W)$ *at each index* i.

2. *This redefines each linear map $\phi_{\mathrm{i},\mathrm{i}+1}$ induced by inclusion to a sum of identity maps of independent one-dimensional subspaces.*

*Proof.* Let the interval decomposition of the persistence module $V$ from $\mathrm{im}(H_\bullet |_{Q(\mathbf{Simp})})$ be:

$$V \cong \bigoplus_{b,d \in Q(\mathbf{Simp}) \setminus G : b<d} I[b,d) \oplus \bigoplus_{b \in G : G \subseteq Q(\mathbf{Simp})} I[b,\infty) \tag{3.77}$$

For each interval $I[b_\mathrm{j}, d_\mathrm{j})$, at each index $a_\mathrm{j} \in [n] : b_\mathrm{j} \le a_\mathrm{j} < d_\mathrm{j}$, let there be a basis vector

$$\mathrm{e}_{[b_\mathrm{j},d_\mathrm{j}),\mathrm{j};a_\mathrm{j}} \in \mathrm{Forget}(W) \tag{3.78}$$

Letting:

$$M_{a_\mathrm{j}} \triangleq \mathrm{span}(\{\mathrm{e}_{[b_\mathrm{j},d_\mathrm{j});\mathrm{j},a_\mathrm{j}} : \exists I[b_\mathrm{j}, d_\mathrm{j}) \subseteq V, a_\mathrm{j} \in [b_\mathrm{j}, d_\mathrm{j})\}) \tag{3.79}$$

Due to the interval decomposition, we must have that:

$$H_\bullet(K_\mathrm{i}) \cong M_\mathrm{i} \tag{3.80}$$

We also have that for any interval $I[b_\mathrm{j}, d_\mathrm{j})$, the spanning set for $a_\mathrm{j} \in [b_\mathrm{j}, d_\mathrm{j})$:

$$\{\mathrm{e}_{[b_\mathrm{j},d_\mathrm{j});\mathrm{j},a_\mathrm{j}} : \exists I[b_\mathrm{j}, d_\mathrm{j}) \subseteq V, a_\mathrm{j} \in [b_\mathrm{j}, d_\mathrm{j})\} \tag{3.81}$$

is a **basis**. This comes from the direct sum.

The maps $\phi_{a_\mathrm{j},a_\mathrm{j}+1} : H_\bullet(K_{a_\mathrm{j}}) \to H_\bullet(K_{a_\mathrm{j}+1})$ induced by inclusion are now $\psi_{a_\mathrm{j},a_\mathrm{j}+1} : M_{a_\mathrm{j}} \to M_{a_\mathrm{j}+1}$.

$$\psi_{a_\mathrm{j},a_\mathrm{j}+1} \triangleq \bigoplus_{a_\mathrm{j} : a_\mathrm{j} \in I[b_\mathrm{j},d_\mathrm{j}), b_\mathrm{j} \le a_\mathrm{j} < a_\mathrm{j}+1 < d_\mathrm{j}} id\, |_{\mathrm{span}(\{\mathrm{e}_{[b_\mathrm{j},d_\mathrm{j});\mathrm{j},a_\mathrm{j}}\})} \oplus \bigoplus_{a_\mathrm{j} : a_\mathrm{j} \in I[b_\mathrm{j},\infty), b_\mathrm{j} \le a_\mathrm{j}} id\, |_{\mathrm{span}(\{\mathrm{e}_{[b_\mathrm{j},\infty);\mathrm{j},a_\mathrm{j}}\})} \tag{3.82}$$

□

**Remark 3.2.10.** *According to Theorem 3.2.9, it must be that the general framework of Question 3.1.1 encompasses the interval decomposition.*

*This is because Gabriel's theorem imposes the requirement of an indecomposable decomposition. This algebraic definition is independent of the original data.*

*Our general framework, however, respects the difference between various generators chosen from the set Forget($W$) of the embedding space $\mathcal{W}$.*

*This means that the geometry of the original data is not lost if $\mathcal{W}$ is chosen to respect this. In fact, a basis that respects the geometry of $\mathcal{W}$ could be chosen to change the death times of the generators.*

In the following we give a simple example illustrating an interval decomposition.

(3.83) [A Simple Illustration of an Interval Decomposition]

$$\begin{array}{ccccccc}
K_1 = \{\bullet_1\} & \hookrightarrow & K_2 = \{\bullet_1, \bullet_2\} & \hookrightarrow & K_3 = \{\bullet_1, \bullet_2, \bullet_1 - \bullet_2\} & \hookrightarrow & K_4 = \{\bullet_1, \bullet_2, \bullet_1 - \bullet_2\} \\
\downarrow & & \downarrow & & \downarrow & & \downarrow \\
H_0(K_1) & \longrightarrow & H_0(K_2) & \longrightarrow & H_0(K_3) & \longrightarrow & H_0(K_4) \\
\cong & & \cong & & \cong & & \cong \\
\{\vec{0}, \bullet_1\} & \underset{\text{rnk=1}}{\longrightarrow} & \{\vec{0}, \bullet_1, \bullet_2\} & \underset{\text{rnk=1}}{\longrightarrow} & \color{red}{\{\vec{0}, \bullet_1 + (\bullet_1 + \bullet_2)\}} & \underset{\text{rnk=1}}{\longrightarrow} & \{\vec{0}, \bullet_1 + (\bullet_1 + \bullet_2)\}
\end{array}$$

$$1 \longrightarrow_{1} \longrightarrow 0$$

$$1 \longrightarrow_{1} \longrightarrow 1$$

**Figure 3.3.** The indecomposable intervals of the persistence module $\{H_0(K_{\mathrm{i}}) \xrightarrow{\phi_{\mathrm{i,j}}} H_0(K_{\mathrm{j}})\}_{\mathrm{i,j}:1 \le \mathrm{i} \le \mathrm{j} \le 4}$

In the commuting diagram of Figure 3.3, there are two independent generators $\mathrm{e}_1, \mathrm{e}_2$ at times 1 and 2 that merge at time 3 via the relation $(\mathrm{e}_1 + \mathrm{e}_2)$ and continue to time 4. Both generators last till time 3 but $\mathrm{e}_2$ is the latest of the generators in the relation $(\mathrm{e}_1 + \mathrm{e}_2)$.

We can check that the indecomposables are $I[2,3), I[1,4)$ since $I[2,3)$ is shortest for generator $\mathrm{e}_2$, making it indecomposable.

### 3.2.11 Matrix Reduction for Computation

Given $Q(\mathbf{Simp}) = (V = \{K_{\mathrm{i}}\}_{\mathrm{i}=0}^{n}, E = \{\mathrm{inc}_{\mathrm{i,j}} : K_{\mathrm{i}} \hookrightarrow K_{\mathrm{j}}\}_{\mathrm{i,j} \in \{0,\ldots,n\}:\mathrm{i} \le \mathrm{j}})$, as a simplex-wise filtration of simplicial complexes $\varnothing = K_0 \subseteq K_1 \subseteq \cdots \subseteq K_n$. In order to compute the interval modules for the induced persistence module for the simplex-wise filtration, we can devise the following "standard algorithm" for persistent homology.

In the order of the filtration $\varnothing \subseteq K_1 \subseteq \cdots \subseteq K_n$, for each dimension $p \geq 0$ and for each simplex $\sigma_{\mathrm{i}} \in K_{\mathrm{i}} \setminus K_{\mathrm{i}-1}$, $K_{\mathrm{i}} \setminus K_{\mathrm{i}-1} \subseteq C_{p+1}(K_n)$, we compute the boundary of $\sigma_{\mathrm{i}}$ as defined in Equation 2.6.7 viewed as a column vector in $\mathbb{R}^n$: $[\partial_{p+1}(\sigma_{\mathrm{i}})] \in \mathbb{R}^n$.

Then, while respecting the causality of the filtration, we check for the independence of $\partial_{p+1}(\sigma_{\mathrm{i}})$ from the subspace $\mathrm{span}(\{\partial_{p+1}(\sigma_{\mathrm{j}})\}_{\mathrm{j}=1}^{\mathrm{i}-1})$. The intention is to compute the intervals as determined by Gabriel's theorem. According to Theorem 3.2.9, we know that these intervals are an answer to "The Question of Persistence across Vector Spaces" in the general framework of Question 3.1.1.

The independence check can easily be accomplished by vector addition from left to right onto column $[\partial_{p+1}(\sigma_{\mathrm{i}})]$ over the partial boundary matrix $[[\partial_{p+1}(\sigma_1)]\|\cdots\|[\partial_{p+1}(\sigma_{\mathrm{i}})]]$ (the concatentation of the first i vectors: $\{[\partial_{p+1}(\sigma_{\mathrm{j}})]\}_{\mathrm{j}=1}^{\mathrm{i}}\}$, viewed as column vectors in $\mathbb{R}^n$, where $n$ is the number of simplicial complexes in the filtration.

The independence test of left to right column addition must respect the arrows provided by the filtration. In matrix terms, for any matrix $M$ these columns are defined as **causal past** columns for column $M[\mathrm{i}]$ of $M$ that respect the nonzeros of column $M[\mathrm{i}]$:

**Definition 3.2.12.** *A column* j *of matrix* $M$ *is a* ***causal past column*** *for column* $M[\mathrm{i}]$ *if* $\mathrm{j} < \mathrm{i}$ *and all nonzero entries of column* j *have index less than or equal to the largest index of a nonzero of column* $M[\mathrm{i}]$*.*

*Similarly, a column* $k$ *of matrix* $M$ *is a* ***causal future column*** *for column* $M[\mathrm{i}]$ *if* $k > \mathrm{i}$ *and all nonzero entries of column* $k$ *have row index greater than or equal to the smallest index of a nonzero of column* $M[\mathrm{i}]$*.*

(3.84) In terms of the $[\partial_{p+1}]$ boundary matrix, this means for column $[\partial_{p+1}(\sigma_{\mathrm{i}})]$ we only add the **causal past** columns $[\partial_{p+1}(\sigma_{\mathrm{j}})] = \sum_{k=1}^{p+1}(-1)^k[(\sigma_{\mathrm{j}})_{-k}]$ for $\mathrm{j} < \mathrm{i}$ that have all summands $[\sigma_l] = [(\sigma_{\mathrm{j}})_{-k}]$ having the property that $l \leq \max_{\sigma_m \in \partial_{p+1}(\sigma_{\mathrm{i}})} m$.

We can then form a submatrix of $M$ involving all of the causal past columns of a column $M[\mathrm{i}]$. We call this concatenation of columns the **causal past submatrix**.

**Definition 3.2.13.** *For the column $M[\mathrm{i}]$ of a matrix $M$, the* ***causal past submatrix*** *of column $M[\mathrm{i}]$, denoted $M_{<M[\mathrm{i}]}$, is defined as the concatenation of all of its causal past columns.*

*Similarly, the* ***causal future submatrix*** *of column* i*, denoted $M_{>M[\mathrm{i}]}$, is defined as the concatenation of all of its causal future columns.*

(3.85) As a concrete example, consider the simplicial complex $K_{12} \triangleq (\{v_{\mathrm{i}}\}_{\mathrm{i}=1}^{6}, \{\sigma_{\mathrm{i}}\}_{\mathrm{i}=1}^{12})$ where the simplices $\sigma_{\mathrm{i}} = v_{\mathrm{i}}, \mathrm{i} = 1, ..., 6$ and

$\sigma_7 = \{v_1, v_7\}, \sigma_8 = \{v_1, v_4\}, \sigma_9 = \{v_1, v_2\}, \sigma_{10} = \{v_1, v_3\}, \sigma_{11} = \{v_2, v_3\}, \sigma_{12} = \{v_1, v_2\}$,

which forms a filtration $\varnothing \subseteq K_1 \subseteq \cdots \subseteq K_{12}$ by $\{\sigma_{\mathrm{i}+1}\} = K_{\mathrm{i}+1} \smallsetminus K_{\mathrm{i}}, \mathrm{i} = 0, ..., 11$.

For $p = 0$, we can visualize the following discretized $6 \times 6$ matrix

$[\partial_1] = [[\partial_1(\sigma_7)] \| \cdots \| [\partial_1(\sigma_{12})]] \in \mathbb{R}^{6\times 6}$, (where the integer index is for time):

$$[\partial_1] = \begin{pmatrix} -[(\sigma_7)_{-1}] & -[(\sigma_8)_{-1}] & -[(\sigma_9)_{-1}] & -[(\sigma_{10})_{-1}] & 0 & -[(\sigma_{12})_{-1}] \\ 0 & 0 & [(\sigma_9)_{-2}] & 0 & -[(\sigma_{11})_{-1}] & [(\sigma_{12})_{-2}] \\ 0 & 0 & 0 & [(\sigma_{10})_{-2}] & [(\sigma_{11})_{-2}] & 0 \\ 0 & [(\sigma_8)_{-2}] & 0 & 0 & 0 & 0 \\ 0 & 0 & 0 & 0 & 0 & 0 \\ [(\sigma_7)_{-2}] & 0 & 0 & 0 & 0 & 0 \end{pmatrix}$$

where the notation $[(\sigma_{\mathrm{i}})_{-\mathrm{j}}]$ denotes the number $1 \in \mathbb{R}$, which indicates the j-th $p$-dimensional face $(\sigma_{\mathrm{i}})_{-\mathrm{j}} \subseteq \sigma_{\mathrm{i}}$ of the dimension $p+1$ simplex $\sigma_{\mathrm{i}}$.

We see in the matrix $[\partial_1]$ in Equation 2.6.7 that the blue submatrix is the causal "past" submatrix whose columns allow for feasible column addition to the red column representation of $\partial(\sigma_{10})$. The green column $[\partial(\sigma_{12})]$ is in the future of the red column and thus relative to column $[\partial(\sigma_{11})]$, this column will never add to column $[\partial(\sigma_{11})]$.

We notice the following property on the matrix:

**Observation 3.2.14.** *For a simplex-wise filtration $\varnothing \subseteq K_1 \subseteq ... \subseteq K_n$,*

*At index* i*, $\sigma_{\mathrm{i}} \in C_{p+1}(K_{\mathrm{i}})$ is either creating a linearly independent homological cycle at index* i *or is a linear combination of $p+1$ dimensional boundaries from the causal past submatrix.*

*Proof.* At each index i there is one of either a creation or destruction event since each index i corresponds to a single rank 1 column that either belongs to $\mathrm{im}(\partial_{p+1})$ (destruction event) or to $\ker(\partial_{p+1})$ (creation event) by the rank-nullity theorem [48].

Since every destruction event at index i has amongst all creation events a unique maximal index j and since the destruction event cannot coincide with a creation event, by symmetry this destruction event at index i is the latest destruction event for creation event at index j. $\square$

To compute the interval modules of Gabriel's theorem for the representation of the homology functor on a filtration:

$$(H_\bullet : \mathbf{Simp} \to \mathbf{Vec}, Q(\mathbf{Simp}) = (V = \{K_\mathrm{i}\}_{\mathrm{i}=0}^n, E = \{\mathrm{inc}_{\mathrm{i},\mathrm{j}} : K_\mathrm{i} \hookrightarrow K_\mathrm{j}\}_{\mathrm{i},\mathrm{j} \in \{0,\dots,n\} : \mathrm{i} \leq \mathrm{j}}) \tag{3.86}$$

we ask "The Question of Persistence across Vector Spaces":

**Question 3.2.1.** *For any $\partial_{\bullet+1}(\sigma_\mathrm{i}) \in C_\bullet(K_n), 1 \leq \mathrm{i} \leq n$, does this introduce a linearly independent generator for $H_{\bullet+1}(K_\mathrm{i})$.*

*If not, then for what times $\mathrm{j} : \mathrm{j} < \mathrm{i}$ when the induced image of a homology generator from $H_\bullet(K_\mathrm{j})$ whose image in $H_\bullet(K_\mathrm{i})$ is linearly dependent to $\partial_{\bullet+1}(\sigma_\mathrm{i})$ in $H_\bullet(K_\mathrm{i})$?*

These two questions are mutually exclusive by the Observation 3.2.14

We can translate these two questions into:

- ($\bullet + 1$-dimensional creation event at index i):

  Is there a set of $\bullet + 1$-dimensional simplices $S_{<\mathrm{i}} \subseteq K_\mathrm{i}$ before index i so that

$$\mathrm{span}(\{\partial_{\bullet+1}(\sigma_\mathrm{i})\} \cup \{\partial_{\bullet+1}(\sigma_\mathrm{j})\}_{\sigma_\mathrm{j} \in S_{<\mathrm{i}}}) \subseteq \ker(\partial_{\bullet+1}(K_\mathrm{i})) \tag{3.87}$$

- ($\bullet + 1$ dimensional destruction event at index i for some $\bullet$-dimensional creation event at index j):

  If not, then there is a creation event at index j whose generator $\sigma_\mathrm{j}$ from $H_\bullet(K_\mathrm{j})$ becomes linearly dependent to $\partial_{\bullet+1}(\sigma_\mathrm{i})$.

We first make the following observation:

**Observation 3.2.15.** *The index* j *of the last* $\bullet$*-dimensional simplex* $\sigma_{\mathrm{j}} \in \partial_{\bullet+1}(\sigma_{\mathrm{i}})$ *always indexes a creation event.*

*Conversely, all creation events occur at the index of the latest simplex* $\sigma_{\mathrm{j}}$ *of the boundary* $\partial_{\bullet+1}(\sigma_{\mathrm{i}})$.

*Proof.* **For the first part,**

Each boundary $\partial_{\bullet+1}(\sigma_{\mathrm{i}})$ is a linear combination of $\bullet$-dimensional faces of a $\bullet$+1-dimensional simplex $\sigma_{\mathrm{i}}$. This is by definition of a simplicial complex. The linear combination of $\bullet$-dimensional faces itself is an independent generator of $H_{\bullet}(K_{\mathrm{i}})$. Thus there is an injection from the set of $\bullet$-dimensional boundaries to the set of all creation events.

**For the second part,**

Every $\bullet$-dimensional creation event at index i for $H_{\bullet}(K_{\mathrm{i}})$ comes from a linear combination of $\bullet$-dimensional boundaries with the index of its latest boundary at index i being its creation event index.

By 1 and 2, all creation events occur at exactly each of the latest simplices of each boundary.

□

For a given destruction event, we would like to make the creation times pair with the destruction times in a well-defined manner, meaning that there is a matching, or one to one correspondence between creation and destruction events.

The creation events occur at particular indices according to Observation 3.2.15. Based on this, we can define the **elder-rule** [49] to provide a matching:

> At index i, the elder-rule finds the earliest of the **latest** $\bullet$-dimensional generators that can connect through a linear combination of boundaries $\partial_{\bullet+1}(\sigma_{\mathrm{j}}), \sigma_{\mathrm{j}} \in S_{<\mathrm{i}}$.

(3.88) Looking back at Example 3.83, we can see that the **last** generator $e_2$, corresponding to point $\bullet_2$ creates the connected component at index 3 and thus is the generator destroyed at index 4.

For the matrix $[\partial^{K_n}_{\bullet+1}]$, we can perform column operations with the columns of the causal past submatrix for $[\partial_{\bullet+1}(\sigma_{\mathrm{i}})]$ to $[\partial_{\bullet+1}(\sigma_{\mathrm{i}})]$ until independence is achieved.

- If $[\partial_{\bullet+1}(\sigma_{\mathrm{i}})]$ is linearly dependent with some columns from the causal past submatrix then there is a $\bullet+1$-dimensional creation event at index i.

- Otherwise, there is a linear combination of $[\partial_{\bullet+1}(\sigma_{\mathrm{i}})]$ and columns from the causal past submatrix which minimizes the maximal nonzero row index of these columns. Let this minimizing row index be $\mathrm{j}^*$.

  We say that there is a $\bullet+1$-dimensional destruction event at index i and it is **paired with** the $\bullet$-dimensional creation event at index $\mathrm{j}^*$.

At index i a creation or a destruction event occurs exclusively since linear independence and dependence are exclusive by Observation 3.2.14.

It will be convenient to define the following function on a column $c$:

$$\mathrm{low}(c) \triangleq \begin{cases} \max_{c[\mathrm{j}]\neq 0} \mathrm{j} & c \neq 0 \\ 0 & c = 0 \end{cases} \tag{3.89}$$

This function $\mathrm{low}(c)$ defines the maximum nonzero index of a column $c$. When $c = 0$, then $\mathrm{low}(c)$ is 0.

The two questions of checking for linear independence can be expressed as the following problem called the matrix reduction problem for persistent homology:

**Problem 3.2.1.** *[Primal Matrix Reduction Problem of Persistent Homology]*

*At index* i, *compute:*

$$\begin{gathered} \min_{[S_{<\mathrm{i}}]\subseteq[\partial_{\bullet+1}]_{<[\partial_{\bullet+1}(\sigma_{\mathrm{i}})]};c_v\in\mathbb{R}\setminus\{0\},\forall v\in col(S_{<\mathrm{i}})} low\Big(\sum_{v\in col(S_{<\mathrm{i}})} c_v v + [\partial_{\bullet+1}(\sigma_{\mathrm{i}})]\Big)\Big) \\ \textit{where } [S_{<\mathrm{i}}] \textit{ is a causal submatrix of } [\partial_{\bullet+1}]_{<[\partial_{\bullet+1}(\sigma_{\mathrm{i}})]} \end{gathered} \tag{3.90}$$

There are many feasible submatrix solutions $[S^*_{<\mathrm{i}}]$ to this optimization problem. The sequence of columns: $([\partial_{\bullet+1}(\sigma_{\mathrm{j}_k})])_k$ from the submatrix $[\partial_{\bullet+1}]_{<[\partial_{p+1}(\sigma_{\mathrm{i}})]}$ form a solution submatrix $[S^{**}_{<\mathrm{i}}]$ by concatenation.

This sequence of columns can be defined inductively as follows:

$$s_0 \triangleq [\partial_{\bullet+1}(\sigma_{\mathrm{i}})], r_0 \triangleq [\partial_{\bullet+1}(\sigma_{\mathrm{i}})] \tag{3.91a}$$

$$r_k \triangleq (r_{k-1} + (-1)^{\bullet+1}[\partial_{\bullet+1}(\sigma_{\mathrm{j}_k})]), s_k \triangleq [\partial_{\bullet+1}(\sigma_{\mathrm{j}_k})], \text{ where } \mathrm{low}([\partial_{\bullet+1}(\sigma_{\mathrm{j}_k})]) = \mathrm{low}(r_{k-1}), k \geq 1 \tag{3.91b}$$

with

$$[S^{**}_{<\mathrm{i}}] \triangleq [s_L \| s_{L-1} \| \cdots \| s_0] \tag{3.92}$$

The index $L$ exists since the $\mathrm{low}(r_k)$ monotonically decreases with increasing $k$ and $\mathrm{low}(r_k) \geq 0$.

**Proposition 3.2.16.** $[S^{**}_{<\mathrm{i}}]$ *is a solution to the minimax problem of Problem 3.90.*

*Proof.* For any $k$: $\mathrm{low}(r_k) < \mathrm{low}(r_{k-1})$

This follows since the lowest one gets zeroed upon adding $(-1)^{\bullet+1}[\partial_{\bullet+1}(\sigma_{\mathrm{j}_k})]$ to $r_{k-1}$.

For the recurrence relation in Equations 3.91, since $k \leq n$, there must exist a maximum $k$. When $k$ achieves this maximum $k^*$, $r_{k^*}$ cannot add with any more columns. When $k^*$ is reached, the minimum value of the minimax problem 3.2.1 at index i must be achieved. The only way to lower $r_{k^*}$ is to add it with another column but there are no more columns with $\mathrm{low}([\partial_{\bullet+1}(\sigma_{\mathrm{j}_k})]) = \mathrm{low}(r_{k-1})$, thus the minimum is achieved.

Thus $[S^{**}_{<\mathrm{i}}]$ is a solution to the minimax Problem 3.2.1. □

**Proposition 3.2.17.** *All indices* i *where the primal matrix reduction problem of persistent homology of Problem 3.2.1 did not minimize to* 0*, are distinct.*

*Proof.* We show that if two linear combinations:

$$r_{\mathrm{i}} = \sum_{v \in col([S^{**}_{<\mathrm{i}}])} c_v v + [\partial_{\bullet+1}(\sigma_{\mathrm{i}})] \tag{3.93}$$

$$r_{\mathrm{j}} = \sum_{v \in col([S^{**}_{<\mathrm{j}}])} c_v v + [\partial_{\bullet+1}(\sigma_{\mathrm{j}})] \tag{3.94}$$

where $\mathrm{low}(r_{\mathrm{i}}) = \mathrm{low}(r_{\mathrm{j}})$, then $\mathrm{low}(r_{\mathrm{j}})$ is suboptimal.

Since $r_i$ is a linear combination of columns to the left of index i and i < j, then $r_i$ is a linear combination of columns to the left of index j.

Thus $r_j$ can still minimize its lowest one by adding with $r_i$ to form another linear combination of columns to the left of j that include $[\partial_{\bullet+1}(\sigma_j)]$ as a summand.

Thus all minimax values found by the primal matrix reduction problem are unique. □

A simple algorithm on the boundary matrix can find this solution. It is given in Algorithm 1. That algorithm can be refined to look more like Gaussian elimination to involve columns

**Algorithm 1:** An algorithm to solve Problem 3.2.1

```
input: ∂, an n × n matrix   /* let ∂[i] denote the ith column of matrix ∂ */
L ← [-1,...,-1];   /* L of length n */
for j= 0,...,n do
    r ← ∂[j]   /* let r[i] denote the ith row of column r */
    while r ≠ 0 and low(r) ≠ −1 do
        for k = j,...,1 do
            if low(r) = low(∂[k]) then
                r ← r − r[low(r)]∂[k]
            end
        end
    end
    if r ≠ 0 then
        L[low(r)] ← j
    end
end
return L
```

that have already solved the minimax problem of Problem 3.2.1. This algorithm is called the "standard matrix reduction algorithm" and given in Algorithm 2.

The standard matrix reduction algorithm still solves the minimax problem since it suffices to minimize the "lowest" nonzero of any column i using columns from the causal past submatrix. We define a column that witnesses the minimax Problem 3.2.1:

**Definition 3.2.18.** *A column which gives the minimizing linear combination of columns from a solution submatrix $[S_{<i}]$ to Problem 3.2.1 is called a* ***reduced column****.*

**Proposition 3.2.19.** *The columns* $R[\mathrm{j}], \forall \mathrm{j}$ *from Algorithm 2 are reduced columns and consist of a linear combination of causal past columns.*

*Proof.* **For the first part:**

The columns $R[\mathrm{i}]$ from Algorithm 2 are a linear combination of columns from $R[\mathrm{j}], \mathrm{j} < \mathrm{i}$. The algorithm adds a column $R[\mathrm{j}]$ to $R[\mathrm{i}]$ if it can match the lowest nonzero. This monotonically raises (decreases the index of) the lowest nonzero since the "if" inside the while loop only checks for the lowest one of any remaining column. Thus, at the termination of the while loop, $R[\mathrm{i}]$ is a reduced column.

**For the second part:**

We prove that the reduced columns are a linear combination of columns from its causal past submatrix by induction:

For $\mathrm{i} = 1$: $R[1] = [\partial(\sigma_1)]$ has no causal past and thus must be reduced.

Induction step for $\mathrm{i} + 1$:

By the algorithm, $R[\mathrm{i}+1]$ is a linear combination of $R[\mathrm{j}], \mathrm{j} < \mathrm{i}+1$. By induction $R[\mathrm{j}]$ is also a linear combination of causal past columns $[\partial(\sigma_k)], k < \mathrm{j}$ for j. By the transitivity of the total order for the filtration, we must have that the causal past columns for j are causal past columns for column $\mathrm{i}+1$. Thus $R[\mathrm{i}+1]$ is a linear combination of causal past columns for column $\mathrm{i}+1$. □

**Algorithm 2:** The standard matrix reduction algorithm

input: $\partial$, an $n \times n$ matrix
$R \leftarrow \partial$ /* let $R[\mathrm{i}]$ denote the ith column of matrix $R$ and $R[\mathrm{i}][\mathrm{j}]$ its jth row */
$L \leftarrow$ [-1,...,-1]; /* $L$ of length $n$ */
**for** j= *1,...,n* **do**
  **while** $R[\mathrm{j}] \neq 0$ *and* $L[low(R[\mathrm{j}])] \neq -1$ **do**
    $c_{\mathrm{j}} \leftarrow (R[\mathrm{j}])[\mathrm{low}(R[\mathrm{j}])]$ ;
    $R[\mathrm{j}] \leftarrow R[\mathrm{j}] - c_{\mathrm{j}} R[L[\mathrm{low}(R[\mathrm{j}])]]$;
  **end**
  **if** $R[\mathrm{j}] \neq 0$ **then**
    $L[\mathrm{low}(R[\mathrm{j}])] \leftarrow \mathrm{j}$;
  **end**
**end**
**return** $L$

**Theorem 3.2.20.** *Let* $(H_\bullet : \textbf{\textit{Simp}} \to \textbf{\textit{Vec}}, Q(\textbf{\textit{Simp}}) = (V = \{K_{\mathrm{j}}\}_{\mathrm{j}=1}^{n}, E = \{inc_{\mathrm{j,i}} : K_{\mathrm{j}} \hookrightarrow K_{\mathrm{i}}\}_{\mathrm{i,j} \in \{0,\dots,n\} : \mathrm{j} \leq \mathrm{i}}))$ *be the homology functor with a filtration and let* $V$ *be its corresponding persistence module.*

*For any* $\bullet \geq 0$*, let the* $\bullet + 1$*-dimensional boundary matrix* $[\partial_{\bullet+1}^{K_n}]$ *be the input to the Primal Matrix Reduction Problem, Problem 3.2.1.*

*For each* $\mathrm{i} \in [n]$*:*

- *The minimax solutions* $(\mathrm{j}^*, \mathrm{i}), \mathrm{j}^* \in [n]$ *forms pairs of indices from* $K_1 \subseteq \cdots \subseteq K_n$ *which are exactly the endpoint indices of the finite intervals* $I[\mathrm{j}^*, \mathrm{i})$ *for* $\bullet$*-dimensional generators of some interval decomposition of* $V$ *from Gabriel's theorem.*

- *The solutions* $(0, \mathrm{i})$ *correspond to the infinite intervals* $I[\mathrm{i}, \infty)$ *for a* $(\bullet + 1)$*-dimensional generator of some interval decomposition* $V$ *from Gabriel's theorem.*

*Proof.* **For the solutions** $(\mathrm{j}^*, \mathrm{i}) : \mathrm{j}^* \neq 0$**:**

We know that each minimax solution involves the generator $\sigma_{\mathrm{j}^*}$, which did not exist before index $\mathrm{j}^*$. This must correspond to a one dimensional interval $I[\mathrm{j}^*, \mathrm{i})$.

We claim that this interval is indecomposable. We can prove this by induction on the number of columns $L$ of $[S_{<\mathrm{i}}^{**}]$.

1. **Base case:**

   For a $\bullet + 1$-dimensional simplex $\sigma_{\mathrm{i}}$, if $L = 1$, then the minimax solution is

   $$(\mathrm{low}([\partial_{\bullet+1}(\sigma_{\mathrm{i}})]), \mathrm{i}) \tag{3.95}$$

   This must be indecomposable because there are no subintervals for the interval $I[\mathrm{low}([\partial_{\bullet+1}(\sigma_{\mathrm{i}})]), \mathrm{i})$.

2. **Induction Step:**

   Assuming we have that for all possible $l \leq L$ that $I[\mathrm{j}^*, \mathrm{i})$ is indecomposable.

   For $L + 1$, if $I[\mathrm{j}^*, \mathrm{i})$ were decomposable as:

   $$I[\mathrm{j}^*, \mathrm{i}) = I[\mathrm{j}^*, k) \oplus I[k, \mathrm{i}) \tag{3.96}$$

where by the induction hypothesis, both intervals $I[\mathrm{j}^*, k), I[k, \mathrm{i})$ are indecomposable. However, this would mean that at index $k$ there is both a generator and a relation. The filtration is simplex-wise, however. Contradiction.

**For the solutions $(0, \mathrm{i})$:**

Of the remaining solutions $(0, \mathrm{i})$, we know by Observation 3.2.14 that there is a creation event $\ker(\partial_{\bullet+1}^{K_\mathrm{i}})$. These cannot be merged with $\partial_{\bullet+1}^{K_n}$. Thus they form one dimensional interval modules with infinite length.

Since every possible generator and relation that span the kernel and image of the boundary map in all dimensions are considered when solving the primal minimax problem. Thus:

$$V' \triangleq \bigoplus_{(\mathrm{j}^*,\mathrm{i}) \text{ solutions to Problem 3.2.1 } \mathrm{j}^* \neq 0} I[\mathrm{j}^*, \mathrm{i}) \oplus \bigoplus_{(0,\mathrm{i}) \text{ solutions to Problem 3.2.1}} I[\mathrm{i}, \infty) \tag{3.97}$$

has an isomorphism with $V$. We need to show that the following diagram commutes up to isomorphism:

$$\begin{array}{ccc} H_\bullet(K_a) & \xrightarrow{\phi_{a,b}} & H_\bullet(K_b) \\ \cong & & \cong \\ \bigoplus_{[\mathrm{j},\mathrm{i})\ni a} I[\mathrm{j}, \mathrm{i}) \oplus \bigoplus_{[\mathrm{i},\infty)\ni a} I[\mathrm{i}, \infty) & \xrightarrow[\psi_{a,b}]{} & \bigoplus_{[\mathrm{j},\mathrm{i})\ni b} I[\mathrm{j}, \mathrm{i}) \oplus \bigoplus_{[\mathrm{i},\infty)\ni b} I[\mathrm{i}, \infty) \end{array} \tag{3.98}$$

where $\phi_{a,b} : H_\bullet(K_a) \to H_\bullet(K_b)$ are induced homomorphisms from inclusion between $K_a$ and $K_b$ for $a < b$.

Certainly, at any time index $a \in [n]$, the vertical isomorphisms between the homology vector spaces follow since the $\bullet$-dimensional generators from time index $\mathrm{j} : \mathrm{j} \leq a$ that do not merge (drop the rank by 1) through $\bullet$-dimensional boundaries must span $H_\bullet(K_a)$.

Furthermore, we have that for any $a, b \in [n], a \leq b$:

$$\operatorname{rank}(\phi_{a,b}) = \operatorname{rank}(\psi_{a,b}) = \dim(H_\bullet(K_a)) - |I[\mathrm{j}, \mathrm{i}) : [\mathrm{j}, \mathrm{i}) \ni a \text{ and } \mathrm{i} \leq a| \tag{3.99}$$

This is because all possible $\bullet$-dimensional boundaries are considered for the intervals.

Thus we have some decomposition as promised by Gabriel's theorem. □

We can also give the following interpretation of each interval in the interval decomposition in terms of the general framework:

**Corollary 3.2.21.** *(Gabriel's Theorem in the context of the general framework of Question 3.1.1)*

*After choosing a basis to match the interval decomposition of Gabriel's theorem from the minimax problem, a generator created at time index $s$ zeros at time index* $\mathrm{i}$ *when:*

1. *(Not Any Shorter) All new independent p-dimensional generators created at time $s' : s \leq s' < \mathrm{i}$ that* ***could merge*** *using a sequence of connected p-dimensional boundaries that connect to the p-dimensional boundary at index* $\mathrm{i}$ *by a rank drop of* $1$ *at time* $\mathrm{j} : \mathrm{j} < \mathrm{i}$ *have already zeroed by this choice of basis.*

2. *(Not Any Longer) The p-dimensional generators from part (1) are all the possible independent generators connected to the p-dimensional boundary at index* $\mathrm{i}$ *through p-dimensional boundaries created at time $s' : s \leq s' < \mathrm{i}$.*

*Proof.* By definition of an indecomposable interval, the interval $I[s, \mathrm{i})$ must start at the time $s$ when an independent generator is introduced and end at some time $\mathrm{i}$ in the future when it merges with other generators through a relation.

**(It cannot be shorter):**

This is the same proof as in the induction step of proof of Theorem 3.2.20.

**(It cannot be longer):**

If it were longer, then the interval would start at $s' : s' < s$. This would mean that the interval would have:

$$I[s', \mathrm{i}) = I[s', s) \oplus I[s, \mathrm{i}) \tag{3.100}$$

This means that it is not indecomposable, which violates the indecomposable interval decomposition of Gabriel's theorem.

This gives the two conditions in the Corollary since the index $\mathrm{i}$ was arbitrary. □

**Topological Interpretation:**

We can restate the matrix reduction problem in the downstairs homological persistence module of the finite simple acyclic subcategory $Q(\mathbf{Simp})$ in terms of the topology of the upstairs filtration. We need the following definition:

**Definition 3.2.22.** *A* ***connected neighborhood*** $N(\sigma)$ *of a simplex* $\sigma \in C_{p+1}(K_n)$ *is some set of simplices where all of its simplices* $\tau \in C_{p+1}(K_n)$ *have that* $\tau$ *belongs to a sequence* $(\tau_{\mathrm{i}})_{\mathrm{i}\in\mathbb{N}}, \tau_0 = \sigma$ *so that* $\partial_{p+1}(\tau_{\mathrm{i}}) \cap \partial_{p+1}(\tau_{\mathrm{i}+1}) \neq \varnothing$ *as an intersection of formal sums.*

*This intersection of* $p$*-dimensional chains is equivalent to saying that* $\tau_{\mathrm{i}}$ *and* $\tau_{\mathrm{i}+1}$ *share a common face.*

There is a special kind of connected neighborhood which forms a cycle (with respect to the boundary map $\partial_{p+1} : C_{p+1}(K_n) \to C_p(K_n)$, we call this a cyclic connected neighborhood. We define this formally here:

**Definition 3.2.23.** *A connected neighborhood* $N(\sigma), \sigma \in C_{p+1}(K_n)$ *is cyclic iff*

$$\sum_{\tau \in N(\sigma)} c_\tau \partial_{p+1}(\tau) = \overrightarrow{0}, \text{ for some coefficients } c_\tau \in \mathbb{R} \setminus \{0\} \tag{3.101}$$

*In other words,* $\sum_{\tau\in N(\sigma)} c_\tau \tau \in ker(\partial_{p+1})$

This allows us to make the following topological restatement of the algebraic question posed in Question 3.2.1:

For $p \geq 1$, we determine if at index i there is a creation event or a destruction event in terms of the topology. Define the set of $p+1$-dimensional simplices $\sigma_{\mathrm{j}}, \mathrm{j} < \mathrm{i}$ which have all its $p$-dimensional faces in the causal past of the simplex $\sigma_{\mathrm{i}}$:

$$C_{\mathrm{i}}^{past} \triangleq \{\sigma_{\mathrm{j}} : \max_{\tau_{k'} \in \partial_{p+1}(\sigma_{\mathrm{j}})} k' \leq \max_{\tau_k \in \partial_{p+1}(\sigma_{\mathrm{i}})} k \text{ and } \mathrm{j} < \mathrm{i}\} \tag{3.102}$$

Similarly, define the set of $p+1$-dimensional simplices $\sigma_{\mathrm{j}}, \mathrm{j} > \mathrm{i}$ which have $p$-dimensional faces in the causal future of the faces of the $p$-dimensional boundary of the $p+1$ dimensional simplex $\sigma_{\mathrm{i}}$:

$$C_{\mathrm{i}}^{future} \triangleq \{\sigma_{\mathrm{j}} : \min_{\tau_{k'} \in \partial_{p+1}(\sigma_{\mathrm{j}})} k' \geq \min_{\tau_k \in \partial_{p+1}(\sigma_{\mathrm{i}})} k \text{ and } \mathrm{j} > \mathrm{i}\} \tag{3.103}$$

We define a connected causal past neighborhood of $p+1$-dimensional simplex $\sigma_{\mathrm{i}}$ as a connected neighborhood that respects the causal past of $\sigma_{\mathrm{i}}$:

$$N_{causal}^{past}(\sigma_{\mathrm{i}}) \triangleq N(\sigma_{\mathrm{i}}) \cap C_{\mathrm{i}}^{past} \tag{3.104}$$

Analogously, we can define the connected causal future neighborhood for $\sigma_{\mathrm{i}}$:

$$N_{causal}^{future}(\sigma_{\mathrm{i}}) \triangleq N(\sigma_{\mathrm{i}}) \cap C_{\mathrm{i}}^{future} \tag{3.105}$$

This feasible region is a surrogate for the causal past and future submatrices used for algebraic linear independence testing on the matrix as mentioned in the matrix reduction problem for persistent homology as given in Problem 3.2.1. This is stated in the following proposition:

**Proposition 3.2.24.** *Let $\sigma_{\mathrm{i}} \in C_{p+1}(K_n)$, then:*

*There exists a set $N_{causal}^{past}(\sigma_{\mathrm{i}})$ which is cyclic iff index* i *is a $p+1$-dimensional creation event*

*Proof.* Let $N_{causal}^{past}(\sigma_{\mathrm{i}})$ be the causal past of $\sigma_{\mathrm{i}}$ and let $[\partial_{p+1}]_{<[\partial_{p+1}(\sigma_{\mathrm{i}})]}$ be a causal past submatrix of column $[\partial_{p+1}(\sigma_{\mathrm{i}})]$.

We show that the concatenation of columns $[\|_{\tau_{\mathrm{j}} \in N_{causal}^{past}(\sigma_{\mathrm{i}})}[\partial_{p+1}(\tau_{\mathrm{j}})]]$ ordered by the index of each $\tau_{\mathrm{j}}$ is a submatrix of $[\partial_{p+1}]_{<[\partial_{p+1}(\sigma_{\mathrm{i}})]}$.

This follows by the induced submatrix map in the following commuting diagram:

$$\begin{array}{ccc}
N_{causal}^{past}(\sigma_{\mathrm{i}}) & \hookrightarrow & C_{\mathrm{i}}^{past} \\
\downarrow{\scriptstyle [\partial_{p+1}(\bullet)]} & & \downarrow{\scriptstyle [\partial_{p+1}(\bullet)]} \\
([\partial_{p+1}(\tau_{\mathrm{j}})])_{\tau_{\mathrm{j}} \in N_{causal}^{past}(\sigma_{\mathrm{i}})} & \xrightarrow{\text{subsequence}} & ([\partial_{p+1}(\sigma_{\mathrm{j}})])_{\sigma_{\mathrm{j}} \in C_{\mathrm{i}}^{past}} \\
\downarrow{\scriptstyle [\|\bullet]} & & \downarrow{\scriptstyle [\|\bullet]} \\
[\|_{\tau_{\mathrm{j}} \in N_{causal}^{past}(\sigma_{\mathrm{i}})}[\partial_{p+1}(\tau_{\mathrm{j}})]] & \xrightarrow[\text{submatrix}]{} & [\partial_{p+1}]_{<[\partial_{p+1}(\sigma_{\mathrm{i}})]}
\end{array} \tag{3.106}$$

The $[\partial_{p+1}(\bullet)] : C_{p+1}(K_n) \to \mathrm{ob}(\mathbf{Mat})$ map takes a set of $p+1$ simplices with the ordering provided by the filtration and maps each simplex to a sequence of column representations of their boundaries.

The map $[\|\bullet] : \text{ob}(\mathbf{Mat})^{\mathbb{N}} \to \text{ob}(\mathbf{Mat})$ takes a sequence of columns and concatenates these columns in the order of the sequence of columns to form a matrix.

The horizontal maps in Diagram 3.106 are as follows:

1. The upper inclusion is by definition of $N^{past}_{causal}(\sigma_{\mathrm{i}})$.

   The remaining two maps are induced by the inclusion:

2. The subsequence relationship is induced by the upper inclusion and

3. The submatrix relationship is induced by the subsequence relationship upstairs

**For the if and only if :**

($\Leftarrow$) For the minimax Problem 3.2.1, if 0 can be achieved, then the solution $[S^{*}_{<\mathrm{i}}]$ consists of columns which have some linear combination equal to column i.

Algorithm 1 finds columns from $[\partial_{p+1}]_{<[\partial_{p+1}(\sigma_{\mathrm{i}})]}$ which match on the "lowest ones", or maximal indices of each column. According to Proposition 3.2.16 this set of columns $S_L \triangleq \{s_k = [\partial_{p+1}(\sigma_{\mathrm{j}_k})] : k = 1, \ldots, L\}$ are the columns of $[S^{**}_{<\mathrm{i}}]$ from Equation 3.92.

Each of these columns $[\partial_{p+1}(\sigma_{\mathrm{j}_k})]$ correspond to a boundary $\partial_{p+1}(\sigma_{\mathrm{j}_k})$. Furthermore, $\text{low}(s_k)$ corresponds to the highest indexed, or latest, $p$-simplex face of $\sigma_{\mathrm{j}_k}$ in the formal sum of $p$-simplices, or $p$-chain, corresponding to $s_k = \partial_{p+1}(\sigma_{\mathrm{j}_k})$. Since for every $k$,

$$\text{low}([\partial_{p+1}(\sigma_{\mathrm{j}_k})]) = \text{low}(r_{k-1}) \tag{3.107}$$

we must have a sequence of $p+1$-dimensional simplices each adjacent $p+1$-dimensional simplex connected by atleast one $p$-dimensional face.

Thus every $p+1$-chain $\sigma_{\mathrm{j}_k}$ is connected by the sequence $(\sigma_{\mathrm{j}_{k'}})_{k'<k}$. Therefore the set of $p+1$-chains $S_L$ forms a connected neighborhood $N(\sigma_{\mathrm{i}})$ as defined in Definition 3.2.23.

This connected neighborhood $N(\sigma_{\mathrm{i}})$ is cyclic since $r_{k^*} = [\vec{0}]$ as a linear combination of all the columns from $\{[\partial_{\bullet+1}(\sigma_{\mathrm{j}_k})] : [\partial_{\bullet+1}(\sigma_{\mathrm{j}_k})] = (-1)^{\bullet+1}(r_{k+1} - r_k)\}_{k=0}^{L}$.

($\Rightarrow$) A cyclic neighborhood $N(\sigma_{\mathrm{i}})$ can be rewritten by taking column representations of the $p$-chains in $N(\sigma_{\mathrm{i}})$ as $\{[\partial(\sigma_{\mathrm{j}})] : \sigma_{\mathrm{j}} \in N(\sigma_{\mathrm{i}})\}$. Since $N(\sigma_{\mathrm{i}})$ is cyclic, there are coefficients $c_{\sigma_{\mathrm{j}}} \in \mathbb{R} \setminus \{0\}$ so that $\sum_{\sigma_{\mathrm{j}} \in N(\sigma_{\mathrm{i}})} c_{\sigma_{\mathrm{j}}} [\partial(\sigma_{\mathrm{j}})] = [\vec{0}]$. Thus, the minimax problem of Problem 3.2.1 can achieve 0. □

### 3.2.25 Time Reversal and Duality

In persistence theory, when given a pair of a functor and a finite simple acyclic subcategory $(F : \mathcal{D} \to \mathbf{Vec}, Q(\mathcal{D}))$, if the functor $F$ is covariant, then the directions of the arrows in the upstairs $Q(\mathcal{D})$ category are preserved in the downstairs category $\mathrm{im}(F\,|_{Q(\mathcal{D})})$. On the other hand, if the functor $F$ is contravariant, then the arrows in the downstairs $\mathrm{im}(F\,|_{Q(\mathcal{D})})$ are all reversed. We call the use of a contravariant functor $F$ on an acyclic subcategory $Q(\mathcal{D})$ as introducing **time reversal through contravariance**.

According to Section 2.4.20, we have the duality functor between finite dimensional vector spaces and finite dimensional dual vector spaces:

$$\bullet^{\perp} : \mathbf{Vec} \to \mathbf{Vec}^{\perp} \tag{3.108}$$

For the case of covariant functors $F : \mathcal{D} \to \mathbf{Vec}$, we can then define the dual functor $F^{\perp} : \mathcal{D} \to \mathbf{Vec}^{\perp}$ as the functor $F^{\perp} \triangleq \bullet^{\perp} \circ F$. The functor $F^{\perp}$ is a contravariant functor due to the following non-commuting diagram on vector spaces $V, W$:

$$\begin{array}{ccc} V & \xrightarrow{\phi} & W \\ \updownarrow{\scriptstyle \bullet^{\perp}} & & \updownarrow{\scriptstyle \bullet^{\perp}} \\ V^{\perp} & \xleftarrow[\phi^{\perp}]{} & W^{\perp} \end{array} \tag{3.109}$$

We define a dual persistence module in terms of dual vector spaces:

**Definition 3.2.26.** *For a pair of a functor with finite simple acyclic subcategory* $(F : \mathcal{D} \to \mathcal{E}, Q(\mathcal{D}) \subseteq \mathcal{D})$,

*If* $V \triangleq im(F\,|_{Q(\mathcal{D})})$ *is a persistence module.*

*A* ***dual persistence module*** $V^{\perp}$ *of the persistence module* $V$ *is the persistence module:*

$$V^{\perp} \triangleq (im(F\,|_{Q(\mathcal{D})}))^{\perp} \tag{3.110}$$

This means the concept of time is reversed in a dual persistence module for vector spaces. We call this the **time reversal through duality** principle.

### Computing Persistent Cohomology

When computing the interval decomposables of Gabriel's theorem for a persistence module induced by a filtration for the covariant $H_\bullet$ functor, we can apply the contravariant dual functor $H_\bullet^\perp$, called the cohomology functor, instead. This reverses the arrows of the downstairs persistence module. The $H_\bullet^\perp$ functor is defined through the dual map $\bullet^\perp$ by $H_\bullet^\perp \triangleq (H_\bullet)^\perp$. Using the definition of the dual map, for a homology vector space $H_\bullet(K)$, define the cohomology vector space as follows:

$$H_\bullet^\perp(K) \triangleq \mathrm{Lin}(H_\bullet(K), \mathbb{R}) \tag{3.111}$$

where $\mathrm{Lin}(H_\bullet(K), \mathbb{R})$ defines the space of all linear maps from the homology vector space $H_\bullet(K)$ to the real numbers. The maps in $H_\bullet^\perp(K)$ are called cocycles of the cohomology vector space. The dual map $\bullet^\perp$ maps homology to cohomology on a basis as follows:

$$\bullet^\perp : H_\bullet(K) \to H_\bullet^\perp(K); \bullet^\perp : \mathrm{e_i} \mapsto \phi_{\mathrm{e_i}}, \tag{3.112a}$$

$$\text{where } \phi_{\mathrm{e_i}}(\mathrm{e_j}) = \begin{cases} 0 & \mathrm{i} \neq \mathrm{j} \\ 1 & \mathrm{i} = \mathrm{j} \end{cases} \text{ for any basis } \{\mathrm{e_i}\}_{\mathrm{i}=1}^n \text{ with } \mathrm{span}(\{\mathrm{e_i}\}_{\mathrm{i}=1}^n) = H_\bullet(K) \tag{3.112b}$$

The dual map $\bullet^\perp$ is the linear extension of this map on bases.

**Observation 3.2.27.** *Using the dual map between homology and cohomology, we can achieve the functor diagram of Equation 3.113 for* $1 \leq \mathrm{i} < \mathrm{j} \leq n$ *of the filtration* $Q(\boldsymbol{Simp}) = (V = \{K_\mathrm{i}\}_{\mathrm{i}=0}^n, E = \{inc_{\mathrm{i,j}} : K_\mathrm{i} \hookrightarrow K_\mathrm{j}\}_{\mathrm{i,j} \in \{0,\dots,n\} : \mathrm{i} \leq \mathrm{j}})$*:*

$$\begin{array}{ccccccc} K_\mathrm{i} & \hookrightarrow & K_{\mathrm{i}+1} & \hookrightarrow & \dots & \hookrightarrow & K_\mathrm{j} \\ \downarrow{\scriptstyle H_\bullet^\perp} & & \downarrow{\scriptstyle H_\bullet^\perp} & & & & \downarrow{\scriptstyle H_\bullet^\perp} \\ H_\bullet^\perp(K_\mathrm{i}) & \longleftarrow & H_\bullet^\perp(K_{\mathrm{i}+1}) & \longleftarrow & \dots & \longleftarrow & H_\bullet^\perp(K_\mathrm{j}) \\ \updownarrow{\scriptstyle \cong} & & \updownarrow{\scriptstyle \cong} & & & & \updownarrow{\scriptstyle \cong} \\ H_\bullet(K_\mathrm{i}) & \longrightarrow & H_\bullet(K_{\mathrm{i}+1}) & \longrightarrow & \dots & \longrightarrow & H_\bullet(K_\mathrm{j}) \end{array} \tag{3.113}$$

Thus persistent cohomology and persistent homology of a filtration are equivalent up to arrow flips by **time reversal through duality**. This means that going forward and backward in time in either cohomology or homology is equivalent by the commutativity of Diagram 3.113. This also means that creation and destruction events at each index i coincide for cohomology and homology.

By the Universal Coefficient Theorem [50] an equivalent way to define cohomology is through the homology of the dual boundary map:

$$H_\bullet^\perp(K_\mathrm{i}) \cong \frac{\ker((\partial_{\bullet+1}^{K_\mathrm{i}})^\perp)}{\mathrm{im}((\partial_\bullet^{K_\mathrm{i}})^\perp)} \tag{3.114}$$

The boundary map is linear and thus can have a dual version as given in Equation 2.115.

If we were to directly compute the cohomology vector spaces from a later time toward an earlier time, we would have to bootstrap compute $H_\bullet(K_n)$ directly without a construction for $K_n$. Then, going back in time, we would delete simplices from the simplicial complex until the simplicial complex becomes empty. Bootstrapping the homological generators at a high dimension is difficult without constructing $K_n$.

Given a boundary matrix $[\partial_{p+1}]$, its algebraic dual is the matrix transpose $[\partial_{p+1}]^T$ called the coboundary matrix. By Proposition 2.4.58, we know that $[\partial_{p+1}]^T = [\partial_{p+1}^\perp]$. In particular, we will denote with a slight abuse of notation the ith column of $[\partial_{p+1}]^T$ by $[\partial_{p+1}^\perp(\sigma_\mathrm{i})]$ where $\sigma_\mathrm{i} \in K_\mathrm{i} \setminus K_{\mathrm{i}-1}$ is the only simplex in this set. The simplex $\sigma_\mathrm{i}$ is being viewed as a covector when composed with the dual boundary map: $\partial_{p+1}^\perp$.

Since the arrow of time flips in the dual persistence module, we define the "high" function on the columns of $[\partial_{p+1}]^T$, which is dual to the low function on a column $c$:

$$\mathrm{high}(c) \triangleq \begin{cases} \min_{c[\mathrm{j}]\neq 0} \mathrm{j} & c \neq [\,\overrightarrow{0}\,] \\ \infty & c = [\,\overrightarrow{0}\,] \end{cases} \tag{3.115}$$

This function at column $c$, $\mathrm{high}(c)$, defines the minimum nonzero index of a column $c$. When $c = [\,\overrightarrow{0}\,]$, then $\mathrm{high}(c)$ is $\infty$.

We show that we can compute from $p = 0$ to each higher homological dimension the time intervals between creation and destruction times. This avoids having to bootstrap the cohomological sequence on $K_n$ for $p \geq 1$. The equivalence between persistent cohomology and persistent homology up to a reversal of time along with Observation 3.2.14 we can define the dual matrix reduction problem:

**Problem 3.2.2.** *[Dual Matrix Reduction Problem of Persistent Homology]*

*For* $\bullet \geq 1$*, at index* $\mathrm{i}$ *which was not a destruction event on* $[\partial_{\bullet+1}]^T$*, compute:*

$$\max_{[S_{>\mathrm{i}}]\subseteq[\partial_{\bullet+1}]^T_{>[\partial^{\perp}_{\bullet+1}(\sigma_{\mathrm{i}})]};c_v\in\mathbb{R}\setminus\{0\},\forall v\in col(S_{>\mathrm{i}})} high\Big(\sum_{v\in col(S_{>\mathrm{i}})} c_v v + [\partial^{\perp}_{\bullet+1}(\sigma_{\mathrm{i}})]\Big)\Big) \tag{3.116}$$
$$\text{where } [S_{>\mathrm{i}}] \text{ is a causal future submatrix of } [\partial_{\bullet+1}]^T_{>[\partial^{\perp}_{\bullet+1}(\sigma_{\mathrm{i}})]}$$

We show that Problem 3.2.2 can also solve for creation and destruction times at any index i as in Problem 3.2.1.

**Theorem 3.2.28.** *For index* $\mathrm{i}$ *and* $p \geq 1$*, the dual matrix reduction Problem 3.2.2 finds an index* $k^* > \mathrm{i}$ *in the future where:*

- *If* $k^* = \infty$*, then there is no destruction index in the future. Thus, this is a* $p$*-dimensional creation event for index* $\mathrm{i}$*.*
- *Otherwise,* $k^*$ *is the latest destruction index for index* $\mathrm{i}$*.*

*This is dual to the primal Problem 3.2.1 where* $k^* < \mathrm{i}$ *is in the past and gives a creation event of dimension* $p$*.*

*Proof.* In the case of persistence, the boundary matrix $[\partial_{p+1}]$ is used to add columns from the causal past of a column i to column i. According to Problem 3.2.1, the $\text{low}(R[\mathrm{i}])$ of a reduced column $R[\mathrm{i}]$ is the last $p$-simplex to create a cycle. Thus $(\text{low}(R[\mathrm{i}]), \mathrm{i})$ is a pair of indices of creation and destruction events.

For Problem 3.2.2, for $p \geq 1$ and an index i which is not a destruction event at index i, we are solving a maximin problem over the causal future submatrix of $[\partial^{\perp}_{p+1}(\sigma_{\mathrm{i}})]$, denoted $[\partial_{p+1}]^T_{>[\partial^{\perp}_{p+1}(\sigma_{\mathrm{i}})]}$, which is the following transposed matrix $[\partial_{p+1}]^T_{>[\partial^{\perp}_{p+1}(\sigma_{\mathrm{i}})]}$ (see Equation 2.128).

In Observation 3.2.14, we showed that for the homological sequence that every index i is either a destruction event or a creation event for the $p$-dimensional homology vector space. Since $Q(\mathbf{Simp}) : K_1 \subseteq ... \subseteq K_n$ is a simplex-wise filtration, by Observation 3.2.14, at index i there is no destruction event so there must be a $p$-dimensional creation event.

For this $p$-dimensional creation event i, we are checking for independence of column $[\partial^{\perp}_{p+1}(\sigma_{\mathrm{i}})]$ from the causal future submatrix. In the maximization of Problem 3.2.2, if we can get $\infty$ as the maximum, then the maximizing $[S^*_{>\mathrm{i}}] \subseteq [\partial_{p+1}]^T_{>[\partial^{\perp}_{p+1}(\sigma_{\mathrm{i}})]}$ submatrix solution, must have

$$\sum_{[\partial^{\perp}_{p+1}(w)] \in col([S^*_{>\mathrm{i}}])} c_w[\partial^{\perp}_{p+1}(w)] + [\partial^{\perp}_{p+1}(\sigma_{\mathrm{i}})] = [\partial^{\perp}_{p+1}(\sum_{[\partial^{\perp}_{p+1}(w)] \in col([S^*_{>\mathrm{i}}])} c_w w + \sigma_{\mathrm{i}})] = [\overrightarrow{0}] \quad (3.117)$$

which means that

$$\tau_{\mathrm{i}} \triangleq \sum_{[\partial^{\perp}_{p+1}(w)] \in col([S^*_{>\mathrm{i}}])} c_w w + \sigma_{\mathrm{i}} \in \ker(\partial^{\perp}_{p+1}(K_{\mathrm{i}})) \quad (3.118)$$

This $p$-dimensional cocycle $\tau_{\mathrm{i}}$ is independent from any $\tau_k, k > \mathrm{i}$ which is also of the same form as Equation 3.118 with i replaced by $k$. This is because $\sigma_{\mathrm{i}}$ cannot belong to $\tau_k$ which is a sum of $p$-simplices from the future.

Thus, we have that at index i, there is a creation event for the cohomology vector space $H^{\perp}_p(K_n)$, which by duality is a creation event for the homology vector space $H_p(K_n)$.

Thus index i is a creation event for $p$-dimensional homology.

If we cannot maximize Problem 3.2.2 to $\infty$, then we have independence of column $[\partial^{\perp}_{p+1}(\sigma_{\mathrm{i}})]$ from its causal future submatrix. The highest nonzero $k^*$ is the latest of the earliest $p+1$ dimensional simplex of a $p+1$ dimensional coboundary incident to $\sigma_{\mathrm{i}} \in C_p(K_{\mathrm{i}})$. In other words, we can write:

$$\sum_{[\partial^{\perp}_{p+1}(w)] \in col([S^*_{>\mathrm{i}}])} c_w[\partial^{\perp}_{p+1}(w)] + [\partial^{\perp}_{p+1}(\sigma_{\mathrm{i}})] = [\partial^{\perp}_{p+1}(\sum_{[\partial^{\perp}_{p+1}(w)] \in col([S^*_{>\mathrm{i}}])} c_w w + \sigma_{\mathrm{i}})] \neq [\overrightarrow{0}] \quad (3.119)$$

where exactly one $w^* \in C_p(K_n)$ satisfying $[\partial^{\perp}_{p+1}(w^*)] \in col([S^*_{>\mathrm{i}}])$ has $w^* = \sigma_{k^*}$.

This means that

$$\tau_{k^*} \triangleq \partial_{p+1}^{\perp}(\sum_{[\partial_{p+1}^{\perp}(w)] \in col([S^*_{>\mathrm{i}}])} c_w w + \sigma_{\mathrm{i}}) \in \mathrm{im}(\partial_{p+1}^{\perp}) \tag{3.120}$$

By the involution property of $\bullet^{\perp}$, maps do not change rank. Thus $\mathrm{im}(\partial_{p+1}^{\perp}) \cong \mathrm{im}(\partial_{p+1})$. This means the dual of $\tau_{k^*}$, denoted $\tau_{k^*}^{\perp}$, has $\tau_{k^*}^{\perp} \in \mathrm{im}(\partial_{p+1})$ with $k^*$ maximized. Thus index $k^*$ is an index of a destruction event for the creation event at index i.

□

# 4. HYPHA: A FRAMEWORK BASED ON SEPARATION OF PARALLELISMS TO ACCELERATE PERSISTENT HOMOLOGY MATRIX REDUCTION

We know that the question of persistence for a filtration of simplicial complexes can be solved by computing an interval indecomposition of a sequence of vector spaces, or persistence module, according to Gabriel's theorem [29]. The interval decomposables only depend on the persistence module, which only depends on the homology vector spaces. It is thus possible to compute persistent homology entirely at the matrix representation level as shown in Theorem 3.2.20. We call any such computation at the matrix representation level "Persistent homology (PH) matrix reduction."

Persistent homology (PH) matrix reduction is an important tool for data analytics in many application areas since it can disentangle the independence of induced homology generators over multiple scales from data with connectivity. Due to its highly irregular execution patterns in computation, it is challenging to gain high efficiency in parallel processing for increasingly large data sets.

Since the induced persistence module of homology vector spaces respects monotonic causality, it appears that we are restricted to many dependencies amongst columns in the boundary matrix. This appears to occlude any hope for enough independence to give a parallel approach and thus makes a GPU based algorithm appear infeasible. This thus brings about the question:

**Question 4.0.1.** *Can persistent homology be computed using the GPU?*

In this section we answer this in the affirmative and discuss how to accelerate the computation of the creation and destruction times for persistent homology using a hybrid GPU/CPU framework from a sparse matrix perspective.

We introduce HYPHA, a HYbrid Persistent Homology matrix reduction Accelerator, to make parallel processing highly efficient on both GPU and multicore. The essential foundation of our algorithm design and implementation is the separation of SIMT and MIMD parallelisms in PH matrix reduction computation. With such a separation, we are able to perform massive parallel scanning operations on GPU in a super-fast manner, which also collects rich information from an input boundary matrix for further parallel reduction operations on multicore with high efficiency. The HYPHA framework may provide a general purpose guidance to high performance computing on multiple hardware accelerators.

To our best knowledge, HYPHA achieves the highest performance in PH matrix reduction execution. Our experiments show speedups of up to 116x against the standard PH algorithm. Compared to the state-of-the-art parallel PH software packages, such as PHAT and DIPHA, HYPHA outperforms their fastest PH matrix reduction algorithms by factor up to ~2.3x.

## 4.1 Introduction

It is important to find and understand the shape of data in multiple dimensions, which is a major research theme of Topological Data Analysis (TDA) [51]. In TDA, the concept of persistent homology can be applied. In addition to providing topological, or qualitative understanding of data, it offers metrically stable and efficiently computable measurements with comparative and analytical insights. Because of its rigorous mathematical foundation and computing feasibility, this type of data analytics has been widely used in various areas, including sensor networks [52], bioinformatics [53], manifold learning [54, 55], deep learning [56] and many others [57].

As data analytics tasks have become increasingly intensive in both scale and computing complexity, researchers have made efforts to develop fast persistent homology algorithms. There are several open source software packages of persistent homology, e.g. JavaPlex [58], PHAT [59], and Dionysus [60], DIPHA [61], Ripser [62], and Eirene [63]. The core algorithm of these software packages is the matrix reduction on simplices. Fig. 4.1 (a) shows a simplicial 2-dimension complex; and accordingly, Fig. 4.1 (b) shows its boundary matrix. Fig. 4.1 also illustrates how to construct a boundary matrix $\partial$ for a simplicial complex with $\mathbb{Z}_2$ coefficients. As mentioned in Section 3.2.11, another rule for construction of a boundary matrix is that a column representing a simplex can only be encoded by the simplices with smaller column indices to respect the causal ordering that the indices represent. As a result, the boundary matrix is an upper triangular matrix. For any column j of $\partial$, low(j) is defined as the greatest row index i that $\partial[i,j]$ is nonzero. In the case that column j is composed of all zeros, low(j) is -1. We will use low(j) and low($\partial$[j]) interchangeably. In Fig. 4.1 (b), we can find low($\partial$[0]), low($\partial$[1]), low($\partial$[2]) are -1, low($\partial$[3]) and low($\partial$[4]) are 2, low($\partial$[5]) is 1, and low($\partial$[6]) is 5. The matrix reduction is to add column i to column j, if low(i) = low(j) and i < j. Here, the column addition executes the exclusive or (XOR) on corresponding entries of two columns. The matrix reduction will end once the low($\cdot$) is injective on the nonnegatives, i.e., for all i and j that low(i) $\neq -1$ and low(j) $\neq -1$, if i≠j, then low(i) $\neq$ low(j).

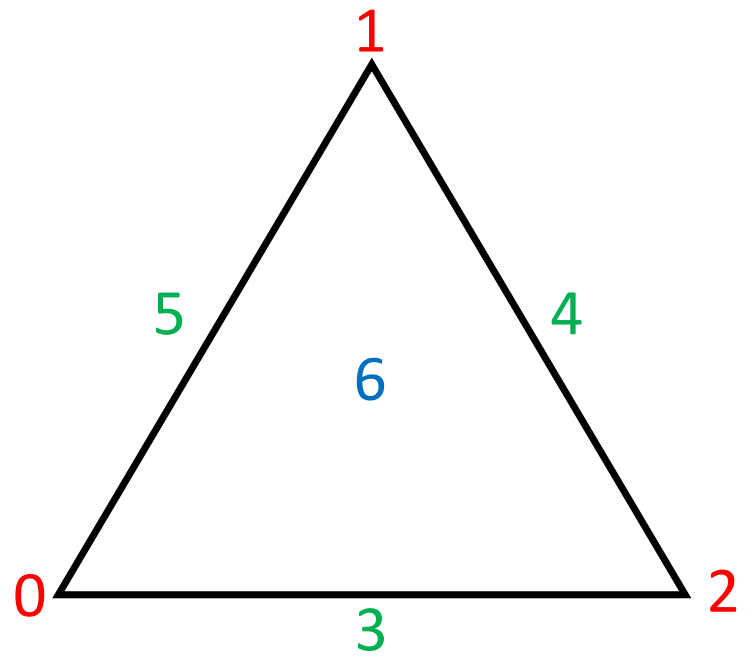

| | 0 | 1 | 2 | 3 | 4 | 5 | 6 |
|---|---|---|---|---|---|---|---|
| 0 | | | | 1 | | 1 | |
| 1 | | | | | 1 | 1 | |
| 2 | | | | 1 | 1 | | |
| 3 | | | | | | | 1 |
| 4 | | | | | | | 1 |
| 5 | | | | | | | 1 |
| 6 | | | | | | | |

**Figure 4.1.** A redrawing of Figure 2.2 with $\mathbb{Z}_2$ coefficients. A simplicial 2-dimension complex composed of 6 simplices. The points 0, 1, 2 are 0-simplices, the line-segments 3, 4, 5 are 1-simplices, and the triangle 6 is a 2-simplex. (b) The corresponding boundary matrix. In the matrix, a column representing a simplex is encoded by the simplices in its boundary, e.g., the triangle 6 has the boundary composed of line-segments 3, 4, and 5.

A basic sequential matrix reduction algorithm (more details in Sec. 4.2) is straightforward and easy to implement. With a set of optimizations [59], such as cache utilization, adopting sparse matrix format (e.g., Compressed Sparse Column (CSC)), using the binary index tree [64], and others, the sequential algorithm is efficient on CPU, especially for medium sized datasets (under 1 million simplices).

Although the sequential execution on a powerful single core CPU can leverage CPU high clock rate, large cache for fast data accesses, and zero synchronization delay, the parallel and scalable PH design is highly desirable. As datasets become increasingly large, PH matrix reduction must be processed in parallel with advanced architecture for high performance.

As Moore's Law [65] along with the Dennard's Scaling Law [66] are ending due to physical limits, to further improve performance, computation needs to be accelerated by additional advanced hardware devices, such as GPU. However, existing parallel PH matrix reduction algorithms (Sec. 4.2.6), e.g., spectral sequence algorithm [67] and chunk algorithm [68], have the following structural issues that may lead to suboptimal performance and hence have hindered their wide usage in practice.

First, it is challenging to parallelize the matrix reduction, as the algorithm itself is highly dependent in column additions. Only the columns with the same *low* values can be added, and such results can be obtained only during the execution at runtime. Second, the additions on real-world datasets are highly skewed. For all the datasets used in this work, we have observed 9.57% - 62.47% columns do not have any additions, while 0.7% - 20.6% columns get 50% column additions. However, existing parallel algorithms do not take the irregularity into implementation consideration. Without distinguishing different computing natures of columns, these parallel algorithms cannot process matrix reduction in a balanced manner. Lastly but most importantly, we observe that existing parallel algorithms cannot fully utilize the power of two effective PH optimizations, i.e., clearing [68, 69] and compression [68] (Sec. 4.2.1), potentially producing a large number of unnecessary column additions, and not fully exploiting SIMT (single instruction, multiple threads) and MIMD (multiple instructions, multiple data) parallelisms in the algorithms on advanced computing systems.

To address these issues, we propose HYPHA, a framework of separation of SIMT and MIMD parallelisms to accelerate persistent homology matrix reduction. We propose a read-only scanning phase of the boundary matrix to quickly identify 0-addition columns and collect rich information for the following matrix preprocessing and parallel matrix reduction. With that, we can unleash the power of clearing and compression to simplify the boundary matrix to a smaller submatrix and then resolve the imbalance problem in parallel matrix reduction. We separate SIMT and MIMD parallelisms of matrix reduction and implement each phase onto their best-fit hardware devices, i.e., the parallel scanning phase on GPU (SIMT), the parallel column addition on multicore (MIMD), and the parallel clearing on GPU (SIMT) and compression on multicore (MIMD). Our contributions are three folds:

- We provide an anatomy of PH matrix reduction algorithms based on large datasets, which provides insights into the separation of two types of parallelisms and computation bottlenecks in order to achieve high performance.

- We propose HYPHA, a framework based on separation of SIMT and MIMD parallelisms to accelerate PH matrix reduction. It includes a novel and effective scanning phase on GPU, a balanced parallel column addition phase on multicore, and an enhanced clearing and compression phase on both hardware devices.

- We carry out the experiments on a HPC cluster with GPUs and compare HYPHA with two state-of-the-art PH software packages, i.e., PHAT and DIPHA, on a set of real-world datasets. To our best knowledge, HYPHA achieves the highest performance compared with other solutions for PH matrix reductions at low cost.

## 4.2 Computing PH matrix reduction

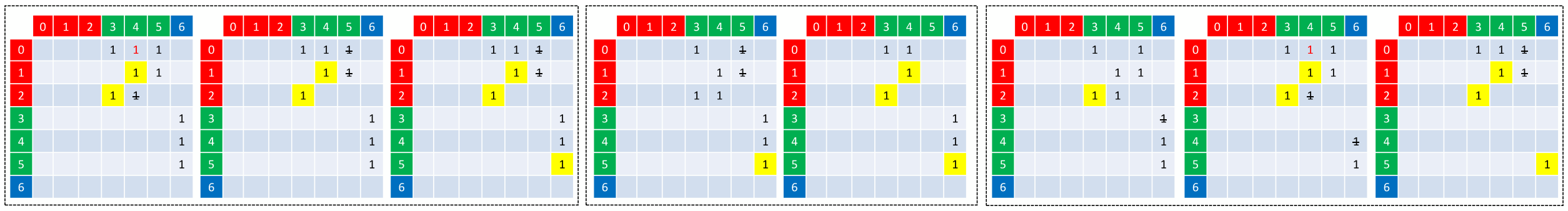

**Figure 4.2.** (a): Standard matrix reduction on the example in Fig. 4.1, where pivots (2,3), (1,4), and (5,6) are found during the execution. (b) With the clearing lemmas 4.2.2 and 4.2.3, the algorithm can zero column 5 of the boundary matrix after finding (5,6) as a pivot. (c) With the compression lemma 4.2.4, the algorithm can zero rows 3 and 4 after finding pivots (2,3) and (1,4), respectively.

Alg. 2 shows the "original" or standard matrix reduction algorithm. The algorithm processes the matrix column by column, from the left to the right. Once a column R[j] is nonzero and low($R$[j]) is found modified in the lookup table L (not the initial value -1), the algorithm knows there is another column R[i] on the left of R[j] and low($R$[i]) is equal to low($R$[j]), and then adds R[i] to R[j]. Otherwise, the algorithm updates the low($R$[j])-th position of the lookup table with the column number j, if R[j] is nonzero.

Fig. 4.2 (a) illustrates the process of matrix reduction on the boundary matrix of Fig. 4.1 (b). The algorithm checks and skips columns 0, 1, and 2 one by one. On columns 3, the algorithm sets L[2] = 3, but doesn't change the boundary matrix. On column 4, the algorithm finds low(R[4]) is 2 and L[2] is 3, and thus adds column 3 to column 4. The column addition updates the column 4 in the partially reduced boundary matrix, as shown in the left sub-figure of Fig. 4.2 (a). After that low($R$[4]) is changed to 1 and L[1] is set to 4. The algorithm then checks column 5, and finds low($R$[5]) is 1 and L[1] is 4. The algorithm adds column 4 to column 5, zeroing column 5, as shown in the right sub-figure of Fig. 4.2 (a). After processing column 6, there is no column which low position can be further changed. The reduction is finished and the matrix is fully reduced. Alg. 2 returns a reduced boundary matrix. In computing PH of the given simplicial complex, we only pay attention to the "pivots" of the reduced matrix. These pivots correspond to persistence pairs of persistent homology [70]. A pivot is defined as the entry (low(j), j) for any column j in the reduced matrix. In Fig. 4.2 (a), the pivots are (2,3), (1,4), and (5,6). The matrix reduction algorithms may generate different reduced boundary matrices for the same input, but the pivots are identical. We will call a column j of a partially reduced boundary matrix with (low(j),j) a pivot fully reduced.

### 4.2.1 Clearing and Compression

There are two effective optimizations for PH matrix reduction, clearing and compression. Clearing [68, 69] can set a column to zero. The original clearing lemma is stated as follows:

**Lemma 4.2.2.** *If (i,j) is a pivot (low(j), j) of a fully reduced column j, then column i can be zeroed. [69]*

The intuition behind the original clearing lemma (Lemma 4.2.2) is that every index is either a creator or destroyer index (each simplex either creates or destroys/zeros a homology class). Pivots (c,d) always have c, a creator index and d, a destroyer index. Thus if a column index is a creator index, it cannot have a pivot in its column and thus must be a zero column in the fully reduced matrix.

Lemma 4.2.2 has an extension as follows:

**Lemma 4.2.3.** *For any nonzero column j, not necessarily fully reduced, column low(j) can be zeroed. [68]*

We will use Lemma 4.2.3, the extension of Lemma 4.2.2 when referring to clearing. Clearing always zeroes the column low(j) on the left of column j of one dimension smaller. Fig. 4.2 (b) illustrates how the algorithm processes the boundary matrix with clearing. For applying clearing, the matrix reduction needs to process columns from higher dimension simplices to lower dimension simplices [69]. In Fig. 4.2 (b), the algorithm starts from the highest simplex, i.e., 2-simplex (column 6), and finds the pivot (5,6). With clearing (Lemma 4.2.3), the algorithm directly zeros column 5. After that, the algorithm processes 1-simplices (columns 3, 4, 5) from the left to the right, and finds column 4 can be reduced by adding column 3. The algorithm stops after processing the 0-simplices (columns 0, 1, 2), which are zero columns. Without clearing (Fig. 4.2 (a)), the algorithm calls the column addition twice, one on column 4 and the other on column 5. With clearing, the algorithm doesn't need the column addition on column 5.

Compression [68] is another technique to optimize PH matrix reduction, which can set a row to zero with the lemma as follows:

**Lemma 4.2.4.** *For any given pivot (i,j), row j can never have a pivot in it. Thus row j can be zeroed. [68]*

The reasoning behind compression is similar to clearing. However, compression zeros a row of index higher dimension instead of lower dimension. Fig. 4.2 (c) shows how the algorithm processes the boundary matrix with compression (Lemma 4.2.4). When the algorithm identifies (2,3) is a pivot, it zeros row 3. And after the column addition on row 4, the algorithm finds (1,4) is a pivot and then zeros row 4.

Clearing and compression are not fully utilized in existing PH software packages. First, clearing (lemma 4.2.3) can be applied in a column-wise parallel manner, without dimension ordering restriction, and without column addition dependency. This technique has not been used in existing software implementations, including in existing parallel algorithms. The sequential algorithm in Fig. 4.2(b), as mentioned in [69] is the way clearing is applied in spectral sequence, see 4.2.6, for example.

Second, a new study [71] introduces a new lemma that can be used for compression as follows:

**Lemma 4.2.5.** *If a column j has an entry (i,j) that is a leftmost nonzero in its row i, then column j must eventually have a pivot. [71]*

In Fig. 4.2(c), compression sets column 4 to zero after finding the pivot (1,4) with one column addition. However, Lemma 4.2.5 tells us the algorithm can directly zero row 4 without identifying the pivot, because of the leftmost non-zero entry (1,4)[1]. This lemma provides us with an opportunity to zero rows as early as possible and eliminate more unnecessary column additions. However, it has not been considered by existing PH software packages.

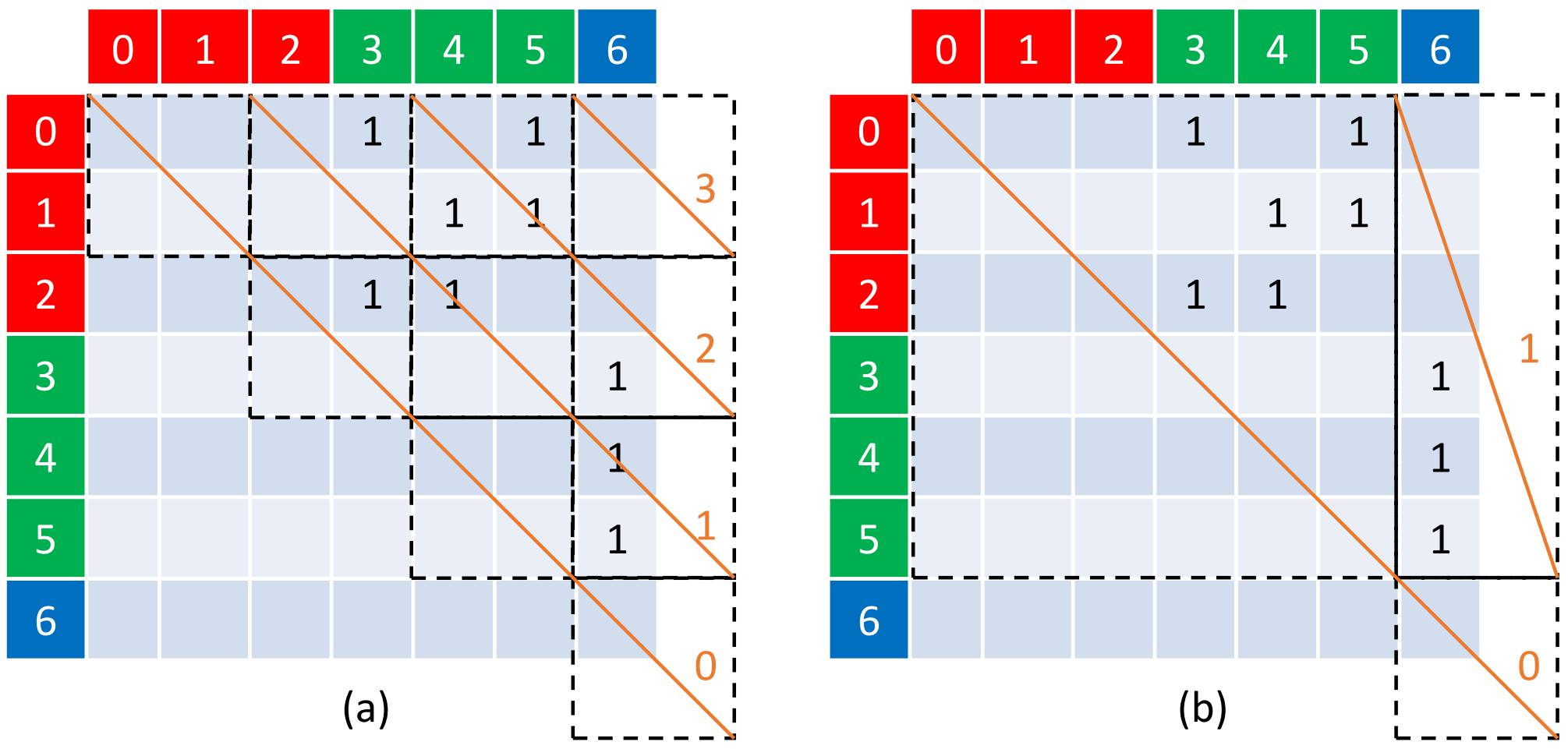

**Figure 4.3.** Two examples of Spectral Sequence-based PH matrix reduction on the boundary matrix in Fig. 4.1. The boundary matrix is divided into multiple tiles, and the tiles on the same diagonal can be processed in parallel. (a) uses 2 × 2 tiles, and four rounds are needed for the matrix reduction. (b) uses 6 × 6 tiles we only draw one 6 × 6 tile to save the limited page space., and two rounds are needed.

[1] ↑This is because no column addition can zero a leftmost non-zero entry, e.g., (1,4) in this case. The non-zero column (column 4) eventually has a pivot, and it can trigger compression due to Lemma 4.2.4.

### 4.2.6 Spectral Sequence Algorithm

Of the parallel algorithms for PH matrix reduction, Spectral Sequence [72] is considered as basic. Since the boundary matrix is the upper triangular matrix, if the N × N boundary matrix is divided into multiple M × M tiles/blocks (M < N), the tiles on the same diagonal can be processed in parallel. As shown in Fig. 4.3 (a), the boundary matrix is divided into ten 2 × 2 tiles. Because each tile only depends on those tiles on the left and below, Spectral Sequence only needs four rounds to finish the process. In each tile, the algorithm can process columns following the standard matrix reduction algorithm, with or without clearing and compression. If the low(j) of column j is found beyond the range of the current tile, the algorithm will continue processing column j with the upper tile in the next round.

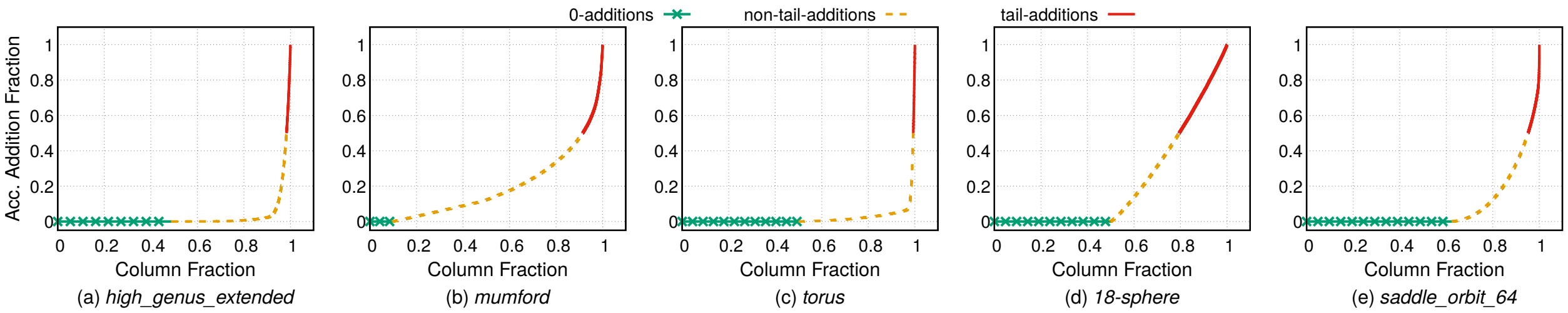

**Figure 4.4.** The column addition distribution in various datasets. *high_genus_extended* is from the PHAT benchmark datasets. *mumford* comes from a 4-skeleton of the Rips filtration with 50 random points from the Mumford dataset [73]. *torus* comes from an alpha shape filtration [74] defined on 10,000 points sampled from the torus embedded in $\mathbb{R}^3$. *18-sphere* comes from a synthesized simplicial complex of a 19-dimensional single simplex with its interior removed. *saddle_orbit_64* comes from a cubical complex generated from a 3D image using a $64^3$ sub-region.

Having inspired by the studies of wavefront loops [75–77], we identify several limitations for Spectral Sequence-based PH matrix reduction. First, the number of tiles that can be processed in parallel decreases from the first round to the last round, leading to load imbalance when scheduling one thread per tile as PHAT [59] does. Second, in processing PH boundary matrices, the load imbalance is also a concern. Because of the nature of sparsity and dependency, the numbers of column additions between tiles are highly skewed and cannot be determined in advance. Therefore, a new scheduling mechanism is needed for load balance. Most importantly, such a tile-based parallel pattern may limit the power of clearing and compression. Assuming the boundary matrix is divided into 6 × 6 tiles, as shown in Fig. 4.3 (b), there are three tiles in total. Two tiles are processed in the first round and one in the second round. The algorithm can only set column 5 to zero by applying clearing in the tile processed in the second round; while at that moment, column 5 has been set to zero in the first round by calling column addition. In this case, the power of clearing is not utilized. Although PHAT [59] can mitigate it by processing each tile multiple times and one dimension at each time (from higher to lower) for enabling clearing, this method aggravates the load imbalance with additional memory access overhead. Therefore, a new parallel framework is needed to unleash the power of clearing too.

### 4.2.7 Anatomy of PH matrix reduction

**Table 4.1.** The number of columns of various data sets and their addition distribution.

| Dataset | # of cols. | 0-adds. (% of cols.) | non-tail-adds. (% of cols.) | tail-adds. (% of cols.) |
|---|---|---|---|---|
| high_genus_extended | $2.76 \times 10^6$ | 48.78 | 49.52 | 1.70 |
| mumford | $2.37 \times 10^6$ | 9.57 | 81.85 | 8.58 |
| torus | $5.58 \times 10^5$ | 50.03 | 49.27 | 0.70 |
| 18-sphere | $1.05 \times 10^6$ | 50.00 | 29.40 | 20.60 |
| saddle_orbit_64 | $2.05 \times 10^6$ | 62.47 | 32.70 | 4.83 |

The running time of a PH matrix reduction is dominated by column additions. To better understand its computational nature, we analyze the column-wise addition distribution for various datasets in the standard algorithm. We count the number of additions for all columns, and with a thorough analysis, we want to understand column addition patterns throughout the matrix.

We look into the *torus* dataset from the PHAT benchmark datasets [59] as an example. After the PH matrix representing the torus is fully reduced and all statistical numbers have been collected, we sort the columns according their column-wise XORs and accumulate the number of XORS (column additions). The result is presented in Fig. 4.4(c), where the x-axis is the column fraction and the y-axis is the accumulated addition fraction. The curve with marks represents the columns unchanged throughout the calculation (column fraction from 0 to 0.5). The solid curve represents the top columns (by column addition count) taking up 50% of all additions; the dashed curve represents the remaining columns. From the experiment, two facts are observed when computing persistence pairs for this torus: 1) about half columns are inherently stable, and 2) half of all additions are performed over a very small portion of columns. Thus, **we define tail-addition columns to be the smallest set of columns that account for more than 50% of the column additions needed to reduce the boundary matrix**. Furthermore, we define 0-addition columns to be columns that do not require column additions. We thus have a partition of three types of columns as *0-addition* columns, *non-tail-addition* columns, and *tail-addition* columns, respectively. The tail phenomenon is an example of a Pareto principle [78] (e.g. 50% of total column additions are due to 1.7% of the columns). The tail phenomenon is also observed in daily Google production systems, where high latency service to a small percentage of customers could dominate overall service performance at large scale [79].

To investigate if the observed characteristics are general, the experiments are extended to other datasets. Some of the datasets are also from the PHAT benchmark datasets and the others are topologically synthesized. The statistical results are also shown in Figure 4.4 (a), (b), (d), and (e) and the numbers are presented in Table 4.1. We observe that for most datasets, there are ~ 50% 0-addition columns (except that *mumford* presented in Figure 4.4 (b) contains only 9.57% 0-addition columns). Furthermore, half of all additions concentrate on at most ~20% columns. In an extreme case, the tail-additions happen on only 0.7% columns (*torus*).

The observed facts shed light upon the structural differences of PH matrix additions. The large portion of 0-addition columns can be massively processed in SIMT mode on GPU. A small percentage of tail-addition columns indicates that sequential algorithms on a powerful single core are still attractive. For the remaining non-tail-additions, MIMD processing on multi-core can be very efficient. Our anatomy study has motivated us to separate additions of various types of columns as a foundation to achieve high performance.

**Table 4.2.** The topological origin of tail-adds. columns, showing the percentage of tail-adds. columns that are creator columns

| Dataset | # of tail-adds. cols. | (% of tail-adds. cols. that are creator) |
|---|---|---|
| high_genus_extended | $4.69 \times 10^4$ | 90.98 |
| mumford | $2.03 \times 10^5$ | 100.00 |
| torus | $3.92 \times 10^3$ | 99.97 |
| 18-sphere | $2.16 \times 10^5$ | 100.00 |
| saddle_orbit_64 | $9.89 \times 10^4$ | 99.98 |

### Topological origins of tail-addition columns

The computing for tail-addition columns is the bottleneck to PH matrix reduction. We find that most tail-addition columns (see table 4.2) are creator columns (columns that are zero when fully reduced and topologically correspond to simplices that generate cycles, e.g. column 5 in Figure 4.1). This is because the columns that usually take the most time to reduce are columns that need to be completely zeroed, requiring many column additions. This also explains the power of the clearing lemma, since it can be used to zero all paired creator columns.

This can also be explained directly through linear algebra:

For the primal version of the matrix reduction problem, the kernel of $[\partial_{p+1}]$ must be computed for each $p \geq 0$. According to Proposition 3.2.24, this requires traversing the $p$-dimensional cycles to determine a cyclic connected $p$-dimensional neighborhood of Definition 3.2.23. This allows us to prove the following proposition on the complexity of computing persistent homology in the primal case:

**Proposition 4.2.8.** *Assume $n_p \geq n_{p-1}, \forall p = 1, ..., dim(K_n)$.*

*The time complexity, denoted by $Time(K_n)$, for computing the primal matrix reduction problem of Problem 3.2.1 is:*

$$\Omega\left(\sum_{p=1}^{dim(K_n)} n_p \log(n_0)\right) \leq Time(K_n) \leq O\left(\sum_{p=1}^{dim(K_n)} n_p n_{p-1} \log(n_0)\right) \tag{4.1}$$

*Proof.* For each $p \geq 1$, we can write out the columns of each boundary matrix $[\partial_p]$ into those that correspond to creation times and those that correspond to destruction times. According to Observation 3.2.14, we know that each column can only correspond exclusively to one of a creation or a destruction time.

The columns that cannot be zeroed by the minimax Equation of Problem 3.2.1 correspond to destruction times. There are $\text{rank}([\partial_p])$ many such columns. The remaining columns correspond to creation times, and there are: $(n_p - \text{rank}([\partial_p]))$ many of them.

We analyze the time to compute the creator and destructor columns individually. We show that in total the reduction over all creator and destructor columns takes time

$O(n_p n_{p-1} \log(n_0))$ with lower bound of $\Omega(n_p \log(n_0))$.

- For creation: A creator column is determined by the linearly independent columns in its causal submatrix. This has an upper bound of $\text{rank}(\partial_p)$ and a lower bound of $\Omega(p+2)$. There are $\dim(\ker(\partial_p))$ many creation columns. Each column addition requires $\Theta(\log(n_0))$ time to compute (e.g. with the bit-tree-pivot column [59] or the fibonocci heap [80] data structure), thus the total complexity over all creation columns is $O(\dim(\ker(\partial_p))\text{rank}(\partial_p)\log(n_0))$ and $\Omega(\dim(\ker(\partial_p))\log(n_0))$.

- For destruction: A column is a destructor column if it does not belong to the kernel of $\partial_p$. Each such column must add with at most $\text{rank}(\partial_p)$ many columns and atleast $\Omega(1)$ columns, thus we have an upper bound of $O(\text{rank}(\partial_p)^2 \log(n_0))$ and lower bound of $\Omega(\text{rank}(\partial_p))$.

We can use the rank-nullity theorem when adding the complexities for the two cases of creator and destructor. Certainly we have the asymptotic upper bound on the computational complexity $\text{Time}(K_n)$:

$$\text{Time}(K_n) = O(\sum_{p=1}^{\dim(K_n)} n_p n_{p-1} \log(n_0)) \tag{4.2}$$

For the lower bound, once again by rank-nullity, we obtain

$$\text{Time}(K_n) = \Omega(\sum_{p=1}^{\dim(K_n)} n_p \log(n_0)) \tag{4.3}$$

Thus:

$$\Omega(\sum_{p=1}^{\dim(K_n)} n_p \log(n_0)) \leq \text{Time}(K_n) \leq O(\sum_{p=1}^{\dim(K_n)} n_p n_{p-1} \log(n_0)) \tag{4.4}$$

□

Besides using the clearing lemma, tail-addition columns can be handled by employing compression (Sec. 4.3.3), introducing parallelism (Sec. 4.3.6), computing cohomology for special cases (Sec. 4.3.3), or using efficient data structures such as bit-tree-pivot column from PHAT[59] (lowering instruction counts). We are able to employ all of these techniques in HYPHA.

## 4.3 The HYPHA Framework

In this section, we introduce our framework, HYPHA, a HYbrid Persistent Homology matrix reduction Accelerator. We first present an overview of HYPHA by comparing it with existing work, and then we go over each phase of HYPHA in detail.

### 4.3.1 Overview

Fig. 4.5 compares HYPHA with existing work in the format of a finite state machine. Fig. 4.5 (a) introduces the framework in existing PH software packages, which is algorithm independent, being either sequential or parallel. The framework starts from the column additions of standard algorithm to update the boundary matrix and the lookup table. Once the lookup table is updated, clearing and/or compression is triggered to zero corresponding columns and rows in the boundary matrix. This process stops at the column addition phase when there is no column that can be changed.

In contrast, the HYPHA framework includes three phases involving the states in Fig. 4.5 (b). The GPU-scan is the start phase to find 0-addition columns and leftmost **1**s. Due to the SIMT execution pattern in this phase, we implement it on GPU in a highly efficient manner. Following that is the clearing and compression phase on multicore CPU to eliminate rows and columns labeled by GPU on multicore. Different with the existing framework that starts from the column additions of the standard algorithm in order to trigger clearing and compression, our framework identifies indices to clear and compress immediately from GPU scan for leveraging its results. With the identified leftmost **1**s and the 0-addition columns, the clearing and compression phase can potentially eliminate much more unnecessary additions (details in Sec. 4.3.3). The third phase is the matrix reduction phase. This reduction of the submatrix is determined after the clearing and compression phase. In this phase, we can enhance the parallel spectral sequence algorithm on multicore with a multi-level scheduling mechanism for load balance (details in Sec. 4.3.6).

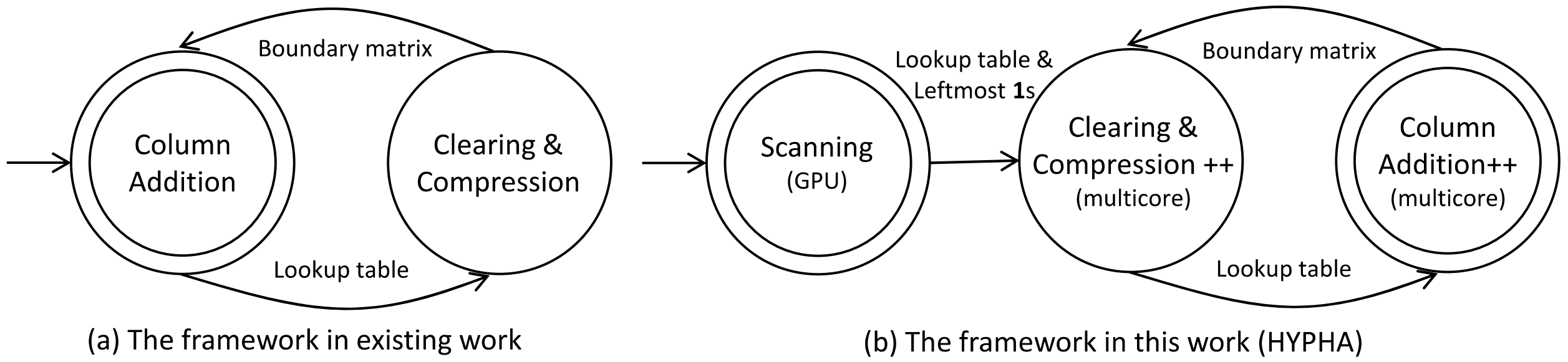

**Figure 4.5.** The frameworks of existing work and HYPHA.

### 4.3.2 GPU-scan Phase

By observing the standard algorithm Alg. 2, we determine that a column is a 0-addition column if its lowest **1** is the leftmost **1** in its row, since there is no column to the column's left that can be added to it. Furthermore, a leftmost **1** implies its column cannot be zeroed, even if that column is reducible in the algorithm (Lemma 4.2.5). This provides us with an opportunity to apply compression on multicore later. Searching the leftmost **1**s is challenging. Launching individual threads for each row and recording the first-met **1** with multithreading on CPU are hardly expected to be efficient since the scan operation itself is simple yet the scale is massive. Compared to millions of columns in a boundary matrix, commonly only tens of CPU cores can be found in a machine. Such a SIMT execution model [81] motivates us to employ GPU accomplishing the task, on which the number of cores is two orders of magnitude higher.

**Algorithm 3:** GPU-scan algorithm

```
Input: ∂
Output: Left, Lookup, u
Left ← {∞} ;
Lookup ← {−1} ;
stable ← {0} ;
u ← ∅ ;
n ← ∂.num_cols ;
dim3 blks(BLK_SIZE, 1, 1) ;
dim3 grds(ceil_div(n, BLK_SIZE), 1, 1) ;
set_leftmost <<<grds,blks>>>(∂, Left, stable) ;
set_lookup <<<grds,blks>>>(∂, Left, Lookup, stable) ;
set_unstable <<<grds,blks>>>(stable, u) ;
return (Left, Lookup, u)
```

It is worth noting that due to sparsity, a boundary matrix is usually stored in a CSC format. To efficiently find out leftmost **1**s at each row over CSC, the GPU-scan algorithm in Alg. 3 includes three steps. In our algorithm, a column that is deemed fully reduced by GPU-scan is defined as a *stable column*; otherwise, it is an *unstable column*. Thus, stable columns are 0-addition columns and columns with all zeros are stable (including columns zeroed by clearing). In Alg. 3, we use **stable** to mark the columns identified as stable and

**Algorithm 4:** Finding the Pivots

```
function __global__ set_leftmost(∂, Left, stable):
    Data: gid: global thread idx
    if gid < n then
        if ∂[gid].length = 0 then
            stable[gid] ← 1 ;
        else
            for rid ← 0 to ∂[gid].length − 1 do
                Left[rid] ← atomicMin(Left[rid], gid) ;
            end
        end
    end
```

the unstable column indices are aggregated in **u**. Alg. 3 also sets up two arrays for the following phase, i.e., **Left** that stores the column indices of leftmost **1**s, indexed by row, and **Lookup** as the lookup table that records the pivots of stable columns.

**Algorithm 5:** Collecting Indices

```
Function __global__ set_lookup(∂, Left, Lookup, stable):
    Data: low(·) comes with ∂
    if gid < n then
        if low(gid) ≠ −1 and Left[low(gid)] = gid then
            Lookup[low(gid)] ← gid ;
            stable[gid] ← 1 ;
            stable[low(gid)] ← 1 ;
    end
```

Alg. 3 initializes global data structures on the GPU side, moves the boundary matrix from CPU to GPU, and launches the kernels one by one. The `set_leftmost` kernel is responsible for searching the leftmost **1**s in the boundary matrix and writing them into **Left**. At this step, the zero columns are also identified and their ids are put into **stable**. The `set_lookup` kernel setups **Lookup**: if the lowest **1** of a column is the leftmost **1** at that row, the entry is a pivot and is put into **Lookup**. We also apply the clearing lemma to zero the low(gid), as shown in Lines 25-29. The last step of `set_unstable` excludes stable columns identified in the prior steps and prepares the unstable column array, i.e., **u**, for the following phases.

**Algorithm 6:** Collect Unstable Column Indices

```
Function __global__ set_unstable(stable, u):
    if gid < n then
        if stable[gid] = 0 then
            u ← u ∪_atomic {gid} ;
    end
```

Fig. 4.6 shows how our algorithm works over the example matrix in Fig. 4.1 (b), in which the numbers in red color are the updated values in this phase. Our algorithm collects three kinds of metadata used for accelerating the following column additions. First, the leftmost **1**s indicate which rows could be zeroed before the calculation. Second, the lookup table keeps track of pivots. Third, we keep track of unstable columns in **u**. The lookup

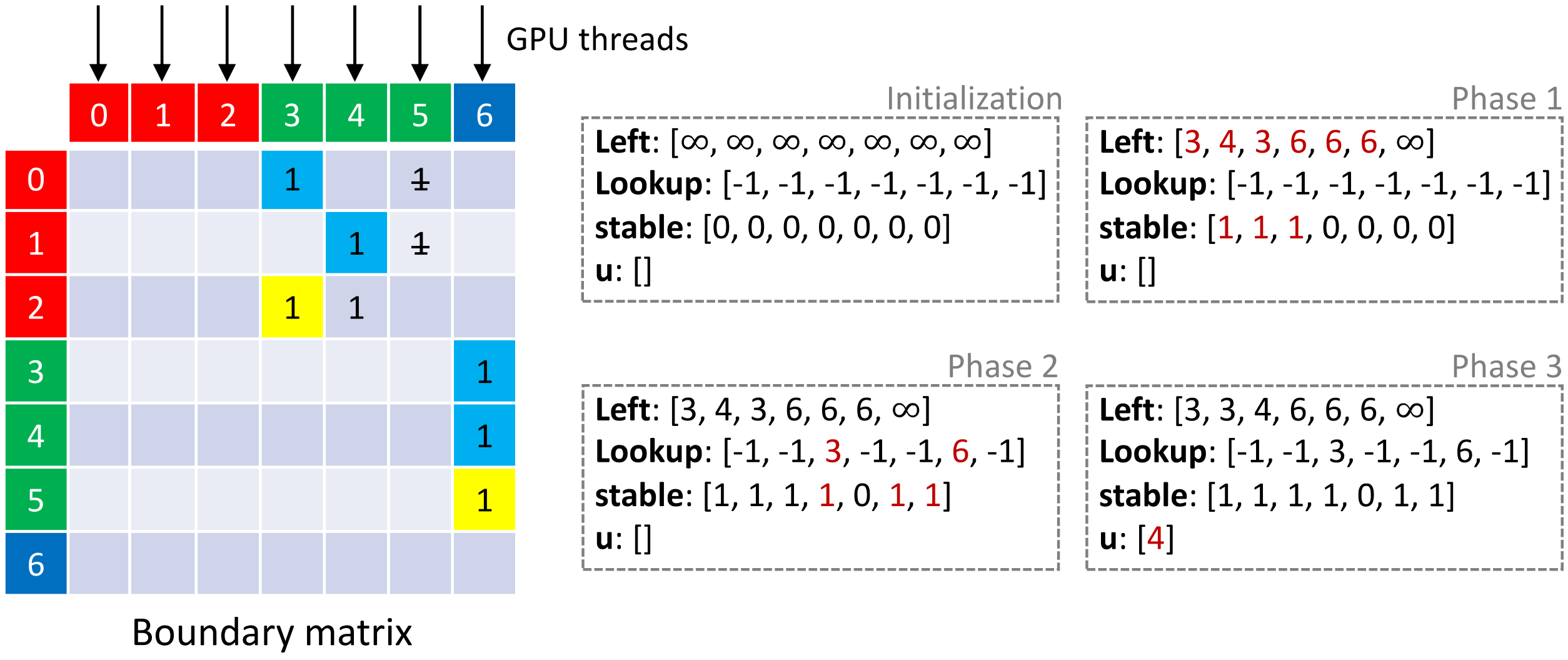

**Figure 4.6.** GPU scans an example boundary matrix.

table itself can be used to speed up the column addition by applying multiple additions in parallel. Additionally, knowing which columns are unstable can reduce the number of memory accesses in the following phases, by performing the iterations over the unstable columns instead of the whole boundary matrix.

**Data transmission**: Employing GPU to scan the boundary matrix introduces additional data transmissions. The complete boundary matrix has to be moved to GPU before the kernel launch, and the scanning results need to be sent back as well. We hide the CPU-GPU data transmission in the file system I/O. When the program reads the boundary matrix from the disk to the main memory, an individual thread is launched to migrate the progressively read data blocks to GPU. With this technique, copying the boundary matrix incurs little overhead. Compared to the boundary matrix, the scanning results are much smaller. The insignificant overhead of GPU to CPU data transmission can be ignored.

### 4.3.3 Clearing and Compression Phase

The key idea of both clearing and compression is that if a pivot (i,j) has been found, persistence pairs like $(x,\mathrm{i})$ and $(\mathrm{j},x)$ will never exist, where $x$ could be any index other than i and j. As a result, discovering a pivot $(i, j)$ means we can safely zero column i (clearing) and row j (compression).

**Algorithm 7:** Clearing (optional if no need for a reduced matrix: if only pivots in the lookup table are desired)

```
Input: R
for cid ← 0 to R.num_cols − 1 do
    if low(cid) ≠ −1 then
        R[low(cid)] ← 0 ;
    end
end
```

The metadata collected in the GPU-scan phase provides us with an opportunity to apply both techniques for preprocessing the boundary matrix before the final matrix reduction phase on multicore. The pivots in the lookup table and the positions of leftmost **1**s spreads in all dimensions. Although we have already identified the columns that we can zero by GPU, we must write the results to a boundary matrix on CPU side (recall the GPU-scan does not write to the matrix). Thus we first apply clearing in parallel on multicore to affect the boundary matrix at little extra cost. Define a submatrix as a matrix restricted to a subset of rows and columns. Knowing the stable columns, this results in an $n \times (n - s)$ submatrix to reduce where n is the number of columns and s is the number of stable columns. Then we apply compression to it, further eliminating d+s' rows where d is the number of unique finite entries in **Left**, and s' is the number of nonzero stable columns.

Every column of the form low(j) for any column j can be zeroed by Lemma 4.2.3. Any elements in the **Left** other than $\infty$ indicates the rows to be zeroed, e.g., we can set all entries at row 3 to 0 if $\mathbf{Left}[2] = 3$, where 2 is arbitrary. Furthermore, if the compression changes a stable column into the state containing only the lowest **1** but all zeros at the other rows, we can safely set all entries to that lowest **1**'s right as zero. This technique has the same effect as a column addition from stable columns to unstable columns without actually performing the column addition. During compression, we utilize all meta data of destroyer indices (columns that must eventually have a pivot) and known pivots from GPU-scan by clearing out all compressible rows while adding stable columns to all unstable columns, zeroing out the right

**Algorithm 8:** Compression

```
Input: R, Left, Lookup, u
Output: Compressed R
C ← {UNKNOWN} ;
FIND-COMPRESSIBLE(R, C, Left, Lookup, u) ;
APPLY-COMPRESSION(R, C, Left, Lookup, u) ;
Procedure APPLY-COMPRESSION(R, C, Left, Lookup, u):
    foreach cid ∈ u do
        foreach rid ∈ R[cid] in decreasing order do
            pivotcol ← Lookup[rid] ;
            if pivotcol = −1 then
                if C[rid] = COMPRESSIBLE then
                    R[cid][rid] ← 0 ;
            else if pivotcol < cid then
                if C[rid] = COMPRESSIBLE then
                    R[cid][rid] ← 0 ;
                else
                    R[cid] ← R[cid] + R[pivotcol] ;
                end
            end
        end
    end
```

of a stable pivot. Our implementation is based on [68] but involves a memoized row-based depth first search for compressible indices and uses Lemma 4.2.5.

The compression algorithm we employ is composed of two stages. FIND-COMPRESSIBLE(·) checks entries of the unstable columns and marks which ones are COMPRESSIBLE. An entry will be compressed if 1) its row index is a column index containing any leftmost **1**, or 2) there is a stable column to its left having all nonzero entries above it compressible. With these two conditions, more indices in addition to the ones identified by the columns with leftmost **1**s are compressible. So we introduce a temporary array **C** to record these compressible row indices. APPLY-COMPRESSION(·) is the procedure which actually zeros the compressible rows (excluding pivots) and zeros all entries to the right of (low(j),j), for j a stable column found by GPU and low(j) an incompressible row, by column addition. Both stages of our compression algorithms are implemented with multithreading. Despite clearing

**Algorithm 9:** Find compressible indices

```
Input: R, C, Left, Lookup, u
Output: Updates C
/* The array C records if a row can be compressed */
foreach cid ∈ u do
    foreach rid ∈ R[cid] do
        if SEARCH(rid) then
            return true
        end
    end
end
Function SEARCH(rid):
    if C[rid] = COMPRESSIBLE then
        return true
    else if C[rid] = INCOMPRESSIBLE then
        return false
    if rid ∈ Left then
        C[rid] ← COMPRESSIBLE ;
        return true
    else if Lookup[rid] > −1 then
        foreach k ∈ R[Lookup[rid]] excluding rid do
            if SEARCH(k) = false then
                C[rid] ← INCOMPRESSIBLE ;
                return false
            end
        end
        C[rid] ← COMPRESSIBLE ;
        return true
    else
        C[rid] ← INCOMPRESSIBLE ;
        return false
    end
```

and compression having been widely adopted, to the best of our knowledge, we are the first employing both optimizations before the matrix reduction phase. This may minimize the additions in later computation, especially for the tail-addition columns.

To check the effects of clearing and compression in HYPHA, we count the column additions in our algorithm, CHUNK [68] and TWIST [69]. Notice that TWIST and spectral sequence in PHAT end up executing the same column additions, with SS in parallel and TWIST sequentially. Furthermore, in TWIST only clearing is applied while in CHUNK both optimizations are adopted. Our experimental results over three datasets are presented in Figure 4.7, where the x-axes are column fraction and the y-axes are accumulated column additions. For the *high_genus_extended* dataset, we can observe that HYPHA significantly lowers the scale of tail-addition columns and has the least total column addition number (49.65M total column additions). For the *mumford* dataset, HYPHA and CHUNK both removes a large portion of column additions compared to the TWIST algorithm, where HYPHA requires 10.73M column additions, CHUNK requires 8.60M ones and TWIST requires 35.66M ones before the boundary matrix has been fully reduced. For the *18-sphere* dataset, HYPHA eliminates all column additions via clearing and compression while TWIST has 19 column additions. Although CHUNK applies both optimization techniques, it still has 74.99K column additions. From the experiments, we can observe that HYPHA is more likely to effectively remove column additions compared to the existing algorithms, especially for the tail-addition columns.

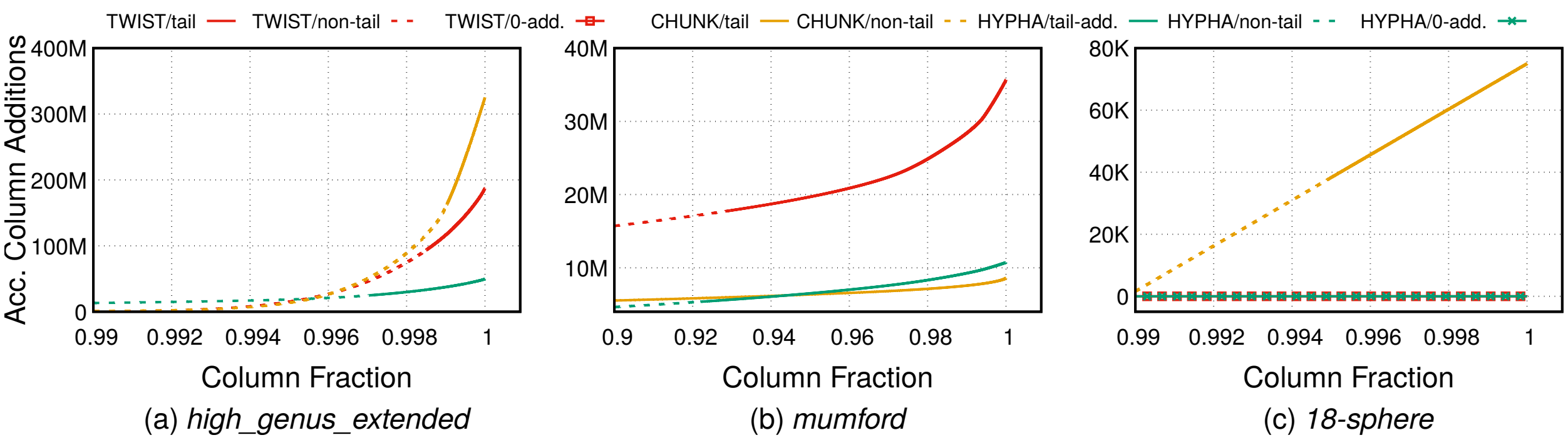

**Figure 4.7.** The accumulated number of column additions in various datasets when using HYPHA, CHUNK and TWIST. SS and TWIST have the same column addition distribution.

### Computing Cohomology

According to Theorem 3.2.28, it is possible to compute the dual matrix reduction problem on each column of a coboundary matrix. We can then show that this is equivalent to computing the standard matrix reduction algorithm of Algorithm 2 on the **anti-transposed boundary matrix**.

**Definition 4.3.4.** *The anti-transposed boundary matrix of a simplicial complex $K$ is the following matrix derived from the boundary matrix $[\partial]$ of Equation 2.6.7:*

$$[\partial]_{anti}^{T} \triangleq [[\partial_0]_{anti}^{T} \| \ldots \| [\partial_{dim(K)}]_{anti}^{T}] \tag{4.5}$$

*where $[\partial_p]_{anti}^{T}$ is the transpose of the p-boundary matrix with jointly reversed row and column indices, meaning that filtration indices reverse as follows:*

$$\hat{\mathrm{i}} : \mathrm{i} \mapsto n - \mathrm{i} + 1 \tag{4.6}$$

*We call $\hat{\mathrm{i}} = n - \mathrm{i} + 1$ the "time reversal" of index* $\mathrm{i}$.

If we reverse a reversed time $\hat{\mathrm{i}}$, then we get back the original index: $\mathrm{i} = \hat{\hat{\mathrm{i}}}$. Thus time reversal is a form of reflection symmetry.

**Proposition 4.3.5.** *For a simplex-wise filtration $Q(\boldsymbol{Simp}) : \varnothing \subseteq K_1 \subseteq \ldots \subseteq K_n$, Algorithm 2 applied to the anti-transposed boundary matrix $[\partial]_{anti}^{T}$ as input computes a matrix $R$ that has nonzero pivots at the pair of row-column indices $(n - \mathrm{i} + 1, n - low(R[\mathrm{i}]) + 1), \forall \mathrm{i} \in [n]$, which correspond to the birth-death times for persistent homology.*

*Proof.* The anti-transposed boundary matrix $[\partial]_{anti}^{T}$ reverses the ordering of the filtration $Q(\mathbf{Simp}) : K_1 \subseteq \ldots \subseteq K_n$ by reversing the row and column indices on each $[\partial_p]^T$ to form $[\partial_p]_{anti}^{T}$.

For each $p \geq 1$ in increasing order of $p$, if the primal Problem 3.2.1 on each column $[\partial_p^{\perp}(\sigma_{\hat{\mathrm{i}}})]$ at reversed time index $\hat{\mathrm{i}}$ of the input matrix $[\partial_p]_{anti}^{T}$, has the solution $\hat{k}^{*}$, then the dual Problem 3.2.2 on column $[\partial_p^{\perp}(\sigma_{\mathrm{i}})]$ at column index $\mathrm{i} = n - \hat{\mathrm{i}} + 1$ of $[\partial_p]^T$ has solution $k^* = n - \hat{k}^* + 1$. This is because a minimax problem is equivalent to a maximin problem when the indices flip sign.

Let $R$ be the fully reduced matrix computed by the standard matrix reduction algorithm, Algorithm 2, on any input matrix. By Proposition 3.2.19, this standard matrix reduction algorithm, Algorithm 2, finds the same $\text{low}(R[\text{i}])$ for every $\text{i} \in [n]$ as $\text{low}(r)$ for every column of the same column index i for the same matrix input to Algorithm 1. Letting this matrix input be $[\partial]^T_{anti}$, we can thus apply the standard matrix reduction algorithm, Algorithm 2, to solve the primal Problem 3.2.1 on $[\partial]^T_{anti}$. This is the dual Problem 3.2.2 up to the time reversal transformation on the indices.

According to Theorem 3.2.28, we have that the times considered by the maximin dual Problem 3.2.2 indexes a $p$-dimensional creation time and the $k^*$ solution is a $p+1$-dimensional destruction time. Since the $p$-dimensional creation times correspond to the columns $[\partial^{\perp}_{p+1}(\sigma_{\text{i}})]$. This means in the standard matrix reduction Algorithm 2 that a column $[\partial^{\perp}_{p+1}(\sigma_{\hat{\text{i}}})]$ at column index $\hat{\text{i}}$ of $[\partial(K_n)]^T_{anti}$ forms a nonzero fully reduced pivot entry at index $(\hat{k^*}, \hat{\text{i}})$ in its fully reduced form as matrix $R$ when the solution $\hat{k}^*$ has $\hat{k}^* > 0$. Rewriting the pair of times of the pivot of $R^{\perp}_{anti}$ into the original times of the filtration through a reversal of the time reversal, we get that $(n - \text{i} + 1, n - k^* + 1)$ are the creation and destruction times. □

Computing persistent cohomology [82, 83] for persistence pairs can potentially result in speedup over directly computing persistent homology with the boundary matrix. This is due to a change in the set of creator columns, replacing the original distribution of tail columns of the boundary matrix. In matrix terms, matrix reduction involves computing persistence with columns representing the coboundaries instead of the boundaries. In PHAT this is performed by constructing the anti-transposed matrix [59, 83] and performing any equivalent PH matrix reduction algorithm on the anti-transposed matrix. The clearing lemma can thus be applied to reducing an anti-transposed matrix. Notice how applying clearing, Lemma 4.2.3, on the anti-transposed boundary matrix is closely related to Lemma 4.2.5. In HYPHA, we are able to perform matrix reduction on antitransposed boundary matrices to still find the original persistence pairs via index transformation after reducing the matrix.

### 4.3.6 Final Phase

For the matrix reduction stage on the extracted submatrix, we choose either a sequential or parallel algorithm. Parallel algorithms do not necessarily outperform sequential ones, considering the high computational dependency among columns and the long computation chain for very few columns, i.e., the tail-addition columns. In our design, the sequential algorithm is derived from the standard algorithm with TWIST, and the parallel algorithm is based on the spectral sequence algorithm but with a multi-level scheduling for load balance. We call our parallel design as the spectral sequence plus (SS+) algorithm, which is designed to handle imbalance column additions to improve the classical spectral sequence (SS) algorithm.

In the SS algorithm, we observe that column additions usually concentrate in a small portion of tiles. Scheduling one thread to process one tile (tile-based scheduling) in the SS algorithm results in imbalance workloads arranged to threads. Therefore, we design our SS+ algorithm as follows: at the beginning of each round of processing tiles along a diagonal, as long as we find the unstable columns locate in only a small portion of tiles and the unstable column number is much higher than the working thread number, we schedule working threads to process unstable columns equally (column-based scheduling); otherwise, we schedule one thread for one tile. By dynamically switching the scheduling between tile- and column-based, SS+ further improves the performance of matrix reduction.

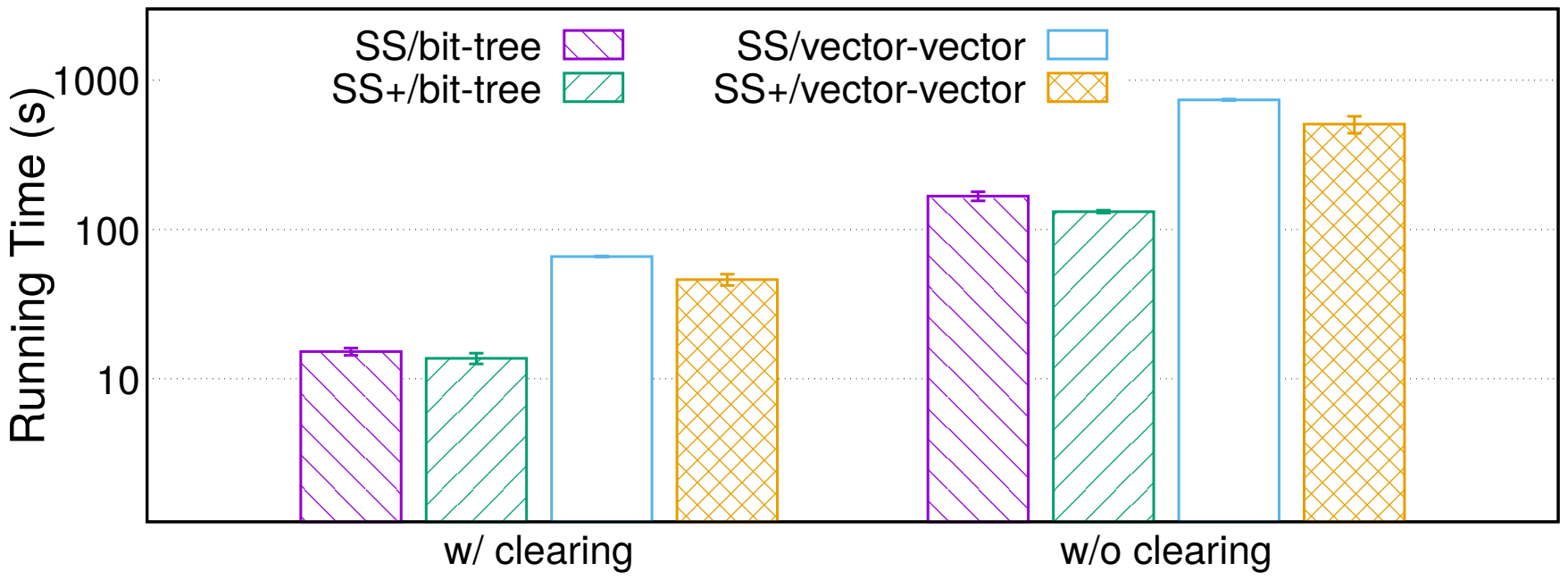

**Figure 4.8.** Running time in seconds of SS+ and SS algorithms over dataset *high_genus_extended*, w/ and w/o clearing, using bit-tree or vector-of-vector as the underlying data structure.

We implement our SS+ algorithm and compare it to the SS algorithm from PHAT. We use two types of underlying data structures to store the boundary matrix, vector-of-vector or bit-tree. Both of them are implemented in PHAT. For vector-of-vector, a column is represented as a vector of C++ and all columns are stored as the elements of another vector. The bit-tree-based method uses the vector-of-vector to store the boundary matrix, but an individual column is transformed to a bit tree [64] when performing the column additions. We also test the performance of two algorithms with and without the clearing technique. The experimental results on the real-world dataset *high_genus_extended* are presented in Fig. 4.8, where the y-axis is the running time in seconds. The system used for the experiment is presented in Section 4.4. The left and right group of data are the performance results with and without clearing, respectively. The pillars show the average performance and the error bars are the standard deviation. We can see that in all cases SS+ outperforms PHAT-SS algorithm. With clearing, SS+ completes the computation in 13.69s (bit-tree) and 46.15s (vector-of-vector), as SS requires on average 15.20s (bit-tree) and 65.81s (vector-of-vector), respectively. With clearing, SS+ can save 9.9% and 29.9% execution time. Without clearing, the average running time of SS+ are 131.58s (bit-tree) and 506.91s (vector-of-vector). These are lower than the ones of SS, which are 167.04s (bit-tree) and 738.19s (vector-of-vector). Without clearing, SS+ can save 21.1% and 31.3% execution time.

**Algorithm 10:** HYPHA

**Input:** $\partial$: an $n \times n$ matrix
**Output:** $\mathbf{R}$

$\mathbf{Left}, \mathbf{Lookup}, \mathbf{u} \leftarrow$ GPU-SCAN($\partial$) ; // after GPU-scan, transfer metadata from GPU to CPU
$\mathbf{R} \leftarrow \partial$ ; // $\mathbf{R}$ is on CPU side
CLEARING($\mathbf{R}$);
COMPRESSION($\mathbf{R}, \mathbf{Left}, \mathbf{Lookup}, \mathbf{u}$);
MATRIX_REDUCTION_ALGORITHM($\mathbf{R}, \mathbf{Lookup}, \mathbf{u}$) ; // we may employ the parallel SS+ as the final matrix reduction algorithm; Sequential algorithms like TWIST may be faster on certain datasets
**return** $\mathbf{R}$

Putting all the algorithms and mechanisms together, we have developed HYPHA (Alg. 10), which is an implementation of the framework in Fig. 4.5 (b). HYPHA starts from the GPU-scan to identify 0-addition columns and leftmost **1**s. With collected results, HYPHA immediately applies clearing and compression. In the final phase, column addition is executed on multicore, with a parallel mode or a sequential one. We use the SS+ or the original SS algorithm for parallel, and the twist algorithm for sequential. Other matrix reduction algorithms can also be embedded into the HYPHA framework to leverage the results of GPU-scan and enhanced clearing/compression.

**Table 4.3.** High-level comparisons of HYPHA with PHAT and DIPHA. Two check marks indicate that HYPHA has the enhanced compression, which eliminates more boundary matrix entries.

| PH Implementation | Sequential | Parallel | Partitioning | | Block | Column | Multi-node | GPU-scan | Clearing | Compression |
|---|---|---|---|---|---|---|---|---|---|---|
| | | | col. | 2d tile | based scheduling | | | | | |
| PHAT-TWIST | ✓ | | | | | | | | ✓ | |
| PHAT-SS | | ✓ | | ✓ | ✓ | | | | ✓ | |
| PHAT-CHUNK | | ✓ | ✓ | ✓ | ✓ | | | | ✓ | ✓ |
| DIPHA | | ✓ | | ✓ | ✓ | | ✓ | | ✓ | |
| HYPHA-TWIST | ✓ | | ✓ | | | | | ✓ | ✓ | ✓ ✓ |
| HYPHA-SS | | ✓ | ✓ | ✓ | ✓ | ✓ | | ✓ | ✓ | ✓ ✓ |

### 4.4 Experimental Results

The experiments are carried out on a HPC cluster, where each node is equipped with 2 Intel Xeon Gold 148 CPUs (40 cores in total), running at 2.4 GHz clock rate, with a 32K L1 cache, a 1024K L2 cache, and shared 28160K L3 cache. There is also a Tesla V100 GPU with 16 GB device DRAM installed in each node.

We compare HYPHA with state-of-the-art parallel software packages, including PHAT [84] and DIPHA [61]. PHAT provides different implementations of PH matrix reduction on a single node, while DIPHA is a distributed implementation for multiple nodes. We label these implementations as PHAT-TWIST, PHAT-SS, PHAT-CHUNK, and DIPHA, and ours as HYPHA-TWIST and HYPHA-SS for sequential and parallel, respectively. Tab. 4.3 shows a high-level comparison of functionalities. For the parallel ones, PHAT-CHUNK, PHAT-SS, DIPHA, and HYPHA-SS follow the 2D tile partitioning; and PHAT-CHUNK and HYPHA-SS can also partition data by columns. HYPHA-TWIST is marked with the column partitioning because of the parallel preprocessing steps, e.g., GPU-scan. The table also shows only HYPHA has the enhanced compression (denoted with two check marks), and only HYPHA-SS has different scheduling policies for load balancing, as discussed in Sec. 4.3.6.

### 4.4.1 GPU-scan Throughput in HYPHA

We first compare the throughput of stable column discovery (including those identified by the clearing lemma) of HYPHA GPU-scan with the throughput of identifying (sequentially scanning through) the set of 0-additions columns in TWIST (including those zeroed by clearing). Specifically, for both GPU-scan and TWIST we measure the number of discovered fully reduced columns divided by the time it takes to process them. This should be a fair comparison since GPU-scan discovers a very similar set of 0-additions columns as TWIST. We calculate the normalized throughput in Figure 4.9, where time for GPU-scan includes memory copy time and destroyer index discovery time. HYPHA GPU-scan shows its high efficiency. On the dataset *18-sphere*, we observe the highest improvement: HYPHA GPU-scan has a factor of 106.11x throughput improvement for identifying fully reduced columns. On the *18-sphere* all but 1 column (more than a 1 million columns) are labeled stable by HYPHA in milliseconds. Best utilizing the massive parallelism provided by GPU, we are able to boost the performance of our scan algorithm. It is certainly worth performing GPU-scan to find metadata of the input boundary matrix such as 0-additions columns. We will next measure full matrix reduction time to get a complete picture of the overall performance.

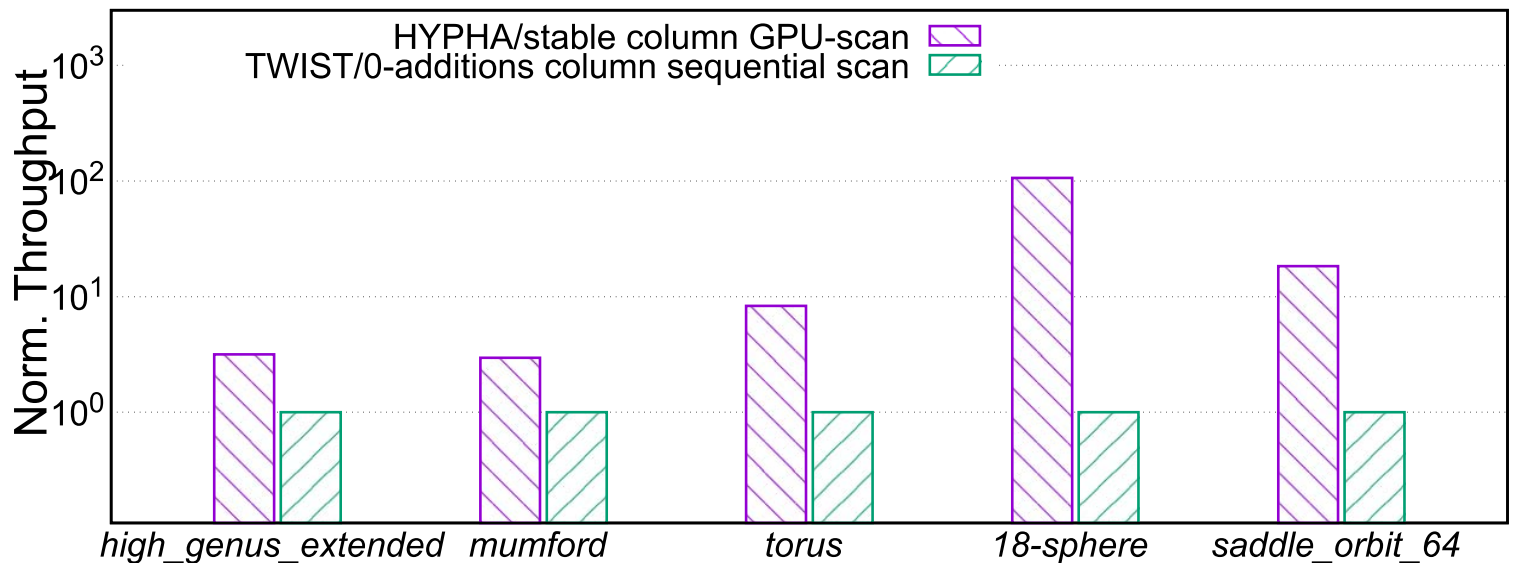

**Figure 4.9.** HYPHA GPU-scan vs. TWIST 0-additions sequential scan throughput, for fully reduced columns, normalized to the throughput of TWIST 0-additions column scanning.

### 4.4.2 Overall Performance Comparisons

We evaluate the overall performance of HYPHA by comparing with PHAT and DIPHA. Following the same experimental setup [61], we run DIPHA on up to 40 nodes, and launch one MPI process on one core of each node. In this case, the large cluster is much more expensive than our light facility of single node of 40 cores with GPU, and more importantly, DIPHA can utilize more cache space than others. In this experiment, we collect the computation time for various algorithms and measure speedup with respect to the standard PH matrix reduction algorithm (Alg. 2) in PHAT. The results are presented in Fig. 4.10, which shows the HYPHA framework achieves the best performance across all datasets. In *high_genus_extended*, *torus*, and *saddle_orbit_64*, HYPHA (HYPHA-TWIST or HYPHA-SS) can achieve 116.01x, 97.38x, and 86.52x speedups, respectively; which are almost two orders of magnitude. For *mumford* and *18-sphere*, the HYPHA framework can still speedup the PH matrix reduction by a factor of 5.01x and 14.16x over the standard algorithm. Among the algorithms of PHAT, the best one depends on the datasets. PHAT-TWIST outperforms the other two algorithms in *torus* (52.61x), *18-sphere* (5.96x) and *saddle_orbit_64* (37.32x) while PHAT-SS and PHAT-CHUNK are the best ones in *high_genus_extended* (53.24x) and *mumford* (4.27x), respectively. DIPHA is not necessarily faster than the standard algorithm implemented in PHAT for the datasets *mumford* and *18-sphere*, due to the overhead of MPI communication. For *high_genus_extended*, *torus*, and *saddle_orbit_64*, DIPHA running on 40 nodes achieves 16.11x, 33.02x, and 82.56x speedups. Overall, HYPHA outperforms the fastest algorithms of PHAT and DIPHA in various datasets by a factor of up to 2.38x (vs. PHAT-TWIST in *18-sphere*), 2.18x (vs. PHAT-SS in *high_genus_extended*), 1.85x (vs. PHAT-TWIST in *torus*), 1.17x (vs. PHAT-CHUNK in *mumford*), and 1.05x (vs. DIPHA-40nodes in *saddle_orbit_64*), respectively.

We profile an example case to understand the sources of the performance gain by HYPHA. Figure 4.11 presents a breakdown of running time of HYPHA-TWIST over the dataset *high_genus_extended*, normalized to PHAT-TWIST. In this experiment, we individually measure the running time for *Pre-processing + 0-additions*, *non-tail-additions* and *tail-additions*, which have been identified as typical computation tasks of PH algorithms in Section 4.2.7. By taking advantage of the powerful GPU scanning, HYPHA-TWIST takes 51.10% less time to processing 0-addition columns and other pre-processing operations. Furthermore, such pre-processing significantly alleviates the computation burden for the later steps. Figure 4.11 also shows that for non-tail-additions and tail-additions, 65.28% and 66.45% computations are removed, respectively.

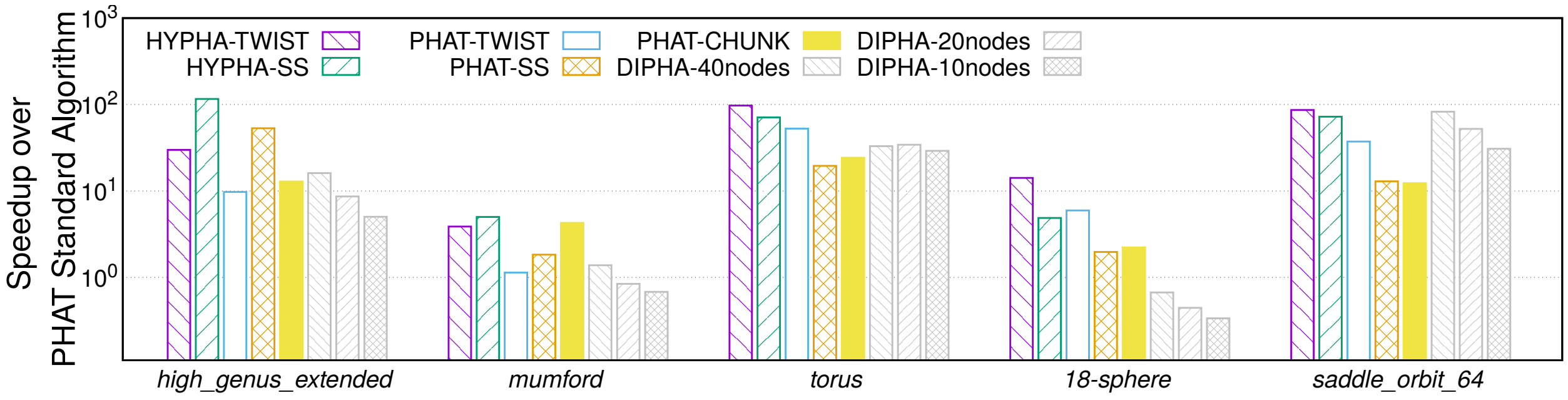

**Figure 4.10.** The speedups of various algorithms over the standard PH reduction algorithm implemented in PHAT.

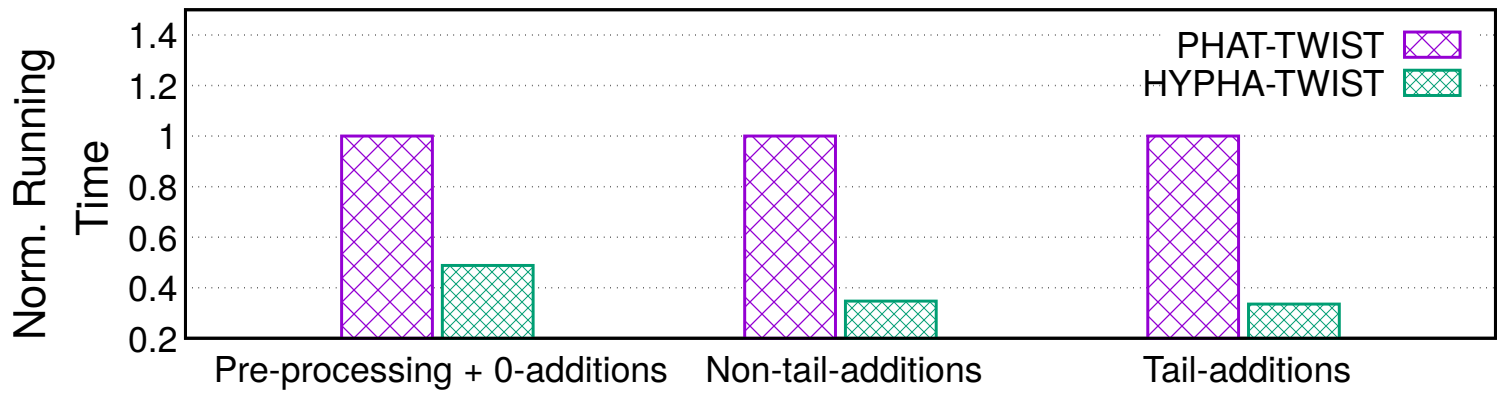

**Figure 4.11.** The time breakdown of HYPHA-TWIST running over the dataset *high_genus_extended*, normalized to PHAT-TWIST.

## 4.5 Discussion

Having made algorithmic and systems efforts in a holistic way, we show that the conventional "one-size-fits-all" approach does not often win. This is because to process PH matrix reduction on a MIMD parallel computer or on a powerful single core machine would not fully exploit rich but two different types of parallelisms exhibited in algorithms, and would not utilize advanced and hybrid devices of both GPU and multicore. Although parallel processing community has impressive accomplishments of solving challenging problems on GPU, an immediate question would be "why not execute the whole PH matrix reduction on GPU?".

There are several reasons for not recommending using GPU alone. First, as the analysis of Sec. 4.2.7 shows, the column addition patterns are highly irregular. For the tail-columns, the data dependency forces the additions to be sequential: only after adding one column and updating the lowest position of the tail-column, we can continue adding the next column until the last one. We have traced the additions on each column of dataset *torus*, and identified the highest number of sequential additions on a single column is ~62000. Compared with the execution performance of a matrix reduction on a single core of CPU, the performance on GPU is underperformed. Second, the column addition needs the runtime memory management to resize buffer, because the addition may add or delete non-zeros on columns. For *torus* that has at most 3 non-zeros in each column at the beginning, the algorithm changes the number of non-zeros of columns to near hundreds and to even tens of thousands at runtime. Although there are several GPU libraries for dynamic graphs and matrices [85–88] we can leverage, the overhead of buffer resize on GPU is still too high in PH matrix reduction. Therefore, HYPHA puts the scan phase on GPU and leaves the highly skewed, dependent, and dynamic column addition phase on multicore.

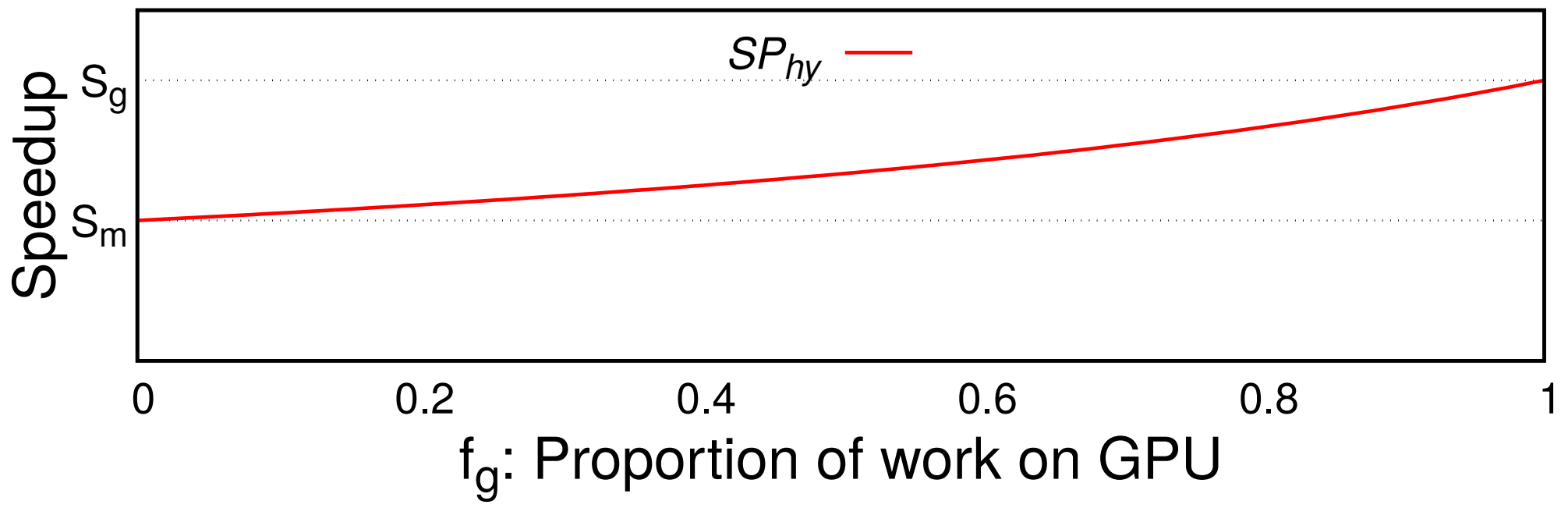

**Figure 4.12.** this illustrates Amdahl's law on a hybrid system's speedup by both GPU and multicore.

### 4.5.1 Separation of Parallelisms Under Amdahl's Law

Amdahl's law quantifies the speedup of a program with a fraction of work $(f)$ to be accelerated in parallel by a factor of $S$ as follows:

$$SP(f,S) = \frac{1}{(1-f)+\frac{f}{S}}$$

. Under this framework, for a hybrid system with GPU and multicore, the speedup is:

$$SP_{hy}(f_g, f_m, S_g, S_m) = \frac{1}{(1-f_g-f_m)+\frac{f_g}{S_g}+\frac{f_m}{S_m}}$$

, where $f_g, f_m$ are the fractions of work for GPU and multicore, respectively; $S_g$ and $S_m$ are the acceleration factors on the fractions of work $f_g, f_m$, respectively. Since our hybrid system only consists of two components, $f_g + f_m = 1$. Thus,

$$SP_{hy} = \frac{S_g \times S_m}{f_g \times S_m + f_m \times S_g}$$

. Fig. 4.12 plots the curve of $SP_{hy}$ that monotonically increases with respect to $f_g$ over [0,1] from $S_m$ to $S_g$. Notice as long as $S_m$ and $S_g$ are both > 1, then $SP_{hy} > 1$. Assuming $S_g > S_m$, the maximum speedup is $S_g$ when we are able to effectively execute an application only by GPU; and the minimum speedup is $S_m$ without involvement of GPU.

Although Fig. 4.12 quantifies execution of an application based on separation of SIMT and MIMD parallelisms conceptually by showing the trajectory of the performance improvement, it may not be able to fully characterize the HYPHA framework. The reason is as follows. Amdahl's Law models the SIMT acceleration $(S_g)$ and MIMD acceleration $(S_m)$ independently and the total performance improvement is proportional to the contributions from the two accelerations, namely in fractions of $f_g$ and $f_m$. In HYPHA, the SIMT and MIMD parallelisms are separated and the execution is independent on GPU and multicore, respectively. However, the GPU scanning is not only fast, but also makes a careful preparation to significantly improve the efficiency of the next two stage reductions. This type of communicative and collaborative computation (see Figure 4.11) may not be modeled by Amdahl's Law.

### 4.6 Related Work

There are several PH software packages to date. Our work focuses on reducing an arbitrary boundary matrix (PH matrix reduction). We selected PHAT and DIPHA to compare against since they involve state of the art efficient parallel algorithms for PH matrix reduction. We briefly overview several other relevant software efforts.

Javaplex [58] is a commonly used software for PH computation due to the breadth of tools it offers for its users. Javaplex is not state of the art in terms of computing performance. Dionysus [60] is a python interfaced C++ library for PH computation. It does not offer any parallelism that we know of. It is faster than Javaplex. It is known to be slower than PHAT and thus we do not compare with it. PHAT [59] is a CPU library written in C++. It offers two potentially parallel algorithms: spectral sequence and chunk [68]. PHAT has efficient data structures for column additions. DIPHA [61] is a distributed computing software that can handle very large data sets (in the billions). Ripser is a time and memory efficient software that computes Vietoris Rips persistence barcodes (persistence pairs) from distance matrices sequentially. Ripser performs very well with its two (there are atleast four) optimizations: clearing [69] and cohomology [83]. There are many datasets (any dataset that is not a filtered Vietoris rips complex) that ripser cannot compute with and so we do not compare with ripser. Eirene [89] uses Morse reduction [90] for an arbitrary filtration to simplify the boundary matrix. GUDHI [91] is a TDA C++ library for computing, amongst many things, the persistent cohomology of certain complexes such as rips or alpha complexes via compressed annotation matrices [92, 93]. In addition, GPU has been used in computational geometry applications, including 3D triangle meshes, [52], and Jaccard similarity for cross-comparing spatial boundaries of segmented objects [94].

Best utilizing the device memory in GPU is an important topic. Efforts for high throughout have been made to dynamically allocate memory [95] and to batch the indexing operation in key value stores [96].

## 4.7 Conclusion

The high performance of HYPHA is achieved by an effective separation of SIMT and MIMD parallelisms, which enables us to make the following three algorithmic and system efforts. First, reduction operations are parallelized in both SIMT and MIMD modes, and executed on the best suitable device of GPU or multicore.

Second, the metadata data structures such as the lookup table that are a byproduct of the GPU scan further improves the efficiency of matrix preprocessing and parallel processing, lowering the computing complexity. Finally, HYPHA cuts data transmission overhead by overlapping the data loading to GPU with the same operations on the multicore side, and reduce the data movement frequency by significantly reducing the number of column additions. Our efforts make HYPHA win both performance and hardware cost (especially compared to a distributed algorithm like DIPHA).

Although HYPHA is developed for PH matrix reduction, it is a framework for high performance computing of irregular execution and data access patterns on hybrid systems. Our methodology of understanding structural issues of algorithms and their mappings to advanced architecture based on a holistic anatomy of a targeted application aims for general-purpose hardware and software design.

# 5. GPU-ACCELERATED COMPUTATION OF VIETORIS-RIPS PERSISTENCE BARCODES

HYPHA is an approach to solve the matrix reduction problem for persistent homology using the GPU in a CPU/GPU hybrid framework. The matrix reduction problem is concerned with computing an interval composition of Gabriel's theorem [29] for a sequence of vector spaces, namely homology vector spaces. We ask the question of how to compute persistent homology when more assumptions on the data are available. In particular, given a finite metric space, it is possible to construct a filtration of simplicial complexes from this finite metric space called the Vietoris-Rips filtration. This introduces a boundary map between adjacent dimensions of connectivity. Applying the $H_\bullet$ functor to this filtration allows us to compute persistence across the downstairs persistence module of homology vector spaces. We then ask the following question:

**Question 5.0.1.** *In the context of the Vietoris-Rips filtration, what are the parallelization opportunities for GPU when computing persistent homology?*

The computation of Vietoris-Rips persistence barcodes is both execution-intensive and memory-intensive. In this paper, we study the computational structure of Vietoris-Rips persistence barcodes, and identify several unique mathematical properties and algorithmic opportunities with connections to the GPU. Mathematically and empirically, we look into the properties of apparent pairs, which are independently identifiable persistence pairs comprising up to 99% of persistence pairs. We give theoretical upper and lower bounds of the apparent pair rate and model the average case. We also design massively parallel algorithms to take advantage of the very large number of simplices that can be processed independently of each other. Having identified these opportunities, we develop a GPU-accelerated software for computing Vietoris-Rips persistence barcodes, called Ripser++. The software achieves up to 30x speedup over the total execution time of the original Ripser and also reduces CPU-memory usage by up to 2.0x. We believe our GPU-acceleration based efforts open a new chapter for the advancement of topological data analysis in the post-Moores Law era.

## 5.1 Vietoris-Rips Filtrations

When computing persistent homology, data is usually represented by a finite metric space $X$, a finite set of points with real-valued distances determined by an underlying metric $d$ between each pair of points. A metric is defined formally below:

**Definition 5.1.1.** *A metric* $d : X \times X \to \mathbb{R}^+$ *on a point set* $X$ *is a real-valued function on pairs of points so that:*

- $d(x, x) = 0, \forall x \in X$ *(identity distance is zero)*
- $d(x, y) > 0$ *if* $x \neq y \in X$ *(distance between differing point is positive)*
- $d(x, y) = d(y, x), \forall x, y \in X$ *(symmetry)*
- $d(x, z) \leq d(x, y) + d(y, z), \forall x, y, z \in X$ *(triangle inequality)*

*We say a metric* $d$ *has* ***no equidistant points*** *if*

$$d(x, y) \neq d(z, w), \forall x, y, z, w \in X \text{ and } (x, y) \neq (z, w) \tag{5.1}$$

The metric space $X$ is defined by its distance matrix $D$, which is defined as $D[\mathrm{i}, \mathrm{j}]$= $d$(point i, point j) with $D[\mathrm{i}, \mathrm{i}] = 0$. This finite metric space $X$ can be considered as a set of samples from a distribution $P(x)$ supported on a i.i.d. set of points $\mathcal{X} \supseteq X$ with a global metric $d_{global}$ on $\mathcal{X}$. With $d_{global}$, we can define the aspect ratio of $X$ as

$$a(X)_{d_{global}} \triangleq \frac{\max_{x,y \in X} d_{global}(x, y)}{\min_{x,y \in X} d_{global}(x, y)} \tag{5.2}$$

There does not have to exist a global metric, however. The finite metric space can be defined on the samples alone.

Define an (abstract) simplicial complex $K$ as a collection of simplices closed under the subset relation, where a simplex $s$ is defined as a subset of $X$. This means that any simplex $s' : s' \subseteq s$ must also belong to $K$. We call a "filtration" as a totally ordered sequence of growing simplicial complexes. A particularly popular and useful [97] filtration is a Vietoris-Rips filtration. On the metric space, we hallucinate simplices over an adjustable threshold $t \in \mathbb{R}$. See Figure 5.1 for an illustration. A subset of points $s \subseteq X$ forms a simplex if they are close enough with respect to $t$. Let

$$\text{Rips}_t(X) = \{\varnothing \neq s \subseteq X \mid \text{diam}(s) \leq t\}, \tag{5.3}$$

where $t \in \mathbb{R}$ and diam$(s)$ is the maximum distance between pairs of points in $s$ as determined by $D$.

**Definition 5.1.2.** *The* ***Vietoris-Rips filtration****, or Rips filtration for short, is defined as the sequence:* $(Rips_t(X))_{t \leq R}$*, indexed by growing* $0 \leq t \leq R, R \in \mathbb{R}$ *where* $Rips_t(X)$ *strictly increases in cardinality for growing* $t$*.*

*A* ***full-Rips filtration*** *is a Vietoris-Rips filtration where* $R = \infty$*.*

The aspect ratio can also be defined on a Vietoris Rips filtration as follows:

$$a(\text{Rips}_t(X))_{d_{global}} \triangleq \frac{R}{\min_{\sigma \subseteq X} \text{diam}(\sigma)} \tag{5.4}$$

**Definition 5.1.3.** *Let a* $k$***-skeleton*** $K_{\text{dim} \leq k}$ *of a simplicial complex* $K$ *is defined as the abstract subcomplex of* $K$ *where every* $p$*-dimensional simplex in* $K_{\text{dim} \leq k}$ *has* $p \leq k$*.*

Certainly $K_{dim \leq k}$ is a subcomplex of $K$ since it is a subset of $K$ which satisfies downward closure. For any $p$-dimensional simplex $\sigma_p \in K_{dim \leq k}$, any $\tau \in \partial_p(\sigma_p)$ belongs to $K_{dim \leq k}$ by definition.

(5.5) The 1-skeleton of a Rips complex $\text{Rips}_t(X)$ at radius $t$ in Euclidean space is a $t$-radius graph. This is the graph of all the points with edges between points that are at most distance $t$ apart.

### 5.1.4 The Simplex-wise Refinement of the Vietoris-Rips Filtration

For computation (see Section 5.2) of Vietoris-Rips persistence barcodes, it is necessary to construct a simplex-wise refinement $S$ of a given filtration $F$. In a filtration, there is a total ordering of the simplicial complexes so that they can be indexed by a single scalar from $\mathbb{R}$.

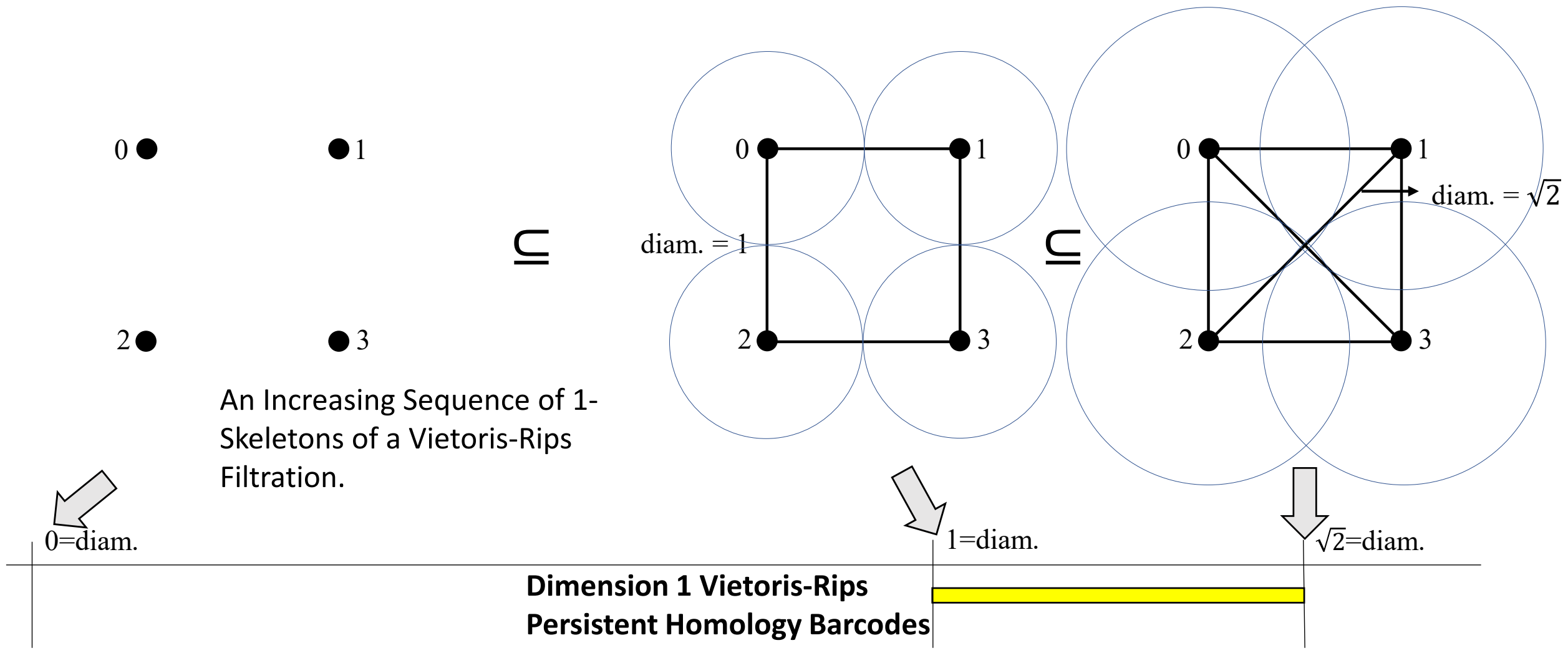

**Figure 5.1.** A filtration on an example finite metric space of four points of a square in the plane. The 1-skeleton, or simplicial complex of only points and unordered pairs of points, at each diameter value where "birth" or "death" occurs is shown. The 1 dimensional Vietoris-Rips barcode is below it: a 1-cycle is "born" at diameter 1 and "dies" at diameter $\sqrt{2}$.

The simplices within each simplicial complex, however, may not have an ordering themselves. Thus, a filtration $F$ is only equivalent to a **partial order** on the simplices of $K$, where $K$ is the largest simplicial complex of $F$. To construct $S$, we assign a total order on the simplices $\{s_\mathrm{i}\}_{\mathrm{i}=1,\ldots,|K|}$ of $K$, extending the partial order induced by $F$ so that the increasing sequence of subcomplexes $S = (K_\mathrm{j} \triangleq \bigcup_{\mathrm{i}\leq \mathrm{j}} \{s_\mathrm{i}\})_{\mathrm{j}=1,\ldots,|K|}$ ordered by inclusion grows subcomplexes by one simplex at a time. There are many ways to order a simplex-wise refinement $S$ of $F$ [98]; in the case of Ripser and Ripser++, we use the following **simplex-wise filtration** ordering criterion on simplices:

1. by increasing diameter: denoted by $\mathrm{diam}(s)$,

2. by increasing dimension: denoted by $\dim(s)$, and

3. by decreasing combinatorial index: denoted by $\mathrm{cidx}(s)$ (equivalently, by decreasing lexicographical order on the decreasing sequence of vertex indices) [99–101].

Every simplex in the simplex-wise refinement will correspond to a "column" in a uniquely associated (co-)boundary matrix for persistence computation. Thus we will use the terms "column" and "simplex" interchangeably to explain our algorithms.

Define *persistence pairs* as a pair of "birth" and "death" simplices from $K$, see [102]. These are the unique simplices $\sigma_{\mathrm{i}} \in K_{\mathrm{i}} \setminus K_{\mathrm{i}-1}, \forall \mathrm{i} = 1, \ldots, n$ occuring during the creation and destruction times for a simplex-wise filtration mentioned in Observation 3.2.14.

**Remark 5.1.5.** *We will show that the combinatorial index maximizes the number of "apparent pairs" when breaking ties under the conditions of Theorem 5.4.2.*

### 5.1.6 The Combinatorial Number System

We use the combinatorial number system to encode simplices. The combinatorial number system is simply a bijection between ordered fixed-length $\mathbb{N}$-tuples and $\mathbb{N}$. It provides a minimal representation of simplices and an easy extraction of simplex facets (see Algorithm 16), cofacets (see Algorithm 15), and vertices. When not mentioned, we assume that all simplices are encoded by their combinatorial index. The bijection is stated as follows:

$$\mathbb{N}^{d+1} \ni (v_d, \ldots, v_0) \iff \binom{v_d}{d+1} + , \ldots, + \binom{v_0}{1} \in \mathbb{N}, v_d > v_{d-1} > \ldots > v_0 \geq 0. \tag{5.6}$$

For a proof of this bijection see [99–101].

**Remark 5.1.7.** *It should be noted that, without mentioning, we will use the notation $(v_d, \ldots, v_0)$ with $v_d > v_{d-1} > \ldots > v_0 \geq 0$ for simplices and a lexicographic ordering on the simplices by* decreasing *sorted vertex indices. One may equivalently view the lexicographic ordering on simplices $(v_0, \ldots, v_d)$ with $0 \leq v_0 < v_1 < \ldots < v_d$ as being in colexicographic order on the given increasingly ordered vertices.*

## 5.2 Computation

The general computation of persistent barcodes involves two inter-relatable stages. One stage is to construct a simplex-wise refinement [59] of the given filtration. The other stage is to "reduce" the corresponding boundary matrix by a "standard algorithm" [67]. In Algorithm 11, let $\text{low}_R(\text{j})$ be the maximum nonzero row of column j, -1 if column j is zero for a given matrix $R$. The pairs $(\text{low}_R(\text{j}), \text{j})$ over all j correspond bijectively to *persistence pairs*.

**Algorithm 11:** Standard Persistent Homology Computation

**Input:** filtered simplicial complex $\boldsymbol{K}$

**Output:** $\boldsymbol{P}$ persistence barcodes

$\boldsymbol{F} \leftarrow \boldsymbol{F_K}$ /* Let $\boldsymbol{F}$ be the filtration of $\boldsymbol{K}$ */<br>
$\boldsymbol{S} \leftarrow$ simplex-wise-refinement$(\boldsymbol{F})$ /* $\boldsymbol{F} = \boldsymbol{S} \circ r$ where $r$ is injective */<br>
$R \leftarrow \partial(\boldsymbol{S})$ ;<br>
**foreach** *column* j *in* $R$ **do**<br>
    **while** $\exists k <$ j *such that* $low_R(\text{j}) = low_R(k)$ **do**<br>
        column j $\leftarrow$ column $k$ + column j<br>
    **end**<br>
    **if** $low_R(\text{j}) \neq -1$ **then**<br>
        $\boldsymbol{P} \leftarrow \boldsymbol{P} \cup r^{-1}([\text{low}_R(\text{j}), \text{j}])$ /* We call the pair $(\text{low}_R(\text{j}), \text{j})$ a pivot in the matrix $R$. */<br>
    **end**<br>
**end**

The construction stage can be optimized [89, 91, 103–105]. Furthermore, all existing persistent homology software efforts are based on the standard algorithm[1, 58–60, 103, 104, 106, 107].

### 5.2.1 The Coboundary Matrix

In Proposition 4.3.5 we show that it is possible to compute the standard matrix reduction of Algorithm 2 on the anti-transposed boundary matrix $[\partial(K_n)]^T_{anti}$. We call this "computing cohomology" since for every $p \geq 0$ the standard algorithm is iterating through $p$-cochains from the dual of the $p$-chain group $C_p(K_n)$. We compute cohomology [82, 83, 93] in Ripser++, like in Ripser, for performance reasons specific to Rips filtrations mentioned in [103] which will be reviewed in Section 5.2.2. For the anti-transpose boundary matrix of a simplex-wise filtration from Definition 4.3.4 we rename it as a **coboundary matrix** here. The reason for the name change is because we will view it as an enumerable ordered set of cofacets. A coboundary matrix consists of coboundaries (each column is made up of the cofacets of the corresponding simplex) where the columns/simplices are ordered in reverse to the order given in Section 5.1.4 (see [83]). If certain columns can be zeroed/cleared [69] in the coboundary matrix, we will still denote the cleared matrix as a coboundary matrix since the matrix reduction does nothing on zero columns (see Algorithm 11).

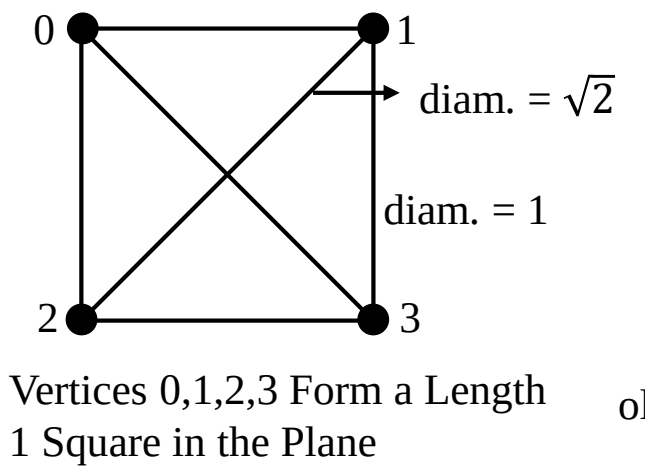

Dim 1 Coboundary Matrix

older →, older ↓

| (diam., simplex) | ($\sqrt{2}$, (21)) | ($\sqrt{2}$, (30)) | (1, (10)) | (1, (20)) | (1, (31)) | (1, (32)) |
|---|---|---|---|---|---|---|
| ($\sqrt{2}$, (210)) | **1** | | **1** | **1** | | |
| ($\sqrt{2}$, (310)) | | **1** | **1** | | **1** | |
| ($\sqrt{2}$, (320)) | | **1** | | **1** | | **1** |
| ($\sqrt{2}$, (321)) | **1** | | | | **1** | **1** |

**Figure 5.2.** The full 1-skeleton for the point cloud of Figure 5.1. Its 1-dimensional coboundary matrix is shown on the right. Let $(\mathrm{e}, (a_d, ..., a_0))$ be a $d$-dimensional simplex with vertices $a_d$,...,$a_0$ such that $v_d > v_{d-1} > ... > v_0 \geq 0$ and diameter $\mathrm{e} \in \mathbb{R}^+$. For example, simplex (1,(10)) has vertices 1 and 0 with diameter 1. The order of the columns/simplices is the reverse of the simplex-wise refinement of the Vietoris-Rips filtration.

### 5.2.2 Computational Structure in Ripser, a Review

The sequential computation in Ripser follows the two stages given in Algorithm 11, with the two stages computed for each dimension's persistence in increasing order. This computation is further optimized with four key ingredients [103]. We use and build on top of all of these four optimizations for high performance, which are:

- 1. The clearing lemma [69] [68] [1],
- 2. Computing cohomology [83] [93],
- 3. Implicit matrix reduction [103], and
- 4. The emergent pairs lemma [103].

In this section we review these four optimizations as well as the enclosing radius condition. Our contributions follow in the next section, Section 5.3. The reader may skip this section if they are very familiar with the concepts from [103].

### 5.2.3 Clearing Lemma

As shown in [108], there is a partition of all simplices into "birth" and "death" simplices. The introduction of a "birth" simplex in a simplex-wise refinement of the Rips filtration creates a homology class. On the other hand, "death" simplices zero a homology class or merge two homology classes upon introduction into the simplex-wise filtration. Death simplices are of exactly one higher dimension than their corresponding birth simplex.

Paired birth and death simplices are represented by a pair of columns in the boundary matrix. For a boundary matrix, the clearing lemma states that paired birth columns must be zero after reduction [68, 69]. Furthermore, death columns are nonzero when fully reduced and their lowest nonzero entry after reduction by the standard algorithm in Algorithm 11, determines its corresponding paired birth column.

This lemma optimizes the matrix reduction stage of computation and is most effective when it is used before any column additions. The clearing lemma is widely used in all persistent homology computation software packages to lower the computation time of matrix reduction. As shown in [1], the smallest set of columns taking up 50% of all column additions in Algorithm 11, also known as "tail columns" are comprised in a large percentage by columns that can be zeroed by the clearing lemma.

### 5.2.4 Computing Cohomology as the Dual Matrix Reduction Problem

It has been proven through linear algebraic techniques that persistence barcodes can be equivalently computed, by the matrix reduction in Algorithm 11, of a coboundary instead of a boundary matrix [83]. This was proven in Theorem 3.2.28. See Section 5.2.1 for the definition of a coboundary matrix.

As shown in [103], in the case of a full-Rips filtration on $n_0$ points of a $(d{+}1)$-skeleton along with the clearing lemma, a significant number of paired creator columns can be eliminated due to the application of clearing to the top $d$-dimensional simplices, which dominate the total number of simplices. Computing cohomology for Vietoris-Rips filtrations, furthermore, significantly lowers the number of columns of a coboundary matrix of dimension $d$ to consider comparing to a boundary matrix of dimension $d$. Thus, the memory allocation needed to represent a sparse column-store matrix is lowered. This is because there are at most $\binom{n_0}{d+2}$ number of $(d+1)$-dimensional simplices (sparsely represented rows) and only $\binom{n_0}{d+1}$ $d$-dimensional simplices (columns). Excessive memory allocation can furthermore become a bottleneck to total execution time.

### 5.2.5 Low Complexity 0-Dimensional Persistence Computation

The 1-skeleton of the Rips complex at dimension 1 is a graph. Viewing the simplex-wise filtration of 1-skeleton Rips subcomplexes, the edges are introduced in order of distance. The edges that form cycles are viewed as 1-dimensional creation events while the remaining edges, called complementary edges, destroy 0-dimensional creation events (the points). This allows us to compute 0-dimensional persistent homology. It also simulates Kruskal's algorithm in exactly the same way. Thus, the edges that are introducing a destruction of a 0-dimensional creation event form an EMST, see Definition 2.5.4, and the complementary edges form a 1-skeleton.

0-dimensional persistence can be computed by a union-find algorithm in Ripser. This algorithm has complexity of $O(\alpha(n) \cdot n_0^2)$, where $n_0$ is the number of points and $\alpha$ is the inverse of the Ackermann's function (essentially a constant). There is no known algorithm that can achieve this low complexity for persistence computation in higher dimensions. This is another reason for computing cohomology. In computing cohomology, clearing is applied from lower dimension to higher dimension (clearing out columns in the higher dimension), the 0th dimension has no cleared simplices and there are very few $d$-dimensional simplices compared to $(d+1)$-dimension simplicies. Thus the 0th dimension should be computed as efficiently as possible without clearing. Since it is more efficient to keep the union-find algorithm on CPU, we focus only on dimension $\geq 1$ persistence computation in this paper and in Ripser++. GPU can offer speedup, especially in the top dimension, in this case.

### 5.2.6 Complexity of the Dual Matrix Reduction Problem

In Proposition 4.2.8 we show the complexity of computing the primal matrix reduction problem of Problem 3.2.2. This has a complexity of $O(\sum_{p=1}^{\dim(K_n)} n_p n_{p-1} \log(n_0))$. As mentioned in Section 5.2.4, the coboundary matrix can eliminate having to store any of the highest dimensional simplices. We show that the time efficiency is also greatly improved. Intuitively, in persistent homology we are computing the kernel of the $p$-dimensional boundary matrix and connecting the kernel with the image of the $p+1$-dimensional boundary matrix. This boundary matrix is "wide" in the case of Rips-filtrations since there are $n_{p+1} = O(\binom{n_0}{p+2})$ $p+1$-dimensional simplices but only $n_p = O(\binom{n_0}{p+1})$ $p$-dimensional simplices. By rank-nullity, we know that $n_{p+1} = \text{rank}([\partial_{p+1}]) + \text{nullity}([\partial_{p+1}])$. However, $\text{rank}([\partial_{p+1}]) \leq \min(n_p, n_{p+1})$. If $n_p \leq n_{p+1}, \forall p \geq 0$, then the dimension of the kernel dominates in complexity. This means it would be beneficial to eliminate having to compute the kernel of the highest dimension boundary $[\partial_{\text{dim}}]$.

**Proposition 5.2.7.** *Assuming $n_p \leq n_{p+1}, \forall p \geq 0$.*

*The time complexity $Time(K_n)$ of computing the dual matrix reduction problem of Problem 3.2.2 is:*

$$\Omega(\sum_{p=1}^{dim(K_n)} n_{p-1} \log(n_0)) \leq Time(K_n) \leq O(\sum_{p=1}^{dim(K_n)} n_{p-1}^2 \log(n_0)) \tag{5.7}$$

*Proof.* For each $p \geq 1$, we can write out the columns of each boundary matrix $[\partial_p]^T$ into those that correspond to creation times and those that correspond to destruction times. According to Observation 3.2.14, we know that each column can only correspond exclusively to one of a creation or a destruction time.

The columns that cannot be zeroed by the maximin Equation of Problem 3.2.2 correspond to destruction times. There are $\text{rank}([\partial_p^\perp]))$ many such columns. The remaining columns correspond to creation times, and there are: $(n_{p-1} - \text{rank}([\partial_p^\perp]))$ many of them. This is exactly the dimension of the kernel of $\partial_p^\perp$.

We analyze the time to compute the creator and destructor columns individually. We show the total time complexity for the reduction over all creator and destructor columns. This analysis is similar to that in the primal case.

- For creation: A creator column is determined by the linearly independent columns in its causal submatrix. This has an upper bound of $\text{rank}(\partial_p^\perp)$ and a lower bound of $\Omega(n_0)$. There are $\dim(\ker(\partial_p^\perp)$ many creation columns. Each column addition requires $\Theta(\log(n_0))$ time to compute by the data structure [59, 80] for XOR operations between columns, thus the total complexity over all creation columns is $O(\dim(\ker(\partial_p^\perp))\text{rank}(\partial_p)\log(n_0))$ and $\Omega(\dim(\ker(\partial_p^\perp))\log(n_0))$.
- For destruction: A column is a destructor column if it does not belong to the kernel of $\partial_p^\perp$. Each such column must add with at most $\text{rank}(\partial_p^\perp)$ many columns and atleast $\Omega(1)$ columns, thus we have an upper bound of $O(\text{rank}^2(\partial_p^\perp)\log(n_0))$ and lower bound of $\Omega(\text{rank}(\partial_p^\perp))$.

Upon adding the total time to reduce a creation column and the total time to reduce a destructor column, we have the asymptotic upper bound on the computational complexity $\text{Time}(K_n)$

$$\text{Time}(K_n) = O(\sum_{p=1}^{\dim(K_n)} n_{p-1}^2 \log(n_0)) \tag{5.8}$$

For the lower bound, once again by rank-nullity, we obtain

$$\text{Time}(K_n) = \Omega(\sum_{p=1}^{\dim(K_n)} n_{p-1} \log(n_0)) \tag{5.9}$$

Thus:

$$\Omega(\sum_{p=1}^{\dim(K_n)} n_{p-1} \log(n_0)) \leq \text{Time}(K_n) \leq O(\sum_{p=1}^{\dim(K_n)} n_{p-1}^2 \log(n_0)) \tag{5.10}$$

□

### 5.2.8 Implicit Matrix Reduction

In Ripser, the coboundary matrix is not fully represented in memory. Instead, the columns or simplices are represented by natural numbers via the combinatorial number system [100, 101, 103] and cofacets are generated as needed and represented by combinatorial indices. This saves on memory allocation along the row-dimension of the coboundary matrix, which scales by a multiplicative factor of $O(n_0)$ more in cardinality than the column-dimension. Furthermore, the generation of cofacets allows us to trade computation for memory. Memory address accesses are replaced by arithmetic and repeated accesses of the much smaller distance matrix and a small binomial coefficient table. Implicit matrix reduction intertwines coboundary matrix construction with matrix reduction.

### 5.2.9 Reduction Matrix vs. Oblivious Matrix Reduction

There are two matrix reduction techniques in Ripser, see Algorithm 11, that can work on top of the implicit matrix reduction. These techniques are applied on a much smaller submatrix of the original matrix in Ripser++ significantly improving performance over full matrix reduction, see Section 5.5.1.

The first is called the reduction matrix technique. This involves storing the column operations on a particular column in a $V$ reduction matrix (see a variant in [92]) by executing the same column operations on $R$ as on the initially identity matrix $V$; $R = \partial \cdot V$ where $\partial$ is the (implicitly represented) boundary operator and where $R$ is the matrix reduction of $\partial$. To obtain a fully reduced column $R_{\mathrm{j}}$ as in Algorithm 11, the nonzeros $V_{\mathrm{i,j}}$ of a column of $V_{\mathrm{j}}$ identify the sum of boundaries $\partial_{\mathrm{i}}$ needed to obtain column $R_{\mathrm{j}} = \Sigma_{\mathrm{i}} \partial_{\mathrm{i}} \cdot V_{\mathrm{i,j}}$.

The second is called the oblivious matrix reduction technique. This involves not caching any previous computation with the $R$ or $V$ matrix, see Algorithm 12 for the reduction of a single column j. Instead, only the boundary matrix pivot indices are stored. A *pivot* is a row column pair $(r, c)$, being the lowest **1** entry of a fully reduced nonzero column $c$ and representing a persistence pair.

In the following, we use the notation $R_{\mathrm{j}}$ to denote a column reduced by the standard algorithm (Algorithm 11) and $R[\mathrm{j}]$ to denote a partially obliviously reduced column. $D_{\mathrm{j}}$ denotes the jth column of the boundary matrix.

**Algorithm 12:** Oblivious Column Reduction

**Input:** j: column to reduce index, $D = \partial$: boundary matrix, lookup[rows 0..j − 1]: lookup table with lookup[$row$] = $col$ if $(row, col)$ is a pivot, -1 otherwise; $low$(j): the maximum row index of any nonzero entry in column j, -1 if the column j is 0.

**Output:** $R$[j] is fully reduced by oblivious reduction and is equivalent to a fully reduced $R_j = D_j$ as in Algorithm 11.

/* Assume columns of index 0 to j − 1 have all been reduced by the oblivious column reduction algorithm */

$R[j] \leftarrow D_j$ ;

**while** $lookup[low(R[j])] \neq -1$ **do**

  $R[j] \leftarrow R[j] + D_{lookup[low(R[j])]}$ ;

**end**

**if** $R[j] \neq 0$ **then**

  $\text{lookup}[low(R[j])] \leftarrow j$ ;

**end**

The reduction matrix technique is correct by the fact that it (re)computes $R_j = \Sigma_i D_i \cdot V_{i,j}$ as needed before adding it with $R_k$ for $k > j$. Thus it involves the same sequence of $(R_j)_j$ to do column additions with $R_k$ as in the standard algorithm in Algorithm 11.

**Lemma 5.2.10.** *Algorithm 2 (oblivious column reduction from Ripser) is equivalent to a reduction of column* j *as in Algorithm 11, namely* $R[j] \leftarrow R[j] + R_i$ *where* $i = lookup[low(R[j])]$.

Our proof of Lemma 5.2.10 is in the Appendix Section 5.10.1.

The reduction matrix technique can lower the column additions (addition of $D_i$ to $R[j]$) needed to reduce any particular column j since after many column additions, many of the nonzeros of $V_j$ will cancel out by modulo 2 arithmetic. This is in contrast to oblivious matrix reduction where there cannot be any cancellation of column additions. Experiments show that datasets with large number of column additions are executed more efficiently with the reduction matrix technique rather than the oblivious matrix reduction technique. In fact, certain difficult datasets will crash on CPU due to too much memory allocation for columns from the oblivious matrix reduction technique but execute correctly in hours by the reduction matrix technique.

### 5.2.11 The Emergent Pairs Lemma

As we generate column cofacets during matrix reduction, we may "skip over" their construction if we can determine that they are "0-addition columns" or have an "emergent pair" [1, 103]. These columns have no column to their left that can add with them. The lemma is stated in [103] and involves a sufficient condition to find a "lowest **1**" or maximal indexed nonzero in a column followed by a check for any columns to its left that can add with it. These nonzero entries correspond to "shortcut pairs" that form a subset of all persistence pairs. We may pair implicit matrix reduction with the emergent pairs lemma to achieve speedup over explicit matrix reduction techniques [1, 59]. However, apparent pairs (see Figure 5.4), when processed massively in parallel by GPU are even more effective for computation than processing the sequentially dependent shortcut pairs (see Figure 5.12).

### 5.2.12 Filtering Out Simplices by Diameter Equal to the "Enclosing Radius"

We define the enclosing radius $R$ as $\min_{x\in X}\max_{y\in X} d(x,y)$, where $d$ is the underlying metric of our finite metric space $X$. If we compute Vietoris-Rips barcodes up to diameter $\leq R$, then the nonzero persistence pairs will not change after the threshold condition is applied [63]. Notice that applying the threshold condition is equivalent to truncating the coboundary matrix to a lower right block submatrix, potentially significantly lowering the size of the coboundary matrix to consider. We prove the following claim used in [107].

**Proposition 5.2.13.** *Computing persistence barcodes for full Rips filtrations with diameter threshold set to the enclosing radius $R$ does not change the nonzero persistence pairs.*

*Proof.* Notice that when we reach the "enclosing radius" $R$ length in the filtration, every point has an edge to one apex point $p \in X$. This means that any $d$-dimensional cycle must actually be a boundary. Thus the following two statements are true. 1. Any persistence interval $[birth, death)$ with $birth \leq R$ must have $death \leq R$. 2. Since no cycles that are not boundaries can form after $R$, there will be no nonzero persistence barcodes $[birth, death)$ with $birth > R$.

By statements 1 and 2, we can restrict all simplices to have diameter $\leq R$ and this does not change any of the nonzero persistence intervals of the full Rips filtration. □

## 5.3 Mathematical and Algorithmic Foundations for GPU Acceleration

### 5.3.1 Overview of GPU-Accelerated Computation

Figure 5.3(a) shows a high-level structure of Ripser, which processes simplices dimension by dimension. In each dimension starting at dimension 1, the filtration is constructed and the clearing lemma is applied followed by a sort operation. The simplices to reduce are further processed in the matrix reduction stage, where the cofacets of each simplex are enumerated to form coboundaries and the column addition is applied iteratively. The matrix reduction is highly dependent among columns.

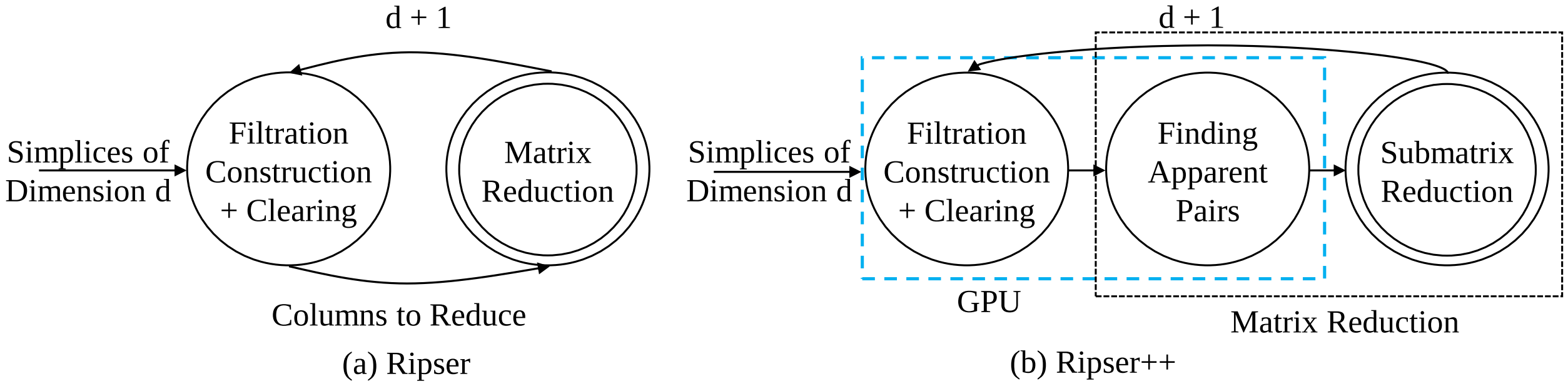

**Figure 5.3.** A High-level computation framework comparison of Ripser and Ripser++ starting at dimension $d \geq 1$ (see Section 5.2.5 for $d$=0). Ripser follows the two stage standard persistence computation of sequential Algorithm 11 with optimizations. In contrast, Ripser++ finds the hidden parallelism in the computation of Vietoris-Rips persistence barcodes, extracts "Finding Apparent Pairs" out from Matrix Reduction, and parallelizes "Filtration Construction with Clearing" on GPU. These two steps are designed and implemented with new parallel algorithms on GPU, as shown in the Figure 5.3(b) with the dashed rectangle.

Running Ripser intensively on many datasets, we have observed its inefficiency on CPU. There are two major performance issues. First, in each dimension, the matrix reduction of Ripser uses an enumeration-and-column-addition style to process each simplex. Although the computation is highly dependent among columns, a large percentage of columns (see Table 5.1 in Section 5.6) do NOT need the column addition. Only the cofacet enumeration and a possible persistence pair insertion (into the hashmap of Ripser) are needed on these columns. In Ripser, a subset of these columns are identified by the “emergent pair” lemma [103] as columns containing “shortcut pairs”. Ripser follows the sequential framework of Figure 5.3(a) to process these columns one by one, where rich parallelisms, stemming from a large percentage of "apparent pairs", are hidden. Second, in the filtration construction with clearing stage, applying the clearing lemma and predefined threshold is independent among simplices. Furthermore, one of the most time consuming part of filtration construction with clearing is sorting. A highly optimized sorting implementation on CPU could be one order of magnitude faster than sorting from the C++ standard library [109]. On GPU, the performance of sorting algorithms can be further improved due to the massive parallelism and the high memory bandwidth of GPU [110, 111].

In our hardware-aware algorithm design and software implementation, we aim to turn these hidden parallelisms and data localities into reality for accelerated computation by GPU for high performance. Utilizing SIMT (single instruction, multiple threads) parallelism and achieving coalesced device memory accesses are our major intentions because they are unique advantages of GPU architecture. Our efforts are based on mathematical foundation, algorithms development, and effective implementations interacting with GPU hardware, which we will explain in this and the following section. Figure 5.3(b) gives a high-level structure of Ripser++, showing the components of Vietoris-Rips barcode computation offloaded to GPU. We will elaborate on our mathematical and algorithmic contributions in this section.

### 5.3.2 Matrix Reduction

Matrix reduction (see Algorithm 11) is a fundamental component of computing Rips barcodes. Its computation can be highly skewed [1], particularly involving very few columns for column additions. Part of the reason for the skewness is the existence of a large number of "apparent pairs." We prove and present the Apparent Pairs Lemma and a GPU algorithm to find apparent pairs in an implicitly represented coboundary matrix. We then design and implement a 2-layer data structure that optimizes the performance of the hashmap storing persistence pairs for subsequent matrix reduction on the non-apparent columns, which we term "submatrix reduction". The design of the 2-layer data structure is centered around the existence of the large number of apparent pairs and their usage patterns during submatrix reduction.

### 5.3.3 The Apparent Pairs Lemma

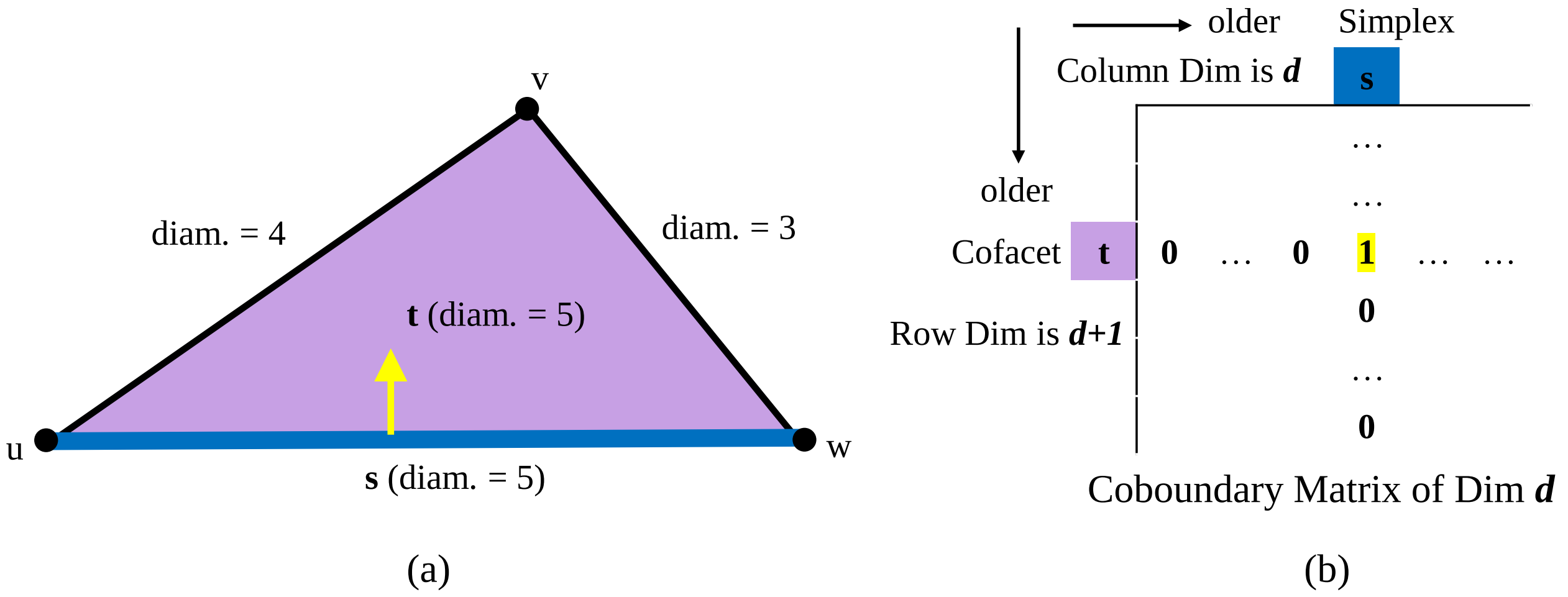

**Figure 5.4.** (a) A dimension 1 0-persistence apparent pair $(s,t)$ on a single 2-dimensional simplex. $s$ is an edge of diameter 5 and $t$ is a cofacet of $s$ with diameter 5. The light arrow denotes the pairing between $s$ and $t$. (b) In the dimension $d$ coboundary matrix, $(s,t)$ is an apparent pair iff entry $(t,s)$ has all zeros to its left and below. We color columns/simplices $s$ with blue and their oldest cofacet $t$ with purple in (a) and (b). See Figure 5.2 for an example 1-dimensional coboundary matrix for a square in the plane.

Apparent pairs of the form $(s,t)$ are particular kinds of persistence pairs, or pairs of "birth" and "death" simplices as mentioned in Sections 5.2.3 and 5.1.4 (see also [102]). In particular they are determined only by the (co-)boundary relations between $s$ and $t$ and the simplex-wise filtration ordering of all simplices.

Apparent pairs [103] show up in the literature by other names such as close pairs [112] or obvious pairs [63] but all are equivalent. Furthermore, apparent pairs are known to form a discrete gradient of a discrete Morse function [103] and have many other mathematical properties.

**Definition 5.3.4.** *A pair of simplices $(s,t)$ is an apparent pair iff:*

1. *$s$ is the youngest facet of $t$ and*
2. *$t$ is the oldest cofacet of $s$.*

We will use the simplex-wise order of Section 5.1.4 for Definition 5.3.4. In a (co-)boundary matrix, a nonzero entry having all zeros to its left and below is equivalent to an apparent pair. Thus apparent pairs do not require the column reduction of Algorithm 11 to be determined. We call a column containing such a nonzero entry as an apparent column. An example of an apparent pair geometrically and computationally is shown in Figure 5.4. Furthermore, apparent pairs have zero persistence in Rips filtrations by property 1 of Definition 5.3.4 and that the diameter of a simplex is determined by its maximum length edge.

In the explicit matrix reduction, where every column of $R$'s nonzeros are stored in memory (see Algorithm 11), it is easy to determine apparent pairs by checking the positions of $s$ and $t$ in the (co-)boundary matrix. However, in the implicit matrix reduction used in Ripser and Ripser++, we need to enumerate cofacets $t$ from $s$ and facets $s'$ from $t$ at runtime. It is necessary to enumerate both, because even if $t$ is found as the oldest cofacet of $s$, $t$ may have facets younger than $s$.

We first notice a property of the facets of a cofacet $t$ of simplex $s$ where $\mathrm{diam}(s) = \mathrm{diam}(t)$. Equivalently, we find a property of the nonzeros along particular rows of the coboundary matrix, allowing for a simple criterion for the order of the nonzeros in a row based on simply computing the diameter and combinatorial index of related columns:

**Proposition 5.3.5.** *Let $t$ be the cofacet of simplex $s$ with $diam(s) = diam(t)$.*

*$s'$ is a strictly younger facet of $t$ than $s$ iff*

1. *$diam(s') = diam(s) = diam(t)$ and*
2. *$cidx(s') < cidx(s)$. ($s'$ is strictly lexicographically smaller than $s$)*

*Proof.* ($\Longrightarrow$) $s'$ as a facet of $t$ implies that $\mathrm{diam}(s') \leq \mathrm{diam}(t) = \mathrm{diam}(s)$. If $s'$ is strictly younger than $s$, then $\mathrm{diam}(s') \geq \mathrm{diam}(s)$. Thus 1. $\mathrm{diam}(s') = \mathrm{diam}(s) = \mathrm{diam}(t)$.

Furthermore, if $s'$ is strictly younger than $s$ and $\text{diam}(s') = \text{diam}(s)$, then the only way for $s'$ to be younger than $s$ is if 2. $\text{cidx}(s') < \text{cidx}(s)$.

($\Longleftarrow$) If $\text{diam}(s') = \text{diam}(s) = \text{diam}(t)$ and $\text{cidx}(s') < \text{cidx}(s)$ then certainly $s'$ is a strictly younger facet of $t$ than $s$ is as a facet of $t$. □

We propose the following lemma to find apparent pairs:

**Lemma 5.3.6** (The Apparent Pairs Lemma)**.** *Given simplex $s$ and its cofacet $t$,*

*1. $t$ is the lexicographically greatest cofacet of $s$ with $diam(s) = diam(t)$, and*

*2. no facet $s'$ of $t$ is strictly lexicographically smaller than $s$ with $diam(s') = diam(s)$,*

*iff $(s,t)$ is an apparent pair.*

*Proof.* ($\Longrightarrow$) Since $\text{diam}(t) \geq \text{diam}(s)$ for all cofacets t, Condition 1 is equivalent to having chosen the cofacet $t$ of $s$ of minimal diameter at the largest combinatorial index, by the filtration ordering we have defined in Section 5.1.4; this implies $t$ is the oldest cofacet of $s$.

Assuming Condition 1, by the negation of the iff in Proposition 5.3.5, there are no simplices $s'$ with $\text{diam}(s') = \text{diam}(s) = \text{diam}(t)$ and $\text{cidx}(s') < \text{cidx}(s)$ iff $s$ is the youngest facet of $t$.

($\Longleftarrow$) If $\text{diam}(t) > \text{diam}(s)$ then there exists a younger $s'$ with same cofacet $t$ and thus $s$ is not the youngest facet of $t$. Thus $(s,t)$ being an apparent pair implies Condition 1. Furthermore, $(s,t)$ being apparent ($s$ the youngest facet of $t$) with Condition 1 ($\text{diam}(s) = \text{diam}(t)$) implies Condition 2 by negating the iff in Proposition 5.3.5.

Thus (Conditions 1 and 2) is equivalent to Definition 5.3.4. □

An elegant algorithmic connection between apparent pairs and the GPU is exhibited by the following Corollary.

**Corollary 5.3.7.** *The Apparent Pairs Lemma can be applied for massively parallel operations on every column $s$ of the coboundary matrix.*

*Proof.* Notice we may generate the cofacets of simplex $s$ and facets of cofacet $t$ of $s$ independently with other simplices $s' \neq s$. □

**Remark 5.3.8.** *The effectiveness of the Apparent Pairs Lemma hinges on an important empirical fact and common dataset property: there are a lot of apparent pairs [1, 103]. In fact, by Table 5.1 in Section 5.6, in many real world datasets up to 99% of persistence pairs over all dimensions are apparent pairs. Theoretically, we further show in Section 5.4 bounds on the number of apparent pairs for a full Rips complex induced by a clique on* $n_0$ *points under a condition depending on the vertex with maximal index, vertex* $n-1$ *(assume 0-indexed vertices). We also perform experiments on random distance matrices with unique entries in Section 5.6.6 and model an approximation to the expected number of apparent pairs. These results further confirm that there are a large number of apparent pairs in a (co-)boundary matrix induced by a Rips filtration.*

Apparent pairs can be found after the clearing lemma and a threshold condition is applied. Thus clearing can be applied as early as possible, even before coboundary matrix reduction so long as the previous dimension's persistence pairs are already found. The following proposition proves this fact. During computation this allows us to eliminate memory space (and thus memory accesses) for columns that will end up zero by the end of computation. In particular, the following proposition justifies the general computation order of Ripser++, namely filtration construction with clearing coming before finding apparent pairs followed by submatrix reduction.

**Proposition 5.3.9.** *The set of apparent pairs does not change if they are found in the coboundary matrix after the clearing lemma.*

*Proof.* We show the number of apparent pairs does not change in a coboundary matrix after the clearing lemma is applied. The number of apparent pairs cannot decrease after clearing since we can only clear birth columns (see Section 5.2.3), while apparent columns are always death columns. We show the number of apparent pairs cannot increase after clearing either.

Consider for contradiction a nonapparent pair $(s, t)$ corresponding to entry $(t, s)$ existing in the coboundary matrix. Let row $t$ not have all zeros to its left before clearing, and let one of the nonzeros in row $t$ belong to a column $s'$ cleared by the clearing lemma. We show that clearing column $s'$ will not make pair $(s, t)$ apparent. This follows since a cleared column $s'$ corresponds to a sequence of column additions with columns to its left in the standard algorithm [69]. Thus there must exist a nonzero entry $(t, s'')$ to the left of entry $(t, s')$, ($s'' < s'$ in the simplex-wise filtration order). Thus the pair $(s, t)$ must stay nonapparent since if it were apparent, then there would be all zeros to the left of entry $(t, s)$. □

### 5.3.10 Finding Apparent Pairs in Parallel on GPU

**Algorithm 13:** Finding Apparent Pairs on GPU

```
Input: C: the simplices to reduce; vertices(·): the vertices of a simplex; diam(·):
       the diameter of a simplex; cidx(·): the combinatorial index of a simplex;
       dist(·): the distance between two vertices; enumerate-facets(·): enumerates
       facets of a simplex.
Output: A: the apparent pair set from the coboundary matrix of dimension dim. ;
            /* global to all threads */
tid: the thread id.  /* local to each thread */
s ← C[tid]  /* Each thread fetches a distinct simplex from the set of simplices */
V ← vertices(s)  /* This only depends on the combinatorial index of s */
foreach cofacet t of s in lexicographically decreasing order do
    foreach v' in V do
        diam(t) ← max(dist(v', v), diam(s))  /* Calculate the diameter of t */
    end
    if diam(t) = diam(s) then
        S ← ∅ ;
        enumerate-facets(t, S)  /* S are facets of t in lexicographical increasing
            order */
        foreach s' in S do
            if diam(s') = diam(s) then
                if cidx(s') = cidx(s) then
                    /* s is the youngest facet of t */
                    A ← A ∪ {(s, t)} ;
                    return ;        /* Exit if (s, t) is apparent or if s' is strictly
                    younger than s */
                end
            end
        end
    end
end
```

Based on Lemma 5.3.6, finding apparent pairs from a cleared coboundary matrix without explicit coboundaries becomes feasible. There is no dependency for identifying an apparent pair as Corollary 5.3.7 states, giving us a unique opportunity to develop an efficient GPU algorithm by exploiting the massive parallelism.

Algorithm 13 shows how a GPU kernel finds all apparent pairs in a massively parallel manner. A GPU thread fetches a distinct simplex $s$ from an ordered array of simplices, represented as a (diameter, cidx) struct in GPU device memory, and uses the Apparent Pairs Lemma to find the oldest cofacet $t$ of $s$ and ensure that $t$ has no younger facet $s'$. If the two conditions of the Apparent Pairs Lemma hold, $(s, t)$ can form an apparent pair. Lastly, the kernel inserts into a data structure containing all apparent pairs in the GPU device memory.

The complexity of one GPU thread is O($log(n_0) \cdot (d$+1)+($n_0$-$d$-1) $\cdot$ ($d$+1)), in which $n_0$ is the number of points and $d$ is the dimension of the simplex $s$. The first term represents a binary search for $d$+1 simplex vertices from a combinatorial index, and the second term says the algorithm checks at most $d$+1 facets of all $n_0$-$d$-1 cofacets of the simplex $s$. Notice that this complexity is linear in $n_0$, the number of points, with dimension $d$ small and constant.

### 5.3.11 Review of Enumerating Cofacets in Ripser

Enumerating cofacets/facets in a lexicographically decreasing/increasing order is substantial to our algorithm. The cofacet enumeration algorithm differs for full Rips computation and for sparse Rips computation. Cofacet enumeration is already implemented in Ripser. Details of the cofacet enumeration from Ripser are presented in Algorithm 14 of Section 5.3.11. For enumerating cofacets of a simplex in the sparse case, we utilize the sparsity of edge relations as in Ripser. Algorithm 15 shows such an enumeration.

In Algorithm 14, we enumerate cofacets of a simplex $s$ by iterating through all vertices $v$ of $X$, a finite metric space. We keep track of an integer $k$. If $v$ matches a vertex of $s$, then we decrement $k$ and $v$ until we can add $\binom{v}{k}$ legally as a binomial coefficient of the combinatorial index of a cofacet of $s$. This algorithm assumes a dense distance matrix, meaning all its entries represent a finite distance between any two points in $X$.

**Algorithm 14:** Enumerating Cofacets of a Simplex

**Input:** $\boldsymbol{X} = \{0, ..., n_0 - 1\}$: a finite metric space; $s$: a simplex with vertices in $\boldsymbol{X}$; vertices$(\cdot)$: the vertices of a simplex; cidx$(\cdot)$: the combinatorial index of a simplex.

**Output:** $\boldsymbol{S}$: the facets of $s$ in lexicographically decreasing order.

```
V ← vertices(s) ;
cidx(s'_high) ← 0 ;
cidx(s'_low) ← cidx(s) ;
while v ∈ X = {0..n − 1} do
    if v ∉ V then
        cidx(s') ← cidx(s'_high) + (v choose k) + cidx(s'_low) ;
        v ← v − 1 ;
    else
        while v ∈ V do
            cidx(s'_high) ← cidx(s'_high) + (v choose k+1)   /* vertices(s'_high) ← vertices(s'_high) ∪ {v} */
            cidx(s'_low) ← cidx(s'_low) − (v choose k)       /* vertices(s'_low) ← vertices(s'_low) − {v}   */
            v ← v − 1 ;
            k ← k − 1 ;
        end
    end
    append(S, s') ;
end
```

### 5.3.12 Computation Induced by Sparse 1-Skeletons

Due to the exponential growth in the number of simplices by dimension during computation of Vietoris-Rips filtrations, which demands a high memory capacity and causes long execution time, we also consider enumerating cofacets induced by sparse distance matrices. A sparse distance matrix $D$ simply means that we consider certain distances between points to be "infinite." This prevents certain edges from contributing to forming a simplex since a simplex's diameter must be finite. Computationally when considering cofacets of a simplex, we only consider finite distance neighbors. This results in a reduction in the number of simplices to consider when performing cofacet enumeration for matrix reduction and filtration construction, potentially saving both execution time and memory space if the distance matrix is "sparse" enough.

There are two uses of sparse distance matrices. One usage is to compute Vietoris-Rips barcodes up to a particular diameter threshold, using a sparsified distance matrix. This results in the same barcodes as in the dense case on a truncated filtration due to the diameter threshold. The second usage is to approximate a finite metric space at the cost of a multiplicative approximation factor on barcode lengths [113–115]. [115] has an algorithm to approximate dense distance matrices with sparse matrices for such barcode approximation and can be used with Ripser++; this approach is a particular necessity for large dense distance matrices where the persistence computation has dimension $d$ such that the resulting filtration size becomes too large.

### 5.3.13 Enumerating Cofacets of Simplices Induced by a Sparse 1-Skeleton in Ripser

The enumeration of cofacets for sparse edge graphs (sparse distance matrices involving few neighboring relations between all vertices) must be changed from Algorithm 14 for performance reasons. The sparsity of neighboring relations can significantly reduce the number of cofacets that need to be searched for. Similar to the inductive algorithm described in [105], cofacet enumeration is in Algorithm 15.

### 5.3.14 Enumerating Facets in Ripser++

Algorithm 16 shows how to enumerate facets of a simplex as needed in Algorithm 13. A facet of a simplex is enumerated by removing one of its vertices. Due to properties of the combinatorial number system, if we remove vertex indices in a decreasing order, the combinatorial indices of the generated facets will increase (the simplices will be generated in lexicographically increasing order). Algorithm 16 is used in Algorithm 13 for GPU in Ripser++ and does not depend on sparsity of the distance matrix since a facet does not introduce any new distance information. In fact, there are only $d+1$ facets of a simplex to enumerate. On GPU we can use shared memory per thread block to cache the few $d+1$ vertex indices.

**Algorithm 15:** Enumerating Cofacets of Simplex for Sparse 1-Skeletons

**Input:** $\boldsymbol{X} = \{0..n_0 - 1\}$: a finite metric space; $s$: a simplex with vertices in $\boldsymbol{X}$; vertices($\cdot$): the vertices of a simplex; cidx($\cdot$): the combinatorial index of a simplex; $\text{cidx}_{vert}(\cdot)$: calculate the combinatorial index from the vertices.

**Output:** $\boldsymbol{S}$: the facets of $s$ in lexicographically decreasing order.

```
V ← vertices(s) ;
fix some v0 ∈ V ⊆ X ;
foreach neighbor v' ≠ v0, v' ∈ X − V of v0 in decreasing order do
    foreach w ∈ V, w ≠ v0 and w ≠ v' do
        if w is a neighbor of v' then
            continue ;          /* Jump to the next iteration lof the inner loop */
        end
        else
            if all neighboring vertices to w are greater than v' then
                return ;        /* There are no more cofacets that can be enumerated */
            end
            else
                goto try_next_vertex ;                        /* Jump to line 13 */
            end
        end
    end
    s' ← cidx_vert(V ∪ v') ;
    append(S, s') ;
    try_next_vertex: ;
end
```

## 5.4 Theoretical Bounds on the Number of Apparent Pairs

Besides being an empirical fact for the existence of a large number of apparent pairs (see Section 5.6), we show theoretically that there are tight upper and lower bounds to the number of apparent pairs for a $d$-dimensional full Rips filtration on a $(d + 1)$-skeleton $X$ induced by the clique on $n_0$ points when the simplices containing maximum point $n-1$ occur oldest in the filtration. We assume the simplex-wise filtration order of 5.1.4 throughout this paper. First we prove a useful property of apparent pairs on $X$ depending on the largest indexed vertex.

**Algorithm 16:** Enumerating Facets of a Simplex

**Input:** $\boldsymbol{X} = \{0, \ldots, n_0 - 1\}$: $n_0$ points of a finite metric space; $s$: a simplex with vertices in $\boldsymbol{X}$; vertices$(\cdot)$: the vertices of a simplex; cidx$(\cdot)$: the combinatorial index of a simplex; last$(\cdot)$: the last simplex of a sequence.

**Output:** $\boldsymbol{S}$: the facets of $s$ in lexicographically increasing order.

```
Function enumerate-facets(s, S):
    V ← vertices(s) ;
    prev ← ∅ ;
    k ← |V| ;
    for v ∈ V ⊆ X in decreasing order do
        if prev ≠ ∅ then
            cidx(s′) ← cidx(last(S)) − C(v, k) + C([prev], k) ;   /* [prev] is the only element of
              the singleton prev */
        end
        else
            cidx(s′) ← cidx(last(S)) − C(v, k) ;
        end
    end
    append(S, s′) ;                                           /* Append s′ to the end of S */
    prev ← {v} ;
    k ← k − 1 ;
```

**Lemma 5.4.1** (Cofacet Lexicographic Property)**.** *For any pair of d-dimensional simplex and its oldest cofacet: $(s,t)$ of a full Rips filtration on a $(d+1)$-skeleton generated by a fully connected graph on $n_0$ points $V = \{0, \ldots, n_0 - 1\}$ where $s$ does not contain the maximum vertex $n_0 - 1$ and all d-dimensional simplices $s'$ containing maximum vertex $n_0 - 1$ have $diam(s') \leq diam(s)$, $t = (v_{d+1}, \ldots, v_0)$ must have $v_{d+1} = n_0 - 1$, where $v_{d+1} > v_d > \ldots > v_0$.*

*Proof.* Let $s = (w_d, \ldots, w_0)$, with $w_d > w_{d-1} > \ldots > w_0$ and $w_d \neq n-1$; The maximum indexed vertex $n-1$ forms a cofacet $t = (n-1, s)$ of $s$. If $\text{diam}(t) > \text{diam}(s)$, then some facet $s'$ of $t$ must be strictly younger (having larger diameter) than $s$ and must contain the vertex $n-1$. $s' = (n-1, s'')$, with $s''$ a facet of $s$. However, $s'$ is actually strictly older than $s$ by the fact that $\text{diam}(s') \leq \text{diam}(s)$ by assumption and $\text{cidx}(s') > \text{cidx}(s)$ since $n-1$ is the largest vertex index. This is a contradiction. Thus $\text{diam}(t) = \text{diam}(s)$ (since also $\text{diam}(s) \leq \text{diam}(t)$) and $\mathrm{cidx}(t)$ is largest amongst all cofacets of $s$ by the existence of point $n-1$ in $t$. Thus the oldest cofacet of $s$ is in fact $t$. $\square$

**Theorem 5.4.2.** *(Equal Diameter Bound on the Number of Apparent Pairs)*

*Let $X$ be a point set of size $n_0$. For a full Rips-filtration $(Rips_t(X))_{t\leq R=\infty}$,*

*If $d$-dimensional simplices $s' \in Rips_\infty(X)$ containing some vertex $x \in X$ have $diam(s') \leq diam(s)$ for all $d$-dimensional simplices $s \in Rips_\infty(X)$ not containing vertex $x \in X$, then the ratio of the number of $(d, d+1)$-dimensional apparent pairs to the total number of $d$-dimensional simplices satisfies the following upper and lower bounds:*

- *upper bound: $(n_0 - d - 1)/n_0$; (tight for all $n_0 \geq d+1$ and $d \geq 1$).*
- *lower bound: $1/(d+2)$; (tight for $d \geq 1$).*

*Proof.* **Upper Bound**:

If $d$-dimensional simplex $s$ has vertex $x$, so do all its cofacets $t$. Thus by Lemma 5.4.1, the set of $(d, d+1)$-dimensional apparent pairs $(s,t)$ must have $t = (v_{d+1}, ..., v_0)$ with $v_{d+1} = x$. There are at most $\binom{n_0-1}{d+1}$ such $d$-dimensional simplices $s$ by counting all possible tuples $s = (v_d, ..., v_{\mathrm{j}}, ..., v_0)$, $v_{\mathrm{j}} \in \{0, ..., n_0 - 1\} \setminus \{x\}$, $s$ a facet of $t$ not including point $x$. Since there are a total of $\binom{n_0}{d+1}$ $d$-dimensional simplices, we divide the two factors and obtain $\frac{\binom{n_0-1}{d+1}}{\binom{n_0}{d+1}} = \frac{(n_0-d-1)}{n_0}$ percentage of $d$-dimensional simplices, each of which are paired up with a $(d+1)$-dimensional simplex as an apparent pair. This percentage forms an upper bound on the number of apparent pairs.

**Tightness of the Upper Bound**:

We show that the upper bound is achievable in the special case where all diameters are equal. The condition of the theorem is certainly still satisfied. In this case we break ties for the simplex-wise refinement of the Rips filtration by considering the lexicographic order of simplices on their decreasing sorted list of vertex indices (See Section 5.1.6).

We exhibit the upper bounding case by forming the corresponding coboundary matrix. The vertex $x$ is set to permuted to $n_0 - 1$. This does not affect the algorithm since it is permutation invariant on the indices of the vertices.

By the lexicographic ordering on simplices, in the coboundary matrix there exists a staircase (moving down and to the right by one row and one column) of apparent pair entries starting from the pair $(s_1, t_1) = ((d, ..., 1, 0), (n_0 - 1, d, ..., 1, 0)) = (s_1, (n_0 - 1, s_1))$ and ending on the pair $(s_{\binom{n_0-1}{d+1}}, t_{\binom{n_0-1}{d+1}}) = ((n_0 - 2, n - 3, ..., n_0 - d - 2), (n_0 - 1, n_0 - 2, ..., n_0 - d - 2)) = (s_{\binom{n_0-1}{d+1}}, (n_0 - 1, s_{\binom{n_0-1}{d+1}}))$. See Figure 5.5, for the case of $n_0 = 5$ and $d = 1$, where the first $\binom{n_0-1}{d+1}$ columns are all apparent.

The staircase certainly is made up of apparent pairs since each entry has all zeros below and to its left, being the lowest nonzero entries of the leftmost columns of the coboundary matrix. Furthermore, the staircase spans all possible apparent pairs since all $\binom{n_0-1}{d+1}$ rows (the upper bound on number of apparent pairs) of the form $(n_0 - 1, s')$ are apparent, $s'$ an arbitrary simplex on the points $\{0, ..., n_0 - 2\}$.

Dim 1 Coboundary Matrix

→ older: lex. incr. (columns); ↓ older: lex. incr. (rows)

| (diam., simplex) | (1, (10)) | (1, (20)) | (1, (21)) | (1, (30)) | (1, (31)) | (1, (32)) | (1, (40)) | (1, (41)) | (1, (42)) | (1, (43)) |
|---|---|---|---|---|---|---|---|---|---|---|
| (1, (210)) | **1** | **1** | **1** | | | | | | | |
| (1, (310)) | **1** | | | **1** | **1** | | | | | |
| (1, (320)) | | **1** | | **1** | | **1** | | | | |
| (1, (321)) | | | **1** | | **1** | **1** | | | | |
| (1, (410)) | **1** | | | | | | **1** | **1** | | |
| (1, (420)) | | **1** | | | | | **1** | | **1** | |
| (1, (421)) | | | **1** | | | | | **1** | **1** | |
| (1, (430)) | | | | **1** | | | **1** | | | **1** |
| (1, (431)) | | | | | **1** | | | **1** | | **1** |
| (1, (432)) | | | | | | **1** | | | **1** | **1** |

**Figure 5.5.** A dimension 1 coboundary matrix of the full Rips filtration of the 2-skeleton on 5 points with all simplices of diameter 1. The yellow highlighted entries above the staircase correspond to apparent pairs.

**Lower bound**:

The largest number of cofacets of a given $d$-dimensional simplex must be $n_0 - d - 1$. Thus we will obtain a lower bound if each set of cofacets of the same diameter can be forced to be disjoint from one another. Thus we seek to find a minimal $k$ s.t. $\binom{n_0}{d+2} \le k \cdot (n_0 - d - 1)$. Upon solving for $k$, divide $k$ by $\binom{n_0}{d+1}$, the number of $d$-dimensional simplices, and we get a lower bounding ratio of $1/(d + 2)$ $d$-dimensional simplices being paired with $(d + 1)$-dimensional simplices as apparent pairs.

**Tightness of the Lower Bound**:

For every dimension $d$, a $(d+1)$-dimension simplex has $d+2$ $d$-dimensional facets. There must be exactly one apparent pair of dimension $(d, d+1)$ in such a $(d+1)$-dimension simplex. For example, if $d = 1$, the 2-simplex $X$ has one 1-simplex paired with it out of all $(3 = d+2)$ 1-simplices in $X$ (see Figure 5.4 and Figure 5.7(a)). □

### 5.4.3 The Theoretical Upper Bound Under Differing Diameters

The theoretical upper bound of Theorem 5.4.2 can still be achieved in dimension 1 with the following diameter assignments (distance matrix); let $d_1 > d_2 > \ldots > d_{n\cdot(n-1)/2} > 0$ be a sequence of diameters. We assign them in increasing lexicographic order on the 1-simplices ordered on the vertices sorted in decreasing order. For example, 1-simplex (10) with vertices 1 and 0 gets assigned $d_1$, 1-simplex (20) with vertices 2 and 0 gets assigned $d_2$. See the following distance matrix, Figure 5.6, for how the diameters are assigned.

Distance Matrix

| Vertices | 0 | 1 | 2 | 3 | 4 |
|---|---|---|---|---|---|
| 0 | 0 | 10 | 9 | 7 | 4 |
| 1 | **10** | 0 | 8 | 6 | 3 |
| 2 | **9** | **8** | 0 | 5 | 2 |
| 3 | **7** | **6** | **5** | 0 | 1 |
| 4 | **4** | **3** | **2** | **1** | 0 |

(a)

Dim 1 Coboundary Matrix

→ older

↓ older:

| (diam., simplex) | (10, (10)) | (9, (20)) | (8, (21)) | (7, (30)) | (6, (31)) | (5, (32)) | (4, (40)) | (3, (41)) | (2, (42)) | (1, (43)) |
|---|---|---|---|---|---|---|---|---|---|---|
| (10, (210)) | **1** | **1** | **1** | | | | | | | |
| (10, (310)) | **1** | | | **1** | **1** | | | | | |
| (10, (410)) | **1** | | | | | | **1** | **1** | | |
| (9, (320)) | | **1** | | **1** | | **1** | | | | |
| (9, (420)) | | **1** | | | | | **1** | | **1** | |
| (8, (321)) | | | **1** | | **1** | **1** | | | | |
| (8, (421)) | | | **1** | | | | | **1** | **1** | |
| (7, (430)) | | | | **1** | | | **1** | | | **1** |
| (6, (431)) | | | | | **1** | | | **1** | | **1** |
| (5, (432)) | | | | | | **1** | | | **1** | **1** |

(b)

**Figure 5.6.** (a) is an example distance matrix with differing edge diameters assigned in decreasing order for increasing lexicographic order on simplices. The barcodes are equivalent up to scaling so long as the distance matrix entries are in the same order (see Observation 5.6.8 in Section 5.6.6); thus we set the distance matrix entries to 1,...,10. (b) is a dimension 1 coboundary matrix of the full Rips filtration of the 2-skeleton on 5 points with simplices having diameters given in (a). The yellow highlighted entries above the staircase correspond to apparent pairs. Notice the coboundary matrix is a row permutation transformation from Figure 5.5.

Since the increasing lexicographic order from the smallest lexicographically ordered simplex $(d, ..., 1, 0)$ is still used on the columns $s$, we still have the same leftmost $\binom{n-1}{d+1}$ simplices/columns (but with a different diameter) as in the diameters all the same case (call this the original case). Furthermore, the oldest cofacet $t$ of the simplex/column $s$ must still be the same $(d+1)$-dimensional cofacet $t$ of $s$ originally. This follows by Lemma 5.4.1, since we assume the lexicographic order on columns is preserved by the diameter assignment and thus that the simplices with vertex $n-1$ are oldest in the filtration (the diameter condition in Lemma 5.4.1 will be satisfied). These oldest cofacets are all different and are the same simplices as in the original case; Since the columns under consideration are leftmost, each cofacet $t$ has as youngest facet the $s$. Thus all $(s, t)$ are apparent pairs. Thus we preserve the same apparent pairs as in the original case.

### 5.4.4 Geometric Interpretation of the Theoretical Upper Bound

Geometrically the theoretical upper bound in dimension 1 of Theorem 5.4.2 is illustrated in Figure 5.7. The construction involves adding in a point at a time in order of its vertex index. By the construction, we can alternatively count the number of apparent pairs on $n_0$ points, $T(n_0)$, by the following recurrence relation:

$$T(n_0) = T(n_0 - 1) + n_0 - 2, \tag{5.11}$$

where $T(3) = 1$ and the $n_0 - 2$ term comes from the number of edges incident to the vertex with maximum index $n_0 - 2$ in the $(n_0 - 1)$-point subcomplex that become apparent when adding the new maximum point of index $n_0 - 1$ to the subcomplex. Solving for $T(n_0)$, we get $T(n_0) = \binom{n_0-1}{2}$ as in Theorem 5.4.2. Thus, assuming the conditions of Theorem 5.4.2, $n_0 - 2$ is maximal in the recurrence.

### 5.4.5 Apparent Pairs are Heavy-Hitters for Large Point Samples

A finite metric space $X$ only requires a metric $d : X \times X \to \mathbb{R}$ on pairs of its points. Assume $d$ can be extended to a global metric on the Euclidean space $\mathbb{R}^{d_{amb}}$. We show that the apparent pairs of the Vietoris-Rips barcode have a high probability of occurring for points sampled from a hypercube.

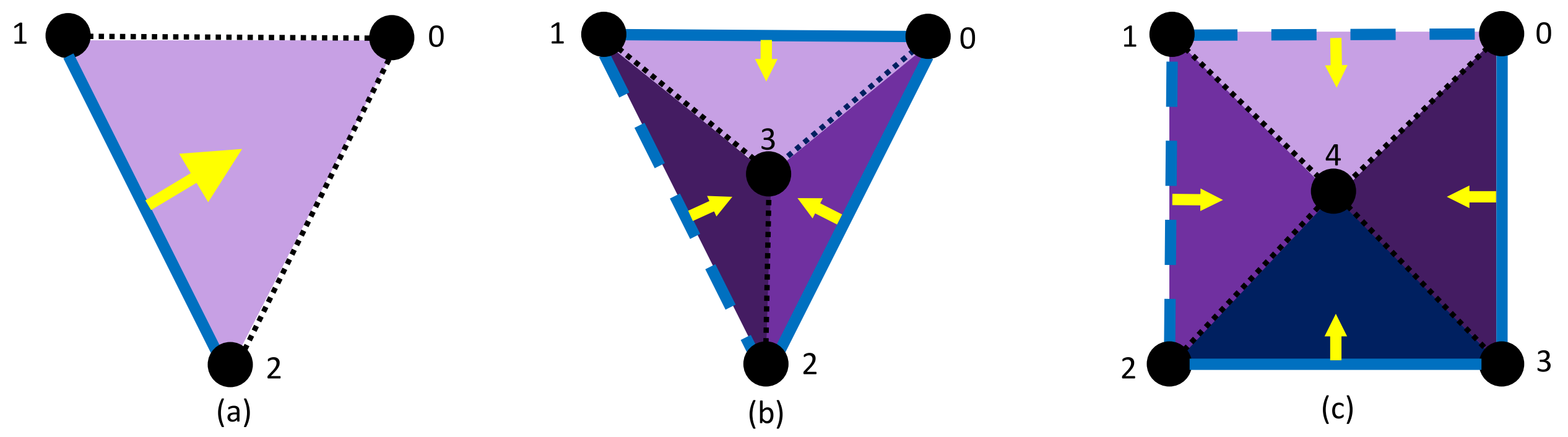

**Figure 5.7.** Geometric interpretation of the theoretical upper bound in Theorem 5.4.2. Edge distances are not to scale. (a),(b),(c) (constructed in this order) show the apparent pairs for $d = 1$ on the planar cone graph centered around the newest apex point: $n_0 - 1$ for $n_0 = 3, 4, 5$ points. The yellow arrows denote the apparent pairs: blue edges paired with purple or navy triangles. The dashed (not dotted) blue edges denote apparent edges from the previous $n_0 - 1$ point subcomplex.

We first give some equivalent characterizations of apparent pairs under various settings.

**In Terms of Raw Point Samples:**

In the following lemma we show that it suffices to check condition (1) of the apparent pair condition for the case of $p = 1$ involving matching edges with triangles.

**Lemma 5.4.6.** *Let $d_{global}$ be the metric on finite metric space $X$ with $n_0$ points no pair of which are equidistant and $p = 1 : p \leq d_{amb}$. Let $R \in \mathbb{R} \cup \{\infty\}$.*

*For any $\sigma = \{x_{k_1}, ..., x_{k_{p+1}}\} \in z_p$, associated with some $p$-dimensional creation event for $z_p \in ker(\partial_p^{K_n})$,*

*There exists a $\tau \in C_{p+1}(K_n) : \partial_{p+1}(\tau) \in im(\partial_{p+1}^{K_n})$ with $(\sigma, \tau)$ an apparent pair if any of the following equivalent conditions hold true:*

1. $\tau \supseteq \sigma$, *s.t.* $diam(\tau) = diam(\sigma)$

2. $\exists x_{k_{p+2}} \in X :$ *s.t.* $\max_{i=1}^{p+1}(d_{global}(x_{k_i}, x_{k_{p+2}})) \leq diam(\sigma) \leq R$

*Proof.* We show that the first condition is sufficient and then that the second condition of the lemma is equivalent to the first condition.

**The first condition is sufficient:**

According to Definition 5.3.4, an apparent pair

$$(\sigma, \tau) \in (\ker(\partial_p^{K_n}) \cap \binom{X}{p+1}) \times (\{\tau : \partial_{p+1}(\tau) \in \mathrm{im}(\partial_{p+1}^{K_n})\} \cap \binom{X}{p+2})) \tag{5.12}$$

has the property that:

1. $\tau$ is the oldest cofacet of $\sigma$ and
2. there is no other $\sigma'$ younger than $\sigma$ which has the same oldest cofacet.

**Satisfying Property 1 of an Apparent Pair:**

We show that $\mathrm{diam}(\tau) \leq \mathrm{diam}(\sigma)$ for some $\tau : \tau \supseteq \sigma$ implies the first condition for $(\sigma, \tau)$ to be an apparent pair:

Let $\tau^*$ be some cofacet of $\sigma$ amongst all $\tau$ with $\mathrm{diam}(\tau) \geq \mathrm{diam}(\sigma)$ that results in $\mathrm{diam}(\tau^*) = \mathrm{diam}(\sigma)$. This is the first property of an apparent pair.

**Satisfying Property 2 of an Apparent Pair:**

In order to satisfy the second property of an apparent pair, the $\sigma$ cannot have any other $\sigma'$ younger than $\sigma$ sharing $\tau^*$ as its oldest cofacet. For $p = 1$ there cannot be a $\sigma'$ when

$$\mathrm{diam}(\tau^*) = \mathrm{diam}(\sigma) \tag{5.13}$$

since $\sigma'$ would have to have

$$\mathrm{diam}(\sigma') < \mathrm{diam}(\sigma) = \mathrm{diam}(\tau^*) \tag{5.14}$$

However, this would mean that $(\sigma', \tau)$ cannot form an apparent pair and thus there are no potential conflicts.

If we can find a tie-breaking order that can replace (3) from Section 5.1.4, the latest amongst $\tau^*$ satisfying $\mathrm{diam}(\tau^*) = \mathrm{diam}(\sigma)$ becomes an oldest cofacet of $\sigma$. This would make $(\sigma, \tau)$ an apparent pair.

Since both conditions are satisfied, the pair $(\sigma, \tau^*)$, is an apparent pair according to Definition 5.3.4.

**The second condition is equivalent to the first:**

Expanding the diameter function on $\tau$ and $\sigma$ as:

$$\mathrm{diam}(\tau) = \max_{\mathrm{i,j}=1}^{p+2}(d(x_{k_\mathrm{i}}, x_{k_\mathrm{j}})) \tag{5.15}$$

and

$$\mathrm{diam}(\sigma) = \max_{\mathrm{i,j}=1}^{p+1}(d(x_{k_\mathrm{i}}, x_{k_\mathrm{j}})), \tag{5.16}$$

the equation $\mathrm{diam}(\tau) \leq \mathrm{diam}(\sigma)$ is equivalent to:

$$\max_{\mathrm{i,j}=1}^{p+2}(d(x_{k_\mathrm{i}}, x_{k_\mathrm{j}})) \leq \max_{\mathrm{i,j}=1}^{p+1}(d(x_{k_\mathrm{i}}, x_{k_\mathrm{j}})) \text{ iff} \tag{5.17a}$$

$$\max_{\mathrm{i,j}=1}^{p+2}(d(x_{k_\mathrm{i}}, x_{k_\mathrm{j}})) = \max(\max_{\mathrm{i}=1}^{p+1}(d(x_{k_\mathrm{i}}, x_{k_{p+2}})), \max_{\mathrm{i,j}=1}^{p+1}(d(x_{k_\mathrm{i}}, x_{k_\mathrm{j}}))) \leq \max_{\mathrm{i,j}=1}^{p+1}(d(x_{k_\mathrm{i}}, x_{k_\mathrm{j}})) \text{ iff} \tag{5.17b}$$

$$\max_{\mathrm{i}=1}^{p+1}(d(x_{k_\mathrm{i}}, x_{k_{p+2}})) \leq \max_{\mathrm{i,j}=1}^{p+1}(d(x_{k_\mathrm{i}}, x_{k_\mathrm{j}})) \tag{5.17c}$$

For any $\sigma \in \ker(\partial_p^{K_n})$, if any point $x \in X$ with $\max_{\mathrm{i}=1}^{p+1} d(x, x_{k_\mathrm{i}}) \leq \mathrm{diam}(\sigma) \leq R$ then amongst all such $x \in X$, we can form a cofacet $\tau : \tau \supseteq \sigma$ with points $\tau = \{x_{k_1}, ..., x_{k_{p+1}}, x^*\}$ that maximizes $\mathrm{diam}(\tau)$. □

In Lemma 5.4.6 the proof required that for $p = 1$ there would be a unique face of $\tau^*$ forming an apparent pair.

Of course, when $p \geq 2$, this uniqueness of the $p$-dimensional face may not hold anymore. This case is not easy to explain and would require understanding the incidences to $\mathrm{im}(\partial_p)$ for each diameter value $t \in \mathbb{R}$.

### $k$-th Order Nearest Neighbors

We would like to rewrite the second sufficient condition of Lemma 5.4.6 in more geometric terms.

In metric geometry the concept of a "nearest neighbor" of a point $x$ is defined as the point that is closest to the point $x$. In terms of an EMST, we know that the edge formed by every point and its **nearest neighbor** must belong to the EMST according to Kruskal's algorithm [116]. These are "apparent pairs" in the sense of persistent homology. They are the first point-edge pairs that must form when computing $H_0(\partial_1^{K_n})$.

In relation to the second sufficient condition of Lemma 5.4.6, we can generalize this definition of a nearest neighbor to its higher order generalization as follows:

**Definition 5.4.7.** *In a metric space $\mathcal{X}$ with metric $d_{global}$, for a set of points $S \subseteq \mathcal{X}$ of size $k \geq 1$ a **$k$-th order nearest neighbor** of $S$ is defined as:*

$$NN(S) \triangleq \arg\min_{x \in \mathcal{X}} (\max_{x_i \in S} (d_{global}(x, x_i))) \tag{5.18}$$

This gives us the following corollary to Lemma 5.4.6.

**Corollary 5.4.8.** *Assuming the conditions of Lemma 5.4.6, namely $p = 1$:*

*$(\sigma, \tau)$ is an apparent pair if:*

$$d_{global}(NN(\sigma), x_i) \leq diam(\sigma), \forall x_i \in \sigma \tag{5.19}$$

*Proof.* This directly follows by the second condition of Lemma 5.4.6. □

Thus, apparent pairs $(\sigma, \tau)$ for $\sigma$ a 1-dimensional simplex can be interpreted as a "conditional 2-th order nearest neighbor criterion" for $\sigma$.

(5.20) **The Relationship between the Theoretical Upper Bound and the Combinatorial Index:**

In Section 5.4.3, there is a proof of the upper bound on the number of apparent pairs under differing diameters. In that section, we showed that for any $p \geq 1$ we have that the point maximally indexed as $n_0 - 1$ over $n_0$ points acts as a $p+1$-nearest neighbor for every $p$-dimensional $\sigma \subseteq \binom{X}{p+1}$.

The distances between pairs of vertices are chosen in a way so that the lexicographic order and these distances are in opposite order. Thus, the largest indexed vertex acts as the tip of a "fat cone" as in Figure 5.7. Since this cone is determined by a generalized nearest neighbor to the previous points, it destroys all existing cycles with the faces of the cone itself.

**In terms of the Euclidean Minimum Spanning Tree:**

For the computation of 0-dimensional persistent homology, we knew that $\{\tau : \partial_1(\tau) \in \mathrm{im}(\partial_1(K_n))\}$ is the formal sum of edges from the Euclidean minimum spanning tree on $X$. See Definition 2.5.4 for the definition of an EMST. The kernel: $\ker(\partial_1(K_n))$ is represented by the formal sum of complementary edges of the spanning tree of $X$.

For Vietoris-Rips complexes, only pairs of points and their distances determine any simplex. We can show that apparent pairs maintain complementary edges on the EMST of a set of $n_0$ points $X$.

**Lemma 5.4.9.** *Let $X \subseteq \mathbb{R}^{n_0 \times d_{amb}}$ be a set of $n_0$ points and $EMST(X)$ be the Euclidean minimum spanning tree on $X$.*

*For any $p \geq 1$, the pair $(\sigma, \tau) \in (z_p, b_{p+1})$, with*

$$(z_p, b_{p+1}) \in (ker(\partial_p^{K_n}) \cap \binom{X}{p+1}) \times (\{\tau : \partial_{p+1}(\tau) \in im(\partial_{p+1}^{K_n})\} \cap \binom{X}{p+2}) \tag{5.21}$$

*is apparent then*

$$u_\sigma, v_\sigma = \arg\max_{u,v \in \sigma} \|u - v\|_2 \tag{5.22}$$

*is a complementary edge.*

*Proof.* For each $p \geq 1$ :

If $(\sigma, \tau)$ is apparent, then according to Lemma 5.4.6, we simply need $\mathrm{diam}(\tau) \leq \mathrm{diam}(\sigma)$ for some $p+1$-dimensional cofacet $\tau : \tau \supseteq \sigma$.

This means that $\tau$ shares the same maximum distance pair of points $\{u_\sigma, v_\sigma\} \subseteq \tau$ as $\sigma \subseteq \tau$. If this pair were not a complementary edge, then it would belong to the EMST. However, by definition of minimum spanning tree, all edges on the EMST are shorter than complementary edges.

□

### Lower Bounds under the Uniform Sampling Condition

Based on our understanding of apparent pairs in terms of the point samples, in the following we specify a generic sampling distribution. This is the uniform independent sampling condition on a hypercube.

**Condition 5.4.10.** *[Uniform Independent Sampling with Constant Ambient Dimension]*

$$P(x) \triangleq P(x \sim U([0,1]^{d_{amb}})) \tag{5.23}$$

We show that there is a lower bound on the probability of achieving an apparent pair for a given $p$-dimensional simplex $\sigma$. This lower bound holds in the case of uniform independent sampling with growing ambient dimension from Condition 5.4.10.

**Lemma 5.4.11.** *Let $n_0 \in \mathbb{N}$ and letting $X$ be a point sample of size $n_0$ from the uniform sampling distribution of Condition 5.4.10.*

*For any p-simplex $\sigma \in ker(\partial_p^{K_n}) \cap \binom{X}{(p+1)}$ with $p = 1$,*

*We then have:*

$$P(\exists \tau : (\sigma, \tau) \text{ is apparent} \,|\, diam(\sigma) = t) \geq 1 - (1 - \frac{t^{d_{amb}}}{(d_{amb}+1)!2^{p+1}})^{n_0} \tag{5.24}$$

*Proof.* Using Lemma 5.4.6, we know that given $p$-dimensional simplex $\sigma \in K_n$ with $\sigma = \{x_1, ..., x_{p+1}\}$ if all $x \in X$ have that $\max_{\mathrm{i}=1}^{p+1} d(x, x_\mathrm{i}) \leq t$. Thus, the point $x \sim P(x)$ has the property of forming a $p+1$-dimensional simplex $\tau$ that pairs with $\sigma$ as an apparent pair if and only if each $x$ belongs to the intersection of the $p+1$ $t$-radius balls centered about $x_\mathrm{i}, \mathrm{i} = 1, ..., p+1$.

Since $P(x)$ is a uniform distribution on the hypercube $([0,1])^{d_{amb}}$, we can take the complementary Euclidean volume of the intersection of $p+1$ balls of radius $t$ in $d_{amb}$ dimensions to get the probability of not having an apparent pairing $(\sigma, \tau)$ for some $\tau$. This gives the complementary probability of a point belonging to this intersection: $1 - P(\bigcap_{\mathrm{i}=1}^{p+1} B(x_\mathrm{i}, t))$.

If none of the $n_0 - (p+1)$ points not in $\sigma$ belong to $\bigcap_{\mathrm{i}=1}^{p+1} B(x_\mathrm{i}, t)$, then there is no cofacet $\tau$ that makes an apparent pair $(\sigma, \tau)$. Thus, the complementary probability is a lower bound to $P(\exists \tau : (\sigma, \tau) \text{ is apparent} \,|\, \mathrm{diam}(\sigma) = t)$.

We now compute this complementary probability:

We know that

$$\mathrm{BB}(\mathrm{Cyl}(\sigma, 2t)) \supseteq \mathrm{Cyl}(\sigma, 2t) \supseteq \bigcap_{\mathrm{i}=1}^{p+1} B(x_\mathrm{i}, t) \supseteq \mathrm{inscpoly}(\{x_\mathrm{i}\}_{\mathrm{i}=1}^{p+1})) \tag{5.25}$$

where $\mathrm{Cyl}(\sigma, 2t) \triangleq \sigma \times [-t, t]$, $\mathrm{BB}(S)$ is the $d_{amb}$-dimensional bounding box of a set of points $S$, and $\mathrm{inscpoly}(\{x_\mathrm{i}\}_{\mathrm{i}=1}^{p+1}))$ is a polytope that is contained in $\bigcap_{\mathrm{i}=1}^{p+1} B(x_\mathrm{i}, t)$. This polytope exists by Proposition 2.5.26.

We have that:

$$\mathrm{BB}(\mathrm{Cyl}(\sigma, 2t)) \leq \frac{(2t)^{d_{amb}}}{2^p} \tag{5.26a}$$

$$\mathrm{vol}(\mathrm{Cyl}(\sigma, 2t)) = \mathrm{vol}(\sigma)(2t) \tag{5.26b}$$

$$\begin{aligned} \mathrm{vol}(\mathrm{inscpoly}(\{x_\mathrm{i}\}_{\mathrm{i}=1}^{p+1})) &\geq \frac{2^{d_{amb}} \mathrm{vol}(\Delta_{d_{amb}+1,t})}{t 2^p} \\ &= \frac{1}{(d_{amb}+1)!} (\frac{t}{\sqrt{2}})^{d_{amb}+1} \frac{2^{d_{amb}}}{2^p} \frac{1}{t} \end{aligned} \tag{5.26c}$$

where in the inequalities for the inscribed polytope volume, the first lower bound is the volume of a $t$-equilateral $d_{amb}$-dimensional simplex with fixed $p$-dimensional face that is reflected across $\binom{d_{amb}+1}{d_{amb}} - (p+1)$ faces. Such a union of simplices is inscribed in $\bigcap_{\mathrm{i}=1}^{p+1} B(x_\mathrm{i}, t)$ where the multiplicative factor $2^{(d_{amb}-p)}$ in the lower bound on $\mathrm{vol}(\mathrm{inscpoly}(\{x_\mathrm{i}\}_{\mathrm{i}=1}^{p+1}))$ comes from the $\binom{d_{amb}+1}{d_{amb}} - (p+1) = d_{amb} - p$ reflections on the faces of the $t$-equilateral $d_{amb}$-dimensional simplex. See Proposition 2.5.26.

This results in the following probabilistic inequalities:

$$\frac{(2t)^{d_{amb}}}{2^p} \geq \mathrm{vol}(\sigma)(2t) \geq \mathrm{vol}(\bigcap_{\mathrm{i}=1}^{p+1} B(x_\mathrm{i}, t)) \geq \frac{1}{(d_{amb}+1)!} (\frac{t}{\sqrt{2}})^{d_{amb}+1} \frac{2^{d_{amb}}}{2^p} \frac{1}{t} \tag{5.27}$$

Thus

$$\frac{(2t)^{d_{amb}}}{2^p} \geq P(\exists \tau : (\sigma, \tau) \text{ is apparent} \mid \mathrm{diam}(\sigma) = t) = 1 - (1 - P(\bigcap_{\mathrm{i}=1}^{p+1} B(x_\mathrm{i}, t)))^{n_0-(p+1)} \tag{5.28a}$$

$$\geq 1 - (1 - \frac{1}{(d_{amb}+1)!}(\frac{t}{\sqrt{2}})^{d_{amb}+1}\frac{2^{d_{amb}}}{2^p}\frac{1}{t})^{n_0-(p+1)} \tag{5.28b}$$

$$\geq 1 - (1 - \frac{t^{d_{amb}}}{(d_{amb}+1)!2^{p+1}})^{n_0} \tag{5.28c}$$

□

**A Remark on the Event of forming an Apparent Pair:**

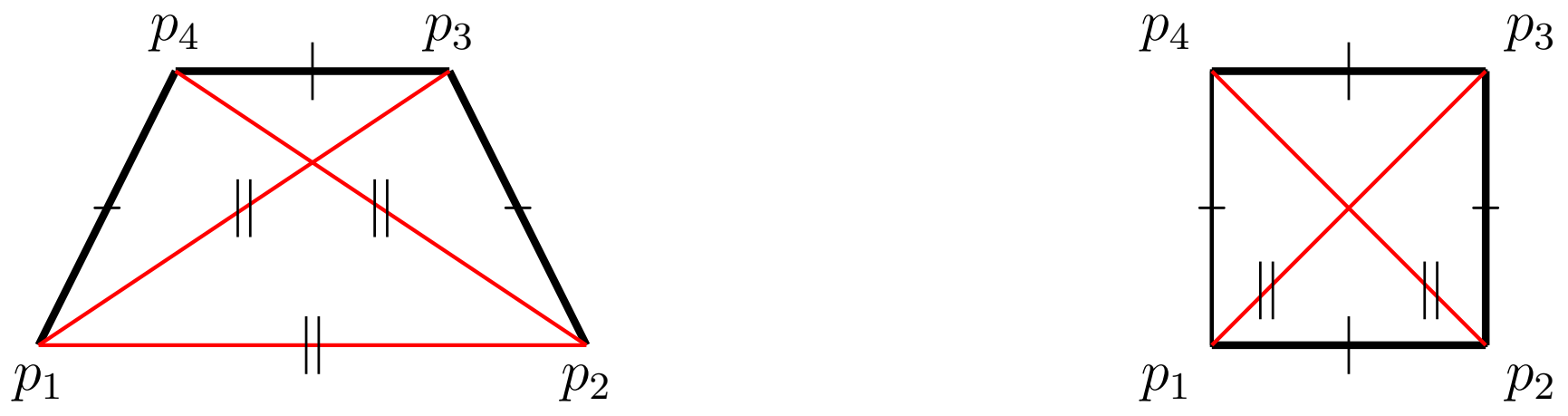

(a) A random trapezoid from 4 i.i.d. sampled points

(b) A random square from 4 i.i.d. sampled points

**Figure 5.8.** A set of 4 i.i.d. points $\{p_1, p_2, p_3, p_4\}$ sampled from some common distribution whose pair of points form edges. All the edges with the same number of ticks have the same length $d$. The edges with more than one tick have length $d' > d$. The left figure has the sample of points that form a trapezoid and the right figure has the sample of points form a square.

**Remark 5.4.12.** *Consider a random sample of 4 points from the uniform distribution in $d_{amb}$ dimensions, a Rips-complex can form by either:*

1. *Adding diagonals before forming a full-Rips complex on 4 points or*

2. *Creating a 4-cycle before closing with diagonals to form a full-Rips complex*

*In the later, case (2), where the cycle cannot close up with its own apparent edge, the event would be considered spurious. For a large number of samples there should be no distinguishable spacing. In particular, with just one more point in the interior of the square, atleast one of the sides of the square becomes apparent. It would be suspected that with a large number of points these spurious cycles would not appear.*

**Remark 5.4.13.** *Furthermore, the conditional event on $\sigma \in z_p$, for some $z_p \in ker(\partial_p^{K_n})$ with $diam(\sigma) = t$ from Lemma 5.4.11:*

$$E_{\sigma,t} \triangleq (\exists \tau : (\sigma, \tau) \text{ is apparent } \mid (diam(\sigma) = t)) \tag{5.29}$$

*is not independent, meaning:*

$$\exists \sigma, \sigma' \in ker(\partial_p^{K_n}), t \in \mathbb{R} : E_{\sigma,t} \not\perp E_{\sigma',t}, \tag{5.30}$$

*For example, in the left figure of Figure 5.1, we have a VR complex on a random sample of 4 points. Knowing that the longest edge $\sigma = \{p_1, p_2\}$ has a $\tau$ with $(\sigma, \tau)$ apparent, it must be the case that all the chords $\sigma'$, e.g. $\sigma' = \{p_1, p_3\}$ must be apparent. Thus the diagonal edges must be apparent when the longest edge in the 4-cycle is apparent.*

*Thus, $E_{\sigma,t} \not\perp E_{\sigma,t'}$.*

According to Lemma 5.4.11, we thus have the following implication:

$$1 - \left(1 - \frac{t^{d_{amb}}}{(d_{amb}+1)!2^{p+1}}\right)^{n_0} \geq q \Rightarrow P(\exists \tau : (\sigma, \tau) \text{ is apparent} \mid \text{diam}(\sigma) = t) \geq q \tag{5.31}$$

Solving for $t$, this gives us the lower bound on all the $t$ so that the implication of Equation 5.31 can hold:

$$t_q^* \triangleq [((d_{amb}+1)!2^{p+1})(1 - (1-q)^{\frac{1}{n_0}})]^{\frac{1}{d_{amb}}} \tag{5.32}$$

Taking $n_0 \to \infty$, we see that $t_q^* \to 0$, so that the expression in Equation 5.24 gives a lower bound of 0 probability. Intuitively this is because when there are unlimited number of points in a compact domain with no lower bound on the closest pair of points, then an arbitrarily large number of "thin", non-apparent $p+1$-dimensional simplices $\sigma$ could occur.

This is why we must condition on having a lower bound on the closest pair of points in any sample $X \sim P(x)^{n_0}$. Since $P(x)$ has compact support, this is equivalent to upper bounding the aspect ratio.

We show in the following theorem a sufficient condition to obtain atleast probability of $q$ chance for any $\sigma \in z_p$, for some $z_p \in \ker(\partial_p^{K_n})$ of being part of an apparent pair.

Let $\rho = a(\text{Rips}_t(X))$, this gives the lower bound on $t$ as $\frac{1}{\rho}$.

We group together the terms in Equation 5.31 involving $\rho$ into a single variable. This defined below:

$$\Delta \triangleq (d_{amb}+1)!2^{p+1}\rho^{(d_{amb})} \tag{5.33}$$

**Theorem 5.4.14.** *Under the same conditions as Lemma 5.4.6 where $X$ is a point sample of size $n_0$ from the uniform sampling of Condition 5.4.10 and $\rho = a(Rips_t(X))$.*

*For any $q \in [0,1]$ and for any p-simplex $\sigma \in ker(\partial_p^{K_n}) \cap \binom{X}{(p+1)}$ with $p = 1$,*

*Let*

$$n^*(q) \triangleq \frac{\log(1-q)}{\log(1-\frac{1}{\Delta})} \tag{5.34}$$

*We must have that*

$$n_0 \geq n^*(q) \Rightarrow P(\exists \tau : (\sigma,\tau) \text{ is apparent}) \geq q \tag{5.35}$$

*Furthermore, if we enforce aspect ratio $\rho = 1$ as in Theorem 5.4.2, we obtain the optimal lower bound on $n^*(q)$.*

*Proof.* Using that $\Delta = (d_{amb}+1)!2^{p+1}\rho^{d_{amb}}$, we have:

$$(1-(1-\frac{1}{\Delta})^{n_0}) \geq q \Rightarrow P(\exists \tau : (\sigma,\tau) \text{ is apparent}) \geq q \tag{5.36}$$

Bringing $n_0$ to one side, we obtain the desired implication from Equation 5.35. □

### 5.4.15 The Expected Complexity of Ripser++

**Theorem 5.4.16.** *Assuming the conditions of Lemma 5.4.11, let*

$$1 - q_{n_0,p} \triangleq (\frac{C}{n_0})^{p\log(1+\frac{1}{\Delta})\log(\Delta)}, p = 1 \tag{5.37}$$

*for some constant $C > 0$.*

*When $n_0$ satisfies*

$$n_0 \geq \Omega(\max(C^{\frac{D\log(1+\frac{1}{\Delta})\log(\Delta)}{D(\log(1+\frac{1}{\Delta})\log(\Delta)-1)}}, D^2\log^2(\Delta)) \tag{5.38}$$

*then the expected Betti numbers of the Rips complex satisfy* $\mathbb{E}[\beta_1] = O(1)$

*This makes the expected work complexity of Ripser++ for* $p = 1$ *to be:*

$$O(\beta_p n_p \log(n_0) + n_{p+1}) \tag{5.39}$$

*with expected depth complexity of*

$$O(\beta_p n_p \log(n_0) + n_0(p+1)) \tag{5.40}$$

*which is the complexity of computing ker*$([\partial_2])$.

*To compute the Betti number only, the expected complexity of Ripser++ is*

$$O(n_{p+1}) \tag{5.41}$$

*and the expected depth complexity is*

$$O(n_0(p+1)) \tag{5.42}$$

*Proof.* **1. Using $q_{n_0,D}$, there is an asymptotically constant Expected Number of Nonapparent Simplices $\sigma$ for column $[\sigma] \in$ ker$([\partial_1])$:**

We can check that

$$C^{\frac{D\log(1+\frac{1}{\Delta})\log(\Delta)}{D(\log(1+\frac{1}{\Delta})\log(\Delta)-1)}} = C^{\frac{p\log(1+\frac{1}{\Delta})\log(\Delta)}{p(\log(1+\frac{1}{\Delta})\log(\Delta)-1)}}, p = 1 \tag{5.43}$$

If $n_0 \geq C^{\frac{D\log(1+\frac{1}{\Delta})\log(\Delta)}{D(\log(1+\frac{1}{\Delta})\log(\Delta)-1)}}$, then we obtain from the definition of $1-q_{n_0,p}$ and Equation 5.43 that for $p = 1$:

$$(1 - q_{n_0,p})n_0^p = n_0^p(\frac{C}{n_0})^{p\log(1+\frac{1}{\Delta})\log(\Delta)} = (\frac{C^{p\log(1+\frac{1}{\Delta})\log(\Delta)}}{n_0^{p(\log(1+\frac{1}{\Delta})\log(\Delta)-1)}}) \tag{5.44a}$$

$$= \frac{C^{D\log(1+\frac{1}{\Delta})\log(\Delta)}}{(n_0)^{D(\log(1+\frac{1}{\Delta})\log(\Delta)-1)}} = O(1) \tag{5.44b}$$

where the last equality with $O(1)$ follows by $n_0 \geq C^{\frac{D\log(1+\frac{1}{\Delta})\log(\Delta)}{D\log(1+\frac{1}{\Delta})\log(\Delta)-1}}$.

Since $n_0^p$ is an upper bound on the number of simplices $\sigma \in \ker(\partial_p^{K_n}) \cap \binom{X}{(p+1)}$ that can be apparent, by union bound, we have the following bound on the expected number of simplices $\sigma$ that cannot be apparent:

$$\mathbb{E}[\mathbf{1}[\sigma \in \ker(\partial_p^{K_n}) \cap \binom{X}{(p+1)} : \nexists\, \tau, (\sigma,\tau) \text{ is apparent}]] \leq (1-q_{n_0,p})n_0^p = O(1) \tag{5.45}$$

**2. Sufficient Condition to achieving $P(\exists\tau : (\sigma,\tau)$ is apparent$) \geq q_{n_0,D}$**

Let $n^*(q_{n_0,D}) = \frac{\log((1-q_{n_0,D})^{-1})}{\log((1-\frac{1}{\Delta})^{-1})}$ from Theorem 5.4.14. We know that:

$$n^*(q_{n_0,D}) = \frac{\log((\frac{n_0}{C})^{D\log(1+\frac{1}{\Delta})\log(\Delta)})}{\log((1-\frac{1}{\Delta})^{-1})} \leq \frac{D\log(1+\frac{1}{\Delta})\log(\Delta)\log(\frac{n_0}{C})}{\log(1+\frac{1}{\Delta})} \tag{5.46a}$$

$$\leq D\log(\Delta)\log(\frac{n_0}{C}) \leq D\log(\Delta)\sqrt{n_0} \tag{5.46b}$$

This allows us to design a sufficient condition for Theorem 5.4.14, namely that:

$$n_0 \geq D\log(\Delta)\sqrt{n_0} \Rightarrow \tag{5.47a}$$

$$n_0 \geq n^*(q_{n_0,D}) = \frac{\log((1-q_{n_0,D})^{-1})}{\log((1-\frac{1}{\Delta})^{-1})} \Rightarrow P(\exists\tau : (\sigma,\tau) \text{ is apparent}) \geq q_{n_0,D} \tag{5.47b}$$

Thus, taking both sufficient conditions: $n_0 \geq D\log(\Delta)\sqrt{n_0}$ and $n_0 \geq C^{\frac{D\log(1+\frac{1}{\Delta})\log(\Delta)}{D\log(1+\frac{1}{\Delta})\log(\Delta)-1}}$, we have that:

$$n_0 \geq \Omega(\max(C^{\frac{D\log(1+\frac{1}{\Delta})\log(\Delta)}{D\log(1+\frac{1}{\Delta})\log(\Delta)-1}}, D^2\log^2(\Delta))) \tag{5.48a}$$

$$\Rightarrow \mathbb{E}[\mathbf{1}[\sigma : [\sigma] \in \ker([\partial_1]), \nexists\, \tau, (\sigma,\tau) \text{ is apparent}]] = O(1) \tag{5.48b}$$

Since $\mathbb{E}[\mathbf{1}[\sigma : [\sigma] \in \ker([\partial_1]), \nexists\, \tau, (\sigma,\tau) \text{ is apparent}]] \geq \mathbb{E}[\beta_1]$, we must have $\mathbb{E}[\beta_1] = O(1)$.

Theorem 3.2.28 gives a complexity bound for computing the dual matrix reduction problem of Problem 3.2.2. Since Ripser++ is also based on dual matrix reduction, we obtain the work complexity bound for Ripser++ is:

$$O(\beta_p n_p \log(n_0) + n_{p+1}) \tag{5.49}$$

where the $O(n_{p+1})$ is the total amount of simplices from the coboundary that must be enumerated.

Since Ripser++ enumerates these coboundaries in parallel, this part of the work complexity becomes $O(n_0(p+1))$, which is the work it takes to enumerate a single coboundary.

This thus gives a depth complexity of $O(\beta_p n_p \log(n_0) + n_0(p+1))$ □

### 5.4.17 From Geometry to Combinatorics through Locality

A space $X$ where every sample $x \in X$ can intervene with an independent **local metric**, meaning that for every $x \in X$ there is a function $d_x : X \to \mathbb{R}^+ \cup \{\infty\}$, is equivalent to a weighted directed graph. Certainly each $x \in X$ has a neighborhood defined by $\{y \in X : d_x(y) \in \mathbb{R}^+\}$, thus we can form the graph

$$G_{X,d_\bullet} \triangleq (X, \{(x,y) : d_x(y) < \infty\}) \tag{5.50}$$

If there is symmetry across local metrics on pairs of samples, meaning: $d_x(y) = d_y(x) \forall x, y \in X$ then $(X, d_x)$ forms a weighted undirected complete graph also defined similarly as

$$G_{X,d_\bullet,sym} = (X, \{(x,y) : d_x(y) < \infty : d_x(y) = d_y(x) \forall x, y \in X\}) \tag{5.51}$$

As in the metric case, we can define an **aspect ratio** on a weighted (un)directed graph $G_{X,d_\bullet}$, denoted $a(G_{X,d_\bullet})_{d_\bullet}$, by:

$$a(G_{X,d_\bullet})_{d_\bullet} \triangleq \frac{\max_{x,y \in X} d_x(y)}{\min_{x,y \in X} d_x(y)} \tag{5.52}$$

In analogy to the case of the uniform distribution on the hypercube, for $G_{X,d_\bullet,sym}$ we can also derive similar probabilistic bounds.

**Lemma 5.4.18.** *For $p = 1$ and $p$-dimensional simplex $\sigma \in ker(\partial_p) \cap \binom{X}{(p+1)}$ with $diam(\sigma) = t$, for $t \in [0,1]$. Let $G_{X,d_\bullet}$ denote a complete weighted undirected graph.*

*If the local metrics $d_x$ are chosen for each $x \in X$ to satisfy $d_x(y) \in U([0,1]), \forall y \in X$ with the only dependency of $d_x(y) = d_y(x)$, then*

$$P(\exists \tau : (\sigma, \tau) \text{ is apparent} \mid diam(\sigma) = t) \geq t^{p+1} \tag{5.53}$$

*Proof.* According to Lemma 5.4.6, $\sigma \in z_p$, for some $z_p \in \ker(\partial_p)$ has some $\tau : \partial_{p+1}(\tau) \in \mathrm{im}(\partial_{p+1}^{K_n})$ with $(\sigma, \tau)$ apparent if there is a $x \in X$ with

$$d_x(x_i) \leq t, \forall x_i \in \sigma \tag{5.54}$$

Such a $x \in X$ can be found if the local metric $d_x$ satisfies the distance bounds of Equation 5.54. By assumption of the chosen local metrics $d_x$, this will occur with probability $t^{p+1}$. This gives the lower bound. □

**Theorem 5.4.19.** *Assuming the conditions of Lemma 5.4.18, let $a(Rips(X))_{d_\bullet} = \rho$ and for $p = 1$,*

$$(1 - q_{n_0,p}) \triangleq (\frac{C}{n_0})^p \tag{5.55}$$

*If $\rho$ satisfies:*

$$\frac{\rho^{-(p+1)}\rho^{-(\binom{p+1}{2}+1)}}{p + 1 + \binom{p+1}{2}} \geq q_{n_0,p}, \tag{5.56}$$

*then the expected Betti numbers of the Rips complex satisfy $\mathbb{E}[\beta_1] = O(1)$.*

*As in Theorem 5.4.14, we get similar expected complexity bounds for $p = 1$.*

*The expected work complexity of Ripser++ is*

$$O(\beta_p n_p \log(n_0) + n_{p+1}) \tag{5.57}$$

*with expected depth complexity of*

$$O(\beta_p n_p \log(n_0) + n_0(p+1)) \tag{5.58}$$

*which is the complexity of computing ker*$([\partial_1])$.

*Proof.* For each $\sigma \in \ker(\partial_p^{K_n}) \cap \binom{X}{(p+1)}$, we can compute that

$$P(\exists \tau : (\sigma,\tau) \text{ is apparent}) \tag{5.59a}$$

$$\geq \int_{t\in[0,1]} P(\exists \tau : (\sigma,\tau) \text{ is apparent} \mid \text{diam}(\sigma) = t) P(\text{diam}(\sigma) = t) dt \tag{5.59b}$$

$$\geq \int_{t\in[0,1]} t^{p+1} t^{\binom{p+1}{2}} dt \tag{5.59c}$$

$$\geq \frac{t^{p+1} t^{\binom{p+1}{2}+1}}{p+1+\binom{p+1}{2}} \tag{5.59d}$$

$$\geq \frac{\rho^{-(p+1)} \rho^{-(\binom{p+1}{2}+1)}}{p+1+\binom{p+1}{2}} \tag{5.59e}$$

$$\geq q_{n_0,p} \tag{5.59f}$$

Taking the complementary event of $E_\sigma \triangleq (\exists \tau : (\sigma,\tau)$ is apparent), and using the definition of $1 - q_{n_0,p}$, we have by union bound that:

$$\mathbb{E}[\beta_1] \leq \mathbb{E}[\mathbf{1}[\sigma \in \ker(\partial_p^{K_n}) : \nexists\, \tau : (\sigma,\tau) \text{ is apparent}]] \leq (1 - q_{n_0,p}) n_0^p \tag{5.60a}$$

$$= \left(1 - \frac{\rho^{-(p+1)} \rho^{-(\binom{p+1}{2}+1)}}{p+1+\binom{p+1}{2}}\right) n_0^p \leq \frac{C n_0^p}{n_0^p} = O(1) \tag{5.60b}$$

The remainder of the proof follows verbatim as in Theorem 5.4.14. □

In Theorem 5.4.19, the bound on the aspect ratio of Equation 5.56, requires a bound on $\rho$. In fact, $\rho$ must be bounded by a constant.

**Corollary 5.4.20.** *In order for Theorem 5.4.19 to hold, the aspect ratio must satisfy:*

$$\rho = O(1) \tag{5.61}$$

*Proof.* Writing out the equality for $1 - q_{n_0,p}$.

$$1 - q_{n_0,p} = 1 - \frac{\rho^{-(p+1)}\rho^{-(\binom{p+1}{2}+1)}}{p + 1 + \binom{p+1}{2}} \leq \frac{C}{n_0^p} \tag{5.62}$$

Taking a lower bound on $1 - q_{n_0,p}$:

$$\exists C' > 0 : 1 - \frac{1}{C'\rho^{p^2}p^2} \leq 1 - q_{n_0,p} \leq \frac{C}{n_0^p} \tag{5.63}$$

Since $p$ is a constant, we must have that for $n_0 \to \infty$ that $\rho \not\to \infty$. If $\rho \to \infty$, then we would have $1 \leq \frac{C}{n_0^p}$ for $n_0 \to \infty$, Contradiction.

Thus, $\rho$ cannot have any dependency with $n_0$. It is thus bounded by a constant. □

**Remark 5.4.21.** *In Theorem 5.4.16, the aspect ratio can depend on $n_0$ and still achieve the expected complexities. This is because in Lemma 5.4.11, we have the following lower bound:*

$$P(\exists \tau : (\sigma, \tau) \textit{ is apparent} \mid \textit{diam}(\sigma) = t) \geq (1 - (1 - \frac{1}{(d_{amb}+1)!2^{p+1}\rho^{d_{amb}}})^{n_0}) \tag{5.64}$$

*For large $n_0$ this goes to 1 even when $\rho$ has a dependency on $n_0$.*

*However, in Lemma 5.4.18, when we assume that each edge obtains a distance i.i.d. uniform from $[0,1]$, we have that:*

$$P(\exists \tau : (\sigma, \tau) \textit{ is apparent} \mid \textit{diam}(\sigma) = t) \geq t^{p+1} \tag{5.65}$$

*This does not go to one with $n_0$, which is the reason for the stronger requirement of $\rho = O(1)$. Without the compactness from the uniform distribution on the points, increasing the number of points does not bring the points closer to each other. This prevents an asymptotic convergence with respect to $n_0$ for the probability of an apparent pair. This puts the constant bound on $\rho$.*

## 5.5 GPU and System Kernel Development for Ripser++

We have laid the mathematical and algorithmic foundation for Ripser++ in Section 5.3. GPU has two distinguished features to deliver high performance compared with CPU. First, GPU has very high memory bandwidth for massively parallel data accesses between computing cores and the on-chip memory (or device memory). Second, Warp is a basic scheduling unit consisting of multiple threads. GPU is able to hide memory access latency by warp switching, benefiting from zero-overhead scheduling by hardware.

To turn the elegant mathematics proofs and parallel algorithms into GPU-accelerated computation for high performance, we must address several technical challenges. First, the capacity of GPU device memory is much smaller than the main memory in CPU. Therefore, GPU memory performance is critical for the success of Ripser++. Second, only a portion of the computation is suitable for GPU acceleration. Ripser++ must be a hybrid system, effectively switching between GPU and CPU, which increases system development complexity. Finally, we aim to provide high performance computing service in the TDA community without a requirement for users to do any GPU programming. Thus, the interface of Riper++ is GPU independent and easy to use. In this section, we will explain how we address these issues for Ripser++.

### 5.5.1 Core System Optimizations

The expected performance gain of finding apparent pairs on GPU comes from not only the parallel computation on thousands of cores but also the concurrent memory accesses at a high bandwidth, where the apparent pairs can be efficiently aggregated. In a sequential context, an apparent pair (a row index and a column index) of the coboundary matrix may be kept in a hashmap as a key-value pair with the complexity of $O(1)$. However building a hashmap is not as fast as constructing a sorted continuous array [117] in parallel. So in our implementation, the apparent pairs are represented by a key-value pair $(t, s)$ where $t$ is the oldest cofacet of simplex $s$ and stored in an aligned continuous array of pairs. This slightly lowers the read performance because we need a binary search to locate a desired apparent pair. But this is a cost-effective implementation since the number of insertions of apparent pairs are actually three orders of magnitude higher than that of reads (See Table 5.3 in Section 5.6) after finding apparent pairs. Figure 5.9 presents how we collect apparent pairs on GPU, where each thread works on a column of coboundary matrix and writes to the output array in parallel.

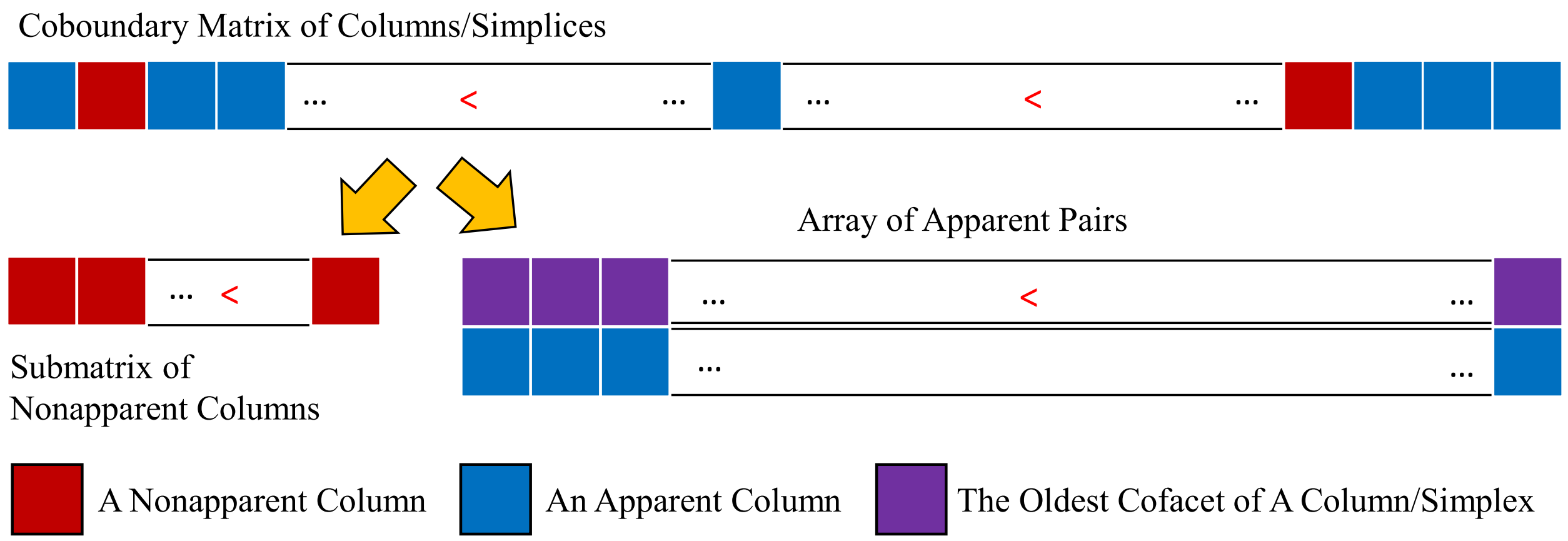

**Figure 5.9.** After finding apparent pairs, we partition the coboundary matrix columns into apparent and nonapparent columns. The apparent columns are sorted by the coboundary matrix row (the oldest cofacet of an apparent column) and stored in an array of pairs; while the nonapparent columns are collected and sorted by coboundary matrix order in another array for submatrix reduction.

On top of the sorted array, we add a hashmap as one more layer to exclusively store persistence pairs discovered during the submatrix reduction. Apparent pairs, and in fact persistence pairs in general can be stored as key-value pairs since no two persistence pairs $(s,t)$ and $(s',t')$ have the possibility of $s = s'$ or $t = t'$, as any equality would contradict Algorithm 11. Figure 5.10 explains our two layer design of a key-value storage data structure for persistence pairs in detail.

### 5.5.2 Filtration Construction with Clearing

Before entering the matrix reduction phase, the input simplex-wise filtration must be constructed and simplified to form coboundary matrix columns. We call this Filtration Construction with Clearing. This requires two steps: filtering and sorting. Both of which can be done in parallel, in fact massively in parallel. Filtering removes simplices that we don't need to reduce as they are equivalent to zeroed columns. As presented in Algorithm 17, these simplices are filtered out: the ones having higher diameters than the threshold (see Section 5.2.12 for the enclosing radius condition that can be applied even when no threshold is explicitly specified) and paired simplices (the clearing lemma [69]).

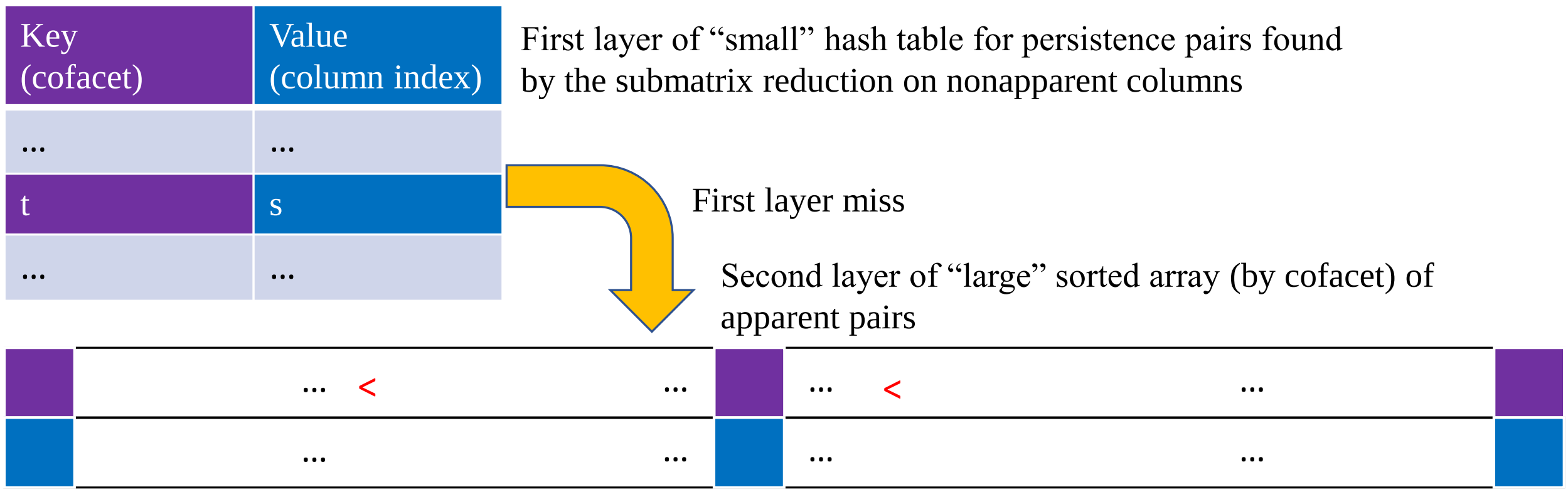

**Figure 5.10.** Two-layer data structure for persistence pairs. Apparent pair insertion to the second layer of the data structure is illustrated in Figure 5.9, followed by persistence pair insertion to a small hashmap during the submatrix reduction on CPU. A key-value read during submatrix reduction involves atmost two steps: first, check the hashmap; second, if the key is not found in the hashmap, use a binary search over the sorted array to locate the key-value pair (see the arrow in the figure).

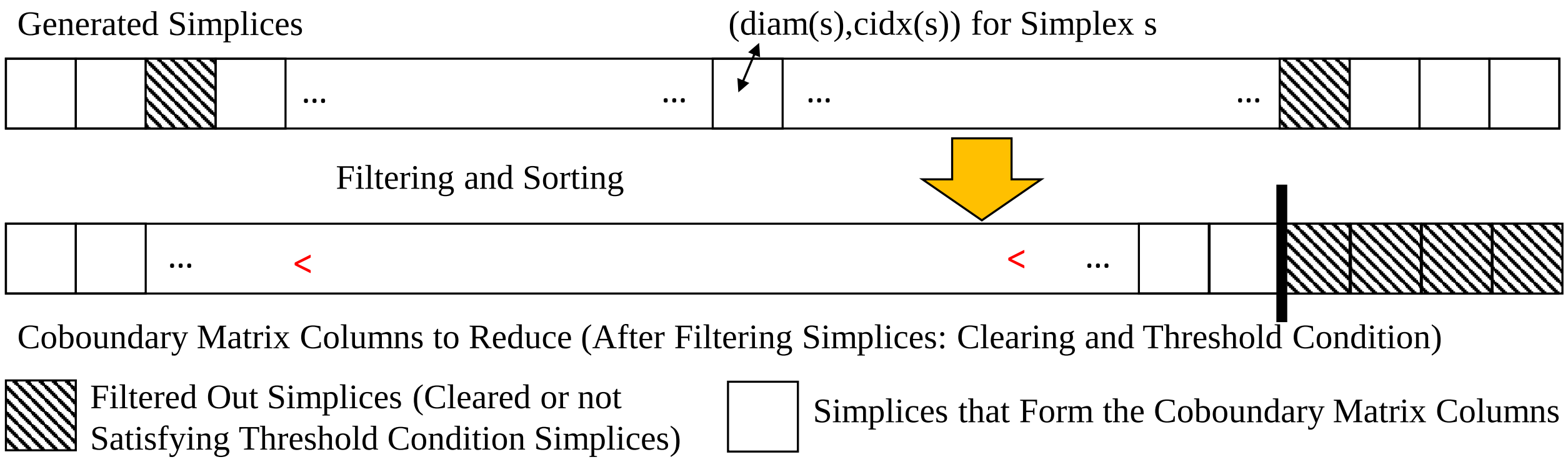

**Figure 5.11.** The Filtration Construction with Clearing Algorithm for Full Rips Filtrations

Sorting in the reverse of the order given in Section 5.1.4 is then conducted over the remaining simplices. This is the order for the columns of a coboundary matrix. The resulting sequence of simplices is then the columns to reduce for the following matrix reduction phase. Algorithm 18 presents how we construct the full Rips filtration with clearing. Our GPU-based algorithms leverage the massive parallelism of GPU threads and high bandwidth data processing in GPU device memory. For a sparse Rips filtration, our construction process uses the cofacet enumeration of Ripser per thread, as in Algorithm 15, and is similar to [105].

**Algorithm 18:** Use GPU for Full Rips Filtration Construction with Clearing

**Input:** $\boldsymbol{P}$, threshold, tid: as defined in Algorithm 17; $n_0$: the number of points; $d$: the current dimension for simplices to construct.

**Output:** $\boldsymbol{C}$: as defined in Algorithm 17.

$\boldsymbol{C} \leftarrow \varnothing$ ;

tid $\leftarrow \{0, \ldots, 0\}$ ;

`filter-columns-kernel`($\boldsymbol{C}$, $\boldsymbol{P}$, threshold, tid) /* $\binom{n_0}{d+1}$ threads launched */

len $\leftarrow$ `GPU-reduction(flagarray)` ;

`GPU-sort`($\boldsymbol{C}$) /* Sort entries of $\boldsymbol{C}$ in coboundary filtration order: decreasing diameters, increasing combinatorial indices; restrict $\boldsymbol{C}$ to indices 0 to len $-$ 1 afterwards. */

### 5.5.3 Warp-based Filtering

There is also a standard technique for filtering on GPU which is warp-based. A warp is a unit of 32 threads that work in SIMT (Single Instruction Multiple Threads) fashion, executing the same instruction on multiple data. This concept is very different from MIMD (Multiple Instruction Multiple Data) parallelism [1]. Warp-filtering does not change the complexity of filtering in $O(N)$, where $N$ is the number of elements to filter; however it can allow for insertion into an array using 32 threads (a unit of a warp) at a time in SIMT fashion. We use warp-based filtering for sparse computation and as an equivalent algorithm to Algorithm 18. Warp-based filtering involves each warp atomically grabbing an offset to

**Algorithm 17:** Filtering the Columns on GPU

**Input:** $\boldsymbol{P}$: the persistence pairs in the form (cofacet, simplex) discovered in the previous dimension; threshold: the max diameter allowed for a simplex; diam(·): the diameter of a simplex; cidx(·): the combinatorial index of a simplex.

**Output:** $\boldsymbol{C}$: an array of simplices, where each element includes a diameter paired with a combinatorial index; tid: an array of flags marking which columns are kept (filtered in).

```
Procedure filter-columns-kernel(C, P, threshold, tid):
    cidx(s) ← tid ;
    if ∄t such that (t, s) ∈ P AND diam(s) ≤ threshold then
        diam(C[tid]) ← diam(s) ;
        cidx(C[tid]) ← cidx(s) ;
        tid[tid] ← 1 ;
    else
        diam(C[tid]) ← −∞ ;
        cidx(C[tid]) ← +∞ ;
        tid[tid] ← 0 ;
    end
```

the output array and communicating within the warp to determine which thread will write what selected array element to the array beginning at the thread's offset within the warp.

### 5.5.4 Using Ripser++

The command line interface for Ripser++ is the same as Ripser to make the usage of Ripser++ as easy as possible to TDA specialists. However, Ripser++ has a `-sparse` option which manually turns on, unlike in Ripser, the sparse computation algorithm for Vietoris-Rips barcode computation involving a sparse number of neighboring relations between points. Python bindings for Ripser++ will be available to allow users to write their own preprocessing code on distance matrices in Python as well as to aid in automating the calling of Ripser++ by removing input file I/O.

## 5.6 Experiments

All experiments are performed on a powerful computing server. It consists of an NVIDIA Tesla V100 GPU that has 5120 FP32 cores and 2560 FP64 cores for single- and double-precision floating-point computation. The GPU device memory is 32 GB High Bandwidth Memory 2 (HBM2) that can provide up to 900 GB/s memory access bandwidth. The server also has two 14 core Intel XEON E5-2680 v4 CPUs (28 cores in total) running at 2.4 GHz with a total of 100 GB of DRAM. The datasets are taken from the original Ripser repository on Github [62] and the repository of benchmark datasets from [57].

### 5.6.1 The Empirical Relationship amongst Apparent Pairs, Emergent Pairs, and Shortcut Pairs

There exists three kinds of persistence pairs of the Vietoris-Rips filtration, in fact for any filtration with a simplex-wise refinement. Using the terminology of [103], these are apparent (Definition 5.3.4) [63, 71, 103, 112], shortcut [103], and emergent pairs [1, 103]. By definition, they are known to form a tower of sets ordered by inclusion (expressed by Equation (5.66)). We will show a further empirical relationship amongst these pairs involving their cardinalities.

$$\overbrace{\underbrace{\text{apparent pairs}}_{\text{large cardinality}} \subseteq \text{shortcut pairs} \subseteq \text{emergent pairs} \subseteq \text{persistence pairs}}^{\text{the difference in cardinalities is "small"}} \tag{5.66}$$

The first empirical property is that the cardinality difference amongst all of the sets of pairs is very small compared to the number of pairs, assuming Ripser's framework of computing cohomology and using the simplex-wise filtration ordering in Section 5.1.4. Thus there exist a very large number of apparent pairs. The second is that the proportion of apparent pairs to columns in the cleared coboundary matrix increases with dimension (see Section 5.7.1 in Appendix), assuming no diameter threshold criterion as in the first property.

Table 5.1 shows the percentage of apparent pairs up to dimension $d$ is extremely high, around 99%. Since the number of columns of a cleared coboundary matrix equals to the number of persistence pairs, the number of nonapparent columns for submatrix reduction is a tiny fraction of the original number of columns in Ripser's matrix reduction phase.

**Table 5.1.** Empirical Results on Apparent, Shortcut, Emergent Pairs

| Datasets | $n$ | $d$ | apparent pairs | shortcut pairs | emergent pairs | all pairs | percentage of apparent pairs |
|---|---|---|---|---|---|---|---|
| *celegans* | 297 | 3 | 317,664,839 | 317,723,916 | 317,723,974 | 317,735,650 | 99.9777139% |
| *dragon1000* | 1000 | 2 | 166,132,946 | 166,160,587 | 166,160,665 | 166,167,000 | 99.9795062% |
| *HIV* | 1088 | 2 | 214,000,996 | 214,030,431 | 214,040,521 | 214,060,736 | 99.9720920% |
| *o3* (sparse: $t = 1.4$) | 4096 | 3 | 43,480,968 | 43,940,030 | 43,940,686 | 44,081,360 | 98.6379912% |
| *sphere_3_192* | 192 | 3 | 54,779,316 | 54,871,199 | 54,871,214 | 54,888,625 | 99.8008531% |
| *Vicsek300_of_300* | 300 | 3 | 330,724,672 | 330,818,491 | 330,818,507 | 330,835,726 | 99.9664323% |

### 5.6.2 Execution Time and Memory Usage

We perform extensive experiments that demonstrate the execution time and memory usage of Ripser++. We further look into the performance of both the apparent pairs search algorithm and the management of persistence pairs in the two layer data structure after finding apparent pairs. Variables $n$ and $d$ for each dataset are the same for all experiments.

Table 5.2 shows the comparisons of execution time and memory usage for computation up to dimension $d$ between Ripser++ and Ripser with six datasets, where R. stands for Ripser and R.++ stands for Ripser++. Memory usage on CPU and total execution time were measured with the `/usr/time -v` command on Linux. GPU memory usage was counted by the total displacement of free memory over program execution.

**Table 5.2.** Total Execution Time and CPU/GPU Memory Usage

| Datasets | n | d | R.++ time | R. time | R.++ GPU mem. | R.++ CPU mem. | R. CPU mem. | Speedup |
|---|---|---|---|---|---|---|---|---|
| *celegans* | 297 | 3 | 7.30 s | 228.56 s | 16.84 GB | 10.53 GB | 23.84 GB | 31.33x |
| *dragon1000* | 1000 | 2 | 5.79 s | 48.98 s | 8.81 GB | 3.75 GB | 5.79 GB | 8.46x |
| *HIV* | 1088 | 2 | 7.11 s | 147.18 s | 11.36 GB | 6.68 GB | 14.59 GB | 20.69x |
| *o3* (sparse: $t = 1.4$) | 4096 | 3 | 11.62 s | 64.18 s | 18.76 GB | 2.77 GB | 3.86 GB | 5.52x |
| *sphere_3_192* | 192 | 3 | 2.43 s | 36.96 s | 2.92 GB | 2.03 GB | 4.32 GB | 15.21x |
| *Vicsek300_of_300* | 300 | 3 | 9.98 s | 248.72 s | 17.53 GB | 11.46 GB | 27.78 GB | 24.92x |

Table 5.2 shows Ripser++ can achieve 5.52x - 31.33x speedups of total execution time over Ripser in the evaluated datasets. The performance improvement mainly comes from massive parallel operations of finding apparent pairs on GPU, and from the fast filtration construction with clearing by GPU using filtering and sorting. We also notice that the speedups of execution time varies in different datasets. That is because the percentages of execution time in the submatrix reduction are different among datasets.

It is well known that the memory usage of full Vietoris-Rips filtration grows exponentially in the number of simplices with respect to the dimension of persistence computation. For example, 2000 points at dimension 4 computation may require $\binom{2000}{4+1} \times 8$ bytes = 2 million GB memory. Algorithmically, we avoid allocating memory in the cofacet dimension and keep the memory requirement of Ripser++ asymptotically same as Ripser. Table 5.2 also shows the memory usage of Ripser++ on CPU and GPU. Ripser++ can actually lower the memory usage on CPU. This is mostly because Ripser++ offloads the process of finding apparent pairs to GPU and the following matrix reduction only works on much fewer columns than that of Ripser (as the submatrix reduction). Table 5.2 also shows that the GPU device memory usage is usually lower than the total memory usage of Ripser. However, in the sparse computation case (dataset *o3*) the algorithm must change; Ripser++ thus allocates memory depending on the hardware instead of the input sizes.

### 5.6.3 Throughput of Apparent Pairs Discovery with Ripser++ vs. Throughput of Shortcut Pairs Discovery in Ripser

Discovering shortcut pairs in Ripser and discovering apparent pairs in Ripser++ account for a significant part of the computation. Thus, we compare the discovery throughput of these two types of pairs in Ripser and Ripser++, respectively. The throughput is calculated as the number of a specific type of pair divided by the time to find and store them. The results are reported in Figure 5.12. We can find for all datasets, our GPU-based solution outperforms the CPU-based algorithm used in Ripser by 4.2x-12.3x. Since the number of the two types of pairs are almost the same (see Table 5.1), such throughput improvement can lead to a significant saving in computation time.

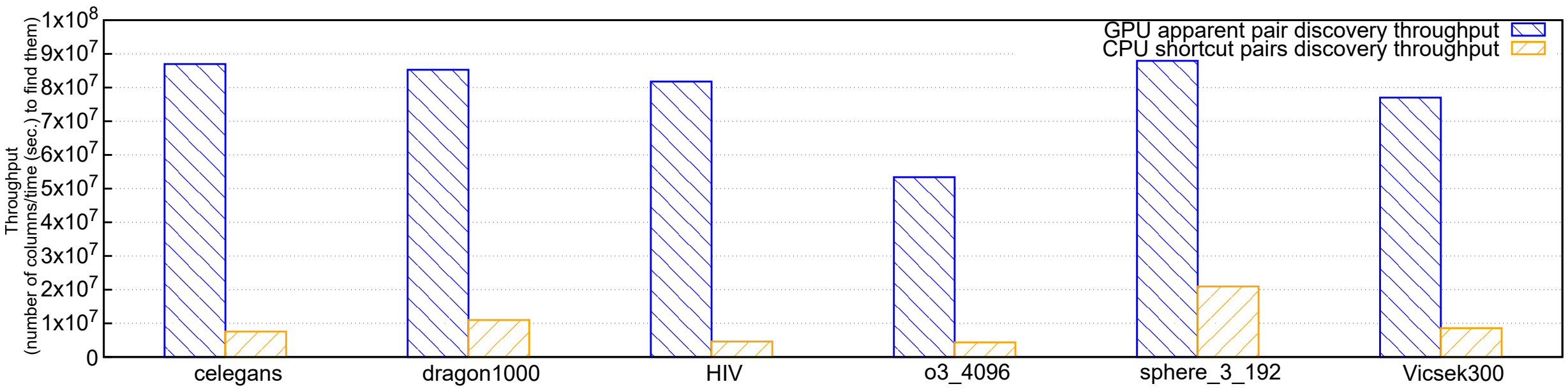

**Figure 5.12.** A comparison of column discovery throughput of apparent pair discovery with Ripser++ vs. Ripser's shortcut pair discovery. The time is greatly reduced due to the parallel algorithm of finding apparent pairs on GPU (see Algorithm 13).

**Table 5.3.** Hashmap Access Throughput, Counts, and Times Comparisons

| Datasets | R.++ write throuput (pairs/s) | R. write throughput (pairs/s) | Num. of R.++ reads to data struct. | Num. of R. reads to hashmap | R.++ read time (s) | R. read time (s) |
|---|---|---|---|---|---|---|
| *celegans* | $7.21 \times 10^8$ | $6.98 \times 10^7$ | $3.22 \times 10^4$ | $5.81 \times 10^8$ | 0.00100 | 11.43 |
| *dragon1000* | $7.62 \times 10^8$ | $6.29 \times 10^7$ | $1.19 \times 10^5$ | $1.12 \times 10^8$ | 0.00460 | 1.28 |
| *HIV* | $7.06 \times 10^8$ | $8.85 \times 10^7$ | $1.57 \times 10^5$ | $3.10 \times 10^8$ | 0.00130 | 5.52 |
| *o3* (sparse: $t = 1.4$) | $4.78 \times 10^8$ | $6.88 \times 10^7$ | $1.65 \times 10^6$ | $8.85 \times 10^7$ | 0.01500 | 0.56 |
| *sphere_3_192* | $7.32 \times 10^8$ | $9.41 \times 10^7$ | $2.71 \times 10^5$ | $9.37 \times 10^7$ | 0.00068 | 0.30 |
| *Vicsek300_of_300* | $6.80 \times 10^8$ | $8.82 \times 10^7$ | $2.12 \times 10^5$ | $5.67 \times 10^8$ | 0.00053 | 10.81 |

### 5.6.4 Two-layer Data Structure for Memory Access Optimizations

Table 5.3 presents the write throughput of persistence pairs in pairs/s in the 2nd and 3rd columns. In Ripser, we use the measured time of writing pairs to the hashmap to divide the total persistence pair number; while in Ripser++, the time includes writing to the two-layer data structure and sorting the array on GPU. The results show that Ripser++ consistently has one order of magnitude higher write throughput than that of Ripser.

Table 5.3 also gives the number of reads in the 4th, 5th, and 6th columns as well as the time consumed in the read operations (in seconds) in the last column. The number of reads in Ripser means the number of reads to its hashmap, while Ripser++ counts the number of reads to the data structure. The reported results confirm that Ripser++ can reduce at least two orders of magnitude memory reads over Ripser. A similar performance improvement can also be observed in the measured read time.

### 5.6.5 Breakdown of Accelerated Components

We breakdown the speedup on the two accelerated components of Vietoris-Rips persistence barcode computation over all dimensions ≥ 1: matrix reduction vs. filtration construction with clearing. Ripser++ accelerates both stages of computation; however, which stage is accelerated more varies. For most datasets, it appears the filtration construction with clearing stage is accelerated more than the matrix reduction stage. This is because this stage is massively parallelized in its entirety while accelerated matrix reduction only parallelizes the finding and aggregation/management of apparent pairs. Speedups on filtration construction with clearing range from 2.87x to 42.67x while speedups on matrix reduction range from 5.90x to 36.71x.

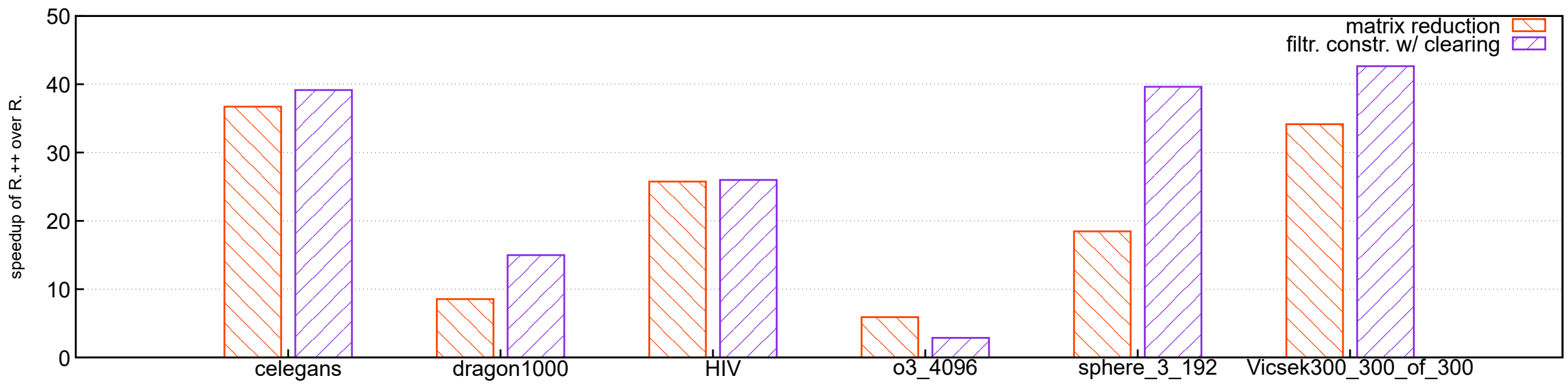

**Figure 5.13.** A breakdown of the speedup of Ripser++ over Ripser for computation beyond dimension 0 into the two stages: matrix reduction and filtration construction with clearing.

### 5.6.6 Experiments on the Apparent Pairs Rate in Dimension 1

We run extensive experiments to analyze the number of apparent pairs in the average case of a random distance matrix input. We make an assumption and one observation about the values of a random distance matrix.

**Assumption 5.6.7.** *In practice, distances between points are almost never exactly equal. Thus we assume the entries of the distance matrix are all different.*

**Observation 5.6.8.** *The persistence barcodes (see Section 5.2 on definition of barcodes) executed by the persistent homology algorithm do not change up to endpoint reassignment if we reassign the distances of the input distance matrix while preserving the total order amongst all distances.*

Thus the setup for our experiments is to uniformly at random sample permutations of the integers 1 through $n(n-1)/2$ to fill the lower triangular portion of a distance matrix, where $n$ is the number of points. We run Ripser++ for $n$ = 50, 100, 200, 400, ..., 1000, ..., 9000 with 10 uniformly random samples with replacement of $n(n-1)/2$-permutations for a fixed random seed for dimension 1 persistence. We consider the general combinatorial case where the distance matrix does not necessarily satisfy the triangle inequality and thus that the set of points may not form a finite metric space. This is still valid input, as Vietoris-Rips barcode computation is dependent only on the edge relations between points (e.g. the 1-skeleton). Figure 5.14 shows the plot for the percentage of apparent pairs with respect to the

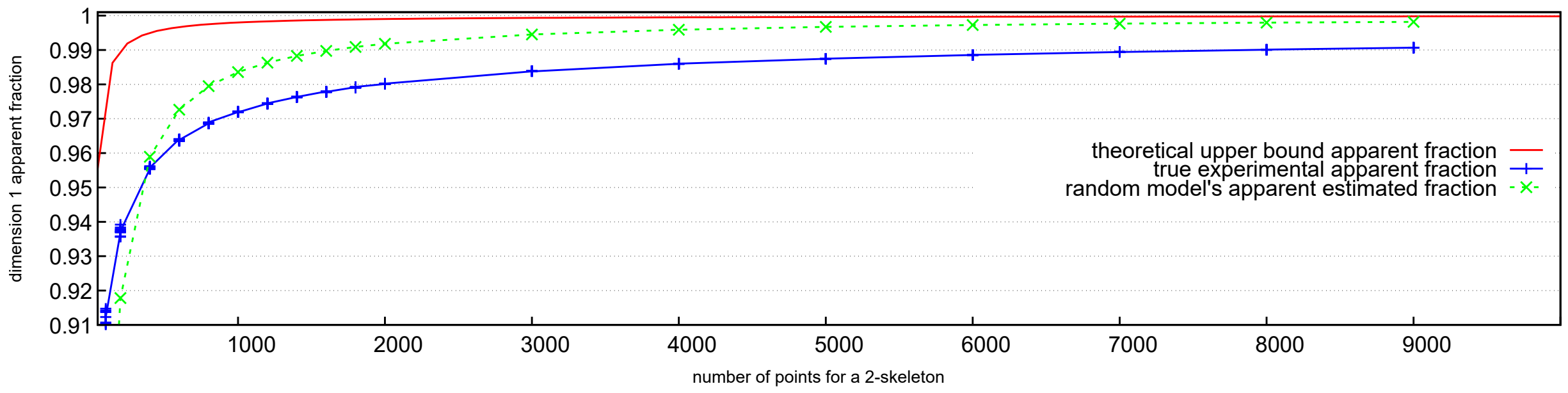

**Figure 5.14.** Three different curves of the apparent fraction for a 1-dimensional coboundary matrix as a function of the number of points. The theoretical upper bound is for the case of all diameters the same, but also can be achieved when all diameters are different. Interestingly, the variance of the apparent fraction for each point count from the experiments is very low even though the lower distance matrix entries are uniformly at random permuted. The dotted curve is the piecewise linear interpolated curve of the random model that matches the shape of the empirical and theoretical curve. For more details about the random mathematical model, see the Section 5.6.11

total number of 1-simplices for a 1-dimensional coboundary matrix (the apparent fraction)

as a function of the number of points of a full 2-skeleton for three different contexts. The first is the theoretical upper bound of $(n-2)/n$ proven in Theorem 5.4.2, the second is the actual percentage found by our experiments, and the third is the percentage predicted by our random model for the case of dimension 1 coboundary matrices (see Section 5.6.11).

From the experiments, we notice that in the average case there is still a large number of apparent pairs and this number is close to and closely guided by the theoretical upper bound found in Theorem 5.4.2. Furthermore, the model's curve and the theoretical upper bound are asymptotic to 1.00 as $n \to \infty$. This is calculated by some algebraic manipulations of radical equations. Furthermore, we performed the true apparent pairs fraction search experiment up to 20000 points, with consistent monotonic behavior toward 1.00. For example, at 10000, 20000 points we obtain 0.991127743, and 0.993733522 average apparent fractions respectively.

### 5.6.9 Algorithm for Randomly Assigning Apparent Pairs on a Full 2-Skeleton

**Algorithm 19:** Algorithm for Random Apparent Pairs Construction

**Input:** $d_\mathrm{i}$: a sequence of diameters with $d_1 > d_2 > \ldots > d_{\binom{n}{2}}$; $K = (V, E, T)$: a full 2-skeleton on $n$ points where $V$ is a set of $n$ vertices, $E$ is a set of $\binom{n}{2}$ edges, and $T$ is a set of $\binom{n}{3}$ triangles.

**Output:** A sequence of apparent pairs of edges and triangles $(\mathrm{e_i}, t_\mathrm{i})$ with $\mathrm{diam}(\mathrm{e_i}) = \mathrm{diam}(t_\mathrm{i}) = d_\mathrm{i}$.

1 **Function** `RandomDiameterAssignment(`$K$`):`

2     **while** *there are triangles left in* $T$ **do**

3         Uniformly at random pick a 1-dimensional simplex $\mathrm{e_i} \in E$ ;

4         Assign edge $\mathrm{e_i}$ a diameter $d_\mathrm{i}$ strictly less than all $d_k$ for $k < \mathrm{i}$ `/* (e.g.,` $d_\mathrm{i} = \binom{n}{2} - \mathrm{i} + 1$`) */`

5         **if** *there are triangles incident to* $\mathrm{e_i}$ **then**

6             Pair up $\mathrm{e_i}$ with its oldest cofacet, the unique triangle $t_\mathrm{i} \in T$ of highest lexicographic order amongst remaining triangles incident to $\mathrm{e_i}$ ;

7             Emit $(\mathrm{e_i}, t_\mathrm{i})$ ;

8             Remove $\mathrm{e_i}$ from $E$ and all triangles $t'_\mathrm{i}$ containing $\mathrm{e_i}$ in their boundary from $T$ since these triangles must all have the same diameter $d_\mathrm{i}$ ;

9         **end**

10     **end**

Algorithm 19 assigns diameters to a subset of the edges in decreasing order so that each diameter value $d_\mathrm{i}$ at iteration i results (if possible) in an apparent pair $(\mathrm{e_i}, t_\mathrm{i})$ with edge $\mathrm{e_i}$ and triangle $t_\mathrm{i}$ both of diameter $d_\mathrm{i}$. In the Algorithm 19 at line 5, since we assume at iteration $\mathrm{i}, \mathrm{i} > \mathrm{i}'$ that $d_\mathrm{i} < d_{\mathrm{i}'}$ and that all triangles of diameter greater than or equal to $d_{\mathrm{i}'}$ have already been removed, at iteration i all remaining cofacets $t'$ of $\mathrm{e_i}$ must have the same diameter as $\mathrm{e_i}$. Recalling Assumption 5.6.7 and Observation 5.6.8, this random algorithm is equivalent to uniformly at random assigning permutations of the numbers $1, \ldots, \binom{n}{2}$ to the lower triangular part of a symmetric distance matrix $D$ and counting the number of apparent pairs in the 1-dimensional coboundary matrix induced by $D$.

### 5.6.10 A Greedy Deterministic Distance Assignment

We consider Algorithm 19 with line 3 changed to pick $e_i \in E$ with maximum number of cofacets remaining, with largest combinatorial index if there is a tie. This greedy deterministic algorithm serves as a lower bound to Algorithm 19. We plot the resulting apparent fraction and experimentally verify that $1/(d+2)$ is a theoretical lower bound for $d = 1$; (notice the theoretical lower bound did not depend on the diameter condition on simplices containing the maximum indexed point of Theorem 5.4.2). It is currently unknown what the theoretical relationship is between the greedy deterministic algorithm and any theoretical lower bounding curve.

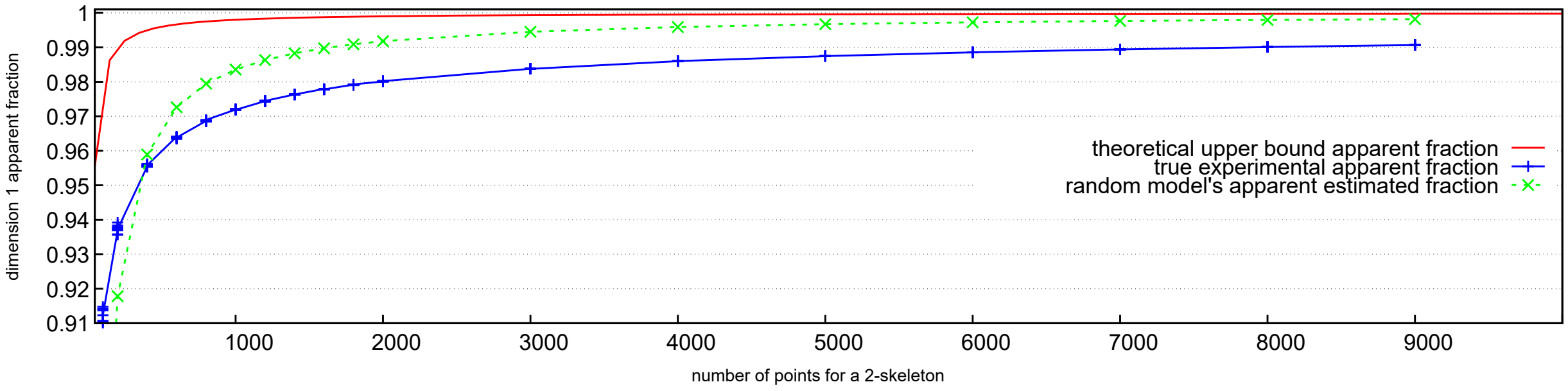

**Figure 5.15.** The deterministic greedy apparent fraction curve. Notice the theoretical lower bound of 0.3333 is confirmed experimentally by the experimental curve. The experiments show the apparent fraction stays within a neighborhood of 0.5 as n gets large enough.

### 5.6.11 A Random Approximation Model for the Number of Apparent Pairs in a Full 2-Skeleton on n Points

We construct a model for the analysis of random Algorithm 19. We model Algorithm 19 by analyzing a modified algorithm. Let there be a full 2-skeleton $K = (V, E, T)$ as in Algorithm 19. Let $E' \subseteq E$ be the subset of edges not including the single point $v \in V$ of highest index: $n-1$ and $T' \subseteq T$ be the subset of triangles induced by $E'$. Modify Algorithm 19 to let $\mathrm{j} \leq \binom{n-1}{2}$ be the fixed number of iterations of the loop, replacing line 8. Modify Algorithm 19 at line 3 to choose uniformly at random from $E'$ instead of $E$, outputting a sequence $C$ with j different edges and having diameters $d_1 > d_2 > \ldots > d_{\mathrm{j}}$.

We pick edges from $E'$ since this ensures that the If in line 4 of Algorithm 19 will always evaluate to true by the existence of triangles containing vertex $n-1$ and thus that there are at least $\|C\| = \mathrm{j}$ number of apparent pairs in $K$. $\|C\| = \mathrm{j}$ is equal to the number of iterations of the algorithm. After choosing j edges, we count how many triangles are still left in $T'$ in expectation.

We define a Bernoulli random variable for each triangle $t \in T'$ of the full 2-skeleton $K$.

$$X_{t,\mathrm{j}} = \begin{cases} 1 & \text{if triangle t} \in T' \text{ is not incident to any edges in } C \\ 0 & \text{otherwise} \end{cases}$$

We notice that for every triangle, the same random variable can be defined on it, all identically distributed.

Let

$$p_{t,\mathrm{j}} = \frac{(\binom{n-1}{2} - 3) \cdot (\binom{n-1}{2} - 4) \cdots (\binom{n-1}{2} - 3 - \mathrm{j} + 1)}{(\binom{n-1}{2} \cdot (\binom{n-1}{2} - 1) \cdots (\binom{n-1}{2} - \mathrm{j} + 1))}$$

be the probability of triangle $t \in T$ not containing any of the j chosen edges in its boundary of 3 edges.

We thus define the random variable $T_{\mathrm{j}} = \Sigma_{t \in T'} X_{t,\mathrm{j}}$ to count the number of triangles remaining after j edges are chosen in sequence.

Taking expectation, we get

$$E[T_{\mathrm{j}}] = \Sigma_{t \in T'} E[X_t] = \Sigma_{t \in T'} 1 \cdot p_{t,\mathrm{j}} = \binom{n-1}{3} \cdot \frac{(\binom{n-1}{2} - 3) \cdot (\binom{n-1}{2} - 4) \cdots (\binom{n-1}{2} - 3 - \mathrm{j} + 1)}{(\binom{n-1}{2} \cdot (\binom{n-1}{2} - 1) \cdots (\binom{n-1}{2} - \mathrm{j} + 1))}$$

by linearity of expectation, the definition of $T'$ and the definition of $p_{t,\mathrm{j}}$.

Set $E[T_{\mathrm{j}}] = \tau$, with $\tau$ the number of triangles reserved to not be incident to the sequence $C$ of j apparent edges of $K$. Then solve for j from the equation $E[T_{\mathrm{j}}] = \tau$ with a numerical equation solver system; then divide j by $\binom{n}{2}$, the total number of edges, and call this the ratio $r_\tau(n)$. Since we are just building a mathematical model to match experiment, we fit our curve $r_\tau(n)$ to the true experimental curve from Figure 5.14. We choose $\tau = 500$ to minimize the least squares error on the sampled values of n in Figure 5.14 (we assume $n, \tau$ s.t. $\binom{n}{3} > \tau$). to the averaged experimental curve, formed by averaging the apparent fraction for each $n$. $\tau$ was chosen in units of 100s due to the high computational cost of solving for j for every $n$. We then obtain the dotted curve in Figure 5.14, $r_{500}(n)$: a function of $n$, the number of points.

The shape of the model's curve, which matches experiment and stays within theoretical bounds is the primary goal of our model. The constant, $\tau$=500, suggests that as the number of points increases, in practice the expected percentage of triangles not a cofacet of an apparent edge decreases to 0 and that the the expected value is approximately a constant value. See Section 5.11 for an equivalent model that counts edges and triangles differently.

## 5.7 The "Width" and "Depth" of Computing Vietoris-Rips Barcodes

We are motivated by a common phenomenon in computation of Vietoris-Rips barcodes found in [1] for the matrix reduction stage: the required sequential computation is concentrated on a very few number of columns in comparison with the filtration size. We further generalize a principle to quantify parallelism in the computation as a guidance for parallel processing.

We define two concepts: "computational width" as a measurement of the amount of independent operations and "computational depth" as a measurement of the amount of dependent operations. Their quotients measure the level of parallelism in computation. We consider rough upper and lower bounds on parallelism for Vietoris-Rips barcode computation using these quotients.

For an upper bound, let the "computational width" be 2 × the number of simplices in the filtration or 2 × the total number of uncleared columns of the coboundary matrix where the 2 comes from the two stages of persistence computation: filtration construction and matrix reduction. (This quantifies the maximum amount of independent columns achievable if all columns were independent of each other). This rationale comes from the existence of a large percentage of columns that are truly independent of each other (e.g. apparent columns during matrix reduction) as well as the independence amongst simplices for their construction and storage during filtration construction (assume the full Rips computation case). Let the "computational depth" be 1 + the amount of column additions amongst columns requiring at least one column addition. (The + 1 is to prevent dividing by zero). In this case, the "computational width" divided by the "computational depth" thus quantifies an upper bound on the amount of parallelism available for computation.

For a lower bound one could similarly consider the "computational width" as the number of apparent columns divided by the "computational depth" as the sum of all column additions amongst columns plus any dependencies during filtration construction.

These bounds along with empirical results on columns additions [1], the percentage of apparent pairs in Table 5.1, and the potentially several orders of magnitude factor difference in number of simplices compared to column additions, suggest that there can potentially be a high level of hidden parallelism suitable for GPU in computing Vietoris-Rips barcodes. We are thus led to aim for two objectives for effective performance optimization:

- 1. To massively parallelize in the "computational width" direction (e.g. parallelize independent simplex-wise operations).
- 2. To exploit locality in the "computational depth" direction (while using sparse representations).

Objective 1 is well achievable on GPU while Objective 2 is known to be best done on CPU. In fact depth execution such as column additions are best done sequentially due to the few number of "tail columns" [1] of the coboundary matrix of Vietoris-Rips filtrations.

Our "width" and "depth" principle gives bounds for developing potential parallel algorithms to accelerate, for example, Vietoris-Rips barcode computation. However, real-world performance improvements must be measured empirically, as in Table 5.2.

### 5.7.1 The Growth of the Proportion of Apparent Pairs

In our experiments on smaller datasets with high dimension and a low number of points, all persistence pairs eventually become trivial at a relatively low dimension compared to the number of points. Theoretically, of course, there always exists some dimension at which the filtration collapses in persistence (all pairs become 0-persistence), e.g. $n$ points with an $n$-1 dimensional simplex. Apparent pairs are a subset of the 0-persistence pairs, we show empirically in Table 5.4 that the proportion of apparent pairs of dimension $d$ to the number of pairs (including infinite persistence "pairs") in dimension $d$ grows with dimension. The only exception to this is the o3 dataset with sparse computation and a restrictive threshold. We believe this outlier is due to the threshold condition. The other datasets were computed without any threshold condition.

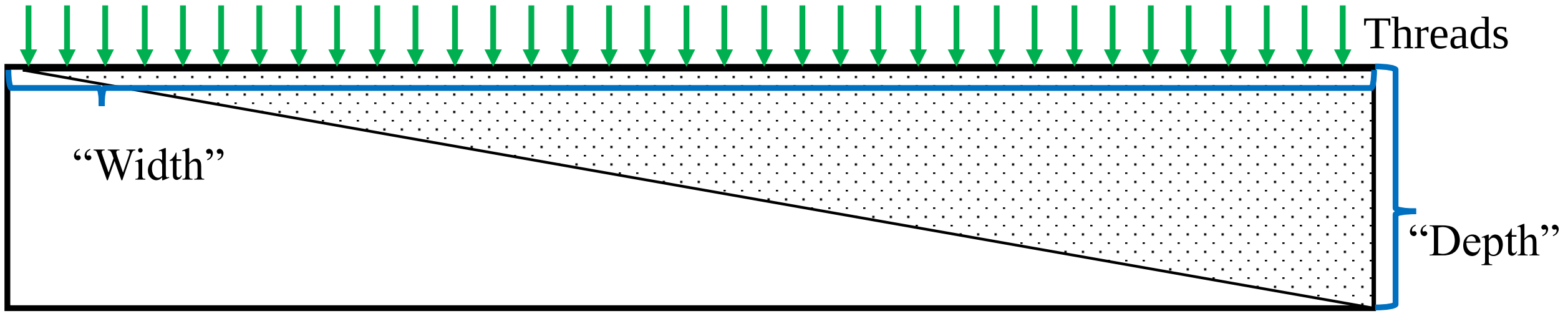

**Figure 5.16.** Illustration of the "width" and "depth" of computation. The area of the triangle represents the total amount of work performed. Furthermore, the "width" divided by the "depth" quantifies the level of parallelism in computation.

**Table 5.4.** Apparent Pair Percentage (Out of All Persistence Pairs) per Dimension

| Datasets | n | d | % apparent in dim 1 | % apparent in dim 2 | % apparent in dim 3 |
|---|---|---|---|---|---|
| *celegans* | 297 | 3 | 99.661017% | 99.962021% | 99.977972% |
| *dragon1000* | 1000 | 2 | 99.937011% | 99.972234% | |
| *HIV* | 1088 | 2 | 99.92071% | 99.972234% | |
| *o3* (sparse: $t = 1.4$) | 4096 | 3 | 98.928064% | 98.651461% | 98.634918% |
| *sphere_3_192* | 192 | 3 | 99.707909% | 99.734764% | 99.802291% |
| *Vicsek300_of_300* | 300 | 3 | 99.849611% | 99.925678% | 99.966999% |

### 5.7.2 Empirical Properties of Filtering by Diameter Threshold and Clearing

We have done an empirical study on the filtration construction with clearing stage of computation. By Table 5.5, except for dragon1000, all full Rips filtrations, after applying the enclosing radius condition (see Section 5.2.12) and clearing (see Section 5.2.3), result in a large percentage of simplices selected for reduction. More importantly, Algorithm 18 sorts all the possible number of simplices for full Rips computation. If the number of simplices selected is close to the number of simplices sorted, the sorting is effective. By our experiments, even in the dragon1000 case, sorting all simplices is still faster than CPU filtering or even warp-based filtering on GPU (see Section 5.5.3).

Table 5.5. Empirical results on Clearing and Threshold Restriction

| Datasets | n | d | possible num. simpl. | num. cols. to reduce | simpl. removed by diameter | cols. cleared | % of simplices sel. for red. |
|---|---|---|---|---|---|---|---|
| *celegans* | 297 | 3 | 322,058,286 | 256,704,712 | 61,576,563 | 3,777,011 | 79.7075322% |
| *dragon1000* | 1000 | 2 | 166,666,500 | 56,110,140 | 110,237,919 | 318,441 | 33.6661177% |
| *HIV* | 1088 | 2 | 214,652,064 | 155,009,693 | 59,123,662 | 518,709 | 72.2143967% |
| *o3* (sparse: $t = 1.4$) | 4096 | 3 | $1.17 \times 10^{13}$ | 44,081,360 | $1.17 \times 10^{13}$ | 4,347,112 | 0.00037604% |
| *sphere_3_192* | 192 | 3 | 55,004,996 | 46,817,416 | 8,159,941 | 1,072,739 | 85.114843% |
| *Vicsek300_of_300* | 300 | 3 | 335,291,125 | 283,441,085 | 47,803,132 | 4,046,908 | 84.5358150% |

By Section 5.2.12, the enclosing radius eliminates simplices when added to a growing simplicial complex, will only contribute 0-persistence pairs. This is equivalent to truncating the coboundary matrix to its bottom right block (or zeroing such rows and columns). There are usually fewer columns zeroed by the clearing lemma than zeroed by the threshold condition, however they correspond to columns that must be completely zeroed during reduction and a lot will form "tail columns" that dominate the time of matrix reduction.

Notice that the o3_4096 dataset has a predefined threshold less than the enclosing radius. The number of possible simplices is several orders of magnitude larger than the actually number of columns needed to reduce after clearing. Thus we must use the sparse option of computation and avoid using Algorithm 18. This means memory is not allocated as a function of the number of possible simplices and is instead allocated with respect to the GPU's memory capacity as we grow the number of columns for matrix reduction. Sorting is not used and a warp-base filtering is used instead (see Section 5.5.3).

## 5.8 Related Persistent Homology Computation Software

In Section 5.2, we briefly introduce several software to compute persistent homology, paying special attention on computing Vietoris-Rips barcodes. In this section, we elaborate more on this topic.

The basic algorithm upon which all such software are based on is given in Algorithm 11. Many optimizations are used by these software, and have made significant progress over the basic computation of Algorithm 11. Amongst all such software, Ripser is known to achieve state of the art time and memory performance in computing Vietoris-Rips barcodes [57, 103]. Thus it should suffice to compare our time and memory performance against Ripser alone. We overview a few, and certainly not all, of the related software besides Ripser.

Gudhi [91] is a software that computes persistent homology for many filtration types. It uses a general data structure called a simplex tree [118, 119] for general simplicial complexes for storage of simplices and related operations on simplices as well as the compressed annotation matrix algorithm [92] for computing persistent cohomology. It can compute Vietoris-Rips barcodes.

Eirene [107] is another software for computing persistent homology. It can also compute Vietoris-Rips barcodes. One of its important features is that it is able to compute cycle representatives. Paper [89] details how Eirene can be optimized with GPU.

Hypha (a hybrid persistent homology matrix reduction accelerator) [1] is a recent open source software for the matrix reduction part of computing persistent homology for explicitly represented boundary matrices, similar in style to [59, 106]. Hypha is one of the first publicly available softwares using GPU. A framework based on the separation of parallelisms is designed and implemented in Hypha due to the existence of atleast two very different execution patterns during matrix reduction. Hypha also finds apparent pairs on GPU and subsequently forms a submatrix on multi-core similar to in Ripser++.

## 5.9 Conclusion

Ripser++ can achieve significant speedup (up to 20x-30x) on representative datasets in our work and thus opens up unprecedented opportunities in many application areas. For example, fast streaming applications [120] or point clouds from neuroscience [121] that spent minutes can now be computed in seconds, significatly advancing the domain fields.

We identify specific properties of Vietoris-Rips filtrations such as the simplicity of diameter computations by individual threads on GPU for Ripser++. Related discussions, both theoretical and empirical, suggest that our approach be applicable to other filtration types such as cubical [106], flag [98], and alpha shapes [91]. We strongly believe that our acceleration methods are widely applicable beyond computing Vietoris-Rips persistence barcodes.

We have described the mathematical, algorithmic, and experimental-based foundations of Ripser++, a GPU-accelerated software for computing Vietoris-Rips persistence barcodes. Computationally, we develop massively parallel algorithms directly tied to the GPU hardware, breaking several sequential computation bottlenecks. These bottlenecks include the Filtration Construction with Clearing stage, Finding Apparent Pairs, and the efficient management of persistence pairs for submatrix reduction. Theoretically we have looked into properties of apparent pairs, including the Apparent Pairs Lemma for massively parallel computation and a Theorem of upper and lower bounds on their significantly large count. Empirically we have performed extensive experiments, showing the true consistent behavior of apparent pairs on both random distance matrices as well as real-world datasets, closely matching the theoretical upper bound we have shown. Furthermore, we have measured the time, memory allocation, and memory access performance of our software against the original Ripser software. We achieve up to 2.0x CPU memory efficiency, besides also significantly reducing execution time. We hope to lead a new direction in the field of topological data analysis, accelerating computation in the post Moore's law era and turning theoretical and algorithmic opportunities into a high performance computing reality.

## 5.10 Some Proofs

### 5.10.1 Oblivious Column Reduction Proof

Recall the following notation: let $D$ and $R$ be as in Algorithm 11, let $D_{\mathrm{j}}$ denote the $\mathrm{j}th$ column of $D$, $R_{\mathrm{j}}$ denote a fully reduced column, i.e. the $\mathrm{j}th$ column of $R$ after Algorithm 11 terminates, and let $R[\mathrm{j}]$ denote the $\mathrm{j}th$ column of $R$ during Algorithm 12, partially reduced.

*Proof.* **base case**: The first nonzero column $j_0$ requires no column additions. $R_{j_0} = D_{j_0}$ is equivalent to a fully reduced column by standard algorithm.

**induction hypothesis**: We have reduced all columns from 0 to $j \geq j_0$ by the oblivious matrix reduction algorithm. Each such column $R[j']$, $j \geq j' \geq j_0$ (initially $D_{j'}$) was reduced by a sum of a sequence of $D_{i'}$, $i' < j'$, equivalent to a sum of a sequence of fully reduced $R_i, i < j'$ from the standard algorithm.

**induction step** : Let the *next* column $k$, to the right of column j, be a nonzero partially reduced column that needs column additions and call it $R[k]$ (initially $D_k$). Let $j = \text{lookup}[low(R[k])]$, the column index with matching lowest **1** with column $k$, that must add with column $k$.

If $R[j] = D_j$, then since column $k$ adds with $D_j$, certainly $R[k] \leftarrow R[k] + (R_j = R[j] = D_j)$

Otherwise if $R[j] \neq D_j$, add column $D_j$ with $R[k]$ and call this new column $R[k]'$ and notice that all nonzeros from $low(R[k]) + 1$ down to $low(R[k]')$ (viewing the column from top to bottom) of the working column $R[k]'$ are now exactly equivalent to the nonzeros from index $low(R[k]) + 1{=}low(R_j) + 1$ down to $low(R[k]') = low(D_j)$ of column $D_j$. This is because $low(R[k])$ is the lowest **1** so all entries below it are zero, so we can recover a block of nonzeros equivalent to a bottom portion of column $D_j$ upon adding $D_j$ to $R[k]$.

We have recovered the exact same nonzeros of column $D_j$ from $low(R_j)+1 = low(R[k])+1$ down to $low(D_j)$. Thus by the oblivious algorithm, before $low(R[k]')$ rises above $low(R[k])$ during column reduction, the sequence of columns to add to $R[k]'$ is equivalent to the sequence of columns to add to $D_j$. By the induction hypothesis on column $D_j$ $R_j = \Sigma_{i<j} D_i$, where the right hand side comes from Algorithm 12. We thus have, $R[k] \leftarrow R[k] + (R_j = \Sigma_{i<j} D_i)$. □

## 5.11 An Equivalent Model to Section 5.6.11 for Analyzing Algorithm 19

Consider the same random variable $T_j$, as before, as the number of triangles in $T'$ not incident to any edge chosen from a sequence of edges $C$ of length j. Recall that in the model, there are only j iterations of the modified Algorithm 19. Consider the recurrence relation:

$$T_{\mathrm{j}} = T_{\mathrm{j}-1} - Y_{\mathrm{j}}$$

where $Y_{\mathrm{j}}$ is the random variable for the number of triangles removed at step j.

Taking expectations on both sides and using linearity of expectation, we get:

$$E[Y_{\mathrm{j}}] = \frac{\Sigma_{u_1 \ldots u_{\mathrm{j}}} Y_{\mathrm{j}}(u_1 \ldots u_{\mathrm{j}-1}, u_{\mathrm{j}})}{(\binom{n-1}{2} - \mathrm{j} + 1) \cdot (\binom{n-1}{2} - \mathrm{j} + 2) \cdots \binom{n-1}{2}}$$

$$= \frac{\Sigma_{u_1 \ldots u_{\mathrm{j}-1}} \Sigma_{u_{\mathrm{j}}} Y_{\mathrm{j}}(u_1 \ldots u_{\mathrm{j}-1}, u_{\mathrm{j}})}{(\binom{n-1}{2} - \mathrm{j} + 1) \cdot (\binom{n-1}{2} - \mathrm{j} + 2) \cdots \binom{n-1}{2}}$$

$$= \frac{\Sigma_{u_1 \ldots u_{\mathrm{j}-1}} 3 \cdot (\binom{n-1}{3} - \Sigma_{k \leq \mathrm{j}-1} Y_k(u_1 \ldots, u_k))}{(\binom{n-1}{2} - \mathrm{j} + 1) \cdot (\binom{n-1}{2} - \mathrm{j} + 2) \cdots \binom{n-1}{2}}$$

by the fact that the total number of triangles incident to all remaining edges at the jth step must be (3 · the remaining number of triangles after j-1 iterations). (Think of the bipartite graph between edges (left nodes) and triangles (right nodes); the total number of remaining bipartite edges at the jth step in this bipartite graph is what we are counting. These bipartite edges represent the triangles incident to the remaining edges or equivalently, the edges incident to the remaining triangles.)

$$= \frac{\frac{\Sigma_{u_1 \ldots u_{\mathrm{j}-1}} 3 \cdot T_{\mathrm{j}-1}}{(\binom{n-1}{2} - \mathrm{j} + 2) \cdots \binom{n-1}{2}}}{\binom{n-1}{2} - \mathrm{j} + 1} = \frac{3 \cdot E[T_{\mathrm{j}-1}]}{\binom{n-1}{2} - \mathrm{j} + 1}$$

We thus have the recurrence relation:

$$E[T_{\mathrm{j}}] = E[T_{\mathrm{j}-1}] \cdot (1 - \frac{3}{\binom{n-1}{2} - \mathrm{j} + 1})$$

with $E[T_0] = \binom{n-1}{3}$. Solving the recurrence, we get:

$$= \binom{n-1}{3} \cdot \Pi_{\mathrm{i}=0}^{\mathrm{j}-1} (1 - \frac{3}{\binom{n-1}{2} - \mathrm{i}}) = \binom{n-1}{3} \cdot \Pi_{\mathrm{i}=0}^{\mathrm{j}-1} \frac{\binom{n-1}{2} - \mathrm{i} - 3}{\binom{n-1}{2} - \mathrm{i}}$$

$$= \binom{n-1}{3} \cdot \frac{(\binom{n-1}{2} - 3) \cdot (\binom{n-1}{2} - 4) \cdots (\binom{n-1}{2} - 3 - \mathrm{j} + 1)}{(\binom{n-1}{2} \cdot (\binom{n-1}{2} - 1) \cdots (\binom{n-1}{2} - \mathrm{j} + 1))}$$

Notice this is the same equation as in Section 5.6.11.

# 6. APPROXIMATING 1-WASSERSTEIN DISTANCE BETWEEN PERSISTENCE DIAGRAMS BY GRAPH SPARSIFICATION

Let $[n] \triangleq \{1, ..., n\}$ be the set of all integers from 1 to $n$ with the usual addition between integers. Let $\overline{[n]} \triangleq [n] \cup \{\infty\}$ be the extension of $[n]$ where the addition with infinity acts like: $\infty = \mathrm{i} + \infty, \forall \mathrm{i} = 1, ..., n$. On this extended set, we can define $t_{\mathrm{i}}, \mathrm{i} \in [n]$ as a real number and $t_\infty \triangleq \infty$. Let the extended plane be $(\mathbb{R} \cup \{\infty\})^2$. Certainly we have that

$$\{(t_{\mathrm{i}}, t_{\mathrm{j}}) : \mathrm{i}, \mathrm{j} \in \overline{[n]}\} \subseteq (\mathbb{R} \cup \{\infty\})^2 \tag{6.1}$$

In persistent homology, the creation and destruction times of the homology generators can be visualized in the extended plane as a persistence diagram. This is defined below:

**Definition 6.0.1.** *Given the pair of the homology functor and a filtration of simplicial complexes which define persistent homology*

$$(H_\bullet : \boldsymbol{Simp} \to \boldsymbol{Vec}, Q(\mathcal{D})) = (\{K_t\}_{t=t_1}^{t_n}, \{inc : K_{t_{\mathrm{i}}} \hookrightarrow K_{t_{\mathrm{j}}}\}_{\mathrm{i},\mathrm{j} \in [n] : \mathrm{i} \leq \mathrm{j}})) \tag{6.2}$$

*A persistence diagram (PD) is the set of points* $P \bigcup \Delta \subseteq (\mathbb{R} \cup \{\infty\})^2$ *where:*
$(\beta, \delta) \in P$ *iff*

- $\beta$ *is a creation time for some dimension* $p$ *and*
- $\delta$ *is its corresponding destruction time for a dimension* $p+1$

*and* $\Delta \triangleq \{\!\{(x,x) : x \in \mathbb{R}\}\!\}$ *where every point* $(x,x) \in \Delta$ *has infinite multiplicity.*

**The Relationship between the Persistence Diagram and the rank:**

The persistence diagram has an alternative explanation. This alternative explanation is in terms of the rank of the induced homomorphisms between homology groups in the persistence module.

The induced homomorphisms are denoted $(H_\bullet)(K_{t_i} \hookrightarrow K_{t_j})$. They are induced by the inclusion between simplicial complexes $K_{t_i} \hookrightarrow K_{t_j}$ by the homology functor $H_\bullet$ (see Figure 2.45).

For the persistence module induced by the homology functor on the filtration $Q(\mathcal{D})$, we would like to measure the rank of the composition of the induced homomorphisms across time.

For these induced homomorphisms, we can define a rank function as follows:

$$\mathrm{rank} : [n] \times \overline{[n]} \to \mathbb{Z} \tag{6.3}$$

be defined through the filtration indices as:

$$\mathrm{rank}(\mathrm{i},\mathrm{j}) \triangleq \begin{cases} \dim(\mathrm{im}(H_\bullet(K_{t_\mathrm{i}} \hookrightarrow K_{t_\mathrm{j}}))) & \mathrm{i} \le \mathrm{j} \le n \\ \dim(\mathrm{im}(H_\bullet(K_{t_\mathrm{i}} \hookrightarrow K_{t_n}))) & \mathrm{j} = \infty \end{cases} \tag{6.4}$$

A persistence diagram can be viewed as the **Möbius inversion**, or discrete derivative, of this rank function.

Möbius inversion [122] states that for a poset $(P, \le)$, for $f : P \to \mathbb{Z}, g : P \to \mathbb{Z}$

$$f(I) = \sum_{J \ge I} g(J) \text{ iff } g(I) = \sum_{J \ge I} f(J)\mu(J, I) \tag{6.5}$$

This follows by the existence of a matrix inverse.

The left hand side is called a discrete integral of $g$ and the right hand side is called a convolution between $f$ and $\mu$, forming a discrete derivative.

Letting the poset be the set of all possible intervals on $\mathcal{T} \triangleq \{t_\mathrm{i}\}_{\mathrm{i}=1}^{n}$, namely

$$\mathrm{Int}(\mathcal{T}) \triangleq (\{[t_\mathrm{i}, t_\mathrm{j})\}_{1 \le \mathrm{i} < \mathrm{j} \le n} \cup \{[t_\mathrm{i}, \infty)\}_{1 \le \mathrm{i} \le n}, \subseteq) \tag{6.6}$$

and choosing

$$f([t_\mathrm{i}, t_\mathrm{j})) \triangleq \mathrm{rank}(\mathrm{i}, \mathrm{j} - 1) \tag{6.7a}$$

$$g([t_\mathrm{i}, t_\mathrm{j})) \triangleq \begin{cases} \mathrm{rank}(\mathrm{i}, \mathrm{j}-1) - \mathrm{rank}(\mathrm{i}, \mathrm{j}) + \mathrm{rank}(\mathrm{i}-1, \mathrm{j}) - \mathrm{rank}(\mathrm{i}-1, \mathrm{j}-1) & \mathrm{i}, \mathrm{j} \in [n], \mathrm{i} \leq \mathrm{j} \\ \mathrm{rank}(\mathrm{i}, \infty) - \mathrm{rank}(\mathrm{i}-1, \infty) & \mathrm{j} = \infty \end{cases} \tag{6.7b}$$

for the Equation 6.5. It can be shown as in [123] that these choices of $f$ and $g$ satisfy the **Möbius inversion** relationship.

This means that there is a Möbius function $\mu$, which only depends on the poset on finite half open intervals and the Möbius inversion relationship.

This is the inclusion-exclusion Möbius function:

$$\mu(J, I) \triangleq \begin{cases} (-1)^{|(J \cap \mathcal{T})| - |(I \cap \mathcal{T})|} & I = [t_\mathrm{i}, t_\mathrm{j}), t_\mathrm{j} < \infty \\ (-1)^{|(J \cap \mathcal{T})| - |(I \cap \mathcal{T})|} & I = [t_\mathrm{i}, \infty) \end{cases} \tag{6.8}$$

The function $g : \mathrm{Int}(\mathcal{T}) \to \mathbb{Z}$ acts like an inclusion-exclusion of the rank function. Its physical meaning on the interval $[t_\mathrm{i}, t_\mathrm{j})$, denoted $g([t_\mathrm{i}, t_\mathrm{j}))$, is that

> *"It counts the number of independent generators that are at index* $\mathrm{i}$ *that still exist at index* $\mathrm{j}-1$ *but do not exist at index* $\mathrm{j}$ ***excluding*** *those previous generators that came from index* $\mathrm{i}-1$ *that still exist at index* $\mathrm{j}-1$ *but do not exist at index* $\mathrm{j}$*."*

This is equivalent to the number of new generators at time $t_\mathrm{i}$ that survive through a composition of induced homomorphisms till the farthest out time $t_\mathrm{j}, t_\mathrm{j} > t_\mathrm{i}$, excluding generators from before time $t_\mathrm{i}$.

We will show in the following Theorem that the persistence diagram is equivalent to the function $g$. We will also show that the persistence diagram produces an interval decomposition. It shows the following equivalence amongst the Möbius inversion, Gabriel's theorem, and Persistence diagrams:

**Theorem 6.0.2.** *Let* $\boldsymbol{Simp}$ *be the category of simplicial complexes with the following finite simple acyclic subcategory:*

$$Q(\boldsymbol{Simp}) = (\{K_{t_{\mathrm{i}}}\}_{\mathrm{i}=1,\ldots,n}, \{inc : K_{t_{\mathrm{i}}} \hookrightarrow K_{t_{\mathrm{j}}} : \mathrm{i},\mathrm{j} \in [n], \mathrm{i} \leq \mathrm{j}, t_{\mathrm{i}} \leq t_{\mathrm{j}}\}) \tag{6.9}$$

*be a totally ordered increasing sequence of times where* $H_\bullet \mid_{Q(Simp)} (t_{\mathrm{i}})$ *is a real valued vector space.*

*Let* $(H_\bullet : \boldsymbol{Simp} \to \boldsymbol{Vec}, Q(\boldsymbol{Simp}))$ *be a data representation functor on a filtration of simplicial complexes where* $im(H_\bullet \mid_{Q(\boldsymbol{Simp})})$ *corresponds to a persistence module* $M$ *of* $n$ *vector spaces and* $n-1$ *linear maps.*

*Let* $rank : [n] \times \overline{[n]} \to \mathbb{Z}$ *be defined through the functor* $H_\bullet$ *by:*

$$rank(\mathrm{i},\mathrm{j}) \triangleq dim(im(H_\bullet \mid_{Q(\boldsymbol{Simp})} (t_{\mathrm{i}} \to t_{\mathrm{j}}))) \tag{6.10}$$

*The following are equivalent up to bijections:*

1. *The Möbius Inversion of the rank map, defined in Equation 6.5:*
   - $Dgm : [n] \times \overline{[n]} \to \mathbb{Z}$
2. *The barcode:*
   - $\mathcal{B} \triangleq \{\!\{[\beta_{\mathrm{j}}, \delta_k) : \mathrm{j}, k \in [n] \setminus B\}\!\} \cup \{\!\{[\alpha_{\mathrm{i}}, \infty) : \mathrm{i} \in B \subseteq [n]\}\!\}$
3. *Persistence Diagram:*
   - $PD \setminus \Delta$ *where* $PD \triangleq \{\!\{(\alpha_{\mathrm{i}}, \infty) : \mathrm{i} \in B \subseteq [n]\}\!\} \cup \{\!\{(\beta_{\mathrm{j}}, \delta_k) : \mathrm{j}, k \in [n] \setminus B\}\!\} \cup \Delta$

   *And imply the following two isomorphic decompositions:*
4. *The indecomposable interval decomposition of* $M$*:*
   - $\bigoplus_{[t_{\mathrm{j}}, t_k) : \mathrm{j}, k \in [n] \setminus B} M_{t_{\mathrm{j}}, t_k} \oplus \bigoplus_{\mathrm{i} \in B \subseteq [n]} M_{t_{\mathrm{i}}, \infty}$
5. *The* $\mathbb{R}[t]$*-graded primary module decomposition:*
   - $(\bigoplus_{\mathrm{i} \in B \subseteq [n]} \Sigma^{t_{\mathrm{i}}} \mathbb{R}[t]) \oplus (\bigoplus_{\mathrm{j}, k \in [n] \setminus B} \Sigma^{t_{\mathrm{j}}} \frac{\mathbb{R}[t]}{(t^{t_k - t_{\mathrm{j}}})})$

*Proof.* **2. iff 3.**

This is by direct one to one correspondence between multisets. Thus, we can say that the indices for the endpoints of the barcode are the primal minimax solutions to the persistent homology matrix reduction problem.

**1. implies 2.**

- For the index j, consider $[t_{\mathrm{i}}, t_{\mathrm{j}}) \notin \mathcal{B}$. We consider two cases for the true bar $[t_{\mathrm{i}'}, t_{\mathrm{j}}) \in \mathcal{B}$:

  1. If $t_{\mathrm{i}'} < t_{\mathrm{i}}$, then the creation event at time index $\mathrm{i}' : \mathrm{i}' < \mathrm{i}$ is destroyed at time index j:

  $$\mathrm{rank}(\mathrm{i}, \mathrm{j}) = \mathrm{rank}(\mathrm{i}, \mathrm{j} - 1) + 1 \tag{6.11}$$

  and similarly since $\mathrm{i}' \leq \mathrm{i} - 1$:

  $$\mathrm{rank}(\mathrm{i} - 1, \mathrm{j}) = \mathrm{rank}(\mathrm{i} - 1, \mathrm{j} - 1) + 1 \tag{6.12}$$

  Thus:

  $$g([t_{\mathrm{i}}, t_{\mathrm{j}})) = 0 \tag{6.13}$$

  2. If $t_{\mathrm{i}'} > t_{\mathrm{i}}$, then since starting at time index i, the creation event at time index $\mathrm{i}' > \mathrm{i}$ does not affect the rank function $\mathrm{rank}(\mathrm{i}, \bullet)$, we have:

  $$\mathrm{rank}(\mathrm{i}, \mathrm{j}) = \mathrm{rank}(\mathrm{i}, \mathrm{j} - 1) \tag{6.14}$$

  and for the same reason:

  $$\mathrm{rank}(\mathrm{i} - 1, \mathrm{j}) = \mathrm{rank}(\mathrm{i} - 1, \mathrm{j} - 1) \tag{6.15}$$

  Thus:

  $$g([t_{\mathrm{i}}, t_{\mathrm{j}})) = 0 \tag{6.16}$$

- For the case of $[t_{\mathrm{i}}, \infty) \notin \mathcal{B}$. We consider two cases for the true bar $[t_{\mathrm{i}'}, \infty) \in \mathcal{B}$:

1. If $i' < i$, then since there is no creation event at time index i:

$$\mathrm{rank}(i, \infty) - \mathrm{rank}(i - 1, \infty) = 0 \tag{6.17}$$

2. If $i' > i$, then since there is no creation event at time index i:

$$\mathrm{rank}(i, \infty) - \mathrm{rank}(i - 1, \infty) = 0 \tag{6.18}$$

**2. implies 1.**

- For a bar $[t_i, t_j) \in \mathcal{B}$, we show that the Möbius inverse $g$ of the rank function is nonzero.

  Starting at index i, we know that at index j there is a sum of $p$-dimensional boundaries from indices $k : k \leq j$ that destroys the $p$-dimensional cycle created at time index i. Since only one $p$-dimensional boundary is introduced at time index j, we have that:

$$\mathrm{rank}(i, j) = \mathrm{rank}(i, j - 1) + 1 \tag{6.19}$$

  Since the index i is a minimizing creation time index for the rank drop at time index j due to a $p$-dimensional boundary, we must have that starting at time index $i - 1$:

$$\mathrm{rank}(i - 1, j) = \mathrm{rank}(i - 1, j - 1) \tag{6.20}$$

  These equations give:

$$g([t_i, t_j)) = 1 \tag{6.21}$$

- Similarly, we will show that for a bar $[t_i, \infty)$, the function $g$ is nonzero.

  Since a $p$-dimensional creation event is introduced at time index i, we must have:

$$\mathrm{rank}(i, \infty) - \mathrm{rank}(i - 1, \infty) = 1 \tag{6.22}$$

As in the finite interval case, we get:

$$g([t_{\mathrm{i}}, \infty)) = 1 \tag{6.23}$$

**4. iff 5.**

This follows by ZC correspondence theorem of [124, 125]

**1., 2., 3. implies 4., 5.:**

Let $M = \{H_\bullet \mid_{Q(\mathbf{Simp})} (t_{\mathrm{i}}) \xrightarrow{v_{t_{\mathrm{i}},t_{\mathrm{j}}}} H_\bullet \mid_{Q(\mathbf{Simp})} (t_{\mathrm{j}})\}_{\mathrm{i},\mathrm{j}\in[n]:\mathrm{i}\leq\mathrm{j},t_{\mathrm{i}}\leq t_{\mathrm{j}}}$ be the persistence module on the total order $(\{t_{\mathrm{i}}\}_{\mathrm{i}=1}^{n}, \leq)$.

By Theorem 3.2.20, all minimax primal solutions $(\mathrm{j}^*, \mathrm{i}), \mathrm{j}^* \neq 0$ with $[\beta_{\mathrm{j}^*}, \delta_{\mathrm{i}}) \in \mathcal{B}$ correspond to indecomposable interval modules $M_{t_{\mathrm{i}},t_{\mathrm{j}}}$. Similarly, the solutions $(0, k)$, with $[\alpha_k, \infty) \in \mathcal{B}$ correspond to the interval modules $M_{\alpha_k,\infty}$.

Collecting the multiset of such intervals:

$$\mathcal{M} \triangleq \{\!\{M_{t_k} : k \in B \subseteq [n]\}\!\} \uplus \{\!\{M_{t_{\mathrm{i}},t_{\mathrm{j}}} : \mathrm{i}, \mathrm{j} \notin B \subseteq [n]\}\!\} \tag{6.24}$$

and then taking the direct sum over $\mathcal{M}$:

$$\bigoplus_{M \in \mathcal{M}} M \tag{6.25}$$

Since all $p$-dimensional generators and all $p$-dimensional boundaries are considered in constructing $\mathcal{M}$, we must obtain Gabriel's interval decomposition of a persistence module. This is also mentioned in Theorem 3.2.20.

□

**Corollary 6.0.3.** *Two different persistence diagrams can give two isomorphic indecomposable interval decompositions.*

*Thus Gabriel's theorem is not enough to characterize creation and destruction times.*

*Proof.* Consider the following two filtrations over $\mathcal{D} = \mathbb{Z}$:

$$K_0 \triangleq \varnothing, K_{\mathrm{i}} \triangleq \{\bullet_{\mathrm{i}}\}, \mathrm{i} \in \{1, 2, 3\} \tag{6.26a}$$

$$K_4 \triangleq K_3 \cup \{\{\bullet_1, \bullet_2\}\} \tag{6.26b}$$

$$K_5 \triangleq K_4 \cup \{\{\bullet_5\}\} \tag{6.26c}$$

$$K_6 \triangleq K_5 \cup \{\{\bullet_6\}\} \tag{6.26d}$$

$$K_7 \triangleq K_6 \cup \{\{\bullet_3, \bullet_5\}\} \tag{6.26e}$$

and:

$$L_0 \triangleq \varnothing, L_\mathrm{i} \triangleq \{\bullet_\mathrm{i}\}, \mathrm{i} \in \{1, 2, 3, 4\} \tag{6.27a}$$

$$L_5 \triangleq L_4 \cup \{\{\bullet_1, \bullet_3\}\} \tag{6.27b}$$

$$L_6 \triangleq L_5 \cup \{\{\bullet_2, \bullet_4\}\} \tag{6.27c}$$

$$L_7 \triangleq L_6 \cup \{\{\bullet_7\}\} \tag{6.27d}$$

For the homology functor $H_0$ over these two filtrations, we have the following two barcodes:

$$\mathcal{B}_K \triangleq \{[2, 4), [5, 7), [6, \infty)\} \tag{6.28}$$

$$\mathcal{B}_L \triangleq \{[3, 5), [4, 6), [7, \infty)\} \tag{6.29}$$

According to Theorem 6.0.2, we know that the two barcodes produce two finite interval decompositions for the two persistence modules $M, N$ induced by the categories

$$(\{H_0(K_\mathrm{i})\}_{\mathrm{i}=0}^{7}, \{H_0(K_\mathrm{i} \hookrightarrow K_\mathrm{j})\}_{\mathrm{i},\mathrm{j} \in \{0,\dots,7\}:\mathrm{i} \le \mathrm{j}}) \tag{6.30}$$

$$(\{H_0(L_\mathrm{i})\}_{\mathrm{i}=0}^{7}, \{H_0(L_\mathrm{i} \hookrightarrow L_\mathrm{j})\}_{\mathrm{i},\mathrm{j} \in \{0,\dots,7\}:\mathrm{i} \le \mathrm{j}}) \tag{6.31}$$

$$M = M_{2,4} \oplus M_{5,7} \oplus M_{6,\infty} \tag{6.32}$$

$$N = N_{3,5} \oplus N_{4,6} \oplus N_{7,\infty} \tag{6.33}$$

Certainly the finite intervals from the two satisfy the isomorphisms

$$M_{2,4} \cong N_{3,5} \cong M_{5,7} \cong N_{4,6} \tag{6.34}$$

the same also holds for the half infinite intervals:

$$M_{6,\infty} \cong N_{7,\infty} \tag{6.35}$$

Thus, $M \cong N$.

However, certainly the barcodes are not equivalent: $\mathcal{B}_K \neq \mathcal{B}_L$. □

Persistence diagrams give a representation of the persistence module in the extended plane. They also happen to play a central role in topological data analysis. It would be natural to compare them in terms of their point-set representation in the extended plane. We can do this with the 1-Wasserstein distance between persistence diagrams. This brings about a question about large-scale computation and its trade-off with accuracy.

In particular, accurate computation of these PD distances for large data sets that render large diagrams may not scale appropriately with the existing methods. The main source of difficulty ensues from the size of the bipartite graph on which a matching needs to be computed for determining these PD distances. We address this problem by making several algorithmic and computational observations in order to obtain an approximation. First, taking advantage of the proximity of PD points, we *condense* them thereby decreasing the number of nodes in the graph for computation. The increase in point multiplicities is addressed by reducing the matching problem to a min-cost flow problem on a transshipment network. Second, we use Well Separated Pair Decomposition to sparsify the graph to a size that is linear in the number of points. Both node and arc sparsifications contribute to the approximation factor where we leverage a lower bound given by the Relaxed Word Mover's distance. Third, we eliminate bottlenecks during the sparsification procedure by introducing parallelism. Fourth, we develop an open source software called [1]PDoptFlow based on our algorithm, exploiting parallelism by GPU and multicore. We perform extensive experiments and show that the actual empirical error is very low. We also show that we can achieve high performance at low guaranteed relative errors, improving upon the state of the arts.

[1]↑https://github.com/simonzhang00/pdoptflow

## 6.1 Introduction

A standard processing pipeline in topological data analysis (TDA) converts data, such as a point cloud or a function on it, to a topological descriptor called the persistence diagram (PD) by a persistence algorithm [102]. See books [67, 126] for a general introduction to TDA. Two PDs are compared by computing a distance between them. By the stability theorem of PDs [127, 128], close distances between shapes or functions on them imply close distances between their PDs; thus, computing diagram distances efficiently becomes important. It can help an increasing list of applications such as clustering [129–131], classification [132–134] and deep learning [135] that have found the use of topological persistence for analyzing data. The 1-Wasserstein ($W_1$) distance is a common distance to compare persistence diagrams; Hera [136] is a widely used open source software for this. Others include [137, 138]. We develop a new approach and its efficient software implementation for computing the 1-Wasserstein distance called here the $W_1$-distance that improves the state-of-the-art.

### 6.1.1 Existing Algorithms and Our Approach

As defined in Section 6.2, the $W_1$-distance between PDs is the assignment problem on a bipartite graph [139], the problem of minimizing the cost of a perfect matching on it. Thus, any algorithm that solves this problem [140–145] can solve the exact $W_1$-distance between PDs problem.

Different algorithms computing $W_1$-distance between PDs have been implemented for open usage which we briefly survey here. For $\epsilon > 0$, the software hera [136] gives a $(1+\epsilon)$ approximation to the $W_1$-distance by solving a bipartite matching problem using the auction algorithm in $\tilde{O}(\frac{n^{2.5}}{\epsilon})$ time. In software GUDHI [104], the problem is solved exactly by leveraging a dense min-cost flow implementation from the POT library [146, 147] to solve the assignment problem. The sinkhorn algorithm for optimal transport has a time complexity of $\tilde{O}(\frac{n^2}{\epsilon^2})$ [148] but requires $O(n^2)$ memory and incurs numerical errors for small $\epsilon$. The $O(n^2)$ memory requirement is demanding for large $n$, especially on GPU. It was shown in [149] that the quadtree [150] and flowtree [151] algorithms can be adapted to achieve a $O(\log \Delta)$ approximation in $O(n \log \Delta)$ memory and time where $\Delta$ is the ratio of the largest pairwise distance between PD points divided by their closest pairwise distance. One has no control over the error with this approach and in practice the approximation factor is large. The sliced Wasserstein Distance achieves an upper bound on the error with a factor of $2\sqrt{2}$ in $O(n^2 \log n)$ time [132]. Table 6.1 shows the complexities and approximation factors of PDoptFlow and other algorithms.

**Table 6.1.** $\hat{n} \leq n$ depends on $n$, the total number of points. A better bound of $\tilde{O}(\hat{n}^2/\epsilon)$ for PDoptFlow is possible with a tighter spanner, see Section 6.3.5 for the reasoning behind our spanner choice.

| Algorithm Complexities and Approximation Factors | | | |
|---|---|---|---|
| Algorithm | Time (Sequential) | Memory | Approx. Bound |
| HERA | $\tilde{O}(\frac{n^{2.5}}{\epsilon})$ | $O(n)$ | $(1+\epsilon)$ |
| dense MCF | $\tilde{O}(n^3)$ | $O(n^2)$ | exact |
| sinkhorn | $\tilde{O}(\frac{n^2}{\epsilon^2})$ | $O(n^2)$ | $\epsilon$ abs. err |
| flowtree, quadtree | $O(n \log \Delta)$ | $O(n \log \Delta)$ | $O(\log \Delta)$ |
| WCD, RWMD | $O(n)$ and $O(n\sqrt{n})$ | $O(n)$ | none |
| sliced Wasserstein | $O(n^2 \log n)$ | $O(n^2)$ | $2\sqrt{2}$ |
| PDoptFlow | $\tilde{O}(\frac{\hat{n}^2}{\epsilon^2})$ | $O(\max(\frac{\hat{n}}{\epsilon^2}, n))$ | $1 + O(\epsilon)$ |

**Our Approach**: We design an algorithm that achieves a $(1 + O(\epsilon))$ approximation to $W_1$-distance. The input to our algorithm is two PDs and a sparsity parameter $s$ with $\epsilon = O(1/s)$.

Our approach is centered around the following theoretical result for the complexity of computing the $W_1$-distance between PDs.

**Theorem 6.1.2.** *(Main Theorem for the Complexity of Computing the $W_1$-distance)*

*Let $\varepsilon > 0$ and $A = \tilde{A} \cup \Delta, B = \tilde{B} \cup \Delta$ two PDs of atmost $n$ points,*

*The $W_1$-distance can be reduced to computing a min-cost flow on a sparse network. This can theoretically be computed in time $O(\frac{1}{\epsilon^2} n \log(n))$*

See Section 6.4 for a proof.

The computational complexity in Theorem 6.1.2 is optimal assuming the EMD problem on $\mathbb{R}^2$ cannot be computed in $O(n^{1+o(1)-\delta})$ time for any $\delta > 0$. See Theorem 6.6.1.

In order to achieve this, the problem is reduced to a min-cost flow problem on a sparsified transshipment network with sparsification determined by $s$. The min-cost flow problem is implemented with the network simplex algorithm. We use two geometric ideas to sparsify the nodes and arcs of the transshipment network. We are able to construct networks of linear complexity while availing high parallelism. This lowers the inherent complexity of the network simplex routine, and enables us to gain speedup using the GPU and multicore executions over existing implementations.

We apply a simple geometric idea called $\delta$-condensation (see Figure 6.4) to reduce the number of nodes in the transshipment network. This approach is synonymous to "grid snapping" [152, 153] or "binning" [130] to a $\delta$-grid where $\delta$ depends on $s$. In order to maintain a $(1 + O(\epsilon))$-approximation, we use a lower bound given by the Relaxed Word Mover's distance [154]. Its naive sequential computation can be a bottleneck for large PDs. We parallelize its computation with parallel nearest neighbor queries to a kd-tree data structure.

In existing flow-based approaches [147, 155] that compute the $W_1$-distance, the cost matrix is stored and processed incurring a quadratic memory complexity. We address this issue by reducing the number of arcs to $O(s^2n)$ using an $s$-well separated pair decomposition ($s$-WSPD) ($s$ is the algorithm's sparsity parameter) where $n$ is the number of nodes. This requires $O(s^2n)$ memory. Moreover, we parallelize WSPD construction in the pre-min-cost flow computation since it is a computational bottleneck. This can run in time $O(polylog(s^2n))$ according to [156]. Thus, the pre-min-cost flow computation of our algorithm incurs $O(s^2n)$ cost. We focus on the $W_1$-distance instead of the general $q$-Wasserstein distance since we can use the triangle inequality for a guaranteed $(1+\epsilon)$-spanner [157].

### 6.1.3 Experimental Results

Table 6.2 and Table 6.3 summarize the results obtained by our approach. First, we detail these results and explain the algorithms later. For our experimental setup and datasets, the reader may refer to Section 6.7. Experiments show that our methods accelerated by GPU and multicore, or even serialized, can outperform state-of-the-art algorithms and software packages. These existing approaches include GPU-sinkhorn [155], HERA [136], and dense network simplex [147] (NtSmplx). We outperform them by an order of magnitude in total execution time on large PDs and for a given low guaranteed relative error. Our approach is implemented in the software PDoptFlow, published at https://github.com/simonzhang00/pdoptflow.

**Table 6.2.** Running times of PDoptFlow, parallel (Ours) and sequential (Ours, Sq.), against HERA, GPU-sinkhorn (S.H.), and Network-Simplex (NtSmplx) for $W_1$-distance; > 32 GB or > .3 TB means out of memory for GPU or CPU respectively.

| $W_1$ Comput. Times (sec.) for Relative Error Bound | | | | |
|---|---|---|---|---|
| | bh | AB | mri | rips |
| Ours (th. error = 0.5 | 8.058s | 0.67s | 18.0s | 48.4s |
| HERA (th. error = 0.5) | 405.02s | 10.46s | 1010.7s | 207.38s |
| Ours (th. error = 0.2) | 29.15s | 1.52s | 51.5s | 154s |
| HERA (th. error = 0.2) | 405.02s | 14.56s | 1256.4s | 342.1s |
| S.H. (emp. err. ≤ 0.5) | >32GB | 3.80s | >32GB | >32GB |
| dense NtSmplx | >.3TB | 5.934s | >.3TB | 354s |
| Ours, Sq. th. err. = 0.5 | 9.13s | 0.88s | 29.3s | 80.16s |
| Ours, Sq. th. err. = 0.2 | 35.69s | 3.03s | 88.85s | 266.48s |

We also perform experiments for the nearest neighbor(NN) search problem on PDs, see Problem 6.6.1 in Section 6.6.2. This means finding the nearest PD from a set of PDs for a given query PD with respect to the $W_1$ metric. Following [149, 151], define recall@1 for a given algorithm as the percentage of nearest neighbor queries that are correct when using that algorithm for distance computation. We also use the phrase "prediction accuracy" synonymously with recall@1. Our experiments are conducted with the reddit dataset; we allocate 100 query PDs and search for their NN amongst the remaining 100 PDs. We find that PDoptFlow at $s = 1$ and $s = 18$ achieve very high NN recall@1 while still being fast, see Table 6.3. Although at $s = 1$ there are no approximation guarantees, PDoptFlow still obtains high recall@1; see Figure 6.7 and Table 6.6 for a demonstration of the low empirical error from our experiments. Other approximation algorithms [149, 151, 154] are incomparable in prediction accuracy though they run much faster.

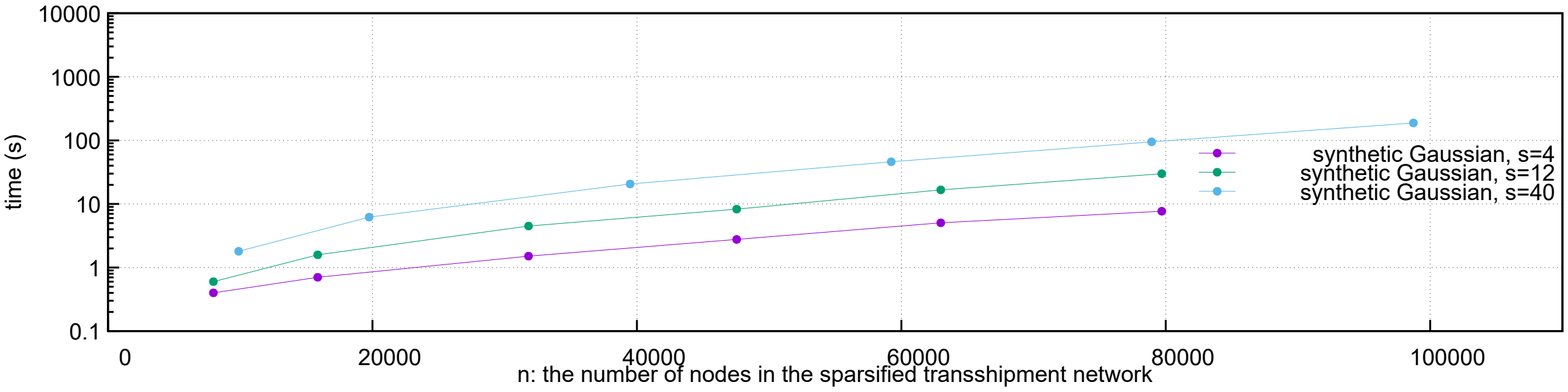

**Figure 6.1.** Plot of the empirical time (log scale) against the number of nodes $n$.

In Table 6.2, we present the total execution times for comparing four pairs of persistence diagrams: bh, AB, mri, rips from Table 6.5 in Section 6.7. The guaranteed relative error bound is given for each column. We achieve 50x, 15.6x, 56x and 4.3x speedup over HERA with the bh, AB, mri and rips datasets at a guaranteed relative error of 0.5. When this error is 0.2, we achieve a speedup of 13.9x, 9.6x, 24.4x, and 2.2x respectively on the same datasets. We achieve a speedup of up to 3.90x and 5.67x on the AB dataset over the GPU-sinkhorn and the NtSmplx algorithm of POT respectively. Execution on rips is aborted early by POT. We also run PDoptFlow sequentially, doing the same total work as our parallel approach does. A slowdown of 1.1x-2.0x is obtained on bh at $\epsilon = 0.5$ and AB at $\epsilon = 0.2$ respectively compared to the parallel execution of PDoptFlow. This suggests most of PDoptFlow's speedup comes from the approximation algorithm design irrespective of the parallelism. The Software from [149] is not in Table 6.2 since its theoretical relative error (2× (height of its quadtree)-1) [149, 151] is not comparable to values (0.5 and 0.2) from Table 6.2. In fact, it has theoretical relative errors of $75, 41, 61, 41$ for bh, AB, mri, rips respectively. FLOWTREE [149] is much faster than PDoptFlow and is less accurate empirically. On these datasets, there is a 10.1x, 2.6x, 18.8x and 90.3x speedup against PDoptFlow($s = 18$) at $\epsilon = 1.3$, for example.

Table 6.3 shows the total time for 100 NN queries amongst 100 PDs in the reddit dataset. The overall prediction accuracies using each of the algorithms are listed. See Section 6.7.1 for details on each of the approximation algorithms. Table 6.3 shows that the algorithms ordered from the fastest to the slowest on average on the reddit dataset are QUADTREE [149, 151], FLOWTREE [149, 151], WCD [154], RWMD [154], PDOPTFLOW($s = 1$), PDOPTFLOW($s = 18$), and HERA [136].

**Table 6.3.** Total time for all 100 NN queries and overall prediction accuracy over 5 dataset splits of 50/50 queries/search PDs; $\epsilon$ is the theoretical relative error.

| NN PD Search for $W_1$ Time and Prediction Accuracy | | |
|---|---|---|
| | avg. time ± std. dev. | avg. recall@1 ± std. dev. |
| QUADTREE: ($\epsilon = 37.8 \pm 0.5$) | 0.46s ± 0.05s | 2.2% ± 0.75% |
| FLOWTREE: ($\epsilon = 37.8 \pm 0.5$) | 4.88s ± 0.2s | 44% ± 4.05% |
| WCD | 8.14s ± 2.0s | 39.8% ± 2.71% |
| RWMD | 17.16s ± 0.97s | 29.8% ± 5.74% |
| PDOPTFLOW(s=1) | 62.6s ± 3.38s | 81% ± 5.2% |
| PDoptFow(s=18): ($\epsilon = 1.4$) | 371.2s ± 85s | 95.4% ± 1.62% |
| HERA: ($\epsilon = 0.01$) | 2014s ± 12.6s | 100% |

Table 6.3 also ranks the algorithms from the most accurate to the least accurate on average as HERA, PDOPTFLOW($s = 18$), PDOPTFLOW($s = 1$), FLOWTREE, WCD, RWMD, and QUADTREE. The average accuracy is obtained by 5 runs of querying the reddit dataset 100 times. PDOPTFLOW($s = 18$) provides a guaranteed 2.3-approximation which can even be used for ground truth distance since it computes 95% of the NNs accurately. Furthermore, it takes only one-fourth the time that HERA takes. Figure 6.2 shows the time-accuracy tradeoff of the seven algorithms in Table 6.3 on the reddit dataset.

Figure 6.1 shows that our overall approach runs empirically in $O(s^2 n^{1.5})$ time for small $s$ ($\leq 40$), although it would be suspected that the true complexity is $O(s^2 n^2)$. The empirical complexity improves with a smaller $s$. The datasets are given by synthetic 2D Gaussian point distributions on the plane acting as PDs. There are total of 10K, 20K, 40K, ... 100K points in the synthetic PDs. We achieve up to 20% reduction in the total number of PD points by $\delta$-condensation. Section 6.10.2 in the Appendix further explains the trend. This partly explains the speedups that Table 6.2 exhibits. For empirical relative errors, see Table 6.6 Section 6.7.

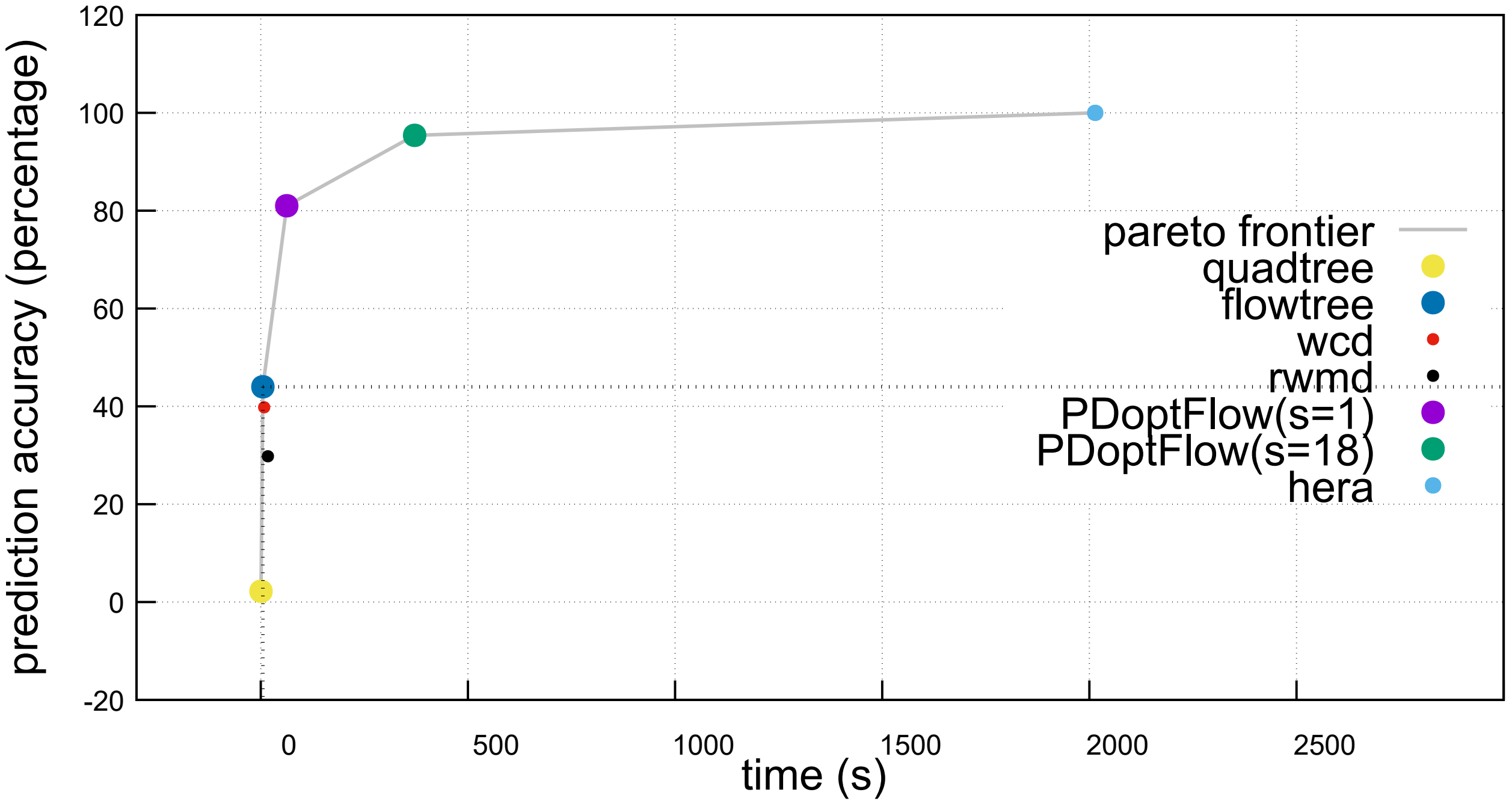

**Figure 6.2.** Pareto frontier of 7 algorithms showing the time and prediction accuracy tradeoff amongst the algorithms from Table 6.3 on the reddit dataset.

The rest of the paper explains our approach, implementation, and further experiments. Here is a table of the notations that follow.

### 6.1.4 Persistence Diagrams

A persistence diagram is a multiset of points in the plane along with the points of infinite multiplicity on the diagonal line $\Delta$ (line with slope 1). See Definition 6.0.1 for the formal explanation. The pairwise distances between diagonal points are assumed to be 0. Each point $(b, d)$, $b \neq d$ in the multiset represents the birth and death time of a topological feature as computed by a persistence algorithm [67, 102]. Diagonal points are introduced to ascertain a stability [67, 127, 158] of PDs.

Notice that for a persistence diagram $P = \tilde{P} \cup \Delta$, the size complexity of $\tilde{P}$ is $n = \sum_{p\geq 0}^{D} \beta_p$. According to Gabriel's theorem [29], the persistence diagram exists up to isomorphism be-tween persistence modules only when the $Q(\mathbf{Simp})$ is a simply laced Dynkin diagram of type $A_n$. For Vietoris Rips persistence, which satisfies Gabriel's theorem, the PD would only be constant size for $p \geq 1$ according to Theorem 5.4.14. Furthermore, for $p = 0$ all creation times are 0 so such points only lie on the $y$-axis. In particular, for most applications the betti numbers should not grow to infinity, with increasing number of samples except for $\beta_0$ which has the property that $\beta_0 \geq \Omega(n_0)$. Thus, we will mostly focus on PDs with support determined by node values instead of distances.

**Table 6.4.** Notations used.

| Notations | |
|---|---|
| Symbol | Meaning |
| $A, B$ | input PDs |
| $\tilde{A}, \tilde{B}$ | multiset of points on $\mathbb{R}^2$ (nondiagonal points of $A$ and $B$) |
| $\Delta$ | set of diagonal points |
| $\tilde{A}_{proj}, \tilde{B}_{proj}$ | multisets of projections of $\tilde{A}, \tilde{B}$ to $\Delta$ |
| $\hat{A}, \hat{B}$ | sets of points corresponding to $\tilde{A}, \tilde{B}$ |
| $\bar{a}, \bar{b}$ | virtual points that represent $\tilde{A}_{proj}$ and $\tilde{B}_{proj}$ |
| $\hat{A}^{\delta}$, $\hat{B}^{\delta}$ | $\delta$ condensation of $\hat{A}$ and $\hat{B}$ |
| $\sigma, c, f$ | supply, cost and flow functions of a transshipment network |
| $L, \delta$ | a lower bound to the $W_1$-distance, additive error |
| WCD, RWMD | word centroid distance, relaxed word movers distance |
| $s, \epsilon, n$ | sparsification factor, theoretical relative error, and $\lVert\tilde{A} \cup \tilde{B}\rVert$ |
| $G(A, B)$ | bipartite transportation network on $\hat{A} \cup \{\bar{b}\}$ and $\hat{B} \cup \{\bar{a}\}$ |
| $G_{\delta}$ | $G(\hat{A}^{\delta} \cup \{\bar{b}\}, \hat{B}^{\delta} \cup \{\bar{a}\})$ |
| $\mathsf{WS}_s(\hat{A}^{\delta} \cup \hat{B}^{\delta})$ | s-WSPD on $(\hat{A}^{\delta} \cup \hat{B}^{\delta})$ |
| $\mathsf{WS}_s^{PD}(A^{\delta}, B^{\delta})$ | sparsified transshipment network induced by $\mathsf{WS}_s(\hat{A}^{\delta} \cup \hat{B}^{\delta})$ |
| $W_1(A, B)$ | ground truth $W_1$-distance |

## 6.2 1-Wasserstein Distance Problem

Given two PDs $A = \tilde{A} \cup \Delta$ and $B = \tilde{B} \cup \Delta$, $\tilde{A}, \tilde{B} \subseteq \mathbb{R}^2 \setminus \Delta$ let

$$W_1(A, B) = \inf_{\Pi: A \to B} \sum_{x_1 \in A} (\lVert x_1 - \Pi(x_1) \rVert_2),$$

where $\Pi$ is a bijection from $A$ to $B$. Notice that this formulation is slightly different from the ones in [67, 126] which takes the $l_1$ and $l_\infty$-norms respectively instead of the $l_2$-norm considered here. It is easy to check that this is equivalent to the following formulation:

$$\inf_{M\subseteq\tilde{A}\times\tilde{B}}\Big(\sum_{(x_1,x_2)\in M}(\|x_1-x_2\|_2)+\sum_{x_1\notin\pi_1(M)}d_\Delta(x_1)+\sum_{x_2\notin\pi_2(M)}d_\Delta(x_2)\Big) \tag{6.36}$$

where $M$ is a partial one-to-one matching between $\tilde{A}$ and $\tilde{B}$; $\pi_1$, $\pi_2$ are the projections of the matching $M$ onto the first and second factors, respectively; $d_\Delta(x)$ is the $l_2$-distance of $x$ to its nearest point on the diagonal $\Delta$. The triangle inequality does not hold among the points on $\Delta$. In that sense, this $W_1$-distance differs from the classical Earth Mover's Distance (EMD) [159] between point sets with the $l_2$ ground metric. Computing $W_1(A,B)$ (Problem 6.2.1) reduces to the problem of finding a minimizing partial matching $M\subseteq\tilde{A}\times\tilde{B}$.

**Problem 6.2.1.** *(PD-EMD)*

*Given two PDs $A$ and $B$, Compute $W_1(A,B)$.*

### 6.2.1 Matching to Min-Cost Flow

Let $\tilde{A}_{proj}$, $\tilde{B}_{proj}$ be the sets of points in $\Delta$ nearest (in $l_2$-distance) to $\tilde{A}$, $\tilde{B}$, respectively. Define the bipartite graph $\mathrm{Bi}(A,B)=(U_1\dot\cup U_2,E)$ where $U_1:=\tilde{A}\cup\tilde{B}_{proj}$ and $U_2:=\tilde{B}\cup\tilde{A}_{proj}$. Define the point $p_{proj}$ to be the nearest point in $l_2$-distance to $p$ in $\Delta$ and let

$$E \;=\; (\tilde{A}\times\tilde{B})\cup\{(p,p_{proj})\}_{p\in\tilde{A}}\cup\{(q_{proj},q)\}_{q\in\tilde{B}}\cup(\tilde{A}_{proj}\times\tilde{B}_{proj}). \tag{6.37}$$

The edge $\mathrm{e}=(p,q)\in E$ has weight (i) 0 if $\mathrm{e}\in\tilde{A}_{proj}\times\tilde{B}_{proj}$, (ii) weight $\|p-q\|_2$ if $p\in\tilde{A}$, $q\in\tilde{B}$, (iii) weight $d_\Delta(p)$ if $q=p_{proj}$, and (iv) weight $d_\Delta(q)$ if $p=q_{proj}$. Because of the edges with cost 0, minimizing the total weight of a perfect matching on $B\mathrm{i}(A,B)$ is equivalent to finding a minimizing partial matching $M\subseteq\tilde{A}\times\tilde{B}$ and thus computing $W_1(A,B)$ in turn.

**Definition 6.2.2.** *Let $G=(V,E,c,\sigma)$ be a* ***transshipment network****. This consists of nodes and directed edges called arcs where we have:*

- *A supply function $\sigma:V(G)\to\mathbb{Z}$*

- *A cost function* $c : V(G) \times V(G) \to \mathbb{R}^+$*, and*

- *An uncapacitated flow function* $f : V(G) \times V(G) \to \mathbb{R}$*, which is defined by the following properites:*

  - *Nonnegativity:* $f(u,v) \geq 0, \forall u, v \in V(G)$
  - *Flow conservation out:* $\sum_w f(u,w) = |\sigma(u)|$ *for all* $u \in V(G)$*,*
  - *Flow conservation in:* $\sum_w f(w,u) = |\sigma(u)|$ *for all* $u \in V(G)$

Define the uncapacitated min-cost flow on $G$ as:

$$\min_{\Sigma_u f(u,v) = \|\sigma(v)\|, \Sigma_v f(u,v) = \|\sigma(u)\|, f(u,v) \geq 0} c(u,v) \cdot f(u,v), \text{where } (u,v) \in E(G). \tag{6.38}$$

Now we describe a construction of the bipartite transshipment network $G(A,B)$ for two PDs $A$ and $B$. Intuitively, $G(A,B)$ is $B\mathrm{i}(A,B)$ with a set instead of multiset representation for the nodes. Let $\pi_A$ and $\pi_B$ denote the mapping of the points in $\tilde{A} \cup \tilde{B}_{proj}$ and $\tilde{B} \cup \tilde{A}_{proj}$ respectively to the nodes in the graph $G(A,B)$. All points with distance 0 are mapped to the same node by $\pi_A$ and $\pi_B$. Since the diagonal points $\tilde{A}_{proj}$ and $\tilde{B}_{proj}$ are assumed to have distance zero, all points in $\tilde{A}_{proj}$ map to a single node, say $\bar{a} = \pi_A(\tilde{A}_{proj})$. Similarly, all points in $\tilde{B}_{proj}$ map to a single node, say $\bar{b} = \pi_B(\tilde{B}_{proj})$ (See Figure 6.3). We call this 0-*condensation* because it does not perturb the non-diagonal PD points. All arcs to or from $\bar{a}$ or $\bar{b}$ form *diagonal arcs*, which are used in our main algorithm.

Let $\hat{A}$ and $\hat{B}$ denote the set of nodes corresponding to the non-diagonal points, that is, $\hat{A} = \pi_A(\tilde{A})$, $\hat{B} = \pi_B(\tilde{B})$. The nodes of $G(A,B)$ are $(\hat{A} \cup \{\bar{b}\}) \dot{\cup} (\hat{B} \cup \{\bar{a}\})$. The vertices of the transshipment network $G(A,B)$ are assigned supplies $\sigma(u) = |\pi_A^{-1}(u)|$ for $u \in U_1$ and $\sigma(v) = -|\pi_B^{-1}(v)|$ for $v \in U_2$. Intuitively, negative supply at a node means that there is a demand for a net flow at that node, which corresponds to a point in $B$. The intuition for positive supply is analogous.

**Proposition 6.2.3.** *There is a perfect matching on* $\mathrm{Bi}(A,B)$ *with* $\|\tilde{A}\| = n_1$ *and* $\|\tilde{B}\| = n_2$ *iff there is a feasible flow of value* $n_1 + n_2$ *in* $G(A,B)$*.*

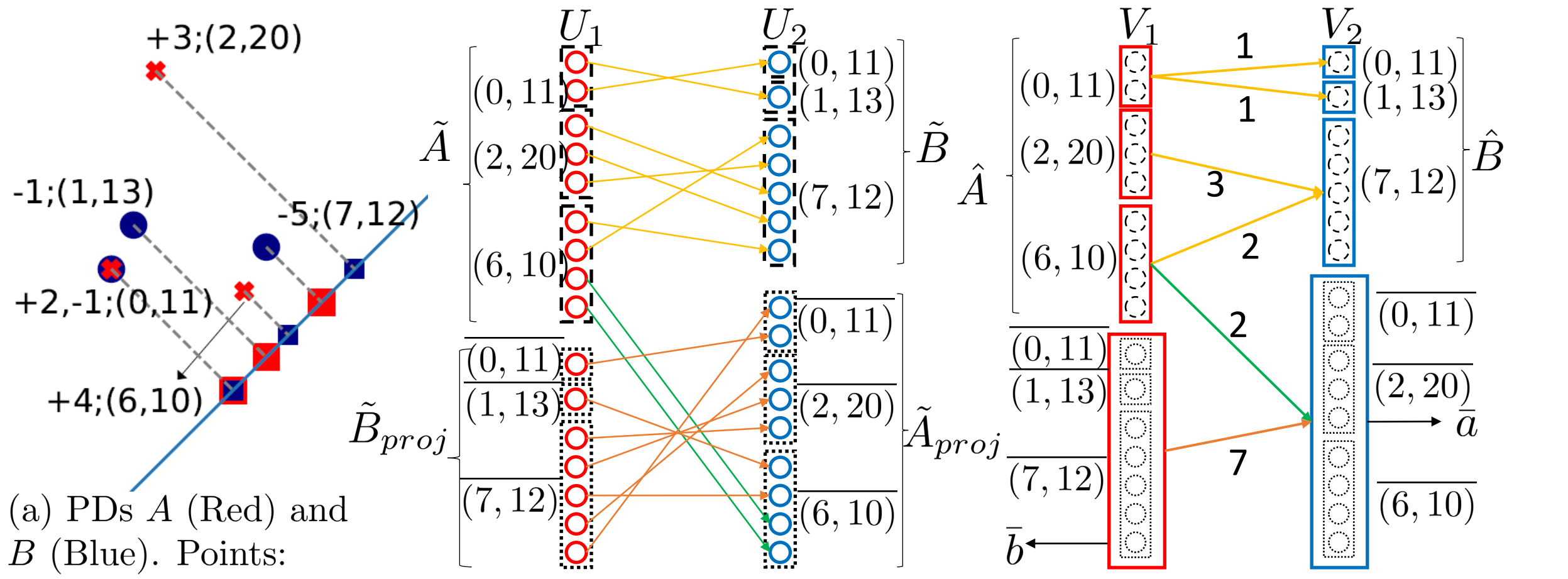

(a) PDs $A$ (Red) and $B$ (Blue). Points: (supply;(x,y))

(b) A Perfect Matching on Bi$(A,B)$

(c) A Flow of Value 9+7 on $G(A,B)$.

**Figure 6.3.** (a) $-5;(7,12)$ means a supply of $-5$ units at point $(7,12)$. (b) Bi$(A,B)$ with the nodes denoted by solid circles. (c) $G(A,B)$, nodes are the solid outer boxes. Supplies in $G(A,B)$ are set by the number of circles inside each box. In (b) and (c), barred-points e.g. $\overline{(7,12)}$ are projections to the diagonal.

*Proof.* $\Rightarrow$ Any perfect matching $\mu$ on Bi$(A,B)$ can be converted to a feasible flow on $G(A,B)$ by assigning a flow between $u \in U_1$ and $v \in U_2$ equal to the number of pairs $(p,\mu(p))$ with $p \in \pi^{-1}(u)$ and $\mu(p) \in \pi^{-1}(v)$. The supplies on $G(A,B)$ are met because of the way $G(A,B)$ is constructed. The value of the flow for the conversion is $n_1+n_2$ since there were that many pairings in the perfect matching.

$\Leftarrow$ Given a feasible flow of value $n_1+n_2$ on $G(A,B)$, we obtain a matching on Bi$(A,B)$ by observing that we can decompose any feasible flow on arc $(u,v) \in \hat{A} \cup \{\bar{b}\} \times \hat{B} \cup \{\bar{a}\}$, into unit flows from $\pi_A^{-1}(u)$ to $\pi_B^{-1}(v)$ with no pair repeating any point from other pairs. Each unit flow corresponds to a pair in the matching. Since the flow has $n_1+n_2$ flow value, there must be the $n_1+n_2$ pairings in the matching, making it perfect. □

Problem 6.2.1 reduces to a min-cost flow problem on $G(A,B)$ by Proposition 6.2.3. A proof based on linear algebra can be found in [130].

### 6.3 Approximating 1-Wasserstein Distance

We formulate the $(1+\epsilon)$-approximate PD-EMD problem in terms of a flow over a graph below:

**Problem 6.3.1.** *($(1+\epsilon)$-approximate PD-EMD)*

*Given two PDs A and B, Compute $W_1(A,B)$.*

In this section we design a $(1+O(\epsilon))$-approximation algorithm for Problem 6.2.1 that first sparsifies the bipartite graph $G(A,B)$ with an algorithm incurring a cost of $\tilde{O}_\epsilon(n)$, where $\tilde{O}_\epsilon$ hides a polylog dependence on $n$ and a polynomial dependence on $\frac{1}{\epsilon}$. Due to the node and edge sparsification, we must then use the min-cost flow formulation of Section 6.2 instead of a bi-partite matching for computing an approximation to the $W_1$-distance. We use the network simplex algorithm to solve the min-cost flow problem because it suits our purpose aptly though theoretically speaking any min-cost flow algorithm can be used.

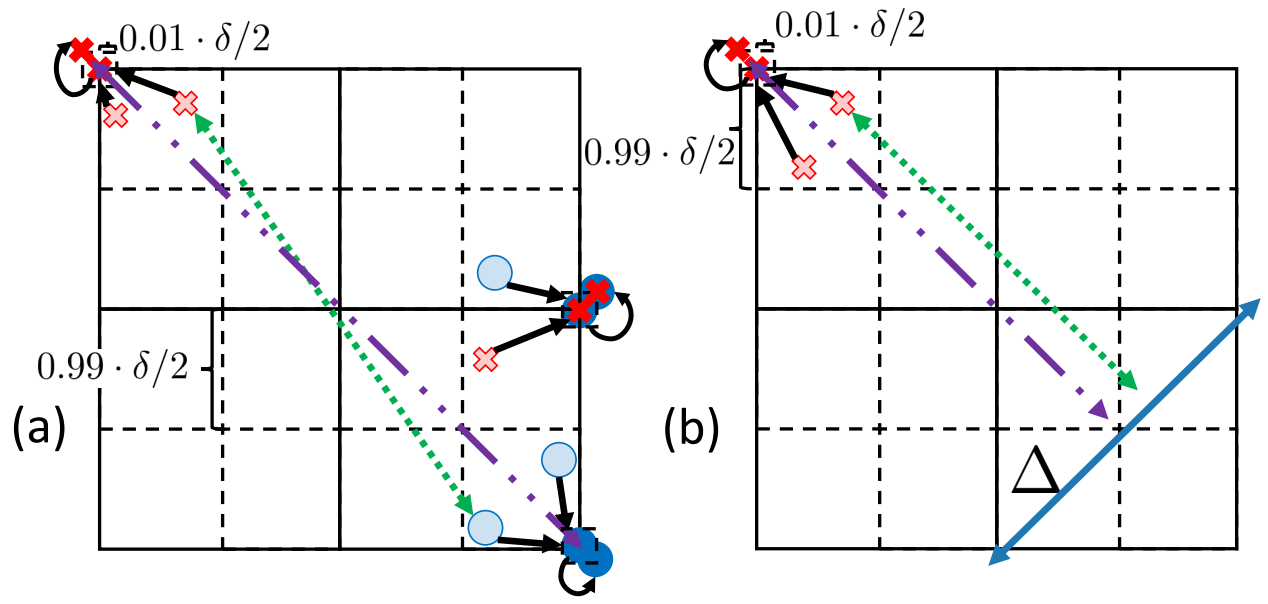

**Figure 6.4.** $\delta$-condensation for (a) matched and (b) unmatched points. Points snapped to their nearest $0.99\delta$-grid point. Points are then perturbed in a $0.01(\frac{\delta}{2})$ neighborhood. Green dotted pairwise distances change to new purple dotted and dashed pairwise distances.

#### 6.3.1 Condensation (Node Sparsification)

As the size of the PD increases, many of its points cluster together since filtration values often become close. This is because the filtration values may come from, for example, scale-free [160] social network data, that is, graphs with power law degree distributions, or voxelized data given as finitely many grid values. See Section 6.3.3 for a discussion on the clustering of PD points and how do they arise from scale-free graphs and voxel data. Figure 6.6 shows such evidences for voxelized data.

We draw upon a common technique for rasterizing the plane by snapping (or pooling) points to an evenly spaced grid to decrease the number of points. We would like to perform this operation within linear time, not affecting any complexity. At the same time, if the PDs provide a low aspect ratio property, then much of the computation is taken care of. The low aspect ratio is certainly possible for the PDs produced by filtrations of graph data, see Section 6.3.3.

We describe here our method. For a $\delta > 0$ and a fraction $k \geq 0.5$ (say $k = 0.99$), we snap nondiagonal points to a $k\delta \cdot \mathbb{Z} \times k\delta \cdot \mathbb{Z}$ lattice (grid). Let $\boldsymbol{\pi}_\delta : \hat{A} \cup \hat{B} \to (k\delta \cdot \mathbb{Z}) \times (k\delta \cdot \mathbb{Z})$ define this snapping of a point to its nearest $\delta$-lattice point where $\boldsymbol{\pi}_\delta((x,y)) = (k\delta \cdot round(\frac{x}{k\delta}), k\delta \cdot round(\frac{y}{k\delta}))$. As discussed in Section 6.10, it is known that the network simplex algorithm performs better on a transshipment network with many different arc lengths than the one with many arcs having the same length. To avoid the symmetry induced by the lattice formed by rasterizing the points, we perturb randomly the combined points. We follow the snapping by $\boldsymbol{\pi}_\delta$ with a random shift of each condensed point by at most $\frac{1-k}{2} \cdot \delta$ in any of the $\pm x$ or $\pm y$ directions; see Figure 6.4. We call the entire procedure as "$\delta$-condensation" or "$\delta$-snapping". The aggregate of the points snapped to a grid point is accounted for by a supply value assigned to it; see Algorithm 21.

**Proposition 6.3.2.** *Let $A$ and $B$ be two PDs and $\epsilon > 0$. For $\delta := \frac{2\epsilon L}{\sqrt{2}(\|\tilde{A}\| + \|\tilde{B}\|)}$ where $L \leq W_1(A,B)$, let the snapping by $\boldsymbol{\pi}_\delta$ followed by a $\delta \cdot (1-k)/2$ random shift on $A$ and $B$ produce $A^\delta$ and $B^\delta$ respectively. Then, $(1-\epsilon)W_1(A,B) \leq W_1(A^\delta, B^\delta) \leq (1+\epsilon)W_1(A,B)$.*

*Proof.* After applying $\boldsymbol{\pi}_\delta$, each point moves in a $\frac{\sqrt{2}\delta}{2}$ neighborhood. Thus for any pair of nondiagonal points $p \in \tilde{A}$ and $q \in \tilde{B}$, the $l_2$-distance between the two points shrinks/grows at most by $\frac{2\sqrt{2}k\delta}{2}$ units. A $\frac{(1-k)\delta}{2}$-perturbation contributes to an error of $\frac{2\sqrt{2}(1-k)\delta}{2}$ units for the $l_2$-distance between $p$ and $q$. Thus, for a pair of nondiagonal points the additive error incurred is $\sqrt{2}\delta$ units. Furthermore, for any nondiagonal point in either diagram, its distance to $\Delta$ can shrink/grow by at most $\frac{\sqrt{2}k\delta}{2} + \frac{\sqrt{2}(1-k)\delta}{2} = \frac{\sqrt{2}\delta}{2}$ units.

Let $m_1$ be the number of pairs of matched nondiagonal points and $m_2$ be the number of unmatched nondiagonal points. Let the $\delta$-condensation of A and B be $A^\delta, B^\delta$ and let $\delta' = \sqrt{2}(m_1 + \frac{m_2}{2})\delta$. To reach the conclusion of the proposition, we want $\delta$ to induce a relative error of $\epsilon$ for $W_1(A^\delta, B^\delta)$ with respect to $W_1(A, B)$ satisfying the following inequalities:

$$(1-\epsilon)W_1(A,B) \leq W_1(A,B) - \delta' \leq W_1(A^\delta, B^\delta) \leq W_1(A,B) + \delta' \leq (1+\epsilon)W_1(A,B) \quad (6.39)$$

Observe that $m_1 + \frac{m_2}{2} = \frac{(\|\tilde{A}\| + \|\tilde{B}\|)}{2}$. Also, we have that $L \leq W_1(A, B)$. These together constrain $\delta$ to satisfy $\sqrt{2}(m_1 + \frac{m_2}{2})\delta = \sqrt{2}\frac{(\|\tilde{A}\| + \|\tilde{B}\|)}{2}\delta \leq \epsilon L \leq \epsilon W_1(A, B)$, which gives the desired value of $\delta$ as stated. □

A lower bound $L$ from Proposition 6.3.2 is needed in order to convert the additive error of $\delta$ to a multiplicative error of $1 \pm \epsilon$. To find the lower bound $L$, we use the Relaxed Word Mover's distance (RWMD) [154] that gives a lower bound for the min-cost flow of $G(A, B)$, hence for $W_1(A, B)$. There are many lower bounds that can be used such as those from [154, 161]. However, we find RWMD to be the most effective in terms of computational time and approximation in general.

**Algorithm 20:** RWMD($\hat{A}$, $\hat{B}$, $c$)

**Input:** Point sets $\hat{A}, \hat{B}$
**Output:** Result of the RWMD computation
1 build-kd-tree on $\hat{B}$ using Euclidean distance on $\mathbb{R}^2$
2 compute $v^* \leftarrow \text{argmin}_{v \in \hat{B}} c(u, v)$; Store $f^{flow,A}(u, v^*)$ for each $u \in \hat{A}$ in parallel;
3 $L_A \leftarrow$ compute sum reduction from line 2;
4 $L_B \leftarrow$ compute lines 1-3 with $\hat{A}$ and $\hat{B}$ swapped;
5 **return** $\max(L_A, L_B)$

Recall that RWMD is a relaxation of one of the two constraints of the min-cost flow problem. If we "relax" or remove the constraint $\sum_v f(u, v) = \|\sigma(u)\|, u \in \hat{B} \cup \{\bar{a}\}$ from the min-cost flow formulation, we obtain the following feasible flow to the min-cost flow with one of its constraints removed

$$f^{low,A}(u,v) = \begin{cases} \|\sigma(u)\| & \text{if } v = \operatorname{argmin}_{v'} c(u,v') \\ 0 & \text{otherwise} \end{cases} \tag{6.40}$$

and evaluate $L_A := \sum_{u,v} c(u,v) \cdot f^{low,A}(u,v)$. Since $W_1(A,B)$ is a feasible solution to the relaxed min-cost flow problem, $L_A \leq W_1(A,B)$. Relaxing the constraint $\sum_u f(u,v) = \|\sigma(v)\|, v \in \hat{A} \cup \{\bar{b}\}$, we can define $f^{low,B}(u,v)$ and $L_B$ similarly.

Our simple parallel algorithm involves computing $L := \max(L_A, L_B)$, the RWMD, by exploiting the geometry of the plane via a kd-tree to perform fast parallel nearest neighbor queries. For $L_A$, we first construct a kd-tree for $\hat{B}$ viewed as points in the plane, then proceed to search in parallel for every $u \in \hat{A}$, its nearest $l_2$-neighbor $v^*$ in $\hat{B}$ while writing the quantity $c(u,v^*) \cdot f^{low,A}(u,v^*)$ to separate memory addresses. Noticing that the closest point to $\bar{b}$, is $\bar{a}$ at cost 0, it suffices to consider the points $\hat{A}$ to compute $L_A$. We then apply a sum-reduction to the array of products, taking $O(\log n)$ depth [162]. We apply a similar procedure for $L_B$. See Algorithm 20.

Since the kd-tree queries each takes $O(\sqrt{n})$ sequential time, we obtain an algorithm with $O(n)$ processors requiring $O(\sqrt{n} + \log n) = O(\sqrt{n})$ depth and $O(n\sqrt{n})$ work.

The algorithm for $\delta$-condensation is given in Algorithm 21. We first gather all the points based on their $x$ and $y$ coordinates called a 0-condensation; see Section 6.2. Then, we compute the RWMD in order to compute $\delta$. This $\delta$ depends on an intermediate relative error of $\epsilon$ for $\delta$-condensation, which depends on the input $s$. The quantity $\epsilon$ is chosen to be less than 1. In particular, we set $\epsilon \leftarrow \frac{8}{s-4}$ if $s \geq 12$ and $\epsilon \leftarrow 1$ otherwise; see line 3 in Algorithm 21. Finally, we snap the points of $\hat{A}$ and $\hat{B}$ to the $\delta$-grid and then perturb the condensed points in a small neighborhood. The resulting sets of points are denoted $\hat{A}^\delta$ and $\hat{B}^\delta$. For each condensed point, we aggregate the supplies of points that are snapped to it. The aggregated supply function is denoted $\sigma_{\hat{A}^\delta \cup \hat{B}^\delta \cup \{\bar{a}\} \cup \{\bar{b}\}}$. The bipartite transshipment network that could be constructed by placing arcs between all nodes from $A^\delta := \hat{A}^\delta \cup \{\bar{b}\}$ to $B^\delta := \hat{B}^\delta \cup \{\bar{a}\}$ is denoted as $G_\delta{:=}G(A^\delta, B^\delta)$. The cost $c_\delta$ is defined on arcs of $G(A^\delta, B^\delta)$ as $c_\delta(u,v) = \|u - v\|_2$

**Algorithm 21:** $\delta$-condensation

**Input:** PDs $A$, $B$, and $s > 0$

**Output:** $(\hat{A}^\delta \cup \hat{B}^\delta, \sigma_{\hat{A}^\delta \cup \hat{B}^\delta \cup \{\bar{a}\} \cup \{\bar{b}\}})$

$(\hat{A}, \bar{b}, \sigma_{\hat{A}}, \hat{B}, \bar{a}, \sigma_{\hat{B}}) \leftarrow$ 0-CONDENSE$(A, B)$;

$L \leftarrow \text{RWMD}(\hat{A}, \hat{B}, c)$ // where $c(\cdot,\cdot)$ is defined in Section 6.2

**if** $s \geq 12$ **then**

  $\epsilon \leftarrow \frac{8}{s-4}$;

**else**

  $\epsilon \leftarrow 1$;

$\delta \leftarrow \frac{2\epsilon L}{\sqrt{2}(\|\hat{A}\| + \|\hat{B}\|)}$;

$(\hat{A}^\delta, \hat{B}^\delta) \leftarrow (\pi_\delta(\hat{A}), \pi_\delta(\hat{B}))$ // snap points of $\hat{A}$, $\hat{B}$ to a common $0.99\delta$-lattice

$\sigma_{\hat{A}^\delta \cup \hat{B}^\delta \cup \{\bar{a}\} \cup \{\bar{b}\}} \leftarrow \begin{cases} \sum_{u \in \pi_\delta^{-1}(v)} \sigma(u) & \text{if } v \in \hat{A}^\delta \cup \hat{B}^\delta \\ \sigma(v) & \text{if } v \in \{\bar{a}\} \cup \{\bar{b}\} \end{cases}$;

Perturb $\hat{A}^\delta \cup \hat{B}^\delta$ in a $\frac{0.01}{2}\delta$-radius square;

**return** $(\hat{A}^\delta \cup \hat{B}^\delta, \sigma_{\hat{A}^\delta \cup \hat{B}^\delta \cup \{\bar{a}\} \cup \{\bar{b}\}})$

for $u \in \hat{A}^\delta$ and $v \in \hat{B}^\delta$. The costs $c_\delta(u, \bar{a})$ and $c_\delta(\bar{b}, v)$ are defined by the $l_2$-distances of $u$ and $v$ to $\Delta$ as in Section 6.2.1. Furthermore, the supply on all points is defined by $\sigma_{A^\delta \cup B^\delta}$. Only the nodes and supplies of this network are constructed.

### 6.3.3 For large $n$, $\delta$-condensation collects the heavy-hitter filtration values

Proposition 21 has $\delta = O(\frac{1}{n})$ where $n$ is the total number of points of both PDs. In particular, assuming $W_1(A, B)$ is bounded, we have $\delta \to 0$ as $n \to \infty$. In order for $\delta$-condensation to scale with $n$, we need to make an appropriate assumption about the empirical distribution of points for our PDs. Define the density for a point set $A \subseteq \mathbb{R}^2$ on a $\delta$-square grid as $\frac{\|A\|}{\|\Gamma_\delta\|}$ where $\|\Gamma_\delta\|$ is the number of nonempty grid cells with points from $A$.

**Proposition 6.3.4.** *For a PD $A$, the fraction of nodes eliminated from $A$ by $\delta$-condensation increases if the density of a PD $A$ on a $\delta$-grid increases.*

*Proof.* For each grid point $p \in \Gamma_\delta$, all points in a $\delta$-square neighborhood centered at $p$ snap to $p$. These new cells partition the plane just like the original grid cells and are a translation of the original grid cells. We consider this translated grid as $\Gamma_\delta$, which can only affect the number of nonempty cells by at most a factor of 4. Say a $\delta$-cell $\mathrm{i} \in \Gamma_\delta$, $\delta$ depending on $\|A\|$, has $c_\mathrm{i}$ points. We get that exactly $c_\mathrm{i}$ points collapse into one point. Thus $c_\mathrm{i} - 1$ points are eliminated. Adding this up over all nonempty cells i, we get that the fraction of nodes eliminated from $A$ is:

$$\frac{\sum_{\mathrm{i} \in \Gamma_\delta} (c_\mathrm{i} - 1)}{\|A\|} = \frac{\|A\| - \|\Gamma_\delta\|}{\|A\|} \tag{6.41}$$

It follows that if the density $\frac{\|A\|}{\|\Gamma_\delta\|}$ increases, we eliminate a larger fraction of nodes as claimed.

□

We give some usages of Proposition 6.3.4. As discussed in Section 6.1.4, we consider the case of filtration values (times) coming from 0-dimensional simplices:

(6.42) In particular, for lower star filtrations on voxel based data, we have that there are only $2^8$ possible number of filtration values to fill up, up to infinitesimal perturbations from the data. We thus have, $\|\Gamma_\delta\| \leq 2^{16}$ for all $\delta$, where $2^{16}$ is a counting bound on the number of pairs of filtration values that lie in $\mathbb{R}^2$. Then, by Proposition 6.3.4, $\delta$-condensation scales well when $n$ is sufficiently large.

This also means that the PD stays under a constant size.

(6.43) For lower star filtrations defined on degree valued nodes of scale free networks, the degree distribution is given by the power law: $P(k) \sim k^{-\gamma}$, $2 < \gamma < 3$ a constant and $k$ the degree of any node. Thus, as $n \to \infty$, we sample at most $n$ times independently from this distribution. We show that the degrees sampled won't depend on the number of samples.

Using the CDF of the power law, we get that:

$$P(k < N(\gamma)) = 1 - k^{-\gamma+1} \geq 0.99 \Rightarrow N(\gamma) = O(1 - 0.99)^{\frac{1}{-\gamma+1}} \tag{6.44}$$

We have shown that with probability 0.99, each sample is bounded by some constant threshold $N(\gamma) = O(1 - 0.99)^{\frac{1}{-\gamma+1}}$ independent of $n$. Hence, $\|\Gamma_\delta\|$ is bounded w.h.p. and by Proposition 6.3.4, we have that $\delta$-condensation eliminates an eventually increasing proportion of nodes w.h.p. as $n \to \infty$.

This means that the size of the PD converges to a constant size with high probability.

### 6.3.5 Well Separated Pair Decomposition(Arc Sparsification)

The node sparsification of $G(A,B)$ gives $G_\delta$ whose arcs are further sparsified. Using Theorem 1 in [157], we bring the quadratic number of arcs down to a linear number by constructing a geometric $(1+\epsilon)$-spanner on the point set $\hat{A}^\delta \cup \hat{B}^\delta$. For a point set $P \subseteq \mathbb{R}^2$, let its complete distance graph be defined with the points in $P$ as nodes where every pair $p,q \in P$, $p \neq q$, is joined by an edge with weight equal to $\|p-q\|_2$. Define a geometric $t$-spanner $S(P)$ as a subgraph of the complete distance graph of $P$ where for any $p,q \in P, p \neq q$, the shortest path distance $d_{SP}(p,q)$ between $p$ and $q$ in $S(P)$ satisfies the condition $d_{SP}(p,q) \leq t \cdot \|p-q\|_2$.

We compute a spanner using the well separated decomposition $s$-WSPD [163, 164]. Notice that there are many other possible spanner constructions such as $\theta$-graphs [165, 166] and others, e.g. [167, 168]. However, experimentally we find that WSPD is effective in practice, and becomes especially effective when $s$ is small. The $\theta$-graphs, for example, can be an order of magnitude slower to compute as implemented in the CGAL software [169]. This is theoretically justified by the $O(\log n)$ factor in the $O(n \log n)$ construction time of $\theta$-graphs when $n > 1024$. An $s$-WSPD is a well known geometric construction that approximates the pairwise distances between points by pairs of "$s$-well-separated" point subsets. Two point subsets $U$ and $V$ are $s$-well separated in $l_2$-norm if there exist two $l_2$ balls of radius $d$ containing $U$ and $V$ that have distance at least $d \cdot s$. An $s$-WSPD of a point set $P \subseteq \mathbb{R}^2$ is a collection of pairs of $s$-well separated subsets of $P$ so that for every pair of points $p,q \in P$, $p \neq q$, there exists a unique pair of subsets $U,V$ in the $s$-WSPD with $U \ni p$ and $V \ni q$. Each subset in an $s$-WSPD is represented by an arbitrary but fixed point in the subset. We can construct a digraph $\mathsf{WS}_s(P)$ from the $s$-WSPD on $P$ by taking the point representatives as nodes and placing biarcs between any two nodes $u,v$, that is, creating both arcs $(u,v)$ and $(v,u)$. It is known [163, 164] that $\mathsf{WS}_s(P)$, viewed as an undirected graph, is a geometric t-spanner for $t = (s+4)/(s-4)$. Putting $t = (1+\epsilon)$, this gives $s = 4 + \frac{8}{\epsilon}$. It was recently shown in [170] that by taking leftmost points as representatives in the well separated subsets, one can improve $t$ to $1 + \frac{4}{s} + \frac{4}{s-2}$. Furthermore, it is also known that $\mathsf{WS}_s(P)$ has $O(s^2 n)$ number of arcs where $n = \|P\|$.

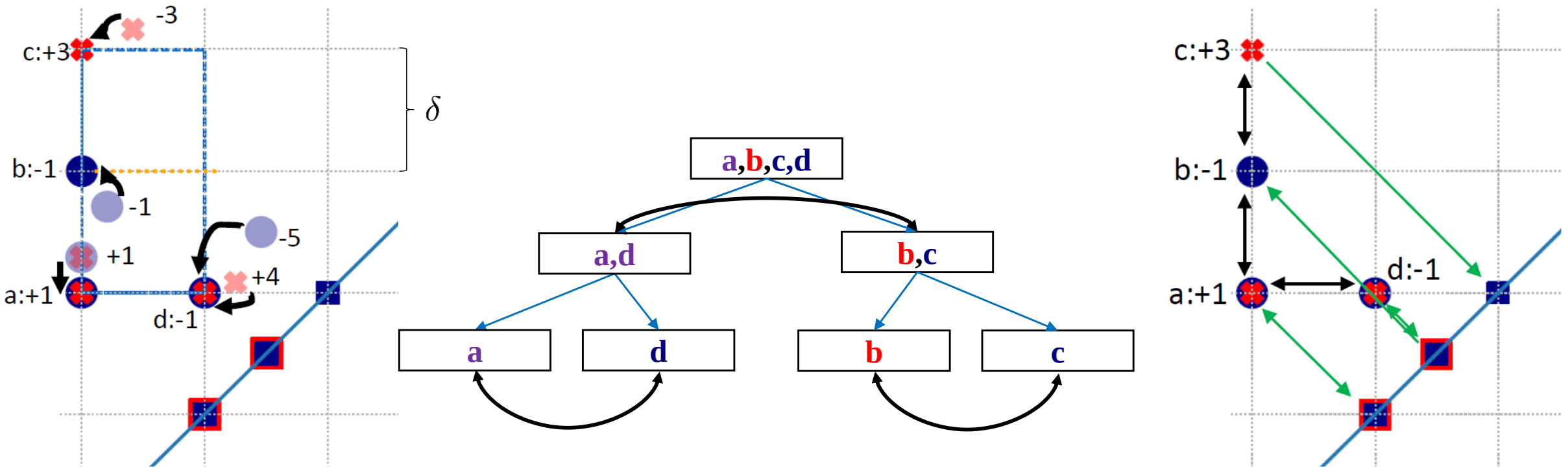

(a) $\delta$-Condensation (b) Split Tree and WSPD pairs (c) $(1+\varepsilon)$-Spanner with Diag. Arcs

**Figure 6.5.** Illustration of Algorithm 22: (a) $\delta$-condensation for the example in Figure 6.3 with the split tree construction on $\hat{A}^\delta \cup \hat{B}^\delta$; (b) WSPD pairs (black biarcs) on the split tree from (a); and (c) the induced transshipment network from the WSPD with the green diagonal arcs included.

Now we describe how we compute an arc sparsification of $G_\delta$. To save notations, we assume the points of $\hat{A}^\delta$ and $\hat{B}^\delta$, the $\delta$-condensation of $\hat{A}$ and $\hat{B}$ respectively, to be nodes also. We compute a $(1+\epsilon)$-spanner $\mathsf{WS}_s(\hat{A}^\delta \cup \hat{B}^\delta)$ via an $s$-WSPD on the points $\hat{A}^\delta \cup \hat{B}^\delta$. Notice that this digraph has all nodes of $G_\delta$ except the two diagonal nodes $\bar{a}$ and $\bar{b}$ which we add to it with all the original arcs from $\bar{a}$ and to $\bar{b}$ having the cost same as in $G_\delta$. Now we assign supplies to nodes in $\mathsf{WS}_s(\hat{A}^\delta \cup \hat{B}^\delta)$ as in $G_\delta$. There is a caveat here. It may happen that points from $\hat{A}^\delta$ and $\hat{B}^\delta$ overlap. Two such overlapped points from two sets are represented with a single point having the supply equal to the supplies of the overlapped points added together. Let $\mathsf{WS}_s^{PD}(A^\delta, B^\delta)$ denote this sparsified transshipment network. Adapting an argument in [157] to our case, we have:

**Theorem 6.3.6.** *Let $f^*$ and $\bar{f}^*$ be the min-cost flow values in $G_\delta$ and $\mathsf{WS}_s^{PD}(A^\delta, B^\delta)$ respectively where $s$ satisfies $\epsilon = \frac{4}{s} + \frac{4}{s-2}$ for some $\epsilon > 0$. Then $f^*$ and $\bar{f}^*$ satisfy $f^* \le \bar{f}^* \le (1+\epsilon) f^*$.*

*Proof.* First, notice that the nodes of $\mathsf{WS}_s^{PD}(A^\delta, B^\delta)$ are exactly the same as in $G(A^\delta, B^\delta) = (A^\delta \dot\cup B^\delta, A^\delta \times B^\delta, c_\delta, \sigma_{A^\delta \cup B^\delta})$ except the overlapped nodes. We can decompose the overlapped nodes back to their original versions in $\hat{A}^\delta$ and $\hat{B}^\delta$ with biarcs of 0-distance between them.

This will also restore the supplies at each node. This does not affect $\bar{f}^*$. Let the cost $c_\delta(u,v)$ in $\mathsf{WS}_s^{PD}(A^\delta, B^\delta)$ be the $l_2$-distance between corresponding points of $u$ and $v$ for $u, v \in \hat{A}^\delta \sqcup \hat{B}^\delta$ (all non-diagonal points pairs). Furthermore, let $c_\delta(\bar{b}, v), v \in \hat{B}^\delta$ and $c_\delta(u, \bar{a}), u \in \hat{A}^\delta$ have cost exactly as in $G(A^\delta, B^\delta)$. Recall that in $\mathsf{WS}_s^{PD}(A^\delta, B^\delta)$ there is no arc between $\hat{A}^\delta$ and $\bar{b}$ nor between $\hat{B}^\delta$ and $\bar{a}$. Treating the costs on the arcs as weights, let the shortest path distance between $u$ and $v$ be $d_{SP}^{\mathsf{WS}}(u,v)$ on $\mathsf{WS}_s^{PD}(A^\delta, B^\delta)$. We already have a $(1+\epsilon)$-spanner $\mathsf{WS}_s(\hat{A}^\delta \cup \hat{B}^\delta)$, and adding the nodes $\bar{a}$ and $\bar{b}$ with the diagonal arcs to form $\mathsf{WS}_s^{PD}(A^\delta, B^\delta)$ still preserves the spanner property, namely

$$d_{SP}(\bar{b}, v) = c_\delta(\bar{b}, v) \leq (1+\epsilon) c_\delta(\bar{b}, v) \text{ for } v \in B^\delta \tag{6.45}$$

and

$$d_{SP}(u, \bar{a}) = c_\delta(u, \bar{a}) \leq (1+\epsilon) c_\delta(u, \bar{a}) \text{ for } u \in A^\delta. \tag{6.46}$$

Let $f$ and $\bar{f}$ denote the respective flows for $f^*$ and $\bar{f}^*$. We can now prove the conclusion of the theorem.

$f^* \leq \bar{f}^*$: $\bar{f}$ can be decomposed into flows along paths from nodes in $A^\delta$ to nodes in $B^\delta$. One can get a flow $\hat{f}$ on $G(A^\delta, B^\delta)$ from $\bar{f}$ by considering a flow on every bipartite arc $(u,v)$ in $G(A^\delta, B^\delta)$ which equals the path decomposition flow from $u$ to $v$ in $\mathsf{WS}_s^{PD}(A^\delta, B^\delta)$. We have

$$f^* = \sum_{(u,v)\in G(A^\delta, B^\delta)} c_\delta(u,v) \cdot f_{uv} \leq \sum_{(u,v)\in G(A^\delta, B^\delta)} c_\delta(u,v) \cdot \hat{f}_{uv} \leq \sum_{(u,v)\in A^\delta \times B^\delta} d_P^{\mathsf{WS}}(u,v) \cdot \hat{f}_{uv} = \bar{f}^*,$$

where $d_P^{\mathsf{WS}}(u,v)$ is the path distance on $\mathsf{WS}_s^{PD}(A^\delta, B^\delta)$ as determined by the flow decomposition. The leftmost inequality follows since $\hat{f}$ is a feasible flow on $G(A^\delta, B^\delta)$ and the rightmost inequality follows since any path length between two nodes $u$ and $v$ is bounded from below by the direct distance $c_\delta(u,v)$ between the points they represent. The last equality follows by the flow decomposition.

$\bar{f}^* \leq (1+\epsilon) f^*$:

$$\bar{f}^* \leq \sum_{(u,v) \in A^\delta \times B^\delta} d_{SP}^{\mathsf{WS}}(u,v) \cdot f_{uv} \leq \sum_{(u,v) \in G(A^\delta, B^\delta)} (1+\epsilon) c_\delta(u,v) \cdot f_{uv} = (1+\epsilon) f^*.$$

The leftmost inequality follows since the flow $f$ of $G(A^\delta, B^\delta)$ sent across shortest paths forms a feasible flow on $\mathsf{WS}_s^{PD}(A^\delta, B^\delta)$. To check this, notice that the supplies are all satisfied for every node in $\mathsf{WS}_s^{PD}(A^\delta, B^\delta)$. Any intermediate node of a shortest path between $u \in A^\delta$ and $v \in B^\delta$ gets a net change of 0 supply. The rightmost inequality follows because $\mathsf{WS}_s^{PD}(A^\delta, B^\delta)$ still satisfies the $(1+\epsilon)$-spanner property as mentioned above. □

**$s$-WSPD Construction:** In order to construct an $s$-WSPD, a hierarchical decomposition such as a split tree or quad tree is constructed. We build a split tree due to its simplicity and high efficiency. A split tree can be computed sequentially with any of the standard algorithms in [163, 164, 171] that runs in $O(n \log n)$ time. It is not a bottleneck in practice. This is because there is only $O(n)$ writing to memory for constructing the tree. A simple construction of the split tree $T$ starts with a bounding box containing the input point set followed by a recursive division that splits a box into two halves by dividing the longest edge of the box in the middle. The split tree construction for a given box stops its recursion when it has one point.

Sequential construction of a WSPD involves collecting all well separated pairs of nodes which represent point subsets from the split tree $T$. This is done by searching for descendant node pairs from each interior node $w$ in $T$. For each pair of descendant nodes $u$ and $v$ reached from $w$, the procedure recursively continues the search on both children of the node amongst $u$ and $v$ that has the larger diameter for its bounding box. When the points corresponding to a pair of nodes $u, v$ become well separated, we collect $(u, v)$ in the WSPD and stop recursion.

The construction of WSPD is the primary bottleneck before the min-cost flow computation. The sequential computation incurs high data movement and also a large hidden constant factor in the complexity. To overcome these difficulties, we compute the WSPD in parallel while still preserving locality of reference, only using $n-1$ threads, and $O(n)$ auxiliary memory. We propose a simple approach on multicore that avoids linked lists or arbitrary pointers as in [171, 172]. A unique thread is assigned to each internal node $w$ in the split tree $T$. Then, we write a prefix sum [173] of the counts of well separated pairs found by each thread. Following this, each thread on $w \in T$ re-searches for well separated pairs and independently writes out its well separated descendant nodes in its memory range as determined by the prefix sum. Recursive calls on split tree node pairs can also be run in parallel as in [156]; doing so requires an unbounded data structure to store the pairs found by each thread such as a 2-layer tree with blocks at its leaves. Such a parallel algorithm can have worst-case depth of $O(\text{polylog}(s^2 n))$ and work complexity of $O(s^2 n)$. In practice, we can gain speedup in our simplified implementation, which does not issue recursive calls at interior nodes and thus has $O(s^2 n)$ depth. This is because significant work can arise at internal nodes near the leaves. For an illustration of the implementation, see Appendix.

### 6.3.7 Min-cost Flow by Network Simplex

Having constructed a sparsified transshipment network, we solve the min-cost flow problem on this network with an efficient implementation of the network simplex (NtSmplx) algorithm.

The NtSmplx algorithm is a graph theoretic version of the simplex algorithm used for linear programming. It involves the search for basic feasible min-cost flow solutions. This is done by successively applying *pivoting* operations to improve the objective function. A pivot involves an interchange of arcs for a spanning tree on the transshipment network. As observed in [174], we also find that the pivot searching phase for the incoming arc during pivoting dominates the runtime of NtSmplx. In particular, it is vital to have an efficient pivot searching algorithm: to quickly find a high quality entering arc that lessens the number of subsequent pivots. Authors in [175] propose an interpolation between Dantzig's greedy pivot rule [176] and Bland's pivot rule [177] by the block search pivot (BSP) algorithm. This implementation for NtSmplx is adopted in [178]. It is found empirically in [174] that the BSP algorithm is very efficient, simple, and results in a low number of degenerate pivots in practice. We use the BSP algorithm in our implementation because of these reasons.

Notice that if dynamic trees [179] are used, the complexity of a pivot search can be brought down to $O(\log n)$ and thus NtSmplx can run in time $\tilde{O}(s^2 n^2)$ [180, 181] on our WSPD spanner.

BSP sacrifices theoretical guarantees for simplicity and efficiency in practice. During computation, degenerate pivots, or pivots that do not make progress in the objective function may appear. There is the possibility of *stalling* or repeatedly performing degenerate pivots for exponentially many iterations. As Section 6.10 in Appendix illustrates, stalling drives the execution to a point where no progress is made. However, our experiments suggest that, before stalling, BSP usually arrives at a very reasonable feasible solution.

We observe that performance of NtSmplx depends heavily on the sparsity of our network. Since a pivot involves forming a cycle with an entering arc and a spanning tree in the network, if the graph is sparse there are few possibilities for this entering arc.

### 6.3.8 Approximation Algorithm

The approximation algorithm is given in Algorithm 22, which proceeds as follows. Given input PDs $A$ and $B$ and the parameter $s > 2$, first we set $\epsilon = \frac{8}{s-4}$. We compute a $\delta$ according to Proposition 6.3.2 using this $\epsilon$ for $s \geq 12$ and setting $\epsilon = 1$ for $2 < s < 12$. Then, we perform a $\delta$-condensation and compute an $s$-WSPD via a split tree construction on $\hat{A}^\delta \cup \hat{B}^\delta$.

We then compute $\mathsf{WS}_s(\hat{A}^\delta \cup \hat{B}^\delta)$ from the $s$-WSPD. It is a $(1+\epsilon')$-spanner for $s > 2$ where $\epsilon' = \frac{4}{s} + \frac{4}{s-2}$. Diagonal nodes along with their arcs are added to this graph as determined by $G_\delta$. This means that we add the nodes $\bar{a}$ and $\bar{b}$ and all arcs from $\hat{A}^\delta$ to $\bar{b}$ and $\bar{a}$ to $\hat{B}^\delta$. This produces $\mathsf{WS}_s^{PD}(A^\delta, B^\delta)$. Figure 6.5 shows our construction. The network simplex algorithm is applied to the sparse network $\mathsf{WS}_s^{PD}(A^\delta, B^\delta)$ to get a distance that approximates the min-cost flow value on $G_\delta$ within a factor of $(1+\epsilon')$ between inputs $A^\delta$ and $B^\delta$. The algorithm still runs for $s > 0$ instead of $s > 2$ since we can still construct a valid transshipment network for optimization. However, there are no guarantees if $s \leq 2$. Nonetheless, empirical error is found to be low and the computation turns out very efficient; see Section 6.7.

**Algorithm 22:** Approximate $W_1$-Distance Algorithm

**Input:** PDs: $A, B$; $s > 2$ the sparsity parameter; $\epsilon = \frac{8}{s-4}$ for $s \geq 12$ and $\epsilon = 1 + \frac{8}{s} + \frac{8}{s-2}$ for $2 < s < 12$

**Output:** A $(1 + O(\epsilon))$-approximation to $W_1$-distance

$(\boldsymbol{P}, \sigma_{\boldsymbol{P}}) \leftarrow \delta$-CONDENSATION$(A, B, s)$ // $\boldsymbol{P} = \hat{A}^\delta \cup \hat{B}^\delta$

$\boldsymbol{T} \leftarrow$ FORM-SPLITTREE$(\boldsymbol{P})$;

nondiag-arcs $\leftarrow$ FORM-WSPD$(\boldsymbol{T}, s)$ // $1 + \epsilon$-spanner

diag-arcs $\leftarrow$ FORM-DIAG-ARCS$(\boldsymbol{P})$ // Diagonal arcs constructed as in Section 6.2.1

$\boldsymbol{G} \leftarrow (\boldsymbol{P}, \text{nondiag-arcs} \cup \text{diag-arcs}, \sigma_{\boldsymbol{P}}, c := \text{dists}(\text{nondiag-arcs} \cup \text{diag-arcs}))$

// $\boldsymbol{G} = (V, E, \sigma_{\boldsymbol{P}}, c)$ is as defined in Definition 6.2.2

**return** *MIN-COST FLOW*$(\boldsymbol{G})$

The time complexity of the algorithm is dominated by the computation of the min-cost flow routine. Thus, all the steps of our algorithm are designed to improve the efficiency of the NtSmplx algorithm. Replacing NtSmplx with the algorithm in [142], a complexity of $\tilde{O}(ns^2 + n^{1.5})$ can be achieved. However, NtSmplx is simpler, more memory efficient, has a reasonable complexity of $\tilde{O}(s^2n^2)$ [179], and is very efficient in practice; see Figure 6.1 and Figure 6.12 in Appendix.

## 6.4 Theoretical Bounds

By Theorem 6.3.6, the spanner achieves a $(1 + \frac{4}{s} + \frac{4}{s-2})$-approximation to the min-cost flow value on the $\delta$-condensed graph. A $\delta$-condensation results in an approximation of the $W_1$-distance with a factor of $(1 \pm (\frac{8}{s-4}))$ for $s \geq 12$ and 2 for $2 < s < 12$. The factor 2 for the range $2 < s < 12$ is obtained by putting $s = 12$ in $\frac{8}{s-4}$ because $s \leq 12$ and we need $\frac{8}{s-4} > 0$. The node and arc sparsifications together guarantee an approximation factor of $((1 + \frac{4}{s} + \frac{4}{s-2})(1 \pm (\frac{8}{s-4}))) \leq (1 + \epsilon)^2$ where $\epsilon = \frac{8}{s-4}$ and $s \geq 12$. For the range $2 < s < 12$, we have $2(1 + \frac{4}{s} + \frac{4}{s-2}) = 1 + \epsilon$ where $\epsilon = 1 + \frac{8}{s} + \frac{8}{s-2}$. We are thus guaranteed a $(1 + O(\epsilon))$-approximation to the $W_1$-distance if $s > 2$ as claimed in Algorithm 22. Hence, we have the following Corollary to Theorem 6.3.6.

**Corollary 6.4.1.** *Let $\epsilon > 0$ and define $s = 4 + \frac{8}{\epsilon}$ for $s \geq 12$. Define $\delta$ in terms of $\epsilon$ as in Proposition 6.3.2. Then, $\bar{f}^*$, the min-cost flow value of $\mathsf{WS}_s^{PD}(A^\delta, B^\delta)$, is a $(1 + O(\epsilon))$-approximation of $W_1(A, B)$.*

This allows us to now prove the main theorem upon which our approach is centered.

**Proof of Main Theorem:**

*Proof.* According to Corollary 6.4.1, the sparse transhipment network $\mathsf{WS}_s^{PD}(A^\delta, B^\delta)$ has a min-cost flow that is a $(1+\varepsilon)$-approximation of $W_1$. We know that computing the min-cost flow can be computed in near linear time [182] and constructing a hierarchical decomposition tree such as a kd-tree, quadtree or split tree takes $O(n \log(n))$ time. Constructing the WSPD geometric spanner takes time complexity of $O(\frac{1}{\varepsilon^2})$ Composing then gives the complexity as stated in the Theorem. □

### 6.4.2 Conditional Lower Bound for $W_1(A, B)$

The $W_1$ distance between persistence diagrams can be viewed as a variation of the Earth mover's distance (EMD) problem from computational geometry. We state the EMD problem here:

**Problem 6.4.1.** *(EMD) Let $V_1, V_2 \subseteq \mathbb{R}^d$ be two point sets of $d$-dimensional Euclidean space. The EMD problem seeks for the minimum value of the following optimization problem:*

$$EMD(V_1, V_2) \triangleq \min_{\sigma: V_1 \to V_2} \sum_{u \in V_1} \|u - \sigma(u)\|_2 \tag{6.47}$$

*where $\sigma : V_1 \to V_2$ is a matching (injective map) between $V_1$ and $V_2$*

**Conjecture 6.5.** *(Constant Dimension EMD Conjecture)*

*For a constant $d \geq 2$, there is no $\delta > 0$ where there is a deterministic algorithm that given two lists of $n$ points from $\mathbb{R}^d$ can compute in $O(n^{2-\delta})$ time the EMD between these two lists.*

This conjecture appears reasonable since a perfect matching over a bipartite graph with an arbitrary cost matrix takes time $\Omega(n^2)$, the size of the input.

Assuming the constant dimension EMD conjecture, we show through the technique of fine-grained reduction [183] that the exact $W_1$ distance between persistence diagrams also cannot be subquadratic unless EMD can be solved in subquadratic time.

**Theorem 6.5.1.** *Let $n > 0$ be an integer and let $\epsilon > 0$*

*If the exact EMD on $\mathbb{R}^2$ and two point sets of size $n$ cannot be computed in time $O(n^{2+o(1)-\delta})$ for any $\delta > 0$, then the computation of $W_1$ between two persistence diagrams of total size $n$ cannot be computed in time $O(n^{2+o(1)-\delta'})$ for some $\delta' > 0$*

*Proof.* We do a $(n^2, n^2)$-fine grained reduction from the exact EMD for $d = 2$ to the $W_1$ problem between persistence diagrams.

**The Reduction:**

Given an input $A, B \subseteq \mathbb{R}^2$,

1. Compute $\text{diam}(A \cup B) = \max_{x,y \in A \cup B} \|x - y\|_2$. This takes time $O(n)$ time.

2. Compute the displacement vectors

$$\mathcal{D} = \{(\|p - p_{proj}(p)\|_2, (p - p_{proj}(p))_x)\}_{p \in A \cup B} \tag{6.48}$$

consisting of (magnitude, direction) pairs where $p_{proj} : \mathbb{R}^2 \to \Delta$ is the projection map to the diagonal $\Delta = \{(x, x) : x \in \mathbb{R}\}$ as given in Section 6.2.1 and $(p)_x$ is the $x$-coordinate of point $p$. This takes time $O(n)$.

3. Amongst all the vectors $(m, r) \in \mathcal{D}$ with $r = (p - p_{proj}(p))_x \leq 0$ find the vector with the largest magnitude $m$, call this maximizer $m^*$. This takes time $O(n)$.

4. Translate all points of point sets $A$ and $B$ by

$$dd \triangleq (-(m^* + n(\epsilon + 1)\text{diam}(A \cup B)), +(m^* + n(\epsilon + 1)\text{diam}(A \cup B))) \tag{6.49}$$

Call these translated point sets $A_t, B_t$. This takes time $O(n)$.

**The Reduction Maintains Correctness:**

**An optimal EMD matching iff an optimal $W_1$ matching**

We claim that $EMD(A, B) = W_1(A_t, B_t)$, namely that the $EMD(A, B)$ and $W_1$ distances don't change under translation by $dd$. In fact, the witnesses to both problems are exactly equal:

Let $\sigma^* : A \to B$ be the optimal EMD matching and let $\sigma_t^* : A_t \to B_t$ be the partial matching witnessing $W_1(A_t, B_t)$. We claim that $\sigma^* = \sigma_t^*$.

This follows since if we introduce any matching $(p, p_{proj}(p)), p \in A_t \cup B_t$ to the diagonal into $\sigma_t^*$ by replacing a match $(p_t, q_t) \in A_t \times B_t$ by the two matches $(p_t, p_{proj}(p_t)), (q_t, p_{proj}(q_t))$ the new $W_1$ cost results in the following inequality:

$$\sum_{u_t \in A_t} \|u_t - \sigma_t^*(u_t)\|_2 - \|p_t - q_t\|_2 + \|p_t - p_{proj}(p_t)\|_2 + \|q_t - p_{proj}(q_t)\|_2 \tag{6.50a}$$

$$\geq \sum_{u_t \in A_t} \|u_t - \sigma_t^*(u_t)\|_2 = W_1(A_t, B_t) \tag{6.50b}$$

This follows since

$$\|p_t - p_{proj}(p_t)\|_2 + \|q_t - p_{proj}(q_t)\|_2 \geq 2(n(\epsilon + 1)\text{diam}(A \cup B)) \tag{6.51a}$$

$$\geq \text{diam}(A \cup B) \geq \|p_t - q_t\|_2, \forall p_t, q_t \in A_t \times B_t \tag{6.51b}$$

Thus $\sigma_t^*$ cannot involve any matchings to the diagonal and thus $W_1(A_t, B_t)$ reduces to the $EMD(A_t, B_t)$. Thus $\sigma^* = \sigma_t^*$. □

### 6.5.2 Conditional Lower Bound for the $(1 + \epsilon)$ case

For an approximate EMD problem, namely a problem where the desired solution is near the original EMD, we can define the following. A $(1 + \epsilon)$-approximate EMD solution is defined by a transshipment network $G = (V_1 \cup V_2, V_1 \cup V_2 \times V_1 \cup V_2, f, c, \mu)$ with uncapacitated flow function $f : V_1 \cup V_2 \times V_1 \cup V_2 \to \mathbb{R}^+$ as defined in Equation 6.2.2, for some cost function $c : V_1 \cup V_2 \times V_1 \cup V_2 \to \mathbb{R}^+$, and some supply function $\mu : V_1 \cup V_2 \to \mathbb{R}$ with $\mu(v) = 1, \mu(w) = -1, \forall v, w \in V_1 \times V_2$ where:

$$EMD(V_1, V_2) \leq \min_{f: V_1 \times V_2 \to \mathbb{R}^+,\text{ a flow}} \sum_{(u,v) \in V_1 \times V_2} c(u, v) f(u, v) \leq (1 + \epsilon) EMD(V_1, V_2) \tag{6.52}$$

This allows us to define the $(1+\epsilon)$-approximate EMD problem:

**Problem 6.5.1.** *The* $(1+\epsilon)$*-approximate EMD problem computes the value*

$\min_{f:V_1\times V_2\to\mathbb{R}^+,\ a\ flow} \sum_{(u,v)\in V_1\times V_2} c(u,v)f(u,v)$ *from Equation 6.52.*

Within the fine-grained complexity framework [184] we show that the $W_1$ distance between PDs and the EMD problem in the plane are reducible to each other in both the exact and $(1+\epsilon)$-approximate cases.

It is known that a $(1+\epsilon)$-approximate EMD in $d$ dimensions can be computed by a randomized algorithm in time $O(n\mathrm{poly}(\frac{1}{\epsilon},\log(n)))$ [185] as well as in deterministic time $O(n(\frac{1}{\epsilon}\log(n))^{O(d)})$ [186]. Certainly the EMD can be solved in $d$ dimensions through a spanner followed by the near linear time min-cost flow algorithm of [182], making a $(1+\epsilon)$-approximate EMD computable in $O(n^{1+o(1)})$ time according to [157]. This is, in fact, faster than the $O(n\mathrm{polylog}(n))$ time algorithm of [186] due to $n^{o(1)}=O(\log(n))$. We hypothesize that for any $\delta>0$ and any $\epsilon>0$, a $(1+\epsilon)$-approximate EMD in constant $d$ dimensions cannot be solved in time $O(n^{1+o(1)-\delta})$.

This is stated in the following conjecture:

**Conjecture 6.6.** *(Constant Dimension* $(1+\epsilon)$*-approximate EMD Conjecture)*

*For a constant* $d\geq 2$*, there is no* $\delta>0$ *such that for all* $\epsilon>0$*, there is a deterministic algorithm that given two lists of* $n$ *points from* $\mathbb{R}^d$ *can compute in* $O(n^{1+o(1)-\delta})$ *time the* $(1+\epsilon)$*-approximate EMD.*

It is known through fine-grained reduction [187] that for any $\delta>0$ if a $(1+\frac{1}{n^\delta})$-approximate EMD in dimensions $\omega(\log(n))$ of Euclidean space cannot be solved in $O(n^{2-\delta})$ time, then the Hitting Sets [188] would be false. However this is separate from the finite dimensional case due to the dependency of $d$ on $n$.

It is presumed that the smaller the dimension $d\geq 2$ that the $(1+\epsilon)$-approximate EMD problem on $\mathbb{R}^d$ would be easier to solve. So there would be no contradiction that the constant dimension version of the problem is solvable in subquadratic time. We show below that assuming the hypothesis that the $(1+\epsilon)$-approximate EMD has an optimal near linear time lower bound complexity, then the $W_1$ distance between PDs has optimal lower bound complexity of near linear time.

**Theorem 6.6.1.** *Let $n > 0$ be an integer and let $\epsilon > 0$*

*If the $(1+\epsilon)$-approximate EMD on $\mathbb{R}^2$ and two point sets of size $n$ cannot be computed in time $O(n^{1+o(1)-\delta})$ for any $\delta > 0$, then a $(1+\epsilon)$-approximate computation of $W_1$ between two persistence diagrams of total size $n$ cannot be computed in time $O(n^{1+o(1)-\delta'})$ for some $\delta' > 0$*

*Proof.* We do a $(n^{1+o(1)}, n^{1+o(1)})$-fine grained reduction from the $(1+\epsilon)$-approximate EMD for $d = 2$ to the $W_1$ problem between persistence diagrams.

**The Reduction (same as in the reduction of Theorem 6.5.1):**

Given an input $A, B \subseteq \mathbb{R}^2$,

1. Compute $\text{diam}(A \cup B) = \max_{x,y \in A \cup B} \|x - y\|_2$. This takes time $O(n)$ time.

2. Compute the displacement vectors

$$\mathcal{D} = \{(\|p - p_{proj}(p)\|_2, (p - p_{proj}(p))_x)\}_{p \in A \cup B} \tag{6.53}$$

consisting of (magnitude, direction) pairs where $p_{proj} : \mathbb{R}^2 \to \Delta$ is the projection map to the diagonal $\Delta = \{(x, x) : x \in \mathbb{R}\}$ as given in Section 6.2.1 and $(p)_x$ is the $x$-coordinate of point $p$. This takes time $O(n)$.

3. Amongst all the vectors $(m, r) \in \mathcal{D}$ with $r = (p - p_{proj}(p))_x \leq 0$ find the vector with the largest magnitude $m$, call this maximizer $m^*$. This takes time $O(n)$.

4. Translate all points of point sets $A$ and $B$ by

$$dd \triangleq (-(m^* + n(\epsilon+1)\text{diam}(A \cup B)), (m^* + n(\epsilon+1)\text{diam}(A \cup B))) \tag{6.54}$$

Call these translated point sets $A_t, B_t$. This takes time $O(n)$.

**The Reduction Maintains Correctness:**

$(1+\epsilon)$**-approximate** $EMD(A, B)$ **iff** $(1+\epsilon)$**-approximate** $W_1(A_t, B_t)$

Let

$$\begin{aligned} G_{t,proj} \triangleq (A_t \cup B_t \cup p_{proj}(A_t) \cup p_{proj}(B_t), \\ (A_t \cup B_t \cup p_{proj}(A_t) \cup p_{proj}(B_t)) \times (A_t \cup B_t \cup p_{proj}(A_t) \cup p_{proj}(B_t)), \hat{f}_t, c_t, \mu_t) \end{aligned} \tag{6.55}$$

and denote

$$G \triangleq ((A \cup B), (A \cup B) \times (A \cup B), f, c, \mu) \tag{6.56}$$

as transhipment networks for the $(1+\epsilon)$-$W_1(A_t, B_t)$ and $(1+\epsilon)$-$EMD(A, B)$ problems, respectively.

$(\Rightarrow)$ :

For a witness flow $f^* : G \to \mathbb{R}^+$ that pushes all $n$ units of flow from $A$ to $B$ that solves the $(1+\epsilon)$-approximate $EMD(A, B)$ problem, define $f_t : G_{t,proj} \to \mathbb{R}^+$ as

$$f_t((u + dd, v + dd)) \triangleq f^*(u, v) \tag{6.57}$$

We claim that this flow $f_t$ obtains a $(1+\epsilon)$-approximate $W_1(A_t, B_t)$ distance. This means that it is the minimizer of the following distance:

$$\hat{W}_1(A_t, B_t) \triangleq \min_{\hat{f}_t : G_{t,proj} \to \mathbb{R}^+ \text{ is a flow}} F_{c_t}(\hat{f}_t) \tag{6.58}$$

where:

$$F_{c_t}(\hat{f}_t) \triangleq \sum_{(u_t, v_t) \in G_{t,proj}} c_t(u_t, v_t) \hat{f}_t(u_t, v_t) \tag{6.59}$$

satisfying:

$$W_1(A_t, B_t) \le \hat{W}_1(A_t, B_t) \le (1+\epsilon) W_1(A_t, B_t) \tag{6.60}$$

We first notice that any flow $\hat{f}_t$ cannot involve any flow to the diagonal. Similar to the proof above, we have that pushing flow towards the diagonal will increase the cost. This can be expressed as:

$$\begin{aligned} &F_{c_t}(\hat{f}_t) - c_t(p, q)\hat{f}_t(p, q)\rho \\ +\rho(\|p - p_{proj}(p)\|_2 \hat{f}_t(p, p_{proj}(p)) &+ \|q - p_{proj}(q)\|_2 \hat{f}_t(p_{proj}(q), q)) \end{aligned} \tag{6.61a}$$

$$\ge F_{c_t}(\hat{f}_t) \tag{6.61b}$$

$$\text{s.t. } \forall \rho : 0 < \rho \le 1, \hat{f}_t(p, p_{proj}(p)) + \hat{f}_t(p_{proj}(q), q) = \hat{f}_t(p, q) \tag{6.61c}$$

Where the inequality of Equation 6.61b comes from the following inequality on the projection distances.

$$\|p - p_{proj}(p)\|_2 \hat{f}_t(p, p_{proj}(p)) + \|q - p_{proj}(q)\|_2 \hat{f}_t(p_{proj}(q), q) \tag{6.62a}$$

$$\geq ((1+\epsilon)n\text{diam}(A_t \cup B_t))\hat{f}_t(p,q) \geq \text{diam}(A_t \cup B_t))\hat{f}_t(p,q) \geq c_t(p,q)\hat{f}_t(p,q) \tag{6.62b}$$

Let

$$G_t \triangleq (A_t \cup B_t, (A_t \cup B_t) \times (A_t \cup B_t), \hat{f}_t |_{G_t}, c_t |_{(A_t \cup B_t)\times(A_t \cup B_t)}, \mu_t |_{A_t \cup B_t}) \tag{6.63}$$

We know that $f_t$ does not involve flow to or from the diagonal. The flow $f_t$ also cannot be improved with flow to or from the diagonal. Furthermore, since $f_t$ is optimal on $G_t$, it must be that $f_t$ is the optimal solution for $\hat{W}_1$.

$(\Leftarrow)$ :

We know that any $\hat{f}_t$ cannot be improved with any flow to or from the diagonal. Thus letting the flow $\hat{f}_t |_{G_t} = \hat{f}_t$ and optimizing $F_{c_t}(\hat{f}_t)$, we get that:

$$f_t((u + dd, v + dd)) \triangleq f^*(u,v) \tag{6.64}$$

is an optimal solution. □

### 6.6.2 Approximate Nearest Neighbor Bound:

Define the following problem using the solution to Problem 6.2.1.

**Problem 6.6.1.** *Given PDs $A_1, \ldots, A_n$ and a query PD $B$, find the nearest neighbor (NN) $A^* = \text{argmin}_{A_i \in \{A_1,\ldots,A_n\}} W_1(B, A_i)$.*

We obtain the following bound on the approximate NN factor of our algorithm, where a $c$-approximate nearest neighbor $A^*$ to query PD $B$ among $A_1 \ldots A_n$ means that $W_1(A^*, B) \leq c \cdot \min_i(W_1(A_i, B))$.

**Theorem 6.6.3.** *Let $4 + \frac{8}{\epsilon} = s \geq 12$. The nearest neighbor of PD $B$ among PDs $A_1, \ldots A_n$ as computed by* PDoptFlow *at sparsity parameter $s$ is a $\frac{(1+\epsilon)^2}{1-\epsilon}$-approximate nearest neighbor in the $W_1$-distance.*

*Proof.* For a given $s$, define $\epsilon = \frac{8}{s-4}$ and an appropriate $\delta$ as in Proposition 6.3.2. Let $A'$ be the nearest neighbor according to PDoptFlow at sparsity parameter $s$ and $B$ be the query PD. Let $f^*_{A',B}$ be the optimal flow between $A'$ and $B$ and let $f^*_{A'^\delta,B^\delta}$ be the optimal flow on the pertrubed $\delta$-grid and let $f^s_{A',B}$ be the optimal flow between them on the sparsified graph with parameter $s$. Let $X$ be the union of all PDs of interest such as $A_1 \ldots A_n$ and $B$. Let $X^\delta$ be the perturbed grid obtained by snapping $X$. Let PDoptFlow$_s$ denote the value of the optimal flow computed by PDoptFlow for sparsity parameter $s$. We have that:

$W_1(A', B) = \sum_{(x,y)\in X\times X} f^*_{A',B} \cdot \|x - y\|_2$
$\leq (\frac{1}{1-\frac{8}{s-4}}) \cdot \sum_{(x',y')\in X^\delta \times X^\delta} f^*_{A'^\delta,B^\delta} \cdot \|x' - y'\|_2$ (lower bound from Proposition 6.3.2)
$\leq (\frac{1}{1-\frac{8}{s-4}}) \cdot \sum_{(x',y')\in \mathsf{WS}_s(\hat{A'}^\delta,\hat{B}^\delta)} f^s_{A'^\delta,B^\delta} \cdot \|x' - y'\|_2$ (optimality of $f^*_{A'^\delta,B^\delta}$)
$= (\frac{1}{1-\frac{8}{s-4}}) \cdot$ PDoptFlow$_s(A'^\delta, B^\delta)$
$\leq (\frac{1}{1-\frac{8}{s-4}}) \cdot \sum_{(x',y')\in \mathsf{WS}_s(\hat{A^*}^\delta,\hat{B}^\delta)} f^s_{A^{*\delta},B^\delta} \cdot \|x' - y'\|_2$ (optimality of $A'$ w.r.t. PDoptFlow$_s$)
$= (\frac{1}{1-\frac{8}{s-4}}) \cdot$ PDoptFlow$_s(A^{*\delta}, B^\delta)$
$\leq (\frac{1}{1-\frac{8}{s-4}}) \cdot (1 + \frac{8}{s-4}) \cdot (1 + \frac{4}{s} + \frac{4}{s-2}) \cdot W_1(A^*, B)$ (by Corollary 6.4.1)
$\leq \frac{(1+\epsilon)^2}{1-\epsilon} \cdot W_1(A^*, B)$ (if $4 + \frac{8}{\epsilon} = s \geq 12$ and by Corollary 6.4.1) □

This bound matches with our experiments described in Section 6.7.1 which show the high NN prediction accuracy of PDoptFlow.

## 6.7 Experiments

All experiments are performed on a high performance computing platform [189]. The node we use is equipped with an NVIDIA Tesla V100 GPU with 32 GB of memory. The node also has a dual Intel Xeon 8268 with a total of 48 cores where 300 GB of CPU DRAM is used for computing. Table 6.5 describes the persistence diagrams data we used for all experiments.

The Athens and Beijing (producing pair AB) are real-world images taken from the public repository of [190]. MRI750 and MRI751 (producing pair mri) are adjacent axial slices of a high resolution 100 micron brain MRI scan taken from the data used in [191]. The images are saved as csv and jpeg files, respectively. The H0 barcodes of the lower star filtration are computed using ripser.py [192]. The MRI scans are perturbed by a small pixel value to remove any pixel symmetry from natural images. The brain and heart (producing pair bh) 3d models are vti [193] files converted from raw data and then converted to a bitmap cubical complex. The brain and heart raw data are from [194] and [195]. The H1 barcodes of the lower star filtration of the bitmap cubical complex are computed with GUDHI [104]. Datasets rips1 and rips2 (producing pair rips) consist of 7000 randomly sampled points from a normal distribution on a 5000 dimensional hypercube of seeds 1 and 2 respectively from the numpy.random module [196]. The Rips barcodes [103] for H1 are computed by RIPSER++ [3]. The reddit dataset is taken from [149] and is made up of 200 PDs built from the extended persistence of graphs from the reddit dataset with node degrees as filtration height values.

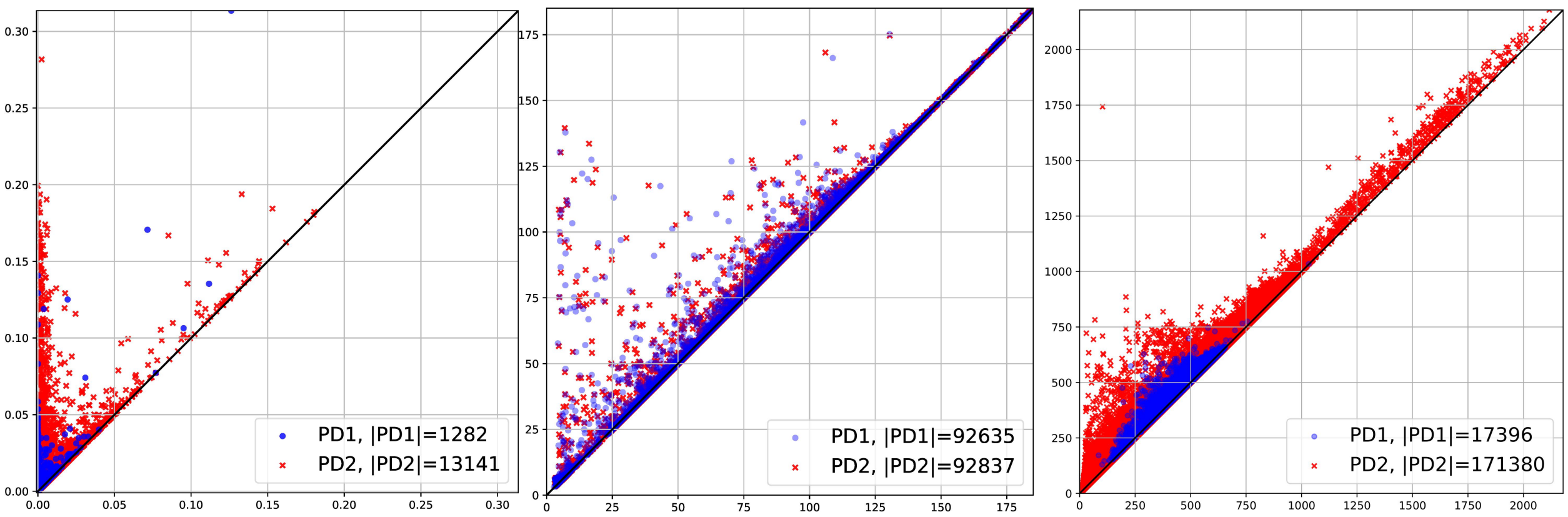

(a) PD1: Athens, PD2: Beijing (b) PD1: MRI750, PD2: MRI751 (c) PD1: brain, PD2: heart

**Figure 6.6.** Some of the persistence diagrams; PD1 is in blue and PD2 is in red.

**Table 6.5.** Datasets used for all experiments.

| Datasets | | | | |
|---|---|---|---|---|
| Name | Multiset Card. | Unique Points | Type of Filtration | Orig. Data |
| Athens | 1281 | 1226 | H0 lower star | csv image |
| Beijing | 13141 | 13046 | H0 lower star | csv image |
| Brain | 17396 | 17291 | H1 low. star cubical | 3d vti file |
| Heart | 171380 | 171335 | H1 low. star cubical | 3d vti file |
| MRI750 | 92635 | 92635 | H0 low. star pertb. | jpg img. |
| MRI751 | 92837 | 92837 | H0 low. star pertb. | jpg img. |
| rips1 | 31811 | 31811 | H1 Rips | pnt. cloud |
| rips2 | 38225 | 38225 | H1 Rips | pnt. cloud |
| Name | Avg. Card. | Avg. Card. | Type of Filtration | Orig. Data |
| reddit | 278.55 | 278.55 | lower/upper star | graphs |

The input to our algorithm contains the parameter $s$ with which we determine a $\delta$ for $\delta$-condensation and construct an $s$-WSPD. The larger the $s$ is, the smaller the average supply of each node in the transshipment network and the denser the network becomes since it has $O(s^2 n)$ number of arcs for $n$ points. Since there is a quadratic dependence on $s$, it is best to use $s \in (0, 18]$ on a conventional laptop for memory capacity reasons. Figure 6.7 shows the empirical dependence of the relative error $\epsilon'$ w.r.t. the parameter $s$. To calculate a tighter theoretical bound $\epsilon$ on the relative error than Corollary 6.4.1, one can solve for $s$ from the expression $1 + \epsilon = (1 + \frac{4}{s} + \frac{4}{s-2}) \cdot (1 + (\frac{8}{s-4}))$

In practice the algorithm performs very well in both time and relative error with $s < 12$, see Figure 6.7 and Table 6.6. Compared to FLOWTREE [149], PDOPTFLOW is surprisingly not that much slower for $n \sim 100K$ and $s \leq 1$ (a very low sparsity factor) and has a smaller relative error. Since the FLOWTREE algorithm only needs one pass through the tree, it is very efficient. On the other hand, our algorithm depends on the cycle structure of the sparsified transshipment network. The relative error of PDOPTFLOW may heavily depend on the amount of $\delta$-condensation; see bh, for example.

The $\delta$-condensation can significantly change the number of nodes in the transshipment network. From the graph $G(A, B)$, the number of nodes in $W_{40}^{PD}(A^\delta, B^\delta)$ can drop by 90%, 82%, 70%, and 2% for the bh, AB, mri, and rips comparisons respectively. The great variability is determined by the clustering of points in the PDs. Since the range of pairwise distances for a random point cloud in high dimensions (5000) is much greater than the distribution of $2^8$ pixel values of a natural image due to the curse of dimensionality, there is almost no clustering of filtration values, see Table 6.5 and Table 6.7. In cases like these it is actually advised not to use $\delta$-condensation and just a spanner instead so that one may get a tighter theoretical approximation bound. More condensation results in higher empirical relative errors and less computing time. Since only the theoretical relative error is known before execution, we compare times for a given theoretical relative error bound as in Table 6.2.

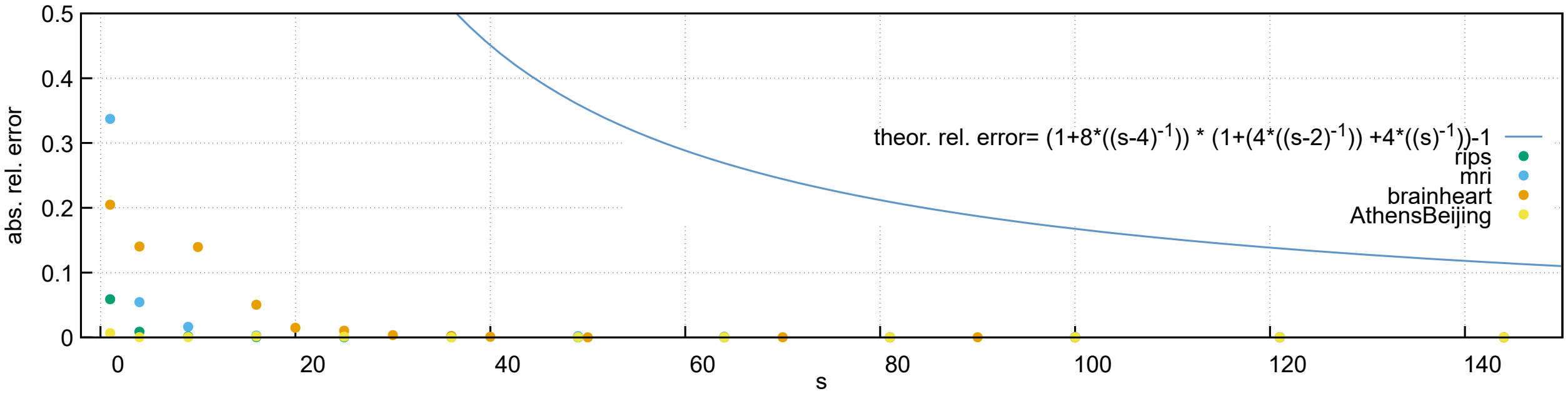

**Figure 6.7.** Convergence of PDoptFlow for $W_1$-distance against the parameter $s$.

**Table 6.6.** Empirical relative error of PDoptFlow and hera.

| $W_1$ Empirical Errors for a given Theoretical Error | | | | |
|---|---|---|---|---|
| PD data sets | Emp. Err. $s$ = 40 (Ours) | Emp. Err. $\epsilon$ = 0.5(hera) | Emp. Err. $s$ = 93 (Ours) | Emp. Err. $\epsilon$ = 0.2(hera) |
| bh | 0.00093 | 0.00028 | 0.00014 | 0.000280 |
| AB | 0.00043 | 0.00101 | 8.6e-5 | 0.000233 |
| mri | 0.00224 | 0.00373 | 0.00077 | 0.001315 |
| rips | 0.00011 | 0.00689 | 3.4e-5 | 0.001770 |

**Table 6.7.** $\delta$-condensation statistics. K: $\times 10^3$, M: $\times 10^6$.

| $W_1$-Distance Computation Stats. for a Guaranteed Rel. Error Bound | | |
|---|---|---|
| PD data sets | %node drop, (#nodes, #arcs) for $W_{40}^{PD}(A^\delta, B^\delta)$ | %node drop, (#nodes, #arcs) for $W_{90}^{PD}(A^\delta, B^\delta)$ |
| bh | 90%,(18K,22M) | 86%,(25K,92M) |
| AB | 82%,(2.5K,1.4M) | 70%,(4.3K,6.1M) |
| mri | 70%,(55K,57M) | 67%,(60K,188M) |
| rips | 2%,(68K,133M) | 0.3%,(69K,468M) |

### 6.7.1 Nearest Neighbor Search Experiments

We perform experiments in regard to Problem 6.6.1. NN search is an important problem in machine learning [151, 197], content based image retrieval [198], in high performance computing [199, 200] and recommender systems [201]. We use the dataset given in [149] which consists of 200 PDs coming from graphs generated by the reddit dataset. Having established ground truth with the guaranteed 0.01 approximation of HERA, we proceed to find the nearest neighbor for a given query PD. Following [151], we consider various approximations to the $W_1$-distance. We experiment with 6 different approximations: the Word Centroid Distance (WCD), RWMD [154], QUADTREE, FLOWTREE [149], PDoptFlow at $s = 1$ and PDoptFlow at $s = 18$ for a guaranteed 2.3 factor approximation. The WCD lower bound is achieved with the observation in [149]. Table 6.3 shows the prediction accuracies and timings of all approximation algorithms considered on the reddit dataset. Sinkhorn or dense network simplex are not considered in our experiments because they require $O(n^2)$ memory. This is infeasible for large PDs in general.

Although PDoptFlow is fast for the error that it can achieve, the computational time to use PDoptFlow for all comparisions is still too costly, however. This suggests combining the 7 considered algorithms to achieve high performance at the best prediction accuracy. One way of combining algorithms is through pipelining, which we discuss next.

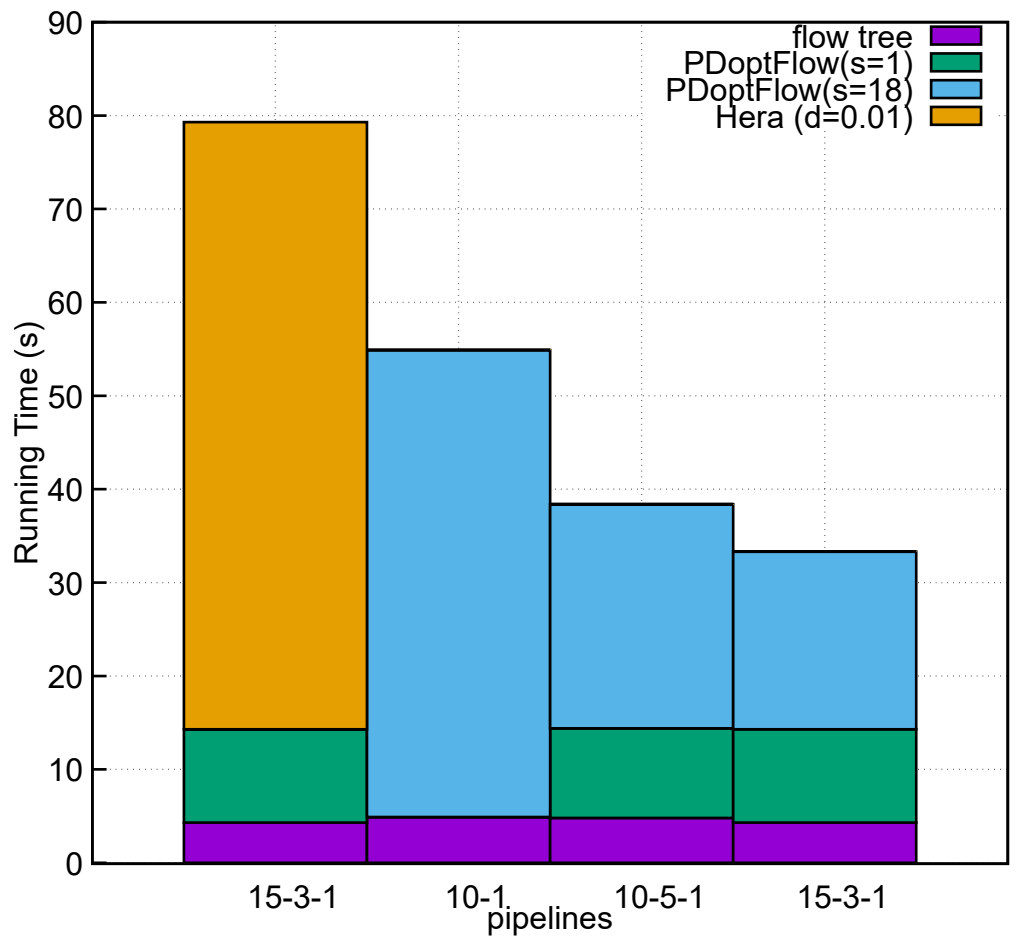

**Figure 6.8.** Pipelines for computing NN.

**Pipelining Approximation Algorithms**: Following [151] and using a distance to compute a set of candidate nearest neighbors, we pipeline these algorithms in increasing order of their accuracy to find the 1-NN with at least 90% accuracy. A pipeline of $k$ algorithms is written as $c_1 - c_2 - \cdots - c_k$ where $c_i$ is the number of output candidates of the ith algorithm in the pipeline.

Since FLOWTREE achieves a better accuracy than RWMD and WCD in less time, we can eliminate WCD and RWMD from any pipeline experiment. This is illustrated by WCD and RWMD not being on the Pareto frontier in Figure 6.2. The QUADTREE algorithm is not worth placing into the pipeline since its accuracy is too low; it prunes the NN as a potential output PD too early. It also can only save on FLOWTREE's time, which is not the bottleneck of the pipeline. In fact, the last stage of computation, which can only be achieved with a high accuracy algorithm such as HERA or PDOPTFLOW, always forms the bottleneck to computing the NN.

Figure 6.8 shows four pipelines involving FLOWTREE, PDOPTFLOW and HERA. The 15-3-1 pipeline consisting of FLOWTREE then PDOPTFLOW(S=1) and then PDOPTFLOW(S=18) was found to be the best in performance through grid search. Three other pipelines computed in the grid search are shown. Each pipeline computes 100 queries with at least 90% accuracy for a random split of the reddit dataset. We measure the total amount of time it takes to compute all 100 queries. For the pipeline 15-3-1 with HERA replacing PDOPTFLOW(S=18), HERA takes 65 seconds on 3 queries, while PDOPTFLOW takes 19 seconds on 3 queries. We find that 82% of the time is spent on only 3% of the PDs for HERA, while 57% of the time is spent if PDOPTFLOW($s = 18$) replaces HERA. We notice that FLOWTREE is able to eliminate a large number of candidate PDs in a very short amount of time though it is not able to complete the task of finding the NN due to its low prediction accuracy. PDOPTFLOW($s = 1$) surprisingly achieves very good times and prediction accuracies without an approximation bound.

### 6.8 Conclusion

We propose a new implementation for computing the $W_1$-distances between persistence diagrams that provides a $1 + O(\epsilon)$ approximation. We achieve a considerable speedup for a given guaranteed relative error in computation by two algorithmic and implementation design choices. First, we exploit geometric structures effectively via $\delta$-condensation and $s$-WSPD, which sparsify the nodes and arcs, respectively, when comparing PDs. Second, we exploit parallelism in our methods with an implementation in GPU and multicore. Finally, we es-tablish the effectiveness of the proposed approaches in practice by extensive experiments. Our software PDoptFlow can achieve an order of magnitude speedup over other existing software for a given theoretical relative error. Furthermore, PDoptFlow overcomes the computational bottleneck to finding the NN amongst PDs and guarantees an $O(1)$ approximate nearest neighbor. One merit of our algorithm is its applicability beyond comparing persistence diagrams. The algorithm is in fact applicable to an unbalanced optimal transport problem on $\mathbb{R}^2$ upon viewing $\bar{b}$ and $\bar{a}$ as creator/destructor and reassigning the diagonal arc distances to the creation/destruction costs.

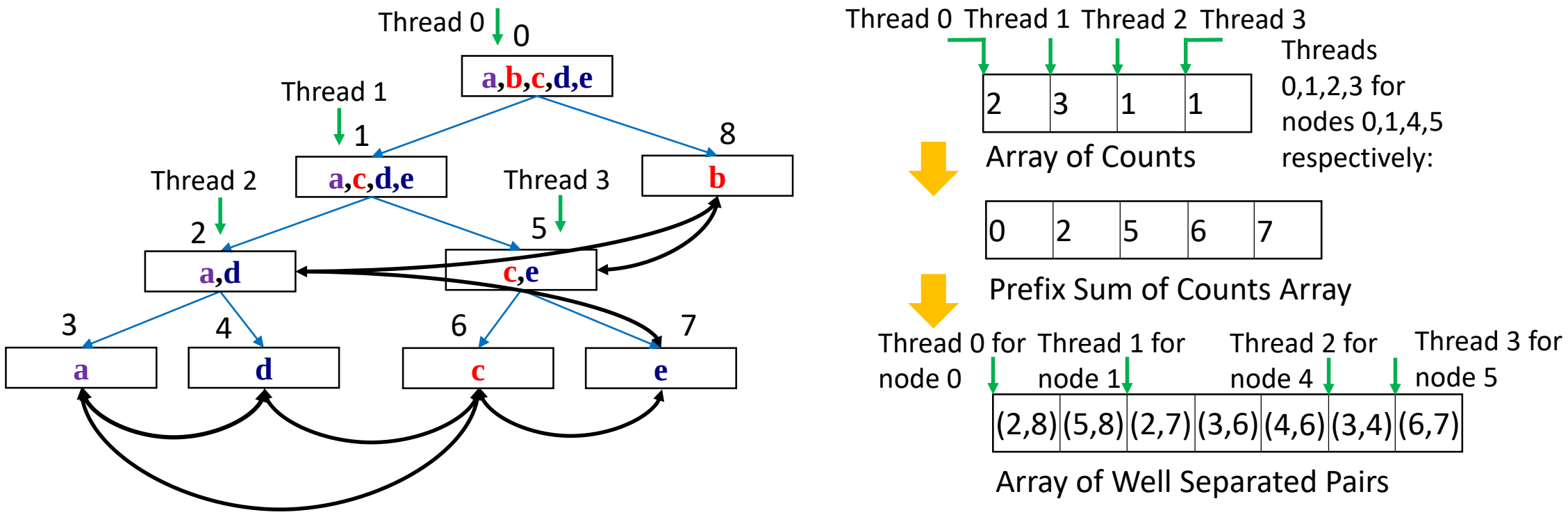

**Figure 6.9.** constructing WSPD in parallel for array from the split tree

## 6.9 More Algorithmic Details:

Here we present the algorithmic and implementation details that are omitted in the main context of the paper.

### 6.9.1 WCD:

We implement the WCD using the Observation in [149] that $OT(A \cup \tilde{B}, B \cup \tilde{A}) \leq 2W_1(A, B)$, where $OT(A \cup \tilde{B}, B \cup \tilde{A})$ is the classical optimal transport distance between $A \cup \tilde{B}$ and $B \cup \tilde{A}$, the sum of distances of their optimal matching, is a 2 approximation to $W_1(A, B)$. Since $\text{WCD}(A \cup \tilde{B}, B \cup \tilde{A}) \leq \text{OT}(A \cup \tilde{B}, B \cup \tilde{A})$, we get $\frac{1}{2}$ WCD$\leq W_1(A, B)$.

FLOWTREE is faster than WCD on reddit due to the small scale of the PDs in that dataset. However asymptotically WCD is much faster on very large datasets since it can be implemented as a $O(\log n)$ depth sum-reduction of coordinates on GPU, similar to QUADTREE.

### 6.9.2 WSPD and Spanner Construction:

Here we present our simplified parallel algorithm for WSPD construction used in our implementation. The purpose of traversing the split tree twice is to parallelize writing out the WSPD, the bottleneck to constructing a WSPD. Although the WSPD is linear in $n$, the number of nodes of $\mathsf{WS}_s^{PD}(A^\delta, B^\delta)$, in practice the size of the WSPD is several orders of magnitude larger than $n$. Thus writing out the WSPD requires a large amount of data movement. Algorithm 24 first finds the number of pairs written out by a thread rooted at some node in the split tree. The computation of counts is in parallel and is mostly arithmetic. Once the counts are accumulated, a prefix sum of the counts is computed and written out to an offsets array. The offsets are then used as starting memory addresses to write out the WSPD pairs for each thread in parallel.

Figure 6.9 illustrates the parallel computation of the WSPD. The prefix sum is computed over the counts determined by each thread. There is a thread per internal node.

In our implementation, we do not actually keep track of the point subsets for each node of the split tree. Instead, we keep track of a single point in each point subset $P \subseteq \hat{A} \cup \hat{B}$ as well as a bounding box of $P$. This constructs the non-diagonal arcs of $\mathsf{WS}_s^{PD}(A, B)$ with minimal data.

**Algorithm 23:** Construct $s$-WSPD-biarcs in parallel

**Input:** $\boldsymbol{T}$: a split tree; $s$: WSPD parameter

**Output:** $s$-WSPD represented by wspd-ptn-pairs as an array

counts $\leftarrow \{0, \ldots, 0\}$ `// Allocate` $O(n)$ `elements`
**for** *node* $w \in \boldsymbol{T}$ *in parallel* **do**
  COUNT-WSPD(tid($w$), $w$.left, $w$.right, $s$, counts) `// tid(·) is injective`
**end**
offsets $\leftarrow$ PREFIXSUM(counts);
$L \leftarrow$ offsets[$w$] `// offsets[w] = sum(counts)`
wspd-ptn-pairs $\leftarrow \{\ldots\}$ `// Allocate` $L$ `elements for wspd-ptn-pairs:` $O(s^2 n)$ `memory`
**for** *node* $w \in \boldsymbol{T}$ *in parallel* **do**
  CONSTRUCT-WSPD(tid($w$), $w$.left, $w$.right, $s$, offsets, wspd-ptn-pairs)
**end**

**Algorithm 24:** Compute WSPD thread counts for offsets

**Input:** tid: thread id; nodes $u$ and $v$ in the split tree; $s$: WSPD parameter; counts: the number of recursive calls made by each thread

**Output:** counts: array of counts, counts[tid] = number of pairs each thread will find

**Function** `count-WSPD(`*tid, u, v, s, counts*`)`:
  **if** *$u$ is $s$-well separated from $v$* **then**
    counts[tid]++ `// Keep track of the number of well-separated pairs associated with tid`
    **return**
  **if** max_*len(BndingBx(u))* > max_*len(BndingBx(v))* **then**
    COUNT-WSPD(tid, $u$.left, $v$, $s$, counts);
    COUNT-WSPD(tid, $u$.right, $v$, $s$, counts);
  **else**
    COUNT-WSPD(tid, $u$, $v$.left, $s$, counts);
    COUNT-WSPD(tid, $u$, $v$.right, $s$, counts);

**Algorithm 25:** Write out WSPD from offsets

**Input:** tid: thread id; nodes $u$ and $v$ in the split tree; $s$: WSPD parameter; offsets: prefix sum of the counts; wspd: writable array of point pairs

**Output:** $s$-WSPD with representatives of point pairs as an array

**Function** `construct-WSPD`(*tid, u, v, s, wspd*):
    **if** *$u$ is $s$-well separated from $v$* **then**
        wspd[offsets[tid]++] $\leftarrow$ ($u$.point, $v$.point) `// All threads write in parallel`
        **return**
    **if** max_*len(BndingBx(u))* > max_*len(BndingBx(v))* **then**
        CONSTRUCT-WSPD(tid, $u$.left, $v$, $s$, wspd);
        CONSTRUCT-WSPD(tid, $u$.right, $v$, $s$, wspd);
    **else**
        CONSTRUCT-WSPD(tid, $u$, $v$.left, $s$, wspd);
        CONSTRUCT-WSPD(tid, $u$, $v$.right, $s$, wspd);

Algorithm 26 shows how to write out the diagonal arcs for $\mathsf{WS}_s^{PD}(A^\delta, B^\delta)$. On line 2 it states that there is a parallelization by prefix sum on arc counts. This computation is similar to the algorithm for WSPD construction. The number of arcs per point is kept track of. A prefix sum is computed after this and the diagonal arcs are written out per point.

**Algorithm 26:** Form diagonal arcs

**Input:** $\boldsymbol{P} = \hat{A}_\delta \cup \hat{B}_\delta$

**Output:** diag-arcs

1 **foreach** *point $p \in \boldsymbol{P}$ parallelized by prefix sum on count of arcs incident on each $p$* **do**
2     **if** *$p$ is from $\hat{A}_\delta$* **then**
3         diag-arcs $\leftarrow$ diag-arcs $\cup\{(p, p_{proj})\}$;
4     **if** *$p$ is from $\hat{B}_\delta$* **then**
5         diag-arcs $\leftarrow$ diag-arcs $\cup\{(p_{proj}, p)\}$;
6 **end**

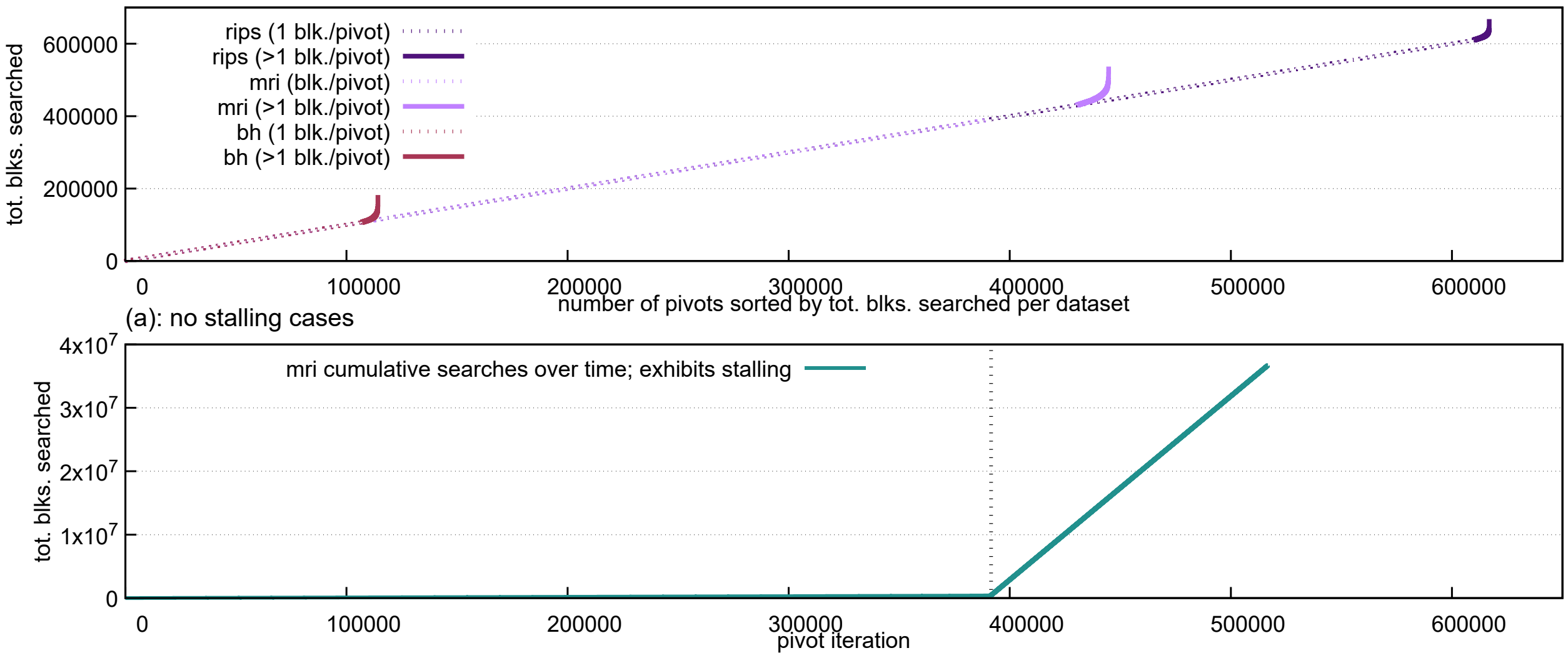

**Figure 6.10.** (a) Plot of no stalling case of the cumulative distribution of blocks searched for rips, mri and brain-heart datasets. (b) Plot of a stalling case for the mri dataset. block size= $\sqrt{m}$

### 6.9.3 Representing the Transshipment Network:

The data structure used to represent the transshipment network significantly affects the performance of network simplex algorithm. Since most of the time of computation is spent on the network simplex algorithm and not the network construction stage, the network data structure is designed to be constructed to be as efficient for arc reading and updating as possible. A so-called static graph representation [178], essentially a compressed sparse row (CSR) [202] format matrix, is used to represent the transshipment network. Thus in order to build a CSR matrix, we must sort the arcs $(u, v)$ first by first node followed by second node in case of ties. This sorting can be over several millions of arcs, see Table 6.2 column 2. For example, for rips at $\epsilon' \leq 0.2$, 468 million arcs must be sorted. ($\epsilon'$ is the guaranteed relative error bound). For a sequential $O(m \log m)$ algorithm, this would form a bottleneck to the entire algorithm before network simplex, making the algorithm $\Omega(m \log(m))$. Thus we sort the arcs on GPU using the standard parallel merge sorting algorithm [162, 203] and achieving a parallel depth complexity of $O(\log m)$.

## 6.10 Computational Behavior of Network Simplex (BSP in practice):

Refer to Section 6.7 and Table 6.5 for dataset information. Figure 6.10 shows two very different computational patterns of the block search pivot based NtSmplx algorithm. Figure 6.10(a) shows the vast majority of cases when there is no stalling. We show the cumulative distribution of blocks searched for the rips, mri and brain-heart datasets at s=20, 49 and 150 respectively. The block sizes are set to the square root of the number of arcs; the block sizes are 6539, 9134 and 8922 respectively. Notice that 98.9%, 96.8% and 93.8% of the pivots involve only a single block being searched, and account for 91.4%, 80.2% and 58.8% of the total blocks searched. Although the pivots are sorted per dataset by the number of blocks searched, the cumulative distribution depending on the pivots computed over execution is almost identical. Figure 6.10(b) shows the relatively rare but severe case of stalling for the mri dataset at s=36, stopped after 10 minutes. Stalling begins at the 391559th arc found.

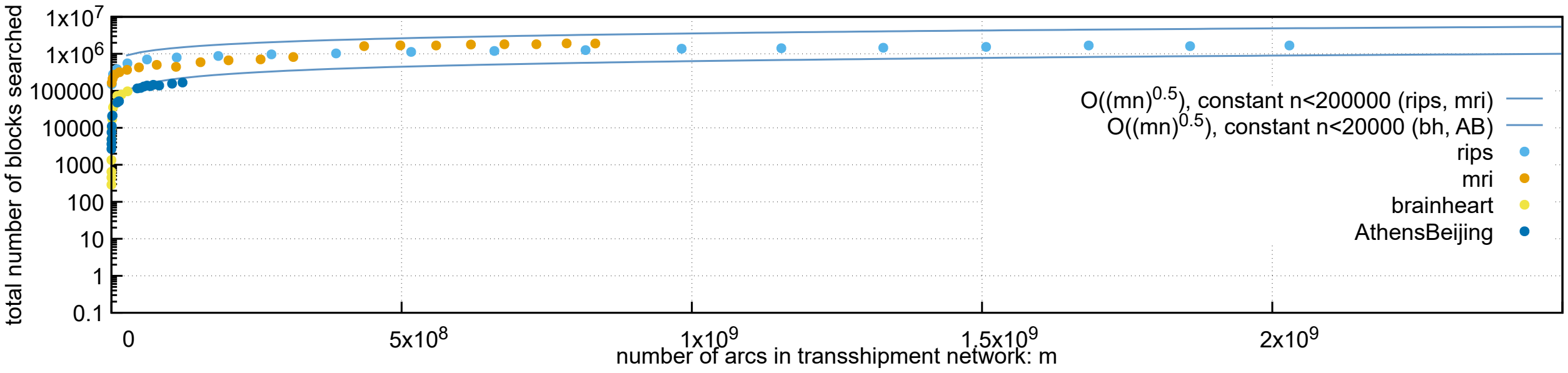

**Figure 6.11.** Plot of the total number of block pivot searches depending on the number of arcs.

Furthermore, we have noticed empirically that repeated tie breaking of reduced costs during pivot searching results in a tendency to stall. In fact, most implementations simply repeatedly choose the smallest indexed arc for tie breaking. After applying lattice snapping by $\pi_\delta$, symmetry is introduced into the pairwise relationships and thus results in many equivalent costs on arcs and subsequent reduced costs. This is why we introduce a small perturbation to the snapped points in order to break this symmetry. This results in much less stalling in practice.

### 6.10.1 Parallelizing Network Simplex Algorithm:

Network simplex is a core algorithm used for many computations, especially exact optimal transport. This introduces a natural question: can we directly parallelize some known network simplex pivot search strategies and gain a performance improvement? We attempted to implement parallel pivot search strategies such as a $O(\log m)$-depth parallel min reduction over all reduced costs on either GPU or multicore, such as in [204]. These approaches did not improve performance over a sequential block pivot search strategy. There was speedup over Dantzig's pivot strategy, where all admissible arcs are checked, however. For GPU, there is an issue of device to host and host to device memory copy. These IO operations dominate the pivot searching phase and are several of orders of magnitude slower than a single block searched sequentially from our experiments. Recall that most searches result in a single block by Figure 6.10. For multicore, there is an issue of thread scheduling which provides too much overhead. In general, it is very difficult to surpass the performance of a sequential search over a single block when the block size can fit in the lower level cache due to the two aforementioned issues. For example, in our experiments the cache size is 28160KB, which should hold $6 \cdot B \cdot 8$ bytes for $B = \sqrt{m}$, the block size, and $m < 3 \times 10^{11}$ where 6 denotes the 6 arrays needed to be accessed to compute the reduced cost and 8 is the number of bytes in a double. This bound on $m$, the number of arcs, should hold for almost all pairs of conceivable input persistence diagrams and $s > 0$ in practice. This does not preclude, however the possibility of efficient parallel pivoting strategies completely since stalling still exists for the sequential block search algorithm.

### 6.10.2 Empirical Complexity:

For each of the datasets from Table 6.5, our experiments illustrated in Figure 6.12, show that for varying $s$ and fixed $n$, our overall approach runs empirically in $O(\sqrt{n}m)$ time, where $m = s^2 n$ with $s$ the WSPD parameter and $n$ the number of nodes in the sparsified transshipment network.

Here we explain in more detail the experiment illustrated in Figure 6.1. We determine the empirical complexity with respect to the number of points on a synthetic Gaussian dataset.

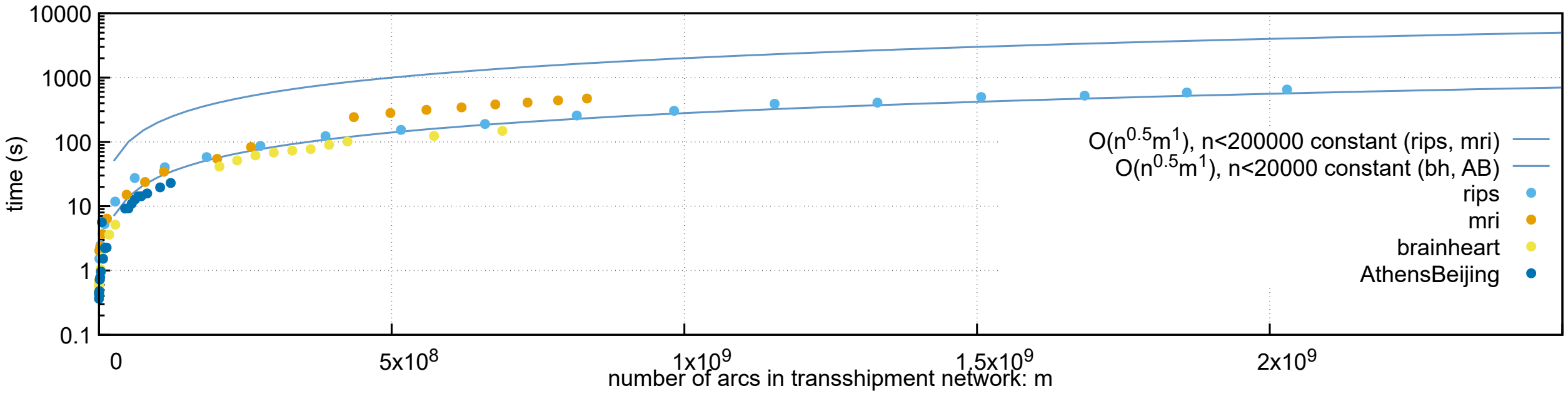

**Figure 6.12.** Plot of the empirical time (log scale) depending on the number of arcs of the sparsified transshipment network for each dataset. $n$ is the number of nodes.

These are not real persistence diagrams and are made up of points randomly distributed on the plane above the diagonal. The points follow a Gaussian distribution. For fixed $s \leq 40$, as a function of $n$ our algorithm empirically is upper bounded by $O(s^2 n^{1.5})$. This is determined through upper bounding the least squares curve fitting. Since we are still solving a linear program, it should not be expected that the empirical complexity can be truly linear, except perhaps under certain dataset conditions. The proximity of points, for certain real persistence diagrams, for example, could be exploited more by $\delta$-condensation. We notice that the empirical complexity is better, the smaller the $s$, including for $s \leq 40$. This is why in Section 6.7 PDoptFlow for $s = 1$ performs so much faster than PDoptFlow for $s = 18$.

### 6.10.3 Stopping Criterion:

Due to the rareness of stalling for given $s$ in practice, our stopping criterion is designed to justify the empirical time bound. If the block size is $\sqrt{m}$, the computation goes like $O(s^2 n^{1.5})$, and each iteration within a block search determines the time, $O(\frac{s^2 n^{1.5}}{s\sqrt{n}}) = O(\sqrt{mn})$ blocks is an upper bound on the number of searched blocks when there is no stalling. Figure 6.11 illustrates this relationship amongst $m$, $n$ and the time. Thus the stopping criterion is set to $C\sqrt{mn} + b$. In practice, $C$ may simply be set to 0 and $b$ set to a large number however it has been empirically found that the stopping citerion goes like $\sqrt{mn}$ blocks for a large number of the various types of real persistence diagrams such as those generated by the persistence algorithm on lower star filtrations induced by images and rips filtrations on random point clouds, to name the types from the experiments.

### 6.10.4 Bounds on Min-Cost Flow:

The $W_1$-distance between PDs is a special case of the unbalanced optimal transport (OT) problem as formulated in [130, 205]. Solving such a problem exactly using min-cost flow is known to take cubic complexity [206] in the number of points. However, affording cubic complexity is usually infeasible in practice and thus we seek a subcubic solution.

There are several approaches to approximating the distance between PDs with $n$ total points. In [149], a $\log \Delta$ approximation is developed, where, $\Delta$ is the aspect ratio, adapting the work of [150] and [151] for persistence diagrams. In [136], the auction matching algorithm performs a $(1+\epsilon)$ approximation, also lowering complexity by introducing geometry into the computation. Geometry lowers a linear search over $O(n)$ points for nearest neighbors to $O(\sqrt{n})$ via kd-tree. This does not lower the theoretical bound below $O(n^{2.5})$, however. Our approach lowers complexity by introducing a geometric spanner [157], using a linear number of arcs between points.

Min-cost flow algorithms can have theoretically very low complexity. The input to min-cost flow is a transshipment network and its output is the minimum cost flow value. Let $m$ be the number of arcs in the transshipment network and $n$ its number of nodes. It was shown that min-cost flow can be found exactly in $\tilde{O}(m+n^{1.5})$ complexity in [142], by network simplex in $\tilde{O}(n^2)$ complexity, in parallel in $\tilde{O}(\sqrt{m})$ and approximated on undirected graphs in [140] in $\tilde{O}(m^{1+o(1)})$ complexity.

Since the number of nodes and arcs of the transshipment network depend directly on the points and pairwise distances respectively, an implication of using a geometric spanner for $(1+O(\epsilon))$ approximation is that the complexity becomes theoretically subcubic and requiring $O_\epsilon(n)$ memory, where $O_\epsilon$ hides a polynomial dependency on $\epsilon$, the relative approximation error. In fact this bound is actually achieved in practice. We show the empirical complexity is actually similar to $O(s^2n^{1.5})$ as shown in Figure 6.12 and Figure 6.1 but only for small $s$.

# 7. GEFL: EXTENDED FILTRATION LEARNING FOR GRAPH CLASSIFICATION

Extended persistence is a technique from topological data analysis to obtain global multiscale topological information from a graph. This includes information about connected components and cycles that are captured by the so-called persistence barcodes. We introduce extended persistence into a supervised learning framework for graph classification. Global topological information, in the form of a barcode with four different types of bars and their explicit cycle representatives, is combined into the model by the readout function which is computed by extended persistence. The entire model is end-to-end differentiable. We use a link-cut tree data structure and parallelism to lower the complexity of computing extended persistence, obtaining a speedup of more than 60x over the state-of-the-art for extended persistence computation. This makes extended persistence feasible for machine learning. We show that, under certain conditions, extended persistence surpasses both the WL[1] graph isomorphism test and 0-dimensional barcodes in terms of expressivity because it adds more global (topological) information. In particular, arbitrarily long cycles can be represented, which is difficult for finite receptive field message passing graph neural networks. Furthermore, we show the effectiveness of our method on real world datasets compared to many existing recent graph representation learning methods.[1]

## 7.1 Introduction

Graph classification is an important task in machine learning. Applications range from classifying social networks to chemical compounds. These applications require global as well as local topological information of a graph to achieve high performance. Message passing graph neural networks (GNNs) are an effective and popular method to achieve this task.

These existing methods crucially lack quantifiable information about the relative prominence of cycles and connected component to make predictions. Extended persistence is an unsupervised technique from topological data analysis that provides this information through a generalization of hierarchical clustering on graphs. It obtains both 1- and 0-dimensional multiscale global homological information.

[1]↑ https://github.com/simonzhang00/GraphExtendedFiltrationLearning

Existing end-to-end filtration learning methods [207, 208] that use persistent homology do not compute extended persistence because of its high computational cost at scale. A general matrix reduction approach [209] has time complexity of $O((n+m)^\omega)$ for graphs with $n$ nodes and $m$ edges where $\omega$ is the exponent for matrix multiplication. We address this by improving upon the work of [210] and introducing a link-cut tree data structure and a parallelism for computation. This allows for $O(\log n)$ update and query operations on a spanning forest with $n$ nodes. We consider the expressiveness of our model in terms of extended persistence barcodes and the cycle representatives. We characterize the barcodes in terms of size, what they measure, and their expressivity in comparison to WL[1] [208]. We show that it is possible to find a filtration where one of its cycle's length can be measured as well as a filtration where the size of each connected component can be measured. We also consider the case of barcodes when no learning of the filtration occurs. We consider several simple examples where our model can perfectly distinguish two classes of graphs that no GNN with expressivity at most that of WL[1] (henceforth called WL[1] bounded GNN) can. Furthermore, we present a case where experimentally 0-dimensional standard persistence [208, 211], the only kind of persistence considered in learning persistence so far, are insufficient for graph classification.

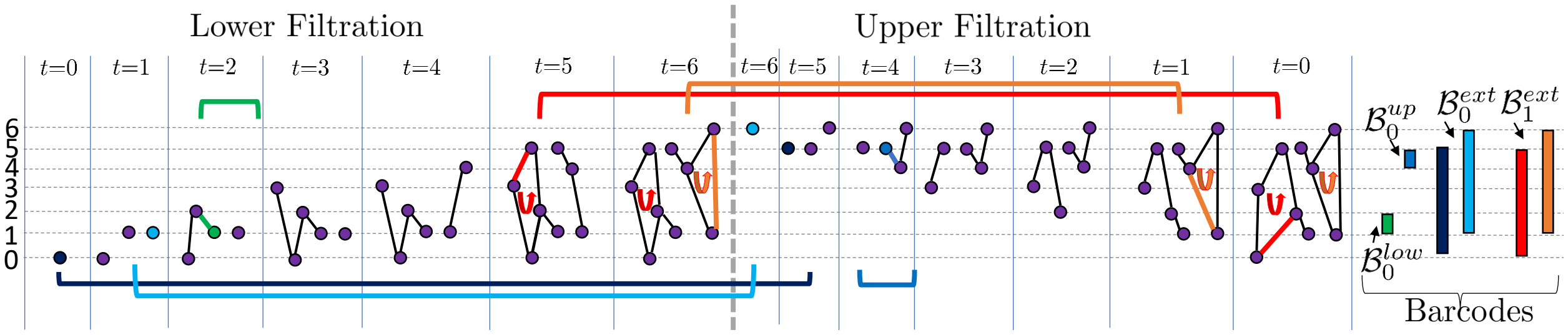

**Figure 7.1.** Lower and upper filtrations for extended persistence and the resulting barcode for a graph. The green bar comes from a pairing of a green edge with a vertex in the lower filtration. Similarily the blue bar in the upper filtration comes from a vertex-edge pairing in the upper filtration. The two dark blue bars count connected components and come from pairs of two vertices. The two red bars count cycles and come from pairs of edges. Both $\mathcal{B}_0^{ext}$ and $\mathcal{B}_1^{ext}$ bars cross from the lower filtration to the upper filtration. The multiset of bars forms the barcode. Cycle reps. are shown in both filtrations.

Our contributions are as follows:

1. We introduce extended persistence and its cycle representatives into the supervised learning framework in an end-to-end differentiable manner, for graph classification.

2. For a graph with $m$ edges and $n$ vertices, we introduce the link-cut tree data structure into the computation of extended persistence, resulting in an $O(m \log n)$ depth and $O(mn)$ work parallel algorithm, achieving more than 60x speedup over the state-of-the-art for extended persistence computation, making extended persistence amenable for machine learning tasks.

3. We analyze conditions and examples upon which extended persistence can surpass the WL[1] graph isomorphism test [212] and 0-dimensional standard persistence and characterize what extended persistence can measure from additional topological information.

4. We perform experiments to demonstrate the feasibility of our approach against standard baseline models and datasets as well as an ablation study on the readout function for a learned filtration.

## 7.2 Background

### 7.2.1 Computational Topology for Graphs

Let $G = (V, E)$ be a graph where $V$ is the set of vertices and $E \subseteq V \times V$ is the set of edges. Let $n = |V|$ and $m = |E|$ be the number of nodes and edges of $G$, respectively. Graphs in our case are undirected and simple, containing at most a single edge between any two vertices. We describe a way to order the nodes and edges of a graph.

Define a **filtration function** $F : G \to \mathbb{R}$ where $F$ has a value in $\mathbb{R}$ on each vertex and edge, denoted by $F(u)$ or $F(\mathrm{e})$ for $u \in V$ or $\mathrm{e} \in E$. A filtration function also has the property that either:

1. (Monotonicity) $F(\mathrm{e}) \geq F(u), F(v)$ when $u, v \in \mathrm{e}, \forall \mathrm{e} \in E$

2. Or (Reverse Monotonicity) $F(\mathrm{e}) \leq F(u), F(v)$ when $u, v \in \mathrm{e}, \forall \mathrm{e} \in E$

Only one of these properties can hold over all the edges.

Given the graph $G = (V, E)$, we can define the $\lambda$-sublevel graph as $G_\lambda = (V_\lambda, E_\lambda)$ w.r.t. $F$ and a $\lambda \in \mathbb{R}$ where $V_\lambda = \{v \in V : F(v) \leq \lambda\}$ and $E_\lambda = \{\mathrm{e} \in E : F(\mathrm{e}) \leq \lambda\}$. Sublevel graphs of $G$ are subgraphs of $G$. If we change $\lambda$ from $-\infty$ to $+\infty$ we obtain an increasing sequence of sublevel graphs $\{G_\lambda\}_{\lambda \in \mathbb{R}}$ which we call a sublevel set filtration. Such a filtration can always be converted into a sequence of subgraphs of $G$: $\varnothing = G_0 \subseteq G_1 \subseteq \ldots \subseteq G_{n+m} = G$ (See [126, Page 102]) s.t. $\sigma_\mathrm{i} = G_{\mathrm{i}+1} \backslash G_\mathrm{i}$ is a single edge or vertex and $F_\mathrm{i} := F(\sigma_\mathrm{i})$. The sequence of vertices and edges $\sigma_0, \sigma_1, \ldots, \sigma_{n+m-1}$ thus obtained is called the index filtration.

Define a vertex-induced lower filtration for a vertex function $f_G : V \to \mathbb{R}$ as an index filtration where a vertex $v$ has a value $F^{low}(v) := f_G(v)$ and any edge $(u, v)$ has the value $F^{low}(u, v) := \max(F^{low}(u), F^{low}(v))$ and $F^{low}_\mathrm{i} \leq F^{low}_{\mathrm{i}+1}$. Similarly define an upper filtration for $f_G$ as an index filtration where $F^{up}(v) := f_G(v)$ and the edge $(u, v)$ has value $F^{up}(u, v) := \min(f_G(u), f_G(v))$ and $F^{up}_\mathrm{i} \geq F^{up}_{\mathrm{i}+1}$.

**Proposition 7.2.2.** *The vertex-induced lower filtration $F^{low} : V \cup E \to \mathbb{R}$ is a filtration function.*

*Similarly, the vertex-induced upper filtration $F^{up} : V \cup E \to \mathbb{R}$ is also a filtration function.*

*Proof.* Certainly

$$F^{low}(\mathrm{e}) = \max(F^{low}(u), F^{low}(v)) \geq F^{low}(u), F^{low}(v) : \mathrm{e} = (u, v), \forall \mathrm{e} \in E \tag{7.1}$$

$$F^{up}(\mathrm{e}) = \min(F^{up}(u), F^{up}(v)) \leq F^{up}(u), F^{up}(v) : \mathrm{e} = (u, v), \forall \mathrm{e} \in E \tag{7.2}$$

proves both claims. □

**Persistent homology**(PH) tracks changes in homological features of a topological space as the sublevel set for a given function grows; see books [67, 126]. For graphs, these features are given by evolution of components and cycles over the intervals determined by pairs of vertices and edges. A vertex $v_\mathrm{i} \in G_{\mathrm{i}+1} \setminus G_\mathrm{i}$ begins a connected component (CC) signalling a birth at filtration value $F^{low}(v_\mathrm{i})$ in zeroth homology group $H_0(G)$ where $G$ is a simplicial complex with $\mathbb{Z}_2$ coefficients. An edge $\mathrm{e}_\mathrm{j} \in G_{\mathrm{j}+1} \setminus G_\mathrm{j}$ may join two components signalling a death of a class in $H_0(G)$ at filtration value $F^{low}(\mathrm{e_j})$, or it may create a cycle signalling a birth in the 1st homology group $H_1(G)$ at filtration value $F^{low}(\mathrm{e_j})$. When a death occurs in $H_0(G)$ by an edge $\mathrm{e_j}$, the youngest of the two components being merged is said to die giving a birth-death pair $(b,d) \triangleq (F^{low}(v_\mathrm{i}), F^{low}(\mathrm{e_j}))$ if the dying component was created by vertex $v_\mathrm{i}$. For cycles, there is no death and thus they have death at $\infty$.

The multiset of birth death pairs $\mathcal{B}(F^{low}) = \{\{(b,d)\}\}$ given by the persistent homology is called the barcode. Each pair $(b,d)$ provides a closed-open interval $[b,d)$, which is called a bar. The **persistence** of each bar $[b,d)$ in a barcode is defined as $|d-b|$. Notice that, both in 0- and 1-dimensional persistence, some bars may have infinite persistence since some components ($H_0(G)$ features) and cycles ($H_1(G)$ features) never die, equivalently, have death at $\infty$.  For any filtration function $F : G \to \mathbb{R}$ there is a corresponding spanning tree that is being constructed through the sequence of sublevel graphs $\{G_\lambda\}_{\lambda\in\mathbb{R}}$ of the graph $G$.

We summarize the above about the nodes and edges used to construct $G$ for the creation and destruction of homology classes as follows:

1. Each new node creates a new connected component.

2. When an edge is added to maintain a spanning tree of $G_\lambda$, a destruction of a connected component class in $H_0(G)$ occurs.

3. Otherwise, the new edge is a complementary edge to an existing spanning tree on $G_\lambda$, see Definition 2.8.17. This contributes a generator to $H_1(G)$.

**Extended persistence($\mathbf{PH_{ext}}$):**

Define a coned space of $G = (V, E)$ by $G$ with each edge and vertex of $G$ extended by a virtual node $\alpha \notin V$:

$$\text{Cone}(G) \triangleq V \cup E \cup \{(\alpha, u, v)\}_{(u,v)\in E} \cup \{(\alpha, u)\}_{u\in V} \tag{7.3}$$

The extra edges and triangles attached to $G$ form cones of the vertices and edges of $G$ respectively. Specifically, the cone of a vertex $u$ is given by the edge $(\alpha, u)$ and the cone of an edge $(u, v)$ is given by the triangle $(\alpha, u, v)$.

We describe how to extend the original lower filtration $F^{low}$ to form the extended upper filtration $F^{\text{ext},up}$. Let $\alpha \notin V$ be the additional vertex for the graph $G$. Define an extended function $f_{G\cup\{\alpha\}}$ whose value is equal to $f_G$ on all vertices except $\alpha$ on which it has a value larger than any other vertices. The upper filtration $F^{\text{ext},up}$ defined on Cone$(G)$ is induced by the min operation and $f_{G\cup\{\alpha\}}$ on the edges and triangles formed by the cones of the vertices and edges of $G$.

This means:

1. $F^{\text{ext},up}(v \cup \{\alpha\}) \coloneqq f_{G\cup\{\alpha\}}(v)$
2. $F^{\text{ext},up}((u, v) \cup \{\alpha\}) \coloneqq \min(f_{G\cup\{\alpha\}}(u) f_{G\cup\{\alpha\}}(v))$
3. and $F^{\text{ext},up}(\{\alpha\}) \coloneqq (\max_{v\in V} f_G(v) + 1)$.

Extended persistence takes an extended filtration $F_{f_G}$ as input, which is obtained by concatenating the lower filtration $F^{low}$ of the graph $G$ and an upper filtration $F^{\text{ext},up}$ of the coned space of $G$ induced by a vertex filtration function $f_G$. This is well-defined since $F^{\text{ext},up}$ is defined on the coned space Cone$(G)$, which has the property that Cone$(G) \supseteq G$. We denote this concatenation by:

$$F_{f_G} \triangleq [F^{low} \| F^{\text{ext},up}] \tag{7.4}$$

Since the cone covers every vertex and edge of $G$, it must fill in all generators of $H_1(G)$ where $G$ is viewed as a simplicial complex with coefficients in $\mathbb{Z}_2$. As a result, in extended persistence all 0- and 1-dimensional features die (bars are finite; see [209] for details).

Four different persistence pairings or bars result from $\mathbf{PH_{ext}}$. The barcode $\mathcal{B}_0^{low}(F_{f_G})$ results from the vertex-edge pairs within the lower filtration, the barcode $\mathcal{B}_0^{up}(F_{f_G})$ results from the vertex-edge pairs within the upper filtration, the barcode $\mathcal{B}_0^{\text{ext}}(F_{f_G})$ results from the vertex-vertex pairs that represent the persistence of connected components born in the lower filtration and die in the upper filtration, and the barcode $\mathcal{B}_1^{\text{ext}}(F_{f_G})$ results from edge-edge pairs that represent the persistence of cycles that are born in the lower filtration and die in the upper filtration. The barcodes $\mathcal{B}_0^{low}(F_{f_G})$, $\mathcal{B}_0^{up}(F_{f_G})$, and $\mathcal{B}_0^{\text{ext}}(F_{f_G})$ represent persistence in the 0th homology $H_0(G)$ where $G$ is viewed as a simplicial complex with coefficients in $\mathbb{Z}_2$. The barcode $\mathcal{B}_1^{\text{ext}}(F_{f_G})$ represents persistence in the 1st homology $H_1(G)$ where $G$ is viewed as a simplicial complex with coefficients in $\mathbb{Z}_2$. We also use the shorthand notation $\mathcal{B}_0^{low}$, $\mathcal{B}_0^{up}$, $\mathcal{B}_0^{\text{ext}}$, and $\mathcal{B}_1^{\text{ext}}$ to denote the barcodes when the dependency on $F_{f_G}$ is implicit. In the TDA literature, $\mathcal{B}_0^{low}(F_{f_G})$, $\mathcal{B}_0^{up}(F_{f_G})$, $\mathcal{B}_0^{\text{ext}}(F_{f_G})$, and $\mathcal{B}_1^{\text{ext}}(F_{f_G})$ also go by the names of $Ord_0, Rel_1, Ext_0, Ext_1$ respectively.

See Figure 7.1 for an illustration of the filtration and barcode one obtains for a simple graph with vertices taking on values from 0...6 denoted by the variable $t$. In particular, at each $t$, we have the filtration subgraph $G_t$ of all vertices and edges of filtration function value less than or equal to $t$. Each line indicates the values 0...6 from the bottom to top. Repetition in the bar endpoints across all bars which appear on the right of Figure 7.1 is highly likely in general due to the fact that there are only $O(n)$ filtration values but $O(m)$ possible bars.

### 7.2.3 Message Passing Graph Neural Networks (MPGNN)

A message passing GNN (MPGNN) convolutional layer takes a vertex embedding $\mathbf{h}_u$ in a finite dimensional Euclidean space and an adjacency matrix $A_G$ as input and outputs a vertex embedding $\mathbf{h}'_u$ for some $u \in V$. The $k$th layer is defined generally as

$$\mathbf{h}_u^{k+1} \leftarrow \text{AGG}(\{\text{MSG}(\mathbf{h}_v^k) \| v \in N_{A_G}(u)\}, \mathbf{h}_u^k), u \in V$$

where $N_{A_G}(u)$ is the neighborhood of $u$. The functions MSG and AGG have different implementations and depend on the type of GNN.

Since there should not be a canonical ordering to the nodes of a GNN in graph classification, a GNN for graph classification should be permutation invariant with respect to node indices. To achieve permutation invariance [213], as well as achieve a global view of the graph, there must exist a readout function or pooling layer in a GNN. The readout function is crucial to achieving power for graph classification. With a sufficiently powerful readout function, a simple 1-layer MPGNN with $O(\Delta)$ number of attributes [214] can compute any Turing computable function, $\Delta$ being the max degree of the graph. Examples of simple readout functions include aggregating the node embeddings, or taking the element-wise maximum of node embeddings [215]. See Section 7.3 for various message passing GNNs and readout functions from the literature.

## 7.3 Related Work

Graph Neural Networks (GNN)s have achieved state of the art performance on graph classification tasks in recent years. For a comprehensive introduction to GNNs, see the survey [216]. In terms of the Weisfeler Lehman (WL) hierarchy, there has been much success and efficiency in GNNs [215, 217, 218] bounded by the WL[1] [219] graph isomorphism test. In recent years, the WL[1] bound has been broken by heterogenous message passing [220], high order GNNs [221], and put into the framework of cellular message passing networks [222]. Furthermore, a sampling based pooling layer is designed in [223]. It has no theoretical guarantees and its code is not publicly available for comparison. Other readout functions include [224–226]. For a full survey on global pooling, see [227].

Topological Data Analysis (TDA) based methods [208, 211, 228–232] that use learning with persistent homology have achieved favorable performance with many conventional GNNs in recent years. All existing methods have been based on 0-dimensional standard persistent homology on separated lower and upper filtrations [211]. We sidestep these known limitations by introducing extended persistence into supervised learning while keeping computation efficient.

A TDA inspired cycle representation learning method in [233] learns the task of knowledge graph completion. It keeps track of cycle bases from shortest path trees and has a $O(|V| \cdot$ $|E| \cdot k)$, $k$ a constant, computational complexity per graph. This high computational cost is addressed in our method by a more efficient algorithm for keeping track of a cycle basis. Furthermore, since the space of cycle bases induced by spanning forests is a strict subset of the space of all possible cycle bases, the extended persistence algorithm can find a cycle basis that the method in [233] cannot.

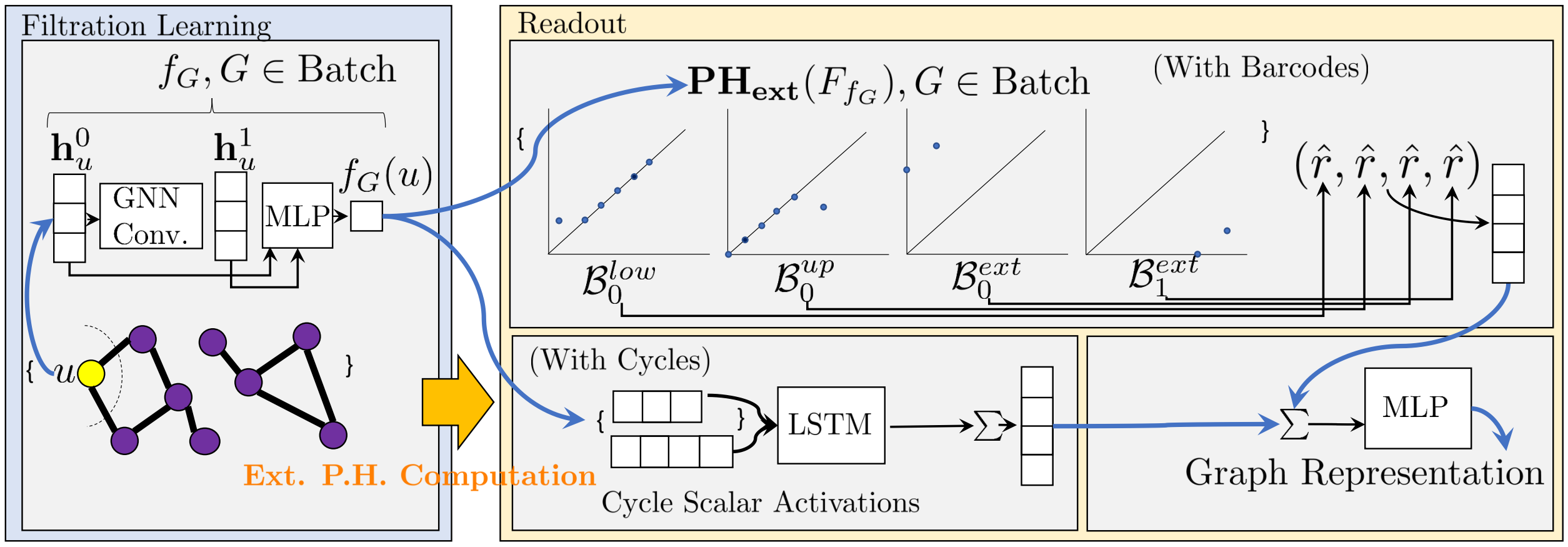

**Figure 7.2.** The extended persistence architecture (bars+cycles) for graph representation learning. The negative log likelihood (NLL) loss is used for supervised classification. The yellow arrow denotes extended persistence computation, which can compute both barcodes and cycle representatives.

On the computational side, fast methods to compute higher dimensional PH using GPUs, a necessity for modern deep learning, have been introduced in [2]. In [231, 234] neural networks have been shown to successfully approximate the persistence diagrams with learning based approach. However, differentiability and parallel extended persistence computation has not been implemented. Given the expected future use of extended persistence in graph data, a parallel differentiable extended persistence algorithm is an advance on its own.

## 7.4 What Kind of Graph Data is Best Described by Persistent Homology?

For any simple graph $G = (V, E)$ we would like to continuously embed it into Euclidean space. We are given a one dimensional ordering of vertices from the reals. Thus, we are looking for a continuous injection from $\mathbb{R}$ to some Euclidean space that also prevents the edges between the Euclidean embeddings from crossing.

A natural map to consider is a space-filling curve such as the Peano curve [235]. This would provide a continuous surjection from a compact interval of vertex values to a hypercube. The space filling curve, however, is not injective according to Petto's theorem [236].

In order to obtain injectivity, we need the embedding to not have edge crossings. For a one dimensional ordering on the vertices, we need at least 3 dimensions. Certainly in 1 dimension, a complete graph on 3 nodes already has edge crossings. In 2 dimensions, Kurtowski's theorem [237] prevents all possible graphs from having a graph embedding.

However, in 3 dimensions there is a canonical embedding for any simple graph $G = (V, E)$ [238]. This embedding is called the moment curve [239, 240] embedding $\rho_V : V \to \mathbb{R}^3$ satisfying

$$\rho_V(v) \triangleq [t_V(v), t_V^2(v), t_V^3(v)] \tag{7.5}$$

where $t_V : V \to \mathbb{R}$ maps the vertices to a total order on the real line. Since the image of $t_V$ is a total order, $t_V$ must be injective. In fact, the ambient dimension 3 can be raised to any $d$ by concatentating with higher degree monomials of $t_V$. The map $t_V$ is arbitrary and defines $\rho_v$. We call the map $t_V$ a **real-valued vertex map**.

A standard result about the moment curve $\gamma = \{(t, t^2, t^3) : t \in \mathbb{R}\}$ is that no four points can form a plane:

**Proposition 7.4.1.** *([240]) No plane intersects the moment curve $\gamma$ in $\mathbb{R}^3$ in more than 3 points. Consequently, every set of 4 distinct points on $\gamma$ cannot span a plane. Moreover, if $\gamma$ intersects a plane $h$ at 3 distinct points, then it crosses $h$ from one side to the other at each intersection*

*Proof.* See [240]. □

We extend $\rho_V : V \to \mathbb{R}^3$ to the edges of $G$ by embedding $\{u, v\} \in E$ to the straight line segment $\{t\rho_V(u) + (1-t)\rho_V(v) : t \in [0,1]\} \subseteq \mathbb{R}^3$. We can denote this by $\rho_E : E \to \mathbb{R}^3$.

By Proposition 7.4.1, we can deduce that there can be no edge crossings for any pair of edges in the image of $\rho_E$ for any $t_V$. Thus, we call $\rho_E$ a moment curve edge embedding.

We can extend these embeddings to any graph. We denote this by the moment curve graph embedding $\rho : \mathrm{ob}(\mathbf{Graph}) \to \mathrm{ob}(\mathbf{Simp})$. Given any graph $G = (V, E)$, we can then define $\rho(G)$ as a 1-dimensional simplicial complex

$$\rho(G) \triangleq \rho_V(V) \cup \rho_E(E) \tag{7.6}$$

The simplicial complex $\rho(G)$ is embeddable in $\mathbb{R}^3$. For definition of simplicial complex see Definition 2.5.27.

As a simplicial complex, the $p$-chain vector space $C_p(\rho(G))$ for $p = 0, 1$ is defined through formal sums of vertices and edges, respectively. Any simplicial complex has a boundary map, we define the boundary map on $\rho(G)$ to respect the ordering determined by $t_V$.

Let this boundary map $\partial_1(\bullet)$ on the 1-dimensional simplicial complex $\rho(G) \subseteq \mathbb{R}^3$ be as defined in Definition 2.6.7 where the order determined by $t_V$ is respected:

$$\partial_1^{\rho(G)}(\mathrm{e}) = -u + v, t_V(u) < t_V(v), \forall \mathrm{e} = \{u, v\} \in E \tag{7.7a}$$

$$\partial_0^{\rho(G)}(v) = \overrightarrow{0}, \forall v \in V \tag{7.7b}$$

### 7.4.2 Symmetries of a Graph with respect to the Homology Group of its Embedding in Euclidean Space

We define here the symmetries on any simplicial complex data $X \in \mathrm{ob}(\mathbf{Simp})$ with respect to the set of transformations that preserve the homology group determined by a boundary map $\partial_p : C_p(X) \to C_{p-1}(X)$ between $p$ and $p-1$ chains of $X$. This is defined as follows:

$$\mathrm{Sym}_{H_p}(X) \triangleq \{g : X \to \mathbb{R}^3 \text{ continuous} : H_p(\partial_p^X) \cong H_p(\partial_p^{g(X)})\} \tag{7.8}$$

where $\cong$ denotes a group isomorphism.

**Proposition 7.4.3.** *$Sym_{H_p}(X)$ contains noninvertible continuous transformations.*

*Thus it does not form a group under function composition.*

*Proof.* Let $X$ be a triangle:

$$X = \{\{u\},\{v\},\{w\},\{u,v\},\{v,w\},\{u,w\},\{u,v,w\}\}, u,v,w \in \mathbb{R}^3 \tag{7.9}$$

Consider the transformation $g : X \to \mathbf{0}$. This transformation collapses the triangle to a single point and still preserves that $\dim(H_0(\partial_0(X))) = 1$ and $\dim(H_p(\partial_p^X) = 0, p \geq 1$.

This map $g$ is noninvertible. □

A more natural set of symmetries for the moment curve graph embedding $\rho : \text{ob}(\mathbf{Graph}) \to \text{ob}(\mathbf{Simp})$ can be defined through graph isomorphism. For any $G = (V, E)$ and $\rho$ a moment curve graph embedding, let the set of continuous maps from $\mathbb{R}^3$ to itself that can preserve the original graph up to isomorphism be defined by:

$$\begin{aligned} \text{Homeo}(\rho) \triangleq \{&g : \mathbb{R}^3 \to \mathbb{R}^3 \text{ is a continuous map} \\ &\text{and for any } G, H \in \text{ob}(\mathbf{Graph}) \text{ the diagram 7.10a commutes}\} \end{aligned}$$

$$\begin{array}{ccc} G & \overset{\cong}{\longleftrightarrow} & H \\ \downarrow{\scriptstyle\rho} & & {\scriptstyle\rho}\downarrow \\ \rho(G) & \underset{g}{\longrightarrow} & \rho(H) \end{array} \tag{7.10a}$$

where $\rho : \text{ob}(\mathbf{Graph}) \to \text{ob}(\mathbf{Simp})$ is a moment curve graph embedding and $\cong$ denotes that graphs $G$ and $H$ are isomorphic.

In our definition we require inducing a graph isomorphism in the commuting diagram 7.10a. This is required in order to respect the fact that $\rho$ is defined on graphs.

**Proposition 7.4.4.** *For $\rho : ob(\boldsymbol{Graph}) \to ob(\boldsymbol{Simp})$ a moment curve graph embedding.*

*All continuous maps in Homeo$(\rho)$ are homeomorphisms on $\rho(G), \forall G \in ob(\boldsymbol{Graph})$.*

*Homeo$(\rho)$ also forms a group under function composition.*

*Proof.* **1. All maps $g \in$ Homeo$(\rho)$ are homeomorphisms on $\rho(G), \forall G \in$ ob(Graph):**

According to Definition 2.6.4, a continuous map $g : \mathbb{R}^3 \to \mathbb{R}^3$ is a homeomorphism on $\rho(G)$ iff it is injective and $g^{-1}$ is also continuous.

**a. $g \in$ Homeo$(\rho)$ is injective:**

According to the commuting diagram:

If $g \in \text{Homeo}(\rho)$, then we must have $G \cong H$. By definition of the graph isomorphism $\cong$, we have that there is a bijective map $\phi : V(G) \to V(H)$ with $\{\phi(u), \phi(v)\} \in E(H)$ iff $\{u, v\} \in E(G)$.

This allows us to extend the bijective homomorphism $\phi$ on vertices to all of $G = (V, E)$ as:

$$\phi(G) = (\phi(V), \{\{\phi(u), \phi(v)\}\}_{\{u,v\} \in E}) = H \tag{7.11}$$

The commuting diagram states that:

$$\rho(\phi(G)) = g(\rho(G)) \tag{7.12}$$

Since both $\rho$ and $\phi$ are injective, their composition is also injective. Thus $g$ on the right hand side must also be injective.

**b.** $g \in$ **Homeo**$(\rho)$ **has** $g^{-1}$ **continuous:**

Let $N(v) \triangleq \{v\} \cup \{\{u, v\} : \{u, v\} \in E\}$ be the open neighborhood of a vertex $v \in V$.

Let $\rho(N(v)) = \{\rho_V(v)\} \cup \{\lambda\rho_V(u) + (1-\lambda)\rho_V(v) : \lambda \in [0,1]\} : \{u, v\} \in E\}$ be the embedding of $N(v)$ through $\rho$ as defined in Definition 7.6.

We show that $g \in \text{Homeo}(\rho)$ is an open map on $\rho(G)$. To do so, it suffices to show that it takes the open embedded neighborhoods $\rho(N(v)) \subseteq \rho(G)$ to embedded neighborhoods $\rho(N(\phi(v))) \subseteq \rho(\phi(G))$. This is because the open sets of an embedded 1-dimensional simplicial complex $\rho(G) \subseteq \mathbb{R}^3$ are determined by the union of open neighborhoods $\rho(N(v)), \forall v \in V$.

This follows since $\phi$ is edge-preserving and $g(\rho(v)) = \rho(\phi(v)), \forall v \in V$.

Composing $g$ on $\rho(N(v))$:

$$g(\rho(N(v))) = \{g(\rho_V(v))\} \cup \{\lambda g(\rho_V(u)) + (1-\lambda)g(\rho_V(v)) : \lambda \in [0,1]\} : \{u, v\} \in E\} \tag{7.13a}$$

$$= \{\rho_V(\phi(v))\} \cup \{\lambda\rho_V(\phi(u)) + (1-\lambda)\rho_V(\phi(v)) : \lambda \in [0,1]\} : \{u, v\} \in E\} \tag{7.13b}$$

$$= \{\rho_V(\phi(v))\} \cup \{\lambda\rho_V(\phi(u)) + (1-\lambda)\rho_V(\phi(v)) : \lambda \in [0,1]\} : \{\phi(u), \phi(v)\} \in E(H)\} \tag{7.13c}$$

$$= \rho(N(\phi(v))) \tag{7.13d}$$

Thus $g$ is an open map.

Since $g \in \text{Homeo}(\rho)$ was arbitrary, we have shown that all such $g$ are homeomorphisms.

**2. Homeo($\rho$) forms a group:**

Checking the four conditions for $\text{Homeo}(\rho)$ being a group under function composition:

- Associativity: This follows by function composition
- Identity element: $\text{id} \in \text{Homeo}(\rho)$ since it is induced by the identity graph automorphism on a graph $G$.
- Inverse element: Since every $g \in \text{Homeo}(\rho)$ is a homeomorphism, we have that every $g \in \text{Homeo}(\rho)$ has $g^{-1} \in \text{Homeo}(\rho)$
- Closure: Two homeomorphisms $g_1, g_2 \in \text{Homeo}(\rho)$ have that $g_1 \circ g_2 \in \text{Homeo}(\rho)$ since both continuous and open maps are closed under composition.

□

We claim that these homeomorphisms are a subgroup of the symmetries of the 1-dimensional simplicial complex $\rho(G)$ with respect to the homology group $H_p(\partial_p^{\rho(G)})$ on $\rho(G) \subseteq \mathbb{R}^3$.

**Theorem 7.4.5.** *For any $\rho : ob(\textbf{\textit{Graph}}) \to ob(\textbf{\textit{Simp}})$ a moment curve graph embedding and any $G \in ob(\textbf{\textit{Graph}})$ then*

$$Homeo(\rho) \subseteq Sym_{H_p}(\rho(G)), \forall p = 0, 1 \tag{7.14}$$

*as a subset.*

*Proof.* For any map $g \in \text{Homeo}(\rho)$ we must have some pair $G, H \in \text{ob}(\textbf{Graph})$ with $G \cong H$, where $\cong$ is a graph isomorphism in the category **Graph**. Let $G = (V_1, E_1), H = (V_2, E_2)$.

Since $G \cong H$, there is a bijective map $\phi : V_1 \to V_2$ with $(\phi(u), \phi(v)) \in E_2, \forall (u, v) \in E_1$.

**Checking $H_p(\partial_p^{\rho(G)}) \cong H_p(\partial_p^{\rho(H)})$:**

**1. For $p = 0$,** we check that $\dim(\frac{\ker(\partial_p^{\rho(G)})}{\text{im}(\partial_{p+1}^{\rho(G)})}) = \dim(\frac{\ker(\partial_p^{\rho(H)})}{\text{im}(\partial_{p+1}^{\rho(H)})})$, which will show that $H_p(\partial_p^{\rho(G)}) \cong H_p(\partial_p^{\rho(H)})$ by linearity of the isomorphism.

**The number of connected components don't change:**

We can check that $\phi(C) \triangleq \{\phi(v)\}_{v \in C}$ is still maximally connected iff $C \subseteq V_1$ is maximally connected:

The subset of vertices $C \subseteq V_1$ is a connected component iff for any $u, v \in C$, there is a sequence of edges $\mathrm{e}_1, ..., \mathrm{e}_k \in E_1$ with $\mathrm{e}_\mathrm{i} \subseteq E_1$ that connects $u$ to $v$, meaning $u \in \mathrm{e}_1, v \in \mathrm{e}_k$ and $\mathrm{e}_\mathrm{i} \cap \mathrm{e}_{\mathrm{i}+1} \neq \varnothing$ and there is no larger $C' \supseteq C$ containing $C$ with the same property.

Applying the isomorphism $\phi$ to the nodes in $C$, we see that the path connectivity property of $C$ is maintained in $H$ due to the edge preserving property of an isomorphism.

Maximality of $\phi(C)$ is achieved since if another node $x \in V_2$ could be added to the set $\phi(C)$, then there is a node $y \in V_2$ with $\{x, y\} \in E_2$. By the edge preserving property of $\phi$, there is the edge $\{u, v\} \in E_1$ iff $\{x, y\} \in E_2$. This violates the maximality of $C \subseteq V_1$. Thus $\phi(C)$ is maximal.

Let $\mathrm{CC}(G')$ be the collection of all connected components of an arbitrary graph $G'$. In particular, let $\mathrm{CC}(G) \subseteq 2^{V_1}, \mathrm{CC}(H) \subseteq 2^{V_2}$ denote the collection of all connected components of $G, H$ respectively.

We know that $\dim(\frac{\ker(\partial_0^{\rho(G)})}{\mathrm{im}(\partial_1^{\rho(G)})}) = |\mathrm{CC}(G)|$. Since any two nodes are considered equivalent by the existence of a path connecting them, of the $n$ linearly independent 0-chains, or nodes, only $|\mathrm{CC}(G)|$ are linearly independent by being disconnected by paths. Certainly this spans the image of $\partial_1$ since all nodes and edges are considered.

Thus,

$$\dim(\frac{\ker(\partial_0^{\rho(G)})}{\mathrm{im}(\partial_1^{\rho(G)})}) = |\mathrm{CC}(G)| = |\mathrm{CC}(H)| = \dim(\frac{\ker(\partial_0^{\rho(H))}}{\mathrm{im}(\partial_1^{\rho(H))}}) \tag{7.15}$$

**2. Similarly, for $p = 1$,** we check that $\dim(\ker(\partial_p^{\rho(G)})) = \dim(\ker(\partial_p^{\rho(H)})$. This shows that $H_p(\partial_p^{\rho(G)}) \cong H_p(\partial_p^{\rho(H)})$ since there are no 2-dimensional simplices in $\rho(\bullet)$ and by linearity of the isomorphism.

**a. The number of nodes don't change:**

Certainly $|V_1| = |V_2|$ since $\phi$ is a bijection on the nodes.

**b. The number of edges don't change:**

The number of edges is not changing due to graph isomorphism: $|E_1| = |E_2|$.

Since $\phi$ is a bijection any $\{x, y\} \in E_2$ has some set $\{u, v\} \subseteq V_1$ with $\phi(u) = x, \phi(v) = y$. By definition of $\phi$ as edge-preserving, $\{u, v\} \in E_1$ iff $\{\phi(u), \phi(v)\} = \{x, y\} \in E_2$.

**c. The number of connected components don't change:**

See proof from the $p = 0$ case.

For the $p = 1$ case:

For any graph $G'$:

$$\dim(\ker(\partial_1(\rho(G')))) = |E(G')| - \operatorname{rank}(\partial_1(\rho(G'))) \tag{7.16}$$

with $\operatorname{rank}(\partial_1(\rho(G'))) = |V(G')| - |\mathrm{CC}(G')|$, which can be proven by induction:

Base case:

$n$ nodes, 0 edges gives a rank of $n$.

Inductive step:

When an edge is added to the graph, it only drops the rank by 1 when it connects two connected components. Otherwise, the rank does not change.

**Equality of ranks:**

Since by 1. $|V_1| = |V_2|$, 2. $|E_1| = |E_2|$, and 3. $|\mathrm{CC}(G)| = |\mathrm{CC}(H)|$ by the isomorphism $\phi$, we must have:

$$\dim(\ker(\partial_p^{\rho(G)})) = \dim(\ker(\partial_p^{\rho(H)}) \tag{7.17}$$

Thus: $H_p(\partial_p^{\rho(G)}) \cong H_p(\partial_p^{\rho(H)})$ ☐

### 7.4.6 Symmetries with respect to a Filtration of a Graph

For a moment curve graph embedding $\rho : \mathrm{ob}(\mathbf{Graph}) \to \mathbb{R}^3$, for each $G = (V, E) \in \mathrm{ob}(\mathbf{Graph})$ there must exist a $t_V : V \to \mathbb{R}$ map to define it. We claim that for any map $t_V$ there is a filtration function $h(t_V) : V \cup E \to \mathbb{R}$. This $h(t_V)$ can be defined by the lower-star filtration. With the lower-star filtration function $h(t_V)$, we can then form a sublevel set filtration $F_G$ of graphs.

$$F_G \triangleq (\{G_{h(t_V)(u)}\}_{u \in V}, \{\mathrm{inc} : G_h \hookrightarrow G_{h'}\}_{h \leq h'}) \tag{7.18}$$

As a filtration, this set of graphs can be totally ordered by the subgraph relationship by appending nodes from $V$ in order of increasing $h(t_V)(\bullet)$.

In persistent homology, an ordering on the data determines the relationships amongst the homology groups of the ordered data. In the case of graph data $G = (V, E)$, we can let this ordering be determined by the map $t_V : V \to \mathbb{R}$. In particular, $F_G$ orders the graph as a filtration of subgraphs. Thus, we can view persistent homology as measuring the homology groups of a filtered 1-dimensional simplicial complex $\rho(G)$ embedded in $\mathbb{R}^3$ ordered by $t_V : V \to \mathbb{R}$.

We define a special subgroup of $\text{Homeo}(\rho)$ that preserves the filtration $F_G$ from the filtration function $h(t_V) : V \to \mathbb{R}$ where the real valued function $t_V : V \to \mathbb{R}$ is defined by $\rho$ via the following definition:

$$\text{FiltHomeo}(\rho, h) \triangleq \{g : \mathbb{R}^3 \to \mathbb{R}^3 \text{ is a continuous map and the diagram 7.10a commutes}\}$$

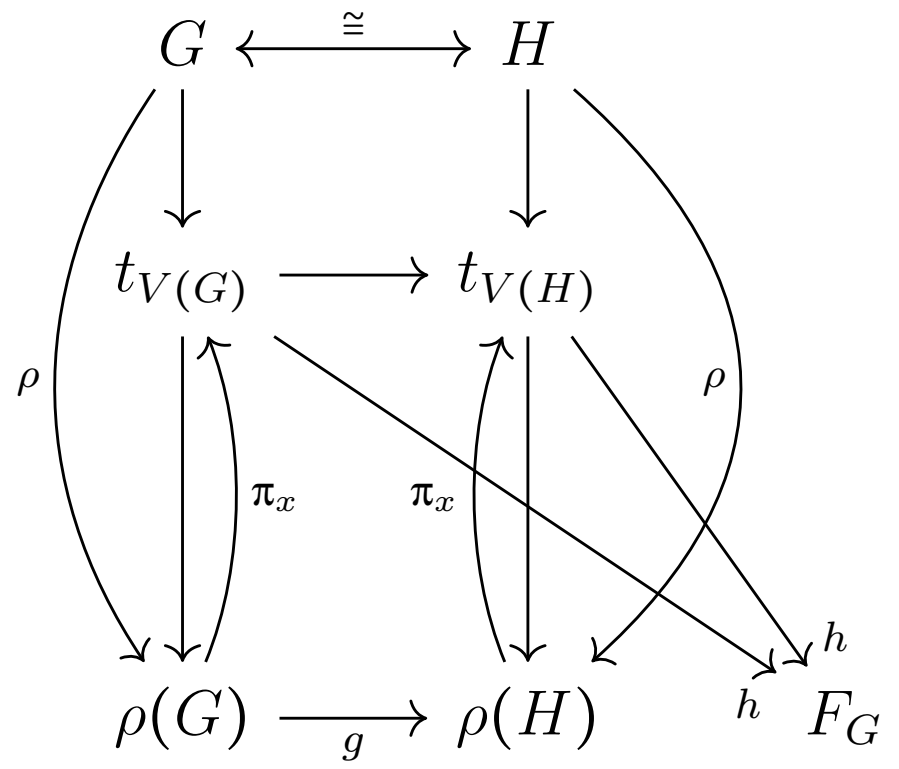

(7.19a)

where $\pi_x : \mathbb{R}^3 \to \mathbb{R}$ is the projection onto the first coordinate of a 3-tuple. These kinds of homeomorphisms do not change the filtration $F_G$. We call any homeomorphism belonging to the group $\text{FiltHomeo}(\rho, h)$ as a filtration preserving homeomorphism.

**Proposition 7.4.7.** *$\textit{FiltHomeo}(\rho, h) \subseteq \textit{Homeo}(\rho) \subseteq \textit{Sym}_{H_p}(\rho(G)), \forall p = 0, 1, \forall G \in ob(\textbf{Graph})$*

*Proof.* The subgroup relationship $\text{FiltHomeo}(\rho, h, G) \subseteq \text{Homeo}(\rho)$ is by definition and the remaining subset relationship is from Theorem 7.4.5.

□

## 7.5 Method

Our method as illustrated in Figure 7.2 introduces extended persistence as the readout function for graph classification. In our method, an upper and lower filtration, represented by a filtration function, coincides with a set of scalar vertex representations from standard message passing GNNs. This filtration function is thus learnable by MPGNN convolutional layers. Learning filtrations was originally introduced in [211] with standard persistence. As we show in Section 7.7 and Section 7.6 arbitrary cycle lengths are hard to distinguish by both standard GNN readout functions [241] as well as standard persistence due to the lack of explicitly tracking paths or cycles. Extended persistence, on the other hand, explicitly computes learned displacements on cycles of some cycle basis as determined by the filtration function as well as explicit cycle representatives.

We represent the map from graphs to learnable filtrations by any message passing GNN layer such as GIN, GCN or GraphSAGE followed by a multi layer perceptron (MLP) as a Jumping Knowledge (JK) [242] layer. The JK layer with concatenation is used since we want to preserve the higher frequencies from the earlier layers [243]. Our experiments demonstrate that fewer MPGNN layers perform better than more MPGNN layers. This prevents oversmoothing [244, 245], which is exacerbated by the necessity of scalar representations.

The readout function, the function that consolidates a filtration into a global graph representation, is determined by computing four types of bars for the extended persistence on the concatenation of the lower and upper filtrations followed by compositions with four rational hat functions $\hat{r}$ as used in [207, 208, 211]. To each of the four types of bars in barcode $\mathcal{B}$, we apply the hat function $\hat{r}$ to obtain a $k$-dimensional vector. The function $\hat{r}$ is defined as:

$$\hat{r}(\mathcal{B}) := \left\{ \sum_{\mathbf{p} \in \mathcal{B}} \frac{1}{1 + \|\mathbf{p} - \mathbf{c_i}\|_1} - \frac{1}{1 + \left\| |r_i| - \|\mathbf{p} - \mathbf{c_i}\|_1 \right\|} \right\}_{i=1}^{k} \tag{7.20}$$

where $r_i \in \mathbb{R}$ and $\mathbf{c_i} \in \mathbb{R}^2$ are learnable parameters. The intent of Equation 7.20 is to have controlled gradients. It is derived from a monotonic function, see [207]. This representation is then passed through MLP layers followed by a softmax to obtain prediction probability vector $\hat{p}_G$ for each graph $G$. The negative log likelihood loss from standard graph classification is then used on these vectors $\hat{p}_G$.

If the filtration values on the nodes and edges are distinct, the extended persistence barcode representation is permutation invariant with respect to node indices. Isomorphic graphs with permuted indices and an index filtration with distinct filtration values will have a unique sorted index filtration. Node filtration values are usually distinct since computed floating points rarely coincide. However to break ties and eliminate any dependence on node indices for edges, implement edge filtration values for lower filtration as $F(u,v) = \max(F(u), F(v)) + \epsilon \cdot \min(F(u), F(v))$ and for upper filtration as $F(u,v) = \min(F(u), F(v)) + \epsilon \cdot \max(F(u), F(v))$, $\epsilon$ being very small.

**Cycle Representatives:** A cycle basis of a graph is a set of cycles where every cycle can be obtained from it by a symmetric difference, or sum, of cycles in the cycle basis. It can be shown that every cycle is a sum of the cycles induced by a spanning forest and the complementary edges. Extended persistence computes the same number of *independent* cycles as in the cycle basis induced from a spanning forest. Thus computing extended persistence results in computing a cycle basis. We can explicitly store the cycle representatives, or sequences of filtration scalars, along with the barcode on graph data. This slightly improves the performance in practice and guarantees cycle length classification for arbitrary lengths. After the cycle representatives are stored, we pass them through a bidirectional LSTM then aggregate these LSTM representation per graph and then sum this graph representation by cycles with the vectorization of the graph barcode by the rational hat function of Equation 7.20, see Figure 7.2. The aggregation of the cycle representations is permutation invariant due to the composition of aggregations [213]. In particular, the sum of the barcode vectorization and the mean of cycle representatives, our method's graph representation, must be permutation invariant. What makes keeping track of cycle representatives unique to standard message passing GNNs is that a finite receptive field message passing GNN would never be able to obtain such cycle representations and certainly not from a well formed cycle basis.

### 7.5.1 Efficient Computation of Extended Persistence

The computation for extended persistence can be reduced to applying a matrix reduction algorithm to a coned matrix as detailed in [67]. In [210], this computation was found to be equivalent to a graph algorithm, which we improve upon.

**Algorithm:**

Our algorithm is as follows and written in Algorithm 27. We perform the 0-dimensional persistence algorithm, $\mathrm{PH}_0$, using the union find data structure in $O(m \log n)$ time and $O(n)$ memory for the upper and lower filtrations in lines 1 and 2. See the Appendix Section 7.12.1 for a description of this algorithm. These two lines generate the vertex-edge pairs for $\mathcal{B}_0^{low}$ and $\mathcal{B}_0^{up}$. We then measure the minimum lower filtration value and maximum upper filtration value of each vertex in the union-find data structure found from the $\mathrm{PH}_0$ algorithm as in lines 3 and 4 using the roots of the union-find data structure $U_{up}$ formed by the algorithm. These produce the vertex-vertex pairs in $\mathcal{B}_0^{\mathrm{ext}}$.

See the Appendix Section 7.12.2 for a more thorough explanation of the link-cut tree implementation and the operations we use on it. We collect the max spanning forest $T$ of negative edges, edges that join components, from the upper filtration by repeatedly applying the link operation $n-1$ times in lines 6-8 in decreasing order of $F_{up}$ values and sort the list of the remaining positive edges, which create cycles in line 9. Then, for each positive edge e $= (u, v)$, in order of the upper filtration (line 10), we find the least common ancestor ($lca$) of $u$ and $v$ in the spanning forest $T$ we are maintaining as in line 11. Next, we apply the parallel primitive [162] of *list ranking* twice, once on the path $u$ to $lca$ and the other on the path $v$ to $lca$ in line 12. List ranking allows a list to populate an array in parallel in logarithmic time. The tensor concatenation of the two arrays is appended to a list of cycle representatives as in line 13. This is so that the cycle maintains order from $u$ to $v$. We then apply an $\textsc{ArgMaxReduceCycle}(T, u, v, lca)$ which finds the edge having a maximum filtration value in the lower filtration on it over the cycle formed by $u, v$ and $lca$. We then cut the spanning forest at the edge $(u', v')$, forming two forests as in line 15. These two forests are then linked together at $(u, v)$ as in line 15. The bar $(F_{low}(u', v'), F_{up}(u, v))$

**Algorithm 27:** Efficient Computation of $\mathbf{PH_{ext}}$

**Input:** $G = (V, E)$, $F_{low}$: lower filtration function, $F_{up}$: upper filtration function
**Output:** $\mathcal{B}_0^{low}, \mathcal{B}_0^{up}, \mathcal{B}_0^{ext}, \mathcal{B}_1^{ext}$, $\mathcal{C}$: cycle reps.

$\mathcal{B}_0^{low}$, $E_{pos}^{low}$, $E_{neg}^{low}$, $U_{low} \leftarrow \mathrm{PH}_0(G, F_{low}, lower)$;
$\mathcal{B}_0^{up}$, $E_{pos}^{up}$, $E_{neg}^{up}$, $U_{up} \leftarrow \mathrm{PH}_0(G, F_{up}, upper)$;
$roots \leftarrow \{\textsc{Get_Union-Find_Roots}(U_{up}, v) \mid v \in V\}$;
$\mathcal{B}_0^{ext} \leftarrow \{\min(roots[v]), \max(roots[v]) \mid v \in V\}$;
$\mathbf{T} \leftarrow \{\}$ ; // empty link-cut tree
$\mathcal{B}_1^{ext} \leftarrow \{\{\}\}$;
$\mathcal{C} \leftarrow \{\}$ ; // empty list of cycle representatives
// Sort $E_{neg}^{up}$ by $\mathrm{PH}_0$ in decreasing order of $F_{up}$ values (descending filtration values)
**foreach** $\mathrm{e} = (u, v) \in E_{neg}^{up}$ **do**
  $\mathbf{T} \leftarrow \textsc{Link}(\mathbf{T}, \mathrm{e}, \{w\})$ ; // $w \notin \mathbf{T}$, $w = u$ or $v$
**end**
// Sort $E_{pos}^{up}$ by $\mathrm{PH}_0$ with respect to $F_{up}$ (descending filtration values)
**foreach** $\mathrm{e} = (u, v) \in E_{pos}^{up}$ **do**
  $lca \leftarrow \textsc{Lca}(\mathbf{T}, u, v)$ ; // Get the least common ancestor of $u$ and $v$ to form a cycle
  $P_1 \leftarrow \textsc{ListRank}(\textsc{Path}(u, lca))$;
  $P_2 \leftarrow \textsc{ListRank}(\textsc{Path}(v, lca))$;
  $\mathcal{C} \leftarrow \mathcal{C} \cup \{F_{up}(P_1) \cup F_{up}(Reverse(P_2))\}$ ; // Keep track of the scalar activations on cycle
  $(u', v') \leftarrow \textsc{ArgmaxReduceCycle}(\mathbf{T}, u, v, lca)$ ; // Find max edge on cycle using $F_{low}$
  $\mathbf{T_1}, \mathbf{T_2} \leftarrow \textsc{Cut}(\mathbf{T}, (u', v'))$;
  $\mathbf{T} \leftarrow \textsc{Link}(\mathbf{T_1}, (u, v), \mathbf{T_2})$;
  $\mathcal{B}_1^{ext} \leftarrow \mathcal{B}_1^{ext} \cup \{(F_{low}(u', v'), F_{up}(u, v))\}$;
**end**
**return** $(\mathcal{B}_0^{low}, \mathcal{B}_0^{up}, \mathcal{B}_0^{ext}, \mathcal{B}_1^{ext}, \mathcal{C})$

is now found and added to the multiset $\mathcal{B}_1^{ext}$. The final output of the algorithm is four types of bars and a list of cycle representatives: $((\mathcal{B}_0^{low}, \mathcal{B}_0^{up}, \mathcal{B}_0^{ext}, \mathcal{B}_1^{ext}), \mathcal{C})$.

**Complexity:**

We improve upon the complexity of [210] by obtaining a $O(mn)$ work $O(m \log n)$ depth algorithm on $O(n)$ processors using $O(n)$ memory. Here $m$ and $n$ are the number of edges and vertices in the input graph. We introduce two ingredients for lowering the complexity, the first is the link-cut dynamic connectivity data structure and the second is the paral-lel primitives of list ranking. The link-cut tree data structure is a dynamic connectivity data structure that can keep track of the spanning forest with $O(\log n)$ amortized time for $\textsc{Link}, \textsc{Cut}, \textsc{Path}, \textsc{Lca}, \textsc{ArgMaxReduce}$. Furthermore, list ranking [246] is an $O(\log n)$ depth and $O(n)$ work parallel algorithm on $O(\frac{n}{\log n})$ processors that determines the distance of each vertex from the start of the path or linked list it is on. In other words, list ranking turns a linked list into an array in parallel. Sorting can be performed in parallel using $O(n \log n)$ work and $O(\log n)$ depth.

Notice that if we do not keep track of cycle representatives (remove lines 12 and 13 from Algorithm 27), then we have an $O(m \log n)$ time sequential algorithm. The repeated calling of the supporting operation $\textsc{Expose}()$ dominates the complexity, see Appendix Section 7.12.2.

## 7.6 Expressivity of Extended Persistence

We prove some properties of extended persistence barcodes. We also find a case where extended persistence with supervised learning can give high performance for graph classification. WL[1] bounded GNNs, on the other hand, are guaranteed to not perform well. *Certainly all such results also apply for the explicit cycle representatives since the min and max on the scalar activations on the cycle form the corresponding bar.*

### 7.6.1 Some Properties

The following Theorem 7.6.2 states some properties of extended persistence. This should be compared with the 0- and 1-dimensional persistence barcodes in the standard persistence. Every vertex and edge is associated with some bar in the standard persistence though they can be both finite or infinite. However, in extended persistence all bars are finite and we form barcodes from an extended filtration of $2m + 2n$ edges and vertices instead of the standard $(m + n)$-lengthed filtration.

**Theorem 7.6.2.** *(Extended Barcode Properties)*

$\mathbf{PH_{ext}}(G)$ *produces four multisets of bars:* $\mathcal{B}_1^{\mathrm{ext}}, \mathcal{B}_0^{\mathrm{ext}}, \mathcal{B}_0^{low}, \mathcal{B}_0^{up}$, *s.t.*

$|\mathcal{B}_1^{\mathrm{ext}}| = \dim H_1 = m - n + C$,

$|\mathcal{B}_0^{ext}| = \dim H_0 = C,$

$|\mathcal{B}_0^{low}| = |\mathcal{B}_0^{upper}| = n - C,$

*where there are $C$ connected components and $\dim H_k$ is the dimension of the $k$th homology group s.t.:*

*1. the $H_1(G)$ bars comes from a cycle basis of $G$ which also constitutes a basis of its fundamental group,*

*2. $\dim H_1(G)$ counts the number of chordless cycles when $G$ is outer-planar, and*

*3. there exists an injective filtration function where the union of the resulting barcodes is strictly more expressive than the histogram produced by the WL[1] graph isomorphism test.*

The barcodes found by extended persistence thus have more degrees of freedom than those obtained from standard persistence. For example, a cycle is now represented by two filtration values rather than just one. Furthermore, the persistence $|d - b|$ of a pair $(b, d) \in \mathcal{B}_1^{\text{ext}}$ or $\mathcal{B}_0^{ext}$ can measure topological significance of a cycle or a connected component respectively through persistence. Thus, extended persistence encodes more information than standard persistence. In Theorem 7.6.2, property 1 says that extended persistence actually computes pairs of edges of cycles in a cycle basis. A modification of the extended persistence algorithm could generate all or count certain kinds of important cycles, see [247]. Property 2 characterizes what extended persistence can count.

We makes some observations on the expressivity of $\mathbf{PH_{ext}}$.

**Observation 7.6.3.** *(Cycle Lengths) For any graph $G$ and chordless cycle $\mathbf{C} \subseteq G$, there exists an injective vertex function $f_G$ on $V$ where $\mathbf{PH_{ext}}$ of the vertex-induced filtration for extended persistence can measure the number of edges along $\mathbf{C}$.*

Such a result cannot hold for learning of the filtration by local message passing from constant node attributes. Thus, for the challenging 2CYCLE graphs dataset in Section 7.10.2, it is a necessity to use the cycle representatives $\mathcal{C}$ for each graph to distinguish pairs of cycles of arbitrary length. This should be compared with Top-$K$ methods, $K$ being a constant hyper parameter such as in [223, 248]. The constant hyper parameter $K$ prevents learning an arbitrarily long cycle length when the node attributes are all the same. Furthermore, a readout function like SUM is agnostic to graph topology and also struggles with learning when presented with an arbitrarily long cycle. This struggle for distinguishing cycles in standard MPGNNs is also reported in [249]. An observation similar to the previous Observation 7.6.3 can also be made for paths measured by $\mathcal{B}_0^{ext}$.

**Observation 7.6.4.** *(Connected Component Sizes) For any graph $G$ and all connected components $\mathbf{CC} \subseteq G$, there exists an injective vertex function $f_G$ defined on $V$ where $\mathbf{PH_{ext}}$ of the vertex-induced filtration for extended persistence can measure the number of vertices in $\mathbf{CC}$.*

We investigate the case where no learning takes place, namely when the filtration values come from a random noise. We observe that even in such a situation some information is still encoded in the extended persistence barcodes with a probability that depends on the graph.

**Observation 7.6.5.** *For any graph $G$ where every edge belongs to some cycle and an extended filtration on it induced by randomly sampled vertex values $x_i \sim U([0,1])$, $\mathbf{PH_{ext}}$ has a $H_1(G)$ bar $[\max_i(x_i), \min_i(x_i)]$ with probability $\sum_{v \in V} \frac{1}{n} \frac{deg(v)}{n-1} = \frac{2m}{n(n-1)}$.*

Notice that for a clique, the probability of finding the bar with maximum possible persistence is 1. It becomes lower for sparser graphs.

**Corollary 7.6.6.** *In Observation 7.6.5, assuming the bar $[\max_i(x_i), \min_i(x_i)]$ exists, the expected persistence of that bar $\mathbb{E}[|\max_i(x_i) - \min_i(x_i)|]$ goes to 1 as $n \to \infty$.*

What Corollary 7.6.6 implies is that, for certain graphs, even when nothing is learned by the GNN filtration learning layers, the longest $\mathcal{B}_1^{ext}$ bar indicates that $n$ is large. This happens for graphs that are randomly initialized with vertex labels from the unit interval and occurs with high probability for dense graphs by Observation 7.6.5. For large $n$, the empirical mean of the longest bar will have persistence near 1. Notice that $\mathcal{B}_1^{ext}$ can measure this even though the number of $H_1(G)$ bars, $m - n + C$, could tell us nothing about $n$.

**Observation 7.6.7.** *Computing a cycle basis of a graph $G$ provides strictly more expressivity for $G$ than any tree-based encoding such as WL[1] or standard persistence.*

*Proof.* Consider the example of two nodes $u, v$ with many nodes $w_\mathrm{i}$, $\mathrm{i} = 1...k$ attached to both $u$ and $v$ through edges $(w_\mathrm{i}, u)$ and $(w_\mathrm{i}, v)$. Call this graph

$$G = (V = \{u, v\} \cup \bigcup_\mathrm{i} \{w_\mathrm{i}\}, E = \bigcup_\mathrm{i} \{(w_\mathrm{i}, u), (w_\mathrm{i}, v)\}) \tag{7.21}$$

If we have a filtration function $f_G$ satisfying $f_G(w_k) > f_G(v), \forall v \in V \setminus \{w_k\}$ then in standard persistence, the creation of all $k$ cycles cannot be distinguished. If a cycle basis is determined, however, then these $k$ cycles are spanned by the cycle basis. Thus the integer coefficients on the linear combination of the cycles in the cycle basis distinguish each of the $k$ cycles.

Since WL[1] views the graph as a collection of rooted trees, it cannot detect cycles. □

There are many cycle bases that can be chosen. A fundamental cycle basis is the cycle basis consisting of the complementary edges of a spanning tree and the cycles formed by the path to the least common ancestor of the nodes that form the complementary edges. This cycle basis completely relies on the shortest paths in the spanning tree and would depend on the graph $G$. The extended cycle basis is a cycle basis amongst all cycle bases directly induced by a reversal, through changing the max to min operation on edges, of the filtration function $f_G$ independent of $G$ itself.

**Proposition 7.6.8.** *Let* $R_\theta(\bullet) : \mathbb{R}^{ob(\boldsymbol{Set})} \to [0, 1]$ *denote a parameterized readout function.*

*Let* $f : ob(\boldsymbol{Graph}) \to \mathbb{R}^{ob(\boldsymbol{Set})}$ *be a real valued vertex map.*

*The composition* $R_\theta(f(v, G))$, *is continuous with respect to perturbations of* $f(v, G)$.

*Proof.* We check for continuity:

For any $\epsilon > 0$, pick $\delta > 0$ so that $\delta < \min_{u,v \in V(G)} \|f(v, G) - f(u, G)\|$

$\|R_\theta(f(v, G) + \delta) - R_\theta(f(v, G))\| < \epsilon$ □

## 7.7 Experiments

We perform experiments of our method on standard GNN datasets. We also perform timing experiments for our extended persistence algorithm, showing impressive scaling. Fi-nally, we investigate cases where experimentally our method distinguishes graphs that other methods cannot, demonstrating how our method learns to surpass the WL[1] bound.

### 7.7.1 Experimental Setup

We perform experiments on a 48 core Intel Xeon Gold CPU machine with 1 TB DRAM equipped with a Quadro RTX 6000 NVIDIA GPU with 24 GB of GPU DRAM.

Hyper parameter information can be found in Table 7.3. For all baseline comparisons, the hyperparameters were set to their repository's standard values. In particular, all training were stopped at 100 epochs using a learning rate of 0.01 with the Adam optimizer. Vertex attributes were used along with vertex degree information as initial vertex labels if offered by the dataset. We perform a fair performance evaluation by performing standard 10-fold cross validation on our datasets. The lowest validation loss is used to determined a test score on a test partition. An average±standard deviation test score over all partitions determines the final evaluation score.

The specific layers of our architecture for the neural network for our filtration function $f_G$ is given by one or two GIN convolutional layers, with the number of layers as determined by an ablation study.

### 7.7.2 Performance on Real World and Synthetic Datasets

We perform experiments with the TUDatasets [250], a standard GNN benchmark. We compare with WL[1] bounded GNNs (GIN, GIN0, GraphSAGE, GCN) from the PyTorch Geometric [251, 252] benchmark baseline commonly used in practice as well as GFL[211], ADGCL [253], and InfoGraph [254], self-supervised methods. Self supervised methods are promising but should not surpass the performance of supervised methods since they do not use the label during representation learning. We also compare with existing topology based methods TOGL [208] and GFL [211]. We also perform an ablation study on the readout function, comparing extended persistence as the readout function with the SUM, AVERAGE, MAX, SORT, and SET2SET [255] readout functions. The hyper parameter $k$ is set to the 10th percentile of all datasets when sorting for the top-$k$ nodes activations. We do not compare with [223] since its code is not available online. The performance numbers are listed in Table 7.1. We are able to improve upon other approaches for almost all cases. The real world datasets include DD, MUTAG, PROTEINS and IMDB-MULTI. DD, PROTEINS, and MUTAG are molecular biology datasets, which emphasize cycles, while IMDB-MULTI is a social network, which emphasize cliques and their connections. We use accuracy as our performance score since it is the standard for the TU datasets.

**Table 7.1.** Average accuracy ± std. dev. of our approach (GEFL) with and without explicit cycle representations, Graph Filtration Learning (GFL), GIN0, GIN, GraphSAGE, GCN, ADGCL, GraphCL and TOGL and a readout ablation study on the four TUDatasets: DD, PROTEINS, IMDB-MULTI, MUTAG as well as the two Synthetic WL[1] bound and Cycle length distinguishing datasets. Numbers in bold are highest in performance; bold-gray numbers show the second highest. The symbol – denotes that the dataset was not compatible with software at the time.

| Experimental Evaluation | | | | | | |
|---|---|---|---|---|---|---|
| avg. acc. ± std. | DD | PROTEINS | IMDB-MULTI | MUTAG | PINWHEELS | 2CYCLES |
| GFL | 75.2 ± 3.5 | 73.0 ± 3.0 | 46.7 ± 5.0 | **87.2 ± 4.6** | 100 ±0.0 | 50.0 ±0.0 |
| **Ours+Bars** | 75.5 ± 2.9 | **74.9 ± 4.1** | **50.3 ± 4.7** | **88.3 ± 7.1** | **100 ±0.0** | 50 ± 0.0 |
| **Ours+bars+cycles** | **75.9 ± 2.0** | **75.2 ± 4.1** | **51.0 ± 4.6** | 86.8 ± 7.1 | **100 ±0.0** | **100 ± 0.0** |
| GIN | 72.6± 4.2 | 66.5 ± 3.8 | 49.8 ± 3.0 | 84.6 ± 7.9 | 50.0 ±0.0 | 50.0 ±0.0 |
| GIN0 | 72.3 ± 3.6 | 67.5 ± 4.7 | 48.7 ± 3.7 | 83.5 ± 7.4 | 50.0 ±0.0 | 50.0 ±0.0 |
| GraphSAGE | 72.6 ± 3.7 | 59.6 ± 0.2 | 50.0 ± 3.0 | 72.4 ± 8.1 | 50.0 ±0.0 | 50.0 ±0.0 |
| GCN | 72.7 ± 1.6 | 59.6 ± 0.2 | 50.0 ± 2.0 | 73.9 ± 9.3 | 50.0 ±0.0 | 50.0 ±0.0 |
| GraphCL | 65.4 ±12 | 62.5 ± 1.5 | 49.6± 0.4 | 76.6 ± 26 | 49.0 ±8.0 | 50.5± 10 |
| InfoGraph | 61.5 ± 10 | 65.5 ± 12 | 40.0 ± 8.9 | 89.1 ± 1.0 | 50.0 ± 0.0 | 50.0 ± 0.0 |
| ADGCL | 74.8± 0.7 | 73.2± 0.3 | 47.4 ± 0.8 | 63.3± 31 | 42.5 ± 19 | 52.5 ± 21 |
| TOGL | 74.7 ± 2.4 | 66.5 ± 2.5 | 44.7 ± 6.5 | - | 47.0 ± 3 | **54.4 ± 5.8** |
| Filt.+SUM | 75.0 ± 3.2 | 73.5± 2.8 | 48.0 ± 2.9 | 86.7± 8.0 | 51.0 ± 11 | 50.0 ± 0.0 |
| Filt.+MAX | 67.6± 3.9 | 68.6± 4.3 | 45.5 ± 3.1 | 70.3± 5.4 | 48.0 ± 4.2 | 50.0 ± 0.0 |
| Filt.+AVG | 69.5± 2.9 | 67.2± 4.2 | 46.7 ± 3.8 | 81.4± 7.9 | 50.0 ± 13 | 50.0 ± 0.0 |
| Filt.+SORT | **76.9± 2.6** | 72.6 ± 4.6 | 49.0± 3.6 | 85.6± 9.2 | 51.0 ± 16 | 50.0 ± 0.0 |
| Filt.+S2S | 69.0 ± 3.3 | 67.8 ± 4.6 | 48.7 ± 4.2 | 86.8 ± 7.1 | 51.0 ± 13 | 50.0 ± 0.0 |

**Table 7.2.** Ablation study on readout functions. The average ROC-AUC ± std. dev. on the ogbg-mol datasets is shown for each readout function. Number coloring is as in Table 7.1

| 10-fold cross validation ablation study on OGBG-MOL datasets by ROC-AUC | | | | | | | |
|---|---|---|---|---|---|---|---|
| avg. score ± std. | **Ours+Bars** | **Ours+Bars +Cycles** | Filt.+SUM | Filt.+MAX | Filt.+AVG | Filt.+SORT | Filt.+Set2Set |
| molbace | **80.0 ± 3.6** | **81.6 ± 3.9** | 79.7 ± 4.6 | 71.9 ± 4.8 | 78.0 ± 3.0 | 78.4 ± 3.3 | 78.2 ± 3.6 |
| molbbbp | 78.0 ± 4.3 | **81.9 ± 3.3** | 76.7 ± 4.9 | 69.8 ± 8.7 | **78.5 ± 4.6** | 76.3 ± 4.3 | 78.0 ± 5.0 |

We also verify that our method surpasses the WL[1] bound, a theoretical property which can be proven, as well as can count cycle lengths when the graph is sparse enough, e.g. when the set of cycles is equal to the cycle basis. This is achieved by the two datasets PINWHEELS and 2CYCLES. See the Appendix Sections 7.10 for the related experimental and dataset details. Both datasets are particularly hard to classify since they contain spurious constant node attributes, with the labels depending completely on the graph connectivity. This removal of node attributes is in simulation of the WL[1] graph isomorphism test, see [212]. Furthermore, doing so is a case considered in [256]. It is known that WL[1], in particular WL[2], cannot determine the existence of cycles of length greater than seven [257, 258].

Table 7.2 shows the ablation study of extended filtration learning on the ogbg datasets [259] OGBG MOLBACE and MOLBBBP. We perform a 10 fold cross validation with the test ROC-AUC score of the lowest validation loss used as the test score. This is performed instead of using the train/val/test split offered by the OGBG dataset in order to keep our evaluation methods consistent with the evaluation of the TUDATASETS and synthetic datasets.

From Section 7.10, we know that there are special cases where extended persistence can distinguish graphs where WL[1] bounded GNNs cannot. We perform experiments to show that our method can surpass random guessing whereas other methods achieve only ~ 50% accuracy on average, which is no better than random guessing. Our high accuracy is guaranteed on PINWHEELS since such graphs are distinguished by counting bars through 0-dim standard persistence. Similarly, 2CYCLES is guaranteed high accuracy when keeping track of cycles and comparing the variance of cycle representations since cycle lengths can be distinguished by a LSTM on different lengthed cycle inputs. Of course, a barcode representation alone will not distinguish cycle lengths.

## 7.8 Conclusion

We introduce extended persistence into the supervised learning framework, bringing in crucial global connected component and cycle measurement information into the graph representations. We address a fundamental limitation of MPGNNs, which is their inability to measure cycles lengths. Our method hinges on an efficient algorithm for computing extended persistence. This is a parallel differentiable algorithm with an $O(m \log n)$ depth $O(mn)$ work complexity and scales impressively over the state-of-the-art. The speed with which we can compute extended persistence makes it feasible for machine learning. Our end-to-end model obtains favorable performance on real world datasets. We also construct cases where our method can distinguish graphs that existing methods struggle with.

## 7.9 Proofs

**Theorem 7.9.1.** *(Theorem 7.6.2)*

$\mathbf{PH_{ext}}(G)$ *produces four types of bars:* $\mathcal{B}_1^{\text{ext}}, \mathcal{B}_0^{\text{ext}}, \mathcal{B}_0^{low}, \mathcal{B}_0^{up}$, *s.t.*

1. $|\mathcal{B}_1^{\text{ext}}| = \dim H_1 = m - n + C$,

2. $|\mathcal{B}_0^{\text{ext}}| = \dim H_0 = C$,

3. $|\mathcal{B}_0^{low}| = |\mathcal{B}_0^{upper}| = n - C$,

*where there are* $C$ *connected components and* $\dim H_k$ *is the dimension of the* $k$*th homology group s.t.:*

1. *the* $H_1(G)$ *barcode comes from a cycle basis of* $G$ *which also constitutes a basis of its fundamental group,*

2. $\dim H_1(G)$ *counts the number of chordless cycles when* $G$ *is outer-planar, and*

3. *there exists an injective filtration function where the union of the resulting barcodes is strictly more expressive than the histogram produced by the WL[1] graph isomorphism test.*

*Proof.* There are $n$ bars with vertex births since every vertex creates exactly one connected component. The number of these bars which are in $\mathcal{B}_0^{\text{ext}}$ is $C$, which counts the number of global connected components. In other words, $\mathcal{B}_0^{\text{ext}} = \dim(H_0(G)) = C$. Thus, we have $n - C = |\mathcal{B}_0^{low}| = |\mathcal{B}_0^{upper}|$.

Considering all $2m$ edges on the extended filtration, every edge gets paired. Furthermore, $n - C$ of the edges in the lower filtration are negative edges paired with vertices that give birth to connected components. Similarly there are $n - C$ edges paired with vertices in the upper filtration. We thus have $\frac{2m-2(n-C)}{2}$ edge-edge pairings in $\mathcal{B}_1^{\text{ext}}$ because every edge gets paired. Thus, $|\mathcal{B}_1^{\text{ext}}| = m - n + C$. Since each bar in $\mathcal{B}_1^{\text{ext}}$ counts a birth of a 1-dimensional homological class which together span the 1-dimensional homological classes in $H_1(G)$, we have that $\dim H_1 = |\mathcal{B}_1^{\text{ext}}|$.

1. All cycle representatives found by the algorithm are a symmetric difference of cycles from a *fundamental* cycle basis, or cycle basis induced from a spanning forest. This follows since the link and cut operations in our algorithm correspond to cycle additions in extended persistence computation. Furthermore, each cycle in the returned cycle representatives is independent because it has a unique cycle from the fundamental cycle basis as a summand. Also, there are $m-n+1$ returned cycle representatives, same as the fundamental cycle basis. Thus the cycle representatives form a cycle basis.

Any cycle basis generates the fundamental group and $H_1(G, \mathbb{Z}_2)$ homology group of graph $G$ [260]

2. By Euler's formula, we have $n-m+F = C+1$ for planar graphs where $F$ is the number of faces of the planar graph as embedded in $\mathbb{S}^2$. For outer planar graphs, since $F-1$ interior faces lie on one hemisphere of $\mathbb{S}^2$ and one exterior face covers the opposite hemisphere, each interior face must be a chordless cycle.

3. This follows directly by the result in [208] stating that 0-dimensional barcodes are more expressive than the WL[1] graph isomorphism test. In extended persistence, $\mathcal{B}_0^{low}$ and $\mathcal{B}_0^{ext}$ are computed. Since all bars in $\mathcal{B}_0^{ext}$ correspond to infinite bars denoted $\mathcal{B}_0^{\infty}$ in the 0-dimensional standard persistence, we have that $\mathcal{B}_0^{low}$ and $\mathcal{B}_0^{ext}$ carry at least the same amount of information as a 0-dimensional barcode as determined by $\mathcal{B}_0^{low}$ and $\mathcal{B}_0^{\infty}$.

□

**Observation 7.9.2.** *(Observation 7.6.3) For any graph $G$ and chordless cycle $\mathbf{C} \subseteq G$, there exists an injective vertex function $f_G$ on $V$ where $\mathbf{PH_{ext}}$ of the vertex-induced filtration for extended persistence can measure the number of edges along $\mathbf{C}$.*

*Proof.* Number the vertices of the cycle $\mathbf{C}$ of length $k$ in descending order and counter clockwise as $n-1...n-k$. For each vertex $u \in \mathbf{C}$, set $f_G(u)$ to be the index of $u$ in the vertex numbering, the lower and upper filtration value on the nodes. Let $f_G^{low}(u) = f_G^{up}(u) = f_G(u)$. For the other vertices, assign arbitrary different values less than $n-k$. The edge values are then assigned $f_G^{up}(\{u,v\}) = \min(f_G^{up}(u), f_G^{up}(v))$ and $f_G^{low}(\{u,v\}) = \max(f_G^{low}(u), f_G^{low}(v))$. Apply $\varepsilon$-perturbation to make the filtration functions $f_G^{low}, f_G^{up}$ injective and to force exactly one edge on the cycle to be positive. Every edge is either positive or negative and all negative edges are on a spanning forest.

In particular, for vertex $u$ and all its incident edges of same upper filtration value, one can subtract different $\varepsilon \in \mathbb{R}^+$ from each edge to impose an order amongst edges with the same value from $f_G^{up}$. Similarly, for $f_G^{low}$ subtract different $\varepsilon \in \mathbb{R}^+$ to each edge to break ties. For the edges on $\mathbf{C}$, do not subtract an $\varepsilon$ in order to force each node i on $\mathbf{C}$ to pair with the largest edge: $(\mathrm{i}-1, \mathrm{i})$ of filtration value i. Since there is a tie for the edges in $\mathcal{C}$ incident to node $n-k$ in the upper filtration, set the edge $f_G^{up}(n-k, n-k+1) := n-k+\varepsilon$. This ensures that edge $(n-k, n-k+1)$ is negative and $(n-1, n-k)$ is positive in the upper filtration since the edges with larger filtration values are paired, or made negative, first. Similarly, there is a tie for the edges in $\mathcal{C}$ incident to node $n-1$ in the lower filtration. Set $f_G^{low}(n-1, n-k) := n-1+\varepsilon$. This ensures that edge $(n-1, n-2)$ is positive in the lower filtration and $(n-1, n-k)$ is negative in the lower filtration. The resulting filtration functions $f_G^{low}$ and $f_G^{up}$ are injective. Furthermore, we then get that every edge on the cycle $\mathbf{C}$ except one: $(n-1, n-k)$, a positive edge, becomes negative in the upper filtration and thus belongs to the negative spanning forest of the upper filtration. The positive edge of smallest value in the upper filtration is edge $(n-1, n-k)$. The extended persistence algorithm, after computing $\mathcal{B}_0^{low}$ and $\mathcal{B}_0^{up}$, pairs the edge $\mathrm{e} = (n-1, n-k)$ with the edge having maximum value in the lower filtration in the cycle $\mathbf{C}$ that e forms with the spanning forest. This paired edge is $(n-1, n-2)$ and has lower filtration value $n-1$. We thus have the bar $[n-1, n-k]$ which encodes the length $k$ of the cycle $\mathbf{C}$.

□

**Observation 7.9.3.** *(Observation 7.6.4) For any graph $G$ and all connected components $\mathbf{CC} \subseteq G$, there exists an injective vertex function $f_G$ defined on $V$ where $\mathbf{PH_{ext}}$ of the vertex-induced filtration for extended persistence can measure the number of vertices in $\mathbf{CC}$.*

*Proof.* For each connected component $\mathbf{CC}$ in $G$, index the vertices in $\mathbf{CC}$ in consecutive order where indices in each connected component remain distinct. Then define $f_G(u)$ equal to the index of $u$ in $G$. Let $f_G(u) = f_G^{low}(u) = f_G^{up}(u)$. By some $\varepsilon$-perturbation, where we break ties amongst edges, we can make these two functions injective on the graph $G$. Since $\mathcal{B}_0^{\mathrm{ext}}$ has each bar $[\min_{u \in \mathbf{CC}} f_G^{low}(u), \max_{u \in \mathbf{CC}} f_G^{up}(u)]$ and since all indices are consecutive, each bar's persistence in $\mathcal{B}_0^{ext}$ measures how many vertices are in the connected component they constitute.

□

**Observation 7.9.4.** *(Observation 7.6.5) For any graph $G$ where every edge belongs to some cycle and an extended filtration on it is induced by randomly sampling vertex values $x_{\mathrm{i}} \sim U([0,1])$, $\mathbf{PH_{ext}}$ has the $H_1(G)$ bar $[\max_{\mathrm{i}}(x_{\mathrm{i}}), \min_{\mathrm{i}}(x_{\mathrm{i}})]$ with probability $\sum_{v \in V} \frac{1}{n} \frac{deg(v)}{n-1} = \frac{2m}{n(n-1)}$.*

*Proof.* Since the probability of finding a given permutation on $n$ vertices sampled uniformly at random without replacement is equivalent to the probability of a given order on the vertices sampled uniformly at randomly $n$ times, it suffices to find the probability of sampling uniformly at random without replacement two vertices that are connected with an edge in $G$.

For a fixed $\sigma \in S_n$, a permutation from the group $S_n$ of permutations on $n$ vertices, we have:

$$\begin{aligned} \frac{1}{n!} &= P(x_n < x_{n-1} < \ldots < x_1, x_{\mathrm{i}} \sim U([0,1])) \\ &= \int_0^1 \int_0^{x_1} \ldots \int_0^{x_{n-1}} dx_n dx_{n-1} \ldots dx_1 = P(\sigma \sim U(S_n)) \end{aligned} \tag{7.22}$$

Let $G = (V, E)$ be the graph with vertex values sampled from a uniform distribution. Let $G' = (V', E')$ be the same graph with vertex values in $\{0, 1, \ldots, n-1\}$ sampled uniformly without replacement. We know that the probability for a given order on these vertices is the same for both graphs. In fact, the two node labelings are in bijection with each other. By the law of total probability and with Equation 7.22:

$$\begin{aligned} & P((\min_{\mathrm{i}} x_{\mathrm{i}}, \max_{\mathrm{i}} x_{\mathrm{i}}) \in E, x_{\mathrm{i}} \sim U([0,1])) \\ &= \sum_{v \in V} (P(v = \max_{\mathrm{i}} x_{\mathrm{i}}, x_{\mathrm{i}} \sim U([0,1])) \cdot P(\min_{\mathrm{i}} x_{\mathrm{i}} \in \mathrm{Nbr}(v) \| v = \max_{\mathrm{i}} x_{\mathrm{i}}, x_{\mathrm{i}} \sim U([0,1]))) \\ &= \sum_{v \in V} (n-1)! \int_0^1 \int_0^{x_1} \ldots \int_0^{x_{n-1}} dx_n dx_{n-1} \ldots dx_1 \cdot deg(v)(n-2)! \int_0^1 \int_0^{x_2} \ldots \int_0^{x_{n-1}} dx_n dx_{n-1} \ldots dx_2 \\ &= \sum_{v \in V'} (P(v = n-1) \cdot P(0 \in \mathrm{Nbr}(v) \mid v = n-1)) = P((n-1, 0) \in E') \\ &= \sum_{v \in V'} \frac{1}{n} \frac{deg(v)}{n-1} \end{aligned}$$

We now show that if $(\min_{\mathrm{i}} x_{\mathrm{i}}, \max_{\mathrm{i}} x_{\mathrm{i}})$ occurs as an edge in $G = (V, E)$, where every edge belongs to some cycle, then the bar $[\max_{\mathrm{i}} x_{\mathrm{i}}, \min_{\mathrm{i}} x_{\mathrm{i}}]$ is guaranteed to occur.

The spanning tree comprised of negative edges that begins the computation for $\mathcal{B}_1^{ext}$ as in line 6 of Algorithm 27 for the $H_1(G)$ barcode computation is a maximum spanning tree. This is because the negative edges are just those found by the Kruskal's algorithm for the 0-dimensional standard persistence applied to an upper filtration. Since e $= (\min_{\mathrm{i}} x_{\mathrm{i}}, \max_{\mathrm{i}} x_{\mathrm{i}})$ has value $\min_{\mathrm{i}} x_{\mathrm{i}}$ in the upper filtration and since every edge belongs to at least one cycle, it cannot be in the maximum spanning tree. Thus e is a positive edge.

Since e is positive in the upper filtration, it will be considered at some iteration of the for loop in line 10 of Algorithm 27. When we consider it, it will form a cycle **C** with the dynamically maintained spanning forest. To form a persistence $H_1(G)$ bar for e, we pair it with the maximum edge in the cycle **C** in the lower filtration. This forms a bar $[\max_i x_i, \min_i x_i]$.

□

**Corollary 7.9.5.** *In Observation 7.6.5, assuming the bar* $[\max_i(x_i), \min_i(x_i)]$ *exists, the expected persistence of that bar,* $\mathbb{E}[|\max_i(x_i) - \min_i(x_i)|]$*, goes to* 1 *as* $n \to \infty$*.*

*Proof.* Define the random variable $X_n = |\max_i x_i - \min_i x_i|$ for $n$ random points drawn uniformly from $[0,1]$. We find $\lim_{n\to\infty} \mathbb{E}[X_n]$. The following sequence of equations follow by repeated substitution.

$$\begin{aligned}\mathbb{E}[X_n] &= n! \int_0^1 \int_0^{x_1} \dots \int_0^{x_{n-1}} (x_1 - x_n) dx_n \dots dx_1 \\ &= n! \int_0^1 \left(\frac{x_1^n}{(n-1)!} - \frac{x_1^n}{n!}\right) dx_1 = \frac{n-1}{n+1}\end{aligned}$$

where the $n!$ comes from symmetry.

Therefore: $\lim_{n\to\infty} \mathbb{E}[X_n] = 1$. □

## 7.10 Demonstrating the Expressivity of Learned Extended Persistence

We present some cases where the classification performance of our method excels. We look for graphs that cannot be distinguished by WL[1] bounded GNNs. We find that pinwheeled cycle graphs and varied length cycle graphs can be perfectly distinguished by learned extended persistence and, in practice, with much better performance than random guessing using our model. See the experiments Section 7.7 to see the empirical results for our method against other methods on this synthetic data.

### 7.10.1 Pinwheeled Cycle Graphs (The PINWHEELS Dataset)

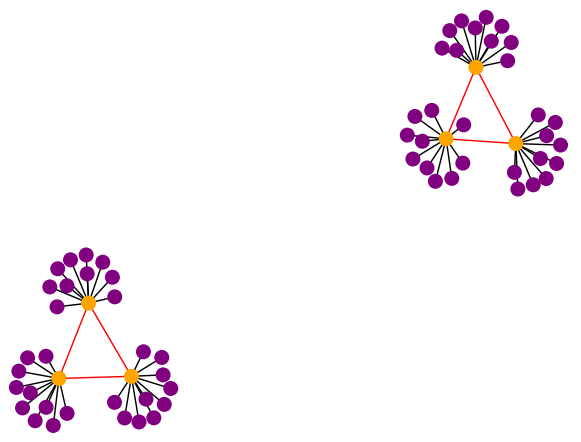

**Figure 7.3.** Class 0: 2 triangles with pinwheel at each vertex.

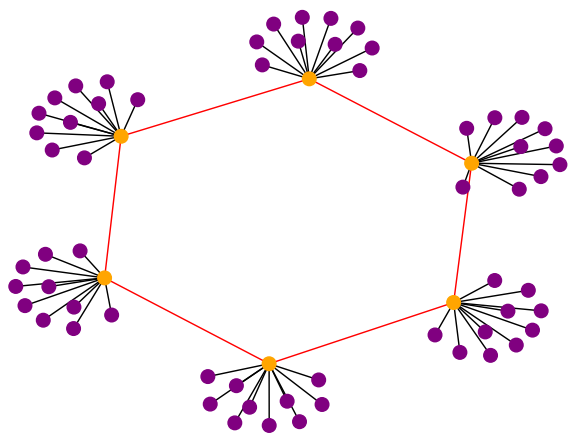

**Figure 7.4.** Class 1: A hexagon with pinwheel at each vertex.

We consider pinwheeled cycle graphs. To form the base skeleton of these graphs, we take the standard counter example to the WL[1] test of 2 triangles and 1 hexagon. We then append pinwheels of a constant number of vertices to the vertices of these base skeletons. The node attributes are set to a spurious constant noise vector. They have no effect on the labels.

It is easy to check that both Class 0 and Class 1 graphs are indistinguishable by WL[1]; see Figures 7.3 and 7.4. Notice that if there are 6 core vertices and edges in the base skeleton and if there are pinwheels of size $k$, then with edge deletions and vertex deletions composed, we have a $1 - (\frac{6}{6k+6})^2$ probability of only deleting a pinwheel edge or vertex and thus not affecting $H_1(G)$. This probability converges to 1 as $k \to \infty$. According to Theorem 7.6.2, $\dim H_1(G)$ measures the number of cycles and $\dim H_0$ measures the number of connected components. If neither of these counts are affected by training during supervised learning, our method is guaranteed to distinguish the two classes simply by counting according to Theorem 7.6.2.

Certainly the pinwheeled cycle graphs, are distinguishable by counts of bars. We check this experimentally by constructing a dataset of 1000 graphs of two classes of graph evenly split. Class 0 is as in Figure 7.3 and involves two triangles with pinwheels of random sizes. Class 1 is as in Figure 7.4 with a hexagon and pinwheels of random sizes attached. We obtain on average 100% accuracy. This is confirmed experimentally in Table 7.1. This matches the performance of GFL [211], since counting bars, or Betti numbers, can also be done through 0-dim. standard persistence. Interestingly TOGL does not achieve a score of 100 accuracy on this dataset. We conjecture this is because their layers are not able to ignore the spurious and in fact misleading constant node attributes.

### 7.10.2 Regular Varied Length Cycle Graphs (The 2CYCLES dataset)

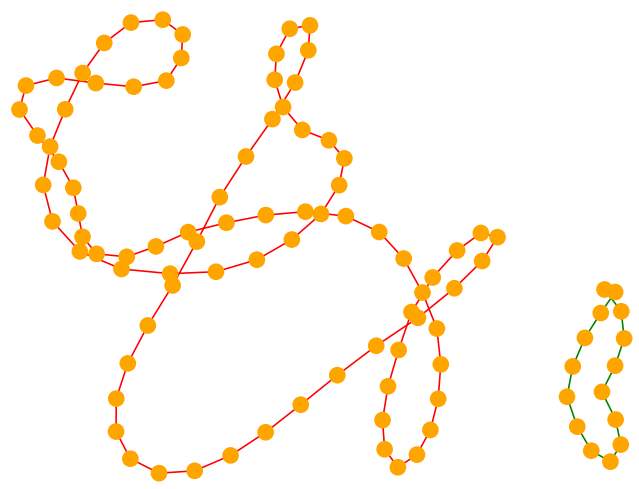

**Figure 7.5.** Class 0: A 15 node cycle and an 85 node cycle.

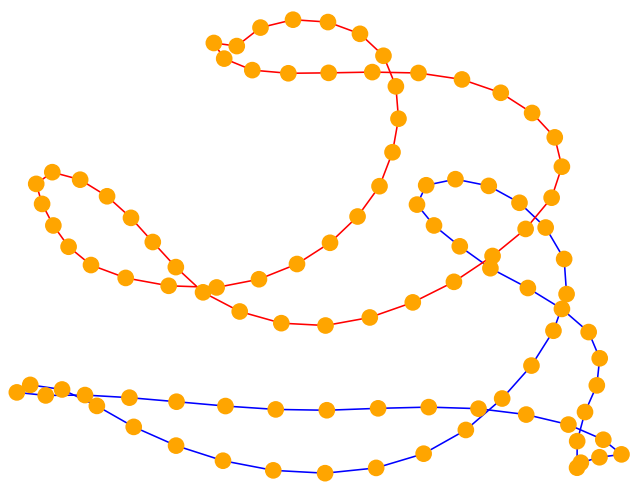

**Figure 7.6.** Class 1: A 50 node cycle with a 50 node cycle.

We further consider varied length cycle graphs. These are graphs that involve two cycles. Class 0 has one short and one long cycle while Class 1 has two near even lengthed cycles. The node attributes are all the same and spurious in this dataset. Extended persistence should do well to distinguish these two classes. We conjecture this based on Observation 7.6.3, which states that there is some filtration that can measure the length of certain cycles.

It is the path length, coming from Observation 7.6.4, which is being measured. The 0-dimensional standard persistence is insufficient for this purpose. The infinite bars of 0-dimensional standard persistence are determined only by a birth time. Furthermore, extended persistence without cycle representatives is also insufficient since a message passing

GNN learns a constant filtration function over the nodes. However, with cycle representatives, or a list of scalar node activations per cycle for each graph, we can easily distinguish the average sequence representation since the pair of sequence lengths are different. In class 0, a short cycle and a long cycle are paired while in class 1, two cycles of medium lengths are paired.

A similar but more challenging dataset to the PINWHEELS dataset, the 2CYCLES dataset, is similar to the necklaces dataset from [208] and is illustrated in Section 7.10.2 but with more misleading node attributes and simplified to two cycles. It involves 400 graphs consisting of two cycles. There are two classes as shown in Figures 7.5 and 7.6.

The experimental performance on 2CYCLES surpasses random guessing while all other methods just randomly guess as stated in Section 7.10.2. Certainly WL[1] bounded GNNs cannot distinguish the two classes in 2CYCLES since they are all regular. As discussed, because GFL and TOGL use learned 0-dimensional standard persistence, these approaches do no better than random guessing on this dataset.

## 7.11 Timing of Extended Persistence Algorithm (without storing cycle representations)

Since the persistence computation, especially extended persistence computation, is the bottleneck to any machine learning algorithm that uses it, it is imperative to have a fast algorithm to compute it. We perform timing experiments with a C++ torch implementation of our fast extended persistence algorithm. In our implementation each graph in the batch has a single thread assigned to it.

Our experiment involves two parameters, the sparsity, or probability, $p$ for the edges of an Erdos-Renyi graph and the number of vertices of such a graph $n$. We plot our speedup over GUDHI, the state of the art software for computing extended persistence, as a function of $p$ with $n$ held fixed. We run GUDHI and our algorithm 5 times and take the average and standard deviation of each run's speedup. Since our algorithm has lower complexity, our speedup is theoretically unbounded. We obtain up to 62x speedup before surpassing 12 hours of computation time for experimentation. The plot is shown in Figure 7.7. The speedup is up to 2.8x, 9x, 24x, and 62x for $n = 200, 500, 1000, 2000$ respectively.

## 7.12 Algorithm and Data Structure Details

Here we detail the algorithmic details of computing extended persistence.

### 7.12.1 The $PH_0(G)$ Algorithm

Algorithm 28 is the union-find algorithm that computes 0 dimensional persistent homology. The algorithm is a single-linkage clustering algorithm [261]. It starts with $n$ nodes, 0 edges, and a union-find data structure [262] on $n$ nodes. The edges are sorted in ascending order if a lower filtration function is given. Otherwise, the edges are sorted in descending order. It then proceeds to connect nearest neighbor clusters, or connected components, in a sequential fashion by introducing edges in order one at a time. Two connected components are nearest to each other if they have two nodes closer to each other than any other pair of connected components. This is achieved by iterating through the edges in sorted order and merging the connected components that they connect. When given a lower filtration function, when a connected component merges with another connected component, the connected component with the larger connected component root value has its root filtration function value a birth time. This birth time is paired with the current edge's filtration value and form a birth death pair. The smaller of the two connected component root values is used as birth time when an upper filtration function is given. The two connected components are subsequently merged in a union-find data structure by the LINK operation.

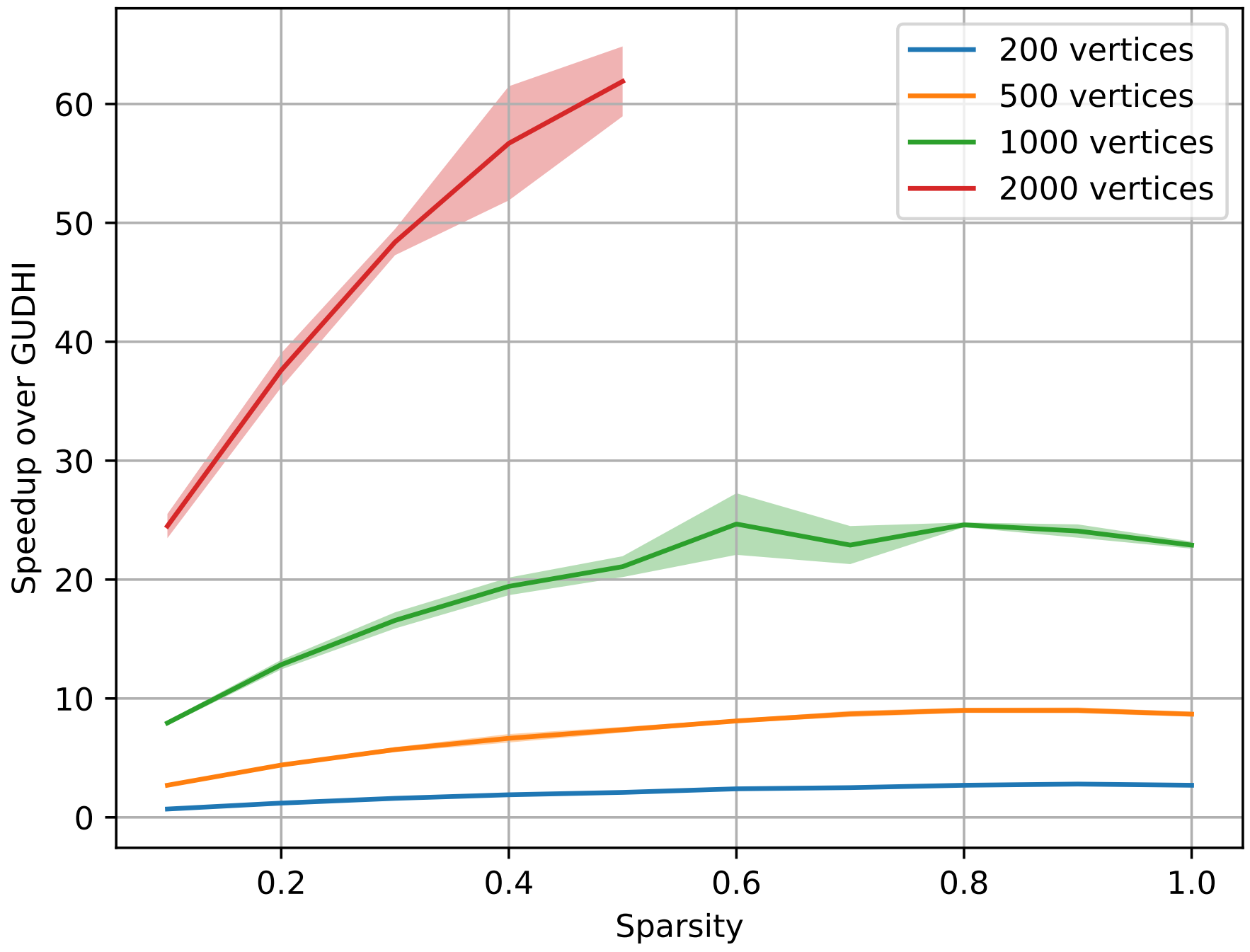

**Figure 7.7.** Average speedup with std. dev. as a function of sparsity $p$ and number of vertices $n$ on Erdos Renyi graphs.

### 7.12.2 A Brief Overview of the Link-Cut Tree Data Structure

The link-cut tree data structure [263] is a well known dynamic connectivity data structure. For modifying the tree of $n$ nodes, it takes $O(\log n)$ amortized time for deleting an edge (cut) and joining two trees (link). Furthermore, it takes $O(\log n)$ amortized time for the composition of associative reductions, such as max, min, sum, on some path from any

**Algorithm 28:** $\mathrm{PH}_0$ Algorithm

**Input:** $G = (V, E)$, $F$: filtration function, *order*: flag to denote an upper or lower filtration

**Output:** $\mathcal{B}_0$, $E_{pos}$, $E_{neg}$, $U$: $H_0(G)$ bars, positive edges, negative edges, and union-find data structure

```
U ← V ;                    // a union-find data structure populated by n unlinked nodes
B_0 ← {} ;                 // A multiset
if order = lower then
    SORT_incr(E);          // increasing w.r.t. F
else
    SORT_decr(E) ;         // decreasing w.r.t F
end
foreach e = (u, v) ∈ E do
    root_u ← U.FIND(u);
    root_v ← U.FIND(v);
    if root_u = root_v then
        E_pos ← E_pos ∪ {e};
    else
        E_neg ← E_neg ∪ {e};
    end
    if order = lower then
        b ← max(F(root_u), F(root_v));
    else
        b ← min(F(root_u), F(root_v));
    end
    d ← F(e);
    B_0 ← B_0 ∪ {{(b, d)}};
    U.LINK(root_u, root_v);
end
return (B_0, E_pos, E_neg, U)
```

node to its root. We may view the link-cut tree data structure as a collection of trees and thus as a forest as well. Details of this forest implementation are omitted.

The link-cut tree decomposes the nodes of a tree $T$ into disjoint preferred paths. A preferred path is a sequence of nodes that strictly decreasing in depth (distance from the root of $T$) on $T$. A path has each consecutive node connected by a single edge. In particular, each node in $T$ has a single preferred child, forming a preferred edge. The maximally connected sequence of preferred edges forms a preferred path. The preferred path decomposition will change as the link-cut tree gets operated on. Each preferred path is in one to one correspondence with a splay tree [264] called an auxiliary tree on the set of nodes in the preferred path. For any node $v$ in a preferred path's auxiliary tree, its left subtree is made up of nodes higher up (closer to the root in $T$) than $v$ and its right subtree is made up of nodes lower (farther from the root in $T$) than $v$. Each auxiliary tree contains a pointer, termed the auxiliary tree's parent-pointer, from its root to the parent of the highest (closest to the root) node in the preferred path associated with the auxiliary tree.

The most important supporting operation to a link-cut tree $T$ is the $\textsc{Expose}()$ operation. The result of $\textsc{Expose}(v)$ for $v \in T$ is the formation of a unique preferred path from the root of $T$ to $v$ with this preferred path's set of nodes forming an auxiliary tree. Furthermore, it results in $v$ to be the root of the auxiliary tree it belongs to. The complexity of $\textsc{Expose}(v)$ is $O(\log n)$. For implementation details, see [263].

Let $T_1, T_2$ be two link-cut trees and $u \in T_1, v \in T_2$ with $u$ a root of $T_1$. Define the operation $\textsc{Link}(T_1, (u, v), T_2)$ as the operation that attaches $T_1$ to $T_2$ by connecting $u$ with $v$ by an edge and outputs the resulting tree. This is achieved by simply calling $\textsc{Expose}(u)$ then $\textsc{Expose}(v)$, which makes $u$ and $v$ the roots of their respective auxiliary trees. In the auxiliary tree of $u$, then set the left child of $u$ to $v$.

Let $T$ be a link-cut tree and $u, v \in T$ connected by an edge with $v$ higher up in $T$, closer to the root of $T$. Define the operation $\textsc{Cut}(T, (u, v))$ as the operation that disconnects $T$ by deleting the edge between $u$ and $v$. This is achieved by simply calling $\textsc{Expose}(u)$ and then making $u$ a root by making the left child of $v$ point to *null*.

Let $T$ be a link-cut tree and $u, v \in T$. Define the operation $\textsc{Lca}(T, (u, v))$ as the operation that finds the least common ancestor of $u$ and $v$ in $T$. This is achieved by calling $\textsc{Expose}(u)$ then $\textsc{Expose}(v)$ and then taking the node pointed to by the parent-pointer of the auxiliary tree of which $u$ is root.

Let $T$ be a link-cut tree and $u, v \in T$ with $v$ higher up in the tree $T$, meaning that it is closer to the root than $u$, and there being a unique path of monotonically changing depth in the tree from $u$ to $v$. Define the operation $\textsc{Path}(u, v)$ as the operation that returns a linked list of the path from $u$ to $v$ in $T$. If $v$ is the root, call $\textsc{Expose}(u)$ and return the linked list formed by the splay tree with $u$ at root. Otherwise, first find the parent $v'$ of $v$. The parent of $v$ can be obtained by calling EXPOSE($v$) then traversing the splay tree it is a root of for its parent in $T$. Call EXPOSE($u$) to form a preferred path from u to the root of $T$ then EXPOSE($v'$) to detach $v'$ from this preferred path. Let SPLAY($u$) be the operation that rotates the unique splay tree, or preferred path, containing $u$ so that $u$ becomes the root of its splay tree. After calling SPLAY($u$), $u$ becomes the root of a linked-list splay tree. It is a linked-list since $u$ is the lowest (farthest from the root) node in its splay tree and the rest of the preferred path is made up of a path of strictly decreasing distance to the root. Return this linked-list splay tree as the resulting path from $u$ to $v$.

Let $T$ be a link-cut tree, $u, v \in T$ with $v$ higher up in the tree than $u$ (it is closer to the root of $T$ than $u$) and there being a unique path of monotonically changing depth in the tree from $u$ to $v$. Define REDUCE($T, u, v, op$) to be an associative reduction on the path from $u$ to $v$. To do this, apply EXPOSE($u$) then EXPOSE($v$), then apply the associative operation on the whole auxiliary tree rooted at $u$, as implemented on a splay tree in [264]. The associative reduction takes $O(\log n)$ time. This splay tree corresponds to the preferred path from $u$ to $v$ formed from the two EXPOSE operations. Notice that EXPOSE($u$) results in a preferred path from $u$ to the root while the second call EXPOSE($v$) detaches the path from $v$ to the root of $T$ from the preferred path of $u$ to the root.

Let $T$ be a link-cut tree, $u, v \in T$ and $lca$ the least common ancestor of $u, v \in T$. Assume the nodes are labeled by a pair of their value and index. Two nodes are compared by their respective values. Define ARGMAXREDUCECYCLE($T, u, v, lca$) as the operation that finds the edge with one of its nodes containing the maximum value on the cycle formed by $u, v$ and $lca$. There are many ways to implement this. We describe a method that maintains the $O(\log n)$ complexity of link-cut tree operations. We first compute $(value(w_1), w_1) :=$ REDUCE($T, u, lca, max$) to find the maximum value node along the path from $u$ to $lca$, then compute $(value(w_2), w_2) :=$ REDUCE($T, v, lca, max$) to find the maximum value node along the path from $v$ to $lca$. Let $w$ to be the maximum valued vertex between $w_1$ and $w_2$. If $w \neq lca(u, v)$, then find the parent $z$ of $w$; otherwise, apply EXPOSE($u$) then EXPOSE($v$) and keep track of the child $z$ of $w$ that gets detached during EXPOSE($v$). Parent of $w$ can be found by EXPOSE($w$) then traversing its splay tree to find the parent of $w \in T$. The edge $(z, w)$ is returned by ARGMAXREDUCECYCLE($T, u, v, lca$)

## 7.13 Cycle Length Distribution of the Cycle Basis found by Extended Persistence Algorithm for Erdos-Renyi Graphs

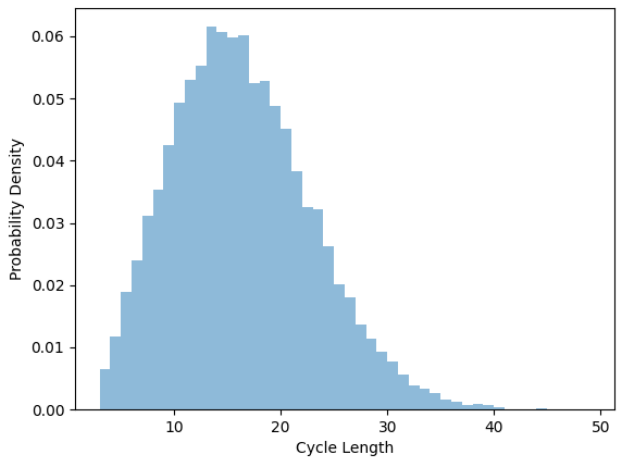

(a) 2000 nodes, 0.01 sparsity

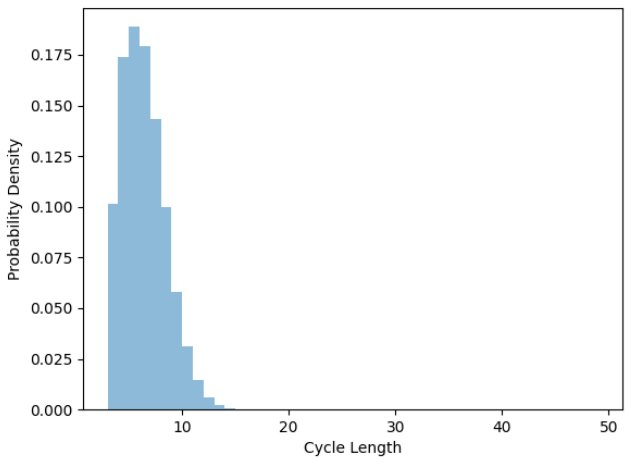

(b) 2000 nodes, 0.1 sparsity

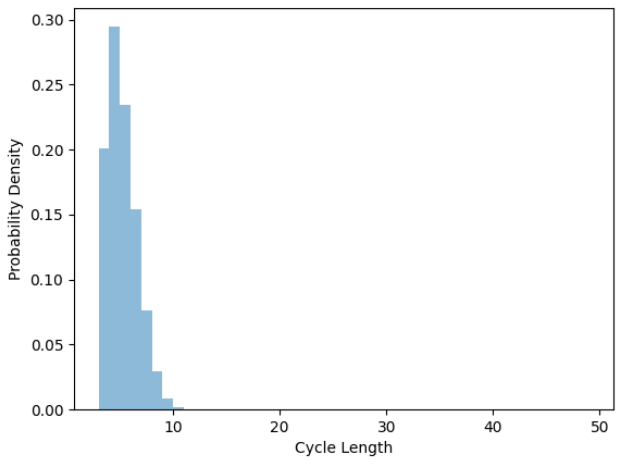

(c) 2000 nodes, 0.3 sparsity

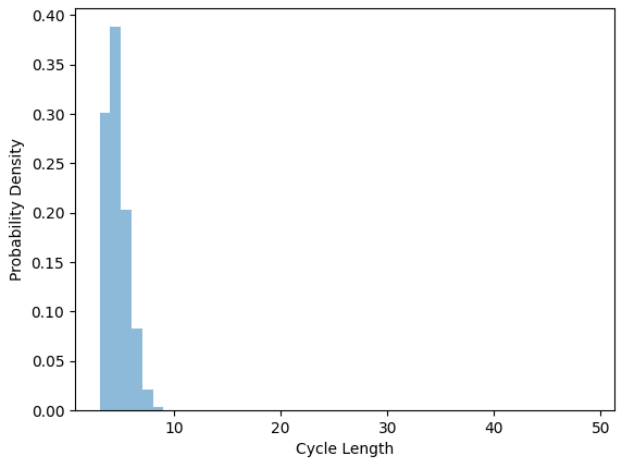

(d) 2000 nodes, 0.5 sparsity

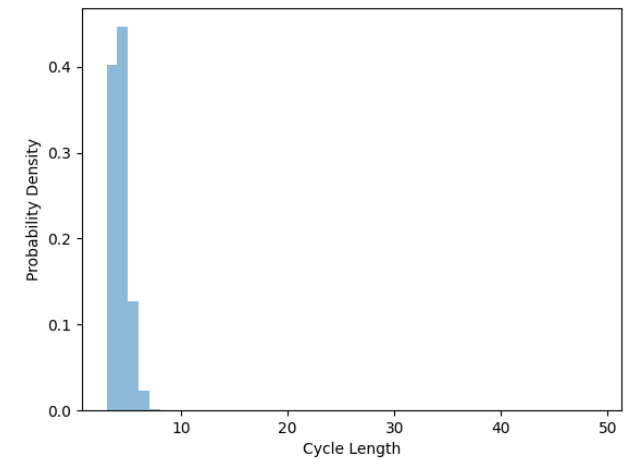

(e) 2000 nodes, 0.7 sparsity

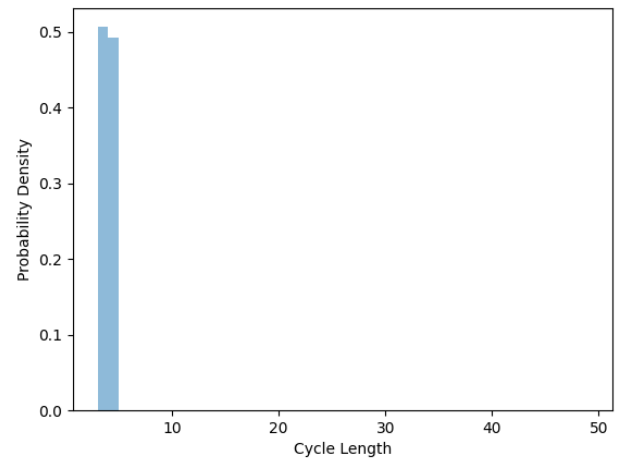

(f) 2000 nodes, 0.9 sparsity

**Figure 7.8.** Cycle length histograms of the cycle representatives output by the extended persistence algorithm on sampled Erdos-Renyi graphs

We perform an experiment to determine the cycle length distribution of cycle representatives output by our algorithm on random Erdos-Renyi graphs. We observe that, as the graph becomes more dense, the distribution of cycles shifts towards very short cycles. We also find that the cycle lengths for most Erdos-Renyi sparsity hyperparameters rarely become very long. We hypothesize that the cycle basis found by the extended persistence algorithm is close to the minimal (in cycle lengths) cycle basis.

For a given node count $n$, edge count $m$, and sparsity hyperparameter, $0 \leq s \leq 1$ which we define as the Erdos-Renyi probability for keeping an edge from a clique on $n$ nodes, we sample three Erdos-Renyi graphs. We collect the multiset of $m - n + 1$ cycle lengths in the cycle basis found by the algorithm. This multiset can be visualized as a histogram. Each histogram is a relative frequency mixture of the three cycle length histograms for each graph. See Figure 7.8 for the histograms we obtained from sampled Erdos-Renyi graphs. Notice that, even for 0.01 sparsity, Erdos-Renyi samples of graphs on 2000 nodes have the average cycle length of 15, which is 0.75% of $n = 2000$.

To put this in perspective, assume that we can relate the Erdos-Renyi sparsity $s$ by $\hat{s} := \frac{m}{n^2}$. For the datasets of our experiments, we have $\hat{s} \approx 0.009, 0.0048, 0.39, 0.062, 0.084, 0.0018, 0.032$,and 0.045 for DD, PROTEINS, IMDB-MULTI, MUTAG, PINWHEELS, 2CYCLES, MOLBACE, and MOLBBBP, respectively. The sparsity estimator is in the range of $0.0018 \leq \hat{s} \leq 0.39$, which tells us that most of the cycle lengths found by our algorithm are short.

## 7.14 Rational Hat Function Visualization

Figure 7.9 and Figure 7.10 visualize the rational hat function for fixed $r$ value and varying $x$ and $y$ values. Notice the boundedness of the plot as $(x, y) \to \infty$. For the theory behind the rational hat function, see [207].

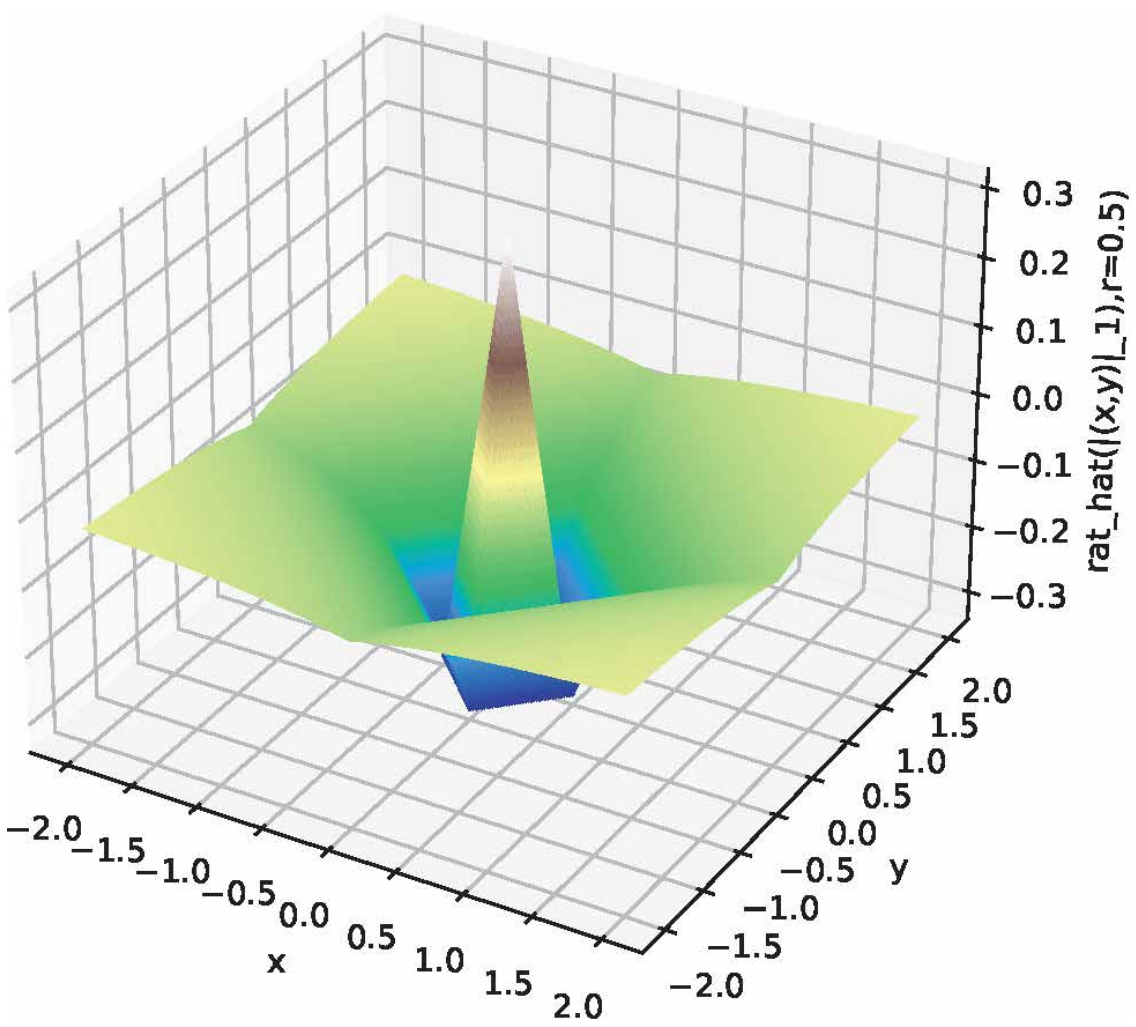

**Figure 7.9.** The function $\hat{r}$, output sliced at one dimension, as a function of $\|(x, y)\|_1$ with $r = 0.5$ from Equation 7.20. The point $(x, y)$ is given by $(x, y) = \mathbf{p} - \mathbf{c}$.

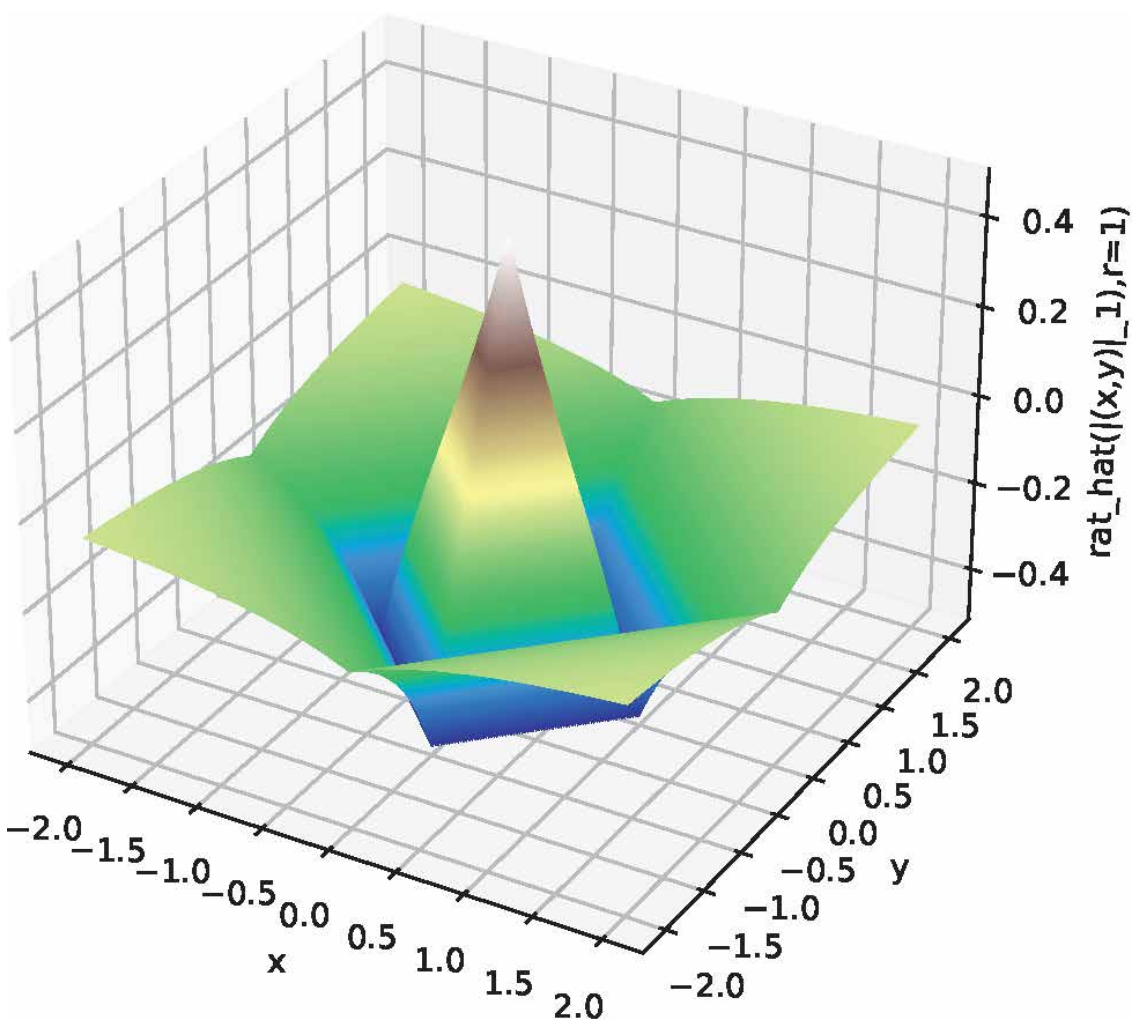

**Figure 7.10.** The function $\hat{r}$, output sliced at one dimension, as a function of $\|(x, y)\|_1$ with $r = 1.0$ from Equation 7.20. The point $(x, y)$ is given by $(x, y) = \mathbf{p} - \mathbf{c}$.

## 7.15 Datasets and Hyperparameter Information

Here are the datasets, both synthetic and real world, used in all of our experiments along with training hyperparameter information.

The barcode vectorization layer, or concatenation of four-rational hat functions, is set to a dimension of 256. The LSTM used on the explicit cycle representatives was set to a 2-layer bidirectional LSTM with single channel inputs and 256 dimensional vector representations. Due to the fact that our algorithm on random Erdos-Renyi graphs rarely encounters long cycles, we set the LSTM layers to a small number like 2 to avoid overfitting.

**Table 7.3.** Dataset statistics and training hyperparameters used for all datasets in scoring experiments of Table 7.1 and Table 7.2

| Dataset and Hyperparameter Information | | | | | | | | |
|---|---|---|---|---|---|---|---|---|
| Dataset | Graphs | Classes | Avg. Vertices | Avg. Edges | lr | Node Attrs.(Y/N) | num. layers | Class ratio |
| DD | 1178 | 2 | 284.32 | 715.66 | 0.01 | Yes | 2 | 691/487 |
| PROTEINS | 1113 | 2 | 39.06 | 72.82 | 0.01 | Yes | 2 | 663/422 |
| IMDB-MULTI | 1500 | 3 | 13.00 | 65.94 | 0.01 | No | 2 | 500/500/500 |
| MUTAG | 188 | 2 | 17.93 | 19.79 | 0.01 | Yes | 1 | 63/125 |
| PINWHEELS | 100 | 2 | 71.934 | 437.604 | 0.01 | No | 2 | 50/50 |
| 2CYCLES | 400 | 2 | 551.26 | 551.26 | 0.01 | No | 2 | 200/200 |
| MOLBACE | 1513 | 2 | 34.09 | 36.9 | 0.001 | Yes | 2 | 822/691 |
| MOLBBBP | 2039 | 2 | 24.06 | 25.95 | 0.001 | Yes | 2 | 479/1560 |

## 7.16 Implementation Dependencies

Our experiments have the following dependencies: python 3.9.1, torch 1.10.1, torch_geometric 2.0.5, torch_scatter 2.0.9, torch_sparse 0.6.13, scipy 1.6.3, numpy 1.21.2, CUDA 11.2, GCC 7.5.0.

## 7.17 Visualization of Graph Filtrations

We visualize the filtration functions $f_G$ learned on graphs $G$ for the datasets: IMDB-MULTI, MUTAG, and REDDIT-BINARY. The value of $f_G(v)$ for each $v \in V$ is shown in each figure.

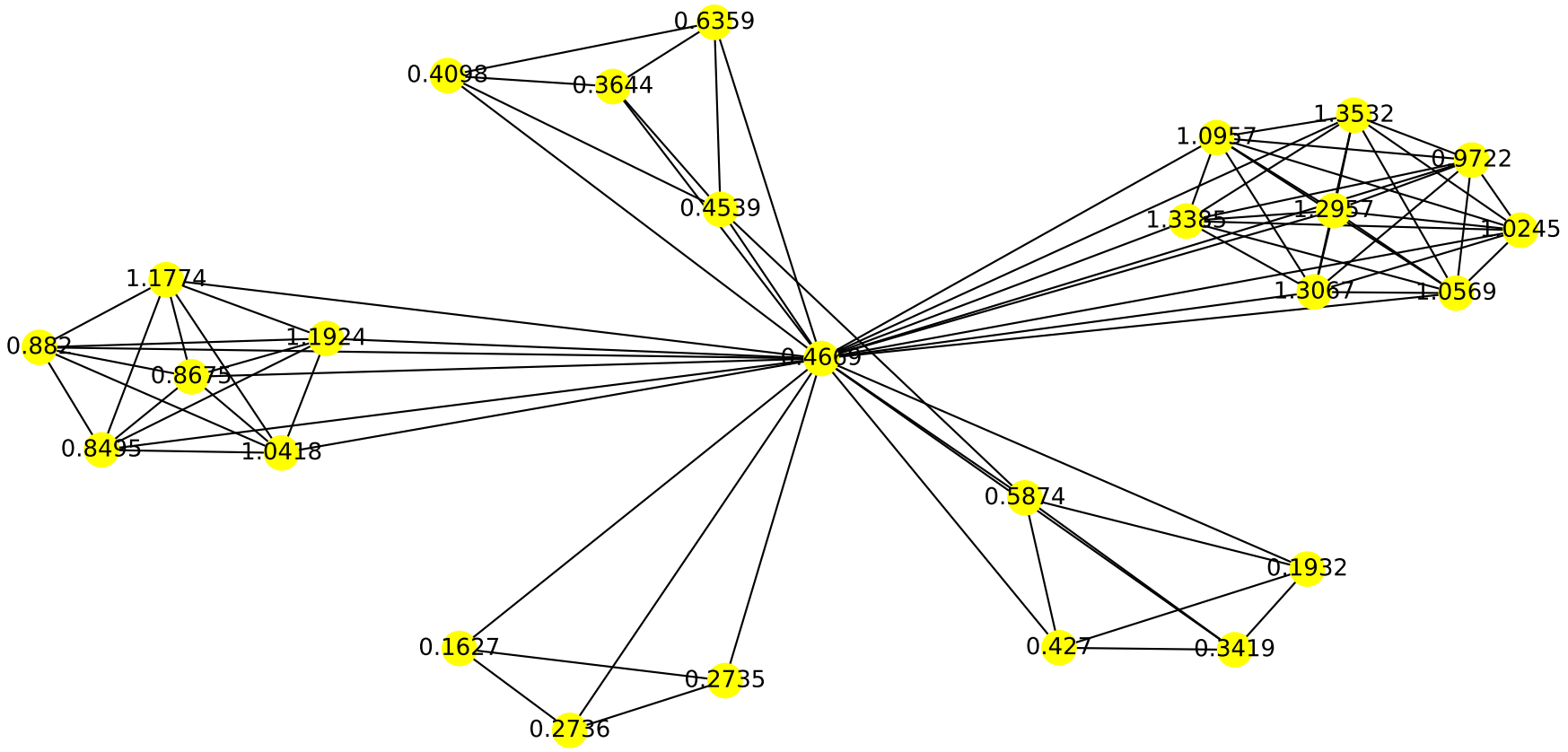

**Figure 7.11.** IMDB-MULTI learned filtration function

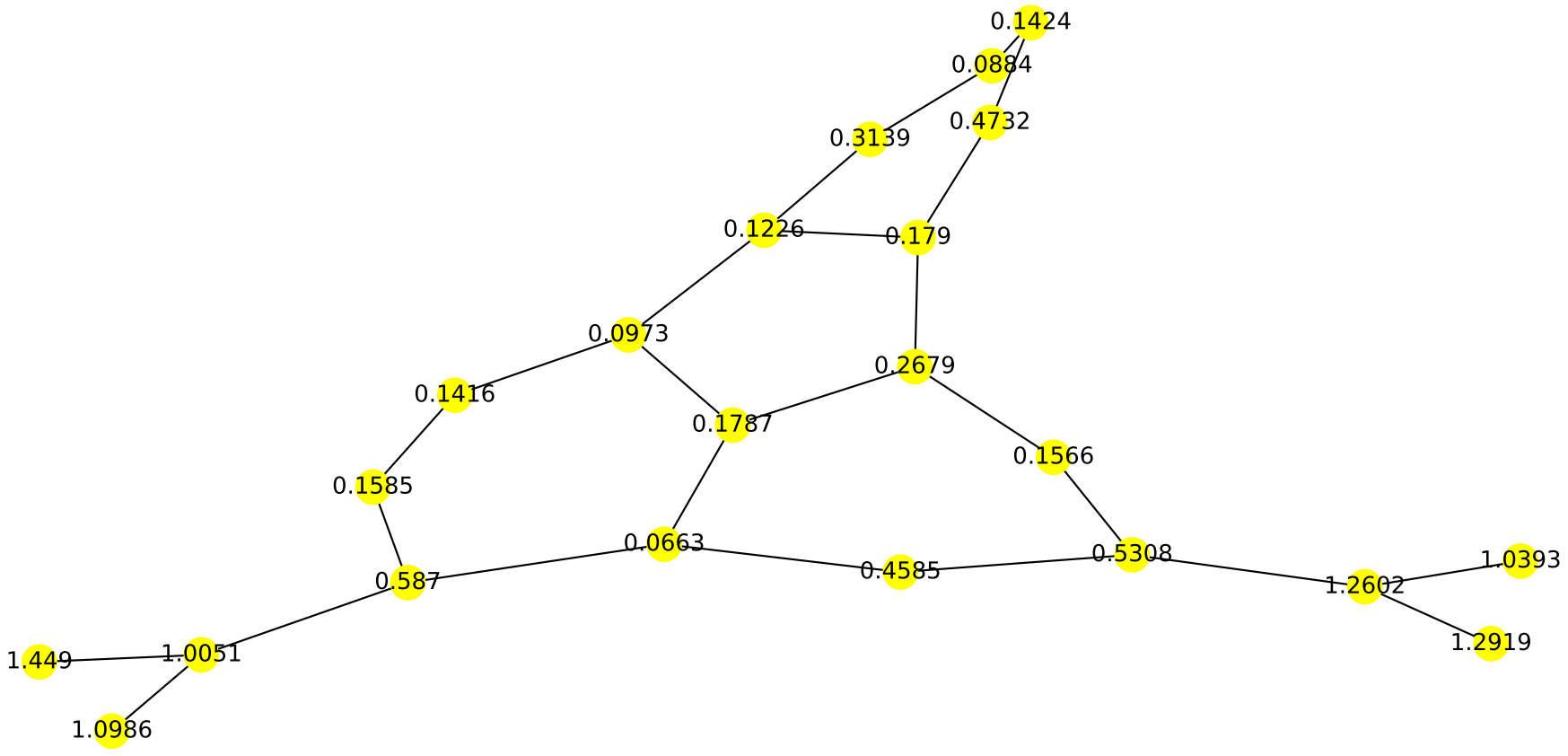

**Figure 7.12.** MUTAG learned filtration function

## 7.18 Additional Experiments

### 7.18.1 Number of Convolutional Layers Experiment

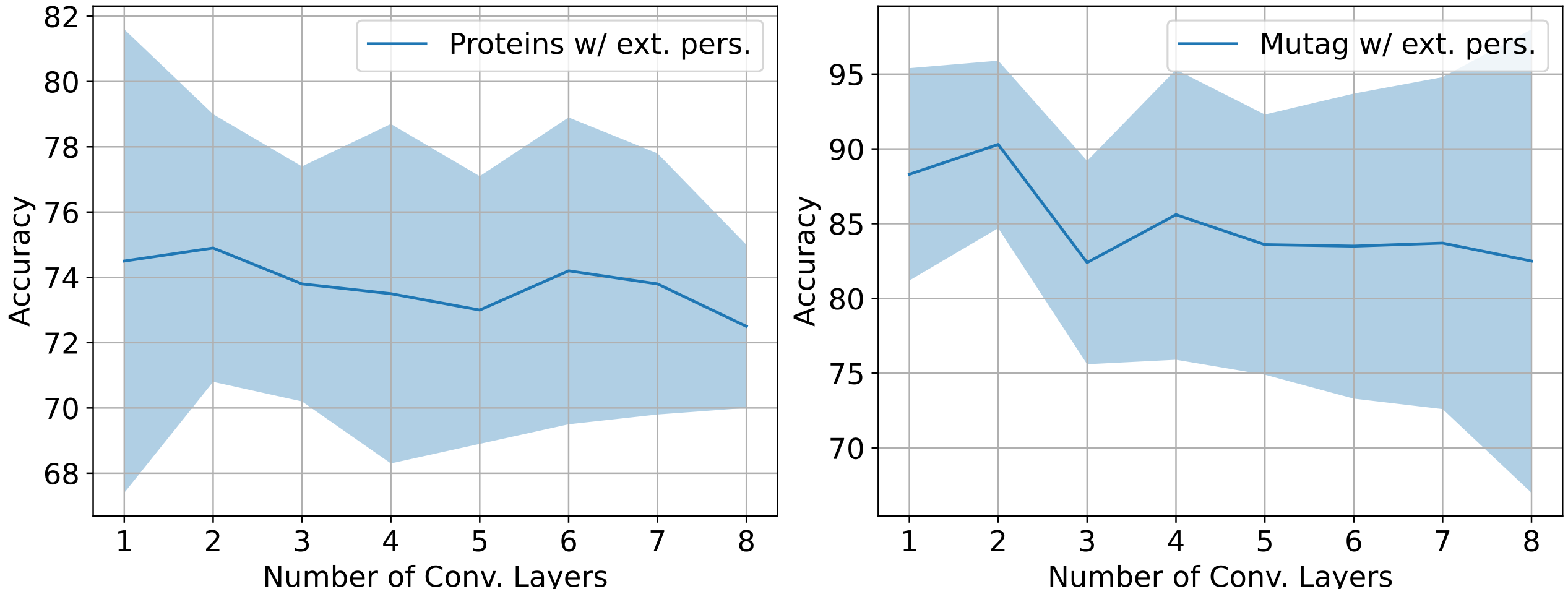

(a) PROTEINS (b) MUTAG

**Figure 7.13.** An exhibit of oversmoothing in the filtration convolutional layers. Plot of the average accuracy with std. dev. as a function of the number of convolutional layers before the Jumping Knowledge MLP and the extended persistence readout. The PROTEINS and MUTAG datasets were used in (a) and (b) respectively.

We also perform an experiment to determine the number of layers in the MPGNN of the filtration function that has the highest performance. Due to oversmoothing [265], which is exacerbated by the required scalar-dimensional vertex embeddings, as we increase the number of layers for the filtration function the performance drops. See Figure 7.13 for an illustration of this phenomenon on the PROTEINS and MUTAG dataset. For these two datasets, two layers perform the best.

### 7.18.2 Macro F1 Experiments

We perform further experiments with the Mutag dataset and evaluate performance with the F1 Macro score due to the class imbalance in the Mutag dataset, see Table 7.3. See Table 7.4 for details of the F1 Macro evaluation.

**Table 7.4.** Mutag Macro F1 Scores

| Model | F1 Macro |
|---|---|
| GEFL-bars | 84.0±14.2 |
| GEFL-bars+cycles | **85.4±10.3** |
| GCN | 69.8 ±15.2 |
| GraphSAGE | 69.5±11.0 |
| GIN0 | 79.9±9.0 |
| GIN | **84.9±6.0** |

### 7.18.3 Experiments on Transferability

We perform an experiment on the transferability of our approach from real world datasets to their edge corrupted versions and compare with the standard GNN baselines. We form a corrupted version of the data by applying a Bernoulli random variable with $p = 0.2$ on each edge of the Mutag and Imdb-multi datasets. We perform zero-shot, one-shot and 10-shot transfer learning. We include an ablation study to compare with models without fine-tuning and with $0, 1, 10$ and 100 epochs of training.

We notice that our model does not directly transfer in either zero or one shot transfer learning. However after 10 epochs of fine-tuning, there appears to be an advantage to transfer learning over raw training. For the Mutag dataset, there is a 77.7 – 64.2 average Macro F1 score difference for a bars only representation model and a 74.8 – 69.8 average Macro F1 score difference for bars with cycle representatives. Furthermore, on Mutag the baseline GNNs do not show any improvement with 10-shot transfer learning. For Imdb-multi there is a 48.3 – 46.1 average accuracy difference for bars only representation model and a 49.7 – 48.0 average accuracy difference for the bars with cycle representatives model. As for Mutag, for Imdb-multi there is no transfer learning improvement for the baselines.

We hypothesize that both versions of our model transfer well since our model captures global topological information with low variance. This is more conducive to transfer learning than an aggregation of each node's local neighborhood information as in the baselines. The baselines result in training a higher variance classifier. Low variance classifiers transfer easier since their decision boundary is easier to adapt to new data. We also hypothesize that the bars only model transfers better than the bars with cycle representatives model due to its even lower variance, no LSTM parameters.

Our model downsamples the graph representation into scalars (a set of sequences of scalars and a set of pairs of scalars) upon computing bars and cycle representatives. Due to our architecture, to accommodate these scalars, this results in a higher bias classifier with few parameters. Due to the bias variance trade off, the variance should be low.

**Table 7.5.** We list here the scores for our transfer learning experiments on Mutag and Imdb-multi. We pretrain on the original Mutag and Imdb-multi by 10-fold cross validation. Then, we fine-tune the pretrained models on corrupted versions of these datasets also by 10-fold cross validation. These two corrupted datasets are obtained by filtering by a Bernoulli random variable of $p = 0.2$ on each edge. This is very likely to introduce different cycle patterns in the data. Fine-tuning is performed for 0, 1 and 10 epochs, while '–' denotes no finetuning.

| Transfer learning results | | | | | | | | |
|---|---|---|---|---|---|---|---|---|
| Dataset | Pretraining (Epochs) | Finetuning (Epochs) | Ours+bars | Ours+bars+cycles | GCN | GraphSage | GIN0 | GIN |
| Mutag-0.2 (F1 Macro) | 100 | – | 83.2±9.6 | 78.3±17.3 | 73.4±10.4 | 74.5±09.1 | 85.6±05.7 | 83.5±11.2 |
| Mutag-0.2 (F1 Macro) | 100 | 0 | 31.4±11.2 | 29.4±11.4 | 32.4±7.9 | 39.9±0.9 | 35.4±7.3 | 35.4±7.3 |
| Mutag-0.2 (F1 Macro) | 0 | – | 37.1 ± 12.2 | 32.7 ± 7.9 | 34.6 ± 8.7 | 36.1 ± 7.8 | 35.8 ± 10.1 | 34.3 ± 7.1 |
| Mutag-0.2 (F1 Macro) | 100 | 1 | 39.9±0.9 | 39.9±0.9 | 39.9±0.9 | 39.9±0.9 | 39.9±0.9 | 39.9±0.9 |
| Mutag-0.2 (F1 Macro) | 1 | – | 39.9 ± 0.9 | 39.9 ± 0.9 | 39.9 ± 0.9 | 39.9 ± 0.9 | 39.9 ± 0.9 | 39.9 ± 0.9 |
| Mutag-0.2 (F1 Macro) | 100 | 10 | 76.7±9.7 | 74.8±10.2 | 39.9±0.9 | 39.7±14.3 | 39.9±0.9 | 39.9±0.9 |
| Mutag-0.2 (F1 Macro) | 10 | – | 64.2 ± 17.9 | 69.8 ± 17.7 | 43.9 ± 9.5 | 44.2 ± 9.7 | 41.8 ± 6.2 | 44.1 ± 13.3 |
| Imdb-multi-0.2 (Acc.) | 100 | – | 49.2±3.9 | 50.3±2.9 | 49.4±2.5 | 50.5±2.6 | 49.6±3.1 | 51.5±3.1 |
| Imdb-multi-0.2 (Acc.) | 100 | 0 | 33.1±1.1 | 32.5±3.1 | 16.8±0.5 | 20.2±4.9 | 24.0±6.1 | 21.0±7.9 |
| Imdb-multi-0.2 (Acc.) | 0 | – | 34.5 ± 4.7 | 34.3 ± 2.4 | 20.2 ± 4.3 | 19.6 ± 4.3 | 20.4 ± 4.3 | 20.6 ± 5.0 |
| Imdb-multi-0.2 (Acc.) | 100 | 1 | 39.7±5.3 | 40.7±5.9 | 44.8±4.4 | 50.5±3.5 | 34.3±4.2 | 40.9±3.4 |
| Imdb-multi-0.2 (Acc.) | 1 | – | 34.3 ± 1.6 | 38.8 ± 4.9 | 47.2 ± 3.5 | 47.8 ± 3.3 | 43.2 ± 3.3 | 45.8 ± 5.5 |
| Imdb-multi-0.2 (Acc.) | 100 | 10 | 48.3±2.2 | 49.7±3.0 | 48.9±3.3 | 50.5±3.5 | 46.7±2.9 | 47.9±4.5 |
| Imdb-multi-0.2 (Acc.) | 10 | – | 46.1 ± 2.1 | 48.0 ± 3.1 | 50.1 ± 3.1 | 49.9 ± 1.9 | 51.5 ± 2.5 | 50.1 ± 2.5 |

# 8. EXPRESSIVE HIGHER-ORDER LINK PREDICTION THROUGH HYPERGRAPH SYMMETRY BREAKING

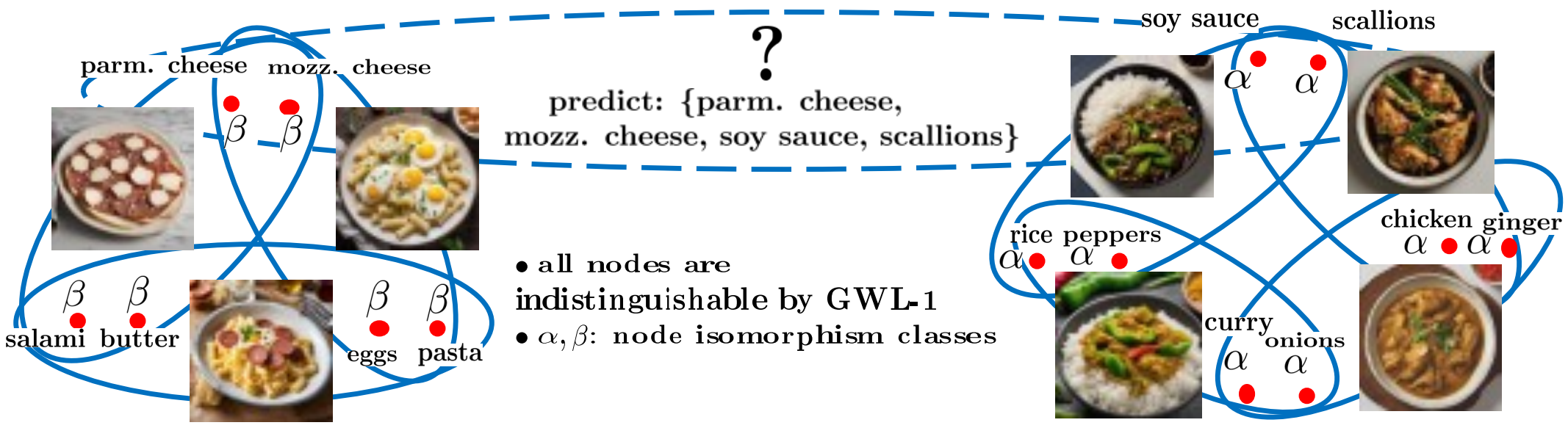

**Figure 8.1.** An illustration of a hypergraph of recipes. The nodes are the ingredients and the hyperedges are the recipes. The task of higher order link prediction is to predict hyperedges in the hypergraph. A negative hyperedge sample would be the dotted hyperedge. The Asian ingredient nodes ($\alpha$) and the European ingredient nodes ($\beta$) form two separate isomorphism classes. However, GWL-1 cannot distinguish between these classes and will predict a false positive for the negative sample.

A hypergraph consists of a set of nodes along with a collection of subsets of the nodes called hyperedges. Higher order link prediction is the task of predicting the existence of a missing hyperedge in a hypergraph. A hyperedge representation learned for higher order link prediction is fully expressive when it does not lose distinguishing power up to an isomorphism. Many existing hypergraph representation learners, are bounded in expressive power by the Generalized Weisfeiler Lehman-1 (GWL-1) algorithm, a generalization of the Weisfeiler Lehman-1 (WL-1) algorithm. The WL-1 algorithm can approximately decide whether two graphs are isomorphic. However, GWL-1 has limited expressive power. In fact, GWL-1 can only view the hypergraph as a collection of trees rooted at each of the nodes in the hypergraph. Furthermore, message passing on hypergraphs can already be computationally expensive, particularly with limited GPU device memory. To address these limitations, we devise a preprocessing algorithm that can identify certain regular subhypergraphs exhibiting symmetry with respect to GWL-1. Our preprocessing algorithm runs once with the time complexity linear in the size of the input hypergraph. During training, we randomly drop the hyperedges of the subhypergraphs identifed by the algorithm and add covering hyperedges to break symmetry. We show that our method improves the expressivity of GWL-1. Our extensive experiments [1] also demonstrate the effectiveness of our approach for higher-order link prediction on both graph and hypergraph datasets with negligible change in computation.

## 8.1 Introduction

Hypergraphs can model complex relationships in real-world networks, extending beyond the pairwise connections captured by traditional graphs. Figure 8.1 is an example hypergraph consisting of recipes of two different types of dishes, which are largely determined by the ingredients to be used. In this hypergraph, the hyperedges are the recipes, and the nodes are the ingredients used in each recipe. The Asian recipes are presented in the right part of the figure, which consist of combinations of the ingredients of soy sauce, scallions, rice, peppers, chicken, ginger, curry and onions. The European recipes are presented in the left part of the figure, which consist of combinations of the ingredients of Parmesan cheese, mozzarella cheese, salami, butter, eggs, and pasta.

Hypergraphs have found applications in diverse fields such as recommender systems [266], visual classification [267], and social networks [268]. Higher-order link prediction is the task of predicting missing hyperedges in a hypergraph. For this task, when the hypergraph is unattributed, it is important to respect the hypergraph's symmetries, or automorphism group. This brings about challenges to learning an expressive view of the hypergraph.

A hypergraph neural network (hyperGNN) is any neural network that learns on a hypergraph. This is in analogy to a graph neural network (GNN), which is a neural network that learns on a graph. Many existing hyperGNN architectures follow a computational message passing model called the Generalized Weisfeiler Lehman-1 (GWL-1) algorithm [269], a hypergraph isomorphism testing approximation algorithm. GWL-1 is a generalization of the message passing algorithm called Weisfeiler Lehman-1 (WL-1) algorithm [212] used for graph isomorphism testing.

GWL-1, like WL-1 on graphs, is limited in how well it can express its input. Specifically, by viewing a hypergraph as a collection of rooted trees, GWL-1 loses topological information of its input hypergraph and thus cannot fully recognize the symmetries, or automorphism group, of the hypergraph. In fact, the hyperGNN views the hypergraph as having false-positive symmetries. For a task like transductive hyperlink prediction, this can result in predicting false-positive hyperlinks as shown in Figure 8.1. Furthermore, such an issue can become even worse during test time since the automorphism group of the hypergraph might change. It is thus important to find a way to improve the expressivity of existing hyperGNNs.

[1] ↑https://github.com/simonzhang00/HypergraphSymmetryBreaking/

Let a hypergraph $\mathcal{H} = (\mathcal{V}, \mathcal{E})$ denote a pair where $\mathcal{V}$ is a set of nodes and $\mathcal{E} \subseteq 2^{\mathcal{V}}$, a collection of subsets of $\mathcal{V}$, indexes a set of hyperedges. GWL-1 views a hypergraph as a collection of trees rooted at the nodes. These rooted trees are formed by viewing each node-hyperedge incidence as an edge and recursively expanding about the nodes and hyperedges alternately. We can recover hyperGNNs by expressing the computation of GWL-1 as a matrix equation. Parameterizing the expression with learnable weights, GWL-1 becomes a neural network, called a hypergraph neural network (hyperGNN), similar to graph neural networks (GNN)s. In practice this is implemented through repeated sparse matrix multiplication.

Computing on a hypergraph can also be very expensive. The subsets $\mathrm{e} \in \mathcal{E}$ that contain a node $v \in \mathcal{V}$ form the neighborhood of $v \in \mathcal{V}$. This means just the neighborhood size of the nodes in hypergraphs can grow exponentially with the number of nodes of the hypergraph. Thus, a computationally more expensive message passing scheme over GWL-1 based hyperGNNs may bring difficulties.

In order to address the issue of the expressivity of hyperGNNS for the task of hyperlink prediction while also respecting the computational complexity of computing on a hypergraph, we devise a method that selectively breaks the symmetry of the hypergraph topology itself coming from the limitations of the hyperGNN architecture. Our method is designed as an efficient preprocessing algorithm that can improve the expressive power of GWL-1 for higher order link prediction. Since the preprocessing only runs once with complexity linear in the input, we do not increase the computational complexity of training.

Similar to a substructure counting algorithm [270], we identify certain symmetries in induced subhypergraphs. However, unlike in existing work where node attributes are modified, such as random noise being appended to the node attributes [271], we directly target and modify the symmetries in the topology. This limits the space for augmentation, which can prevent extreme perturbations of the data. The algorithm identifies a cover of the hypergraph by disjoint connected components whose nodes are indistinguishable by GWL-1. During training, we randomly replace the hyperedges of the identified symmetric regular induced subhypergraphs with single hyperedges that cover the nodes of each subhypergraph. We show that our method of hyperedge hallucination to break symmetry can increase the expressivity of existing hypergraph neural networks both theoretically and experimentally.

**Contributions.** In the context of hypergraph representation learning and hyperlink prediction, we have a method that can break the symmetries introduced by conventional hypergraph neural networks. Conventional hypergraph neural networks are based on the GWL-1 algorithm on hypergraphs. However, the GWL-1 algorithm on hypergraphs views the hypergraph as a collection of rooted trees. This introduces false positive symmetries. We summarize our contributions in this work as follows:

- **Provide a formal analysis of GWL-1 on hypergraphs** from the perspective of algebraic topology. Our analysis offers a novel characterization of the expressive power and limitations of GWL-1. By leveraging concepts from algebraic topology, we establish a precise connection between GWL-1 and the universal covers of hypergraphs, providing deeper insights into the algorithm's behavior on complex hypergraph structures.

- **Devise an efficient hypergraph preprocessing algorithm** to identify false positive symmetries of GWL-1. We propose a linear time preprocessing algorithm to identify specific regular subhypergraphs that exhibit symmetry with respect to GWL-1, which are potential sources of expressivity limitations.

- **Introduce a data augmentation scheme** that leverages the preprocessing algorithm's output to improve GWL-1's expressivity. During training, we randomly modify the hyperedges of the identified symmetric subhypergraphs, effectively breaking symmetries that GWL-1 cannot distinguish. This approach enhances the model's ability to capture fine-grained structural information without significantly increasing computational complexity.

- **Provide formal analysis and performance guarantees** for our method. We rigorously prove how our approach improves the expressivity of GWL-1 under certain conditions. These theoretical results offer valuable insights into the circumstances under which our method can enhance hypergraph representation learning, providing a solid foundation for its practical application.

- **Perform extensive experiments on real-world datasets** to demonstrate the effectiveness of our approach. Our comprehensive evaluation spans various hypergraph and graph datasets, showcasing consistent improvements across different hypergraph neural network architectures for higher-order link prediction tasks. These empirical results validate the practical utility of our method and its ability to enhance existing models with minimal computational overhead.

## 8.2 Background

The following notation is used throughout the paper:

Let $\mathbb{N} \triangleq \{0,1,...\}, \mathbb{Z} \triangleq \{...,-1,0,1,...\}, \mathbb{Z}^+ \triangleq \{1,...\}$, and $\mathbb{R}$ denote the naturals, integers, positive integers, and real numbers respectively. Let $[n] \triangleq \{1,...,n\} \subseteq \mathbb{Z}^+$ denote the integers from 1 to $n \in \mathbb{Z}^+$.

For a set $A$, let $\binom{A}{k}$ denote the set of all subsets of $A$ of size $k$. Given the set $A$, a multiset is defined by $\tilde{A} \triangleq (A, m)$, $m : A \to \mathbb{Z}^+$. A set is also a multiset with $m = \mathbf{1}$. A submultiset $\tilde{B} \subseteq \tilde{A}$ is defined by $\tilde{B} \triangleq (B, m')$ with $B \subseteq A$ and $m'(\mathrm{e}) \leq m(\mathrm{e}), \forall \mathrm{e} \in B$. A multiset with its elements explicitly enumerated with the double curly brace notation: $\{\{a, a, a, ...\}\}$. The cardinality of a multiset $\tilde{A}$, is defined as $|\tilde{A}| \triangleq \sum_{\mathrm{e} \in A} m(\mathrm{e})$. For two multisets $\tilde{A} = (A, m_A), \tilde{B} = (B, m_B)$, we may define their multiset sum by $\tilde{A} \sqcup \tilde{B} \triangleq (A \cup B, m_{A \cup B} = m_A + m_B)$.

For two pairs of multisets $\tilde{A} = (\tilde{A}_1, \tilde{A}_2), \tilde{B} = (\tilde{B}_1, \tilde{B}_2)$, denote their multiset sum by their elementwise multiset sum: $\tilde{A} \sqcup \tilde{B} \triangleq (\tilde{A}_1 \sqcup \tilde{B}_1, \tilde{A}_2 \sqcup \tilde{B}_2)$. Similarly, for two pairs of sets $A = (A_1, A_2), B = (B_1, B_2)$, denote their union by their elementwise union: $A \cup B \triangleq (A_1 \cup B_1, A_2 \cup B_2)$.

Let $P_t \triangleq P(\bullet; t)$ denote a probability distribution parameterized by $t \in \mathbb{R}$. The distribution $P_t$ has some domain $\mathcal{D}$, which we denote by $\mathrm{dom}(P_t)$. The notation $\mathrm{supp}(P_t) \triangleq \{x \in \mathrm{dom}(P_t) : P_t(x) > 0\}$ denotes the **support** of a distribution $P_t$.

A Bernoulli distribution $P_p \triangleq P(X; p)$ is a probability distribution parameterized by a $p : 0 < p < 1$ with the following definition:

$$P_p(X = k) = \begin{cases} p & k = 1 \\ 1 - p & k = 0 \end{cases} \tag{8.1}$$

The random variable $Bernoulli(p) \sim P_p$ is called a Bernoulli random variable.

### 8.2.1 Group Theory

To better study hypergraphs, we use some concepts of group theory. Here we give a brief introduction to some of them. For more details, see [16].

**Definition 8.2.2.** *A* group $G$ *is a set equipped with a binary operation "$*$" that satisfies the following four properties:*

1. ***Closure**: For every $a, b \in G$, the result of the operation $a * b$ is also in $G$.*
2. ***Associativity**: For every $a, b, c \in G$, $(a * b) * c = a * (b * c)$.*
3. ***Identity Element**: There exists an element $\mathrm{e} \in G$ such that for every $a \in G$, $\mathrm{e} * a = a * \mathrm{e} = a$.*
4. ***Inverse Element**: For each $a \in G$, there exists an element $b \in G$ such that $a * b = b * a = \mathrm{e}$ (such element $b$ is unique for $a$ and is often denoted as $a^{-1}$.)*

#### Permutation Groups

A *permutation group* is a group where the elements are permutations of a set, and the group operation is the composition of these permutations. Permutations are bijective functions that rearrange the elements of a set.

**Group Isomorphism and Automorphism**

Two groups $G$ and $H$ are called *isomorphic* if there exists a bijective function $\phi : G \to H$ such that for all $a, b \in G, \phi(a * b) = \phi(a) * \phi(b)$. This means that $G$ and $H$ have the same group structure, even if their elements are different. Such bijective funtions are called *isomorphisms.* If $G = H$, an isomorphism to a group itself is called an *automorphism.* The set of all automorphisms of a group $G$ forms a group, with group operations given by compositions. Such group is called the *automorphism group* of $G$, denoted as $Aut(G)$.

**Group Action**

A group action is a formal way in which a group $G$ operates on a set $X$. Formally, a group action is a map $G \times X \to X$ (denoted $(g, x) \mapsto g \cdot x$) that satisfies the following two properties:

1. For all $x \in X$, $\mathrm{e} \cdot x = x$ where e is the identity element in $G$.
2. For all $g, h \in G$ and $x \in X$, $(gh) \cdot x = g \cdot (h \cdot x)$.

Group actions are useful for studying the symmetry of objects, as they describe how the elements of a group move or transform the elements of a set.

**Stabilizer**

In the context of group actions, the stabilizer of an element $x$ in a set $X$ (where a group $G$ acts on $X$) is the set of elements in $G$ that leave $x$ fixed. Formally, the stabilizer of $x$ in $G$ is given by:

$$Stab_G(x) = \{g \in G : g \cdot x = x\}$$

### 8.2.3 Isomorphisms on Higher Order Structures

In this section, we go over what a hypergraph is and how this structure is represented as tensors. We then define what a hypergraph isomorphism is.

A hypergraph is a generalization of a graph. Hypergraphs allow for all possible subsets over a set of vertices, called hyperedges. We can thus formally define a hypergraph as:

**Definition 8.2.4.** *An undirected hypergraph is a pair* $\mathcal{H} = (\mathcal{V}, \mathcal{E})$ *consisting of a set of vertices* $\mathcal{V}$ *and a set of hyperedges* $\mathcal{E} \subseteq 2^{\mathcal{V}}$ *where* $2^{\mathcal{V}}$ *is the power set of the vertex set* $\mathcal{V}$*.*

For a given hypergraph $\mathcal{H} = (\mathcal{V}, \mathcal{E})$, a hypergraph $\mathcal{G} = (\mathcal{V}', \mathcal{E}')$ is a **subhypergraph** of $\mathcal{H}$ if $\mathcal{V}' \subseteq \mathcal{V}$ and $\mathcal{E}' \subseteq \mathcal{E}$.

A subhypergraph **induced** by $\mathcal{W} \subseteq \mathcal{V}$ is defined as $(\mathcal{W}, \mathcal{F} = 2^{\mathcal{W}} \cap \mathcal{E})$. Similarly, for a subset of hyperedges $\mathcal{F} \subseteq \mathcal{E}$, a subhypergraph **induced** by $\mathcal{F}$ is defined as $(\bigcup_{\mathrm{e} \in \mathcal{F}} \mathrm{e}, \mathcal{F})$

For a given hypergraph $\mathcal{H}$, we also use $\mathcal{V}_{\mathcal{H}}$ and $\mathcal{E}_{\mathcal{H}}$ to denote the sets of vertices and hyperedges of $\mathcal{H}$ respectively. According to the definition, a hyperedge is a nonempty subset of the vertices. A hypergraph with all hyperedges the same size $d$ is called a $d$-uniform hypergraph. A 2-uniform hypergraph is an undirected graph, or just graph.

When viewed combinatorially, a hypergraph can include some symmetries that are captured by isomorphisms. These isomorphisms are defined by bijective structure preserving maps.

**Definition 8.2.5.** *For two hypergraphs* $\mathcal{H}$ *and* $\mathcal{D}$*, a structure preserving map* $\rho : \mathcal{H} \to \mathcal{D}$ *is a pair of maps* $\rho = (\rho_{\mathcal{V}} : \mathcal{V}_{\mathcal{H}} \to \mathcal{V}_{\mathcal{D}}, \rho_{\mathcal{E}} : \mathcal{E}_{\mathcal{H}} \to \mathcal{E}_{\mathcal{D}})$ *such that* $\forall \mathrm{e} \in \mathcal{E}_{\mathcal{H}}, \rho_{\mathcal{E}}(\mathrm{e}) \triangleq \{\rho_{\mathcal{V}}(v_{\mathrm{i}}) : v_{\mathrm{i}} \in \mathrm{e}\} \in \mathcal{E}_{\mathcal{D}}$*. A hypergraph isomorphism is a structure preserving map* $\rho = (\rho_{\mathcal{V}}, \rho_{\mathcal{E}})$ *such that both* $\rho_{\mathcal{V}}$ *and* $\rho_{\mathcal{E}}$ *are bijective. Two hypergraphs are said to be isomorphic, denoted as* $\mathcal{H} \cong \mathcal{D}$*, if there exists an isomorphism between them. When* $\mathcal{H} = \mathcal{D}$*, an isomorphism* $\rho$ *is called an automorphism on* $\mathcal{H}$*. All the automorphisms form a group, which we denote as* $Aut(\mathcal{H})$*.*

A graph isomorphism is the special case of a hypergraph isomorphism between 2-uniform hypergraphs according to Definition 8.2.5.

A *neighborhood* $N(v) \triangleq (\bigcup_{v \in \mathrm{e}} \mathrm{e}, \{\mathrm{e} : v \in \mathrm{e}\})$ of a node $v \in \mathcal{V}$ of a hypergraph $\mathcal{H} = (\mathcal{V}, \mathcal{E})$ is the subhypergraph of $\mathcal{H}$ induced by the set of all hyperedges incident to $v$. The *degree* of $v$ is denoted $\deg(v) = |\mathcal{E}_{N(v)}|$. A simple but very common symmetric hypergraph is of importance to our task, namely the neighborhood-regular hypergraph, or just regular hypergraph.

**Definition 8.2.6.** *A neighborhood-regular hypergraph is a hypergraph where all neighborhoods of each node are isomorphic to each other.*

A $d$-uniform neighborhood of $v$ is the set of all hyperedges of size $d$ in the neighborhood of $v$. Thus, in a neighborhood-regular hypergraph, all nodes have their $d$-uniform neighborhoods of the same cardinality for all $d \in \mathbb{N}$.

**Representing Higher Order Structures as Tensors** : There are many data stuctures one can define on a higher order structure like a hypergraph. An $n$-order tensor [272], as a generalization of an adjacency matrix on graphs can be used to characterize the higher order connectivities. For simplicial complexes, which are hypergraphs where all subsets of a hyperedge are also hyperedges, a Hasse diagram, which is a multipartite graph induced by the poset relation of subset amongst hyperedges, or simplices, differing in exactly one node, is a common data structure [26]. Similarly, the star expansion matrix [273] can be used to characterize hypergraphs up to isomorphism.

In order to define the star expansion matrix, we define the star expansion bipartite graph.

**Definition 8.2.7** (star expansion bipartite graph). *Given a hypergraph $\mathcal{H} = (\mathcal{V}, \mathcal{E})$, the* star expansion bipartite graph $\mathcal{B}_{\mathcal{V},\mathcal{E}}$ *is the bipartite graph with vertices $\mathcal{V} \bigsqcup \mathcal{E}$ and edges* $\{(v, \mathrm{e}) \in \mathcal{V} \times \mathcal{E} : v \in \mathrm{e}\}$.

**Definition 8.2.8.** *The star expansion incidence matrix $H$ of a hypergraph $\mathcal{H} = (\mathcal{V}, \mathcal{E})$ is the* $|\mathcal{V}| \times 2^{|\mathcal{V}|}$ 0-1 *incidence matrix $H$ where $H_{v,\mathrm{e}} = 1$ iff $v \in \mathrm{e}$ for $(v, \mathrm{e}) \in \mathcal{V} \times \mathcal{E}$ for some fixed orderings on both $\mathcal{V}$ and $2^{\mathcal{V}}$.*

In practice, as data to machine learning algorithms, the matrix $H$ is sparsely represented by its nonzero entries.

To study the symmetries of a given hypergraph $\mathcal{H} = (\mathcal{V}, \mathcal{E})$, we consider the permutation group on the vertices $\mathcal{V}$, denoted as $\mathrm{Sym}(\mathcal{V})$, which acts jointly on the rows and columns of star expansion adjacency matrices. We assume the rows and columns of a star expansion adjacency matrix have some canonical ordering, say lexicographic ordering, given by some prefixed ordering of the vertices. Therefore, each hypergraph $\mathcal{H}$ has a unique canonical matrix representation $H$.

We define the action of a permutation $\pi \in \mathrm{Sym}(\mathcal{V})$ on a star expansion adjacency matrix $H$:

$$(\pi \cdot H)_{v,\mathrm{e}=(u_1 \dots v \dots u_k)} \triangleq H_{\pi^{-1}(v),\pi^{-1}(\mathrm{e})=(\pi^{-1}(u_1)\dots\pi^{-1}(v)\dots\pi^{-1}(u_k))} \tag{8.2}$$

Based on the group action, consider the stabilizer subgroup of $\mathrm{Sym}(\mathcal{V})$ on an incidence matrix $H$:

$$Stab_{\mathrm{Sym}(\mathcal{V})}(H) = \{\pi \in \mathrm{Sym}(\mathcal{V}) \mid \pi \cdot H = H\} \tag{8.3}$$

For simplicity we omit the lower index of $\mathrm{Sym}(\mathcal{V})$ when the permutation group is clear from context. It can be checked that $Stab(H) \subseteq \mathrm{Sym}(\mathcal{V})$ is a subgroup. Intuitively, $Stab(H)$ consists of all permutations that fix $H$. These are equivalent to hypergraph automorphisms on the original hypergraph $\mathcal{H}$.

**Proposition 8.2.9.** *$Aut(\mathcal{H}) \cong Stab(H)$ are equivalent as isomorphic groups.*

We can also define a notion of isomorphism between $k$-node sets using the stabilizers on $H$.

**Definition 8.2.10.** *For a given hypergraph $\mathcal{H}$ with star expansion matrix $H$, two $k$-node sets $S, T \subseteq \mathcal{V}$ are called* isomorphic*, denoted as $S \simeq T$, if $\exists \pi \in Stab(H), \pi(S) = T$ and $\pi(T) = S$.*

Such isomorphism is an equivalance relation on $k$-node sets. When $k = 1$, we have isomorphic nodes, denoted $u \cong_{\mathcal{H}} v$ for $u, v \in \mathcal{V}$. Node isomorphism is also studied as the so-called structural equivalence in [274]. Furthermore, when $S \simeq T$ we can then say that there is a matching amongst the nodes in sets $S$ and $T$ so that matched nodes are isomorphic.

### 8.2.11 Invariance and Expressivity

For a given hypergraph $\mathcal{H} = (\mathcal{V}, \mathcal{E})$, we want to do hyperedge prediction on $\mathcal{H}$, which is to predict missing hyperedges from $k$-node sets for $k \geq 2$. Let $|\mathcal{V}| = n$, $|\mathcal{E}| = m$, and $H \in \mathbb{Z}_2^{n \times 2^n}$ be the star expansion adjacency matrix of $\mathcal{H}$. To do hyperedge prediction, we study $k$-node representations $g : \binom{\mathcal{V}}{k} \times \mathbb{Z}_2^{n \times 2^n} \to \mathbb{R}^d$ that map $k$-node sets of hypergraphs to $d$-dimensional Euclidean space. Ideally, we want a most-expressive $k$-node representation for hyperedge prediction, which is intuitively a $k$-node representation that is injective on $k$-node set isomorphism classes from $\mathcal{H}$. We break up the definition of most-expressive $k$-node representation into possessing two properties, as follows:

**Definition 8.2.12.** *Let $g : \binom{\mathcal{V}}{k} \times \mathbb{Z}_2^{n\times 2^n} \to \mathbb{R}^d$ be a $k$-node representation on a hypergraph $\mathcal{H}$. Let $H \in \mathbb{Z}_2^{n\times 2^n}$ be the star expansion adjacency matrix of $\mathcal{H}$ for $n$ nodes. The representation $g$ is $k$-node most expressive if $\forall S, S' \subseteq \mathcal{V}$, $|S| = |S'| = k$, the following two conditions are satisfied:*

1. *$g$ is **k-node invariant**: $\exists \pi \in Stab(H), \pi(S) = S' \implies g(S,H) = g(S',H)$*

2. *$g$ is **k-node expressive** $\nexists \pi \in Stab(H), \pi(S) = S' \implies g(S,H) \neq g(S',H)$*

The first condition of a most expressive $k$-node representation states that the representation must be well defined on the $k$ nodes up to isomorphism. The second condition requires the injectivity of our representation. These two conditions mean that the representation does not lose any information when doing prediction for missing $k$-sized hyperedges on a set of $k$ nodes.

We can also define the symmetry group of a $k$-node representation map $g$ on $H$ as the set of all permutations on $\mathcal{V}$ that make the representation map $g$ $k$-node invariant. This is formally defined below:

**Definition 8.2.13.** *For $g : \binom{\mathcal{V}}{k} \times \mathbb{Z}_2^{n\times 2^n} \to \mathbb{R}^d$ a $k$-node representation on a hypergraph $\mathcal{H}$,*

$$Sym(g(H)) \triangleq \{\pi \in Sym(\mathcal{V}) : \forall S, S' \in \binom{\mathcal{V}}{k}, \pi(S) = S' \Rightarrow g(S,H) = g(S',H)\} \tag{8.4}$$

### 8.2.14 Generalized Weisfeiler-Lehman-1

We describe a generalized Weisfeiler-Lehman-1 (GWL-1) hypergraph isomorphism test similar to [269, 275] based on the WL-1 algorithm for graph isomorphism testing. There have been many parameterized variants of the GWL-1 algorithm implemented as neural networks, see Section 8.3.

Let $H$ be the star expansion matrix for a hypergraph $\mathcal{H}$. We define the GWL-1 algorithm as the following two step procedure on $H$ at iteration number $\mathrm{i} \geq 0$.

$$
\begin{aligned}
f_{\mathrm{e}}^{0} \leftarrow \{\}, h_{v}^{0} \leftarrow \{\} \\
f_{\mathrm{e}}^{\mathrm{i}+1} \leftarrow \{\!\{(f_{\mathrm{e}}^{\mathrm{i}}, h_{v}^{\mathrm{i}})\}\!\}_{v \in \mathrm{e}}, \forall \mathrm{e} \in \mathcal{E}_{\mathcal{H}}(H) \\
h_{v}^{\mathrm{i}+1} \leftarrow \{\!\{(h_{v}^{\mathrm{i}}, f_{\mathrm{e}}^{\mathrm{i}+1})\}\!\}_{v \in \mathrm{e}}, \forall v \in \mathcal{V}_{\mathcal{H}}(H)
\end{aligned}
\tag{8.5}
$$

This is slightly different from the algorithm presented in [269] at the $f_{\mathrm{e}}^{\mathrm{i}+1}$ update step. Our update step involves an edge representation $f_{\mathrm{e}}^{\mathrm{i}}$, which is not present in their version. Thus our version of GWL-1 is more expressive than that in [269]. However, they both possess some of the same issues that we identify. We denote $f_{\mathrm{e}}^{\mathrm{i}}(H)$ and $h_{v}^{\mathrm{i}}(H)$ as the hyperedge and node ith iteration GWL-1, called i-GWL-1, values on an unattributed hypergraph $\mathcal{H}$ with star expansion $H$. If GWL-1 is run to convergence then we omit the iteration number i. We also mean this when we say $\mathrm{i} = \infty$.

For a hypergraph $\mathcal{H}$ with star expansion matrix $H$, GWL-1 is strictly more expressive than WL-1 on $A = H \cdot D_{\mathrm{e}}^{-1} \cdot H^{T}$ with $D_{\mathrm{e}} = diag(H^{T} \cdot \mathbf{1}_{n})$, the node to node adjacency matrix, also called the clique expansion of $\mathcal{H}$. This follows since a triangle with its 3-cycle boundary: $T$ and a 3-cycle $C_3$ have exactly the same clique expansions. Thus WL-1 will give the same node values for both $T$ and $C_3$. GWL-1 on the star expansions $H_T$ and $H_{C_3}$, on the other hand, will identify the triangle as different from its bounding edges.

Let $f^{\mathrm{i}}(H) \triangleq [f_{\mathrm{e}_1}^{\mathrm{i}}(H), \cdots f_{\mathrm{e}_m}^{\mathrm{i}}(H)]$ and $h^{\mathrm{i}}(H) \triangleq [h_{v_1}^{\mathrm{i}}(H), \cdots h_{v_n}^{\mathrm{i}}(H)]$ be two vectors whose entries are ordered by the column and row order of $H$, respectively.

**Proposition 8.2.15.** *The update steps $f^{\mathrm{i}}(H)$ and $h^{\mathrm{i}}(H)$ of GWL-1 are permutation equivariant; For any $\pi \in Sym(\mathcal{V})$, let: $\pi \cdot f^{\mathrm{i}}(H) \triangleq [f_{\pi^{-1}(\mathrm{e}_1)}^{\mathrm{i}}(H), \cdots, f_{\pi^{-1}(\mathrm{e}_m)}^{\mathrm{i}}(H)]$ and $\pi \cdot h^{\mathrm{i}}(H) \triangleq [h_{\pi^{-1}(v_1)}^{\mathrm{i}}(H), \cdots h_{\pi^{-1}(v_n)}^{\mathrm{i}}(H)]$:*

$$
\forall \mathrm{i} \in \mathbb{N}, \pi \cdot f^{\mathrm{i}}(H) = f^{\mathrm{i}}(\pi \cdot H), \ \pi \cdot h^{\mathrm{i}}(H) = h^{\mathrm{i}}(\pi \cdot H)
\tag{8.6}
$$

Define the operator $AGG$ as a $k$-set map to representation space $\mathbb{R}^d$. Define the following representation of a $k$-node subset $S \subseteq \mathcal{V}$ of hypergraph $\mathcal{H}$ with star expansion matrix $H$:

$$h^{\mathrm{i}}(S, H) \triangleq AGG[\{h_v^{\mathrm{i}}(H)\}_{v \in S}] \tag{8.7}$$

where $h_v^{\mathrm{i}}(H)$ is the node value of i-GWL-1 on $H$ for node $v$. The representation $h(S, H)$ preserves hyperedge isomorphism classes as shown below:

**Proposition 8.2.16.** *Let $h^{\mathrm{i}}(S, H) = AGG[\{h_v^{\mathrm{i}}(H)\}_{v \in S}]$ with an injective $AGG$. The representation $h^{\mathrm{i}}(S, H)$ is $k$-node invariant but not necessarily $k$-node expressive for $S$ a set of $k$ nodes.*

It follows that we can guarantee a $k$-node invariant representation by using GWL-1. For deep learning, we parameterize $AGG$ as a universal set learner.

The node representations $h_v^{\mathrm{i}}(H)$ are also parameterized and rewritten into a message passing hypergraph neural network with matrix equations [269]. For example, the HNHN [276] hyperGNN architecture can be viewed as a parameterization of GWL-1:

$$\begin{aligned} X_E^{(l)} &= \sigma(H^T X_V^{(l)} W_E^{(l)} + b_E^{(l)}) \\ X_V^{(l+1)} &= \sigma(H X_E^{(l)} W_V^{(l)} + b_V^{(l)}) \end{aligned} \tag{8.8}$$

where $\sigma$ is a nonlinearity, $X_V, X_E$ are vector representations of $h^{\mathrm{i}}(H)$ and $f^{\mathrm{i}}(H)$ respectively, and $W_E, W_V, b_E, b_V$ are learnable weight matrices. Setting $b_E^{(l)} = 0, b_V^{(l)} = 0$, we maintain permutation equivariance as in Proposition 8.2.15. Furthermore, the HNHN equations of Equation 8.8 become in direct analogy to the steps of GWL-1 from Equation 8.5.

## 8.3 Related Work and Existing Issues

There are many hyperlink prediction methods. Most message passing based methods for hypergraphs are based on the GWL-1 algorithm. These include [267, 269, 276–281]. Examples of message passing based approaches that incorporate positional encodings on hypergraphs include SNALS [282]. The paper [283] uses a pair-wise node attention mechanism to do higher order link prediction. For a survey on hyperlink prediction, see [284].

Various methods have been proposed to improve the expressive power of GNNs due to symmetries in graphs. In [285], substructure labeling is formally analyzed. One of the methods analyzed includes labeling fixed radius ego-graphs as in [220, 286]. Other methods include appending random node features [271], labeling breadth-first and depth-first search trees [287] and encoding substructures [288, 289]. All of the previously mentioned methods depend on a fixed subgraph radius size. This prevents capturing symmetries that span long ranges across the graph. [290] proposes to add metric information of each node relative to all other nodes to improve WL-1. This would be very computationally expensive on hypergraphs.

Cycles are a common symmetric substructure. There are many methods that identify this symmetry. Cy2C [291] is a method that encodes cycles to cliques. It has the issue that if the the cycle-basis algorithm is not permutation invariant, isomorphic graphs could get different cycle bases and thus get encoded by Cy2C differently, violating the invariance of WL-1. Similarly, the CW Network [222] is a method that attaches cells to cycles to improve upon the distinguishing power of WL-1 for graph classification. However, inflating the input topology with cells as in [222] would not work for link predicting since it will shift the hyperedge distribution to become much denser. Other works include cell attention networks [292] and cycle basis based methods [5]. For more related work, see the Appendix.

Data augmentation is a commonly used approach to improve robustness to distribution shifts [293], recognize symmetries [294], and handle data imbalance [295]. In the graph domain, a priori knowledge of the data distribution can inform rule-based data augmentations. In molecular classification, prior knowledge of the physical meaning of the data can be used to augment graphs [296]. Data augmentations can also be learned through data generation methods. For example a link prediction neural network GAE [297] can be used to propose edges to to improve node classification [298]. For a survey on graph data augmentation, see [299].

## 8.4 A Characterization of GWL-1

A hypergraph can be represented by a bipartite graph $\mathcal{B}_{\mathcal{V},\mathcal{E}}$ from $\mathcal{V}$ to $\mathcal{E}$ where there is an edge $(v, \mathrm{e})$ in the bipartite graph iff node $v$ is incident to hyperedge e. This bipartite graph is called the star expansion bipartite graph.

We introduce a more structured version of graph isomorphism called a 2-color isomorphism to characterize hypergraphs. It is a map on 2-colored graphs, which are graphs that can be colored with two colors so that no two nodes in any graph with the same color are connected by an edge. We define a 2-colored isomorphism formally here:

**Definition 8.4.1.** *A* 2*-colored isomorphism is a graph isomorphism on two* 2*-colored graphs that preserves node colors. It is denoted by* $\cong_c$*.*

A 2-colored isomorphism from a graph $G$ to itself is called a 2-colored automorphism. The set of all 2-colored automorphisms on $G$ is denoted $Aut_c(G)$.

A bipartite graph always has a 2-coloring. In this paper, we canonically fix a 2-coloring on all star expansion bipartite graphs by assigning red to all the nodes in the node partition and and blue to all the nodes in the hyperedge partition. See Figure 8.2(a) as an example. We let $\mathcal{B}_{\mathcal{V}}, \mathcal{B}_{\mathcal{E}}$ be the red and blue colored nodes in $\mathcal{B}_{\mathcal{V},\mathcal{E}}$ respectively.

**Proposition 8.4.2.** *We have two hypergraphs* $(\mathcal{V}_1, \mathcal{E}_1) \cong (\mathcal{V}_2, \mathcal{E}_2)$ *iff* $\mathcal{B}_{\mathcal{V}_1,\mathcal{E}_1} \cong_c \mathcal{B}_{\mathcal{V}_2,\mathcal{E}_2}$ *where* $\mathcal{B}_{\mathcal{V}_\mathrm{i},\mathcal{E}_\mathrm{i}}$ *is the star expansion bipartite graph of* $(\mathcal{V}_\mathrm{i}, \mathcal{E}_\mathrm{i})$

We define a topological object for a graph originally from algebraic topology called a universal cover:

**Definition 8.4.3.** *([300]) The* universal covering *of a connected graph* $G$ *is a (potentially infinite) graph* $\tilde{G}$ *together with a map* $p_G : \tilde{G} \to G$ *such that:*

1. $\forall x \in \mathcal{V}(\tilde{G})$, $p_G |_{N(x)}$ *is an isomorphism onto* $N(p_G(x))$*.*

2. $\tilde{G}$ *is simply connected (a tree)*

We call such $p_G$ the *universal covering map* and $\tilde{G}$ the *universal cover* of $G$. A covering graph is a graph that satisfies property 1 but not necessarily 2 in Definition 8.4.3. The universal covering $\tilde{G}$ is essentially unique [300] in the sense that it can cover all connected covering graphs of $G$. Furthermore, define a rooted isomorphism $G_x \cong H_y$ as an isomorphism between graphs $G$ and $H$ that maps $x$ to $y$ and vice versa. Let $\tilde{G}^{\mathrm{i}}_{\tilde{x}}$ denote the rooted universal cover $\tilde{G}_{\tilde{x}}$ with every leaf of depth (number of edges) exactly $\mathrm{i} \in \mathbb{Z}^+$. It is a known result that:

**Theorem 8.4.4.** *([301]) Let $G$ and $H$ be two connected graphs. Let $p_G : \tilde{G} \to G, p_H : \tilde{H} \to H$ be the universal covering maps of $G$ and $H$ respectively. For any $\mathrm{i} \in \mathbb{N}$, for any two nodes $x \in G$ and $y \in H$: $\tilde{G}^{\mathrm{i}}_{\tilde{x}} \cong \tilde{G}^{\mathrm{i}}_{\tilde{y}}$ iff the WL-1 algorithm assigns the same value to nodes $x = p_G(\tilde{x})$ and $y = p_H(\tilde{y})$.*

We generalize the second result stated above about a topological characterization of WL-1 for GWL-1 for hypergraphs. In order to do this, we need to generalize the definition of a universal covering to suite the requirements of a bipartite star expansion graph. To do this, we lift $\mathcal{B}_{\mathcal{V},\mathcal{E}}$ to a 2-colored tree universal cover $\tilde{\mathcal{B}}_{\mathcal{V},\mathcal{E}}$ where the red/blue nodes of $\mathcal{B}_{\mathcal{V},\mathcal{E}}$ are lifted to red/blue nodes in $\tilde{\mathcal{B}}_{\mathcal{V},\mathcal{E}}$. Furthermore, the labels $\{\}$ are placed on the blue nodes corresponding to the hyperedges in the lift and the labels $\{\}$ are placed on all its corresponding red nodes in the lift. Let $(\tilde{\mathcal{B}}^k_{\mathcal{V},\mathcal{E}})_{\tilde{x}}$ denote the $k$-hop rooted 2-colored subtree $\tilde{\mathcal{B}}_{\mathcal{V},\mathcal{E}}$ with root $\tilde{x}$ and $p_{\mathcal{B}_{\mathcal{V},\mathcal{E}}}(\tilde{x}) = x$ for any $x \in \mathcal{V}(\mathcal{B}_{\mathcal{V},\mathcal{E}})$.

**Theorem 8.4.5.** *Let $\mathcal{H}_1 = (\mathcal{V}_1, \mathcal{E}_1)$ and $\mathcal{H}_2 = (\mathcal{V}_2, \mathcal{E}_2)$ be two connected hypergraphs. Let $\mathcal{B}_{\mathcal{V}_1,\mathcal{E}_1}$ and $\mathcal{B}_{\mathcal{V}_2,\mathcal{E}_2}$ be two canonically colored bipartite graphs for $\mathcal{H}_1$ and $\mathcal{H}_2$ (vertices colored red and hyperedges colored blue). Let $p_{\mathcal{B}_{\mathcal{V}_1,\mathcal{E}_1}} : \tilde{\mathcal{B}}_{\mathcal{V}_1,\mathcal{E}_1} \to \mathcal{B}_{\mathcal{V}_1,\mathcal{E}_1}, p_{\mathcal{B}_{\mathcal{V}_2,\mathcal{E}_2}} : \tilde{\mathcal{B}}_{\mathcal{V}_2,\mathcal{E}_2} \to \mathcal{B}_{\mathcal{V}_2,\mathcal{E}_2}$ be the universal coverings of $\mathcal{B}_{\mathcal{V}_1,\mathcal{E}_1}$ and $\mathcal{B}_{\mathcal{V}_2,\mathcal{E}_2}$ respectively. For any $\mathrm{i} \in \mathbb{Z}^+$, for any of the nodes $x_1 \in \mathcal{B}_{\mathcal{V}_1}, \mathrm{e}_1 \in \mathcal{B}_{\mathcal{E}_1}$ and $x_2 \in \mathcal{B}_{\mathcal{V}_2}, \mathrm{e}_2 \in \mathcal{B}_{\mathcal{E}_2}$:*

- $(\tilde{\mathcal{B}}^{2\mathrm{i}-1}_{\mathcal{V}_1,\mathcal{E}_1})_{\tilde{\mathrm{e}}_1} \cong_c (\tilde{\mathcal{B}}^{2\mathrm{i}-1}_{\mathcal{V}_2,\mathcal{E}_2})_{\tilde{\mathrm{e}}_2}$ *iff* $f^{\mathrm{i}}_{\mathrm{e}_1} = f^{\mathrm{i}}_{\mathrm{e}_2}$
- $(\tilde{\mathcal{B}}^{2\mathrm{i}}_{\mathcal{V}_1,\mathcal{E}_1})_{\tilde{x}_1} \cong_c (\tilde{\mathcal{B}}^{2\mathrm{i}}_{\mathcal{V}_2,\mathcal{E}_2})_{\tilde{x}_2}$ *iff* $h^{\mathrm{i}}_{x_1} = h^{\mathrm{i}}_{x_2}$,

*with $f^{\mathrm{i}}_{\bullet}, h^{\mathrm{i}}_{\bullet}$ the $\mathrm{i}$th GWL-1 values for the hyperedges and nodes respectively where $\mathrm{e}_1 = p_{\mathcal{B}_{\mathcal{V}_1,\mathcal{E}_1}}(\tilde{\mathrm{e}}_1)$, $x_1 = p_{\mathcal{B}_{\mathcal{V}_1,\mathcal{E}_1}}(\tilde{x}_1)$, $\mathrm{e}_2 = p_{\mathcal{B}_{\mathcal{V}_2,\mathcal{E}_2}}(\tilde{\mathrm{e}}_2)$, $x_2 = p_{\mathcal{B}_{\mathcal{V}_2,\mathcal{E}_2}}(\tilde{x}_2)$.*

Theorem 8.4.5 states that a 2-colored isomorphism is maintained during each step of the GWL-1 algorithm. Thus we can view the GWL-1 algorithm on a hypergraph as equivalent to computing a universal cover of the star expansion bipartite graph up to a 2-colored isomorphism. We can thus deduce from Theorems 8.4.5, 8.4.4 that GWL-1 reduces to computing WL-1 on the bipartite graph up to the 2-colored isomorphism.

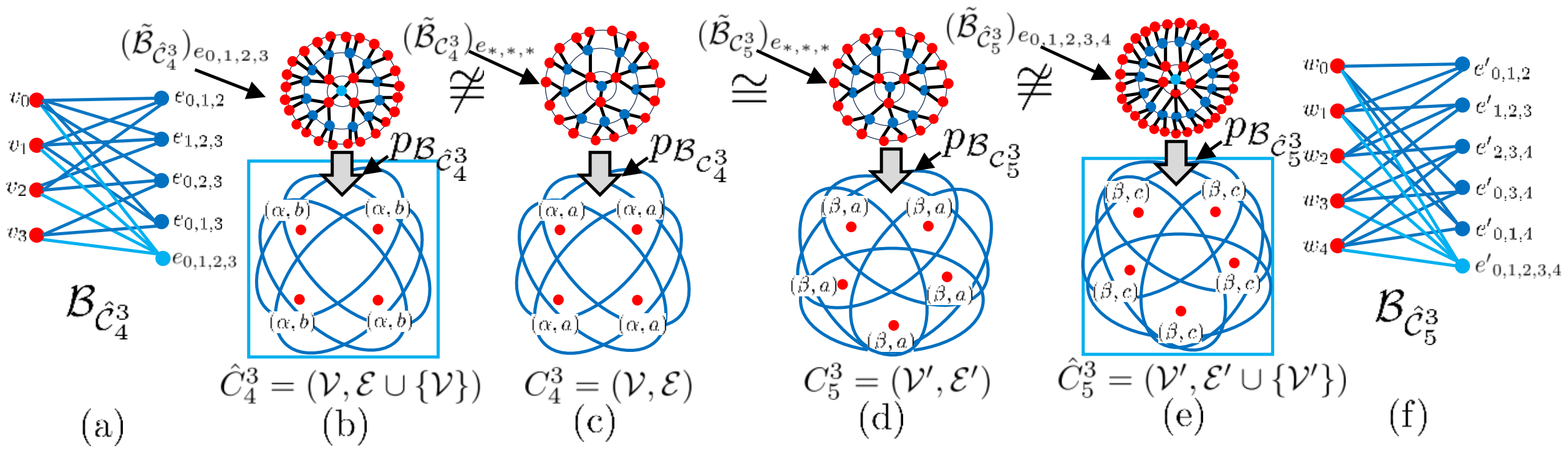

**Figure 8.2.** An illustration of hypergraph symmetry breaking. (c,d) 3-regular hypergraphs $C_4^3$, $C_5^3$ with 4 and 5 nodes respectively and their corresponding universal covers centered at any hyperedge $(\tilde{\mathcal{B}}_{C_4^3})_{e_{*,*,*}}, (\tilde{\mathcal{B}}_{C_5^3})_{e_{*,*,*}}$ with universal covering maps $p_{\mathcal{B}_{C_4^3}}, p_{\mathcal{B}_{C_5^3}}$. (b,e) the hypergraphs $\hat{C}_4^3, \hat{C}_5^3$, which are $C_4^3, C_5^3$ with $4, 5$-sized hyperedges attached to them and their corresponding universal covers and universal covering maps. (a,f) are the corresponding bipartite graphs of $\hat{C}_4^3, \hat{C}_5^3$. (c,d) are indistinguishable by GWL-1 and thus will give identical node values by Theorem 8.4.5. On the other hand, (b,e) gives node values which are now sensitive to the the order of the hypergraphs $4, 5$, also by Theorem 8.4.5.

**Corollary 8.4.6.** *Let $\mathcal{H}_1 = (\mathcal{V}_1, \mathcal{E}_1)$ and $\mathcal{H}_2 = (\mathcal{V}_2, \mathcal{E}_2)$ be two connected hypergraphs. Let $\mathcal{B}_{\mathcal{V}_1,\mathcal{E}_1}$ and $\mathcal{B}_{\mathcal{V}_2,\mathcal{E}_2}$ be two canonically colored bipartite graphs for $\mathcal{H}_1$ and $\mathcal{H}_2$ (vertices colored red and hyperedges colored blue). Let $p_{\mathcal{B}_{\mathcal{V}_1,\mathcal{E}_1}} : \tilde{\mathcal{B}}_{\mathcal{V}_1,\mathcal{E}_1} \to \mathcal{B}_{\mathcal{V}_1,\mathcal{E}_1}, p_{\mathcal{B}_{\mathcal{V}_2,\mathcal{E}_2}} : \tilde{\mathcal{B}}_{\mathcal{V}_2,\mathcal{E}_2} \to \mathcal{B}_{\mathcal{V}_2,\mathcal{E}_2}$ be the universal coverings of $\mathcal{B}_{\mathcal{V}_1,\mathcal{E}_1}$ and $\mathcal{B}_{\mathcal{V}_2,\mathcal{E}_2}$ respectively. For any $\mathrm{i} \in \mathbb{Z}^+$,*

- $(\tilde{\mathcal{B}}^{2\mathrm{i}-1}_{\mathcal{V}_1,\mathcal{E}_1})_{\tilde{v}_1} \cong_c (\tilde{\mathcal{B}}^{2\mathrm{i}-1}_{\mathcal{V}_2,\mathcal{E}_2})_{\tilde{v}_2}$ *iff* $g^{\mathrm{i}}_{v_1} = g^{\mathrm{i}}_{v_2}$

*with $g^{\mathrm{i}}_{v_1}, g^{\mathrm{i}}_{v_2}$ the* i*-th WL-1 values for $v_1, v_2 \in \mathcal{V}(\mathcal{B}_{\mathcal{V}_1,\mathcal{E}_1}), \mathcal{V}(\mathcal{B}_{\mathcal{V}_2,\mathcal{E}_2})$ respectively where $v_1 = p_{\mathcal{B}_{\mathcal{V}_1,\mathcal{E}_1}}(\tilde{v}_1)$, $v_2 = p_{\mathcal{B}_{\mathcal{V}_2,\mathcal{E}_2}}(\tilde{v}_2)$.*

See Figure 8.2 for an illustration of the universal covering of the corresponding bipartite graphs for two 3-uniform neighborhood regular hypergraphs.

### 8.4.7 A Limitation of GWL-1

Due to the equivalence of GWL-1 to viewing the hypergraph $\mathcal{H}$ as a collection of rooted trees, we can show that the automorphism group of $\mathcal{H}$ is a subgroup of the automorphism of this collection of rooted trees. This is stated in the following proposition:

**Theorem 8.4.8.** *Let $h^L : [\mathcal{V}]^1 \times \mathbb{Z}_2^{n\times 2^n} \to \mathbb{R}^d$ be the L-GWL-1 representation of nodes for hypergraph $\mathcal{H}$ in Equation 8.7, then*

$$Aut(\mathcal{H}) \cong Stab(H) \subseteq Sym(h^L(H)) \cong Aut_c(\tilde{\mathcal{B}}^{2L}_{\mathcal{V},\mathcal{E}}), \forall L \geq 1 \tag{8.9}$$

By Proposition 8.2.9, $Aut(\mathcal{H}) \cong Stab(H)$. The subgroup relationship follows by definition of the symmetry group of a representation map given in Definition 8.2.13 and the equivariance of $L$-GWL-1 due to Proposition 8.2.15. The last group isomorphism follows by the equivalence between $L$-GWL-1 and the universal cover of the star expansion bipartite graph $\mathcal{B}_{\mathcal{V},\mathcal{E}}$ up to $2L$-hops.

Since there are more automorphisms over the GWL-1 view of the hypergraph, many false positive symmetries might exist. Consider the following example. For two neighborhood-regular hypergraphs $C_1$ and $C_2$, the red/blue colored universal covers $\tilde{B}_{C_1}, \tilde{B}_{C_2}$ of the star expansions of $C_1$ and $C_2$ are isomorphic, with the same GWL-1 values on all nodes. However, two neighborhood-regular hypergraphs of different order become distinguishable if a single hyperedge covering all the nodes of each neighborhood-regular hypergraph is added. Furthermore, deleting the original hyperedges, does not change the node isomorphism classes of each hypergraph. Referring to Figure 8.2, consider the hypergraph $\mathcal{C} = C_4^3 \sqcup C_5^3$, the hypergraph with two 3-regular hypergraphs $C_4^3$ and $C_5^3$ acting as two connected components of $\mathcal{C}$. As shown in Figure 8.2, the node representations of the two hypergraphs are identical due to Theorem 8.4.5.

Given a hypergraph $\mathcal{H}$, we define a special induced subhypergraph $\mathcal{R} \subseteq \mathcal{H}$ whose node set GWL-1 cannot distinguish from other such special induced subhypergraphs.

**Definition 8.4.9.** *A* $L$-GWL-1 symmetric *induced subhypergraph* $\mathcal{R} \subset \mathcal{H}$ *of* $\mathcal{H}$ *is a connected induced subhypergraph determined by* $\mathcal{V}_{\mathcal{R}} \subseteq \mathcal{V}_{\mathcal{H}}$*, some subset of nodes that are all indistinguishable amongst each other by L-GWL-1:*

$$h_u^L(H) = h_v^L(H), \forall u, v \in \mathcal{V}_{\mathcal{R}} \tag{8.10}$$

*When* $L = \infty$*, we call such* $\mathcal{R}$ *a GWL-1 symmetric induced subhypergraph. Furthermore, if* $\mathcal{R} = \mathcal{H}$*, then we say* $\mathcal{H}$ *is* GWL-1 symmetric.

This definition is similar to that of a symmetric graph from graph theory [302], except that isomorphic nodes are determined by the GWL-1 approximator instead of an automorphism. The following observation follows from the definitions.

**Observation 8.4.10.** *A hypergraph* $\mathcal{H}$ *is GWL-1 symmetric if and only if it is L-GWL-1 symmetric for all* $L \geq 1$ *if and only if* $\mathcal{H}$ *is neighborhood regular.*

Our goal is to find GWL-1 symmetric induced subhypergraphs in a given hypergraph and break their symmetry without affecting any other nodes.

## 8.5 Method

Our goal is to learn from a training hypergraph and then predict higher order links in a temporally later hypergraph transductively. This can be formulated as follows:

**Problem 8.5.1.** ***Hyperlink Transductive Learning:*** *Let* $\mathcal{V}$ *be* $n$ *nodes.*

1. *Given a training hypergraph sampled at time* $t_{tr} \in \mathbb{R}$*:*

   *For* $(\mathcal{V}, (\mathcal{E}_{tr})_{gt}) \sim P(\mathcal{H}; t_{tr})$*, learn a boolean predictor* $\hat{h}$ *on hypergraph input so that:*

2. *For a testing hypergraph sampled at time* $t_{te} \in \mathbb{R}, t_{te} > t_{tr}$ *:*

   *For* $(\mathcal{V}, \mathcal{E}_{te}) \sim P(\mathcal{H}; t_{te})$ *and* $\mathcal{E}_{te} \subseteq (\mathcal{E}_{te})_{gt}$*,*

   $\hat{h}((\mathcal{V}, \mathcal{E}_{te}), \mathrm{e})$ *predicts whether* $\mathrm{e} \in (\mathcal{E}_{te})_{gt} \setminus \mathcal{E}_{te}, \forall \mathrm{e} \in 2^{\mathcal{V}}$

We will assume that the unobservable hyperedges are of the same size $k$ so that we only need to predict on $k$-node sets. In order to preserve the most information while still respecting topological structure, we aim to start with an invariant multi-node representation to predict hyperedges and increase its expressiveness, as defined in Definition 8.2.12. For input hypergraph $\mathcal{H}$ and its matrix representation $H$, to do the prediction of a missing hyperedge on node subsets, we use a multi-node representation $h(S,H)$ for $S \subseteq \mathcal{V}(H)$ as in Equation 8.7 due to its simplicity, guaranteed invariance, and improve its expressivity. We aim to not affect the computational complexity since message passing on hypergraphs is already quite expensive, especially on GPU memory.

Our method is a preprocessing algorithm that operates on the input hypergraph. In order to increase expressivity, we search for potentially indistinguishable regular induced subhypergraphs so that they can be replaced with hyperedges that span the subhypergraph to break the symmetries that prevent GWL-1 from being more expressive. We devise an algorithm, which is shown in Algorithm 29. It takes as input a hypergraph $\mathcal{H}$ with star expansion matrix $H$. The idea of the algorithm is to identify nodes of the same GWL-1 value that are maximally connected and use this collection of node subsets to break the symmetry of $\mathcal{H}$.

First we introduce some combinatorial definitions for hypergraph data that we will use in our algorithm:

**Definition 8.5.1.** *A hypergraph $\mathcal{H} = (\mathcal{V}, \mathcal{E})$ is **connected** if $\mathcal{B}_{\mathcal{V},\mathcal{E}}$ is a connected graph.*

*A **connected component** of $\mathcal{H}$ is a connected induced subhypergraph which is not properly contained in any connected subhypergraph of $\mathcal{H}$.*

Our preprocessing algorithm is explicitly given in Algorithm 29. We describe it in words here:

**Algorithm:**

1. For a given $L \in \mathbb{Z}^+$ and any $L$-GWL-1 node value $c_L$, the first step is to construct the induced subhypergraph $\mathcal{H}_{c_L}$ from the $L$-GWL-1 class of nodes:

$$\mathcal{V}_{c_L} \triangleq \{v \in \mathcal{V} : c_L = h_v^L(H)\}, \tag{8.11}$$

   where $h_v^L$ denotes the $L$-GWL-1 class of node $v$.

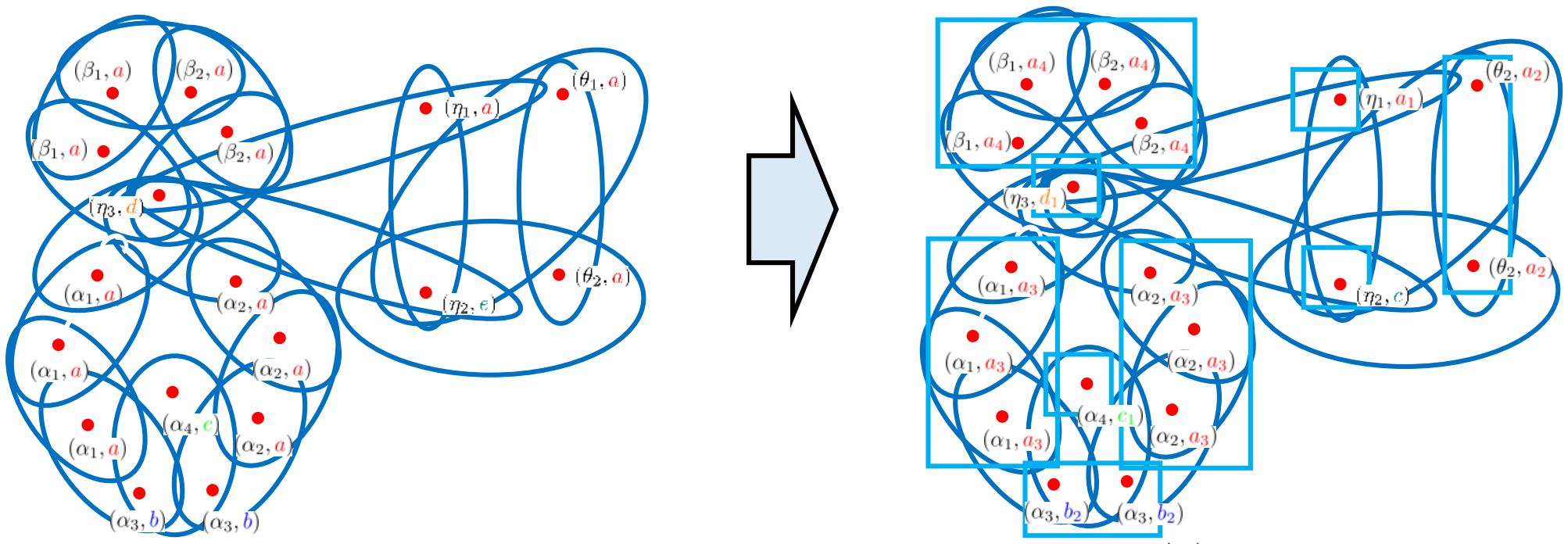

(a) (b)

**Figure 8.3.** An illustration of Algorithm 29 for 1-GWL-1.
In (a) a hypergraph is shown. Each node is labeled with a pair. The left part of the pair in Greek alphabet is its isomorphism class. The right part of the pair in Latin alphabet is its 1-GWL-1 class, which is determined by its neighborhood of hyperedges. In (b) the multi-hypergraph formed by covering the original hypergraph by hyperedges (light blue boxes) which are determined by the connected components of 1-GWL-1 indistinguishable node sets. The nodes can now be relabeled by 1-GWL-1. All hyperedges can be assigned learnable weights. For downstream training, the new hyperedges are randomly added and the existing hyperedges within each new hyperedge are randomly dropped.

2. We then compute the connected components of $\mathcal{H}_{c_L}$. Denote $\mathcal{C}_{c_L}$ as the set of all connected components of $\mathcal{H}_{c_L}$. If $L = \infty$, then drop $L$. Each of these connected components is a subhypergraph of $\mathcal{H}$, denoted $\mathcal{R}_{c_L,\mathrm{i}}$ where $\mathcal{R}_{c_L,\mathrm{i}} \subseteq \mathcal{H}_{c_L} \subseteq \mathcal{H}$ for $\mathrm{i} = 1...|\mathcal{C}_{c_L}|$.

3. Gather the collection of node sets and hyperedge sets formed by $\mathcal{R}_{c_L,\mathrm{i}}, \mathrm{i} = 1...|\mathcal{C}_{c_L}|$.

We call the subhypergraphs in $\mathcal{C}_{c_L}$ found by the symmetry finding Algorithm 29 as **maximally connected $L$-GWL-1 equal valued subhypergraphs**.

We will use the output hyperedges of the preprocessing algorithm to "softly" cover the input hypergraph. This will introduce multiplicities to the hyperedges. We formally define such a generalization of a hypergraph, called a multi-hypergraph here:

**Algorithm 29:** A Symmetry Finding Algorithm

**Data:** Hypergraph $\mathcal{H} = (\mathcal{V}, \mathcal{E})$, represented by its star expansion matrix $H$. $L \in \mathbb{Z}^+$ is the number of iterations to run GWL-1.

**Result:** A pair of collections: $(\mathcal{R}_V = \{\mathcal{V}_{R_\mathrm{j}}\}, \mathcal{R}_E = \cup_\mathrm{j}\{\mathcal{E}_{R_\mathrm{j}}\})$ where $R_\mathrm{j}$ are disconnected subhypergraphs exhibiting symmetry in $\mathcal{H}$ that are indistinguishable by $L$-GWL-1.

$U_L \leftarrow h_v^L(H); \mathcal{G}_L \leftarrow \{U_L[v] : \forall v \in \mathcal{V}\}$ /* $U_L[v]$ is the $L$-GWL-1 value of node $v \in \mathcal{V}$. */

$\mathcal{B}_{\mathcal{V}_\mathcal{H}, \mathcal{E}_\mathcal{H}} \leftarrow \mathrm{Bipartite}(\mathcal{H})$ /* Construct the bipartite graph from $\mathcal{H}$. */

$\mathcal{R}_V \leftarrow \{\}$; $\mathcal{R}_E \leftarrow \{\}$

**for** $c_L \in \mathcal{G}_L$ **do**

  $\mathcal{V}_{c_L} \leftarrow \{v \in \mathcal{V} : U_L[v] = c_L\}, \mathcal{E}_{c_L} \leftarrow \{\mathrm{e} \in \mathcal{E} : u \in \mathcal{V}_{c_L}, \forall u \in \mathrm{e}\}$

  $\mathcal{C}_{c_L} \leftarrow \mathrm{ConnectedComponents}(\mathcal{H}_{c_L} = (\mathcal{V}_{c_L}, \mathcal{E}_{c_L}))$

  **for** $\mathcal{R}_{c_L,\mathrm{i}} \in \mathcal{C}_{c_L}$ **do**

    $\mathcal{R}_V \leftarrow \mathcal{R}_V \cup \{\mathcal{V}_{\mathcal{R}_{c_L,\mathrm{i}}}\}$; $\mathcal{R}_E \leftarrow \mathcal{R}_E \cup \mathcal{E}_{\mathcal{R}_{c_L,\mathrm{i}}}$

  **end**

**end**

**return** $(\mathcal{R}_V, \mathcal{R}_E)$

**Definition 8.5.2.** *A **multi**-hypergraph $\mathcal{H} = (\mathcal{V}, \tilde{\mathcal{E}})$ is a pair consisting of a set of nodes $\mathcal{V}$ and a multiset of hyperedges $\tilde{\mathcal{E}} \triangleq (\mathcal{E}, m)$ where $\mathcal{E} \subseteq 2^\mathcal{V}$ and $m : \mathcal{E} \to \mathbb{Z}^+$ is a multiplicity function.*

The star expansion incidence matrix of a multi-hypergraph $\mathcal{H}$ can now be defined as:

**Definition 8.5.3.** *The star expansion incidence matrix $H$ of a multi-hypergraph $\mathcal{H} = (\mathcal{V}, \tilde{\mathcal{E}})$ is the $|\mathcal{V}| \times (2^{|\mathcal{V}|} \times \mathbb{Z}^+)$ (infinite) 0-1 incidence matrix $H$ where $H_{v,\mathrm{e}} = 1$ iff $v \in \mathrm{e}$ for $(v, \mathrm{e}) \in \mathcal{V} \times \tilde{\mathcal{E}}$ for some fixed orderings on both $\mathcal{V}$ and $2^\mathcal{V} \times \mathbb{Z}^+$.*

In practice, of course, the multiplicity function of $\mathcal{H}$ is bounded and only the nonzeros of this matrix are kept track of. Thus the star expansion incidence matrix is actually a sparse finite matrix.

On a multi-hypergraph $\mathcal{H} = (\mathcal{V}, \mathcal{E})$, we can define the degree of a vertex $v \in \mathcal{V}$ in terms of the cardinality of a multiset: $\deg(v) \triangleq |\{\{\mathrm{e} : \mathrm{e} \ni v\}\}| = \sum_{\mathrm{e} \ni v} m(\mathrm{e})$ where $m(\mathrm{e})$ is the multiplicity, or number of repeated occurrences, of e in $\{\{\mathrm{e} : \mathrm{e} \ni v\}\}$.

Letting $\mathcal{E}_{\mathcal{H}}(H)$ be the multiset of hyperedges of $\mathcal{H}$ and $\mathcal{V}_{\mathcal{H}}(H)$ the set of nodes of $\mathcal{H}$, message passing through incidences is still well defined. We summarize this in the following proposition:

**Proposition 8.5.4.** *The GWL-1 algorithm of Equation 8.5 can be computed on a multi-hypergraph* $\mathcal{H}$

*Thus,* $h_v^{\mathrm{i}}(H)$ *and* $h^{\mathrm{i}}(S, H)$ *are well defined on its star incidence matrix* $H$*.*

This implication of Proposition 8.5.4 is that GWL-1 based hyperGNNs can learn on multi-hypergraphs using the star expansion incidence matrix representations.

**Downstream Training:** After executing Algorithm 29, we collect its output $(\mathcal{R}_V, \mathcal{R}_E)$. During training, for each $\mathrm{i} = 1, ..., |\mathcal{C}_{c_L}|$ we randomly perturb $\mathcal{R}_{c_L,\mathrm{i}}$ to form a random multi-hypergraph $\hat{\mathcal{H}}_L$ by:

- Attaching (with multiplicity) a single hyperedge that covers $\mathcal{V}_{\mathcal{R}_{c_L,\mathrm{i}}}$ with probability $q_{\mathrm{i}}$ and not attaching with probability $1 - q_{\mathrm{i}}$.
- All the hyperedges in $\mathcal{R}_{c_L,\mathrm{i}}$ are dropped or kept with probability $p$ and $1-p$ respectively.

Our method is similar to the concept of adding virtual nodes [303] in graph representation learning. This is due to the equivalence between virtual nodes and hyperedges by Proposition 8.4.2. For a guarantee of improving expressivity, see Lemma 8.5.9 and Theorems 8.5.10, 8.5.11. For an illustration of the data augmentation, see Figure 8.2.

Alternatively, downstream training using the output of Algorithm 29 can be done. Similar to subgraph NNs, this is done by applying an ensemble of models [304–306], with each model trained on transformations of $\mathcal{H}$ with its symmetric subhypergraphs randomly replaced. This, however, is computationally expensive.

**Illustration:** In Figure 8.3 an illustration of Algorithm 29 is shown. For the hypergraph shown in Figure 8.3 (a), two graph cycles are glued at a single node while a separate hypergraph is glued at that same node. With 1-GWL-1, there is a large class of nodes labeled "a" that are indistinguishable. Each of the connected components of these 1-GWL-1 node classes is covered by a single hyperedge (light blue box) to form a multi-hypergraph. We show in Section 8.5.5 that in this multihypergraph, the nodes express more of the original hypergraph. The data augmentation procedure during downstream training can then be applied to the blue boxes and the original hyperedges within each blue box separately.

### 8.5.5 Algorithm Guarantees

We show some guarantees for the output of Algorithm 29. We will assume the following notation to denote the relevant substructures on a single hypergraph found by Algorithm 29.

**Notation:**

Let $\mathcal{H} = (\mathcal{V}, \mathcal{E})$ be a hypergraph with star expansion matrix $H$ as before.

Let $(\mathcal{R}_{\mathcal{V}}, \mathcal{R}_{\mathcal{E}})$ be the output of Algorithm 29 on $H$ for $L \in \mathbb{Z}^+$. We call this collection of nodes and hyperedges as GWL-1 symmetric induced components. Let:

$$\hat{\mathcal{H}}_L \triangleq (\mathcal{V}, \mathcal{E} \sqcup \mathcal{R}_V) \tag{8.12}$$

be the multi-hypergraph formed from $\mathcal{H}$ after adding all the hyperedges from $\mathcal{R}_{\mathcal{V}}$ and let $\hat{H}_L$ be the star expansion matrix of the resulting multi-hypergraph $\hat{\mathcal{H}}_L$. Let:

$$V_{c_L,s} \triangleq \{v \in \mathcal{V}_{c_L} : v \in R, R \in \mathcal{C}_{c_L}, |\mathcal{V}_R| = s\} \tag{8.13}$$

be the set of all nodes of $L$-GWL-1 class $c_L$ belonging to a connected component in $\mathcal{C}_{c_L}$ of $s \geq 1$ nodes in $\mathcal{H}_{c_L}$, the induced subhypergraph of $L$-GWL-1. Let:

$$\mathcal{G}_L \triangleq \{h_v^L(H) : v \in \mathcal{V}\} \tag{8.14}$$

be the set of all $L$-GWL-1 values on $H$. Let:

$$\mathcal{S}_{c_L} \triangleq \{|\mathcal{V}_{\mathcal{R}_{c_L,\mathrm{i}}}| : \mathcal{R}_{c_L,\mathrm{i}} \in \mathcal{C}_{c_L}\} \tag{8.15}$$

be the set of node set sizes of the connected components in $\mathcal{H}_{c_L}$.

**Properties of $(\mathcal{R}_{\mathcal{V}}, \mathcal{R}_{\mathcal{E}})$ from Algorithm 29:**

In the following proposition, we show that the subhypergraphs found by Algorithm 29 are GWL-1 symmetric. This should be evident by line 5 of Algorithm 29 where an induced subgraph of the bipartite graph is formed from nodes of the same $L$-GWL-1 values.

**Proposition 8.5.6.** *If $L = \infty$, for any GWL-1 node value $c$ computed on $\mathcal{H}$, all connected component subhypergraphs $\mathcal{R}_{c,\mathrm{i}} \in \mathcal{C}_c$ are GWL-1 symmetric as hypergraphs.*

Algorithm 29 outputs a partition of the entire hypergraph. This follows from the fact that GWL-1 already covers the entire hypergraph after a single hop.

**Proposition 8.5.7.** *If $L \geq 1$, the output $(\mathcal{R}_{\mathcal{V}}, \mathcal{R}_{\mathcal{E}})$ of Algorithm 29 partitions a subgraph of $\mathcal{H}$, meaning:*

$$\mathcal{V} = \sqcup_{V \in \mathcal{R}_{\mathcal{V}}} V \text{ and } \mathcal{E} \supset \sqcup_{E \in \mathcal{R}_{\mathcal{E}}} E \tag{8.16}$$

The output of Algorithm 29 can additionally be used to guarantee improvement in expressing the hypergraph:

**Prediction Guarantees:**

In order to guarantee that the GWL-1 symmetric components $\mathcal{R}_{c,\mathrm{i}}$ found by Algorithm 29 carry additional information, there needs to be a separation between them to prevent an intersection between the rooted trees computed by GWL-1. We define what it means for two node subsets to be sufficiently separated via the shortest hyperedge path distance between nodes in $\mathcal{V}$ as follows:

**Definition 8.5.8.** *Two subsets of nodes $\mathcal{U}_1, \mathcal{U}_2 \subseteq \mathcal{V}$ are **sufficiently $L$-separated** if:*

$$\min_{v_1 \in \mathcal{U}_1, v_2 \in \mathcal{U}_2} d(v_1, v_2) > L \tag{8.17}$$

*where $d(v_1, v_2) \triangleq \min_{\mathrm{e}_1 \ldots \mathrm{e}_k \in \mathcal{E}, v_1 \in \mathrm{e}_1, v_2 \in \mathrm{e}_k} k$ is the shortest hyperedge path distance from $v_1 \in \mathcal{V}$ to $v_2 \in \mathcal{V}$.*

*A collection of node subsets $\mathcal{C} \subseteq 2^{\mathcal{V}}$ is **sufficiently $L$-separated** if all pairs of node subsets are **sufficiently $L$-separated**.*

Our definition of sufficiently $L$-separated is similar in nature to that of well separation between point sets [307] in Euclidean space. Assuming that the $\mathcal{C}_{c_L}$ are sufficiently $L$-separated from each other, intuitively meaning that no two nodes from two separate $\mathcal{V}_{\mathcal{R}_{c_L},\mathrm{i}} \in \mathcal{R}_V$ are within $L$ hyperedges away, then the cardinality of each component $|\mathcal{V}_{\mathcal{R}_{c_L},\mathrm{i}}|$ is recognizable. This is stated in the following lemma:

**Lemma 8.5.9.** *If $L \in \mathbb{Z}^+$ is small enough so that after running Algorithm 29 on $L$, for any $L$-GWL-1 node class $c_L$ on $\mathcal{V}$ the collection of $\mathcal{C}_{c_L}$ is* ***sufficiently $L$-separated****,*

*then after forming $\hat{\mathcal{H}}_L$, the new $L$-GWL-1 node classes of $\mathcal{V}_{\mathcal{R}_{c_L},\mathrm{i}}$ for $\mathrm{i} = 1 \ldots \mathcal{C}_{c_L}$ in $\hat{\mathcal{H}}_L$ are all the same class $c'_L$ but are distinguishable from $c_L$ depending on $|\mathcal{V}_{\mathcal{R}_{c_L},\mathrm{i}}|$.*

We can then use this lemma to show that under certain conditions for large hypergraphs, augmenting the hypergraph $\mathcal{H}$ to the multi-hypergraph $\hat{\mathcal{H}}_L$ will give a guarantee on the number of pairs of $k$-node sets that become distinguished which were indistinguishable as sets of GWL-1 values.

**Theorem 8.5.10.** *Let $|\mathcal{V}| = n, L \in \mathbb{Z}^+$ and $vol(v) \triangleq \sum_{\mathrm{e} \in \mathcal{E}:\mathrm{e} \ni v} |\mathrm{e}|$ and assuming that the collection of node subsets $\mathcal{C}_{c_L}$ is sufficiently $L$-separated.*

*If $vol(v) = O(\log^{\frac{1-\epsilon}{4L}} n), \forall v \in \mathcal{V}$ for any constant $\epsilon > 0$; $|\mathcal{S}_{c_L}| \leq S, \forall c_L \in \mathcal{C}_L$, $S$ constant, and $|V_{c_L,s}| = O(\frac{n^\epsilon}{\log^{\frac{1}{2k}}(n)}), \forall s \in \mathcal{C}_{c_L}$, then for $k \in \mathbb{Z}^+$ and $k$-tuple $C = (c_{L,1} \ldots c_{L,k}), c_{L,\mathrm{i}} \in \mathcal{G}_L, \mathrm{i} = 1..k$ there exists $\omega(n^{2k\epsilon})$ many pairs of $k$-node sets $S_1 \not\simeq S_2$ such that $(h_u^L(H))_{u \in S_1} = (h_{v \in S_2}^L(H)) = C$, as ordered $k$-tuples, while $h(S_1, \hat{H}_L) \neq h(S_2, \hat{H}_L)$ also by $L$ steps of GWL-1.*

The conditions of Theorem 8.5.10 assume an arbitrarily large hypergraph that is sparse and for every $c_L \in \mathcal{G}_L$ has $\mathcal{C}_{c_L}$ sufficiently $L$-separated, has a bounded set of node set sizes from $\mathcal{S}_{c_L}$ and a controlled growth for each node set size from $\mathcal{S}_{c_L}$. The idea behind the proof of Theorem 8.5.10 is that under these conditions, there are enough isomorphic rooted trees computed by $L$-GWL-1. It can then be shown that over all pairs of $k$-sets of nodes with elementwise isomorphic rooted trees, that they can be distinguished by the component size they belong in. We give a simple example hypergraph that illustrates the condition of Theorem 8.5.10.

**Example:** A simple example of a hypergraph that statisfies the conditions of Theorem 8.5.10 is a union of many disconnected hypergraphs $\mathcal{H} = \cup_{\mathrm{i}} \mathcal{H}_{\mathrm{i}} = (\mathcal{V}, \mathcal{E})$ with $|\mathcal{V}_{\mathcal{H}_{\mathrm{i}}}| \leq S$ where $S < \infty$ is a small constant independent of $n = |\mathcal{V}| \geq S$. Such a hypergraph could be a social

network where the nodes are user instances and the hyperedges are private groups. The disconnected hypergraphs represent disconnected communities where a user can only belong to a single community.

We show that our algorithm increases expressivity (Definition 8.2.12) for $h(S, H)$ of Equation 8.7.

**Theorem 8.5.11.** *(Invariance and Expressivity) If $L = \infty$, GWL-1 enhanced by Algorithm 29 is still invariant to node isomorphism classes of $\mathcal{H}$ and can be strictly more expressive than GWL-1 to determine node isomorphism classes.*

Proving expressivity from Theorem 8.5.9 follows from the added information of component sizes viewable by each node in its vicinity. Proving the invariance from Theorem 8.5.9 follows by a proof by contradiction, which uses the maximality of the connected components found in Algorithm 29.

When training on a hypergraph with the data augmentations, every possible symmetry observed from the random augmentations will be learned. Thus the symmetry group of $h^L$ on the random matrix $\hat{H}_L$ is isomorphic to the intersection of all symmetries over each augmentation sample. This is expressed as follows:

$$\text{Sym}(h^L(\hat{H}_L)) \triangleq \bigcap_{\hat{H}'_L \sim P(\hat{H}_L)} \text{Sym}(h^L(\hat{H}'_L)) \tag{8.18}$$

We show that the intersection over all symmetries of the estimated multi-hypergraph $\hat{\mathcal{H}}_L$ has fewer symmetries than that of the $L$-GWL-1 view of the hypergraph as a collection of rooted trees. We call this **symmetry breaking**.

**Proposition 8.5.12.** *The multi-hypergraph $\hat{\mathcal{H}}_L$ breaks the symmetry of the L-GWL-1 view of the hypergraph $\mathcal{H}$:*

$$Sym(h^L(\hat{H}_L)) \subseteq Aut_c(\tilde{\mathcal{B}}^{2L}_{\mathcal{V},\mathcal{E}}), \forall L \geq 1 \tag{8.19}$$

This follows by the fact that the identity augmentation is in the support of the distribution of random augmentations and that $\text{Sym}(h^L(\hat{H}_L)) \cong Aut_c(\tilde{\mathcal{B}}^{2L}_{\mathcal{V},\mathcal{E}(\hat{\mathcal{H}})})$ by Theorem 8.4.8.

Since hyperGNNs represent each node $v \in \mathcal{V}$ by message passing through the neighbors in the rooted tree $(\tilde{\mathcal{B}}_{\mathcal{V},\mathcal{E}})_v$ at $v$. If probabilities are assigned between nodes, then $T$ layers of a hyperGNN can be viewed as computing the random walk probability of ending on any node starting from some uniformly chosen node. We define these terms in the following:

**Definition 8.5.13.** *([308]) A **random walk** on a (multi) hypergraph $\mathcal{H} = (\mathcal{V}, \mathcal{E})$ is a Markov chain with state space $\mathcal{V}$ with transition probabilities $P_{u,v} \triangleq \sum_{\mathrm{e} \supset \{u,v\}: \mathrm{e} \in \mathcal{E}} \frac{\omega(\mathrm{e})}{deg(u)|\mathrm{e}|}$, where $\omega(\mathrm{e}) : \mathcal{E} \to [0,1]$ is some discrete probability distribution on the hyperedges. When not specified, this is the constant 1 function.*

Assuming $\mathcal{H}$ is connected, let $X_t \in \mathcal{V}$ denote the state of the Markov chain at step $t$ with $P(X_0 = v) = \frac{1}{|\mathcal{V}|}, \forall v \in \mathcal{V}$. Letting $t \to \infty$, this probability converges to the stationary distribution on the nodes $\mathcal{V}$, which is independent of the time. This is expressed in the following definition:

**Definition 8.5.14.** *A **stationary distribution** $\pi : \mathcal{V} \to [0,1]$ for a Markov chain with transition probabilities $P_{u,v}$ is defined by the relationship $\sum_{u \in \mathcal{V}} P_{u,v} \pi(u) = \pi(v)$.*

*For a (multi) hypergraph random walk we have the closed form: $\pi(v) = \frac{deg(v)}{\sum_{u \in \mathcal{V}} deg(u)}$ for $v \in \mathcal{V}$ assuming $\mathcal{H}$ is a connected (multi) hypergraph.*

For the downstream training, we show that there are Bernoulli hyperedge drop/attachment probabilities $p, q_{\mathrm{i}}$ respectively for each $\mathcal{R}_{c_L,\mathrm{i}}$ so that the stationary distribution doesn't change. This shows that our data augmentation can still preserve the low frequency random walk signal.

**Proposition 8.5.15.** *For a connected hypergraph $\mathcal{H} = (\mathcal{V}, \mathcal{E})$, let $(\mathcal{R}_V, \mathcal{R}_E)$ be the output of Algorithm 29 on $\mathcal{H}$. Then there are Bernoulli probabilities $p, q_{\mathrm{i}}$ for $\mathrm{i} = 1, ..., |\mathcal{R}_V|$ for attaching a covering hyperedge so that $\hat{\pi}$ is an unbiased estimator of $\pi$.*

The intuition for Proposition 8.5.15 is that if a hyperedge is added to cover a connected subhypergraph $\mathcal{R}_{c_L,\mathrm{i}}$ containing at least one hyperedge, then allowing any of the hyperedges in $\mathcal{R}_{c_L,\mathrm{i}}$ to drop is enough to keep the estimated stationary distribution $\hat{\pi}$ unbiased.

**Time Complexity:**

Proposition 8.5.16 provides the time complexity of our algorithm.

**Proposition 8.5.16.** *(Complexity) Algorithm 29 runs in time $O(nnz(H)L+(n+m))$, which is order linear in the size of the input star expansion matrix $H$ for hypergraph $\mathcal{H}=(\mathcal{V},\mathcal{E})$, if $L$ is independent of $nnz(H)$, where $n=|\mathcal{V}|$, $nnz(H)=vol(\mathcal{V})\triangleq\sum_{v\in\mathcal{V}}deg(v)$ and $m=|\mathcal{E}|$.*

Since Algorithm 29 runs in time linear in the size of the input when $L$ is constant, in practice it only takes a small fraction of the training time for hypergraph neural networks.

## 8.6 Evaluation

We evaluate our method on higher order link prediction with many of the standard hypergraph neural network methods. Due to potential class imbalance, we measure the PR-AUC of higher order link prediction on the hypergraph datasets. These datasets are: CAT-EDGE-DAWN, CAT-EDGE-MUSIC-BLUES-REVIEWS, CONTACT-HIGH-SCHOOL, CONTACT-PRIMARY-SCHOOL, EMAIL-EU, CAT-EDGE-MADISON-RESTAURANTS.

**Table 8.1.** Transductive hyperedge prediction PR-AUC scores on six different hypergraph datasets. The highest scores per hyperGNN architecture (row) is colored. Red text denotes the highest average scoring method. Orange text denotes a two-way tie and brown text denotes a three-way tie. All datasets involve predicting hyperedges of size 3.

| PR-AUC $\uparrow$ | Baseline | Ours | Baseln.+edrop |
|---|---|---|---|
| HGNN | $0.98 \pm 0.03$ | $0.99 \pm 0.08$ | $0.96 \pm 0.02$ |
| HGNNP | $0.98 \pm 0.02$ | $0.98 \pm 0.09$ | $0.96 \pm 0.10$ |
| HNHN | $0.98 \pm 0.01$ | $0.96 \pm 0.07$ | $0.97 \pm 0.04$ |
| HyperGCN | $0.98 \pm 0.07$ | $0.98 \pm 0.11$ | $0.98 \pm 0.03$ |
| UniGAT | $0.99 \pm 0.06$ | $0.99 \pm 0.03$ | $0.99 \pm 0.07$ |
| UniGCN | $0.99 \pm 0.00$ | $0.99 \pm 0.03$ | $0.99 \pm 0.08$ |
| UniGIN | $0.87 \pm 0.02$ | $0.86 \pm 0.10$ | $0.85 \pm 0.08$ |
| UniSAGE | $0.86 \pm 0.04$ | $0.86 \pm 0.05$ | $0.84 \pm 0.09$ |

(a) CAT-EDGE-DAWN

| PR-AUC $\uparrow$ | Baseline | Ours | Baseln.+edrop |
|---|---|---|---|
| HGNN | $0.90 \pm 0.13$ | $1.00 \pm 0.00$ | $0.90 \pm 0.13$ |
| HGNNP | $0.90 \pm 0.09$ | $1.00 \pm 0.07$ | $1.00 \pm 0.03$ |
| HNHN | $0.90 \pm 0.09$ | $0.91 \pm 0.02$ | $0.90 \pm 0.08$ |
| HyperGCN | $1.00 \pm 0.00$ | $1.00 \pm 0.03$ | $1.00 \pm 0.02$ |
| UniGAT | $0.90 \pm 0.06$ | $1.00 \pm 0.03$ | $1.00 \pm 0.06$ |
| UniGCN | $1.00 \pm 0.01$ | $0.91 \pm 0.01$ | $0.82 \pm 0.09$ |
| UniGIN | $0.90 \pm 0.12$ | $0.95 \pm 0.06$ | $0.90 \pm 0.11$ |
| UniSAGE | $0.90 \pm 0.16$ | $1.00 \pm 0.08$ | $0.90 \pm 0.17$ |

(b) CAT-EDGE-MUSIC-BLUES-REVIEWS

| PR-AUC $\uparrow$ | Baseline | Ours | Baseln.+ edrop |
|---|---|---|---|
| HGNN | $0.96 \pm 0.10$ | $0.98 \pm 0.05$ | $0.96 \pm 0.04$ |
| HGNNP | $0.96 \pm 0.05$ | $0.98 \pm 0.09$ | $0.97 \pm 0.07$ |
| HNHN | $0.96 \pm 0.02$ | $0.97 \pm 0.08$ | $0.97 \pm 0.06$ |
| HyperGCN | $0.93 \pm 0.05$ | $0.98 \pm 0.07$ | $0.96 \pm 0.09$ |
| UniGAT | $0.96 \pm 0.01$ | $0.98 \pm 0.14$ | $0.97 \pm 0.04$ |
| UniGCN | $0.96 \pm 0.04$ | $0.96 \pm 0.11$ | $0.96 \pm 0.09$ |
| UniGIN | $0.97 \pm 0.03$ | $0.97 \pm 0.11$ | $0.96 \pm 0.05$ |
| UniSAGE | $0.96 \pm 0.10$ | $0.96 \pm 0.10$ | $0.96 \pm 0.02$ |

(c) CONTACT-HIGH-SCHOOL

| PR-AUC $\uparrow$ | Baseline | Ours | Baseln.+edrop |
|---|---|---|---|
| HGNN | $0.95 \pm 0.03$ | $0.96 \pm 0.01$ | $0.95 \pm 0.03$ |
| HGNNP | $0.95 \pm 0.02$ | $0.96 \pm 0.09$ | $0.96 \pm 0.07$ |
| HNHN | $0.94 \pm 0.07$ | $0.97 \pm 0.10$ | $0.95 \pm 0.05$ |
| HyperGCN | $0.97 \pm 0.01$ | $0.97 \pm 0.05$ | $0.96 \pm 0.08$ |
| UniGAT | $0.95 \pm 0.02$ | $0.98 \pm 0.07$ | $0.98 \pm 0.02$ |
| UniGCN | $0.96 \pm 0.00$ | $0.97 \pm 0.14$ | $0.97 \pm 0.10$ |
| UniGIN | $0.95 \pm 0.09$ | $0.97 \pm 0.02$ | $0.95 \pm 0.05$ |
| UniSAGE | $0.96 \pm 0.08$ | $0.95 \pm 0.05$ | $0.96 \pm 0.02$ |

(d) CONTACT-PRIMARY-SCHOOL

| PR-AUC $\uparrow$ | Baseline | Ours | Baseln.+edrop |
|---|---|---|---|
| HGNN | $0.95 \pm 0.07$ | $0.97 \pm 0.08$ | $0.96 \pm 0.07$ |
| HGNNP | $0.95 \pm 0.07$ | $0.96 \pm 0.02$ | $0.96 \pm 0.01$ |
| HNHN | $0.94 \pm 0.01$ | $0.97 \pm 0.02$ | $0.95 \pm 0.06$ |
| HyperGCN | $0.92 \pm 0.01$ | $0.94 \pm 0.06$ | $0.94 \pm 0.08$ |
| UniGAT | $0.94 \pm 0.08$ | $0.98 \pm 0.14$ | $0.97 \pm 0.08$ |
| UniGCN | $0.97 \pm 0.08$ | $0.97 \pm 0.14$ | $0.97 \pm 0.06$ |
| UniGIN | $0.93 \pm 0.07$ | $0.94 \pm 0.11$ | $0.93 \pm 0.09$ |
| UniSAGE | $0.93 \pm 0.07$ | $0.93 \pm 0.08$ | $0.92 \pm 0.04$ |

(e) EMAIL-EU

| PR-AUC $\uparrow$ | Baseline | Ours | Baseln.+edrop |
|---|---|---|---|
| HGNN | $0.75 \pm 0.09$ | $0.85 \pm 0.09$ | $0.71 \pm 0.14$ |
| HGNNP | $0.83 \pm 0.09$ | $0.85 \pm 0.08$ | $0.85 \pm 0.04$ |
| HNHN | $0.72 \pm 0.09$ | $0.82 \pm 0.03$ | $0.74 \pm 0.09$ |
| HyperGCN | $0.87 \pm 0.08$ | $0.83 \pm 0.05$ | $1.00 \pm 0.07$ |
| UniGAT | $0.80 \pm 0.09$ | $0.83 \pm 0.03$ | $0.78 \pm 0.05$ |
| UniGCN | $0.84 \pm 0.08$ | $0.89 \pm 0.10$ | $0.71 \pm 0.07$ |
| UniGIN | $0.69 \pm 0.14$ | $0.76 \pm 0.05$ | $0.61 \pm 0.11$ |
| UniSAGE | $0.72 \pm 0.11$ | $0.71 \pm 0.10$ | $0.64 \pm 0.10$ |

(f) CAT-EDGE-MADISON-RESTAURANTS

These datasets range from representing social interactions as they develop over time to collections of reviews to drug combinations before overdose. We also evaluate on the AMHERST41 dataset, which is a graph dataset. All of our datasets are unattributed hypergraphs/graphs.

**Data Splitting:** For the hypergraph datasets, each hyperedge in it is paired with a timestamp (a real number). These timestamps are a physical time for which a higher order interaction, represented by a hyperedge, occurs. We form a train-val-test split by letting the train be the hyperedges associated with the 80th percentile of timestamps, the validation be the hyperedges associated with the timestamps in between the 80th and 85th percentiles. The test hyperedges are the remaining hyperedges. The train validation and test datasets thus form a partition of the nodes. We do the task of hyperedge prediction for sets of nodes of size 3, also known as triangle prediction. Half of the size 3 hyperedges in each of train, validation and test are used as positive examples. For each split, we select random subsets of nodes of size 3 that do not form hyperedges for negative sampling. We maintain positive/negative class balance by sampling the same number of negative samples as positive samples. Since the test distribution comes from later time stamps than those in training, there is a possibility that certain datasets are out-of-distribution if the hyperedge distribution changes.

For the graph dataset, the single graph is deterministically split into 80/5/15 for train/-val/test. We remove 10% of the edges in training and let them be positive examples $P_{tr}$ to predict. For validation and test, we remove 50% of the edges from both validation and test to set as the positive examples $P_{val}, P_{te}$ to predict. For train, validation, and test, we sample $|P_{tr}|, |P_{val}|, |P_{te}|$ negative link samples from the links of train, validation and test.

**Table 8.2.** PR-AUC on graph dataset AMHERST41. Each column is a comparison of the baseline PR-AUC scores against the PR-AUC score for our method (first row) applied to a standard hyperGNN architecture. The coloring scheme is the same as in Table 8.1.

| PR-AUC ↑ | HGNN | HGNNP | HNHN | HyperGCN | UniGAT | UniGCN | UniGIN | UniSAGE |
|---|---|---|---|---|---|---|---|---|
| Ours | 0.73 ± 0.10 | 0.61 ± 0.05 | 0.64 ± 0.06 | 0.71 ± 0.09 | 0.72 ± 0.08 | 0.70 ± 0.08 | 0.73 ± 0.03 | 0.73 ± 0.06 |
| hyperGNN Baseline | 0.62 ± 0.09 | 0.62 ± 0.10 | 0.63 ± 0.04 | 0.71 ± 0.07 | 0.70 ± 0.06 | 0.69 ± 0.07 | 0.73 ± 0.06 | 0.73 ± 0.09 |
| hyperGNN Baseln.+edrop | 0.61 ± 0.03 | 0.61 ± 0.03 | 0.61 ± 0.09 | 0.71 ± 0.06 | 0.71 ± 0.02 | 0.69 ± 0.05 | 0.73 ± 0.09 | 0.73 ± 0.04 |
| APPNP | 0.42 ± 0.07 | 0.42 ± 0.07 | 0.42 ± 0.07 | 0.42 ± 0.07 | 0.42 ± 0.07 | 0.42 ± 0.07 | 0.42 ± 0.07 | 0.42 ± 0.07 |
| APPNP+edrop | 0.42 ± 0.03 | 0.42 ± 0.03 | 0.42 ± 0.03 | 0.42 ± 0.03 | 0.42 ± 0.03 | 0.42 ± 0.03 | 0.42 ± 0.03 | 0.42 ± 0.03 |
| GAT | 0.49 ± 0.06 | 0.49 ± 0.06 | 0.49 ± 0.06 | 0.49 ± 0.06 | 0.49 ± 0.06 | 0.49 ± 0.06 | 0.49 ± 0.06 | 0.49 ± 0.06 |
| GAT+edrop | 0.49 ± 0.06 | 0.49 ± 0.06 | 0.49 ± 0.06 | 0.49 ± 0.06 | 0.49 ± 0.06 | 0.49 ± 0.06 | 0.49 ± 0.06 | 0.49 ± 0.06 |
| GCN2 | 0.56 ± 0.12 | 0.56 ± 0.12 | 0.56 ± 0.12 | 0.56 ± 0.12 | 0.56 ± 0.12 | 0.56 ± 0.12 | 0.56 ± 0.12 | 0.56 ± 0.12 |
| GCN2+edrop | 0.54 ± 0.02 | 0.54 ± 0.02 | 0.54 ± 0.02 | 0.54 ± 0.02 | 0.54 ± 0.02 | 0.54 ± 0.02 | 0.54 ± 0.02 | 0.54 ± 0.02 |
| GCN | 0.40 ± 0.03 | 0.40 ± 0.03 | 0.40 ± 0.03 | 0.40 ± 0.03 | 0.40 ± 0.03 | 0.40 ± 0.03 | 0.40 ± 0.03 | 0.40 ± 0.03 |
| GCN+edrop | 0.65 ± 0.04 | 0.65 ± 0.04 | 0.65 ± 0.04 | 0.65 ± 0.04 | 0.65 ± 0.04 | 0.65 ± 0.04 | 0.65 ± 0.04 | 0.65 ± 0.04 |
| GIN | 0.73 ± 0.10 | 0.73 ± 0.10 | 0.73 ± 0.10 | 0.73 ± 0.10 | 0.73 ± 0.10 | 0.73 ± 0.10 | 0.73 ± 0.10 | 0.73 ± 0.10 |
| GIN+edrop | 0.73 ± 0.10 | 0.73 ± 0.10 | 0.73 ± 0.10 | 0.73 ± 0.10 | 0.73 ± 0.10 | 0.73 ± 0.10 | 0.73 ± 0.10 | 0.73 ± 0.10 |
| GraphSAGE | 0.44 ± 0.01 | 0.44 ± 0.01 | 0.44 ± 0.01 | 0.44 ± 0.01 | 0.44 ± 0.01 | 0.44 ± 0.01 | 0.44 ± 0.01 | 0.44 ± 0.01 |
| GraphSAGE+edrop | 0.44 ± 0.10 | 0.44 ± 0.10 | 0.44 ± 0.10 | 0.44 ± 0.10 | 0.44 ± 0.10 | 0.44 ± 0.10 | 0.44 ± 0.10 | 0.44 ± 0.10 |

### 8.6.1 Architecture and Training

Our algorithm serves as a preprocessing step for selective data augmentation. Given a single training hypergraph $\mathcal{H}$, the Algorithm 29 is applied and during training, the identified hyperedges of the symmetric induced subhypergraphs of $\mathcal{H}$ are randomly replaced with single hyperedges that cover all the nodes of each induced subhypergraph. Each symmetric subhypergraph has a $p = 0.5$ probability of being selected. To get a large set of symmetric subhypergraphs, we run 2 iterations of GWL-1.

We implement $h(S, H)$ from Equation 8.7 as follows. Upon extracting the node representations from the hypergraph neural network, we use a multi-layer-perceptron (MLP) on each node representation, sum across such compositions, then apply a final MLP layer after the aggregation. We use the binary cross entropy loss on this multi-node representation for training. We always use 5 layers of hyperGNN convolutions, a hidden dimension of 1024, and a learning rate of 0.01.

### 8.6.2 Higher Order Link Prediction Results

We show in Table 8.1 the comparison of PR-AUC scores amongst the baseline methods of HGNN, HGNNP, HNHN, HyperGCN, UniGIN, UniGAT, UniSAGE, their hyperedge dropped versions, and "Our" method, which preprocesses the hypergraph to break symmetry during training. For the hyperedge drop baselines, there is a uniform 50% chance of dropping any hyperedge. We use the Laplacian eigenmap [309] positional encoding on the clique expansion of the input hypergraph. This is common practice in (hyper)link prediction and required for using a hypergraph neural network on an unattributed hypergraph.

We show in Table 8.2 the PR-AUC scores on the AMHREST41. Along with hyperGNN architectures we use for the hypergraph experiments, we also compare with standard GNN architectures: APPNP [310], GAT [311], GCN2 [312], GCN [218], GIN [215], and GraphSAGE [217]. For every hyperGNN/GNN architecture, we also apply drop-edge [313] to the input graph and use this also as baseline. The number of layers of each GNN is set to 5 and the hidden dimension at 1024. For APPNP and GCN2, one MLP is used on the initial node positional encodings.

Overall, our method performs well across a diverse range of higher order network datasets. We observe that our method can often outperform the baseline of not performing any data perturbations as well as the same baseline with uniformly random hyperedge dropping. Our method has an added advantage of being explainable since our algorithm works at the data level. There was also not much of a concern for computational time since our algorithm runs in time $O(nnz(H) + n + m)$, which is optimal since it is the size of the input.

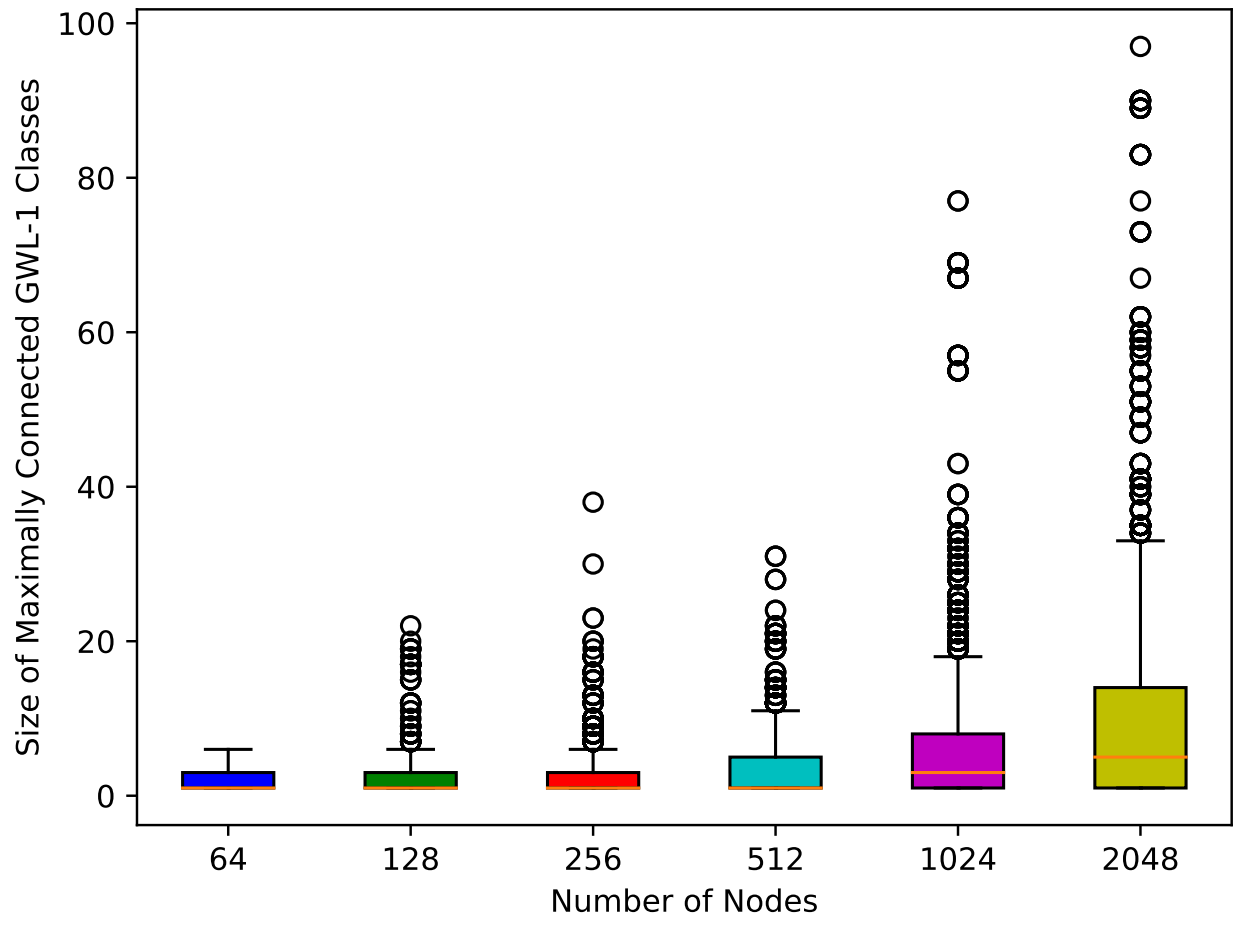

(a) Boxplot of the sizes of the connected components with equal GWL-1 node values from the hy-MMSBM sampling algorithm where there are three independent communities.

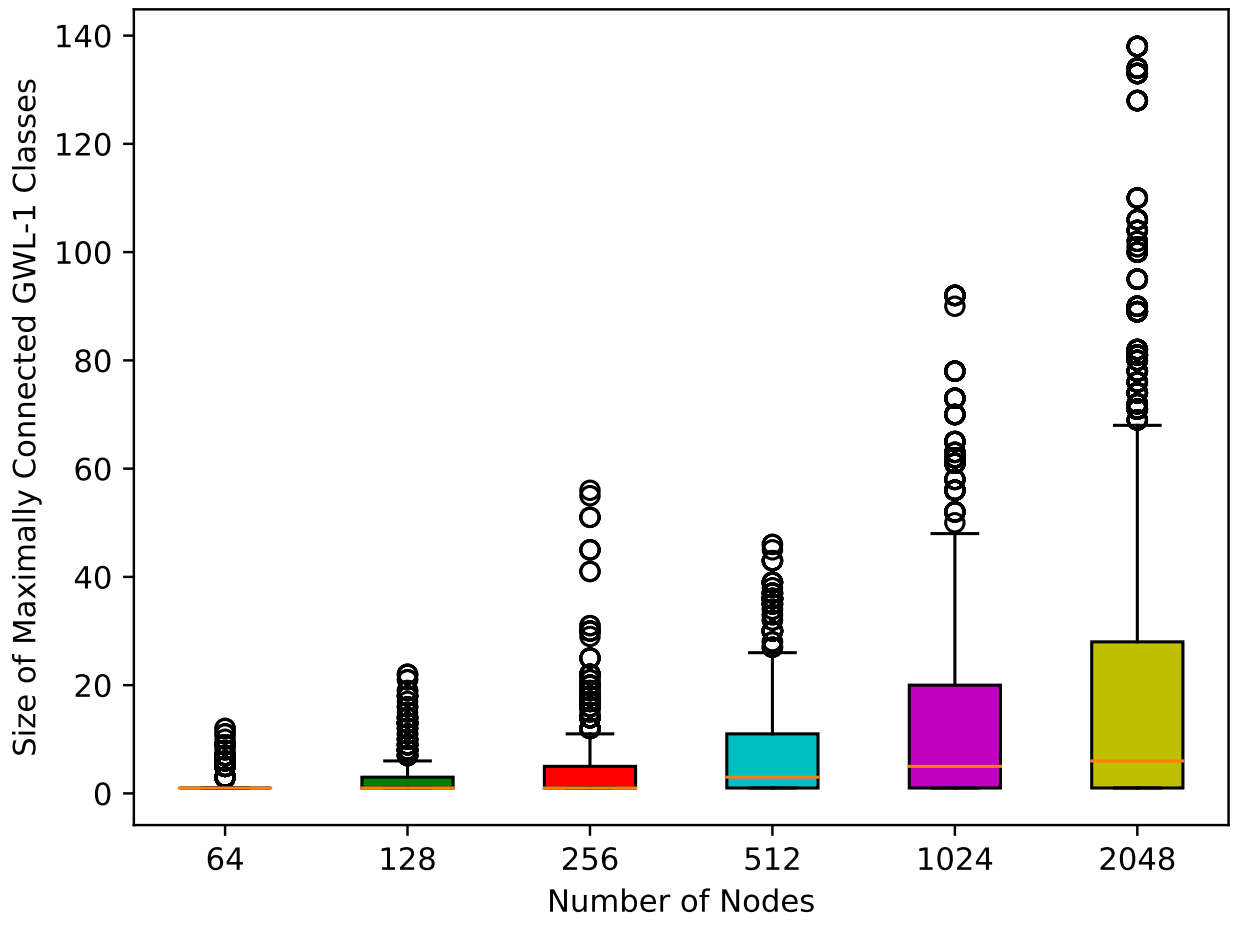

(b) Boxplot of the sizes of the connected components with equal GWL-1 node values from the hy-MMSBM sampling algorithm where any two of the three communities can communicate.

**Figure 8.4.** Experiment on the relationship between the sizes of connected components of equal GWL-1 node values and the communication between communities.

### 8.6.3 Empirical Observations on the Components Discovered by the Algorithm

According to Proposition 8.5.7, we know that the symmetry finding algorithm always covers the hypergraph and thus that we can generate on the order of $O(2^{|\mathcal{V}|})$ counterfactual multi-hypergraphs. It is known that a large set of data augmentations during learning improves learner generalization. The cover by GWL-1 symmetric components follows a distribution depending on the data.

We show in Figure 8.4 the distributions for the component sizes over all GWL-1 symmetric connected components for samples from the Hy-MMSBM model [314]. This is a model to sample hypergraphs with community structure. In Figure 8.4a we sample hypergraphs with 3 isolated communities, meaning that there is 0 chance of any interconnections between any two communities. In Figure 8.4b we sample hypergraphs with 3 communities where every node in a community has a weight of 1 to stay in its community and a weight of 0.2 to move out to any other community. We plot the boxplots as a function of increasing number of nodes. We notice that the more communication there is between communities for more nodes there is more spread in possible connected component sizes. Isolated communities should make for predictable clusters/connected components.

## 8.7 Discussion

Our proposed data augmentation method uses symmetry breaking to handle the symmetries induced by GWL-1 based hyperGNNs. Symmetry breaking provides some guarantees that the symmetries induced by the hyperGNN are brought closer to the symmetries of the training hypergraph. These include hyperlink prediction guarantees. In addition, symmetry breaking prepares the hyperGNN for an automorphism group different from the symmetries it views during training. This brings about an important question regarding the invariant automorphism group across training and testing, namely:

*What is the relationship between the symmetries of the training and testing hypergraphs?*

We answer this question in the context of a temporal shift from training to testing. The term "temporal shift" is usually used in the context of distribution shifts [315]. We define it in terms of transductive hyperlink prediction.

A temporal shift from training to testing means that the testing hyperedges have a temporal future relationship with the training hyperedges.

We formalize a hypergraph in terms of a physical system that can change over time with the following assumptions.

**Assumption 8.7.1.** *If we view a hypergraph as a physical system of particles where only the nodes have mass, then the task of transductive higher order link prediction is on a closed system [316]. This means no mass can enter or leave this system.*

*Given n nodes $\mathcal{V}$, we can view each node $v \in \mathcal{V}$ along with the hypergraph $\mathcal{H} = (\mathcal{V}, \mathcal{E})$ it belongs in as a microstate $(v, \mathcal{H})$ of a statistical ensemble.*

*In [317] the entropy of a graph is defined through the orbits of the automorphism group. We generalize this to a node based definition for hypergraphs. Our definition uses a similar probability on microstates. We define the **hypergraph topological entropy** of a hypergraph $\mathcal{H} = (\mathcal{V}, \mathcal{E})$ by:*

$$\begin{aligned} S &\triangleq -\sum_{v \in \mathcal{V}} p_v(\mathcal{H}) \log(p_v(\mathcal{H})); \\ \text{with } p_v(\mathcal{H}) &\triangleq \frac{n_v(\mathcal{H})}{\sum_{v \in \mathcal{V}} n_v(\mathcal{H})} \end{aligned} \tag{8.20}$$

*where $n_v(\mathcal{H}) \triangleq |\{u \in \mathcal{V} : u \cong_{\mathcal{H}} v\}|$ is the number of nodes $u \in \mathcal{V}$ isomorphic to node $v$, including $v$ itself, (See Definition 8.2.10).*

**Assumption 8.7.2.** ***(Second Law of Thermodynamics )**: The second law of thermodynamics [318] states that the entropy of a closed system must increase over time. This is denoted by the following equation:*

$$\Delta S > 0 \tag{8.21}$$

*This law can be used in terms of the hypergraph topological entropy as defined in Assumption 8.7.1.*

We also assume the following random model on the hypergraph we predict on. Let $\bullet = \text{tr}, \text{te}$ represent temporally ordered training and testing distributions where $(\mathcal{E}_\bullet)_{gt}$ are the hyperedges of $(\mathcal{H}_\bullet)_{gt}$.

**Assumption 8.7.3.** *For each node $v \in \mathcal{V}$, let $X_v$ be independent Bernoulli random variables of some probability $q_v \in [0,1]$. Let $|(\mathcal{E}_{\text{te}})_{gt}|$ be the number of testing hyperedges of $(\mathcal{H}_{\text{te}})_{gt}$. It is a random variable defined by:*

$$f((X_v)_{v \in \mathcal{V}}) = \sum_{\mathrm{e} \subseteq \mathcal{V}} \Pi_{u \in \mathrm{e}} X_u \tag{8.22}$$

*Assume further that $|(\mathcal{E}_{\text{te}})_{gt}|, (\mathcal{V}, (\mathcal{E}_{\text{te}})_{gt}) \sim P(\mathcal{H}; t_{\text{te}})$ only depends on $n$.*

*Let $\tilde{X}_v$ be a Bernoulli random variable with probability $q_{max}$. Assume that $f$ satisfies*

$$P(f(\tilde{X}_{v \neq u}, ..., do(\tilde{X}_u = 1), ..., \tilde{X}_{v \neq u}) - f(\tilde{X}_{v \neq u}, ..., do(\tilde{X}_u = 0), ..., \tilde{X}_{v \neq u}) \leq 2) \leq n^{-\omega(1)}, \forall u \in \mathcal{V} \tag{8.23}$$

These assumptions show that with high probability there are node isomorphism classes which decrease in size:

**Theorem 8.7.4.** *Under Assumptions 8.7.1, 8.7.3, and the Second Law of Thermodynamics on the hypergraph viewed as a closed system:*

$$\exists \mathcal{U} \subseteq \mathcal{V}, \mathcal{U} \neq \varnothing, \text{ so that: } n_v((\mathcal{H}_{tr})_{gt}) > n_v((\mathcal{H}_{\text{te}})_{gt}), \forall v \in \mathcal{U}, \text{ with probability } 1 - O(\frac{1}{\sqrt{n}}) \tag{8.24}$$

Theorem 8.7.4 states that the ground truth node isomorphism classes for the nodes must shrink with probability on order $1 - O(\frac{1}{\sqrt{n}})$. Thus, for large $n \gg 0$, we have that with high probability that the nodes $v \in \mathcal{U} \subseteq \mathcal{V}, \mathcal{U} \neq \varnothing$ have $n_v((\mathcal{H}_{tr})_{gt}) > n_v((\mathcal{H}_{\text{te}})_{gt})$.

We can recognize this property of $(\mathcal{H}_{\text{te}})_{gt}$ by shrinking the automorphism group that the hypergraph encoder recognizes from the training hypergraph. According to Proposition 8.10.40, symmetry breaking as given in Equation 8.18, does this.

Our method breaks the symmetry of a GWL-1 based hyperGNN. This, of course, is not of importance if the symmetry group of the GWL-1 based hyperGNN on some training hypergraph is already the trivial group. Nonetheless, our symmetry breaking method is theoretically beneficial. Our method can also be used within other downstream learning methods such as feature averaging [319], and ensemble methods, as mentioned in Section 8.5.

### 8.7.5 The Temporal Shift

We can summarize the distribution shift in the tranductive hypergraph learning problem of Problem 8.5.1 in the following diagram:

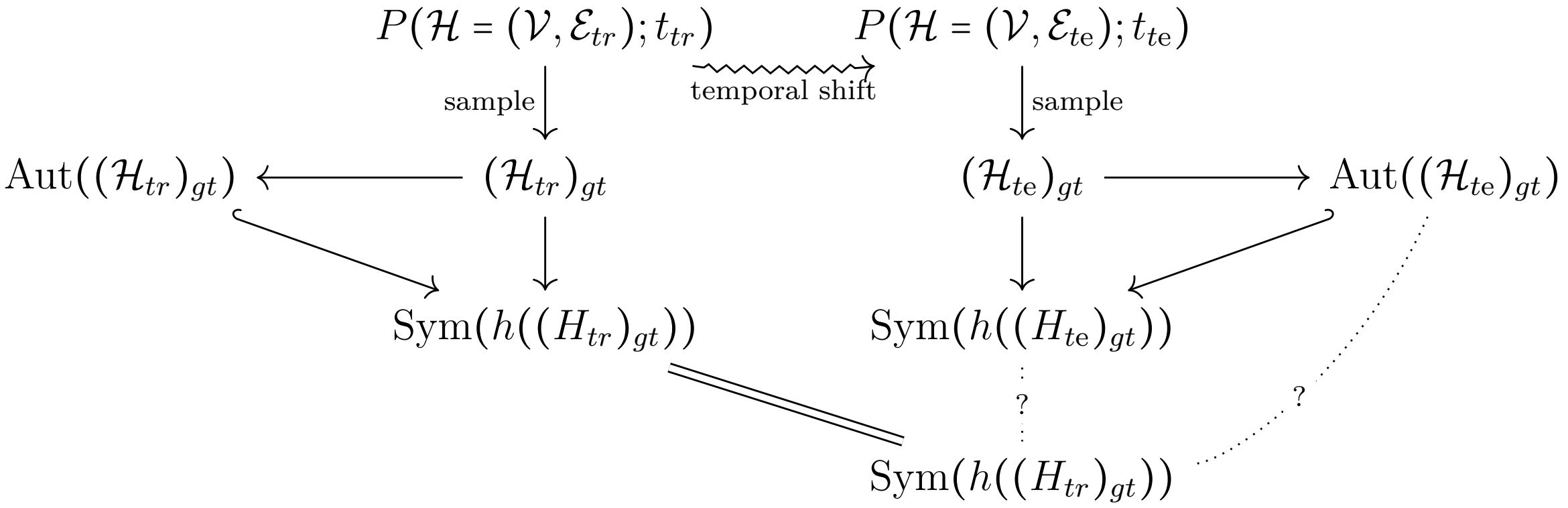

(8.25)

In Diagram 8.25, we have at the top the temporal shift in the hypergraph distribution from training to testing. For a training and testing hypergraph, we can form their automorphism groups. For a given hypergraph encoder, we also have the related symmetry group of some learnable encoder on the training and testing hypergraphs.

During testing, the symmetry group of the encoder is time-lagged, or **persisting**. Thus the encoder has an incorrect view of testing data.

## 8.8 Conclusion

Many existing hyperGNN architectures are based on the GWL-1 algorithm, which is a hypergraph isomorphism testing algorithm. We have characterized and identified the limitations of GWL-1. GWL-1 views the hypergraph as a collection of rooted trees. This means that hyperGNNs recognize more symmetries than the natural automorphisms of the training hypergraph. In fact, maximally connected subsets of nodes that share the same value of GWL-1, which act like regular hypergraphs, are indistinguishable. To address this issue while respecting the structure of a hypergraph, we have devised a preprocessing algorithm that identifies all such connected components. These components cover the hypergraph and allow for downstream data augmentation of symmetry breaking by training on a random multi-hypergraph. We show that this approach improves the expressivity of a hyperGNN learner, including in the case of hyperlink prediction. We perform extensive experiments to evaluate the effectiveness of our approach and make empirical observations about the output of the algorithm on hypergraph data.

## 8.9 More Background

We discuss in this section about the basics of graph representation learning and link prediction. Graphs are hypergraphs with all hyperedges of size 2. Simplicial complexes and hypergraphs are generalizations of graphs. We also discuss more related work.

### 8.9.1 Graph Neural Networks and Weisfeiler-Lehman 1

The Weisfeiler-Lehman (WL-1) algorithm is an isomorphism testing approximation algorithm. It involves repeatedly message passing all nodes with their neighbors, a step called node label refinement. The WL-1 algorithm never gives false negatives when predicting whether two graphs are isomorphic. In other words, two isomorphic graphs are always indistinguishable by WL-1.

The WL-1 algorithm is the following successive vertex relabeling applied until convergence on a graph $G = (X, A)$ (a pair of the set of node attributes and the graph's adjacency structure):

$$\begin{aligned} h_v^0 &\leftarrow X_v, \forall v \in \mathcal{V}_G \\ h_v^{i+1} &\leftarrow \{\!\!\{(h_v^i, h_u^i)\}\!\!\}_{u \in Nbr_A(v)}, \forall v \in \mathcal{V}_G \end{aligned} \tag{8.26}$$

The algorithm terminates after the vertex labels converge. For graph isomorphism testing, the concatenation of the histograms of vertex labels for each iteration is output as the graph representation. Since we are only concerned with node isomorphism classes, we ignore this step and just consider the node labels $h_v^i$ for every $v \in \mathcal{V}_\mathcal{C}$.

The WL-1 isomorphism test can be characterized in terms of rooted tree isomorphisms between the universal covers for connected graphs [301]. There have also been characterizations of WL-1 in terms of counting homomorphisms [320] as well as the Wasserstein Distance [321] and Markov chains [322].

A graph neural network (GNN) is a message passing based node representation learner modeled after the WL-1 algorithm. It has the important inductive bias of being equivariant to node indices. As a neural model of the WL-1 algorithm, it learns neural weights common across all nodes in order to obtain a vector representation for each node. A GNN must use some initial node attributes in order to update its neural weights. There are many variations on GNNs, including those that improve the distinguishing power beyond WL-1. For two surveys on the GNNs and their applications, see [216, 323].

### 8.9.2 Link Prediction

The task of link prediction on graphs involves the prediction of the existence of links. There are two kinds of link prediction. There is transductive link prediction where the same nodes are used for all of train validation and testing. There is also inductive link prediction where the test validation and training nodes can all be disjoint. Some existing works on link prediction include [324]. Higher order link prediction is a generalization of link prediction to hypergraph data.

A common way to do link prediction is to compute a node-based GNN and for a pair of nodes, aggregate, similar to in graph auto encoders [297], the node representations in any target pair in order to obtain a 2-node representation. Such aggregations are of the form:

$$h(S = \{u, v\}) = \sigma(h_u \cdot h_v) \tag{8.27}$$

where $S$ is a pair of nodes. As shown in Proposition 8.10.14, this guaranteems an equivariant 2-node representation but can often give false predictions even with a fully expressive node-based GNN [325]. A common remedy for this problem is to introduce positional encodings such as SEAL [326] and DistanceEncoding [256]. Positional encodings encode the relative distances amongst nodes via a low distortion embedding for example. In the related work section we have gone over many of these embeddings. We have also used these in our evaluation since they are common practice and must exist to compute a hypergraph neural network if there are no ground truth node attributes. According to [327], fully expressive pairwise node representations, as defined by 2-node invariance and expressivity, can be represented by some fully expressive positional embedding, which is a positional embedding that is injective on the node pair isomorphism classes. It is not clear how one would achieve this in practice, however. Another remedy is to increase the expressive power of WL-1 to WL-2 for link prediction [328].

### 8.9.3 More Related Work

The work of [329] also does a data augmentation scheme. It considers randomly dropping edges and generating data through a generative model on hypergraphs. The work of [330] also performs data augmentation on a hypergraph so that homophilic relationships are maintained. It does this through contrastive losses at the node to node, hyperedge to hyperedge and intra hyperedge level. Neither of these methods provide guarantees for their data augmentations.

As mentioned in the main text, an ensemble of neural networks can be used with a dropout [331] like method on the output of the Algorithm. Subgraph neural networks [304, 306] are ensembles of models on subgraphs of the input graph.

Some more of the many existing hypergraph neural network architectures include: [332–337].

## 8.10 Proofs

In this section we provide the proofs for all of the results in the main paper along with some additional theory.

### 8.10.1 Hypergraph Isomorphism

We first repeat the definition of a hypergraph and its corresponding matrix representation called the star expansion matrix::

**Definition 8.10.2.** *An undirected hypergraph is a pair $\mathcal{H} = (\mathcal{V}, \mathcal{E})$ consisting of a set of vertices $\mathcal{V}$ and a set of hyperedges $\mathcal{E} \subseteq 2^{\mathcal{V}}$ where $2^{\mathcal{V}}$ is the power set of the vertex set $\mathcal{V}$.*

**Definition 8.10.3.** *The star expansion incidence matrix $H$ of a hypergraph $\mathcal{H} = (\mathcal{V}, \mathcal{E})$ is the $|\mathcal{V}| \times 2^{|\mathcal{V}|}$ 0-1 incidence matrix $H$ where $H_{v,\mathrm{e}} = 1$ iff $v \in \mathrm{e}$ for $(v, \mathrm{e}) \in \mathcal{V} \times \mathcal{E}$ for some fixed orderings on both $\mathcal{V}$ and $2^{\mathcal{V}}$.*

We recall the definition of an isomorphism between hypergraphs:

**Definition 8.10.4.** *For two hypergraphs $\mathcal{H}$ and $\mathcal{D}$, a structure preserving map $\rho : \mathcal{H} \to \mathcal{D}$ is a pair of maps $\rho = (\rho_{\mathcal{V}} : \mathcal{V}_{\mathcal{H}} \to \mathcal{V}_{\mathcal{D}}, \rho_{\mathcal{E}} : \mathcal{E}_{\mathcal{H}} \to \mathcal{E}_{\mathcal{D}})$ such that $\forall \mathrm{e} \in \mathcal{E}_{\mathcal{H}}, \rho_{\mathcal{E}}(\mathrm{e}) \triangleq \{\rho_{\mathcal{V}}(v_{\mathrm{i}}) : v_{\mathrm{i}} \in \mathrm{e}\} \in \mathcal{E}_{\mathcal{D}}$. A hypergraph isomorphism is a structure preserving map $\rho = (\rho_{\mathcal{V}}, \rho_{\mathcal{E}})$ such that both $\rho_{\mathcal{V}}$ and $\rho_{\mathcal{E}}$ are bijective. Two hypergraphs are said to be isomorphic, denoted as $\mathcal{H} \cong \mathcal{D}$, if there exists an isomorphism between them. When $\mathcal{H} = \mathcal{D}$, an isomorphism $\rho$ is called an automorphism on $\mathcal{H}$. All the automorphisms form a group, which we denote as $Aut(\mathcal{H})$.*

The action of $\pi \in \mathrm{Sym}(\mathcal{V})$ on the star expansion adjacency matrix $H$ is repeated here for convenience:

$$(\pi \cdot H)_{v,\mathrm{e}=(u_1 \ldots v \ldots u_k)} \triangleq H_{\pi^{-1}(v),\pi^{-1}(\mathrm{e})=(\pi^{-1}(u_1)\ldots\pi^{-1}(v)\ldots\pi^{-1}(u_k))} \tag{8.28}$$

Based on the group action, consider the stabilizer subgroup of $\mathrm{Sym}(\mathcal{V})$ on the star expansion adjacency matrix $H$ defined as follows:

$$Stab_{\mathrm{Sym}(\mathcal{V})}(H) = \{\pi \in \mathrm{Sym}(\mathcal{V}) : \pi \cdot H = H\} \tag{8.29}$$

For simplicity we omit the lower index when the permutation group is clear from the context. It can be checked that $Stab(H) \leq \mathrm{Sym}(\mathcal{V})$ is a subgroup. Intuitively, $Stab(H)$ consists of all permuations that leave $H$ fixed.

For a given hypergraph $\mathcal{H} = (\mathcal{V}, \mathcal{E})$, there is a relationship between the group of hypergraph automorphisms $Aut(\mathcal{H})$ and the stabilizer group $Stab(H)$ on the star expansion adjacency matrix.

**Proposition 8.10.5.** *$Aut(\mathcal{H}) \cong Stab(H)$ are equivalent as isomorphic groups.*

*Proof.* Consider $\rho \in Aut(\mathcal{H})$, define the map $\Phi : \rho \mapsto \pi := \rho \mid_{\mathcal{V}(\mathcal{H})}$. The group element $\pi \in \mathrm{Sym}(\mathcal{V})$ acts as a stabilizer of $H$ since for any entry $(v, \mathrm{e})$ in $H$, $H_{\pi^{-1}(v),\pi^{-1}(\mathrm{e})} = (\pi \cdot H)_{v,\mathrm{e}} = 1$ iff $\pi^{-1}(\mathrm{e}) \in \mathcal{E}_{\mathcal{H}}$ iff $\mathrm{e} \in \mathcal{E}_{\mathcal{H}}$ iff $H_{v,\mathrm{e}} = 1 = H_{\pi\circ\pi^{-1}(v),\pi\circ\pi^{-1}(\mathrm{e})}$. Since $(v, \mathrm{e})$ was arbitrary, $\pi$ preserves the positions of the nonzeros.

We can check that $\Phi$ is a well defined injective homorphism as a restriction map. Furthermore it is surjective since for any $\pi \in Stab(H)$, we must have $H_{v,\mathrm{e}} = 1$ iff $(\pi \cdot H)_{v,\mathrm{e}} = H_{\pi^{-1}(v),\pi^{-1}(\mathrm{e})} = 1$ which is equivalent to $v \in \mathrm{e} \in \mathcal{E}$ iff $\pi(v) \in \pi(\mathrm{e}) \in \mathcal{E}$ which implies $\mathrm{e} \in \mathcal{E}$ iff $\pi(\mathrm{e}) \in \mathcal{E}$. Thus $\Phi$ is a group isomorphism from $Aut(\mathcal{H})$ to $Stab(H)$ □

In other words, to study the symmetries of a given hypergraph $\mathcal{H}$, we can equivalently study the automorphisms $Aut(\mathcal{H})$ and the stabilizer permutations $Stab(H)$ on its star expansion adjacency matrix $H$. Intuitively, the stabilizer group $0 \leq Stab(H) \leq \mathrm{Sym}(\mathcal{V})$ characterizes the symmetries in a graph. When the graph has rich symmetries, say a complete graph, $Stab(H) = \mathrm{Sym}(\mathcal{V})$ can be as large as the whole permutaion group.

Nontrivial symmetries can be represented by isomorphic node sets which we define as follow:

**Definition 8.10.6.** *For a given hypergraph $\mathcal{H}$ with star expansion matrix $H$, two $k$-node sets $S, T \subseteq \mathcal{V}$ are called* isomorphic*, denoted as $S \simeq T$, if $\exists \pi \in Stab(H), \pi(S) = T$ and $\pi(T) = S$.*

When $k = 1$, we have isomorphic nodes, denoted $u \cong_{\mathcal{H}} v$ for $u, v \in \mathcal{V}$. Node isomorphism is also studied as the so-called structural equivalence in [274]. Furthermore, if $S \simeq T$ we can then say that there is a matching amongst the nodes in the two node subsets so that matched nodes are isomorphic.

**Definition 8.10.7.** *A $k$-node representation $h$ is **k-permutation equivariant** if:*

*for all $\pi \in Sym(\mathcal{V})$, $S \in 2^{\mathcal{V}}$ with $|S| = k$: $h(\pi \cdot S, H) = h(S, \pi \cdot H)$*

**Proposition 8.10.8.** *If $k$-node representation $h$ is $k$-permutation equivariant, then $h$ is $k$-node invariant.*

*Proof.* given $S, S' \in \mathcal{C}$ with $|S| = |S'| = k$,

if there exists a $\pi \in Stab(H)$ (meaning $\pi \cdot H = H$) and $\pi(S) = S'$ then

$$\begin{aligned} h(S', H) &= h(S', \pi \cdot H) \text{ (by } \pi \cdot H = H) \\ &= h(S, H) \text{ (by } k\text{-permutation equivariance of } h \text{ and } \pi(S) = S') \end{aligned} \tag{8.30}$$

□

We revisit the definition of the symmetry group of a $k$-node representation map on hypergraph $\mathcal{H}$.

**Definition 8.10.9.** *For $h : [\mathcal{V}]^k \times \mathbb{Z}_2^{n \times 2^n} \to \mathbb{R}^d$ a $k$-node representation on a hypergraph $\mathcal{H}$,*

$$Sym(h) \triangleq \{\pi \in Sym(\mathcal{V}) : \pi(S) = S' \Rightarrow h(S, H) = h(S', H)\} \tag{8.31}$$

### 8.10.10 Properties of GWL-1

Here are the steps of the GWL-1 algorithm on the star expansion matrix $H$ is repeated here for convenience:

$$\begin{aligned} f_{\mathrm{e}}^{0} &\leftarrow \{\}, h_v^0 \leftarrow \{\} \\ f_{\mathrm{e}}^{\mathrm{i}+1} &\leftarrow \{\!\{(f_{\mathrm{e}}^{\mathrm{i}}, h_v^{\mathrm{i}})\}\!\}_{v \in \mathrm{e}}, \forall \mathrm{e} \in \mathcal{E}(H) \\ h_v^{\mathrm{i}+1} &\leftarrow \{\!\{(h_v^{\mathrm{i}}, f_{\mathrm{e}}^{\mathrm{i}+1})\}\!\}_{v \in \mathrm{e}}, \forall v \in \mathcal{V}(H) \end{aligned} \tag{8.32}$$

Where $\mathcal{E}(H)$ denotes the nonzero columns of $H$ and $\mathcal{V}(H)$ denotes the rows of $H$.

We make the following observations about each of the two steps of the GWL-1 algorithm:

**Observation 8.10.11.**

$$\{\!\{(f_{\mathrm{e}}^{\mathrm{i}}, h_v^{\mathrm{i}})\}\!\}_{v \in \mathrm{e}} = \{\!\{({f'}_{\mathrm{e}}^{\mathrm{i}}, {h'}_v^{\mathrm{i}})\}\!\}_{v \in \mathrm{e}} \textit{ iff } (f_{\mathrm{e}}^{\mathrm{i}}, \{\!\{h_v^{\mathrm{i}}\}\!\}_{v \in \mathrm{e}}) = ({f'}_{\mathrm{e}}^{\mathrm{i}}, \{\!\{{h'}_v^{\mathrm{i}}\}\!\}_{v \in \mathrm{e}}) \forall \mathrm{e} \in \mathcal{E}(H) \textit{ and} \tag{8.33a}$$

$$\{\!\{(h^{\mathrm{i}}_v, f^{\mathrm{i+1}}_{\mathrm{e}}\}\!\}_{v\in\mathrm{e}} = \{\!\{(h'^{\mathrm{i}}_v, f'^{\mathrm{i+1}}_{\mathrm{e}}\}\!\}_{v\in\mathrm{e}} \textit{ iff } (h^{\mathrm{i}}_v, \{\!\{f^{\mathrm{i+1}}_{\mathrm{e}}\}\!\}_{v\in\mathrm{e}}) = (h'^{\mathrm{i}}_v, \{\!\{f'^{\mathrm{i+1}}_{\mathrm{e}}\}\!\}_{v\in\mathrm{e}}) \forall v \in \mathcal{V}(H) \tag{8.33b}$$

*Proof.* Equation 8.33a follows since

$$\{\!\{(f^{\mathrm{i}}_{\mathrm{e}}, h^{\mathrm{i}}_v)\}\!\}_{v\in\mathrm{e}} = \{\!\{(f'^{\mathrm{i}}_{\mathrm{e}}, h'^{\mathrm{i}}_v)\}\!\}_{v\in\mathrm{e}} \forall \mathrm{e} \in \mathcal{E}(H) \tag{8.34a}$$

$$\text{iff } f^{\mathrm{i}}_{\mathrm{e}} = f'^{\mathrm{i}}_{\mathrm{e}} \text{ and } \{\!\{h^{\mathrm{i}}_v\}\!\}_{v\in\mathrm{e}} = \{\!\{h'^{\mathrm{i}}_v\}\!\}_{v\in\mathrm{e}} \forall \mathrm{e} \in \mathcal{E}(H) \tag{8.34b}$$

$$\text{iff } (f^{\mathrm{i}}_{\mathrm{e}}, \{\!\{h^{\mathrm{i}}_v\}\!\}_{v\in\mathrm{e}}) = (f'^{\mathrm{i}}_{\mathrm{e}}, \{\!\{h'^{\mathrm{i}}_v\}\!\}_{v\in\mathrm{e}}) \forall \mathrm{e} \in \mathcal{E}(H) \tag{8.34c}$$

For Equation 8.33b, we have:

$$\{\!\{(h^{\mathrm{i}}_v, f^{\mathrm{i+1}}_{\mathrm{e}}\}\!\}_{v\in\mathrm{e}} = \{\!\{(h'^{\mathrm{i}}_v, f'^{\mathrm{i+1}}_{\mathrm{e}}\}\!\}_{v\in\mathrm{e}} \forall v \in \mathcal{V}(H) \tag{8.35a}$$

$$\text{iff } \{\!\{(h^{\mathrm{i}}_v, \{\!\{(f^{\mathrm{i}}_{\mathrm{e}}, h^{\mathrm{i}}_u)\}\!\}_{u\in\mathrm{e}})\}\!\}_{v\in\mathrm{e}} = \{\!\{(h'^{\mathrm{i}}_v, \{\!\{(f'^{\mathrm{i}}_{\mathrm{e}}, h'^{\mathrm{i}}_u)\}\!\}_{u\in\mathrm{e}})\}\!\}_{v\in\mathrm{e}} \forall v \in \mathcal{V}(H) \tag{8.35b}$$

$$\text{iff } h^{\mathrm{i}}_v = h'^{\mathrm{i}}_v \text{ and } \{\!\{(f^{\mathrm{i}}_{\mathrm{e}}, h^{\mathrm{i}}_u)\}\!\}_{u\in\mathrm{e}, v\in\mathrm{e}} = \{\!\{(f'^{\mathrm{i}}_{\mathrm{e}}, h'^{\mathrm{i}}_u)\}\!\}_{u\in\mathrm{e}, v\in\mathrm{e}} \forall v \in \mathcal{V}(H) \tag{8.35c}$$

$$\text{iff } h^{\mathrm{i}}_v = h'^{\mathrm{i}}_v \text{ and } \{\!\{f^{\mathrm{i+1}}_{\mathrm{e}}\}\!\} = \{\!\{f'^{\mathrm{i+1}}_{\mathrm{e}}\}\!\} \forall v \in \mathcal{V}(H) \tag{8.35d}$$

These follow by the definition of multiset equality and since there is no loss of information upon factoring out a constant tuple entry of each pair in the multisets. □

**Proposition 8.10.12.** *The update steps of GWL-1:* $f^{\mathrm{i}}(H) \triangleq [f^{\mathrm{i}}_{\mathrm{e}_1}(H), \cdots f^{\mathrm{i}}_{\mathrm{e}_m}(H)]$ *and* $h^{\mathrm{i}}(H) \triangleq [h^{\mathrm{i}}_{v_1}(H), \cdots h^{\mathrm{i}}_{v_n}(H)]$*, are permutation equivariant; in other words, For any* $\pi \in Sym(\mathcal{V})$*, let* $\pi \cdot f^{\mathrm{i}}(H) \triangleq [f^{\mathrm{i}}_{\pi^{-1}(\mathrm{e}_1)}(H), \cdots, f^{\mathrm{i}}_{\pi^{-1}(\mathrm{e}_m)}(H)]$ *and* $\pi \cdot h^{\mathrm{i}}(H) \triangleq [h^{\mathrm{i}}_{\pi^{-1}(v_1)}(H), \cdots h^{\mathrm{i}}_{\pi^{-1}(v_n)}(H)]$*, we have* $\forall \mathrm{i} \in \mathbb{N}, \pi \cdot f^{\mathrm{i}}(H) = f^{\mathrm{i}}(\pi \cdot H)$ *and* $\pi \cdot h^{\mathrm{i}}(H) = h^{\mathrm{i}}(\pi \cdot H)$

*Proof.* We prove by induction on i:

Base case, $\mathrm{i} = 0$:

$[\pi \cdot f^0(H)]_{\mathrm{e}=\{v_1 \ldots v_k\}} = \{\} = f^0_{\pi^{-1}(\mathrm{e})=\{\pi^{-1}(v_1) \ldots \pi^{-1}(v_k)\}}(H) = f^0_{\mathrm{e}}(\pi \cdot H)$ since the $\pi$ cannot affect a list of empty sets and the definition of the action of $\pi$ on $H$ as defined in Equation 8.28.

$[\pi \cdot h^0(H)]_v = [\pi \cdot X]_v = X_{\pi^{-1}(v)} = h^0_{\pi^{-1}(v)}(H) = h^0_v(\pi \cdot H)$ by definition of the group action $\mathrm{Sym}(\mathcal{V})$ acting on the node indices of a node attribute tensor as defined in Equation 8.28.

Induction Hypothesis:

$$[\pi \cdot f^{i}(H)]_{e} = f^{i}_{\pi^{-1}(e)}(H) = f^{i}_{e}(\pi \cdot H) \text{ and } [\pi \cdot h^{i}(H)]_{v} = h^{i}_{\pi^{-1}(v)}(H) = h^{i}_{v}(\pi \cdot H) \tag{8.36}$$

Induction Step:

$$\begin{aligned}
[\pi \cdot h^{i+1}(H)]_{v} &= \{\!\{([\pi \cdot h^{i}(H)]_{v}, [\pi \cdot f^{i+1}(H)]_{e})\}\!\}_{v \in e} \\
&= \{\!\{([\pi \cdot h^{i}(H)]_{v}, \{\!\{([\pi \cdot f^{i}(H)]_{e}, [\pi \cdot h^{i}(H)]_{u})\}\!\}_{u \in e})\}\!\}_{v \in e} \\
&= \{\!\{h^{i}_{v}(\pi \cdot H), \{\!\{(f^{i}_{e}(\pi \cdot H), h^{i}_{u}(\pi \cdot H))\}\!\}_{u \in e}\}\!\}_{v \in e} \\
&= h^{i+1}_{v}(\pi \cdot H)
\end{aligned} \tag{8.37}$$

$$\begin{aligned}
[\pi \cdot f^{i+1}(H)]_{e} &= \{\!\{([\pi \cdot f^{i}(H)]_{e}, [\pi \cdot h^{i}(H)]_{v})\}\!\}_{v \in e} \\
&= \{\!\{(f^{i}_{e}(\pi \cdot H), h^{i}_{v}(\pi \cdot H)\}\!\}_{v \in e} \\
&= f^{i+1}_{e}(\pi \cdot H)
\end{aligned} \tag{8.38}$$

□

**Definition 8.10.13.** *Let $h : [\mathcal{V}]^{k} \times \mathbb{Z}_2^{n \times 2^n} \to \mathbb{R}^d$ be a k-node representation on a hypergraph $\mathcal{H}$. Let $H \in \mathbb{Z}_2^{n \times 2^n}$ be the star expansion adjacency matrix of $\mathcal{H}$ for n nodes. The representation h is k-node most expressive if $\forall S, S' \subseteq \mathcal{V}$, $|S| = |S'| = k$, the following two conditions are satisfied:*

1. *h is **k-node invariant**: $\exists \pi \in Stab(H), \pi(S) = S' \implies h(S,H) = h(S',H)$*
2. *h is **k-node expressive** $\nexists \pi \in Stab(H), \pi(S) = S' \implies h(S,H) \neq h(S',H)$*

Let $AGG$ be a permutation invariant map from a set of node representations to $\mathbb{R}^d$.

**Proposition 8.10.14.** *Let $h(S,H) = AGG_{v \in S}[h^{i}_{v}(H)]$ with injective AGG and $h^{i}_{v}$ permutation equivariant. The representation $h(S,H)$ is k-node invariant but not necessarily k-node expressive for S a set of k nodes.*

*Proof.* $\exists \pi \in Stab(H)$ s.t. $\pi(S) = S', \pi \cdot H = H$

$\Rightarrow \pi(v_{\mathrm{i}}) = v'_{\mathrm{i}}$ for i = 1...$|S|, \pi \cdot H = H$

$\Rightarrow h^{\mathrm{i}}_{\pi(v)}(H) = h^{\mathrm{i}}_v(\pi \cdot H) = h^{\mathrm{i}}_v(H)$ (By permutation equivariance of $h^{\mathrm{i}}_v$ and $\pi \cdot H = H$)

$\Rightarrow AGG_{v \in S}[h^{\mathrm{i}}_v(H)] = AGG_{v' \in S'}[h^{\mathrm{i}}_{v'}(H)]$ (By Proposition 8.10.8 and AGG being permutation invariant)

The converse, that $h(S, H)$ is $k$-node expressive, is not necessarily true since we cannot guarantee $h(S, H) = h(S', H)$ implies the existence of a permutation that maps $S$ to $S'$ (see [338]). □

A hypergraph can be represented by a bipartite graph $\mathcal{B}_{\mathcal{V},\mathcal{E}}$ from $\mathcal{V}$ to $\mathcal{E}$ where there is an edge $(v, \mathrm{e})$ in the bipartite graph iff node $v$ is incident to hyperedge e. This bipartite graph $\mathcal{B}_{\mathcal{V},\mathcal{E}}$ is called the star expansion bipartite graph.

We introduce a more structured version of graph isomorphism called a 2-color isomorphism to characterize hypergraphs. It is a map on 2-colored graphs, which are graphs that can be colored with two colors so that no two nodes in any graph with the same color are connected by an edge. We define a 2-colored isomorphism formally here:

**Definition 8.10.15.** *A 2-colored isomorphism is a graph isomorphism on two 2-colored graphs that preserves node colors. In particular, between two graphs $G_1$ and $G_2$ the vertices of one color in $G_1$ must map to vertices of the same color in $G_2$. It is denoted by $\cong_c$.*

A bipartite graph must always have a 2-coloring. In fact, the 2-coloring with all the nodes in the node bipartition colored red and all the nodes in the hyperedge bipartition colored blue forms a canonical 2-coloring of $\mathcal{B}_{\mathcal{V},\mathcal{E}}$. Assume that all star expansion bipartite graphs are canonically 2-colored.

**Proposition 8.10.16.** *We have two hypergraphs $(\mathcal{V}_1, \mathcal{E}_1) \cong (\mathcal{V}_2, \mathcal{E}_2)$ iff $\mathcal{B}_{\mathcal{V}_1,\mathcal{E}_1} \cong_c \mathcal{B}_{\mathcal{V}_2,\mathcal{E}_2}$ where $\mathcal{B}_{\mathcal{V},\mathcal{E}}$ is the star expansion bipartite graph of $(\mathcal{V}, \mathcal{E})$*

*Proof.* Denote $L(\mathcal{B}_{\mathcal{V}_{\mathrm{i}},\mathcal{E}_{\mathrm{i}}})$ as the left hand (red) bipartition of $\mathcal{B}_{\mathcal{V}_{\mathrm{i}},\mathcal{E}_{\mathrm{i}}}$ to represent the nodes $\mathcal{V}_{\mathrm{i}}$ of $(\mathcal{V}_{\mathrm{i}}, \mathcal{E}_{\mathrm{i}})$ and $R(\mathcal{B}_{\mathcal{V}_{\mathrm{i}},\mathcal{E}_{\mathrm{i}}})$ as the right hand (blue) bipartition of $\mathcal{B}_{\mathcal{V}_{\mathrm{i}},\mathcal{E}_{\mathrm{i}}}$ to represent the hyperedges $\mathcal{E}_{\mathrm{i}}$ of $(\mathcal{V}_{\mathrm{i}}, \mathcal{E}_{\mathrm{i}})$. We use the left/right bipartition and $\mathcal{V}_{\mathrm{i}}/\mathcal{E}_{\mathrm{i}}$ interchangeably since they are in bijection.

$\Rightarrow$ If there is an isomorphism $\pi : \mathcal{V}_1 \rightarrow \mathcal{V}_2$, this means

- $\pi$ is a bijection and
- has the structure preserving property that $(u_1 ... u_k) \in \mathcal{E}_1$ iff $(\pi(u_1)...\pi(u_k)) \in \mathcal{E}_2$.

We may induce a 2-colored isomorphism $\pi^* : \mathcal{V}(\mathcal{B}_{\mathcal{V}_1,\mathcal{E}_1}) \rightarrow \mathcal{V}(\mathcal{B}_{\mathcal{V}_1,\mathcal{E}_1})$ so that $\pi^* |_{L(\mathcal{B}_{\mathcal{V}_1,\mathcal{E}_1})} = \pi$ where equality here means that $\pi^* |_{L(\mathcal{B}_{\mathcal{V}_1,\mathcal{E}_1})}$ acts on $L(\mathcal{B}_{\mathcal{V}_1,\mathcal{E}_1})$ the same way that $\pi$ does on $\mathcal{V}_1$. Furthermore $\pi^*$ has the property that $\pi^* |_{R(\mathcal{B}_{\mathcal{V}_1,\mathcal{E}_1})} (u_1 ... u_k) = (\pi(u_1)...\pi(u_k)), \forall (u_1 ... u_k) \in \mathcal{E}_1$, following the structure preserving property of isomorphism $\pi$.

The map $\pi^*$ is a bijection by definition of being an extension of a bijection.

The map $\pi^*$ is also a 2-colored map since it maps $L(\mathcal{B}_{\mathcal{V}_1,\mathcal{E}_1})$ to $L(\mathcal{B}_{\mathcal{V}_2,\mathcal{E}_2})$ and $R(\mathcal{B}_{\mathcal{V}_1,\mathcal{E}_1})$ to $R(\mathcal{B}_{\mathcal{V}_2,\mathcal{E}_2})$.

We can also check that the map is structure preserving and thus a 2-colored isomorphism since $(u_i, (u_1 ... u_i ... u_k)) \in \mathcal{E}(\mathcal{B}_{\mathcal{V}_1,\mathcal{E}_1}), \forall i = 1...k$ iff $(u_i \in \mathcal{V}_1$ and $(u_1 ... u_i ... u_k) \in \mathcal{E}_1)$ iff $\pi(u_i) \in \mathcal{V}_2$ and $(\pi(u_1)...\pi(u_i)...\pi(u_k)) \in \mathcal{E}_2$ iff $(\pi^*(u_i), (\pi^*(u_1, ... u_i, ... u_k)) \in \mathcal{E}(\mathcal{B}_{\mathcal{V}_2,\mathcal{E}_2}), \forall i = 1...k$. This follows from $\pi$ being structure preserving and the definition of $\pi^*$.

$\Leftarrow$ If there is a 2-colored isomorphism $\pi^* : \mathcal{B}_{\mathcal{V}_1,\mathcal{E}_1} \rightarrow \mathcal{B}_{\mathcal{V}_2,\mathcal{E}_2}$ then it has the properties that

- $\pi^*$ is a bijection,
- (is 2-colored): $\pi^* |_{L(\mathcal{B}_{\mathcal{V}_1,\mathcal{E}_1})} : L(\mathcal{B}_{\mathcal{V}_1,\mathcal{E}_1}) \rightarrow L(\mathcal{B}_{\mathcal{V}_2,\mathcal{E}_2})$ and $\pi^* |_{R(\mathcal{B}_{\mathcal{V}_1,\mathcal{E}_1})}: R(\mathcal{B}_{\mathcal{V}_1,\mathcal{E}_1}) \rightarrow R(\mathcal{B}_{\mathcal{V}_2,\mathcal{E}_2})$
- (it is structure preserving): $(u_i, (u_1 ... u_i ... u_k)) \in \mathcal{E}(\mathcal{B}_{\mathcal{V}_1,\mathcal{E}_1}), \forall i = 1...k$ iff $(\pi^*(u_i), \pi^*(u_1 ... u_i ... u_k)) \in \mathcal{E}(\mathcal{B}_{\mathcal{V}_2,\mathcal{E}_2}), \forall i = 1...k$ .

This then means that we may induce a $\pi : \mathcal{V}_1 \rightarrow \mathcal{V}_2$ so that $\pi = \pi^* |_{L(\mathcal{B}_{\mathcal{V}_1,\mathcal{E}_1})}$.

We can check that $\pi$ is a bijection since $\pi$ is the 2-colored bijection $\pi^*$ restricted to $L(\mathcal{B}_{\mathcal{V}_1,\mathcal{E}_1})$, thus remaining a bijection.

We can also check that $\pi$ is structure preserving. This means that $(u_1 ... u_k) \in \mathcal{E}_1$ iff $(u_i, (u_1 ... u_i ... u_k)) \in \mathcal{E}(\mathcal{B}_{\mathcal{V}_1,\mathcal{E}_1}) \forall i = 1...k$ iff $(\pi^*(u_i), (\pi^*(u_1 ... u_i ... u_k))) \in \mathcal{E}(\mathcal{B}_{\mathcal{V}_2,\mathcal{E}_2}) \forall i = 1...k$ iff $(\pi^*(u_1 ... u_k)) \in R(\mathcal{B}_{\mathcal{V}_2,\mathcal{E}_2})$ iff $(\pi(u_1)...\pi(u_k)) \in \mathcal{E}_2$ □

We define a topological object for a graph originally from algebraic topology called a universal cover:

**Definition 8.10.17.** *([300]) A universal covering of a connected graph $G$ is a (potentially infinite) graph $\tilde{G}$, s.t. there is a map $p_G : \tilde{G} \to G$ called the universal covering map where:*

1. *$\forall x \in \mathcal{V}(\tilde{G})$, $p_G|_{N(x)}$ is an isomorphism onto $N(p_G(x))$.*
2. *$\tilde{G}$ is simply connected (a tree)*

A covering graph is a graph that satisfies property 1 but not necessarily property 2 in Definition 8.10.17. It is known that a universal covering $\tilde{G}$ covers all the graph covers of the graph $G$. Let $T_x^r$ denote a tree with root $x$ where every node has depth $r$. Furthermore, define a rooted isomorphism $G_x \cong H_y$ as an isomorphism between graphs $G$ and $H$ that maps $x$ to $y$ and vice versa. We will use the following result to prove a characterization of GWL-1:

**Lemma 8.10.18.** *([301]) Let $T$ and $S$ be trees and $x \in V(T)$ and $y \in V(S)$ be their vertices of the same degree with neighborhoods $N(x) = \{x_1, ..., x_k\}$ and $N(y) = \{y_1, ..., y_k\}$. Let $r \geq 1$. Suppose that $T_x^{r-1} \cong S_y^{r-1}$ and $T_{x_i}^r \cong S_{y_i}^r$ for all $i \leq k$. Then $T_x^{r+1} \cong S_y^{r+1}$.*

A universal cover of a 2-colored bipartite graph is still 2 colored. When we lift nodes $v$ and hyperedge nodes e to their universal cover, we keep their respective red and blue colors.

Define a rooted colored isomorphism $T_{\tilde{e}_1}^k \cong_c T_{\tilde{e}_2}^k$ as a colored tree isomorphism where blue/red node $\tilde{e}_1/\tilde{v}_1$ maps to blue/red node $\tilde{e}_2/\tilde{v}_2$ and vice versa.

In fact, Lemma 8.10.18 holds for 2-colored isomorphisms, which we show below:

**Lemma 8.10.19.** *Let $T$ and $S$ be 2-colored trees and $x \in V(T)$ and $y \in V(S)$ be their vertices of the same degree with neighborhoods $N(x) = \{x_1, ..., x_k\}$ and $N(y) = \{y_1, ..., y_k\}$. Let $r \geq 1$. Suppose that $T_x^{r-1} \cong_c S_y^{r-1}$ and $T_{x_i}^r \cong_c S_{y_i}^r$ for all $i \leq k$. Then $T_x^{r+1} \cong_c S_y^{r+1}$.*

*Proof.* Certainly 2-colored isomorphisms are rooted isomorphisms on 2-colored trees. The converse is true if the roots match in color since recursively all descendants of the root must match in color.

If $T_x^{r-1} \cong_c S_y^{r-1}$ and $T_{x_i}^r \cong_c S_{y_i}^r$ for all $i \leq k$ and $N(x) = \{x_1 ... x_k\}, N(y) = \{y_1 .. y_k\}$, the roots $x$ and $y$ must match in color. The neighborhoods $N(x)$ and $N(y)$ then must both be of the opposing color. Since rooted colored isomorphisms are rooted isomorphisms, we must have $T_x^{r-1} \cong S_y^{r-1}$ and $T_{x_i}^r \cong S_{y_i}^r$ for all $\mathrm{i} \le k$. By Lemma 8.10.18, we have $T_x^{r+1} \cong S_y^{r+1}$. Once the roots match in color, a rooted tree isomorphism is the same as a rooted 2-colored tree isomorphism. Thus, since $x$ and $y$ share the same color, $T_x^{r+1} \cong_c S_y^{r+1}$ □

**Theorem 8.10.20.** *Let $\mathcal{H}_1 = (\mathcal{V}_1, \mathcal{E}_1)$ and $\mathcal{H}_2 = (\mathcal{V}_2, \mathcal{E}_2)$ be two connected hypergraphs. Let $\mathcal{B}_{\mathcal{V}_1,\mathcal{E}_1}$ and $\mathcal{B}_{\mathcal{V}_2,\mathcal{E}_2}$ be two canonically colored bipartite graphs for $\mathcal{H}_1$ and $\mathcal{H}_2$ (vertices colored red and hyperedges colored blue)*

*For any $\mathrm{i} \in \mathbb{Z}^+$, for any of the nodes $x_1 \in \mathcal{B}_{\mathcal{V}_1}, \mathrm{e}_1 \in \mathcal{B}_{\mathcal{V}_1,\mathcal{E}_1}$ and $x_2 \in \mathcal{B}_{\mathcal{V}_1}, \mathrm{e}_2 \in \mathcal{B}_{\mathcal{V}_2,\mathcal{E}_2}$:*

- $(\tilde{\mathcal{B}}^{2\mathrm{i}-1}_{\mathcal{V}_1,\mathcal{E}_1})_{\tilde{\mathrm{e}}_1} \cong_c (\tilde{\mathcal{B}}^{2\mathrm{i}-1}_{\mathcal{V}_2,\mathcal{E}_2})_{\tilde{\mathrm{e}}_2}$ *iff* $f^{\mathrm{i}}_{\mathrm{e}_1} = f^{\mathrm{i}}_{\mathrm{e}_2}$
- $(\tilde{\mathcal{B}}^{2\mathrm{i}}_{\mathcal{V}_1,\mathcal{E}_1})_{\tilde{x}_1} \cong_c (\tilde{\mathcal{B}}^{2\mathrm{i}}_{\mathcal{V}_2,\mathcal{E}_2})_{\tilde{x}_2}$ *iff* $h^{\mathrm{i}}_{x_1} = h^{\mathrm{i}}_{x_2}$,

*with $f^{\mathrm{i}}_{\bullet}, h^{\mathrm{i}}_{\bullet}$ the* i*th GWL-1 values for the hyperedges and nodes respectively where $\mathrm{e}_1 = p_{\mathcal{B}_{\mathcal{V}_1,\mathcal{E}_1}}(\tilde{\mathrm{e}}_1)$, $x_1 = p_{\mathcal{B}_{\mathcal{V}_1,\mathcal{E}_1}}(\tilde{x}_1)$, $\mathrm{e}_2 = p_{\mathcal{B}_{\mathcal{V}_1,\mathcal{E}_1}}(\tilde{\mathrm{e}}_2)$, $x_2 = p_{\mathcal{B}_{\mathcal{V}_1,\mathcal{E}_1}}(\tilde{x}_2)$. The maps $p_{\mathcal{B}_{\mathcal{V}_1,\mathcal{E}_1}} : \tilde{\mathcal{B}}_{\mathcal{V}_1,\mathcal{E}_1} \to \mathcal{B}_{\mathcal{V}_1,\mathcal{E}_1}, p_{\mathcal{B}_{\mathcal{V}_2,\mathcal{E}_2}} : \tilde{\mathcal{B}}_{\mathcal{V}_2,\mathcal{E}_2} \to \mathcal{B}_{\mathcal{V}_2,\mathcal{E}_2}$ are the universal covering maps of $\mathcal{B}_{\mathcal{V}_1,\mathcal{E}_1}$ and $\mathcal{B}_{\mathcal{V}_2,\mathcal{E}_2}$ respectively.*

*Proof.* We prove by induction:

Let $T^k_{\tilde{\mathrm{e}}_1} := (\tilde{\mathcal{B}}^k_{\mathcal{V}_1,\mathcal{E}_1})_{\tilde{\mathrm{e}}_1}$ where $\tilde{\mathrm{e}}_1$ is a pullback of a hyperedge, meaning $p_{\mathcal{B}_{\mathcal{V}_1,\mathcal{E}_2}}(\tilde{\mathrm{e}}_1) = \mathrm{e}_1$. Similarly, let $T^k_{\tilde{\mathrm{e}}_2} := (\tilde{\mathcal{B}}^k_{\mathcal{V}_2,\mathcal{E}_2})_{\tilde{\mathrm{e}}_2}$, $T^k_{\tilde{x}_1} := (\tilde{\mathcal{B}}^k_{\mathcal{V}_1,\mathcal{E}_1})_{\tilde{x}_1}$, $T^k_{\tilde{x}_2} := (\tilde{\mathcal{B}}^k_{\mathcal{V}_2,\mathcal{E}_2})_{\tilde{x}_2}$, $\forall k \in \mathbb{N}$, where $\tilde{\mathrm{e}}_1, \tilde{\mathrm{e}}_2, \tilde{x}_1, \tilde{x}_2$ are the respective pullbacks of $\mathrm{e}_1, \mathrm{e}_2, x_1, x_2$.

Define an (2-colored) isomorphism of multisets of graphs to mean that there exists a bijection between the two multisets so that each graph in one multiset is (2-colored) isomorphic with exactly one other element in the other multiset.

By Observation 8.10.11 we can rewrite GWL-1 as:

$$f^0_{\mathrm{e}} \leftarrow \{\}, h^0_v \leftarrow \{\} \tag{8.39}$$

$$f^{\mathrm{i}+1}_{\mathrm{e}} \leftarrow (f^{\mathrm{i}}_{\mathrm{e}}, \{\!\{h^{\mathrm{i}}_v\}\!\}_{v \in \mathrm{e}}) \forall \mathrm{e} \in \mathcal{E}_{\mathcal{H}} \tag{8.40}$$

$$h^{\mathrm{i}+1}_v \leftarrow (h^{\mathrm{i}}_v, \{\!\{f^{\mathrm{i}+1}_{\mathrm{e}}\}\!\}_{v \in \mathrm{e}}) \forall v \in \mathcal{V}_{\mathcal{H}} \tag{8.41}$$

Base Case i = 1:

$$T^1_{\tilde{e}_1} \cong_c T^1_{\tilde{e}_2} \text{ iff } (T^0_{\tilde{e}_1} \cong_c T^0_{\tilde{e}_2} \text{ and } \{\!\{T^0_{\tilde{x}_1}\}\!\}_{\tilde{x}_1 \in N(\tilde{e}_1)} \cong_c \{\!\{T^0_{\tilde{x}_2}\}\!\}_{\tilde{x}_2 \in N(\tilde{e}_2)}) \text{ (By Lemma 8.10.19)} \tag{8.42a}$$

$$\text{iff } (f^0_{e_1} = f^0_{e_2} \text{ and } \{\!\{h^0_{x_1}\}\!\} = \{\!\{h^0_{x_2}\}\!\}) \text{ (By Equation 8.39)} \tag{8.42b}$$

$$\text{iff } f^1_{e_1} = f^1_{e_2} \text{ (By Equation 8.40)} \tag{8.42c}$$

$$T^2_{\tilde{x}_1} \cong_c T^2_{\tilde{x}_2} \text{ iff } (T^0_{\tilde{x}_1} \cong_c T^0_{\tilde{x}_2} \text{ and } \{\!\{T^1_{\tilde{e}_1}\}\!\}_{\tilde{e}_1 \in N(\tilde{x}_1)} \cong_c \{\!\{T^1_{\tilde{e}_2}\}\!\}_{\tilde{e}_2 \in N(\tilde{x}_2)}) \text{ (By Lemma 8.10.19)} \tag{8.43a}$$

$$\text{iff } (h^0_{e_1} = h^0_{e_2} \text{ and } \{\!\{f^1_{x_1}\}\!\} = \{\!\{f^1_{x_2}\}\!\}) \text{ (By Equation 8.39)} \tag{8.43b}$$

$$\text{iff } f^1_{e_1} = f^1_{e_2} \text{ (By Equation 8.41)} \tag{8.43c}$$

Induction Hypothesis: For i ≥ 1, $T^{2i-1}_{\tilde{e}_1} \cong_c T^{2i-1}_{\tilde{e}_2}$ iff $f^i_{e_1} = f^i_{e_2}$ and $T^{2i}_{\tilde{x}_1} \cong_c T^{2i}_{\tilde{x}_2}$ iff $h^i_{x_1} = h^i_{x_2}$

Induction Step:

$$T^{2i+1}_{\tilde{e}_1} \cong_c T^{2i+1}_{\tilde{e}_2} \text{ iff } (T^{2i-1}_{\tilde{e}_1} \cong_c T^{2i-1}_{\tilde{e}_2} \text{ and } \{\!\{T^{2i}_{\tilde{x}_1}\}\!\}_{\tilde{x}_1 \in N(\tilde{e}_1)} \cong_c \{\!\{T^{2i}_{\tilde{x}_2}\}\!\}_{\tilde{x}_2 \in N(\tilde{e}_2)}) \text{ (By Lemma 8.10.19)} \tag{8.44a}$$

$$\text{iff } (f^i_{e_1} = f^i_{e_2} \text{ and } \{\!\{h^i_{x_1}\}\!\} = \{\!\{h^i_{x_2}\}\!\}) \text{ (By Induction Hypothesis)} \tag{8.44b}$$

$$\text{iff } f^{i+1}_{e_1} = f^{i+1}_{e_2} \text{ (By Equation 8.40)} \tag{8.44c}$$

$$T^{2i}_{\tilde{x}_1} \cong_c T^{2i}_{\tilde{x}_2} \text{ iff } (T^{2i-2}_{\tilde{x}_1} \cong_c T^{2i-2}_{\tilde{x}_2} \text{ and } \{\!\{T^{2i-1}_{\tilde{e}_1}\}\!\}_{\tilde{e}_1 \in N(\tilde{x}_1)} \cong_c \{\!\{T^{2i-1}_{\tilde{e}_2}\}\!\}_{\tilde{e}_2 \in N(\tilde{x}_2)}) \text{ (By Lemma 8.10.19)} \tag{8.45a}$$

$$\text{iff } (h^i_{e_1} = h^i_{e_2} \text{ and } \{\!\{f^i_{x_1}\}\!\} = \{\!\{f^i_{x_2}\}\!\}) \text{ (By Equation 8.39)} \tag{8.45b}$$

$$\text{iff } h^i_{x_1} = h^i_{x_2} \text{ (By Equation 8.41)} \tag{8.45c}$$

□

We write here the theorem characterizing the WL-1 algorithm on a graph by the graph's universal cover.

**Theorem 8.10.21.** *([301]) Let $G$ and $H$ be two connected graphs. Let $p_G : \tilde{G} \to G, p_H : \tilde{H} \to H$ be the universal covering maps of $G$ and $H$ respectively. For any $\mathrm{i} \in \mathbb{N}$, for any two nodes $x \in G$ and $y \in H$: $\tilde{G}^{\mathrm{i}}_{\tilde{x}} \cong \tilde{G}^{\mathrm{i}}_{\tilde{y}}$ iff the WL-1 algorithm assigns the same value to nodes $x = p_G(\tilde{x})$ and $y = p_H(\tilde{y})$.*

It follows immediately from Theorem 8.10.21 that the WL-1 algorithm on colored star expansion bipartite graphs of two hypergraphs corresponds to constructing their universal covers.

**Corollary 8.10.22.** *Let $\mathcal{H}_1 = (\mathcal{V}_1, \mathcal{E}_1)$ and $\mathcal{H}_2 = (\mathcal{V}_2, \mathcal{E}_2)$ be two connected hypergraphs. Let $\mathcal{B}_{\mathcal{V}_1,\mathcal{E}_1}$ and $\mathcal{B}_{\mathcal{V}_2,\mathcal{E}_2}$ be two canonically colored bipartite graphs for $\mathcal{H}_1$ and $\mathcal{H}_2$ (vertices colored red and hyperedges colored blue). Let $p_{\mathcal{B}_{\mathcal{V}_1,\mathcal{E}_1}} : \tilde{\mathcal{B}}_{\mathcal{V}_1,\mathcal{E}_1} \to \mathcal{B}_{\mathcal{V}_1,\mathcal{E}_1}, p_{\mathcal{B}_{\mathcal{V}_2,\mathcal{E}_2}} : \tilde{\mathcal{B}}_{\mathcal{V}_2,\mathcal{E}_2} \to \mathcal{B}_{\mathcal{V}_2,\mathcal{E}_2}$ be the universal coverings of $\mathcal{B}_{\mathcal{V}_1,\mathcal{E}_1}$ and $\mathcal{B}_{\mathcal{V}_2,\mathcal{E}_2}$ respectively. For any $\mathrm{i} \in \mathbb{Z}^+$,*

- $(\tilde{\mathcal{B}}^{2\mathrm{i}-1}_{\mathcal{V}_1,\mathcal{E}_1})_{\tilde{v}_1} \cong_c (\tilde{\mathcal{B}}^{2\mathrm{i}-1}_{\mathcal{V}_2,\mathcal{E}_2})_{\tilde{v}_2}$ *iff* $g^{\mathrm{i}}_{v_1} = g^{\mathrm{i}}_{v_2}$

*with $g^{\mathrm{i}}_{v_1}, g^{\mathrm{i}}_{v_2}$ the* i*-th WL-1 values for $v_1, v_2 \in \mathcal{V}(\mathcal{B}_{\mathcal{V}_1,\mathcal{E}_1}), \mathcal{V}(\mathcal{B}_{\mathcal{V}_2,\mathcal{E}_2})$ respectively where $v_1 = p_{\mathcal{B}_{\mathcal{V}_1,\mathcal{E}_1}}(\tilde{v}_1)$, $v_2 = p_{\mathcal{B}_{\mathcal{V}_2,\mathcal{E}_2}}(\tilde{v}_2)$.*

*Furthermore, for* $\mathrm{i} \geq 2$,

- $g^{2+\mathrm{i}}_{u_1} = g^{2+\mathrm{i}}_{u_2}$ *iff* $h^{1+\mathrm{i}}_{u_1} = h^{1+\mathrm{i}}_{u_2}, \forall u_1 \in L(\mathcal{B}_{\mathcal{V}_1,\mathcal{E}_1}), \forall u_2 \in L(\mathcal{B}_{\mathcal{V}_2,\mathcal{E}_2})$ *and*
- $g^{2+\mathrm{i}}_{\mathrm{e}_1} = g^{2+\mathrm{i}}_{\mathrm{e}_2}$ *iff* $f^{2+\mathrm{i}}_{\mathrm{e}_1} = h^{2+\mathrm{i}}_{\mathrm{e}_2}, \forall \mathrm{e}_1 \in R(\mathcal{B}_{\mathcal{V}_1,\mathcal{E}_1}), \forall \mathrm{e}_2 \in R(\mathcal{B}_{\mathcal{V}_2,\mathcal{E}_2})$

*Proof.* The first equivalence $(\tilde{\mathcal{B}}^{2\mathrm{i}-1}_{\mathcal{V}_1,\mathcal{E}_1})_{\tilde{v}_1} \cong_c (\tilde{\mathcal{B}}^{2\mathrm{i}-1}_{\mathcal{V}_2,\mathcal{E}_2})_{\tilde{v}_2}$ iff $g^{\mathrm{i}}_{v_1} = g^{\mathrm{i}}_{v_2}$ follows directly by Theorem 8.10.22.

The successive two equivalences follow by Theorem 8.10.20 and the first equivalence. □

**Observation 8.10.23.** *If the node values for nodes $x$ and $y$ from GWL-1 for* i *iterations on two hypergraphs $\mathcal{H}_1$ and $\mathcal{H}_2$ are the same, then for all* j *with* $0 \leq \mathrm{j} \leq \mathrm{i}$, *the node values for GWL-1 for* j *iterations on $x$ and $y$ also agree. In particular*

*$textdeg(x) =$*

*$textdeg(y)$.*

*Proof.* There is a 2-color isomorphism on subtrees $(\tilde{\mathcal{B}}^{\mathrm{j}}_{\mathcal{V}_1,\mathcal{E}_1})_{\tilde{x}}$ and $(\tilde{\mathcal{B}}^{\mathrm{j}}_{\mathcal{V}_2,\mathcal{E}_2})_{\tilde{y}}$ of the i-hop subtrees of the universal covers rooted about nodes $x \in \mathcal{V}_1$ and $y \in \mathcal{V}_2$ for $0 \leq \mathrm{j} \leq \mathrm{i}$ since $(\tilde{\mathcal{B}}^{\mathrm{i}}_{\mathcal{V}_1,\mathcal{E}_1})_{\tilde{x}} \cong_c (\tilde{\mathcal{B}}^{\mathrm{i}}_{\mathcal{V}_2,\mathcal{E}_2})_{\tilde{y}}$. By Theorem 8.10.20, we have that GWL-1 returns the same value for $x$ and $y$ for each $0 \leq \mathrm{j} \leq \mathrm{i}$ . □

**Theorem 8.10.24.** *Let $h^L : [\mathcal{V}]^1 \times \mathbb{Z}_2^{n \times 2^n} \to \mathbb{R}^d$ be the L-GWL-1 representation of nodes for hypergraph $\mathcal{H}$ in Equation 8.7, then*

$$Aut(\mathcal{H}) \cong Stab(H) \subseteq Sym(h^L(H)) \cong Aut_c(\tilde{\mathcal{B}}^{2L}_{\mathcal{V},\mathcal{E}}), \forall L \geq 1 \tag{8.46}$$

*Proof.* 1. $Aut(\mathcal{H}) \cong Stab(H)$ follows by Proposition 8.10.5.

2. $Stab(H) \subseteq \mathrm{Sym}(h^L(H))$ follows by definition of the symmetry group of a representation map given in Definition 8.10.9 and the equivariance of $L$-GWL-1 due to Proposition 8.10.12. We know that

$$h^L(S, H) \triangleq AGG(\{h^L_v\}_{v \in S}) \tag{8.47}$$

for $AGG : \mathbf{Set} \to \mathbb{R}^d$ an injective set representation map and that $S$ has cardinality 1. For any $\pi \in Stab(H) \subseteq \mathrm{Sym}(\mathcal{V})$, we must have $\pi \cdot H = H$ we check that $\pi$ satisfies:

$$\pi(u) = v \Rightarrow h(u, H) = h(v, H), \forall u, v \in \mathcal{V} \tag{8.48}$$

If $\pi(u) = v$ and $\pi \cdot H = H$, we can use the equivariance of $h^L$ to get right hand necessary condition: $h^L(u, H) = h^L(u, \pi(H)) = h^L(\pi(u), H) = h^L(v, H)$

3. The last group isomorphism follows by the equivalence between $L$-GWL-1 and the universal cover up to $2L$-hops given in Theorem 8.10.20.

For any $\pi \in \mathrm{Sym}(h^L(H))$, it must satisfy $\pi(u) = v \Rightarrow h(u, H) = h(v, H), \forall u, v \in \mathcal{V}$. We map each $\pi$ to a 2-colored isomorphism $\Phi : \pi \mapsto \phi_c$ which is the 2-colored isomorphism determined by the Theorem 8.10.20:

$$(\tilde{\mathcal{B}}^{2L}_{\mathcal{V},\mathcal{E}})_{\tilde{u}} \cong_c (\tilde{\mathcal{B}}^{2L}_{\mathcal{V},\mathcal{E}})_{\widetilde{\pi(u)})} \text{ iff } h^L_u = h^L_{\pi(u)} = h^L(u, H) = h^L(\pi(u), H), \forall u \in \mathcal{V} \tag{8.49}$$

Certainly the map $\Phi : \pi \mapsto \phi_c$ is a homomorphism because:

1. $\Phi$ maps the identity to identity:

$$(\tilde{\mathcal{B}}^{2L}_{\mathcal{V},\mathcal{E}})_{\tilde{u}} \cong_c (\tilde{\mathcal{B}}^{2L}_{\mathcal{V},\mathcal{E}})_{\tilde{u})} \text{ iff } h^L_u = h^L(u,H), \forall u \in \mathcal{V} \tag{8.50}$$

can have only one 2-colored isomorphism determining $(\tilde{\mathcal{B}}^{2L}_{\mathcal{V},\mathcal{E}})_{\tilde{u}} \cong_c (\tilde{\mathcal{B}}^{2L}_{\mathcal{V},\mathcal{E}})_{\tilde{u}}$, which is the identity.

2. $\Phi$ perserves composition: $\pi_2 \circ \pi_1 \mapsto (\phi_2)_c \circ (\phi_1)_c$

By definition of $\pi_1$ and $\pi_2$:

$$\pi_1(u) = v \Rightarrow h(u,H) = h(v,H) \text{ iff } (\tilde{\mathcal{B}}^{2L}_{\mathcal{V},\mathcal{E}})_{\tilde{u}} \cong_c (\tilde{\mathcal{B}}^{2L}_{\mathcal{V},\mathcal{E}})_{\widetilde{\pi_1(u)}}, \forall u \in \mathcal{V} \tag{8.51a}$$

$$\pi_2(v) = w \Rightarrow h(v,H) = h(w,H) \text{ iff } (\tilde{\mathcal{B}}^{2L}_{\mathcal{V},\mathcal{E}})_{\tilde{v}} \cong_c (\tilde{\mathcal{B}}^{2L}_{\mathcal{V},\mathcal{E}})_{\widetilde{\pi_1(w)}}, \forall v \in \mathcal{V} \tag{8.51b}$$

Combining, we get:

$$(\tilde{\mathcal{B}}^{2L}_{\mathcal{V},\mathcal{E}})_{\tilde{u}} \cong_c (\tilde{\mathcal{B}}^{2L}_{\mathcal{V},\mathcal{E}})_{\widetilde{\pi_1(u)}} \cong_c (\tilde{\mathcal{B}}^{2L}_{\mathcal{V},\mathcal{E}})_{\widetilde{\pi_2(u)}} \cong (\tilde{\mathcal{B}}^{2L}_{\mathcal{V},\mathcal{E}})_{\widetilde{\pi_2 \circ \pi_1(u)}} \tag{8.52}$$

where the first isomorphism is $(\phi_1)_c$ on $(\tilde{\mathcal{B}}^{2L}_{\mathcal{V},\mathcal{E}})_{\tilde{u}}$, the second isomorphism is from $(\phi_2)_c$ on $(\tilde{\mathcal{B}}^{2L}_{\mathcal{V},\mathcal{E}})_{\tilde{u}}$ and the third isomorphism is $(\phi_1)_c \circ (\phi_1)_c$ on $(\tilde{\mathcal{B}}^{2L}_{\mathcal{V},\mathcal{E}})_{\widetilde{\pi_1(u)}}$ □

**Proposition 8.10.25.** *If GWL-1 cannot distinguish two connected hypergraphs $H_1$ and $H_2$ then HyperPageRank will not either.*

*Proof.* HyperPageRank is defined on a hypergraph with star expansion matrix $H$ as the following stationary distribution $\Pi$:

$$\lim_{n \to \infty} (D_v^{-1} \cdot H \cdot D_{\mathrm{e}}^{-1} \cdot H^T)^n = \Pi \tag{8.53}$$

If $H$ is a connected bipartite graph, $\Pi$ must be the eigenvector of $(D_v^{-1} \cdot H \cdot D_{\mathrm{e}}^{-1} \cdot H^T)$ for eigenvalue 1. In other words, $\Pi$ must satisfy

$$(D_v^{-1} \cdot H \cdot D_{\mathrm{e}}^{-1} \cdot H^T) \cdot \Pi = \Pi \tag{8.54}$$

By Theorem 1 of [269], we know that the UniGCN defined by:

$$h_{\mathrm{e}}^{\mathrm{i}+1} \leftarrow \phi_2(h_{\mathrm{e}}^{\mathrm{i}}, h_v^{\mathrm{i}}) = W_{\mathrm{e}} \cdot H^T \cdot h_v^{\mathrm{i}} \tag{8.55a}$$

$$h_v^{\mathrm{i}+1} \leftarrow \phi_1(h_v^{\mathrm{i}}, h_{\mathrm{e}}^{\mathrm{i}+1}) = W_v \cdot H \cdot h_{\mathrm{e}}^{\mathrm{i}+1} \tag{8.55b}$$

for constant $W_{\mathrm{e}}$ and $W_v$ weight matrices, is equivalent to GWL-1 provided that $\phi_1$ and $\phi_2$ are both injective as functions. Without injectivity, we can only guarantee that if UniGCN distinguishes $H_1, H_2$ then GWL-1 distinguishes $H_1, H_2$. In fact, each matrix power of order $n$ in Equation 8.53 corresponds to $h_v^n$ so long as we satisfy the following constraints:

$$W_{\mathrm{e}} \leftarrow D_{\mathrm{e}}^{-1}, W_v \leftarrow D_v^{-1} \text{ and } h_v^0 \leftarrow I \tag{8.56}$$

We show that the matrix powers are UniGCN under the constraints of Equation 8.56 by induction:

Base Case: $n = 0$: $h_v^0 = I$

Induction Hypothesis: $n > 0$:

$$(D_v^{-1} \cdot H \cdot D_{\mathrm{e}}^{-1} \cdot H^T)^n = h_v^n \tag{8.57}$$

Induction Step:

$$(D_v^{-1} \cdot H \cdot h_{\mathrm{e}}^n) \tag{8.58a}$$

$$= (D_v^{-1} \cdot H \cdot ((D_{\mathrm{e}}^{-1} \cdot H^T) \cdot h_v^n)) \tag{8.58b}$$

$$= (D_v^{-1} \cdot H \cdot D_{\mathrm{e}}^{-1} \cdot H^T) \cdot (D_v^{-1} \cdot H \cdot D_{\mathrm{e}}^{-1} \cdot H^T)^n \tag{8.58c}$$

$$= (D_v^{-1} \cdot H \cdot D_{\mathrm{e}}^{-1} \cdot H^T)^{n+1} = h_v^{n+1} \tag{8.58d}$$

Since we cannot guarantee that the maps $\phi_1$ and $\phi_2$ are injective in Equation 8.58b, it must be that the output $h_v^n$, coming from UniGCN with the constraints of Equation 8.56, is at most as powerful as GWL-1.

In general, injectivity preserves more information. For example, if $\phi_1$ is injective and if $\phi'_1$ is an arbitrary map (not guaranteed to be injective) then:

$$\phi_1(h_1) = \phi_1(h_2) \Rightarrow h_1 = h_2 \Rightarrow \phi'_1(h_1) = \phi'_1(h_2) \tag{8.59}$$

HyperpageRank is exactly as powerful as UniGCN under the constraints of Equation 8.56. Thus HyperPageRank is at most as powerful as GWL-1 in distinguishing power. □

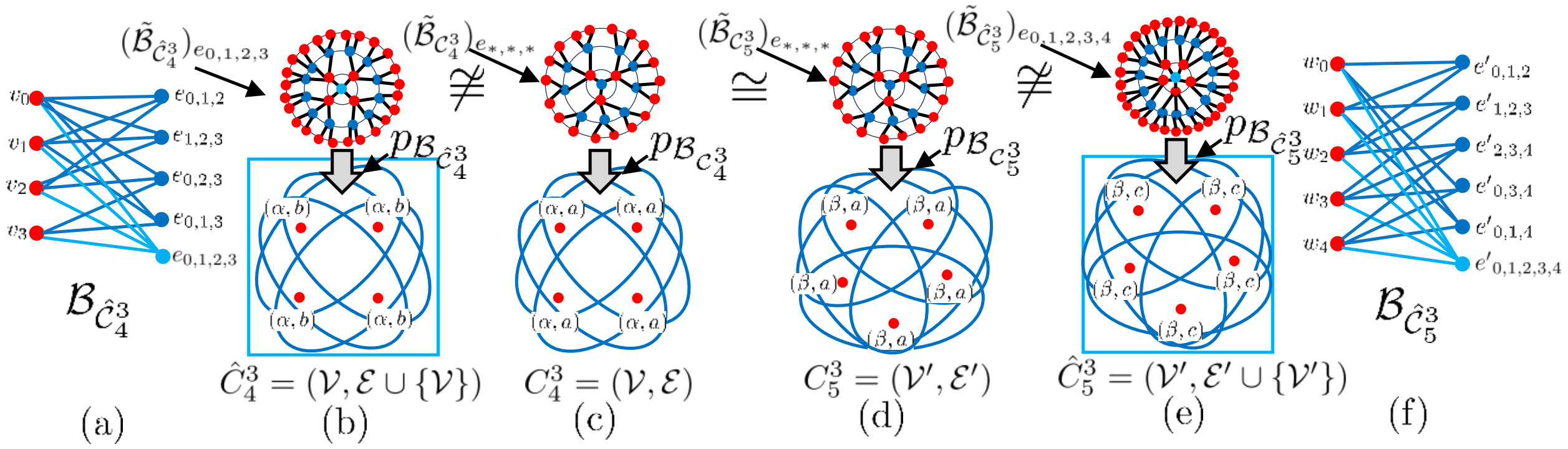

**Figure 8.5.** An illustration of hypergraph symmetry breaking. (c,d) 3-regular hypergraphs $C^3_4$, $C^3_5$ with 4 and 5 nodes respectively and their corresponding universal covers centered at any hyperedge $(\tilde{\mathcal{B}}_{C^3_4})_{e_{*,*,*}}, (\tilde{\mathcal{B}}_{C^3_5})_{e_{*,*,*}}$ with universal covering maps $p_{\mathcal{B}_{C^3_4}}, p_{\mathcal{B}_{C^3_5}}$. (b,e) the hypergraphs $\hat{C}^3_4, \hat{C}^3_5$, which are $C^3_4, C^3_5$ with $4,5$-sized hyperedges attached to them and their corresponding universal covers and universal covering maps. (a,f) are the corresponding bipartite graphs of $\hat{C}^3_4, \hat{C}^3_5$. (c,d) are indistinguishable by GWL-1 and thus will give identical node values by Theorem 8.10.20. On the other hand, (b,e) gives node values which are now sensitive to the the order of the hypergraphs $4, 5$, also by Theorem 8.10.20.

### 8.10.26 Method

We repeat here from the main text the symmetry finding algorithm:

We also repeat here for convenience some definitions used in the proofs. Given a hypergraph $\mathcal{H} = (\mathcal{V}, \mathcal{E})$, let

$$\mathcal{V}_{c_L} := \{v \in \mathcal{V} : c_L = h^L_v(H)\} \tag{8.60}$$

be the set of nodes of the same class $c_L$ as determined by $L$-GWL-1. Let $\mathcal{H}_{c_L}$ be an induced subgraph of $\mathcal{H}$ by $\mathcal{V}_{c_L}$.

**Definition 8.10.27.** *A $L$-*GWL-1 symmetric *induced subhypergraph $\mathcal{R} \subset \mathcal{H}$ of $\mathcal{H}$ is a connected induced subhypergraph determined by $\mathcal{V}_{\mathcal{R}} \subseteq \mathcal{V}_{\mathcal{H}}$, some subset of nodes that are all indistinguishable amongst each other by L-GWL-1:*

$$h^L_u(H) = h^L_v(H), \forall u, v \in \mathcal{V}_{\mathcal{R}} \tag{8.61}$$

**Algorithm 30:** A Symmetry Finding Algorithm

**Data:** Hypergraph $\mathcal{H} = (\mathcal{V}, \mathcal{E})$, represented by its star expansion matrix $H$. $L \in \mathbb{Z}^+$ is the number of iterations to run GWL-1.

**Result:** A pair of collections: $(\mathcal{R}_V = \{\mathcal{V}_{R_\mathrm{j}}\}, \mathcal{R}_E = \cup_\mathrm{j}\{\mathcal{E}_{R_\mathrm{j}}\})$ where $R_\mathrm{j}$ are disconnected subhypergraphs exhibiting symmetry in $\mathcal{H}$ that are indistinguishable by $L$-GWL-1.

$E_{deg} \leftarrow \{\{\deg(v) : v \in \mathrm{e}\} : \forall \mathrm{e} \in \mathcal{E}\}$

$U_L \leftarrow h_v^L(H); \mathcal{G}_L \leftarrow \{U_L[v] : \forall v \in \mathcal{V}\}$ ; /* $U_L[v]$ is the $L$-GWL-1 value of node $v \in \mathcal{V}$. */

$\mathcal{B}_{\mathcal{V}_\mathcal{H}, \mathcal{E}_\mathcal{H}} \leftarrow Bipartite(\mathcal{H})$ /* Construct the bipartite graph from $\mathcal{H}$. */

$\mathcal{R}_V \leftarrow \{\}$; $\mathcal{R}_E \leftarrow \{\}$

**for** $c_L \in \mathcal{G}_L$ **do**

  $\mathcal{V}_{c_L} \leftarrow \{v \in \mathcal{V} : U_L[v] = c_L\}, \mathcal{E}_{c_L} \leftarrow \{\mathrm{e} \in \mathcal{E} : u \in \mathcal{V}_{c_L}, \forall u \in \mathrm{e}\}$

  $\mathcal{C}_{c_L} \leftarrow \text{ConnectedComponents}(\mathcal{H}_{c_L} = (\mathcal{V}_{c_L}, \mathcal{E}_{c_L}))$

  **for** $\mathcal{R}_{c_L,\mathrm{i}} \in \mathcal{C}_{c_L}$ **do**

    $\mathcal{R}_V \leftarrow \mathcal{R}_V \cup \{\mathcal{V}_{\mathcal{R}_{c_L,\mathrm{i}}}\}$; $\mathcal{R}_E \leftarrow \mathcal{R}_E \cup \mathcal{E}_{\mathcal{R}_{c_L,\mathrm{i}}}$

  **end**

**end**

**return** $(\mathcal{R}_V, \mathcal{R}_E)$

*When $L = \infty$, we call such $\mathcal{R}$ a GWL-1 symmetric induced subhypergraph. Furthermore, if $\mathcal{R} = \mathcal{H}$, then we say $\mathcal{H}$ is* GWL-1 symmetric.

**Definition 8.10.28.** *A neighborhood-regular hypergraph is a hypergraph where all neighborhoods of each node are isomorphic to each other.*

**Observation 8.10.29.** *A hypergraph $\mathcal{H}$ is GWL-1 symmetric if and only if it is $L$-GWL-1 symmetric for all $L \geq 1$ if and only if $\mathcal{H}$ is neighborhood regular.*

*Proof.*

**1. First if and only if** :

By Theorem 8.10.20, GWL-1 symmetric hypergraph $\mathcal{H} = (\mathcal{V}, \mathcal{E})$ means that for every pair of nodes $u, v \in \mathcal{V}$, $(\tilde{\mathcal{B}}_{\mathcal{V},\mathcal{E}})_{\tilde{u}} \cong_c (\tilde{\mathcal{B}}_{\mathcal{V},\mathcal{E}})_{\tilde{v}}$. This implies that for any $L \geq 1$, $(\tilde{\mathcal{B}}^{2L}_{\mathcal{V},\mathcal{E}})_{\tilde{u}} \cong_c (\tilde{\mathcal{B}}^{2L}_{\mathcal{V},\mathcal{E}})_{\tilde{v}}$ by restricting the rooted isomorphism to $2L$-hop rooted subtrees, which means that $h_u^L(H) = h_v^L(H)$. The converse is true since $L$ is arbitrary. If there are no cycles, we can just take the isomorphism for the largest. Otherwise, an isomorphism can be constructed for $L = \infty$ by infinite extension.

**2. Second if and only if** :

Let $p_{\mathcal{B}_{\mathcal{V},\mathcal{E}}}$ be the universal covering map for $\mathcal{B}_{\mathcal{V},\mathcal{E}}$. Denote $\tilde{v}, \tilde{u}$ by the lift of some nodes $v, u \in \mathcal{V}$ by $p_{\mathcal{B}_{\mathcal{V},\mathcal{E}}}$.

Let $(\tilde{N}(\tilde{u}))_{\tilde{u}}$ be the rooted bipartite lift of $(N(u))_u$. If $\mathcal{H}$ is $L$-GWL-1 symmetric for all $L \geq 1$ then with $L = 1$, $(\tilde{\mathcal{B}}^2_{\mathcal{V},\mathcal{E}})_{\tilde{u}} \cong_c (\tilde{N}(u))_{\tilde{u}} \cong_c (\tilde{N}(v))_{\tilde{v}} \cong_c (\tilde{\mathcal{B}}^2_{\mathcal{V},\mathcal{E}})_{\tilde{v}}$, iff $(N(u))_u \cong (N(v))_{\tilde{v}}, \forall u, v \in \mathcal{V}$ since $N(u)$ and $N(v)$ are cycle-less for any $u, v \in \mathcal{V}$. For the converse, assume all nodes $v \in \mathcal{V}$ have $(N(v))_v \cong (N^1)_x$ for some 1-hop rooted tree $(N^1)_x$ rooted at node $x$, independent of any $v \in \mathcal{V}$. We prove by induction that for all $L \geq 1$ and for all $v \in \mathcal{V}$, $(\tilde{\mathcal{B}}^{2L}_{\mathcal{V},\mathcal{E}})_{\tilde{v}} \cong_c (\tilde{N}^{2L})_x$ for a $2L$-hop tree $(\tilde{N}^{2L})_x$ rooted at node $x$.

Base case: $L = 1$ is by assumption.

Inductive step: If $(\tilde{\mathcal{B}}^{2L}_{\mathcal{V},\mathcal{E}})_{\tilde{v}} \cong_c (N^{2L})_x$, we can form $(\tilde{\mathcal{B}}^{2L+2}_{\mathcal{V},\mathcal{E}})_{\tilde{v}}$ by attaching $(\tilde{N}(\tilde{u}))_{\tilde{u}}$ to each node $\tilde{u}$ in the $2L$-th layer of $(\tilde{\mathcal{B}}^{2L}_{\mathcal{V},\mathcal{E}})_{\tilde{v}} \cong_c (\tilde{N}^{2L})_x$. Each $(\tilde{N}(u))_{\tilde{u}}$ is independent of the root $\tilde{v}$ since every $u \in \mathcal{V}$ has $(\tilde{N}(\tilde{u}))_{\tilde{u}} \cong_c (\tilde{N}^2)_x$ iff $(N(\tilde{u}))_{\tilde{u}} \cong (N^1)_x$ for an $x$ independent of $u \in \mathcal{V}$. This means $(\tilde{\mathcal{B}}^{2L+2}_{\mathcal{V},\mathcal{E}})_{\tilde{v}} \cong_c (\tilde{N}^{2L+2})_x$ for the same root node $x$ where $(\tilde{N}^{2L+2})_x$ is constructed in the same manner as $(\tilde{\mathcal{B}}^{2L}_{\mathcal{V},\mathcal{E}})_{\tilde{v}}, \forall v \in \mathcal{V}$.

□

**Proposition 8.10.30.** *Let $\mathcal{H} = (\mathcal{V}, \tilde{\mathcal{E}})$ be a multi-hypergraph.*

*A multi-hypergraph isomorphism, like for hypergraphs, is defined by a structure preserving map $(\rho_{\mathcal{V}} : \mathcal{V}_{\mathcal{H}} \to \mathcal{V}_{\mathcal{D}}, \rho_{\tilde{\mathcal{E}}} : \mathcal{E}_{\mathcal{H}} \to \mathcal{E}_{\mathcal{D}})$ but where $\rho_{\tilde{\mathcal{E}}}$ is a bijection between multisets.*

*The star expansion bipartite graph of the multi-hypergraph $\mathcal{H}$: $\mathcal{B}_{\mathcal{V},\tilde{\mathcal{E}}}$, is defined as before as the bipartite graph with vertices $\mathcal{V} \sqcup \tilde{\mathcal{E}}$ and edges $\{(v, \mathrm{e}) \in \mathcal{V} \times \tilde{\mathcal{E}} \mid v \in \mathrm{e}\}$.*

*With these definitions on multi-hypergraphs, Proposition 8.10.16, Theorem 8.10.20, and Corollary 8.10.22 also hold for multi-hypergraphs.*

*Proof.* In the proposition, theorem and corollary, replace the set $\mathcal{E}$ with the multiset $\tilde{\mathcal{E}}$ and the proofs become identical. □

**Algorithm Guarantees** Continuing with the notation, as before, let $\mathcal{H} = (\mathcal{V}, \mathcal{E})$ be a hypergraph with star expansion matrix $H$ and let $(\mathcal{R}_{\mathcal{V}}, \mathcal{R}_{\mathcal{E}})$ be the output of Algorithm 29 on $H$ for $L \in \mathbb{Z}^+$. Denote $\mathcal{C}_{c_L}$ as the set of all connected components of $\mathcal{H}_{c_L}$:

$$\mathcal{C}_{c_L} \triangleq \{C_{c_L} : \text{conn. comp. } C_{c_L} \text{ of } \mathcal{H}_{c_L}\} \tag{8.62}$$

If $L = \infty$, then drop the $L$. Thus, the hypergraphs represented by $(\mathcal{R}_V, \mathcal{R}_E)$ come from $\mathcal{C}_{c_L}$ for each $c_L$. Let:

$$\hat{\mathcal{H}}_L \triangleq (\mathcal{V}, \mathcal{E} \cup \mathcal{R}_V) \tag{8.63}$$

be $\mathcal{H}$ after adding all the hyperedges from $\mathcal{R}_{\mathcal{V}}$ and let $\hat{H}_L$ be the star expansion matrix of the resulting multi-hypergraph $\hat{\mathcal{H}}_L$. Let:

$$\mathcal{G}_L \triangleq \{h_v^L(H) : v \in \mathcal{V}\} \tag{8.64}$$

be the set of all $L$-GWL-1 values on $H$. Let:

$$V_{c_L,s} \triangleq \{v \in \mathcal{V}_{c_L} : v \in R, R \in \mathcal{C}_{c_L}, |\mathcal{V}_R| = s\} \tag{8.65}$$

be the set of all nodes of $L$-GWL-1 class $c_L$ belonging to a connected component in $\mathcal{C}_{c_L}$ of $s \geq 1$ nodes in $\mathcal{H}_{c_L}$, the induced subhypergraph of $L$-GWL-1. Let:

$$\mathcal{S}_{c_L} \triangleq \{|\mathcal{V}_{\mathcal{R}_{c_L,\mathrm{i}}}| : \mathcal{R}_{c_L,\mathrm{i}} \in \mathcal{C}_{c_L}\} \tag{8.66}$$

be the set of node set sizes of the connected components in $\mathcal{H}_{c_L}$.

**Proposition 8.10.31.** *If $L = \infty$, for any GWL-1 node value $c$ for $\mathcal{H}$, the connected component induced subhypergraphs $\mathcal{R}_{c,\mathrm{i}}$, for $\mathrm{i} = 1 \ldots |\mathcal{C}_c|$ are GWL-1 symmetric and neighborhood-regular.*

*Proof.* Let $p_{\mathcal{B}_{\mathcal{V},\mathcal{E}}}$ be the universal covering map for $\mathcal{B}_{\mathcal{V},\mathcal{E}}$. Denote $\tilde{v}, \tilde{u}, \tilde{v}', \tilde{u}'$ by the lift of some nodes $v, u, v', u' \in \mathcal{V}$ by $p_{\mathcal{B}_{\mathcal{V},\mathcal{E}}}$.

Let $L = \infty$ and let $\mathcal{H}_c = (\mathcal{V}_c, \mathcal{E}_c)$. For any i, since $u, v \in \mathcal{V}_c$, $(\tilde{\mathcal{B}}_{\mathcal{V},\mathcal{E}})_u \cong_c (\tilde{\mathcal{B}}_{\mathcal{V},\mathcal{E}})_v$ for all $u, v \in \mathcal{V}_{\mathcal{R}_{c,\mathrm{i}}}$. Since $\mathcal{R}_{c,\mathrm{i}}$ is maximally connected we know that every neighborhood $N_{\mathcal{H}_c}(u)$ for $u \in \mathcal{V}_c$ induced by $\mathcal{H}_c$ has $N_{\mathcal{H}_c}(u) \cong N(u) \cap \mathcal{H}_c$. Since $L = \infty$ we have that $N_{\mathcal{H}_c}(u) \cong N_{\mathcal{H}_c}(v), \forall u, v \in \mathcal{V}_{R_{c,\mathrm{i}}}$ since otherwise WLOG there are $u', v' \in \mathcal{V}_{R_{c,\mathrm{i}}}$ with $N_{\mathcal{H}_c}(u') \ncong N_{\mathcal{H}_c}(v')$ then WLOG there is some hyperedge $\mathrm{e} \in \mathcal{E}_{N_{\mathcal{H}_c}(u')}$ with some $w \in \mathrm{e}$, $w \neq u'$ where e cannot be in isomorphism with any $\mathrm{e}' \in \mathcal{E}_{N_{\mathcal{H}_c}(v')}$. For two hyperedges to be in isomorphism means that their constituent nodes can be bijectively mapped to each other by a restriction of an isomorphism $\phi$ between $N_{\mathcal{H}_c}(u'), N_{\mathcal{H}_c}(v')$ to one of the hyperedges. This means that $(\tilde{\mathcal{B}}_{\mathcal{V}\setminus\{u'\},\mathcal{E}})_w$ is the rooted universal covering subtree centered about $w$ not passing through $u'$ that is connected to $u' \in (\tilde{\mathcal{B}}_{\mathcal{V},\mathcal{E}})_{u'}$ by e. However, $v'$ has no e and thus cannot have a $T_x$ for $x \in \mathcal{V}_{(\tilde{N}(v'))_v}$ satisfying $T_x \cong_c (\tilde{\mathcal{B}}_{\mathcal{V}\setminus\{u'\},\mathcal{E}})_w$ with $x$ connected to $v'$ by a hyperedge $\mathrm{e}'$ isomorphic to e in its neighborhood in $(\tilde{\mathcal{B}}_{\mathcal{V},\mathcal{E}})_{v'}$. This contradicts that $(\tilde{\mathcal{B}}_{\mathcal{V},\mathcal{E}})_{u'} \cong_c (\tilde{\mathcal{B}}_{\mathcal{V},\mathcal{E}})_{v'}$.

We have thus shown that all nodes in $\mathcal{V}_c$ have isomorphic induced neighborhoods. By the Observation 8.10.29, this is equivalent to saying that $\mathcal{R}_{c,\mathrm{i}}$ is GWL-1 symmetric and neighborhood regular. □

We show the partitioning of $\mathcal{H}$ by Algorithm 12:

**Proposition 8.10.32.** *If $L \geq 1$, the output $(\mathcal{R}_\mathcal{V}, \mathcal{R}_\mathcal{E})$ of Algorithm 29 partitions a subgraph of $\mathcal{H}$, meaning:*

$$\mathcal{V} = \sqcup_{V \in \mathcal{R}_\mathcal{V}} V \text{ and } \mathcal{E} \supset \sqcup_{E \in \mathcal{R}_\mathcal{E}} E \tag{8.67}$$

*Proof.* **1.** $\mathcal{V} = \sqcup_{V \in \mathcal{R}_\mathcal{V}} V$**:**

For a given subset of nodes $\mathcal{U} \subseteq \mathcal{V}$, let $\text{ConnectedComponents}_\mathcal{H}(\mathcal{U})$ be the collection of node subsets of $\mathcal{U}$ where each node subset forms a connected component in $\mathcal{H}$.

Let $h_v^L(H)$ denote the $L$-GWL-1 node value of $v \in \mathcal{V}$.

Let $\mathcal{R}_\mathcal{V}(L) \triangleq \bigcup_{v \in \mathcal{V}} \text{ConnectedComponents}_\mathcal{H}(\{u \in \mathcal{V} : h_u^L(H) = h_v^L(H)\})$ denote the collection of node sets of common $L$-GWL-1 values for a given $L$.

By the definition of $\mathcal{R}_\mathcal{V}$, since every connected component is size atleast 1 and every node is considered, we must have $\bigcup_{v \in \mathcal{V}} \mathcal{R}_\mathcal{V} = \mathcal{V}$. Since each connected component of a given GWL-1 value is maximal, meaning there is no superset of nodes that is connected, no two connected components can intersect through either nodes or hyperedges. Furthermore, a single node can belong to only one GWL-1 value, thus the values form a partition of $\mathcal{V}$. This proves $\mathcal{V} = \sqcup_{V \in \mathcal{R}_\mathcal{V}} V$.

**2.** $\mathcal{E} \supset \sqcup_{E \in \mathcal{R}_\mathcal{E}} E$**:**

This follows since $\mathcal{R}_\mathcal{E}$ are the hyperedges of each connected component spanned by $\mathcal{R}_\mathcal{V}$. These connected components form disconnected subhypergraphs of $\mathcal{H}$. □

**Prediction Guarantees:**

In order to guarantee that the GWL-1 symmetric components $\mathcal{R}_{c,\mathrm{i}}$ found by Algorithm 12 carry additional information, there needs to be a separation between them to prevent an intersection between the rooted trees computed by GWL-1. We redefine from the main paper what it means for two node subsets to be sufficiently separated via the shortest hyperedge path distance between nodes in $\mathcal{V}$ as follows:

**Definition 8.10.33.** *Two subsets of nodes $\mathcal{U}_1, \mathcal{U}_2 \subseteq \mathcal{V}$ are **sufficiently $L$-separated** if:*

$$\min_{v_1 \in \mathcal{U}_1, v_2 \in \mathcal{U}_2} d(v_1, v_2) > L \tag{8.68}$$

*where $d(v_1, v_2) \triangleq \min_{\mathrm{e}_1 \ldots \mathrm{e}_k \in \mathcal{E}, v_1 \in \mathrm{e}_1, v_2 \in \mathrm{e}_k} k$ is the shortest hyperedge path distance from $v_1 \in \mathcal{V}$ to $v_2 \in \mathcal{V}$.*

*A collection of node subsets $\mathcal{C} \subseteq 2^\mathcal{V}$ is **sufficiently $L$-separated** if all pairs of node subsets are **sufficiently $L$-separated**.*

Our definition of sufficiently $L$-separated is similar in nature to that of well separation between point sets [307] in Euclidean space.

We give another definition that will be useful for the proof of the following lemma:

**Definition 8.10.34.** *A star graph $N_x$ is defined as a tree rooted at $x$ of depth 1. The root $x$ is the only node that can have degree more than 1.*

Assuming that the $\mathcal{C}_{c_L}$ are sufficiently $L$-separated from each other, intuitively meaning that no two nodes from two separate $\mathcal{V}_{\mathcal{R}_{c_L,\mathrm{i}}} \in \mathcal{R}_V$ are within $L$ hyperedges away, then the cardinality of each component $|\mathcal{V}_{\mathcal{R}_{c_L,\mathrm{i}}}|$ is recognizable.

**Lemma 8.10.35.** *If $L \in \mathbb{Z}^+$ is small enough so that after running Algorithm 29 on $L$, for any $L$-GWL-1 node class $c_L$ on $\mathcal{V}$ the collection of $\mathcal{C}_{c_L}$ is **sufficiently** $L$**-separated**,*

*then after forming $\hat{\mathcal{H}}_L$, the new $L$-GWL-1 node classes of $\mathcal{V}_{\mathcal{R}_{c_L,\mathrm{i}}}$ for $\mathrm{i} = 1...\mathcal{C}_{c_L}$ in $\hat{\mathcal{H}}_L$ are all the same class $c'_L$ but are distinguishable from $c_L$ depending on $|\mathcal{V}_{\mathcal{R}_{c_L,\mathrm{i}}}|$.*

*Proof.* After running Algorithm 29 on $\mathcal{H} = (\mathcal{V}, \mathcal{E})$, let $\hat{\mathcal{H}}_L = (\hat{\mathcal{V}}_L, \hat{\mathcal{E}}_L \triangleq \mathcal{E} \cup \bigsqcup_{c_L,\mathrm{i}}\{\mathcal{V}_{\mathcal{R}_{c_L,\mathrm{i}}}\})$ be the hypergraph formed by attaching a hyperedge to each $\mathcal{V}_{\mathcal{R}_{c_L,\mathrm{i}}}$.

For any $c_L$, a $L$-GWL-1 node class, let $\mathcal{R}_{c_L,\mathrm{i}}, \mathrm{i} = 1...|\mathcal{C}_{c_L}|$ be a connected component subhypergraph of $\mathcal{H}_{c_L}$. Over all $(c_L, \mathrm{i})$ pairs, all the $\mathcal{R}_{c_L,\mathrm{i}}$ are disconnected from each other and for each $c_L$ each $\mathcal{R}_{c_L,\mathrm{i}}$ is maximally connected on $\mathcal{H}_{c_L}$.

Upon covering all the nodes $\mathcal{V}_{\mathcal{R}_{c_L,\mathrm{i}}}$ of each induced connected component subhypergraph $\mathcal{R}_{c_L,\mathrm{i}}$ with a single hyperedge $\mathrm{e} = \mathcal{V}_{\mathcal{R}_{c_L,\mathrm{i}}}$ of size $s = |\mathcal{V}_{\mathcal{R}_{c_L,\mathrm{i}}}|$, we claim that every node of class $c_L$ becomes $c_{L,s}$, a $L$-GWL-1 node class depending on the original $L$-GWL-1 node class $c_L$ and the size of the hyperedge $s$.

Consider for each $v \in \mathcal{V}_{\mathcal{R}_{c_L,\mathrm{i}}}$ the $2L$-hop rooted tree $(\tilde{\mathcal{B}}^{2L}_{\mathcal{V},\mathcal{E}})_{\tilde{v}}$ for $p_{\mathcal{B}_{\mathcal{V},\mathcal{E}}}(\tilde{v}) = v$. Also, for each $v \in \mathcal{V}_{\mathcal{R}_{c_L,\mathrm{i}}}$, define the tree

$$T_{\mathrm{e}} \triangleq (\tilde{\mathcal{B}}^{2L-1}_{\hat{\mathcal{V}}\setminus\{v\},\hat{\mathcal{E}}})_{\tilde{\mathrm{e}}} \tag{8.69}$$

We do not index the tree $T_{\mathrm{e}}$ by $v$ since it does not depend on $v \in \mathcal{V}_{\mathcal{R}_{c_L,\mathrm{i}}}$. We prove this in the following.

**proof for: $T_{\mathrm{e}}$ does not depend on $v \in \mathcal{V}_{\mathcal{R}_{c_L,\mathrm{i}}}$:**

Let node $\tilde{\mathrm{e}}$ be the lift of $\mathrm{e}$ to $(\tilde{\mathcal{B}}^{2L-1}_{\hat{\mathcal{V}},\hat{\mathcal{E}}})_{\tilde{\mathrm{e}}}$. Define the star graph $(N(\tilde{\mathrm{e}}))_{\tilde{\mathrm{e}}}$ as the 1-hop neighborhood of $\tilde{\mathrm{e}}$ in $(\tilde{\mathcal{B}}^{2L-1}_{\hat{\mathcal{V}},\hat{\mathcal{E}}})_{\tilde{\mathrm{e}}}$. We must have:

$$(\tilde{\mathcal{B}}^{2L-1}_{\hat{\mathcal{V}},\hat{\mathcal{E}}})_{\tilde{\mathrm{e}}} \cong_c ((N(\tilde{\mathrm{e}}))_{\tilde{\mathrm{e}}} \sqcup \bigsqcup_{\tilde{u}\in\mathcal{V}_{N(\tilde{\mathrm{e}})}\setminus\{\tilde{\mathrm{e}}\}} (\tilde{\mathcal{B}}^{2L-2}_{\mathcal{V},\mathcal{E}})_{\tilde{u}})_{\tilde{\mathrm{e}}} \tag{8.70}$$

Define for each node $v \in \mathrm{e}$ with lift $\tilde{v}$:

$$(N(\tilde{\mathrm{e}}, \tilde{v}))_{\tilde{\mathrm{e}}} \triangleq (\mathcal{V}_{(N(\tilde{\mathrm{e}}))_{\tilde{\mathrm{e}}}} \setminus \{\tilde{v}\}, \mathcal{E}_{(N(\tilde{\mathrm{e}}))_{\tilde{\mathrm{e}}}} \setminus \{(\tilde{\mathrm{e}}, \tilde{v})\})_{\tilde{\mathrm{e}}} \tag{8.71}$$

The tree $(N(\tilde{\mathrm{e}},\tilde{v}))_{\tilde{\mathrm{e}}}$ is a star graph with the node $\tilde{v}$ deleted from $(N(\tilde{\mathrm{e}}))_{\tilde{\mathrm{e}}}$. The star graphs $(N(\tilde{\mathrm{e}},\tilde{v}))_{\tilde{\mathrm{e}}} \subseteq (N(\tilde{\mathrm{e}}))_{\tilde{\mathrm{e}}}$ do not depend on $\tilde{v}$ as long as $\tilde{v} \in \mathcal{V}_{(N(\tilde{\mathrm{e}}))_{\tilde{\mathrm{e}}}}$. In other words,

$$(N(\tilde{\mathrm{e}},\tilde{v}))_{\tilde{\mathrm{e}}} \cong_c (N(\tilde{\mathrm{e}},\tilde{v}'))_{\tilde{\mathrm{e}}}, \forall \tilde{v},\tilde{v}' \in \mathcal{V}_{(N(\tilde{\mathrm{e}}))_{\tilde{\mathrm{e}}}} \setminus \{\tilde{\mathrm{e}}\} \tag{8.72}$$

Since the rooted tree $(\tilde{\mathcal{B}}^{2L-1}_{\hat{\mathcal{V}},\hat{\mathcal{E}}})_{\tilde{\mathrm{e}}}$, where $\tilde{\mathrm{e}}$ is the lift of e by universal covering map $p_{\mathcal{B}_{\mathcal{V},\mathcal{E}}}$, has all pairs of nodes $\tilde{u},\tilde{u}' \in \tilde{\mathrm{e}}$ in it with $(\tilde{\mathcal{B}}^{2L}_{\mathcal{V},\mathcal{E}})_{\tilde{u}} \cong_c (\tilde{\mathcal{B}}^{2L}_{\mathcal{V},\mathcal{E}})_{\tilde{u}'}$, which implies

$$(\tilde{\mathcal{B}}^{2L-2}_{\mathcal{V},\mathcal{E}})_{\tilde{u}} \cong_c (\tilde{\mathcal{B}}^{2L-2}_{\mathcal{V},\mathcal{E}})_{\tilde{u}'}, \forall \tilde{u},\tilde{u}' \in \tilde{\mathrm{e}} \tag{8.73}$$

By Equations 8.73, 8.72, we thus have:

$$(\tilde{\mathcal{B}}^{2L-1}_{\hat{\mathcal{V}}\setminus\{v\},\hat{\mathcal{E}}})_{\tilde{\mathrm{e}}} \cong_c ((N(\tilde{\mathrm{e}},\tilde{v}))_{\tilde{\mathrm{e}}} \sqcup \bigsqcup_{\tilde{u}\in\mathcal{V}_{(N(\tilde{\mathrm{e}},\tilde{v}))_{\tilde{\mathrm{e}}}}\setminus\{\tilde{\mathrm{e}}\}} (\tilde{\mathcal{B}}^{2L-2}_{\mathcal{V},\mathcal{E}})_{\tilde{u}})_{\tilde{\mathrm{e}}} \tag{8.74}$$

This proves that $T_{\mathrm{e}}$ does not need to be indexed by $v \in \mathcal{V}_{\mathcal{R}_{c_L},\mathrm{i}}$.

We continue with the proof that all nodes in $\mathcal{V}_{\mathcal{R}_{c_L},\mathrm{i}}$ become the $L$-GWL-1 node class $c_{L,s}$ for $s = |\mathcal{V}_{\mathcal{R}_{c_L},\mathrm{i}}|$.

Since every $v \in \mathcal{V}_{\mathcal{R}_{c_L},\mathrm{i}}$ becomes connected to a hyperedge $\mathrm{e} = \mathcal{V}_{\mathcal{R}_{c_L},\mathrm{i}}$ in $\hat{\mathcal{H}}$, we must have:

$$(\tilde{\mathcal{B}}^{2L}_{\hat{\mathcal{V}},\hat{\mathcal{E}}})_{\tilde{v}} \cong_c ((\tilde{\mathcal{B}}^{2L}_{\mathcal{V},\mathcal{E}})_{\tilde{v}} \cup_{(\tilde{v},\tilde{\mathrm{e}})} T_{\mathrm{e}})_{\tilde{v}}, \forall v \in \mathcal{V}_{\mathcal{R}_{c_L},\mathrm{i}} \tag{8.75}$$

The notation $((\tilde{\mathcal{B}}^{2L}_{\mathcal{V},\mathcal{E}})_{\tilde{v}} \cup_{(\tilde{v},\tilde{\mathrm{e}})} T_{\mathrm{e}})_{\tilde{v}}$ denotes a tree rooted at $\tilde{v}$ that is the attachment of the tree $T_{\mathrm{e}}$ rooted at $\tilde{\mathrm{e}}$ to the node $\tilde{v}$ by the edge $(\tilde{v},\tilde{\mathrm{e}})$. As is usual, we assume $\tilde{v},\tilde{\mathrm{e}}$ are the lifts of $v \in \mathcal{V}, \mathrm{e} \in \mathcal{E}$ respectively. We only need to consider the single e since $L$ was chosen small enough so that the $2L$-hop tree $(\tilde{\mathcal{B}}^{2L}_{\hat{\mathcal{V}},\hat{\mathcal{E}}})_{\tilde{v}}$ does not contain a node $\tilde{u}$ satisfying $p_{\mathcal{B}_{\mathcal{V},\mathcal{E}}}(\tilde{u}) = u$ with $u \in \mathcal{V}_{\mathcal{R}_{c_L},\mathrm{j}}$ for all $\mathrm{j} = 1...|\mathcal{C}_{c_L}|, \mathrm{j} \neq \mathrm{i}$.

Since $T_{\mathrm{e}}$ does not depend on $v \in \mathcal{V}_{\mathcal{R}_{c_L},\mathrm{i}}$,

$$(\tilde{\mathcal{B}}^{2L}_{\hat{\mathcal{V}},\hat{\mathcal{E}}})_{\tilde{u}} \cong_c (\tilde{\mathcal{B}}^{2L}_{\hat{\mathcal{V}},\hat{\mathcal{E}}})_{\tilde{v}}, \forall u, v \in \mathcal{V}_{\mathcal{R}_{c_L},\mathrm{i}} \tag{8.76}$$

This shows that $h_u^L(\hat{H}) = h_v^L(\hat{H}), \forall u, v \in \mathcal{V}_{\mathcal{R}_{c_L,\mathrm{i}}}$ by Theorem 8.10.20. Furthermore, since each $v \in \mathcal{V}_{\mathcal{R}_{c_L,\mathrm{i}}} \subseteq \hat{\mathcal{V}}$ in $\hat{\mathcal{H}}$ is now incident to a new hyperedge $\mathrm{e} = \mathcal{V}_{\mathcal{R}_{c_L,\mathrm{i}}}$, we must have that the $L$-GWL-1 class $c_L$ of $\mathcal{V}_{\mathcal{R}_{c_L,\mathrm{i}}}$ on $\mathcal{H}$ is now distinguishable by $|\mathcal{V}_{\mathcal{R}_{c_L,\mathrm{i}}}|$.

□

We will need the following definition to prove the next lemma.

**Definition 8.10.36.** *A partial universal cover of hypergraph $\mathcal{H} = (\mathcal{V}, \mathcal{E})$ with an unexpanded induced subhypergraph $\mathcal{R}$, denoted $U(\mathcal{H}, \mathcal{R})_{\mathcal{V},\mathcal{E}}$ is a graph cover of $\mathcal{B}_{\mathcal{V},\mathcal{E}}$ where we freeze $\mathcal{B}_{\mathcal{V}_\mathcal{R},\mathcal{E}_\mathcal{R}} \subseteq \tilde{\mathcal{B}}_{\mathcal{V},\mathcal{E}}$ as an induced subgraph.*

*A $l$-hop rooted partial universal cover of hypergraph $\mathcal{H} = (\mathcal{V}, \mathcal{E})$ with an unexpanded induced subhypergraph $\mathcal{R}$, denoted $(U^l(\mathcal{H}, \mathcal{R})_{\mathcal{V},\mathcal{E}})_{\tilde{u}}$ for $u \in \mathcal{V}$ or $(U^l(\mathcal{H}, \mathcal{R})_{\mathcal{V},\mathcal{E}})_{\tilde{\mathrm{e}}}$ for $\mathrm{e} \in \mathcal{E}$, where $\tilde{v}, \tilde{\mathrm{e}}$ are lifts of $v, \mathrm{e}$, is a rooted graph cover of $\mathcal{B}_{\mathcal{V},\mathcal{E}}$ where we freeze $\mathcal{B}_{\mathcal{V}_\mathcal{R},\mathcal{E}_\mathcal{R}} \subseteq \tilde{\mathcal{B}}_{\mathcal{V},\mathcal{E}}$ as an induced subgraph.*

**Lemma 8.10.37.** *Assuming the same conditions as Lemma 8.10.35, where $\mathcal{H} = (\mathcal{V}, \mathcal{E})$ is a hypergraph and for all $L$-GWL-1 node classes $c_L$ with connected components $\mathcal{R}_{c_L,\mathrm{i}}$, as discovered by Algorithm 29, so that $L \geq diam(\mathcal{R}_{c_L,\mathrm{i}})$. Instead of only adding the hyperedges $\{\mathcal{V}_{\mathcal{R}_{c_L,\mathrm{i}}}\}_{c_L,\mathrm{i}}$ to $\mathcal{E}$ as stated in the main paper, let $\hat{\mathcal{H}}_\dagger \triangleq (\mathcal{V}, (\mathcal{E} \setminus \mathcal{R}_E) \sqcup \mathcal{R}_V)$, meaning $\mathcal{H}$ with each $\mathcal{R}_{c_L,\mathrm{i}}$ for $\mathrm{i} = 1...|\mathcal{C}_{c_L}|$ having all of its hyperedges dropped and with a single hyperedge that covers $\mathcal{V}_{\mathcal{R}_{c_L,\mathrm{i}}}$ and let $\hat{\mathcal{H}} = (\mathcal{V}, \mathcal{E} \sqcup \mathcal{R}_V)$ then:*

*The GWL-1 node classes of $\mathcal{V}_{\mathcal{R}_{c_L,\mathrm{i}}}$ for $\mathrm{i} = 1...|\mathcal{C}_{c_L}|$ in $\hat{\mathcal{H}}$ are all the same class $c'_L$ but are distinguishable from $c_L$ depending on $|\mathcal{V}_{\mathcal{R}_{c_L,\mathrm{i}}}|$.*

*Proof.* For any $c_L$, a $L$-GWL-1 node class, let $\mathcal{R}_{c_L,\mathrm{i}}, \mathrm{i} = 1...|\mathcal{C}_{c_L}|$ be a connected component subhypergraph of $\mathcal{H}_{c_L}$. These connected components are discovered by the algorithm. Over all $(c_L, \mathrm{i})$ pairs, all the $\mathcal{R}_{c_L,\mathrm{i}}$ are disconnected from each other. Upon arbitrarily deleting all hyperedges in each such induced connected component subhypergraph $\mathcal{R}_{c_L,\mathrm{i}}$ and adding a single hyperedge of size $s = |\mathcal{V}_{\mathcal{R}_{c_L,\mathrm{i}}}|$, we claim that every node of class $c_L$ becomes $c_{L,s}$, a $L$-GWL-1 node class depending on the original $L$-GWL-1 node class $c_L$ and the size of the hyperedge $s$.

Define the subhypergraph made up of the disconnected components $\mathcal{R}_{c_L,\mathrm{i}}$ as:

$$\mathcal{R} := \bigcup_{c,\mathrm{i}} \mathcal{R}_{c_L,\mathrm{i}} \tag{8.77}$$

Since $L \geq diam(\mathcal{R}_{c_L,\mathrm{i}})$, we can construct the $2L$-hop rooted partial universal cover with unexpanded induced subhypergraph $\mathcal{R}$, denoted by $(U^{2L}(\mathcal{H},\mathcal{R})_{\mathcal{V},\mathcal{E}})_{\tilde{v}}, \forall v \in \mathcal{V}$ of $\mathcal{H}$ as given in Definition 8.10.36.

Denote the hyperedge nodes, or right hand nodes of the bipartite graph by $\mathcal{B}(\mathcal{V}_{\mathcal{R}},\mathcal{E}_{\mathcal{R}})$ by $R(\mathcal{B}(\mathcal{V}_{\mathcal{R}},\mathcal{E}_{\mathcal{R}}))$. Their corresponding hyperedges are $\mathcal{E}_{\mathcal{R}} \subseteq \mathcal{E}(U(\mathcal{H},\mathcal{R})) \subseteq \mathcal{E}$. Since each $\mathcal{R}_{c_L,\mathrm{i}}$ is maximally connected, for any nodes $u, v \in \mathcal{V}_{\mathcal{R}}$ we have:

$$(U^{2L}(\mathcal{H},\mathcal{R})_{\tilde{u}} \setminus R(\mathcal{B}(\mathcal{V}_{\mathcal{R}},\mathcal{E}_{\mathcal{R}})))_{\tilde{u}} \cong_c (U^{2L}(\mathcal{H},\mathcal{R})_{\tilde{v}} \setminus R(\mathcal{B}(\mathcal{V}_{\mathcal{R}},\mathcal{E}_{\mathcal{R}})))_{\tilde{v}} \tag{8.78}$$

by Proposition 8.10.31, where $U^{2L}(\mathcal{H},\mathcal{R})_{\tilde{v}} \setminus R(\mathcal{B}(\mathcal{V}_{\mathcal{R}},\mathcal{E}_{\mathcal{R}}))$ denotes removing the nodes $R(\mathcal{B}(\mathcal{V}_{\mathcal{R}},\mathcal{E}_{\mathcal{R}}))$ from $U^{2L}(\mathcal{H},\mathcal{R})_{\tilde{v}}$. This follows since removing $R(\mathcal{B}(\mathcal{V}_{\mathcal{R}},\mathcal{E}_{\mathcal{R}}))$ removes an isomorphic neighborhood of hyperedges from each node in $\mathcal{V}_{\mathcal{R}}$. This requires assuming maximal connectedness of each $\mathcal{R}_{c_L,\mathrm{i}}$. Upon adding the hyperedge

$$\mathrm{e}_{c_L,\mathrm{i}} \triangleq \mathcal{V}_{\mathcal{R}_{c_L,\mathrm{i}}} \tag{8.79}$$

covering all of $\mathcal{V}_{\mathcal{R}_{c_L,\mathrm{i}}}$ after the deletion of $\mathcal{E}_{\mathcal{R}_{c_L,\mathrm{i}}}$ for every $(c_L, \mathrm{i})$ pair, we see that any node $u \in \mathcal{V}_{\mathcal{R}_{c_L,\mathrm{i}}}$ is connected to any other node $v \in \mathcal{V}_{\mathcal{R}_{c_L,\mathrm{i}}}$ through $\mathrm{e}_{c_L,\mathrm{i}}$ in the same way for all nodes $u, v \in \mathcal{V}_{\mathcal{R}_{c_L,\mathrm{i}}}$. In fact, we claim that all the nodes in $\mathcal{V}_{\mathcal{R}_{c_L,\mathrm{i}}}$ still have the same GWL-1 class.

We can write the multi-hypergraph $\hat{\mathcal{H}}_\dagger$ equivalently as $(\mathcal{V}, \bigsqcup_{c_L,\mathrm{i}}(\mathcal{E} \setminus \mathcal{E}(\mathcal{R}_{c_L,\mathrm{i}}) \sqcup \{\!\{\mathrm{e}_{c_L,\mathrm{i}}\}\!\}))$, which is the multi-hypergraph formed by the algorithm. The replacement operation on $\mathcal{H}$ can be viewed in the universal covering space $\tilde{\mathcal{B}}_{\mathcal{V},\mathcal{E}}$ as taking $U(\mathcal{H},\mathcal{R})$ and replacing the frozen subgraph $\mathcal{B}_{\mathcal{V}_{\mathcal{R}},\mathcal{E}_{\mathcal{R}}}$ with the star graphs $(N_{\hat{\mathcal{H}}_\dagger}(\tilde{\mathrm{e}}_{c_L,\mathrm{i}}))_{\tilde{\mathrm{e}}_{c_L,\mathrm{i}}}$ of root node $\tilde{\mathrm{e}}_{c_L,\mathrm{i}}$ determined by hyperedge $e_{c_L,i}$ for each connected component indexed by $(c_L, i)$. Since the star graphs $(N_{\hat{\mathcal{H}}_\dagger}(\tilde{e}_{c_L,i}))_{\tilde{e}_{c_L,i}}$ are cycle-less, we have that:

$$(U(\mathcal{H},\mathcal{R}) \setminus R(\mathcal{B}(\mathcal{V}_\mathcal{R},\mathcal{E}_\mathcal{R}))) \cup \bigcup_{c_L,i}(N_{\hat{\mathcal{H}}_\dagger}(\tilde{e}_{c_L,i}))_{\tilde{e}_{c_L,i}} \cong_c \tilde{\mathcal{B}}_{\mathcal{V}_{\hat{\mathcal{H}}_\dagger},\mathcal{E}_{\hat{\mathcal{H}}_\dagger}} \tag{8.80}$$

Viewing Equation 8.80 locally, by our assumptions on $L$, for any $v \in \mathcal{V}_{\mathcal{R}_{c_L,i}}$, we must also have:

$$(U^{2L}(\mathcal{H},\mathcal{R})_{\tilde{v}} \setminus R(\mathcal{B}(\mathcal{V}_\mathcal{R},\mathcal{E}_\mathcal{R}))) \bigcup(N_{\hat{\mathcal{H}}_\dagger}(\tilde{e}_{c,i}))_{\tilde{e}_{c,i}} \cong_c \tilde{\mathcal{B}}_{\mathcal{V}_{\hat{\mathcal{H}}_\dagger},\mathcal{E}_{\hat{\mathcal{H}}_\dagger}} \tag{8.81}$$

We thus have $(\tilde{\mathcal{B}}^{2L}_{\mathcal{V}_{\hat{\mathcal{H}}_\dagger},\mathcal{E}_{\hat{\mathcal{H}}_\dagger}})_{\tilde{u}} \cong_c (\tilde{\mathcal{B}}^{2L}_{\mathcal{V}_{\hat{\mathcal{H}}_\dagger},\mathcal{E}_{\hat{\mathcal{H}}_\dagger}})_{\tilde{v}}$ for every $u, v \in \mathcal{V}_{\mathcal{R}_{c_L,i}}$ with $\tilde{u}, \tilde{v}$ being the lifts of $u, v$ by $p_{\mathcal{B}_{\mathcal{V},\mathcal{E}}}$, since $(U^{2L}(\mathcal{H},\mathcal{R})_{\tilde{u}} \setminus R(\mathcal{B}(\mathcal{V}_\mathcal{R},\mathcal{E}_\mathcal{R})))_{\tilde{u}} \cong_c (U^{2L}(\mathcal{H},\mathcal{R})_{\tilde{v}} \setminus R(\mathcal{B}(\mathcal{V}_\mathcal{R},\mathcal{E}_\mathcal{R})))_{\tilde{v}}$ for every $u, v \in \mathcal{V}_{\mathcal{R}_{c_L,i}}$ as in Equation 8.78. These rooted universal covers now depend on a new hyperedge $e_{c_L,i}$ and thus depend on its size $s$.

This proves the claim that all the nodes in $\mathcal{V}_{\mathcal{R}_{c_L,i}}$ retain the same $L$-GWL-1 node class by changing $\mathcal{H}$ to $\hat{\mathcal{H}}_\dagger$ and that this new class is distinguishable by $s = |\mathcal{V}_{\mathcal{R}_{c_L,i}}|$. In otherwords, the new class can be determined by $c_s$. Furthermore, $c_{L,s}$ on the hyperedge $e_{c_L,i}$ cannot become the same class as an existing class due to the algorithm. □

**Theorem 8.10.38.** *Let $|\mathcal{V}| = n, L \in \mathbb{Z}^+$ and $vol(v) \triangleq \sum_{e \in \mathcal{E}: e \ni v}|e|$ and assuming that the collection of node subsets $\mathcal{C}_{c_L}$ is sufficiently $L$-separated.*

*If $vol(v) = O(\log^{\frac{1-\epsilon}{4L}} n), \forall v \in \mathcal{V}$ for any constant $\epsilon > 0$; $|\mathcal{S}_{c_L}| \leq S, \forall c_L \in \mathcal{C}_L$, $S$ constant, and $|V_{c_L,s}| = O(\frac{n^\epsilon}{\log^{\frac{1}{2k}}(n)}), \forall s \in \mathcal{C}_{c_L}$ , then for $k \in \mathbb{Z}^+$ and $k$-tuple $C = (c_{L,1}...c_{L,k}), c_{L,i} \in \mathcal{G}_L, i = 1..k$ there exists $\omega(n^{2k\epsilon})$ many pairs of $k$-node sets $S_1 \not\simeq S_2$ such that $(h^L_u(H))_{u \in S_1} = (h^L_{v \in S_2}(H)) = C$, as ordered $k$-tuples, while $h(S_1, \hat{H}_L) \neq h(S_2, \hat{H}_L)$ also by $L$ steps of GWL-1.*

*Proof.*

**1. Constructing forests from the rooted universal cover trees** :

The first part of the proof is similar to the first part of the proof of Theorem 2 of [338].

Consider an arbitrary node $v \in \mathcal{V}$ and denote the $2L$-hop tree rooted at $v$ from the universal cover as $(\tilde{\mathcal{B}}^{2L}_{\mathcal{V},\mathcal{E}})_v$ as in Theorem 8.10.20. As each node $v \in \mathcal{V}$ has volume $vol(v) = \sum_{v \in e}|e| = O(\log^{\frac{1-\epsilon}{4L}} n)$, then every edge $e \in \mathcal{E}$ has $|e| = O(\log^{\frac{1-\epsilon}{4L}} n)$ and for all $v \in \mathcal{V}$ we have that $\deg(v) = O(\log^{\frac{1-\epsilon}{4L}} n)$, we can say that every node in $(\tilde{\mathcal{B}}^{2L}_{\mathcal{V},\mathcal{E}})_{\tilde{v}}$ has degree $d = O(\log^{\frac{1-\epsilon}{4L}} n)$.

Thus, the number of nodes in $(\tilde{\mathcal{B}}^{2L}_{\mathcal{V},\mathcal{E}})_{\tilde{v}}$, denoted by $|\mathcal{V}((\tilde{\mathcal{B}}^{2L}_{\mathcal{V},\mathcal{E}})_{\tilde{v}}|$, satisfies $|\mathcal{V}((\tilde{\mathcal{B}}^{2L}_{\mathcal{V},\mathcal{E}})_{\tilde{v}}| \leq \sum_{i=0}^{2L} d^{i} = O(d^{2L}) = O(\log^{\frac{1-\epsilon}{2}} n)$. We set $K \triangleq \max_{v \in V} |\mathcal{V}((\tilde{\mathcal{B}}^{2L}_{\mathcal{V},\mathcal{E}})_{\tilde{v}}|$ as the maximum number of nodes of $(\tilde{\mathcal{B}}^{2L}_{\mathcal{V},\mathcal{E}})_{\tilde{v}}$ and thus $K = O(\log^{\frac{1-\epsilon}{2}} n)$. For all $v \in \mathcal{V}$, expand trees $(\tilde{\mathcal{B}}^{2L}_{\mathcal{V},\mathcal{E}})_{\tilde{v}}$ to $\overline{(\tilde{\mathcal{B}}^{2L}_{\mathcal{V},\mathcal{E}})_{\tilde{v}}}$ by adding $K - |\mathcal{V}((\tilde{\mathcal{B}}^{2L}_{\mathcal{V},\mathcal{E}})_{\tilde{v}}|$ independent nodes. Then, all $\overline{(\tilde{\mathcal{B}}^{L}_{\mathcal{V},\mathcal{E}})_{\tilde{v}}}$ have the same number of nodes, which is $K$, becoming forests instead of trees.

**2. Counting $|\mathcal{G}_L|$:**

Next, we consider the number of non-isomorphic forests over $K$ nodes. Actually, the number of non-isomorphic graphs over K nodes is bounded by $2^{\binom{K}{2}} = exp(O(\log^{\frac{1-\epsilon}{2}} n)) = o(n^{1-\epsilon})$. Therefore, due to the pigeonhole principle, there exist $\frac{n}{o(n^{1-\epsilon})} = \omega(n^{\epsilon})$ many nodes $v$ whose $\overline{(\tilde{\mathcal{B}}^{L}_{\mathcal{V},\mathcal{E}})_{\tilde{v}}}$ are isomorphic to each other. Denote $\mathcal{G}_L$ as the set of all $L$-GWL-1 values. Denote the set of these nodes as $\mathcal{V}_{c_L}$, which consist of nodes whose $L$-GWL-1 values are all the same value $c_L \in \mathcal{G}_L$ after $L$ iterations of GWL-1 by Theorem 8.10.20. For a fixed $L$, the sets $\mathcal{V}_{c_L}$ form a partition of $\mathcal{V}$, in other words, $\bigsqcup_{c_L \in \mathcal{G}_L} \mathcal{V}_{c_L} = \mathcal{V}$. Next, we focus on looking at $k$-sets of nodes that are not equivalent by GWL-1.

For any $c_L \in \mathcal{G}_L$, there is a partition $\mathcal{V}_{c_L} = \bigsqcup_s V_{c_L,s}$ where $V_{c_L,s}$ is the set of nodes all of which have $L$-GWL-1 class $c_L$ and that belong to a connected component of size $s$ in $\mathcal{H}_{c_L}$. Let $\mathcal{S}_{c_L} \triangleq \{|\mathcal{V}_{\mathcal{R}_{c_L,j}}| : \mathcal{R}_{c_L,j} \in \mathcal{C}_{c_L}\}$ denote the set of sizes $s \geq 1$ of connected component node sets of $\mathcal{H}_{c_L}$. We know that $|\mathcal{S}_{c_L}| \leq S$ where $S$ is independent of $n$.

**3. Computing the lower bound:**

Let $Y$ denote the number of pairs of $k$-node sets $S_1 \not\simeq S_2$ such that $(h^L_u(H))_{u \in S_1} = (h^L_v(H))_{v \in S_2} = C = (c_{(L,1)} \ldots c_{(L,k)})$, as ordered tuples, from $L$-steps of GWL-1. Since if any pair of nodes $u, v$ have the same $L$-GWL-1 values $c_L$, then they become distinguishable by the size of the connected component in $\mathcal{H}_{c_L}$ that they belong to. We can lower bound $Y$ by counting over all pairs of $k$ tuples of nodes $((u_1 \ldots u_k), (v_1 \ldots v_k)) \in (\prod_{i=1}^k \mathcal{V}_{c_{(L,i)}}) \times (\prod_{i=1}^k \mathcal{V}_{c_{(L,i)}})$ that both have $L$-GWL-1 values $(c_{(L,1)} \ldots c_{(L,k)})$ where there is atleast one $i \in \{1..k\}$ where $u_i$ and $v_i$ belong to different sized connected components $s_i, s'_i \in \mathcal{S}_{c_{(L,i)}}$ with $s_i \neq s'_i$. We have:

$$Y \geq \frac{1}{k!}[\sum_{\substack{((s_i)_{i=1}^k, (s'_i)_{i=1}^k) \in [(\prod_{i=1}^k \mathcal{S}^L_{c_{(L,i)}})]^2 \\ :(s_i)_{i=1}^k \neq (s'_i)_{i=1}^k}} \prod_{i=1}^k |V^l_{(c_{(L,i)}),s_i}||V^l_{(c_{(L,i)}),s'_i}|] \tag{8.82a}$$

$$= \frac{1}{k!}[\prod_{i=1}^{k}(\sum_{s_i \in \mathcal{S}_{c_i}^{L}} |V^{l}_{(c_{(L,i)}),s_i}|)^2 - \sum_{(s_i)_{i=1}^{k} \in \prod_{i=1}^{k} \mathcal{S}^{L}_{(c_{(L,i)})}} (\prod_{i=1}^{k} |V^{l}_{(c_{(L,i)}),s_i}|^2)] \tag{8.82b}$$

Using the fact that for each i $\in \{1...k\}$, $|\mathcal{V}_{c_{(L,i)}}| = \sum_{s_i \in \mathcal{S}_{c_{(L,i)}}} |V_{(c_{(L,i)}),s_i}|$ and by assumption $|V_{(c_{(L,i)}),s_i}| = O(\frac{n^{\epsilon}}{\log^{\frac{1}{2k}} n})$ for any $s_i \in \mathcal{S}_{c_{(L,i)}}$, thus we have:

$$Y \geq \omega(n^{2k\epsilon}) - O(|S|^k \frac{n^{2k\epsilon}}{\log n})] = \omega(n^{2k\epsilon}) \tag{8.83}$$

□

(8.84) **Example:** A simple example of a hypergraph that statisfies the conditions of Theorem 8.10.38 is a union of many disconnected hypergraphs $\mathcal{H} = \cup_i \mathcal{H}_i = (\mathcal{V}, \mathcal{E})$ with $|\mathcal{V}_{\mathcal{H}_i}| \leq S$ where $S < \infty$ is a small constant independent of $N = |\mathcal{V}| \geq S$. Such a hypergraph could be a social network where the nodes are user instances and the hyperedges are private groups. The disconnected hypergraphs represent disconnected communities where a user can only belong to a single community.

Even though Theorem 8.10.38 does not depend on the cardinality of a set of disconnected hypergraphs $\mathcal{H}_i$ indistinguishable by GWL-1, due to the disconnected nature of $\mathcal{H}$ and the small size of its components, there is a large chance of obtaining a large number of such components. We give a very rough estimate of this in the following:

Assuming that $\mathcal{H} = \cup_i \mathcal{H}_i$ has each $\mathcal{H}_i$ i.i.d. sampled from a distribution of $s$-uniform $d$-regular hypergraphs of $n$ nodes, denoted $\mathcal{R}_{n,s,d}$: $P(\mathcal{H}_i = \mathcal{R}_{n,s,d})$. If the parameters $(n, s, d)$ for $\mathcal{R}_{n,s,d}$ satisfy $nd = |\mathcal{E}|s$ where $|\mathcal{E}| \in \mathbb{Z}^+$, then a well defined hypergraph is formed. This distribution can be factorized as follows:

$$P(\mathcal{H}_i = \mathcal{R}_{n,s,d}) = P(\deg(v) = d \mid r = s, |\mathcal{V}| = n, nd \mod s \equiv 0) P(r = s \mid |\mathcal{V}| = n) P(|\mathcal{V}| = n) \tag{8.85}$$

where:

$$P(\deg(v) = d \mid r = s, |\mathcal{V}(\mathcal{H}_\mathrm{i})| = n, nd \mod s \equiv 0) \geq \frac{1}{\binom{n-1}{s-1}^d} \geq \frac{1}{\binom{n}{s}^d} \geq \frac{1}{S^{dS}}, \forall v \in \mathcal{V}_{\mathcal{H}_\mathrm{i}} \tag{8.86a}$$

$$P(r = s \mid |\mathcal{V}(\mathcal{H}_\mathrm{i})| = S) = \frac{1}{n} \geq \frac{1}{S}, P(|\mathcal{V}(\mathcal{H}_\mathrm{i})| = n) = \frac{1}{S} \text{ for } n : n \leq S \leq N \tag{8.86b}$$

and we have that

$$P(\mathcal{H}_1 \text{ is neighborhood regular}) \geq P(\mathcal{H}_1 \text{ is a cycle graph}) \geq \frac{1}{S}^{2S+2} \tag{8.87}$$

and that:

$$P(h(\mathcal{H}_\mathrm{i}, \mathcal{H}) = h(\mathcal{H}_1, \mathcal{H}), \mathcal{H}_1 \text{ is neighborhood regular}) \tag{8.88a}$$

$$\geq P(h(\mathcal{H}_\mathrm{i}, \mathcal{H}) = h(\mathcal{H}_1, \mathcal{H}), \mathcal{H}_1 \text{ is a cycle graph of length } S) \tag{8.88b}$$

$$\geq P(\mathcal{H}_\mathrm{i} \text{ is a cycle graph of length } S) P(\mathcal{H}_1 \text{ is a cycle graph of length } S) \tag{8.88c}$$

$$\geq \frac{1}{S}^{2(2S+2)}, \forall \mathrm{i} > 1 \tag{8.88d}$$

where a cycle graph is a 2-uniform hypergraph where each node has degree 2.

Since we sample each $\mathcal{H}_\mathrm{i}$ i.i.d., the indicator random variable is a Bernoulli random variable. By Hoeffding's inequality on the sum of Bernoulli random variables, we get:

$$Pr(\sum_{\mathrm{i}=1}^{m} \mathbf{1}[\mathcal{H}_\mathrm{i} \text{ is neighborhood regular and } h(\mathcal{H}_\mathrm{i}, \mathcal{H}) = h(\mathcal{H}_1, \mathcal{H})] \geq (\frac{m}{S^{4S+4}} + t)) \leq \mathrm{e}^{\frac{-2t^2}{m}} \tag{8.89}$$

where $\sum_{\mathrm{i}=1}^{m} |\mathcal{V}_{\mathcal{H}_\mathrm{i}}| = |\mathcal{V}|$. This means that with large number of samples $m$, or large $n$, it is possible for the number of regular hypergraphs $\mathcal{H}_\mathrm{i}$ equivalent up to GWL-1 to be atleast of order $\Omega(\sqrt{m}) + \frac{m}{S^{4S+4}}$ with high probability. This is one of the simplest examples that demonstrates Theorem 8.10.38.

For the following proof, we will denote $\cong_{\mathcal{H}}$ as a node or hypergraph automorphism with respect to a hypergraph $\mathcal{H}$.

**Theorem 8.10.39.** *(Invariance and Expressivity) If $L = \infty$, GWL-1 enhanced by Algorithm 29 is still invariant to node isomorphism classes of $\mathcal{H}$ and can be strictly more expressive than GWL-1 to determine node isomorphism classes.*

*Proof.*

**1. Expressivity**:

Let $L \in \mathbb{Z}^+$ be arbitrary. We first prove that $L$-GWL-1 enhanced by Algorithm 29 is strictly more expressive for node distinguishing than $L$-GWL-1 on some hypergraph(s). Let $C_4^3$ and $C_5^3$ be two 3-regular hypergraphs from Figure 8.2. Let $\mathcal{H} = C_4^3 \bigsqcup C_5^3$ be the disjoint union of the two regular hypergraphs. $L$ iterations of GWL-1 will assign the same node class to all of $\mathcal{V}_{\mathcal{H}}$. These two subhypergraphs can be distinguished by $L$-GWL-1 for $L \geq 1$ after editing the hypergraph $\mathcal{H}$ from the output of Algorithm 29 and becoming $\hat{\mathcal{H}} = \hat{C}_4^3 \cup \hat{C}_5^3$. This is all shown in Figure 8.2. Since $L$ was arbitrary, this is true for $L = \infty$.

**2. Invariance**:

For any hypergraph $\mathcal{H}$, let $\hat{\mathcal{H}} = (\hat{\mathcal{V}}, \hat{\mathcal{E}})$ be $\mathcal{H}$ modified by the output of Algorithm 29 by adding hyperedges to $\mathcal{V}_{\mathcal{R}_{c,i}}$. GWL-1 remains invariant to node isomorphism classes of $\mathcal{H}$ on $\hat{\mathcal{H}}$.

**a. Case 1** (node $u \in \mathcal{V}$ has its class $c$ changed to class $c_s$):

Let $L \in \mathbb{Z}^+$ be arbitrary. For any node $u$ with $L$-GWL-1 class $c$ changed to $c_s$ in $\hat{\mathcal{H}}$, if $u \cong_{\mathcal{H}} v$ for any $v \in \mathcal{V}$, then the GWL-1 class of $v$ must also be $c_s$. In otherwords, both $u$ and $v$ belong to $s$-sized connected components in $\mathcal{H}_c$ We prove this by contradiction.

Say $u$ belong to a $L$-GWL-1 symmetric induced subhypergraph $S$ with $|\mathcal{V}_S| = s$.

**i.** Say $v$ is originally of $L$-GWL-1 class $c$ and changes to $L$-GWL-1 $c_{s'}$ for $s' < s$ on $\hat{\mathcal{H}}$, WLOG.

If this is the case then $v$ belongs to a $L$-GWL-1 symmetric induced subhypergraph $S'$ with $|\mathcal{V}_{S'}| = s'$. Since there is a $\pi \in Aut(\mathcal{H})$ with $\pi(u) = v$ and since $s' < s$, by the pigeonhole principle some node $w \in \mathcal{V}_S$ must have $\pi(w) \notin \mathcal{V}_{S'}$. Since $S$ and $S'$ are maximally connected, $\pi(w)$ cannot share the same $L$-GWL-1 class as $w$. Thus, it must be that $(\tilde{B}^{2L}_{\mathcal{V},\mathcal{E}})_{\widetilde{\pi(w)}} \ncong_c (\tilde{B}^{2L}_{\mathcal{V},\mathcal{E}})_{\tilde{w}}$ where $\tilde{w}, \widetilde{\pi(w)}$ are the lifts of $w, \pi(w)$ by universal covering map $p_{\mathcal{B}_{\mathcal{V},\mathcal{E}}}$. However $w$ and $\pi(w)$ both belong to $L$-GWL-1 class $c$ in $\mathcal{H}$, meaning $(\tilde{B}^{2L}_{\mathcal{V},\mathcal{E}})_{\widetilde{\pi(w)}} \cong_c (\tilde{B}^{2L}_{\mathcal{V},\mathcal{E}})_{\tilde{w}}$, contradiction.

**ii.** Say node $v \in \mathcal{V}$ has its class $c$ unchanged.

The argument for when $v$ does not change its class $c$ after the algorithm, follows by noticing that since $c$ is the GWL-1 node class of $u$, $c_s$ is the GWL-1 node class of $v$ and $c \neq c_s$. Thus we must have $u \ncong_{\mathcal{H}} v$ once again by the contrapositive of Theorem 8.10.20. This also gives a contradiciton.

Since $L$ was arbitrary, the contradiction must be true for $L = \infty$.

**b. Case 2** (node $u \in \mathcal{V}$ has its class $c$ unchanged):

Now assume $L = \infty$. Let $p_{\mathcal{B}_{\mathcal{V},\mathcal{E}}}$ be the universal covering map of $\mathcal{B}_{\mathcal{V},\mathcal{E}}$. For all other nodes $u' \cong_{\mathcal{H}} v'$ for $u', v' \in \mathcal{V}$ unaffected by the replacement, meaning they do not belong to any $\mathcal{R}_{c,\mathrm{i}}$ discovered by the algorithm, if the rooted universal covering tree rooted at node $\tilde{u}'$ connects to any node $\tilde{w}$ in $l$ hops in $(\tilde{\mathcal{B}}^l_{\mathcal{V},\mathcal{E}})_{\tilde{u}'}$ where $p_{\mathcal{B}_{\mathcal{V},\mathcal{E}}}(\tilde{u}') = u', p_{\mathcal{B}_{\mathcal{V},\mathcal{E}}}(\tilde{w}) = w$ and where $w$ has any class $c$ in $\mathcal{H}$, then $\tilde{v}'$ must also connect to a node $\tilde{z}$ in $l$ hops in $(\tilde{\mathcal{B}}^l_{\mathcal{V},\mathcal{E}})_{\tilde{u}'}$ where $p_{\mathcal{B}_{\mathcal{V},\mathcal{E}}}(\tilde{z}) = z$ and $w \cong_{\mathcal{H}} z$. Furthermore, if $w$ becomes class $c_s$ in $\mathcal{H}$ due to the algorithm, then $z$ also becomes class $c_s$ in $\hat{\mathcal{H}}$. This will follow by the previous result on isomorphic $w$ and $z$ both of class $c$ with $w$ becoming class $c_s$ in $\hat{\mathcal{H}}$.

Since $L = \infty$: For any $w \in \mathcal{V}$ connected by some path of hyperedges to $u' \in \mathcal{V}$, consider the smallest $l$ for which $(\tilde{B}^l_{\mathcal{V},\mathcal{E}})_{\tilde{u}'}$, the $l$-hop universal covering tree of $\mathcal{H}$ rooted at $\tilde{u}'$, the lift of $u'$, contains the lifted $\tilde{w}$ of $w \in \mathcal{V}$ with GWL-1 node class $c$ at layer $l$. Since $u' \cong_{\mathcal{H}} v'$ by $\boldsymbol{\pi}$. We can use $\boldsymbol{\pi}$ to find some $z = \boldsymbol{\pi}(w)$.

We claim that $\tilde{z}$ is $l$ hops away from $\tilde{v}'$. Since $u' \cong_{\mathcal{H}} v'$ due to some $\boldsymbol{\pi} \in Aut(\mathcal{H})$ with $\boldsymbol{\pi}(u') = v'$, using Proposition 8.10.8 for singleton nodes and by Theorem 8.10.20 we must have $(\tilde{\mathcal{B}}^l_{\mathcal{V},\mathcal{E}})_{\tilde{u}'} \cong_c (\tilde{\mathcal{B}}^l_{\mathcal{V},\mathcal{E}})_{\tilde{v}'}$ as isomorphic rooted universal covering trees due to an induced isomorphism $\tilde{\boldsymbol{\pi}}$ of $\boldsymbol{\pi}$ where we define an induced isomorphism $\tilde{\boldsymbol{\pi}} : (\tilde{\mathcal{B}}_{\mathcal{V},\mathcal{E}})_{\tilde{u}'} \to (\tilde{\mathcal{B}}_{\mathcal{V},\mathcal{E}})_{\tilde{v}'}$ between rooted universal covers $(\tilde{\mathcal{B}}_{\mathcal{V},\mathcal{E}})_{\tilde{u}'}$ and $(\tilde{\mathcal{B}}_{\mathcal{V},\mathcal{E}})_{\tilde{v}'}$ for $\tilde{u}', \tilde{v}' \in \mathcal{V}(\tilde{\mathcal{B}}_{\mathcal{V},\mathcal{E}})$ as $\tilde{\boldsymbol{\pi}}(\tilde{a}) = \tilde{b}$ if $\boldsymbol{\pi}(a) = b$ $\forall a, b \in \mathcal{V}(\mathcal{B}_{\mathcal{V},\mathcal{E}})$ connected to $u'$ and $v'$ respectively and $p_{\mathcal{B}_{\mathcal{V},\mathcal{E}}}(\tilde{a}) = a$, $p_{\mathcal{B}_{\mathcal{V},\mathcal{E}}}(\tilde{b}) = b$. Since $l$ is the shortest path distance from $\tilde{u}'$ to $\tilde{w}$, there must exist some shortest (as defined by the path length in $\mathcal{B}_{\mathcal{V},\mathcal{E}}$) path $P$ of hyperedges from $u'$ to $w$ with no cycles. Using $\boldsymbol{\pi}$, we must map $P$ to another acyclic shortest path of the same length from $v'$ to $z$. This path correponds to a $l$ length shortest path from $\tilde{v}'$ to $\tilde{z}$ in $(\tilde{\mathcal{B}}_{\mathcal{V},\mathcal{E}})_{\tilde{v}'}$.

If $w$ has GWL-1 class $c$ in $\mathcal{H}$ that doesn't become affected by the algorithm, then $z$ also has GWL-1 class $c$ in $\mathcal{H}$ since $w \cong_{\mathcal{H}} z$.

If $w$ has class $c$ and becomes $c_s$ in $\hat{\mathcal{H}}$, by the previous result, since $w \cong_{\mathcal{H}} z$ we must have the GWL-1 classes $c'$ and $c''$ of $w$ and $z$ in $\hat{\mathcal{H}}$ be both equal to $c_s$.

The node $w$ connected to $u'$ was arbitrary and so both $\tilde{w}$ and the isomorphism induced $\tilde{z}$ are $l$ hops away from $\tilde{u}'$ and $\tilde{v}'$ respectively, with the same GWL-1 class $c'$ in $\hat{\mathcal{H}}$, thus $(\tilde{\mathcal{B}}_{\hat{\mathcal{V}},\hat{\mathcal{E}}})_{\tilde{u}'} \cong_c (\tilde{\mathcal{B}}_{\hat{\mathcal{V}},\hat{\mathcal{E}}})_{\tilde{v}'}$.

We have thus shown, if $u \cong_{\mathcal{H}} v$ for $u, v \in \mathcal{H}$, then in $\hat{\mathcal{H}}$ we have $h_u^L(\hat{H}) = h_v^L(\hat{H})$ using the duality between universal covers and GWL-1 from Theorem 8.10.20 and Proposition 8.10.30. □

Here we redefine the symmetry group of the $L$-GWL-1 node representation map as in the main paper:

$$\mathrm{Sym}(h^L(\hat{H}_L)) \triangleq \bigcap_{\hat{H}'_L \sim P(\hat{H}_L)} \mathrm{Sym}(h^L(\hat{H}'_L)) \tag{8.90}$$

**Proposition 8.10.40.** *The multi-hypergraph $\hat{\mathcal{H}}_L$ breaks the symmetry of the L-GWL-1 view of the hypergraph $\mathcal{H}$:*

$$Sym(h^L(\hat{H}_L)) \subseteq Aut_c(\tilde{\mathcal{B}}^{2L}_{\mathcal{V},\mathcal{E}}), \forall L \geq 1 \tag{8.91}$$

*Proof.* By Equation 8.90 we have that $\mathrm{Sym}(h^L(\hat{H}_L)) = \bigcap_{\hat{H}'_L \sim P(\hat{H}_L)} \mathrm{Sym}(h^L(\hat{H}'_L))$. Since the original incidence matrix $H$ belongs to $supp(P(\hat{H}_L))$, we must have $\bigcap_{\hat{H}'_L \sim P(\hat{H}_L)} \mathrm{Sym}(h^L(\hat{H}'_L)) \subseteq \mathrm{Sym}(h^L(H))$. Since $\mathrm{Sym}(h^L(H)) \cong Aut_c(\tilde{\mathcal{B}}^{2L}_{\mathcal{V},\mathcal{E}})$, we thus have:

$$\mathrm{Sym}(h^L(\hat{H}_L)) = \bigcap_{\hat{H}'_L \sim P(\hat{H}_L)} \mathrm{Sym}(h^L(\hat{H}'_L)) \subseteq \mathrm{Sym}(h^L(H)) \cong Aut_c(\tilde{\mathcal{B}}^{2L}_{\mathcal{V},\mathcal{E}}), \forall L \geq 1 \tag{8.92}$$

which proves the symmetry breaking statement. □

**Proposition 8.10.41.** *(Complexity) Let $H$ be the star expansion matrix of $\mathcal{H}$. Algorithm 29 runs in time $O(L \cdot nnz(H) + (n + m))$, the size of the input hypergraph when viewing L as constant, where $n$ is the number of nodes, $nnz(H) = vol(\mathcal{V}) \triangleq \sum_{v \in \mathcal{V}} deg(v)$ and $m$ is the number of hyperedges.*

*Proof.* Computing $E_{deg}$, which requires computing the degrees of all the nodes in each hyperedge takes time $O(nnz(H))$. The set $E_{deg}$ can be stored as a hashset datastructure. Constructing this takes $O(nnz(H))$. Computing GWL-1 takes $O(L \cdot nnz(H))$ time assuming a constant $L$ number of iterations. Constructing the bipartite graphs for $H$ takes time $O(nnz(H) + n + m)$ since it is an information preserving data structure change. Define for each $c \in C$, $n_c := |\mathcal{V}_c|, m_c := |\mathcal{E}_c|$. Since the classes partition $\mathcal{V}$, we must have:

$$n = \sum_{c \in C} n_c; m = \sum_{c \in C} m_c; nnz(H) = \sum_{c \in C} nnz(H_c) \tag{8.93}$$

where $H_c$ is the star expansion matrix of $\mathcal{H}_c$. Extracting the subgraphs can be implemented as a masking operation on the nodes taking time $O(n_c)$ to form $\mathcal{V}_c$ followed by searching over the neighbors of $\mathcal{V}_c$ in time $O(m_c)$ to construct $\mathcal{E}_c$. Computing the connected components for $\mathcal{H}_c$ for a predicted node class $c$ takes time $O(n_c + m_c + nnz(H_c))$. Iterating over each connected component for a given $c$ and extracting their nodes and hyperedges takes time $O(n_{c_{\mathrm{i}}} + m_{c_{\mathrm{i}}})$ where $n_c = \sum_{\mathrm{i}} n_{c_{\mathrm{i}}}, m_c = \sum_{\mathrm{i}} m_{c_{\mathrm{i}}}$. Checking that a connected component has size at least 3 takes $O(1)$ time. Computing the degree on $\mathcal{H}$ for all nodes in the connected component takes time $O(n_{c_{\mathrm{i}}})$ since computing degree takes $O(1)$ time. Checking that the set of node degrees of the connected component doesn't belong to $E_{deg}$ can be implemented as a check that the hash of the set of degrees is not in the hashset datastructure for $E_{deg}$.

Adding up all the time complexities, we get the total complexity is:

$$O(nnz(H)) + O(nnz(H) + n + m) + \sum_{c \in C}\Big(O(n_c + m_c + nnz(H_c)) + \sum_{\text{conn. comp. i of } \mathcal{H}_c} O(n_{c_{\mathrm{i}}} + m_{c_{\mathrm{i}}})\Big) \tag{8.94a}$$

$$= O(nnz(H) + n + m) + \sum_{c \in C}\big(O(n_c + m_c + nnz(H_c)) + O(n_c + m_c)\big) \tag{8.94b}$$

$$= O(nnz(H) + n + m) \tag{8.94c}$$

□

**Proposition 8.10.42.** *For a connected hypergraph $\mathcal{H}$, let $(\mathcal{R}_V, \mathcal{R}_E)$ be the output of Algorithm 29 on $\mathcal{H}$. Then there are Bernoulli probabilities $p, q_{\mathrm{i}}$ for $\mathrm{i} = 1 \ldots |\mathcal{R}_V|$ for attaching a covering hyperedge so that $\hat{\pi}$ is an unbiased estimator of $\pi$.*

*Proof.* Let $\mathcal{C}_{c_L} = \{\mathcal{R}_{c_L,\mathrm{i}}\}_\mathrm{i}$ be the maximally connected components induced by the vertices with $L$-GWL-1 values $c_L$. The set of vertex sets $\{\mathcal{V}(\mathcal{R}_{c_L,\mathrm{i}})\}$ and the set of all hyperedges $\cup_\mathrm{i}\{\mathcal{E}(\mathcal{R}_{c_L,\mathrm{i}})\}$ over all the connected components $\mathcal{R}_{c_L,\mathrm{i}}$ for $\mathrm{i} = 1...\mathcal{C}_{c_L}$ form the pair $(\mathcal{R}_V, \mathcal{R}_E)$.

For a hypergraph random walk on connected $\mathcal{H} = (\mathcal{V}, \mathcal{E})$, its stationary distribution $\pi$ on $\mathcal{V}$ is given by the closed form:

$$\pi(v) = \frac{\deg(v)}{\sum_{u \in \mathcal{V}} \deg(u)} \tag{8.95}$$

for $v \in \mathcal{V}$.

Let $\hat{\mathcal{H}} = (\mathcal{V}, \hat{\mathcal{E}})$ be the random multi-hypergraph as determined by $p$ and $q_\mathrm{i}$ for $\mathrm{i} = 1...|\mathcal{R}_V|$. These probabilities determine $\hat{\mathcal{H}}$ by the following operations on the hypergraph $\mathcal{H}$:

- Attaching a single hyperedge that covers $\mathcal{V}_{\mathcal{R}_{c_L,\mathrm{i}}}$ with probability $q_\mathrm{i}$ and not attaching with probability $1 - q_\mathrm{i}$.
- All the hyperedges in $\mathcal{R}_{c_L,\mathrm{i}}$ are dropped/kept with probability $p$ and $1 - p$ respectively.

**1. Setup:**

Let $\deg(v) \triangleq |\{\mathrm{e} : \mathrm{e} \ni v\}|$ for $v \in \mathcal{V}(\mathcal{H})$ and $\deg(\mathcal{S}) \triangleq \sum_{u \in \mathcal{V}(\mathcal{S})} \deg(u)$ for $\mathcal{S} \subseteq \mathcal{H}$ a subhypergraph.

Let

$$Bernoulli(p) \triangleq \begin{cases} 1 & \text{prob. } p \\ 0 & \text{prob. } 1 - p \end{cases} \tag{8.96}$$

and

$$Binom(n, p) \triangleq \sum_{\mathrm{i}=1}^{n} Bernoulli(p) \tag{8.97}$$

Define for each $v \in \mathcal{V}$, $C(v)$ to be the unique $\mathcal{R}_{c_L,\mathrm{i}}$ where $v \in \mathcal{R}_{c_L,\mathrm{i}}$ This means that we have the following independent random variables:

$$X_\mathrm{e} \triangleq Bernoulli(1 - p), \forall \mathrm{e} \in \mathcal{E} \text{ (i.i.d. across all } \mathrm{e} \in \mathcal{E}) \tag{8.98a}$$

$$X_{C(v)} \triangleq Bernoulli(q_\mathrm{i}) \tag{8.98b}$$

As well as the following constant, depending only on $C(v)$:

$$m_{C(v)} \triangleq \sum_{u \in \mathcal{V} \setminus C(v)} \deg(u) \tag{8.99}$$

where $C(v) \subseteq \mathcal{V}, \forall v \in \mathcal{V}$

Let $\hat{\pi}$ be the stationary distribution of $\hat{H}$. Its expectation $\mathbb{E}[\hat{\pi}]$ can be written as:

$$\hat{\pi}(v) \triangleq \frac{\sum_{\mathrm{e} \ni v : \mathrm{e} \in \mathcal{E}} X_{\mathrm{e}} + X_{C(v)}}{m_{C(v)} + \sum_{\mathrm{e} \ni u : u \in C(v), \mathrm{e} \in \mathcal{E}} X_{\mathrm{e}} + X_{C(v)}} \tag{8.100}$$

Letting

$$N_v \triangleq \sum_{\mathrm{e} \ni v : \mathrm{e} \in \mathcal{E}} X_{\mathrm{e}} = Binom(\deg(v), 1-p) \tag{8.101a}$$

$$N \triangleq N_v + X_{C(v)} \tag{8.101b}$$

$$D \triangleq m_{C(v)} + \sum_{\mathrm{e} \ni u : u \in C(v), \mathrm{e} \in \mathcal{E}} X_{\mathrm{e}} + X_{C(v)} \tag{8.101c}$$

$$C \triangleq D - \Big(\sum_{\mathrm{e} \ni v : \mathrm{e} \in \mathcal{E}} |\mathrm{e}|\Big) N_v - m_{C(v)} = \sum_{\mathrm{e} \ni u : v \notin \mathrm{e}, u \in C(v), \mathrm{e} \in \mathcal{E}} X_{\mathrm{e}} - m_{C(v)} = Binom(F_v, 1-p) - m_{c(v)} \tag{8.101d}$$

$$\text{where } F_v \triangleq |\{\mathrm{e} \ni u : v \notin \mathrm{e}, u \in C(v), \mathrm{e} \in \mathcal{E}\}|$$

and so we have: $\hat{\pi}(v) = \frac{N}{D}$

We have the following joint independence $N_v \perp X_{C(v)} \perp C$ due to the fact that each random variable describes disjoint hyperedge sets.

**2. Computing the Expectation:**

Writing out the expectation with conditioning on the joint distribution $P(D, N_v, X_{C(v)})$, we have:

$$\mathbb{E}[\hat{\pi}(v)] = \sum_{b=0}^{1} \sum_{\mathrm{j}=0}^{\deg(v)} \sum_{k=m_{C(v)}}^{\deg(C(v))} \mathbb{E}[\hat{\pi}(v) \mid D = k, N = \mathrm{j}] P(D = k, N_v = \mathrm{j}, X_{C(v)} = b) \tag{8.102a}$$

$$= \sum_{b=0}^{1} \sum_{\mathrm{j}=0}^{\deg(v)} \sum_{k=m_{C(v)}}^{\deg(C(v))} \frac{1}{k} \mathbb{E}[N \mid D = k, N_v = \mathrm{j}, X_{C(v)} = b] P(D = k, N_v = \mathrm{j}, X_{C(v)} = b) \tag{8.102b}$$

$$= \sum_{b=0}^{1} \sum_{j=0}^{\deg(v)} \sum_{k=m_{C(v)}}^{\deg(C(v))} \frac{j+b}{k} P(D=k, N_v = j, X_{C(v)} = b) \tag{8.102c}$$

$$= \sum_{b=0}^{1} \sum_{j=0}^{\deg(v)} \sum_{k=m_{C(v)}}^{\deg(C(v))} \frac{j+b}{k} P(D=k) P(N_v = j) P(X_{C(v)} = b) \tag{8.102d}$$

$$= \sum_{b=0}^{1} \sum_{j=0}^{\deg(v)} \sum_{k=m_{C(v)}}^{\deg(C(v))} \frac{j+b}{k} P(C = k - \deg(v)j - m_{C(v)}) P(N_v = j) P(X_{C(v)} = b) \tag{8.102e}$$

$$\begin{aligned} = \sum_{b=0}^{1} \sum_{j=0}^{\deg(v)} \sum_{k=m_{C(v)}}^{\deg(C(v))} \frac{j+b}{k} P(Binom(F_v, 1-p), 1-p) = k - \deg(v)j - m_{C(v)}) \\ \cdot P(Binom(\deg(v), 1-p) = j) \cdot P(Bernoulli(q_i) = b) \end{aligned} \tag{8.102f}$$

$$= \sum_{b=0}^{1} \sum_{j=0}^{\deg(v)} \sum_{k=m_{C(v)}}^{\deg(C(v))} \frac{j+b}{k} \binom{F_v}{k - \deg(v)j - m_{C(v)}} (\frac{1}{2})^{F_v} \cdot \binom{\deg(v)}{j} (\frac{1}{2})^{\deg(v)} \cdot P(Bernoulli(q_i) = b) \tag{8.102g}$$

$$\begin{aligned} = \sum_{b=0}^{1} \sum_{j=0}^{\deg(v)} \sum_{k=m_{C(v)}}^{\deg(C(v))} \frac{j+b}{k} \binom{F_v}{k - \deg(v)j - m_{C(v)}} (\frac{1}{2})^{F_v} \\ \cdot \binom{\deg(v)}{j} (\frac{1}{2})^{\deg(v)} \cdot P(Bernoulli(q_i) = b) \end{aligned} \tag{8.102h}$$

$$= \sum_{j=0}^{\deg(v)} \sum_{k=m_{C(v)}}^{\deg(C(v))} \binom{F_v}{k - \deg(v)j - m_{C(v)}} (\frac{1}{2})^{F_v} \cdot \binom{\deg(v)}{j} (\frac{1}{2})^{\deg(v)} [(1-q_i)\frac{j}{k} + q_i \frac{j+1}{k}] \tag{8.102i}$$

$$= \sum_{j=0}^{\deg(v)} \sum_{k=m_{C(v)}}^{\deg(C(v))} \binom{F_v}{k - \deg(v)j - m_{C(v)}} (1-p)^{F_v - (k - \deg(v))} p^{k-\deg(v)} \cdot \binom{\deg(v)}{j} (1-p)^{j} p^{\deg(v) - j} [\frac{j}{k} + q_i \frac{1}{k}] \tag{8.102j}$$

$$= C_1(p) + q_i C_2(p) \tag{8.102k}$$

**3. Pick $p$ and $q_i$:**

We want to find $p$ and $q_i$ so that $\mathbb{E}[\hat{\pi}(v)] = C_1(p) + q_i C_2(p) = \pi(v)$

We know that for a given $p \in [0,1]$, we must have:

$$q_i = \frac{\pi(v) - C_1(p)}{C_2(p)} \tag{8.103}$$

In order for $q_i \in [0,1]$, must have $\pi(v) \geq C_1(p)$ and $\pi(v) - C_1(p) \leq C_2(p)$.

**a. Pick $p$ sufficiently large:**

Notice that

$$0 \leq C_1(p) \leq O(\mathbb{E}[\frac{1}{Binom(F_v, 1-p) + m_{C(v)}}]\cdot\mathbb{E}[Binom(\deg(v), 1-p)]) = O(\frac{1}{m_{C(v)}}\deg(v)(1-p)) \tag{8.104}$$

and that

$$0 \leq C_1(p) \leq O(C_2(p)) \tag{8.105}$$

for $p \in [0,1]$ sufficiently large. This is because

$$C_1(p) \leq O(\frac{1}{m_{C(v)}}\deg(v)(1-p)) \tag{8.106}$$

and

$$\Omega(\frac{1}{m_{C(v)}}\deg(v)(1-p)) \leq C_2(p) \tag{8.107}$$

Piecing these two inequalities together gets the desired inequality 8.105.

We can then pick a $p \in [0,1]$ even larger than the previous $p$ so that for the $C' > 0$ which gives $C_1(p) \leq \frac{C'}{m_{C(v)}}\deg(v)(1-p)$, we achieve

$$C_1(p) \leq \frac{C'}{m_{C(v)}}\deg(v)(1-p) < \pi(v) = \frac{\deg(v)}{m_{C(v)} + \sum_{u \in C(v)} \deg(u)} \tag{8.108}$$

We then have that there exists a $s > 1$ so that

$$sC_1(p) = \pi(v) \tag{8.109}$$

Using this relationship, we can then prove that for a sufficiently large $p \in [0,1]$, we must have a $q_i \in [0,1]$

**b. $p \in [0,1]$ sufficiently large implies** $q_i \geq 0$:

We thus have $q_i \geq 0$ since its numerator is nonnegative:

$$\pi(v) - C_1(p) = (s-1)C_1(p) \geq 0 \Rightarrow q_i \geq 0 \tag{8.110}$$

**c.** $p \in [0,1]$ **sufficiently large implies** $q_{\mathrm{i}} \leq 1$:

$$\pi(v) - C_1(p) = sC_1(p) - C_1(p) = (s-1)C_1(p) \leq C_2(p) \Rightarrow q_{\mathrm{i}} \leq 1 \tag{8.111}$$

□

**Theorem 8.10.43.** *Under Assumptions 8.7.1, 8.7.3, and the Second Law of Thermodynamics on the hypergraph viewed as a closed system:*

$$\exists \mathcal{U} \subseteq \mathcal{V}, \mathcal{U} \neq \varnothing, \text{ so that: } n_v((\mathcal{H}_{tr})_{gt}) > n_v((\mathcal{H}_{\mathrm{te}})_{gt}), \forall v \in \mathcal{U}, \text{ with probability } 1 - O(\frac{1}{\sqrt{n}}) \tag{8.112}$$

*Proof.* By the second law of thermodynamics, we must have that between the training hypergraph $(\mathcal{H}_{tr})_{gt} = (\mathcal{V}, (\mathcal{E}_{tr})_{gt})$ and testing hypergraph $(\mathcal{H}_{\mathrm{te}})_{gt} = (\mathcal{V}, (\mathcal{E}_{\mathrm{te}})_{gt})$ which is temporally later than $(\mathcal{H}_{tr})_{gt}$,

$$\Delta S = -\sum_{v \in \mathcal{V}} p_v((\mathcal{H}_{\mathrm{te}})_{gt}) \log(p_v((\mathcal{H}_{\mathrm{te}})_{gt})) + \sum_{v \in \mathcal{V}} p_v((\mathcal{H}_{tr})_{gt}) \log(p_v((\mathcal{H}_{tr})_{gt})) > 0 \tag{8.113a}$$

If we take an upper bound on $\Delta S$, we get the following consequence:

$$0 < \Delta S \leq \sum_{v \in \mathcal{V}} \log(\frac{p_v((\mathcal{H}_{tr})_{gt})}{p_v((\mathcal{H}_{\mathrm{te}})_{gt})}) \Rightarrow p_v((\mathcal{H}_{tr})_{gt}) > p_v((\mathcal{H}_{\mathrm{te}})_{gt}), \forall v \in \mathcal{U}, \text{ for some } \mathcal{U} \subseteq \mathcal{V}, \mathcal{U} \neq \varnothing \tag{8.114}$$

If $n_v((\mathcal{H}_{tr})_{gt}) > n_v((\mathcal{H}_{\mathrm{te}})_{gt}), \forall v \in \mathcal{U} \subseteq \mathcal{V}$, we can conclude that some nodes will shrink the size of their ground truth isomorphism class.

We show that this occurs with high probability by bounding the complementary case.

**Nodes rarely increase their isomorphism class size:**

For the complementary case, $n_v((\mathcal{H}_{tr})_{gt}) \leq n_v((\mathcal{H}_{\mathrm{te}})_{gt}), \forall v \in \mathcal{U} \subseteq \mathcal{V}$, we show that under Assumption 8.7.3, node isomorphism class cardinality growth occurs with low probability due to an anti-concentration bound on multivariate polynomials [339].

In this complementary case, we must have that the nodes in $\mathcal{U} \subseteq \mathcal{V}$ increased the number of nodes isomorphic to them. This implies that each of these nodes $v \in \mathcal{U}$ must have changed their degree vector upon changing $(\mathcal{H}_{tr})_{gt}$ to $(\mathcal{H}_{\text{te}})_{gt}$. This change requires that some other node $u \in \mathcal{V}, u \neq v$ obtains a degree vector equal to the degree vector of $v$.

Let us define this space of all possible testing hyperedge sets with this necessary condition:

$$\text{DegVec}_{\text{eq}}(v,(\mathcal{H}_{\text{te}})_{gt}) \triangleq \{E \in supp(P((\mathcal{E}_{\text{te}})_{gt};t_{\text{te}})) : \exists u \in \mathcal{V}, \text{degvec}_{(\mathcal{V},E)}(v) = \text{degvec}_{(\mathcal{V},E)}(u), u \neq v\ \} \tag{8.115}$$

We can express the relationship between the change in $n_v$ with membership in $\text{DegVec}_{\text{eq}}(v,(\mathcal{H}_{\text{te}})_{gt})$:

$$n_v((\mathcal{H}_{tr})_{gt}) \leq n_v((\mathcal{H}_{\text{te}})_{gt}) \Rightarrow (\mathcal{E}_{\text{te}})_{gt} \in \text{DegVec}_{\text{eq}}(v,(\mathcal{H}_{\text{te}})_{gt}) \tag{8.116}$$

Letting

$$x_{min}(v) = \min_{E \in \text{DegVec}_{\text{eq}}(v,(\mathcal{H}_{\text{te}})_{gt})} |E|, \forall v \in \mathcal{V} \tag{8.117a}$$

and

$$x_{max}(v) = \max_{E \in \text{DegVec}_{\text{eq}}(v,(\mathcal{H}_{\text{te}})_{gt})} |E|, \forall v \in \mathcal{V} \tag{8.117b}$$

We can then say that the number of testing hyperedges $|(\mathcal{E}_{\text{te}})_{gt}|$ is in the interval $[x_{min}(v), x_{max}(v)]$:

$$(\mathcal{E}_{\text{te}})_{gt} \in DegVec_{\text{eq}}(v,(\mathcal{H}_{\text{te}})_{gt}) \Rightarrow x_{min}(v) \leq f((X_u)_{u\in\mathcal{V}}) \leq x_{max}(v) \tag{8.118}$$

Let the diameter of the interval $[x_{min}(v), x_{max}(v)]$ be given by $D$:

$$D \triangleq x_{max}(v) - x_{min}(v) \tag{8.119}$$

This only depends on $n$ since the minimizers and maximizers $E_{x_{min}}, E_{x_{max}}$ of Equations 8.117a and 8.117b satisfy: $E_{x_{min}}, E_{x_{max}} \in \text{DegVec}_{\text{eq}}(v,(\mathcal{H}_{\text{te}})_{gt})$ and thus belong to the $supp(P(E;t_{\text{te}}))$. We know that any $E \in supp(P(E;t_{\text{te}}))$ has that $|E|$ depends only on $n$ by Assumption 8.7.3.

Define the median $M$ on $[x_{min}(v), x_{max}(v)]$ by:

$$M \triangleq \frac{x_{max}(v) + x_{min}(v)}{2} \tag{8.120}$$

We thus have that:

$$x_{min}(v) \leq f((X_u)_{u \in \mathcal{V}}) \leq x_{max}(v) \Rightarrow |f((X_u)_{u \in \mathcal{V}}) - M| \leq D \tag{8.121}$$

Piecing together Equations 8.116, 8.118, and 8.121, we have by monotonicity:

$$P(n_v((\mathcal{H}_{tr})_{gt}) \leq n_v((\mathcal{H}_{te})_{gt})) \leq P(x_{min}(v) \leq f((X_u)_{u \in \mathcal{V}}) \leq x_{max}(v)) \tag{8.122a}$$

$$\leq P(|f((X_u)_{u \in \mathcal{V}}) - M| < D) \tag{8.122b}$$

$$\leq P(|M - f((\tilde{X}_u)_{u \in \mathcal{V}})| < D) \leq O(\frac{1}{\sqrt{n}}), \forall v \in \mathcal{U} \subseteq \mathcal{V} \tag{8.122c}$$

Where the last inequality comes from the anti-concentration bound of Theorem 1.2 of [339], which states:

$$P(|f((\tilde{X}_v)_{v \in \mathcal{V}}) - x| < s) \leq O(\frac{1}{\sqrt{n}}), \forall x \in \mathbb{R} \tag{8.123}$$

for any $s > 0$ which may depend on $n$ and any $x \in \mathbb{R}$. Setting $x := M, s := D$, gives the last inequality.

□

## 8.11 Additional Experiments

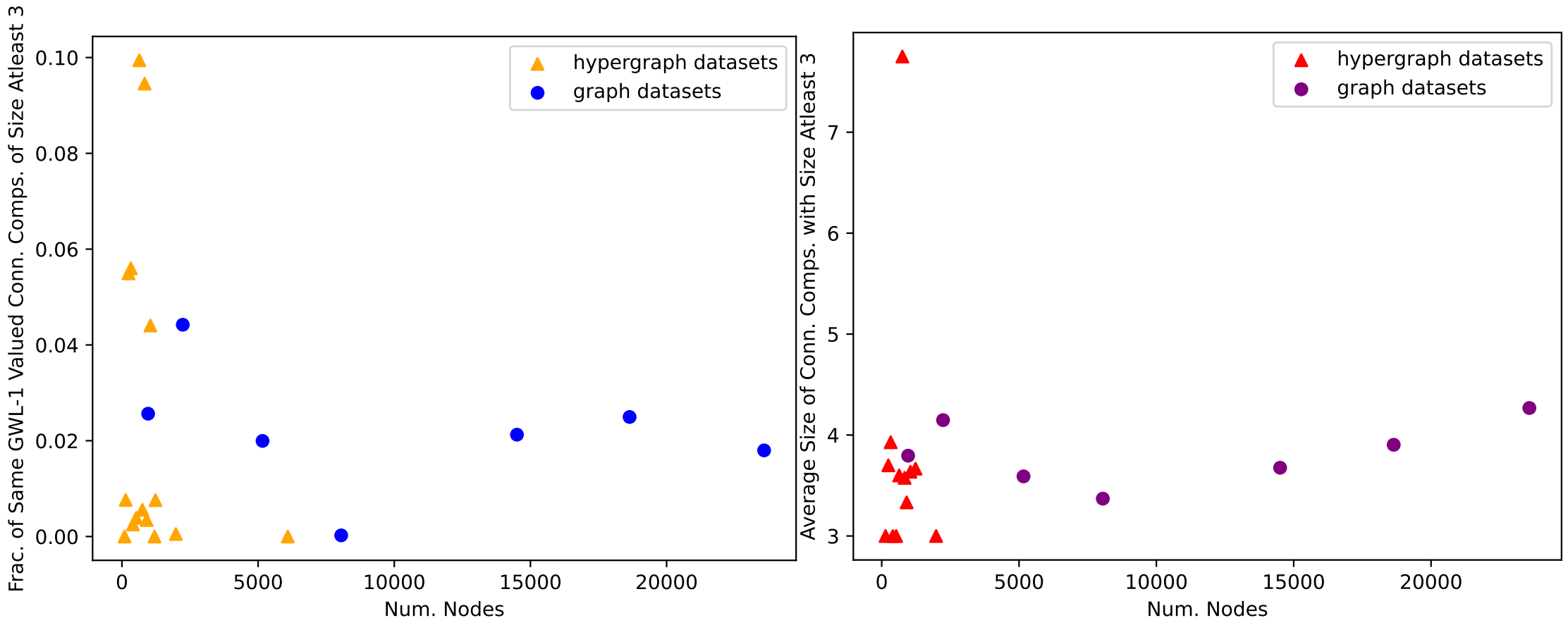

(a) Fraction of components of size atleast 3 selected by Algorithm 29.

(b) Average size of components of size atleast 3 from Algorithm 29.

**Figure 8.6.** Correlation between components and number of nodes.

As we are primarily concerned with symmetries in a hypergraph, we empirically measure the size and frequency of the components found by the Algorithm for real-world datasets. For the real-world datasets listed in Appendix 8.12, in Figure 8.6a, we plot the fraction of connected components of the same $L$-GWL-1 value ($L = 2$) that are atleast 3 in cardinality from Algorithm 29 as a function of the number of nodes of the hypergraph. For these datasets, it is much more common for the connected components to be of sizes 1 and 2. On the right, in Figure 8.6b we show the distribution of the sizes of the connected components found by Algorithm 29. We see that, on average, the connected components are at least an order of magnitude smaller compared to the total number of nodes. Common to both plots, the graph datasets appear to have more nodes and a consistent fraction and size of components, while the hypergraph datasets have higher variance in the fraction of components, which is expected since there are more possibilities for the connections in a hypergraph.

We show below the critical difference diagrams [340] of the average PR-AUC score ranks (percentiles) for two datasets from Table 8.1 across all the downstream hyperGNNs. Each plot contains the PR-AUC ranks of the three compared approaches: baseline, 50% hyperedge drop, and Our method. The bars in each plot represents statistical insignificance between the two approaches according to 4 runs of each experiment.

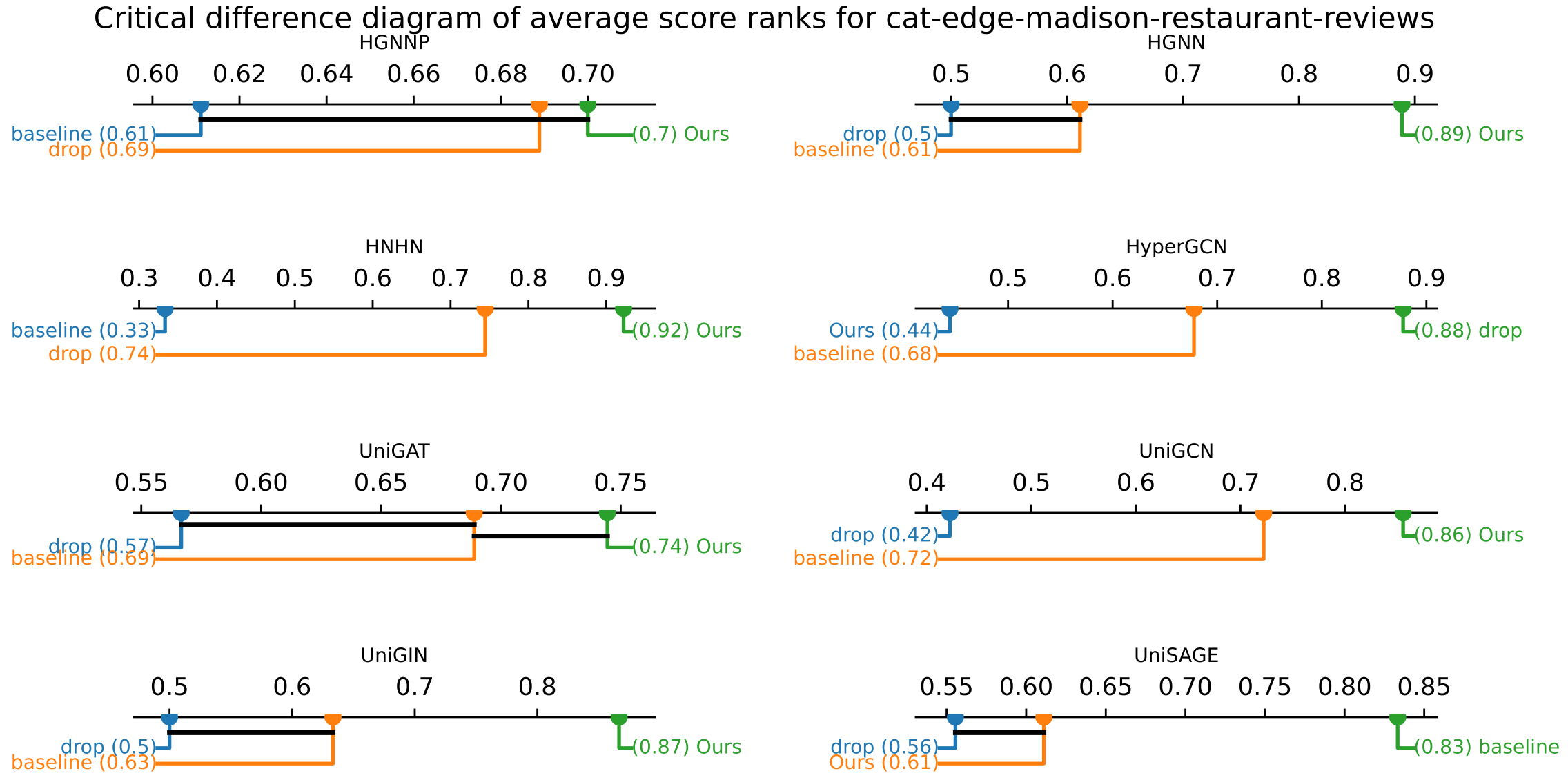

**Figure 8.7.** Critical difference diagrams of CAT-EDGE-MADISON-RESTAURANT-REVIEWS

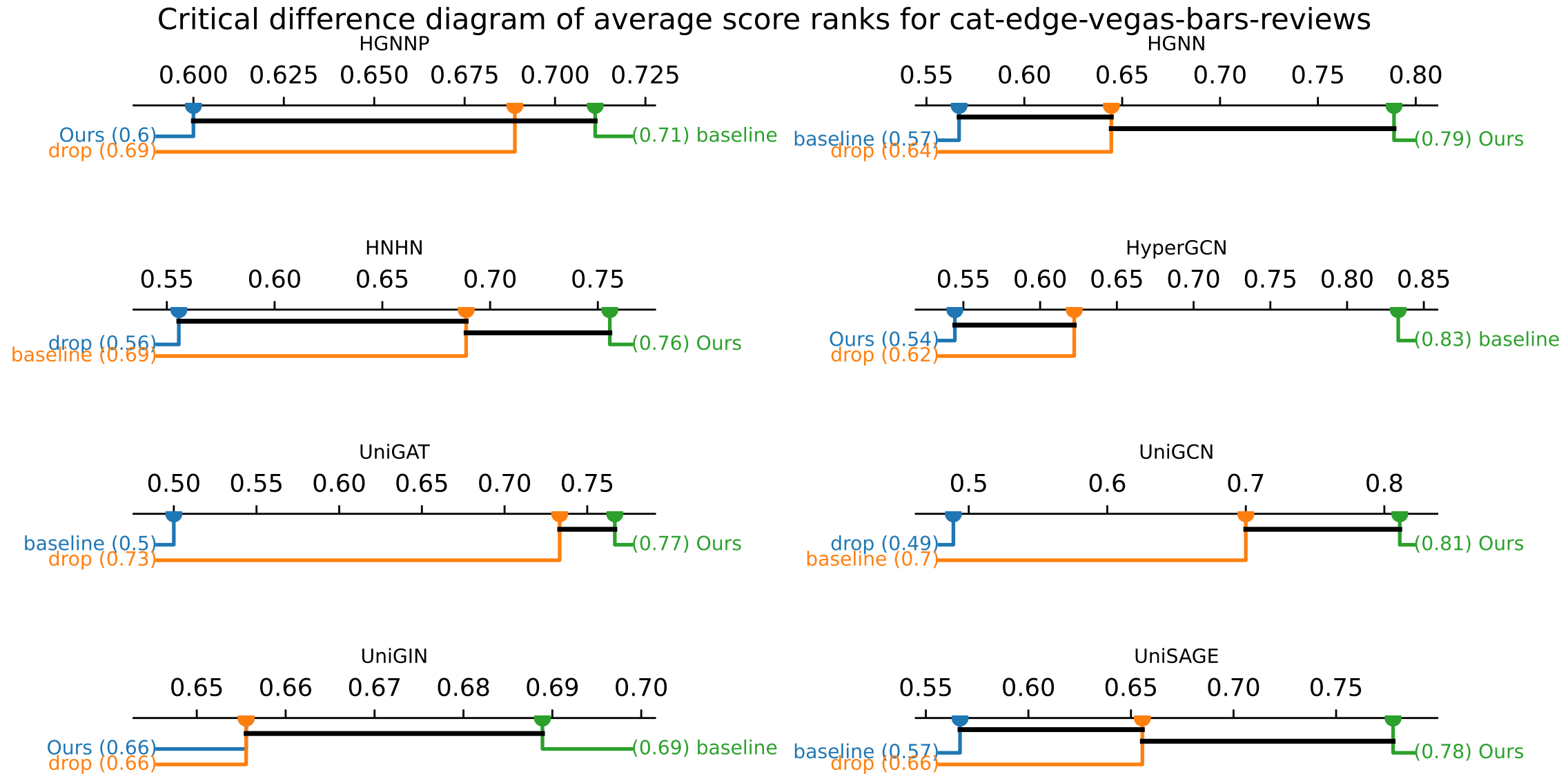

**Figure 8.8.** Critical difference diagrams of CAT-EDGE-VEGAS-BARS-REVIEWS

### 8.11.1 Additional Experiments on Hypergraphs

In Table 8.3, we show the PR-AUC scores for four additional hypergraph datasets, cat-edge-Brain, cat-edge-vegas-bar-reviews, WikiPeople-0bi, and JF17K for predicting size 3 hyperedges.

**Table 8.3.** PR-AUC on four other hypergraph datasets. The top average scores for each hyperGNN method, or row, is colored. Red scores denote the top scores in a row. Orange scores denote a two way tie and brown scores denote a threeway tie.

| PR-AUC ↑ | Baseline | Ours | Baseln.+edrop |
|---|---|---|---|
| HGNN | 0.75 ± 0.01 | 0.79 ± 0.11 | 0.74 ± 0.09 |
| HGNNP | 0.75 ± 0.05 | 0.78 ± 0.10 | 0.74 ± 0.12 |
| HNHN | 0.74 ± 0.04 | 0.74 ± 0.02 | 0.74 ± 0.05 |
| HyperGCN | 0.74 ± 0.09 | 0.50 ± 0.07 | 0.50 ± 0.12 |
| UniGAT | 0.73 ± 0.07 | 0.81 ± 0.10 | 0.81 ± 0.09 |
| UniGCN | 0.78 ± 0.04 | 0.81 ± 0.09 | 0.71 ± 0.08 |
| UniGIN | 0.74 ± 0.09 | 0.74 ± 0.03 | 0.74 ± 0.07 |
| UniSAGE | 0.74 ± 0.03 | 0.74 ± 0.12 | 0.74 ± 0.01 |

(a) cat-edge-Brain

| PR-AUC ↑ | Baseline | Ours | Baseln.+edrop |
|---|---|---|---|
| HGNN | 0.95 ± 0.10 | 0.99 ± 0.04 | 0.96 ± 0.09 |
| HGNNP | 0.95 ± 0.06 | 0.96 ± 0.09 | 0.96 ± 0.08 |
| HNHN | 1.00 ± 0.08 | 0.99 ± 0.09 | 0.95 ± 0.10 |
| HyperGCN | 0.76 ± 0.03 | 0.67 ± 0.14 | 0.68 ± 0.09 |
| UniGAT | 0.87 ± 0.07 | 1.00 ± 0.09 | 0.99 ± 0.08 |
| UniGCN | 0.99 ± 0.07 | 0.96 ± 0.09 | 0.92 ± 0.05 |
| UniGIN | 0.98 ± 0.06 | 0.96 ± 0.08 | 0.95 ± 0.06 |
| UniSAGE | 0.94 ± 0.05 | 0.98 ± 0.07 | 0.97 ± 0.07 |

(b) cat-edge-vegas-bar-reviews

| PR-AUC ↑ | Baseline | Ours | Baseln.+edrop |
|---|---|---|---|
| HGNN | 0.52 ± 0.01 | 0.57 ± 0.08 | 0.54 ± 0.10 |
| HGNNP | 0.52 ± 0.03 | 0.54 ± 0.07 | 0.54 ± 0.06 |
| HNHN | 0.73 ± 0.03 | 0.73 ± 0.07 | 0.73 ± 0.00 |
| HyperGCN | 0.54 ± 0.05 | 0.55 ± 0.02 | 0.49 ± 0.10 |
| UniGAT | 0.49 ± 0.09 | 0.54 ± 0.04 | 0.53 ± 0.04 |
| UniGCN | 0.46 ± 0.08 | 0.68 ± 0.08 | 0.51 ± 0.08 |
| UniGIN | 0.73 ± 0.09 | 0.73 ± 0.01 | 0.73 ± 0.02 |
| UniSAGE | 0.73 ± 0.06 | 0.73 ± 0.02 | 0.73 ± 0.08 |

(c) WikiPeople-0bi

| PR-AUC ↑ | Baseline | Ours | Baseln.+edrop |
|---|---|---|---|
| HGNN | 0.59 ± 0.04 | 0.63 ± 0.04 | 0.42 ± 0.09 |
| HGNNP | 0.71 ± 0.07 | 0.63 ± 0.07 | 0.57 ± 0.04 |
| HNHN | 0.73 ± 0.04 | 0.73 ± 0.03 | 0.73 ± 0.04 |
| HyperGCN | 0.59 ± 0.05 | 0.58 ± 0.09 | 0.48 ± 0.01 |
| UniGAT | 0.61 ± 0.07 | 0.61 ± 0.04 | 0.51 ± 0.08 |
| UniGCN | 0.58 ± 0.00 | 0.60 ± 0.03 | 0.59 ± 0.02 |
| UniGIN | 0.80 ± 0.04 | 0.77 ± 0.08 | 0.75 ± 0.05 |
| UniSAGE | 0.79 ± 0.02 | 0.77 ± 0.08 | 0.74 ± 0.01 |

(d) JF17K

### 8.11.2 Experiments on Graph Data

We show in Tables 8.4, 8.5, 8.6 the PR-AUC test scores for link prediction on some nonattributed graph datasets. The train-val-test splits are predefined for FB15K-237 and for the other graph datasets a single graph is deterministically split into 80/5/15 for train/-val/test. We remove 10% of the edges in training and let them be positive examples $P_{tr}$ to predict. For validation and test, we remove 50% of the edges from both validation and test to set as the positive examples $P_{val}, P_{te}$ to predict. For train, validation, and test, we sample $1.2|P_{tr}|, 1.2|P_{val}|, 1.2|P_{te}|$ negative link samples from the links of train, validation and test. Along with hyperGNN architectures we use for the hypergraph experiments, we also compare with standard GNN architectures: APPNP [310], GAT [311], GCN2 [312], GCN [218], GIN [215], and GraphSAGE [217]. For every hyperGNN/GNN architecture, we also apply drop-edge [313] to the input graph and use this also as baseline. The number of layers of each GNN is set to 5 and the hidden dimension at 1024. For APPNP and GCN2, one MLP is used on the initial node positional encodings. Since graphs do not have any hyperedges beyond size 2, graph neural networks fit the inductive bias of the graph data more easily and thus may perform better than hypergraph neural network baselines more often than expected.

**Table 8.4.** PR-AUC on graph dataset JOHNSHOPKINS55. Each column is a comparison of the baseline PR-AUC scores against the PR-AUC score for our method (first row) applied to a standard hyperGNN architecture. Red color denotes the highest average score in the column. Orange color denotes a two-way tie in the column, and brown color denotes a three-or-more-way tie in the column.

| PR-AUC ↑ | HGNN | HGNNP | HNHN | HyperGCN | UniGAT | UniGCN | UniGIN | UniSAGE |
|---|---|---|---|---|---|---|---|---|
| Ours | 0.71 ± 0.04 | 0.71 ± 0.09 | 0.69 ± 0.09 | 0.75 ± 0.14 | 0.75 ± 0.09 | 0.74 ± 0.09 | 0.65 ± 0.08 | 0.65 ± 0.07 |
| hyperGNN Baseline | 0.68 ± 0.00 | 0.69 ± 0.06 | 0.67 ± 0.02 | 0.75 ± 0.04 | 0.74 ± 0.02 | 0.74 ± 0.00 | 0.65 ± 0.05 | 0.64 ± 0.08 |
| hyperGNN Baseln.+edrop | 0.67 ± 0.02 | 0.70 ± 0.07 | 0.66 ± 0.00 | 0.75 ± 0.03 | 0.73 ± 0.08 | 0.74 ± 0.05 | 0.63 ± 0.01 | 0.64 ± 0.03 |
| APPNP | 0.40 ± 0.03 | 0.40 ± 0.03 | 0.40 ± 0.03 | 0.40 ± 0.03 | 0.40 ± 0.03 | 0.40 ± 0.03 | 0.40 ± 0.03 | 0.40 ± 0.03 |
| APPNP+edrop | 0.40 ± 0.13 | 0.40 ± 0.13 | 0.40 ± 0.13 | 0.40 ± 0.13 | 0.40 ± 0.13 | 0.40 ± 0.13 | 0.40 ± 0.13 | 0.40 ± 0.13 |
| GAT | 0.49 ± 0.03 | 0.49 ± 0.03 | 0.49 ± 0.03 | 0.49 ± 0.03 | 0.49 ± 0.03 | 0.49 ± 0.03 | 0.49 ± 0.03 | 0.49 ± 0.03 |
| GAT+edrop | 0.51 ± 0.05 | 0.51 ± 0.05 | 0.51 ± 0.05 | 0.51 ± 0.05 | 0.51 ± 0.05 | 0.51 ± 0.05 | 0.51 ± 0.05 | 0.51 ± 0.05 |
| GCN2 | 0.50 ± 0.09 | 0.50 ± 0.09 | 0.50 ± 0.09 | 0.50 ± 0.09 | 0.50 ± 0.09 | 0.50 ± 0.09 | 0.50 ± 0.09 | 0.50 ± 0.09 |
| GCN2+edrop | 0.56 ± 0.07 | 0.56 ± 0.07 | 0.56 ± 0.07 | 0.56 ± 0.07 | 0.56 ± 0.07 | 0.56 ± 0.07 | 0.56 ± 0.07 | 0.56 ± 0.07 |
| GCN | 0.73 ± 0.02 | 0.73 ± 0.02 | 0.73 ± 0.02 | 0.73 ± 0.02 | 0.73 ± 0.02 | 0.73 ± 0.02 | 0.73 ± 0.02 | 0.73 ± 0.02 |
| GCN+edrop | 0.73 ± 0.01 | 0.73 ± 0.01 | 0.73 ± 0.01 | 0.73 ± 0.01 | 0.73 ± 0.01 | 0.73 ± 0.01 | 0.73 ± 0.01 | 0.73 ± 0.01 |
| GIN | 0.73 ± 0.06 | 0.73 ± 0.06 | 0.73 ± 0.06 | 0.73 ± 0.06 | 0.73 ± 0.06 | 0.73 ± 0.06 | 0.73 ± 0.06 | 0.73 ± 0.06 |
| GIN+edrop | 0.73 ± 0.01 | 0.73 ± 0.01 | 0.73 ± 0.01 | 0.73 ± 0.01 | 0.73 ± 0.01 | 0.73 ± 0.01 | 0.73 ± 0.01 | 0.73 ± 0.01 |
| GraphSAGE | 0.73 ± 0.08 | 0.73 ± 0.08 | 0.73 ± 0.08 | 0.73 ± 0.08 | 0.73 ± 0.08 | 0.73 ± 0.08 | 0.73 ± 0.08 | 0.73 ± 0.08 |
| GraphSAGE+edrop | 0.73 ± 0.09 | 0.73 ± 0.09 | 0.73 ± 0.09 | 0.73 ± 0.09 | 0.73 ± 0.09 | 0.73 ± 0.09 | 0.73 ± 0.09 | 0.73 ± 0.09 |

**Table 8.5.** PR-AUC on graph dataset FB15K-237. Each column is a comparison of the baseline PR-AUC scores against the PR-AUC score for our method (first row) applied to a standard hyperGNN architecture. Red color denotes the highest average score in the column. Orange color denotes a two-way tie in the column, and brown color denotes a three-or-more-way tie in the column.

| PR-AUC ↑ | HGNN | HGNNP | HNHN | HyperGCN | UniGAT | UniGCN | UniGIN | UniSAGE |
|---|---|---|---|---|---|---|---|---|
| Ours | 0.66 ± 0.06 | 0.78 ± 0.02 | 0.63 ± 0.07 | 0.82 ± 0.10 | 0.75 ± 0.05 | 0.74 ± 0.03 | 0.75 ± 0.03 | 0.75 ± 0.06 |
| hyperGNN Baseline | 0.65 ± 0.06 | 0.65 ± 0.06 | 0.65 ± 0.04 | 0.82 ± 0.09 | 0.74 ± 0.04 | 0.74 ± 0.05 | 0.75 ± 0.03 | 0.77 ± 0.01 |
| hyperGNN Baseln.+edrop | 0.65 ± 0.09 | 0.65 ± 0.00 | 0.64 ± 0.05 | 0.82 ± 0.00 | 0.72 ± 0.00 | 0.74 ± 0.07 | 0.73 ± 0.03 | 0.72 ± 0.07 |
| APPNP | 0.72 ± 0.10 | 0.72 ± 0.10 | 0.72 ± 0.10 | 0.72 ± 0.10 | 0.72 ± 0.10 | 0.72 ± 0.10 | 0.72 ± 0.10 | 0.72 ± 0.10 |
| APPNP+edrop | 0.71 ± 0.05 | 0.71 ± 0.05 | 0.71 ± 0.05 | 0.71 ± 0.05 | 0.71 ± 0.05 | 0.71 ± 0.05 | 0.71 ± 0.05 | 0.71 ± 0.05 |
| GAT | 0.64 ± 0.06 | 0.64 ± 0.06 | 0.64 ± 0.06 | 0.64 ± 0.06 | 0.64 ± 0.06 | 0.64 ± 0.06 | 0.64 ± 0.06 | 0.64 ± 0.06 |
| GAT+edrop | 0.61 ± 0.09 | 0.61 ± 0.09 | 0.61 ± 0.09 | 0.61 ± 0.09 | 0.61 ± 0.09 | 0.61 ± 0.09 | 0.61 ± 0.09 | 0.61 ± 0.09 |
| GCN2 | 0.66 ± 0.03 | 0.66 ± 0.03 | 0.66 ± 0.03 | 0.66 ± 0.03 | 0.66 ± 0.03 | 0.66 ± 0.03 | 0.66 ± 0.03 | 0.66 ± 0.03 |
| GCN2+edrop | 0.65 ± 0.10 | 0.65 ± 0.10 | 0.65 ± 0.10 | 0.65 ± 0.10 | 0.65 ± 0.10 | 0.65 ± 0.10 | 0.65 ± 0.10 | 0.65 ± 0.10 |
| GCN | 0.69 ± 0.03 | 0.69 ± 0.03 | 0.69 ± 0.03 | 0.69 ± 0.03 | 0.69 ± 0.03 | 0.69 ± 0.03 | 0.69 ± 0.03 | 0.69 ± 0.03 |
| GCN+edrop | 0.71 ± 0.06 | 0.71 ± 0.06 | 0.71 ± 0.06 | 0.71 ± 0.06 | 0.71 ± 0.06 | 0.71 ± 0.06 | 0.71 ± 0.06 | 0.71 ± 0.06 |
| GIN | 0.73 ± 0.03 | 0.73 ± 0.03 | 0.73 ± 0.03 | 0.73 ± 0.03 | 0.73 ± 0.03 | 0.73 ± 0.03 | 0.73 ± 0.03 | 0.73 ± 0.03 |
| GIN+edrop | 0.56 ± 0.07 | 0.56 ± 0.07 | 0.56 ± 0.07 | 0.56 ± 0.07 | 0.56 ± 0.07 | 0.56 ± 0.07 | 0.56 ± 0.07 | 0.56 ± 0.07 |
| GraphSAGE | 0.46 ± 0.15 | 0.46 ± 0.15 | 0.46 ± 0.15 | 0.46 ± 0.15 | 0.46 ± 0.15 | 0.46 ± 0.15 | 0.46 ± 0.15 | 0.46 ± 0.15 |
| GraphSAGE+edrop | 0.47 ± 0.01 | 0.47 ± 0.01 | 0.47 ± 0.01 | 0.47 ± 0.01 | 0.47 ± 0.01 | 0.47 ± 0.01 | 0.47 ± 0.01 | 0.47 ± 0.01 |

**Table 8.6.** PR-AUC on graph dataset AIFB. Each column is a comparison of the baseline PR-AUC scores against the PR-AUC score for our method (first row) applied to a standard hyperGNN architecture. Red color denotes the highest average score in the column. Orange color denotes a two-way tie in the column, and brown color denotes a three-or-more-way tie in the column.

| PR-AUC ↑ | HGNN | HGNNP | HNHN | HyperGCN | UniGAT | UniGCN | UniGIN | UniSAGE |
|---|---|---|---|---|---|---|---|---|
| Ours | 0.79 ± 0.11 | 0.73 ± 0.10 | 0.73 ± 0.02 | 0.85 ± 0.07 | 0.75 ± 0.10 | 0.84 ± 0.09 | 0.72 ± 0.03 | 0.72 ± 0.12 |
| hyperGNN Baseline | 0.72 ± 0.07 | 0.72 ± 0.07 | 0.72 ± 0.06 | 0.85 ± 0.05 | 0.75 ± 0.09 | 0.84 ± 0.05 | 0.72 ± 0.07 | 0.72 ± 0.06 |
| hyperGNN Baseln.+edrop | 0.72 ± 0.05 | 0.72 ± 0.08 | 0.72 ± 0.06 | 0.85 ± 0.07 | 0.73 ± 0.09 | 0.84 ± 0.06 | 0.72 ± 0.03 | 0.72 ± 0.07 |
| APPNP | 0.81 ± 0.12 | 0.81 ± 0.12 | 0.81 ± 0.12 | 0.81 ± 0.12 | 0.81 ± 0.12 | 0.81 ± 0.12 | 0.81 ± 0.12 | 0.81 ± 0.12 |
| APPNP+edrop | 0.80 ± 0.05 | 0.80 ± 0.05 | 0.80 ± 0.05 | 0.80 ± 0.05 | 0.80 ± 0.05 | 0.80 ± 0.05 | 0.80 ± 0.05 | 0.80 ± 0.05 |
| GAT | 0.50 ± 0.02 | 0.50 ± 0.02 | 0.50 ± 0.02 | 0.50 ± 0.02 | 0.50 ± 0.02 | 0.50 ± 0.02 | 0.50 ± 0.02 | 0.50 ± 0.02 |
| GAT+edrop | 0.33 ± 0.02 | 0.33 ± 0.02 | 0.33 ± 0.02 | 0.33 ± 0.02 | 0.33 ± 0.02 | 0.33 ± 0.02 | 0.33 ± 0.02 | 0.33 ± 0.02 |
| GCN2 | 0.83 ± 0.05 | 0.83 ± 0.05 | 0.83 ± 0.05 | 0.83 ± 0.05 | 0.83 ± 0.05 | 0.83 ± 0.05 | 0.83 ± 0.05 | 0.83 ± 0.05 |
| GCN2+edrop | 0.78 ± 0.04 | 0.78 ± 0.04 | 0.78 ± 0.04 | 0.78 ± 0.04 | 0.78 ± 0.04 | 0.78 ± 0.04 | 0.78 ± 0.04 | 0.78 ± 0.04 |
| GCN | 0.73 ± 0.14 | 0.73 ± 0.14 | 0.73 ± 0.14 | 0.73 ± 0.14 | 0.73 ± 0.14 | 0.73 ± 0.14 | 0.73 ± 0.14 | 0.73 ± 0.14 |
| GCN+edrop | 0.75 ± 0.08 | 0.75 ± 0.08 | 0.75 ± 0.08 | 0.75 ± 0.08 | 0.75 ± 0.08 | 0.75 ± 0.08 | 0.75 ± 0.08 | 0.75 ± 0.08 |
| GIN | 0.73 ± 0.00 | 0.73 ± 0.00 | 0.73 ± 0.00 | 0.73 ± 0.00 | 0.73 ± 0.00 | 0.73 ± 0.00 | 0.73 ± 0.00 | 0.73 ± 0.00 |
| GIN+edrop | 0.73 ± 0.10 | 0.73 ± 0.10 | 0.73 ± 0.10 | 0.73 ± 0.10 | 0.73 ± 0.10 | 0.73 ± 0.10 | 0.73 ± 0.10 | 0.73 ± 0.10 |
| GraphSAGE | 0.46 ± 0.15 | 0.46 ± 0.15 | 0.46 ± 0.15 | 0.46 ± 0.15 | 0.46 ± 0.15 | 0.46 ± 0.15 | 0.46 ± 0.15 | 0.46 ± 0.15 |
| GraphSAGE+edrop | 0.47 ± 0.01 | 0.47 ± 0.01 | 0.47 ± 0.01 | 0.47 ± 0.01 | 0.47 ± 0.01 | 0.47 ± 0.01 | 0.47 ± 0.01 | 0.47 ± 0.01 |

## 8.12 Dataset and Hyperparameters

Table 8.7 lists the datasets and hyperparameters used in our experiments. All datasets are originally from [341] or are general hypergraph datasets provided in [342, 343]. We list the total number of hyperedges $|\mathcal{E}|$, the total number of vertices $|\mathcal{V}|$, the positive to negative label ratios for train/val/test, and the percentage of the connected components searched over by our algorithm that are size atleast 3. A node isomorphism class is determined by our isomorphism testing algorithm. By Proposition 8.10.8 we can guarantee that if two nodes are in separate isomorphism classes by our isomorphism tester, then they are actually nonisomorphic.

We use 1024 dimensions for all hyperGNN/GNN layer latent spaces, 5 layers for all hypergraph/graph neural networks, and a common learning rate of 0.01. Exactly 2000 epochs are used for training.

The hyperGNN architecture baselines are described in the follwoing:

- HGNN [267] A neural network that generalizes the graph convolution to hypergraphs where there are hyperedge weights. Its architecture can be described by the following update step for the $l+1$-layer from the $l$th layer:

$$X^{(l+1)} = \sigma(D_v^{-\frac{1}{2}} H W D_{\mathrm{e}}^{-1} H^T D_v^{-\frac{1}{2}} X^{(l)} W^{(l)}) \tag{8.124}$$

  where $D_v \in \mathbb{R}^{n\times n}$ is the diagonal node degree matrix, $D_{\mathrm{e}} \in \mathbb{R}^{m\times m}$ is the diagonal hyperedge degree matrix, $H \in \mathbb{R}^{n\times m}$ is the star incidence matrix, $W$ is the diagonal hyperedge weight matrix, $X^{(l)} \in \mathbb{R}^{n\times d}$ is a node signal matrix, $W^{(l)} \in \mathbb{R}^{d\times d}$ is a weight matrix, and $\sigma$ is a nonlinear activation. Following the matrix products, as a message passing neural network, HGNN is GWL-1 based since the nodes pass to the hyperedges and back.

- HGNNP [275] is an improved version of HGNN where asymmetry is introduced into the message passing weightings to distinguish the vertices from the hyperedges. This is also a GWL-1 based message passing neural network. It is described by the following node signal update equation:

$$X^{(l+1)} = \sigma(D_v^{-1} H W D_e^{-1} H^T X^{(l)} W^{(l)}) \tag{8.125}$$

where the matrices are exactly the same as from HGNN.

- HyperGCN [277] computes GCN on a clique expansion of a hypergraph. This has an updateable adjacency matrix defined as follows:

$$A^{(l)}_{i,j} = \begin{cases} 1 & (i,j) \in E^{(l)} \\ 0 & (i,j) \notin E^{(l)} \end{cases} \tag{8.126}$$

where

$$E^{(l)} = \{(i_e, j_e) = argmax_{i,j \in e} |X^{(l)}_i - X^{(l)}_j| : e \in \mathcal{E}\} \tag{8.127}$$

$$X^{(l+1)}_v = \sigma(\sum_{u \in N(v)} ([A^{(l)}]_{v,u} X^{(l)}_u W^{(l)})) \tag{8.128}$$

The $X^{(l)} \in \mathbb{R}^{n \times d}$ is the node signal matrix at layer $l$, the $W^{(l)} \in \mathbb{R}^{d \times d}$ is the weight matrix at layer $l$, and $\sigma$ is some nonlinear activation. This architecture has less expressive power than GWL-1.

- HNHN [276] This is like HGNN but where the message passing is explicitly broken up into two hyperedge to node and node to hyperedge layers.

$$X^{(l)}_E = \sigma(H^T X^{(l)}_V W^{(l)}_E + b^{(l)}_E) \tag{8.129a}$$

and

$$X^{(l+1)}_V = \sigma(H X^{(l)}_E W^{(l)}_V + b^{(l)}_V) \tag{8.129b}$$

where $H \in \mathbb{R}^{n \times m}$ is the star expansion incidence matrix, $W^{(l)}_E, W^{(l)}_V \in \mathbb{R}^{d \times d}, b^{(l)}_E \in \mathbb{R}^m, b^{(l)}_V \in \mathbb{R}^n$ are weights and biases, $X^{(l)}_E, X^{(l)}_V$ are the hyperedge and node signal matrices at layer $l$, and $\sigma$ is a nonlinear activation function. The bias vectors prevent HNHN from being permutation equivariant.

- UniGNN [269] The idea is directly related to generalizing WL-1 GNNs to Hypergraphs. Define the following hyperedge representation for hyperedge $\mathrm{e} \in \mathcal{E}$:

$$h_{\mathrm{e}}^{(l)} = \frac{1}{|\mathrm{e}|} \sum_{u \in \mathrm{e}} X_u^{(l)} \tag{8.130}$$

  - UniGCN: a generalization of GCN to hypergraphs

$$X_v^{(l)} = \frac{1}{\sqrt{d_v}} \sum_{\mathrm{e} \ni v} \frac{1}{\sqrt{d_{\mathrm{e}}}} W^{(l)} h_{\mathrm{e}}^{(l)} \tag{8.131}$$

  - UniGAT: a generalization of GAT to hypergraphs

$$\alpha_{u\mathrm{e}} = \sigma(a^T [X_{\mathrm{i}}^{(l)} W^{(l)}; X_{\mathrm{j}}^{(l)} W^{(l)}]) \tag{8.132a}$$

$$\tilde{\alpha}_{u\mathrm{e}} = \frac{\mathrm{e}^{\alpha_{ue}}}{\sum_{v \in \mathrm{e}} \mathrm{e}^{\alpha_{ve}}} \tag{8.132b}$$

$$X_v^{(l+1)} = \sum_{\mathrm{e} \ni v} \alpha_{v\mathrm{e}} h_{\mathrm{e}} W^{(l)} \tag{8.132c}$$

  - UniGIN: a generalization of GIN to hypergraphs

$$X_v^{(l+1)} = ((1 + \epsilon) X_v^{(l)} + \sum_{\mathrm{e} \ni v} h_{\mathrm{e}}) \tag{8.133}$$

  - UniSAGE: a generalization of GraphSAGE to hypergraphs

$$X_v^{(l+1)} = (X_v^{(l)} + \sum_{\mathrm{e} \ni v} (h_{\mathrm{e}})) \tag{8.134}$$

All positional encodings are computed from the training hyperedges before data augmentation. The loss we use for higher order link prediction is the Binary Cross Entropy Loss for all the positive and negatives samples. Hypergraph neural network implementations were mostly taken from https://github.com/iMoonLab/DeepHypergraph, which uses the Apache License 2.0.

**Table 8.7.** Dataset statistics and training hyperparameters used for all datasets in scoring all experiments.

| Dataset Information | | | | | | |
|---|---|---|---|---|---|---|
| Dataset | $\lvert\mathcal{E}\rvert$ | $\lvert\mathcal{V}\rvert$ | $\frac{\Delta_{+,tr}}{\Delta_{-,tr}}$ | $\frac{\Delta_{+,val}}{\Delta_{-,val}}$ | $\frac{\Delta_{+,te}}{\Delta_{-,te}}$ | % of Sel. Conn. Comps. |
| cat-edge-DAWN | 87,104 | 2,109 | 8,802/10,547 | 1,915/2,296 | 1,867/2,237 | 0.05% |
| email-Eu | 234,760 | 998 | 1,803/2,159 | 570/681 | 626/749 | 0.6% |
| contact-primary-school | 106,879 | 242 | 1,620/1,921 | 461/545 | 350/415 | 9.3% |
| cat-edge-music-blues-reviews | 694 | 1,106 | 16/19 | 7/6 | 3/3 | 0.14% |
| cat-edge-vegas-bars-reviews | 1,194 | 1,234 | 72/86 | 12/14 | 11/13 | 0.7% |
| contact-high-school | 7,818 | 327 | 2,646/3,143 | 176/208 | 175/205 | 5.6% |
| cat-edge-Brain | 21,180 | 638 | 13,037/13,817 | 2,793/3,135 | 2,794/3,020 | 9.9% |
| johnshopkins55 | 298,537 | 5,163 | 29,853/35,634 | 9,329/11,120 | 27,988/29,853 | 2.0% |
| AIFB | 46,468 | 8,083 | 4,646/5,575 | 1,452/1,739 | 4,356/5,222 | 0.02% |
| amherst41 | 145,526 | 2,234 | 14,552/17,211 | 4,547/5,379 | 16,125/13,643 | 4.4% |
| FB15k-237 | 272,115 | 14,505 | 27,211/32,630 | 8,767/10,509 | 10,233/12,271 | 2.1% |
| WikiPeople-0bi | 18,828 | 43,388 | 27,211/32,630 | 10,254/12,301 | 1,164/1,396 | 0.05% |
| JF17K | 76,379 | 28,645 | 11,907/14,287 | 1,341/1,608 | 1,341/1,608 | 0.6% |

We describe here some more information about each dataset we use in our experiments as provided by [341]: Here is some information about the hypergraph datasets:

- [343] CAT-EDGE-DAWN: Here nodes are drugs, hyperedges are combinations of drugs taken by a patient prior to an emergency room visit and edge categories indicate the patient disposition (e.g., "sent home", "surgery", "released to detox").
- [344–346] EMAIL-EU: This is a temporal higher-order network dataset, which here means a sequence of timestamped simplices, or hyperedges with all its node subsets existing as hyperedges, where each simplex is a set of nodes. In email communication, messages can be sent to multiple recipients. In this dataset, nodes are email addresses at a European research institution. The original data source only contains (sender, receiver, timestamp) tuples, where timestamps are recorded at 1-second resolution. Simplices consist of a sender and all receivers such that the email between the two has the same timestamp. We restricted to simplices that consist of at most 25 nodes.
- [347] CONTACT-PRIMARY-SCHOOL: This is a temporal higher-order network dataset, which here means a sequence of timestamped simplices where each simplex is a set of nodes. The dataset is constructed from interactions recorded by wearable sensors by people at a primary school. The sensors record interactions at a resolution of 20 seconds (recording all interactions from the previous 20 seconds). Nodes are the people and simplices are maximal cliques of interacting individuals from an interval.
- [348]CAT-EDGE-VEGAS-BARS-REVIEWS: Hypergraph where nodes are Yelp users and hyperedges are users who reviewed an establishment of a particular category (different types of bars in Las Vegas, NV) within a month timeframe.
- [344, 349] CONTACT-HIGH-SCHOOL: This is a temporal higher-order network dataset, which here means a sequence of timestamped simplices where each simplex is a set of nodes. The dataset is constructed from interactions recorded by wearable sensors by people at a high school. The sensors record interactions at a resolution of 20 seconds (recording all interactions from the previous 20 seconds). Nodes are the people and simplices are maximal cliques of interacting individuals from an interval.

- [350] CAT-EDGE-BRAIN: This is a graph whose edges have categorical edge labels. Nodes represent brain regions from an MRI scan. There are two edge categories: one for connecting regions with high fMRI correlation and one for connecting regions with similar activation patterns.

- [351]JOHNSHOPKINS55: Non-homophilous graph datasets from the facebook100 dataset.

- [352]AIFB: The AIFB dataset describes the AIFB research institute in terms of its staff, research groups, and publications. The dataset was first used to predict the affiliation (i.e., research group) for people in the dataset. The dataset contains 178 members of five research groups, however, the smallest group contains only four people, which is removed from the dataset, leaving four classes.

- [351]AMHERST41: Non-homophilous graph datasets from the facebook100 dataset.

- [353]FB15K-237: A subset of entities that are also present in the Wikilinks database [354] and that also have at least 100 mentions in Freebase (for both entities and relationships). Relationships like !/people/person/nationality which just reverses the head and tail compared to the relationship /people/person/nationality are removed. This resulted in 592,213 triplets with 14,951 entities and 1,345 relationships which were randomly split.

- [355]WIKIPEOPLE-0BI: The Wikidata dump was downloaded and the facts concerning entities of type human were extracted. These facts are denoised. Subsequently, the subsets of elements which have at least 30 mentions were selected. And the facts related to these elements were kept. Further, each fact was parsed into a set of its role-value pairs. The remaining facts were randomly split into training set, validation set and test set by a percentage of 80%:10%:10%. All binary relations are removed for simplicity. This modifies WikiPeople to WikiPeople-0bi.

- [356]JF17K: The full Freebase data in RDF format was downloaded. Entities involved in very few triples and the triples involving String, Enumeration Type and Numbers were removed. A fact representation was recovered from the remaining triples. Facts from meta-relations having only a single role were removed. From each meta-relation containing more than 10,000 facts, 10,000 facts were randomly selected.

### 8.12.1 Timings

We perform experiments on a cluster of machines equipped with AMD MI100s GPUs and 112 shared AMD EPYC 7453 28-Core Processors with 2.6 PB shared RAM. We show here the times for computing each method. The timings may vary heavily for different machines as the memory we used is shared and during peak usage there is a lot of paging. We notice that although our data preprocessing algorithm involves seemingly costly steps such as GWL-1, connected connected components etc. The complexity of the entire preprocessing algorithm is linear in the size of the input as shown in Proposition 8.10.41. Thus these operations are actually very efficient in practice as shown by Tables 8.12 and 8.14 for the hypergraph and graph datasets respectively. The preprocessing algorithm is run on CPU while the training is run on GPU for 2000 epochs.

**Table 8.8.** Timings for our method broken up into the preprocessing phase and training phases (2000 epochs) for the hypergraph datasets.

| Timings (hh:mm) ± (s) | | |
|---|---|---|
| Method | Preprocessing Time | Training Time |
| HGNN | 2m:45s±108s | 35m:9s±13s |
| HGNNP | 1m:52s±0s | 35m:16s±0s |
| HNHN | 1m:55s±0s | 35m:0s±1s |
| HyperGCN | 1m:50s±0s | 58m:17s±79s |
| UniGAT | 1m:54s±0s | 1h:19m:34s±0s |
| UniGCN | 1m:50s±2s | 35m:19s±2s |
| UniGIN | 1m:50s±1s | 35m:12s±1288s |
| UniSAGE | 1m:51s±0s | 35m:16s±0s |

(a) CAT-EDGE-DAWN

| Timings (hh:mm) ± (s) | | |
|---|---|---|
| Method | Preprocessing Time | Training Time |
| HGNN | 1.72s±5s | 2m:11s±11s |
| HGNNP | 1.42s±0s | 2m:10s±0s |
| HNHN | 1.99s±0s | 3m:43s±2s |
| HyperGCN | 1.47s±2s | 4m:12s±3s |
| UniGAT | 1.85s±0s | 3m:54s±287s |
| UniGCN | 2.93s±0s | 3m:15s±19s |
| UniGIN | 2.24s±0s | 3m:17s±18s |
| UniSAGE | 2.04s±0s | 3m:13s±3s |

(b) CAT-EDGE-MUSIC-BLUES-REVIEWS

**Table 8.9.** Timings for our method broken up into the preprocessing phase and training phases (2000 epochs) for the hypergraph datasets.

| Timings (hh:mm) ± (s) | | |
|---|---|---|
| Method | Preprocessing Time | Training Time |
| HGNN | 4.17s±0s | 2m:34s±1954s |
| HGNNP | 4.54s±0s | 2m:41s±53s |
| HNHN | 3.06s±0s | 2m:27s±15s |
| HyperGCN | 1.81s±1s | 2m:27s±0s |
| UniGAT | 1.91s±0s | 2m:27s±306s |
| UniGCN | 2.84s±0s | 2m:30s±72s |
| UniGIN | 3.20s±0s | 2m:27s±1189s |
| UniSAGE | 1.65s±0s | 2m:27s±0s |

(a) CAT-EDGE-VEGAS-BARS-REVIEWS

| Timings (hh:mm) ± (s) | | |
|---|---|---|
| Method | Preprocessing Time | Training Time |
| HGNN | 5.84s±1s | 6m:49s±8s |
| HGNNP | 5.82s±0s | 9m:8s±19s |
| HNHN | 5.95s±0s | 8m:21s±19s |
| HyperGCN | 5.74s±0s | 10m:16s±1s |
| UniGAT | 8.80s±0s | 2m:31s±282s |
| UniGCN | 6.35s±0s | 6m:9s±957s |
| UniGIN | 5.99s±0s | 10m:41s±43s |
| UniSAGE | 5.97s±0s | 9m:50s±0s |

(b) CONTACT-PRIMARY-SCHOOL

**Table 8.10.** Timings for our method broken up into the preprocessing phase and training phases (2000 epochs) for the hypergraph datasets.

| Timings (hh:mm) ± (s) | | |
|---|---|---|
| Method | Preprocessing Time | Training Time |
| HGNN | 23.25s±1s | 25m:41s±17s |
| HGNNP | 23.25s±0s | 19m:52s±49s |
| HNHN | 24.27s±1s | 5m:12s±63s |
| HyperGCN | 24.00s±0s | 21m:16s±0s |
| UniGAT | 14.27s±0s | 5m:13s±243s |
| UniGCN | 25.44s±0s | 5m:51s±1019s |
| UniGIN | 13.71s±1s | 19m:10s±3972s |
| UniSAGE | 14.08s±2s | 36m:29s±5s |

(a) EMAIL-EU

| Timings (hh:mm) ± (s) | | |
|---|---|---|
| Method | Preprocessing Time | Training Time |
| HGNN | 4.89s±6s | 1m:27s±8s |
| HGNNP | 2.12s±0s | 2m:42s±30s |
| HNHN | 2.12s±0s | 2m:39s±42s |
| HyperGCN | 2.11s±0s | 40.11s±3s |
| UniGAT | 2.13s±0s | 3m:18s±8s |
| UniGCN | 2.11s±0s | 3m:21s±2s |
| UniGIN | 2.11s±0s | 2m:24s±70s |
| UniSAGE | 2.11s±0s | 2m:8s±49s |

(b) CAT-EDGE-MADISON-RESTAURANTS

**Table 8.11.** Timings for our method broken up into the preprocessing phase and training phases (2000 epochs) for the hypergraph datasets.

| Timings (hh:mm) ± (s) | | |
|---|---|---|
| Method | Preprocessing Time | Training Time |
| HGNN | 15.11s±4s | 4m:59s±1s |
| HGNNP | 12.72s±0s | 2m:29s±0s |
| HNHN | 12.17s±0s | 3m:6s±0s |
| HyperGCN | 12.47s±0s | 49.25s±0s |
| UniGAT | 12.74s±0s | 2m:1s±1s |
| UniGCN | 12.50s±0s | 2m:29s±3s |
| UniGIN | 12.57s±0s | 2m:16s±3s |
| UniSAGE | 12.67s±0s | 1m:50s±29s |

(a) CONTACT-HIGH-SCHOOL

| Timings (hh:mm) ± (s) | | |
|---|---|---|
| Method | Preprocessing Time | Training Time |
| HGNN | 11.34s±10s | 4m:24s±6s |
| HGNNP | 6.02s±0s | 4m:13s±2s |
| HNHN | 6.01s±0s | 5m:31s±1s |
| HyperGCN | 6.32s±0s | 1m:33s±0s |
| UniGAT | 6.04s±0s | 4m:11s±0s |
| UniGCN | 5.79s±0s | 4m:12s±0s |
| UniGIN | 6.64s±1s | 3m:4s±1s |
| UniSAGE | 5.79s±0s | 3m:2s±0s |

(b) CAT-EDGE-BRAIN

**Table 8.12.** Timings for our method broken up into the preprocessing phase and training phases (2000 epochs) for the hypergraph datasets.

| Timings (hh:mm) ± (s) | | |
|---|---|---|
| Method | Preprocessing Time | Training Time |
| HGNN | 3m:30s±5s | 1h:29m:33s±6s |
| HGNNP | 3m:34s±1s | 1h:48m:57s±1s |
| HNHN | 3m:41s±1s | 2h:9m:34s±1s |
| HyperGCN | 3m:24s±1s | 58m:27s±1s |
| UniGAT | 3m:50s±1s | 4h:21m:24s±1s |
| UniGCN | 3m:38s±1s | 29m:14s±1s |
| UniGIN | 3m:50s±1s | 27m:50s±1s |
| UniSAGE | 3m:41s±1s | 27m:22s±1s |

(a) WikiPeople-0bi

| Timings (hh:mm) ± (s) | | |
|---|---|---|
| Method | Preprocessing Time | Training Time |
| HGNN | 8m:11s±52s | 37m:18s±9s |
| HGNNP | 7m:34s±1s | 47m:56s±1s |
| HNHN | 6m:21s±1s | 49m:33s±1s |
| HyperGCN | 8m:20s±1s | 28m:25s±1s |
| UniGAT | 10m:40s±1s | 1h:54m:36s±1s |
| UniGCN | 7m:25s±1s | 2h:40m:20s±1s |
| UniGIN | 10m:37s±1s | 2h:48m:35s±1s |
| UniSAGE | 6m:58s±1s | 2h:35m:4s±1s |

(b) JF17K

**Table 8.13.** Timings for our method broken up into the preprocessing phase and training phases (2000 epochs) for the graph datasets.

| Timings (hh:mm) ± (s) | | |
|---|---|---|
| Method | Preprocessing Time | Training Time |
| HGNN | 11m:14s±75s | 53m:21s±2845s |
| HGNNP | 11m:10s±21s | 1h:34m:25s±35s |
| HNHN | 5m:15s±395s | 1h:35m:15s±419s |
| HyperGCN | 33.98s±0s | 5m:8s±0s |
| UniGAT | 1m:59s±120s | 2h:2m:47s±25s |
| UniGCN | 34.37s±0s | 1h:17m:38s±2s |
| UniGIN | 34.05s±0s | 1h:16m:38s±7s |
| UniSAGE | 34.36s±0s | 1h:16m:34s±3s |

(a) johnshopkins55

| Timings (hh:mm) ± (s) | | |
|---|---|---|
| Method | Preprocessing Time | Training Time |
| HGNN | 17m:9s±164s | 12m:38s±549s |
| HGNNP | 15m:34s±61s | 20m:26s±124s |
| HNHN | 15m:31s±83s | 18m:11s±30s |
| HyperGCN | 15m:46s±32s | 4m:17s±80s |
| UniGAT | 1m:27s±6s | 16m:30s±0s |
| UniGCN | 15m:57s±24s | 18m:42s±170s |
| UniGIN | 16m:14s±73s | 16m:22s±39s |
| UniSAGE | 8m:42s±610s | 8m:49s±324s |

(b) AIFB

**Table 8.14.** Timings for our method broken up into the preprocessing phase and training phases (2000 epochs) for the hypergraph datasets.

| Timings (hh:mm) ± (s) | | |
|---|---|---|
| Method | Preprocessing Time | Training Time |
| HGNN | 4m:1s±11s | 22m:30s±1177s |
| HGNNP | 3m:53s±4s | 39m:30s±3s |
| HNHN | 3m:16s±22s | 44m:7s±71s |
| HyperGCN | 3m:35s±23s | 5m:22s±25s |
| UniGAT | 11.92s±0s | 1h:51m:53s±123s |
| UniGCN | 3m:20s±6s | 39m:18s±51s |
| UniGIN | 3m:21s±8s | 38m:3s±0s |
| UniSAGE | 11.27s±0s | 58m:48s±956s |

(a) amherst41

| Timings (hh:mm) ± (s) | | |
|---|---|---|
| Method | Preprocessing Time | Training Time |
| HGNN | 3m:32s±9s | 1h:19m:5s±4684s |
| HGNNP | 3m:26s±10s | 2h:19m:44s±3586s |
| HNHN | 3m:27s±0s | 1h:55m:48s±22s |
| HyperGCN | 3m:28s±0s | 10m:31s±18s |
| UniGAT | 3m:24s±5s | 3h:50m:24s±91s |
| UniGCN | 3m:19s±4s | 1h:39m:46s±13s |
| UniGIN | 3m:17s±0s | 1h:36m:47s±35s |
| UniSAGE | 3m:25s±13s | 1h:37m:16s±102s |

(b) FB15k-237

# 9. SUMMARY AND FUTURE DIRECTIONS

This dissertation has covered various problem settings for computing and learning on combinatorial data.

In Chapter 2, we have gone over the requisite math background. This involves many branches of mathematics that have been developed in the past century. In Chapter 3, we provide a general framework for **persistence** across vector spaces. We develop this framework from a causality-based perspective of data with the language of category theory and algebra. In Chapter 4, we consider persistent homology on a filtration of simplicial complexes as a matrix reduction problem. We show the first GPU-based computation of the matrix reduction problem. In Chapter 5, we show a GPU-based algorithm for computing persistent homology in the case of the metric-induced Vietoris-Rips complex. The success of this approach is centered around the concept of a large number of independent higher-order nearest neighbors for point sets. In Chapter 6, we introduce an 2D embedding for the intervals computable from a persistent homology matrix reduction. These can be compared with the 1-Wasserstein distance between them. This is a variant of the Earth Mover's Distance problem. We conjecture that our reduction from computing the EMD problem via a geometric spanner is optimal. In Chapter 7, we investigate graph learning. We use the persistent homology framework to enhance the expressivity of graph neural networks, which learn on graphs. We introduce extended persistent homology for graphs as a way to obtain information about the cycle basis of a graph. In Chapter 8, we study the hyperlink prediction problem. We identify false positive symmetries recognized by hypergraph neural network encoders and devise a method to address these issues. Our method also addresses the persistence of the encoder's symmetry group learned from training in the context of a temporal shift.

## 9.1 Future Directions:

As we have introduced in the third chapter, we can view persistence across vector spaces as a general theory for measuring algebraic information across time. This framework is applicable to a wide range of problems. In this dissertation we discussed computing and learning on combinatorial data. We covered connected data types ranging from simplicial complexes, graphs, hypergraphs etc.

Central to persistence across vector spaces is that of a matrix representation. With this, the question of persistence can be tractably computed in a well-defined manner. Some questions of importance include:

1. How to define the matrix?

2. How can this matrix be constructed efficiently?

3. What physical meaning is represented by the matrix?

An interesting question for the persistence framework is the choice of the **data representation functor**. This functor is a view of the data. In this dissertation, we have investigated the homology functor. This encodes the disconnections between adjacent connectivity dimensions. We have also provided examples of other ways to view the persistence framework.

Of importance to the functor is that of being able to identify persistent 1-dimensional subspaces in the downstairs vector spaces such as in the interval decomposition of Gabriel's theorem. This prevents ambiguity in understanding the physical meaning of individual vectors. It is also paramount to have a common embedding for any algebraic equivalence classes that can still respect the geometry of the data. We mention this in Section 3.1 and introduce the **SpanVec** category. This is used in the definition of the **data representation functor**.

We believe that there are plenty of problems yet to be solved in this framework. This dissertation focuses on combinatorial data, namely the algebraic information obtained from connectivity. There are other data modalities of course and connections to other areas of computer science.

# VITA

Simon Zhang is a Ph.D. student in the Computer Science Department at Purdue University. He received his Masters in Computer Science and Engineering from the Ohio State University in 2020 and has a Bachelors in Computer Science from Cornell University. He was born in San Antonio, Texas.